Doctoral Thesis

# Advanced Feedback Linearization Control for Tiltrotor UAVs: Gait Plan, Controller Design, and Stability Analysis

(ティルトローターUAV におけるフィードバック線形化制御の応用: 歩行計画，制御器設計および安定性解析)

Zhe Shen

沈 哲

A dissertation submitted in partial fulfillment of the requirements for the degree of

Doctor of Philosophy

The University of Tokyo

Program Authorized to Offer Degree: Department of Aeronautics and Astronautics

# Declaration of Authorship

I, Zhe Shen, avow that the work contained in this thesis, *Advanced Feedback Linearization Control for Tiltrotor UAVs: Gait Plan, Controller Design, and Stability Analysis*, is solely my own and, to my understanding and belief, does not feature any material that has already been published or created by another person, apart from where proper citation is made in the text.

If any of my work is based on research conducted by myself or others, I have indicated this in the text and have provided appropriate citations in the references.

Furthermore, this thesis contains elements from the previously published papers which I and my colleagues co-authored:

**Shen, Z.**; Tsuchiya, T. Singular Zone in Quadrotor Yaw–Position Feedback Linearization. *Drones* **2022**, *6*, 20, doi:doi.org/10.3390/drones6040084.

**Shen, Z.**; Tsuchiya, T. Gait Analysis for a Tiltrotor: The Dynamic Invertible Gait. *Robotics* **2022**, *11*, 33, doi:10.3390/robotics11020033.

**Shen, Z.**; Tsuchiya, T. Cat-Inspired Gaits for a Tilt-Rotor—From Symmetrical to Asymmetrical. *Robotics* **2022**, *11*, 60, doi:10.3390/robotics11030060.

**Shen, Z.**; Ma, Y.; Tsuchiya, T. Four-Dimensional Gait Surfaces for a Tilt-Rotor—Two Color Map Theorem. *Drones* **2022**, *6*, 103, doi:10.3390/drones6050103.

**Shen, Z.**; Ma, Y.; Tsuchiya, T. Feedback Linearization-Based Tracking Control of a Tilt-Rotor with Cat-Trot Gait Plan. *International Journal of Advanced Robotic Systems* **2022**, *19*, 17298806221109360, doi:10.1177/17298806221109360.

**Shen, Z.**; Ma, Y.; Tsuchiya, T. Stability Analysis of a Feedback-Linearization-Based Controller with Saturation: A Tilt Vehicle with the Penguin-Inspired Gait Plan. *arXiv preprint arXiv:2111.14456* **2021**.

**Shen, Z.**; Tsuchiya, T. State Drift and Gait Plan in Feedback Linearization Control of a Tilt Vehicle. In Proceedings of the Computer Science & Information Technology (CS & IT); Academy & Industry Research Collaboration Center (AIRCC): Vienna, Austria, March 19 **2022**; Vol. 12, pp. 1–17.

**Shen, Z.**; Tsuchiya, T. The Robust Gait of a Tilt-Rotor and Its Application to Tracking Control -- Application of Two Color Map Theorem. *International Conference on Control, Automation and Systems (ICCAS)* **2022**.

**Shen, Z.**; Ma, Y.; Tsuchiya, T. Generalized Two Color Map Theorem -- Complete Theorem of Robust Gait Plan for a Tilt-Rotor. *arXiv preprint arXiv:2206.13422*. **2022**.

**Shen, Z.**; Tsuchiya, T. Tracking Control for a Tilt-rotor with Input Constraints by Robust Gaits. *IEEE Aerospace Conference* **2023.**

It has been confirmed that I have the appropriate permissions from the respective publishers to include these papers in my thesis.

I hereby declare that, should any aspect of this dissertation have been submitted for any other degree or qualification at another institution, that this has been communicated in writing to the Dean of the Graduate School.

Zhe Shen
5/3/2023

*På trods af forskelle i race, farve og ideologi er afstanden fra ethvert land til universet den samme.*

*Regardless of variations in race, color, or ideology between countries, the distance to the cosmos remains the same.*

尽管存在种族、肤色与意识形态的不同，从任何国家出发到宇宙的距离都是相同的。

人種や色、思想の違いはあっても、どの国から宇宙への距離も同じです。

# Acknowledgement

I am deeply indebted to my family for their unwavering support throughout my academic journey. Without their encouragement, patience, and love, I would not have been able to achieve my dreams and goals.

I would also like to express my sincere appreciation to my friend, Yu Dong Ma. As I embarked on my academic pursuit in Japan, he was the one who generously offered his assistance and companionship.

His intelligence and wisdom have consistently amazed and inspired me. I will always cherish the time we spent together, and I cannot thank him enough for being a true friend.

In addition, I am immensely grateful to my advisor for his invaluable guidance, encouragement, and support. Furthermore, I would like to thank the American and Korean researchers in my laboratory for their insightful feedback and helpful suggestions.

Lastly, I would like to acknowledge the significance of the United Nations' "***All Human Beings Are Born Free and Equal***" provision, which provides the necessary legal protection for academic freedom. It has allowed me to pursue my research interests and express my ideas without fear or discrimination. In conclusion, I owe a great deal of gratitude to all those who have supported, inspired, and challenged me throughout my academic journey. Thank you.

# Abstract

Ryll's tiltrotor, mounted with the extra 4 motors to vary the directions of the thrusts (with respect to its body-fixed frame), includes eight inputs. Different from the conventional quadrotor, the directions of the thrust in the tiltrotor are adjustable during the flight. Therefore, the total control force can be generated along more than one direction in a tiltrotor.

Advanced with the lateral forces, controlling tiltrotors attracted increasing attentions in the past decade. One of the most attractive control methods of stabilizing this vehicle is Feedback Linearization. By inverting the dynamics, this method transfers the nonlinear system to a linear one to accommodate the further control methods. Three challenges, however, can hinder the application of Feedback Linearization: over-intensive control signals, singular decoupling matrix, and saturation. Activating any of these three issues can challenge the stability proof.

To solve these three challenges, first, this research proposed the drone gait plan. The gait plan was initially used to figure out the control problems in quadruped (four-legged) robots; applying this approach, accompanied by Feedback Linearization, the quality of the control signals was enhanced. Then, we proposed the concept of unacceptable attitude curves, which are not allowed for the tiltrotor to travel to. The Two Color Map Theorem was subsequently established to enlarge the supported attitude for the tiltrotor. These theories were employed in the tiltrotor tracking problem with different references. Notable improvements in the control signals were witnessed in the tiltrotor simulator.

Finally, we explored the control theory, the stability proof of the novel mobile robot (tilt vehicle) stabilized by Feedback Linearization with saturation. Instead of adopting the tiltrotor model, which is over-complicated, we designed a conceptual mobile robot (tilt-car) to analyze the stability proof. The stability proof (stable in the sense of Lyapunov) was found for a mobile robot (tilt vehicle) controlled by Feedback Linearization with saturation for the first time.

The success tracking result with the promising control signals in the tiltrotor simulator demonstrates the advances of our control method. Also, the Lyapunov candidate and the tracking result in the mobile robot (tilt-car) simulator confirm our deductions of the stability proof. These results reveal that these three challenges in Feedback Linearization are solved, to some extents.

# Contents

# Chapter 1

# Introduction

### 1. Background of The Tiltrotor

In the past decade, tiltrotors have attracted great interest. Tiltrotors are a novel type of quadrotor [1–8], wherein the axes of the propellers tilt, imparting the ability to change the direction of each thrust [9]. The tilt-rotor quadrotor [8,10–12], which is also referred as the thrust vectoring quadrotor [13,14], is a novel type of the quadrotor. Augmented with the additional mechanical structure [15,16], it is able to provide the lateral force [17].

Comparing with the conventional quadrotor [18–21], the tilt-rotors [22–26] provide the lateral forces, which are not applicable to the collinear/coplanar platforms (e.g., conventional quadrotors). The additional mechanical structures (usually tilting motors) mounted on the arms of the tilt-rotor provide the possibility of changing the direction of each thrust or 'tilting'. As a consequence, the number of inputs increases to eight (four magnitudes of the thrust and four directions of the thrusts) [17].

Among the designs of the tilt-rotor, Ryll's model, the tilt-rotor with eight inputs, received great attention in the last decade [17,24]. It has been an exact decade since the first time Ryll's tilt-rotor was put forward and stabilized [24]. Comparing with the conventional quadrotor, this UAV attracts attentions since its unique capability of generating the lateral forces.

As shown in Fig. 1 [9], the directions of the thrusts are able to be adjusted by changing the tilting angles, $\alpha_1, \alpha_2, \alpha_3, \alpha_4$, during the flight. Here, the direction of each thrust is shown in the corresponding yellow plane. For details in the kinematics, [1,2,4,5,7] are recommended. The primary subject in this thesis is Ryll's tiltrotor.

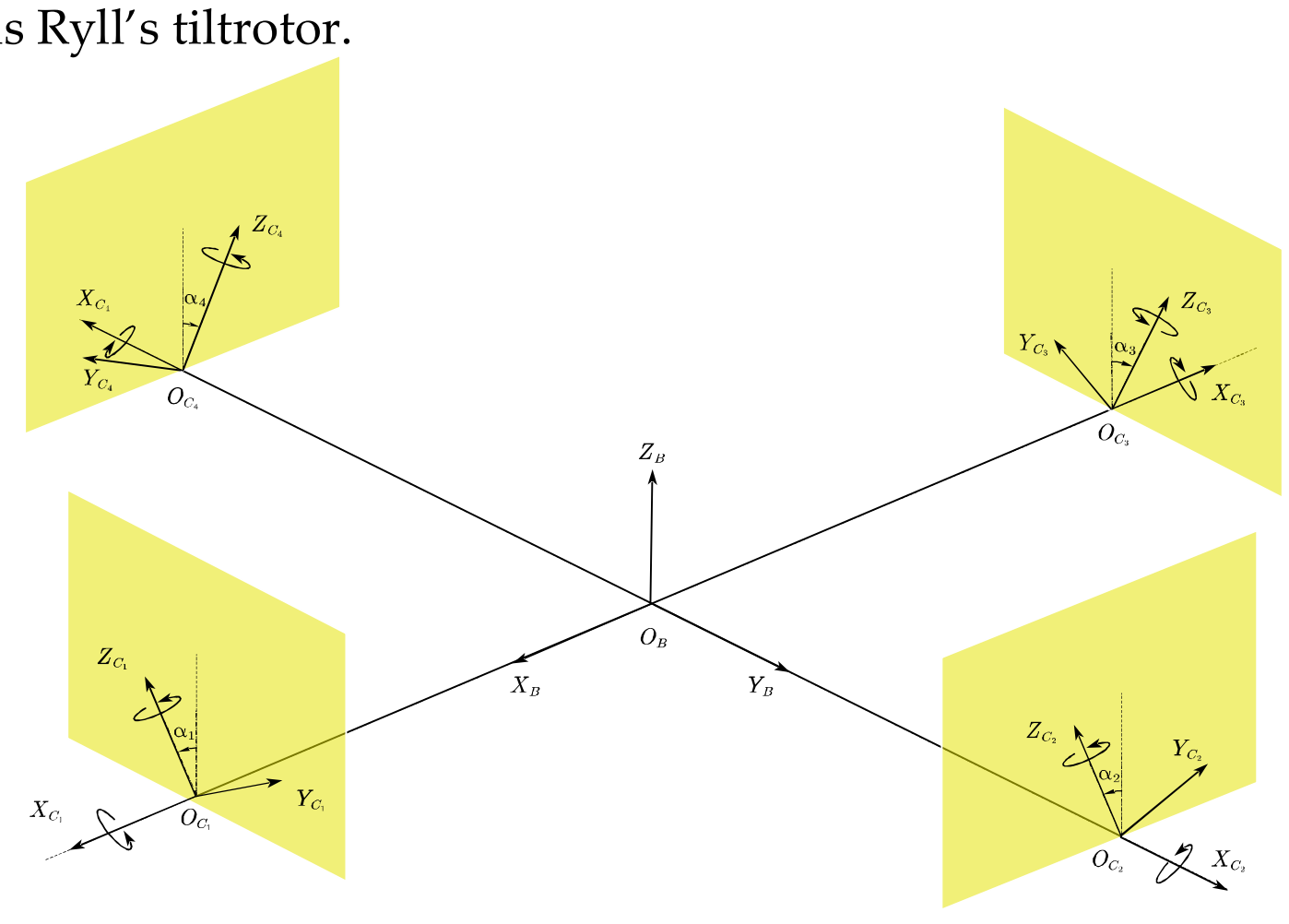

**Figure 1.** The sketch of the tilt-rotor.

Obviously, since the total control force in a conventional quadrotor can be produced only along one direction; maneuvers in which rotation and translation are completely independent are precluded to such platforms [23]. For example, the platform of a conventional quadrotor has to move through a hostile and cluttered ambient or resist a wind gust while keeping a desired attitude [23]. With the introduction of the direction-adjustable thrust in the tiltrotor, it is mechanically possible to change the structure of the tiltrotor during the flight, which provides the possibility of keeping a desired attitude during the flight while tracking the reference.

Another parallel UAV is the tilted-rotor. In a tilted-rotor, the directions of the thrusts are fixed during the flight; these directions, which are constant, are predefined before the flight. Therefore, the direction of the total force can be changed by selecting the intensity of the force produced by each propeller, without the need of reorienting the whole vehicle. Further, based on the trajectories, the relevant optimal structure of the tilted-rotor can be designed [23]. This optimized structure, however, is not allowed to be modified during the flight. On the other hand, the tiltrotor makes it possible.

Typical control methods to stabilize a tiltrotor include LQR and PID [27–29], Backstepping and Sliding Mode [30–34], Feedback Linearization [35–41], Optimal Control [42–44], Adaptive Control [45,46], etc. Among them, Feedback Linearization [47–49] explicitly decouples the nonlinear parts and assists utilizing the over-actuated properties. This approach is not only effective for the tiltrotor but also for the tiltrotor with pre-determined tilting angles [50]. With such benefits in Feedback Linearization, the result witnesses the satisfying behaviors such as tracking with a rapid response for the sophisticated references [35,36,40].

With its unique character in linearizing the nonlinear system [51–55]; several degrees of freedom can be subsequently controlled independently by feedback linearization. These degrees of freedom can be selected as attitude and altitude [4,6,7], position and yaw angle [47,56–58], attitude only [59], etc.

In a conventional quadrotor, the reason for not assigning all six degrees of freedom as independent controlled variables is that the number of inputs is four, marking the maximum number of degrees of freedom to be independently controlled [24] less than the entire degrees of freedom. Typical feedback linearization requires an invertible decoupling matrix [60,61], which is always satisfied for a quadrotor with attitude–altitude independent output choice, while the position-yaw independent output choice witnesses a singular decoupling matrix for some attitudes [58,62].

On the other hand, some studies stabilize Ryll's tilt-rotor [10,15,63], a novel UAV with eight inputs, by controlling all degrees of freedom independently and simultaneously. The relevant decoupling matrices with six inputs and eight inputs have been proven invertible, marking the feasibility of the application of feedback linearization.

Section 2 exemplifies the further advantages of the tiltrotor. And the disadvantages of the feedback linearization in stabilizing the tiltrotor is detailed in Section 3.

## 2. Hovering Experiment in The Gust (Simulation)

One of the typical demonstrations of the unique advantages of the tiltrotor was the attitude-tracking simulation, where the tiltrotor was expected to track the attitude reference while maintaining at a fixed position in the space [24].

Due to the limit in the mechanical structure, the conventional quadrotor is not able to change the attitude while being positioned at a setpoint by the controller; the displacement in the space is inevitable to achieve the desired change in the attitude.

To further demonstrate the advantages of the tiltrotor, we design the hovering experiments in the gust, which is mimicked by the constant acceleration, $1m/s^2$, along an unchangeable direction, the positive direction of $x - axis$, in the inertial frame. The tiltrotor or the conventional quadrotor is expected to be hovering in three experiments with specific settings in each:

1) The tiltrotor adopting the 'perfect' combination of the tilting angles, $\alpha_1, \alpha_2, \alpha_3, \alpha_4$, while being controlled by Feedback Linearization.

2) The tiltrotor adopting the 'imperfect' combination of the tilting angles, $\alpha_1, \alpha_2, \alpha_3, \alpha_4$, while being controlled by Feedback Linearization.

3) The conventional quadrotor controlled by Feedback Linearization.

The resulting attitudes after being stabilized are depicted in Figure 2 to Figure 4, respectively; while the details of the experiments are omitted but can be found later Chapters.

It can be clearly seen that the tiltrotor adopting the 'perfect' combination of the tilting angles, $\alpha_1, \alpha_2, \alpha_3, \alpha_4$, receives the least Euler angles in attitude (Figure 2); the roll angle, pitch angle, and yaw angles of it are close to zero. While the Euler angles in attitude of the conventional quadrotor (Figure 4) are larger than the tiltrotor's; maintaining the position with zero Euler angles with the existence of the gust is not possible in this case.

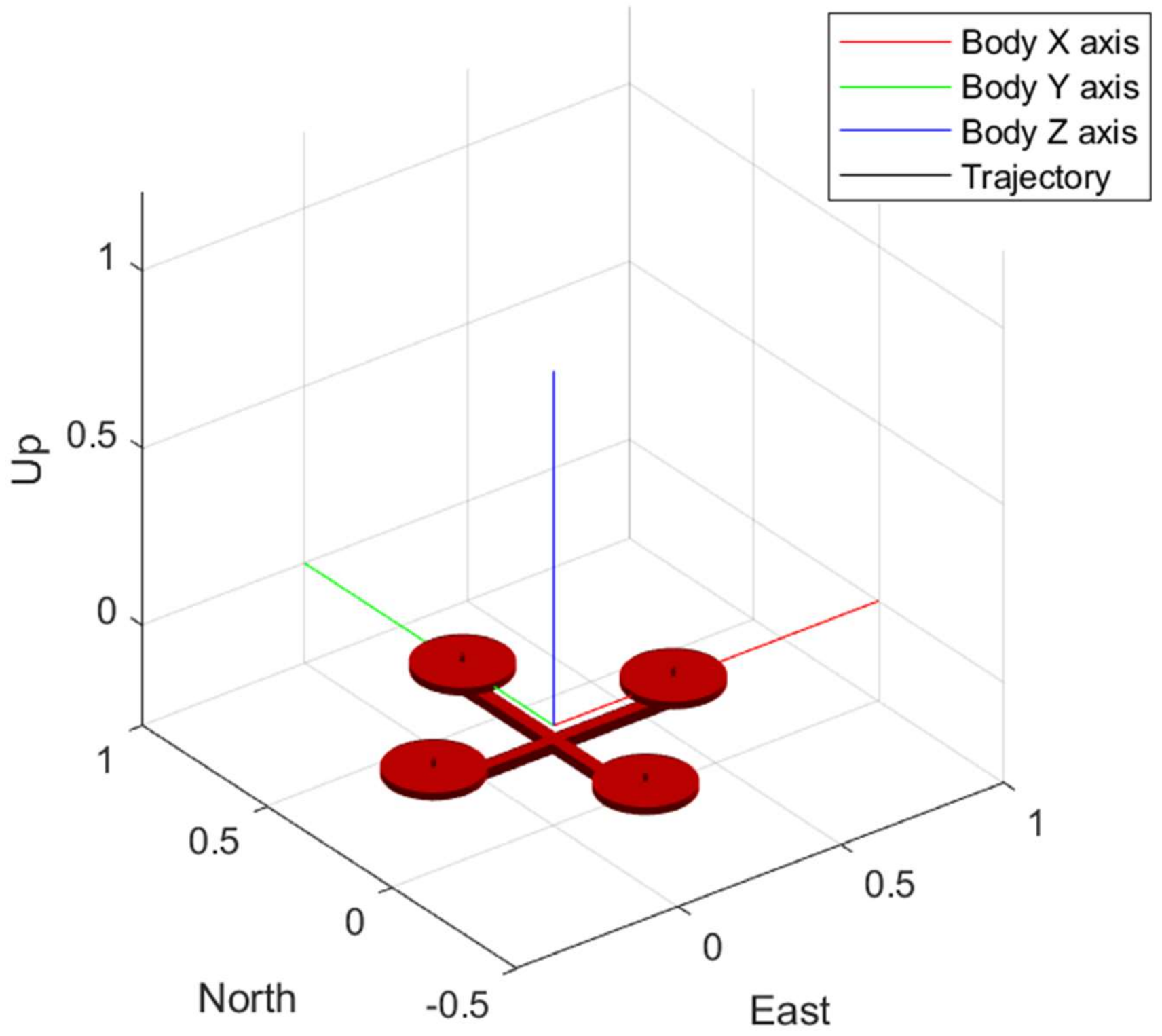

**Figure 2.** The attitude of the tiltrotor which adopts the 'perfect' combination of the tilting angles.

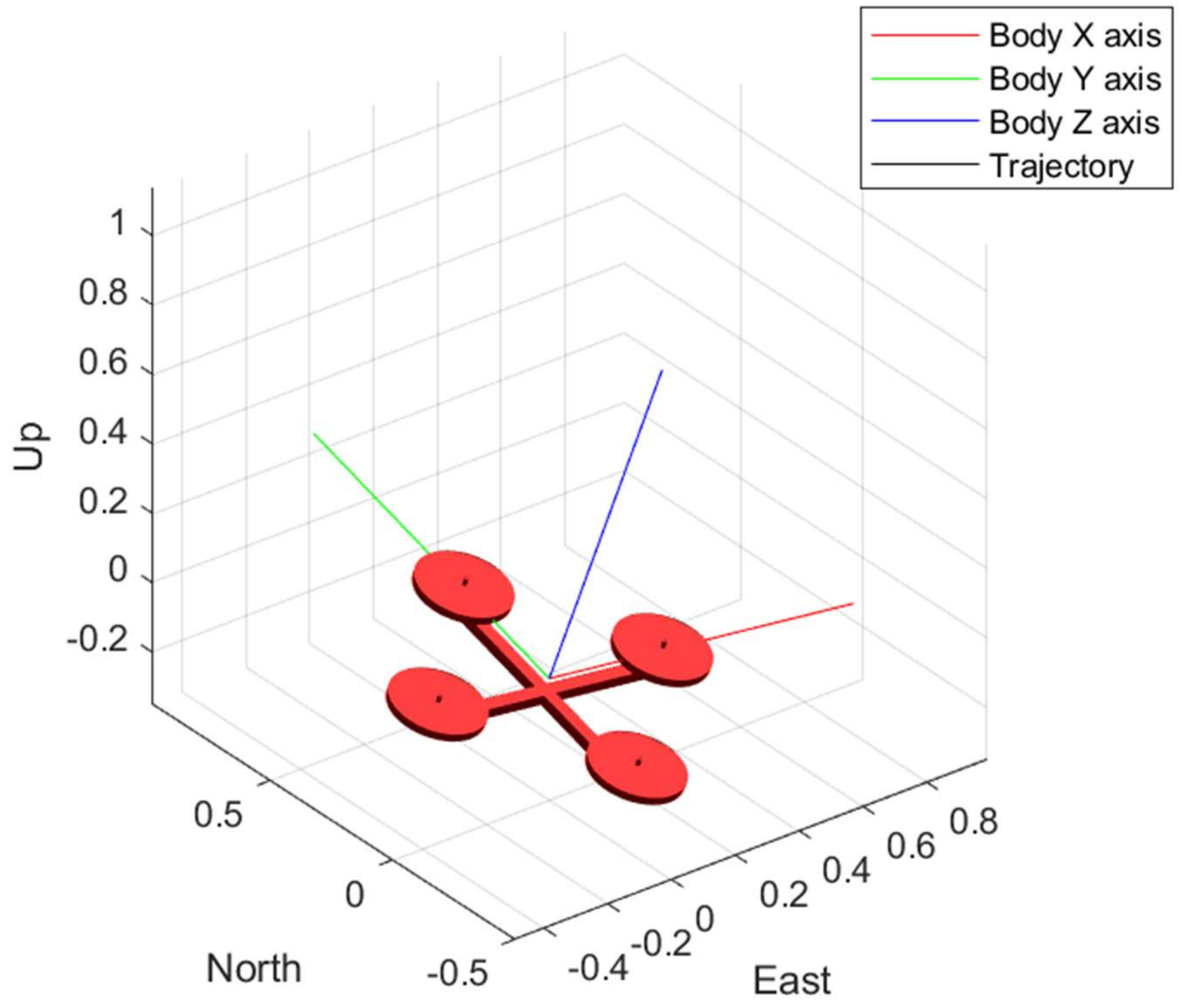

**Figure 3.** The attitude of the tiltrotor which adopts the 'imperfect' combination of the tilting angles.

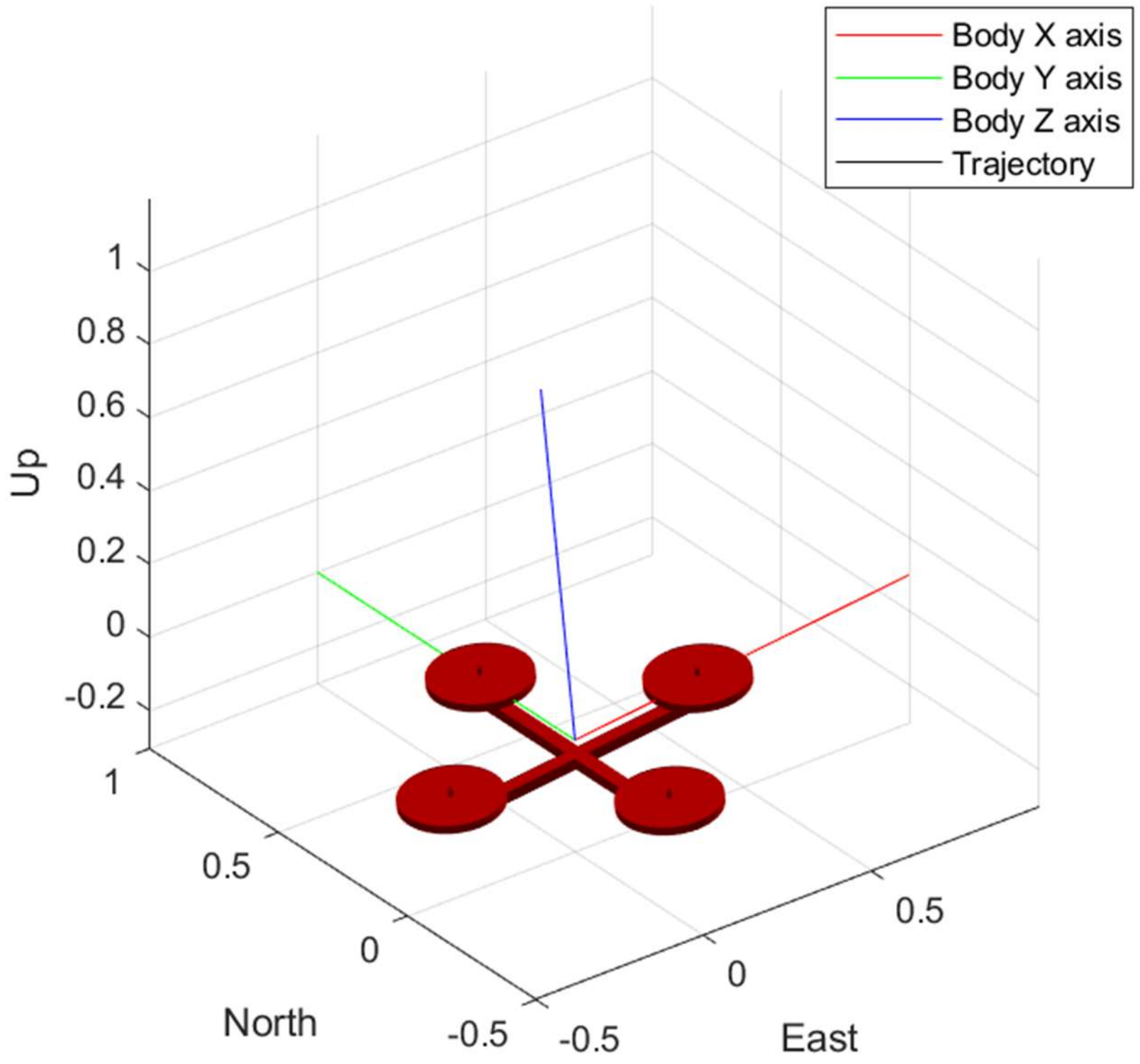

**Figure 4.** The attitude of the conventional quadrotor.

Interestingly, the tiltrotor adopting the 'imperfect' combination of the tilting angles, $\alpha_1, \alpha_2, \alpha_3, \alpha_4$, may receive the larger Euler angles in attitude (Figure 3) than the quadrotor. Therefore, it is more than necessary to plan than the *gait* (the combination of the tilting angles, $\alpha_1, \alpha_2, \alpha_3, \alpha_4$).

Note that the tilted-rotor is not able to change the tilting angles during the flight. As the consequence, the attitude of the tilted-rotor can be unexpected if following a sequencing reference without modifying the gait. On the other hand, the tiltrotor provides the possibility.

Despite of the merit of the tiltrotor, the application of Feedback Linearization, the most popular method in stabilizing the tiltrotor, can be challenged.

## 3. Challenges in Applying Feedback Linearization to The Tiltrotor

With the higher demands on the performance of the tiltrotors and the conventional quadrotor (four inputs) in trajectory tracking, the effects of the controllers based on the linearization at the equilibrium state are reduced [58]. Several nonlinear controllers have been developed to stabilize the quadrotor. References [54,62,64–66] applied feedback linearization equipped with a PID controller. References [67–70] employed the backstepping method, which guarantees stability using the Lyapunov criteria. References [71–74] developed a sliding mode controller. References [75–78] utilized MPC.

Among these controllers, the feedback linearization method is relatively special, since it transfers the original nonlinear system into a linear one compatible with the linear controllers. With the development of the concept of the tiltrotor [10,24,79] in the last decade, feedback linearization [10,54,79–81] has reclaimed its high popularity [58].

Although feedback linearization yields convenience in designing the exterior loop controller, several problems may hinder its application [58]. One requirement is that the control signal and the state variable should not activate the constraints; hitting the input saturation or non-negative constraint are strictly prohibited in the exchange to hold the relevant stability criteria. To meet this requirement, References [82–85] put forward Reference Governor. Note that hitting a boundary does not necessarily result in instability, the corresponding stability criteria for these cases can be hard to trace or generalize [22,23,86].

Another potential issue is the singular decoupling matrix; a singular matrix can block the application of the feedback linearization, though some methods, e.g., dynamic approximation [56], are put forward to avert this issue.

Interestingly, the singular decoupling matrix does not universally exist in the dynamics of a quadrotor or tiltrotor. The conclusion can differ from the choices of the independent outputs directly tracked, even for the same type of tiltrotor or quadrotor [58].

It was proved in Reference [24] that feedback linearization produces no singular decoupling matrix for tiltrotors (eight inputs) if all the degrees of freedom (six) are tracked independently. Notice that this solution can cause a state drift phenomenon [87].

For the same dynamics (eight-input tiltrotor), Reference [9] points out that feedback linearization is hindered by a singular decoupling matrix at several specific attitudes, if only four degrees of freedom (attitude and altitude) are controlled independently. These attitudes are calculated and visualized in Reference [9].

Another issue is the intensive change in the tilting angles while applying feedback linearization [17]; the resulting changes in the tilting angles can be too large or too intensive. Notice that this intensive change in the tilting angles is not unique in the feedback linearization, e.g., PID [79].

Generally [88], the eight inputs are fully assigned by a united control rule, which makes the number of degrees of freedom less than [24] or equal to [89] the number of inputs. Indeed, these approaches avoid the under-actuated system. Further, the decoupling matrix in this scenario is invertible within the interested attitude region while applying feedback linearization. However, the adverse effect is the intensive change in the tilting angles mentioned beforehand, which may not be desired in application [17].

For example, Figure 5 is the tilting angle history in a hovering problem by feedback linearization for a tilt-rotor [9]. The tilting angles are supposed to change in high frequencies and on a large scale at the beginning.

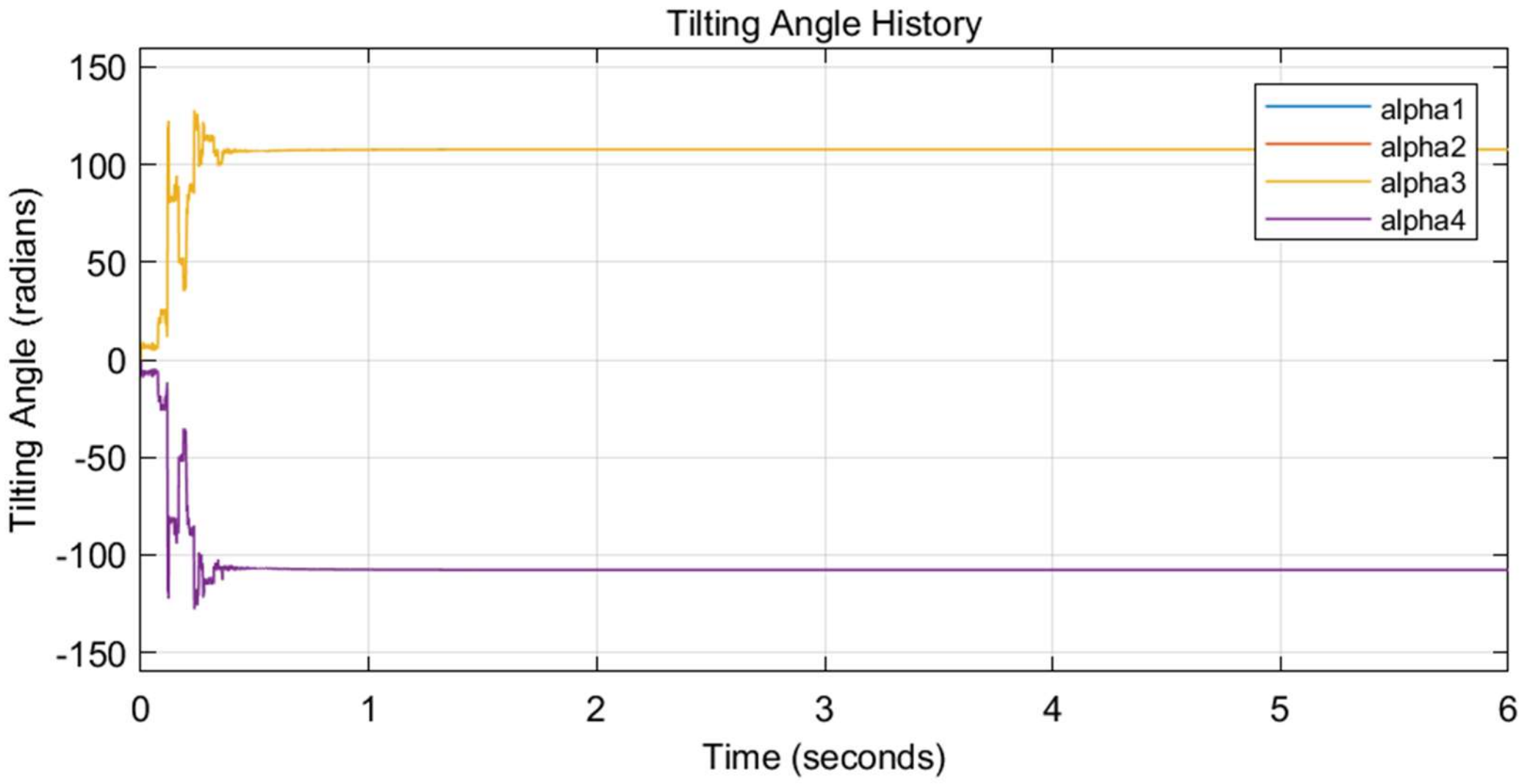

**Figure 5.** The tilting angle history of a tilt-rotor while hovering.

These over-intensive changes in the tilting angles are resulted by the nature of feedback linearization. At each sampling time, the controller calculates the inverse of the dynamics, the result of which can be discontinuous, driving the tilting angles to change intensively.

## 4. Dissertation's Structure

The rest of this dissertation is structured as follows: Chapter 2 deduces the singular zone in quadrotor yaw–position feedback linearization. It gives the reason why the independent variables in feedback linearization of the tiltrotor are picked as attitude-altitude rather than yaw-position. This attitude-altitude decoupling feedback linearization is adopted in Chapter 3 to deal with the over-intensive control signals in the traditional tiltrotor feedback linearization. The concept of gait plan is initially established in the same chapter. This chapter, however, does not give the solution to the position-tracking problem. To advance this novel feedback linearization method, Chapter 4 deduces the advanced position-attitude decoupler for the tiltrotor. This advanced decoupler makes the position-tracking possible for the tiltrotor.

To evaluate the quality (robustness) of the gait, Chapter 5 establishes the concept of the "unacceptable attitude curve" which visualizes the attitudes where the tiltrotor is not allowed to travel to. A novel theorem, Two Color Map Theorem, to generate the robust gaits is created in Chapter 6. This theorem serves as a guide to creating the gaits which are robust to the disturbances in the attitudes. The complete the Two Color Map Theorem is depicted in Chapter 7 with providing the tiltrotor tracking simulations. Chapter 8 generalizes the Two Color Map Theorem; the robust gaits are allowed to be planned in increasing patterns.

Chapter 9 tracks the references adopting the robust gait while considering the saturations. Several admissible references are calculated applying to the constraints introduced by the non-negative and maximum thrusts.

Notice that the stability analysis is intensively addressed throughout this dissertation; Chapter 2 and Chapter 5 give the stability proof of the controller applying feedback linearization. Besides, Chapter 4 thoroughly discusses the sources of the state errors.

To further investigate the properties of the robot with the tilt-structure in feedback linearization, a novel mobile robot is put forward and stabilized in Section 10. The disadvantages called state drift is reported in the application of feedback linearization in stabilizing the mobile robot. Section 11 gives the stability proof to this mobile robot with the tilt-structure while suffering from the activation of the saturation. The system is proved stable in the sense of Lyapunov. It is the first time that the mobile robot with the tilt-structure, being controlled with feedback linearization, is proved stable in the presence of the activated saturation constraints.

Last but not least, the discussions on the limitations of this research are discussed in Chapter 12 where the effect of the parameter change to the Two Color Map is analyzed. Chapter 13 gives the conclusions of this research and the discussions on the further works.

**References**

1. Ryll, M.; Bülthoff, H.H.; Giordano, P.R. Modeling and Control of a Quadrotor UAV with Tilting Propellers. In Proceedings of the Proceedings - IEEE International Conference on Robotics and Automation; 2012.
2. Senkul, F.; Altug, E. Adaptive Control of a Tilt-Roll Rotor Quadrotor UAV. In Proceedings of the 2014 International Conference on Unmanned Aircraft Systems, ICUAS 2014 - Conference Proceedings; 2014.
3. Senkul, F.; Altug, E. Modeling and Control of a Novel Tilt - Roll Rotor Quadrotor UAV. In Proceedings of the 2013 International Conference on Unmanned Aircraft Systems, ICUAS 2013 - Conference Proceedings; 2013.
4. Nemati, A.; Kumar, M. Modeling and Control of a Single Axis Tilting Quadcopter. In Proceedings of the Proceedings of the American Control Conference; 2014.
5. Junaid, A. Bin; Sanchez, A.D.D.C.; Bosch, J.B.; Vitzilaios, N.; Zweiri, Y. Design and Implementation of a Dual-Axis Tilting Quadcopter. Robotics 2018, 7, doi:10.3390/robotics7040065.
6. Andrade, R.; Raffo, G. V.; Normey-Rico, J.E. Model Predictive Control of a Tilt-Rotor UAV for Load Transportation. In Proceedings of the 2016 European Control Conference, ECC 2016; 2016.
7. Nemati, A.; Kumar, R.; Kumar, M. Stabilizing and Control of Tilting-Rotor Quadcopter in Case of a Propeller Failure. In Proceedings of the ASME 2016 Dynamic Systems and Control Conference, DSCC 2016; 2016; Vol. 1.
8. Anderson, R.B.; Marshall, J.A.; L'Afflitto, A. Constrained Robust Model Reference Adaptive Control of a Tilt-Rotor Quadcopter Pulling an Unmodeled Cart. IEEE Trans. Aerosp. Electron. Syst. 2021, 57, 39–54, doi:10.1109/TAES.2020.3008575.
9. Shen, Z.; Tsuchiya, T. Gait Analysis for a Tiltrotor: The Dynamic Invertible Gait. Robotics 2022, 11, 33, doi:10.3390/robotics11020033.
10. Ryll, M.; Bulthoff, H.H.; Giordano, P.R. A Novel Overactuated Quadrotor Unmanned Aerial Vehicle: Modeling, Control, and Experimental Validation. IEEE Trans. Contr. Syst. Technol. 2015, 23, 540–556, doi:10.1109/TCST.2014.2330999.
11. Jin, S.; Bak, J.; Kim, J.; Seo, T.; Kim, H.S. Switching PD-Based Sliding Mode Control for Hovering of a Tilting-Thruster Underwater Robot. PLOS ONE 2018, 13, e0194427, doi:10.1371/journal.pone.0194427.
12. Kumar, R.; Sridhar, S.; Cazaurang, F.; Cohen, K.; Kumar, M. Reconfigurable Fault-Tolerant Tilt-Rotor Quadcopter System.; American Society of Mechanical Engineers: Atlanta, Georgia, USA, September 30 2018; p. V003T37A008.

13. Invernizzi, D.; Giurato, M.; Gattazzo, P.; Lovera, M. Comparison of Control Methods for Trajectory Tracking in Fully Actuated Unmanned Aerial Vehicles. IEEE Transactions on Control Systems Technology 2021, 29, 1147–1160, doi:10.1109/TCST.2020.2992389.
14. Invernizzi, D.; Lovera, M. Trajectory Tracking Control of Thrust-Vectoring UAVs. Automatica 2018, 95, 180–186, doi:10.1016/j.automatica.2018.05.024.
15. Imamura, A.; Miwa, M.; Hino, J. Flight Characteristics of Quad Rotor Helicopter with Thrust Vectoring Equipment. Journal of Robotics and Mechatronics 2016, 28, 334–342, doi:10.20965/jrm.2016.p0334.
16. Ryll, M.; Bulthoff, H.H.; Giordano, P.R. First Flight Tests for a Quadrotor UAV with Tilting Propellers. In Proceedings of the 2013 IEEE International Conference on Robotics and Automation; IEEE: Karlsruhe, Germany, May 2013; pp. 295–302.
17. Shen, Z.; Tsuchiya, T. Cat-Inspired Gaits for a Tilt-Rotor—From Symmetrical to Asymmetrical. Robotics 2022, 11, 60, doi:10.3390/robotics11030060.
18. Ícaro Bezerra Viana; Luiz Manoel Santos Santana; Raphael Ballet; Davi Antônio dos Santos; Luiz Carlos Sandoval Góes Experimental Validation of a Trajectory Tracking Control Using the AR.Drone Quadrotor.; Fortaleza, Ceará, Brasil, 2016.
19. Merheb, A.-R.; Noura, H.; Bateman, F. Emergency Control of AR Drone Quadrotor UAV Suffering a Total Loss of One Rotor. IEEE/ASME Trans. Mechatron. 2017, 22, 961–971, doi:10.1109/TMECH.2017.2652399.
20. Lee, T.; Leok, M.; McClamroch, N.H. Geometric Tracking Control of a Quadrotor UAV on SE(3). In Proceedings of the 49th IEEE Conference on Decision and Control (CDC); IEEE: Atlanta, GA, December 2010; pp. 5420–5425.
21. Luukkonen, T. Modelling and Control of Quadcopter. Independent research project in applied mathematics, Espoo 2011, 22, 22.
22. Horla, D.; Hamandi, M.; Giernacki, W.; Franchi, A. Optimal Tuning of the Lateral-Dynamics Parameters for Aerial Vehicles With Bounded Lateral Force. IEEE Robot. Autom. Lett. 2021, 6, 3949–3955, doi:10.1109/LRA.2021.3067229.
23. Franchi, A.; Carli, R.; Bicego, D.; Ryll, M. Full-Pose Tracking Control for Aerial Robotic Systems With Laterally Bounded Input Force. IEEE Trans. Robot. 2018, 34, 534–541, doi:10.1109/TRO.2017.2786734.
24. Ryll, M.; Bulthoff, H.H.; Giordano, P.R. Modeling and Control of a Quadrotor UAV with Tilting Propellers. In Proceedings of the 2012 IEEE International Conference on Robotics and Automation; IEEE: St Paul, MN, USA, May 2012; pp. 4606–4613.
25. Nemati, A.; Kumar, R.; Kumar, M. Stabilizing and Control of Tilting-Rotor Quadcopter in Case of a Propeller Failure.; American Society of Mechanical Engineers: Minneapolis, Minnesota, USA, October 12 2016; p. V001T05A005.
26. Junaid, A.; Sanchez, A.; Bosch, J.; Vitzilaios, N.; Zweiri, Y. Design and Implementation of a Dual-Axis Tilting Quadcopter. Robotics 2018, 7, 65, doi:10.3390/robotics7040065.
27. Badr, S.; Mehrez, O.; Kabeel, A.E. A Design Modification for a Quadrotor UAV: Modeling, Control and Implementation. Advanced Robotics 2019, 33, doi:10.1080/01691864.2018.1556116.
28. Bhargavapuri, M.; Patrikar, J.; Sahoo, S.R.; Kothari, M. A Low-Cost Tilt-Augmented Quadrotor Helicopter : Modeling and Control. In Proceedings of the 2018 International Conference on Unmanned Aircraft Systems, ICUAS 2018; 2018.
29. Badr, S.; Mehrez, O.; Kabeel, A.E. A Novel Modification for a Quadrotor Design. In Proceedings of the 2016 International Conference on Unmanned Aircraft Systems, ICUAS 2016; 2016.
30. Jiang, X. ying; Su, C. li; Xu, Y. peng; Liu, K.; Shi, H. yuan; Li, P. An Adaptive Backstepping Sliding Mode Method for Flight Attitude of Quadrotor UAVs. Journal of Central South University 2018, 25, doi:10.1007/s11771-018-3765-0.
31. Jin, S.; Kim, J.; Kim, J.W.; Bae, J.H.; Bak, J.; Kim, J.; Seo, T.W. Back-Stepping Control Design for an Underwater Robot with Tilting Thrusters. In Proceedings of the Proceedings of the 17th International Conference on Advanced Robotics, ICAR 2015; 2015.
32. Kadiyam, J.; Santhakumar, M.; Deshmukh, D.; Seo, T.W. Design and Implementation of Backstepping Controller for Tilting Thruster Underwater Robot. In Proceedings of the International Conference on Control, Automation and Systems; 2018; Vol. 2018-October.
33. Scholz, G.; Popp, M.; Ruppelt, J.; Trommer, G.F. Model Independent Control of a Quadrotor with Tiltable Rotors: IEEE/ION PLANS 2016, April 11-14, Savannah, Georgia, United States of America. In Proceedings of the Proceedings of the IEEE/ION Position, Location and Navigation Symposium, PLANS 2016; 2016.
34. Phong Nguyen, N.; Kim, W.; Moon, J. Observer-Based Super-Twisting Sliding Mode Control with Fuzzy Variable Gains and Its Application to Overactuated Quadrotors. In Proceedings of the Proceedings of the IEEE Conference on Decision and Control; 2019; Vol. 2018-December.
35. Kumar, R.; Nemati, A.; Kumar, M.; Sharma, R.; Cohen, K.; Cazaurang, F. Tilting-Rotor Quadcopter for Aggressive Flight Maneuvers Using Differential Flatness Based Flight Controller. In Proceedings of the ASME 2017 Dynamic Systems and Control Conference, DSCC 2017; 2017; Vol. 3.

36. Saif, A.-W.A. Feedback Linearization Control of Quadrotor with Tiltable Rotors under Wind Gusts. International Journal of Advanced and Applied Sciences 2017, 4, 150–159, doi:10.21833/ijaas.2017.010.021.
37. Offermann, A.; Castillo, P.; De Miras, J.D. Control of a PVTOL* with Tilting Rotors*. In Proceedings of the 2019 International Conference on Unmanned Aircraft Systems, ICUAS 2019; 2019.
38. Rajappa, S.; Bulthoff, H.H.; Odelga, M.; Stegagno, P. A Control Architecture for Physical Human-UAV Interaction with a Fully Actuated Hexarotor. In Proceedings of the IEEE International Conference on Intelligent Robots and Systems; 2017; Vol. 2017-September.
39. Scholz, G.; Trommer, G.F. Model Based Control of a Quadrotor with Tiltable Rotors. Gyroscopy and Navigation 2016, 7, doi:10.1134/S2075108716010120.
40. Ryll, M.; Bülthoff, H.H.; Giordano, P.R. A Novel Overactuated Quadrotor Unmanned Aerial Vehicle: Modeling, Control, and Experimental Validation. IEEE Transactions on Control Systems Technology 2015, 23, doi:10.1109/TCST.2014.2330999.
41. Elfeky, M.; Elshafei, M.; Saif, A.W.A.; Al-Malki, M.F. Quadrotor Helicopter with Tilting Rotors: Modeling and Simulation. In Proceedings of the 2013 World Congress on Computer and Information Technology, WCCIT 2013; 2013.
42. Park, S.; Lee, J.; Ahn, J.; Kim, M.; Her, J.; Yang, G.H.; Lee, D. ODAR: Aerial Manipulation Platform Enabling Omnidirectional Wrench Generation. IEEE/ASME Transactions on Mechatronics 2018, 23, doi:10.1109/TMECH.2018.2848255.
43. Magariyama, T.; Abiko, S. Seamless 90-Degree Attitude Transition Flight of a Quad Tilt-Rotor UAV under Full Position Control. In Proceedings of the IEEE/ASME International Conference on Advanced Intelligent Mechatronics, AIM; 2020; Vol. 2020-July.
44. Falanga, D.; Kleber, K.; Mintchev, S.; Floreano, D.; Scaramuzza, D. The Foldable Drone: A Morphing Quadrotor That Can Squeeze and Fly. IEEE Robotics and Automation Letters 2019, 4, doi:10.1109/LRA.2018.2885575.
45. Lu, D.; Xiong, C.; Zeng, Z.; Lian, L. Adaptive Dynamic Surface Control for a Hybrid Aerial Underwater Vehicle with Parametric Dynamics and Uncertainties. IEEE Journal of Oceanic Engineering 2020, 45, doi:10.1109/JOE.2019.2903742.
46. Antonelli, G.; Cataldi, E.; Arrichiello, F.; Giordano, P.R.; Chiaverini, S.; Franchi, A. Adaptive Trajectory Tracking for Quadrotor MAVs in Presence of Parameter Uncertainties and External Disturbances. IEEE Transactions on Control Systems Technology 2018, 26, doi:10.1109/TCST.2017.2650679.
47. Lee, D.; Jin Kim, H.; Sastry, S. Feedback Linearization vs. Adaptive Sliding Mode Control for a Quadrotor Helicopter. Int. J. Control Autom. Syst. 2009, 7, 419–428, doi:10.1007/s12555-009-0311-8.
48. Al-Hiddabi, S.A. Quadrotor Control Using Feedback Linearization with Dynamic Extension. In Proceedings of the 2009 6th International Symposium on Mechatronics and its Applications; Sharjah, United Arab Emirates, March 2009; pp. 1–3.
49. Mukherjee, P.; Waslander, S. Direct Adaptive Feedback Linearization for Quadrotor Control. In Proceedings of the AIAA Guidance, Navigation, and Control Conference; Minneapolis, Minnesota, August 13 2012.
50. Rajappa, S.; Ryll, M.; Bulthoff, H.H.; Franchi, A. Modeling, Control and Design Optimization for a Fully-Actuated Hexarotor Aerial Vehicle with Tilted Propellers. In Proceedings of the Proceedings - IEEE International Conference on Robotics and Automation; 2015; Vol. 2015-June.
51. Chen, C.-C.; Chen, Y.-T. Feedback Linearized Optimal Control Design for Quadrotor With Multi-Performances. IEEE Access 2021, 9, 26674–26695, doi:10.1109/ACCESS.2021.3057378.
52. Lian, S.; Meng, W.; Lin, Z.; Shao, K.; Zheng, J.; Li, H.; Lu, R. Adaptive Attitude Control of a Quadrotor Using Fast Nonsingular Terminal Sliding Mode. IEEE Transactions on Industrial Electronics 2022, 69, 1597–1607, doi:10.1109/TIE.2021.3057015.
53. Chang, D.E.; Eun, Y. Global Chartwise Feedback Linearization of the Quadcopter With a Thrust Positivity Preserving Dynamic Extension. IEEE Trans. Automat. Contr. 2017, 62, 4747–4752, doi:10.1109/TAC.2017.2683265.
54. Martins, L.; Cardeira, C.; Oliveira, P. Feedback Linearization with Zero Dynamics Stabilization for Quadrotor Control. J Intell Robot Syst 2021, 101, 7, doi:10.1007/s10846-020-01265-2.
55. Kuantama, E.; Tarca, I.; Tarca, R. Feedback Linearization LQR Control for Quadcopter Position Tracking. In Proceedings of the 2018 5th International Conference on Control, Decision and Information Technologies (CoDIT); IEEE: Thessaloniki, April 2018; pp. 204–209.
56. Taniguchi, T.; Sugeno, M. Trajectory Tracking Controls for Non-Holonomic Systems Using Dynamic Feedback Linearization Based on Piecewise Multi-Linear Models. IAENG Int. J. Appl. Math. 2017, 47, 339–351.
57. Mutoh, Y.; Kuribara, S. Control of Quadrotor Unmanned Aerial Vehicles Using Exact Linearization Technique with the Static State Feedback. J. Autom. Control Eng 2016, 4, 340–346, doi:10.18178/joace.4.5.340-346.

58. Shen, Z.; Tsuchiya, T. Singular Zone in Quadrotor Yaw–Position Feedback Linearization. Drones 2022, 6, 20, doi:doi.org/10.3390/drones6040084.
59. Voos, H. Nonlinear Control of a Quadrotor Micro-UAV Using Feedback-Linearization. In Proceedings of the 2009 IEEE International Conference on Mechatronics; IEEE: Malaga, Spain, 2009; pp. 1–6.
60. Rajappa, S.; Ryll, M.; Bulthoff, H.H.; Franchi, A. Modeling, Control and Design Optimization for a Fully-Actuated Hexarotor Aerial Vehicle with Tilted Propellers. In Proceedings of the 2015 IEEE International Conference on Robotics and Automation (ICRA); IEEE: Seattle, WA, USA, May 2015; pp. 4006–4013.
61. Systems Engineering Department, KFUPM, Dhahran 31261, Saudi Arabia; Saif, A.-W.A. Feedback Linearization Control of Quadrotor with Tiltable Rotors under Wind Gusts. Int. j. adv. appl. sci. 2017, 4, 150–159, doi:10.21833/ijaas.2017.010.021.
62. Mistler, V.; Benallegue, A.; M'Sirdi, N.K. Exact Linearization and Noninteracting Control of a 4 Rotors Helicopter via Dynamic Feedback. In Proceedings of the Proceedings 10th IEEE International Workshop on Robot and Human Interactive Communication. ROMAN 2001 (Cat. No.01TH8591); IEEE: Paris, France, September 2001; pp. 586–593.
63. Scholz, G.; Trommer, G.F. Model Based Control of a Quadrotor with Tiltable Rotors. Gyroscopy Navig. 2016, 7, 72–81, doi:10.1134/S2075108716010120.
64. Bolandi, H.; Rezaei, M.; Mohsenipour, R.; Nemati, H.; Smailzadeh, S.M. Attitude Control of a Quadrotor with Optimized PID Controller. ICA 2013, 04, 335–342, doi:10.4236/ica.2013.43039.
65. Bouabdallah, S.; Noth, A.; Siegwart, R. PID vs LQ Control Techniques Applied to an Indoor Micro Quadrotor. In Proceedings of the 2004 IEEE/RSJ International Conference on Intelligent Robots and Systems (IROS) (IEEE Cat. No.04CH37566); IEEE: Sendai, Japan, 2004; Vol. 3, pp. 2451–2456.
66. Wang, S.; Polyakov, A.; Zheng, G. Quadrotor Stabilization under Time and Space Constraints Using Implicit PID Controller. Journal of the Franklin Institute 2022, 359, 1505–1530, doi:10.1016/j.jfranklin.2022.01.002.
67. Bouabdallah, S.; Siegwart, R. Backstepping and Sliding-Mode Techniques Applied to an Indoor Micro Quadrotor. In Proceedings of the Proceedings of the 2005 IEEE International Conference on Robotics and Automation; IEEE: Barcelona, Spain, 2005; pp. 2247–2252.
68. Madani, T.; Benallegue, A. Backstepping Control for a Quadrotor Helicopter. In Proceedings of the 2006 IEEE/RSJ International Conference on Intelligent Robots and Systems; IEEE: Beijing, China, October 2006; pp. 3255–3260.
69. Chen, F.; Lei, W.; Zhang, K.; Tao, G.; Jiang, B. A Novel Nonlinear Resilient Control for a Quadrotor UAV via Backstepping Control and Nonlinear Disturbance Observer. Nonlinear Dyn 2016, 85, 1281–1295, doi:10.1007/s11071-016-2760-y.
70. Liu, P.; Ye, R.; Shi, K.; Yan, B. Full Backstepping Control in Dynamic Systems With Air Disturbances Optimal Estimation of a Quadrotor. IEEE Access 2021, 9, 34206–34220, doi:10.1109/ACCESS.2021.3061598.
71. Xu, R.; Ozguner, U. Sliding Mode Control of a Quadrotor Helicopter. In Proceedings of the Proceedings of the 45th IEEE Conference on Decision and Control; IEEE: San Diego, CA, USA, 2006; pp. 4957–4962.
72. Runcharoon, K.; Srichatrapimuk, V. Sliding Mode Control of Quadrotor. In Proceedings of the 2013 The International Conference on Technological Advances in Electrical, Electronics and Computer Engineering (TAEECE); IEEE: Konya, Turkey, May 2013; pp. 552–557.
73. Luque-Vega, L.; Castillo-Toledo, B.; Loukianov, A.G. Robust Block Second Order Sliding Mode Control for a Quadrotor. Journal of the Franklin Institute 2012, 349, 719–739, doi:10.1016/j.jfranklin.2011.10.017.
74. Xu, L.; Shao, X.; Zhang, W. USDE-Based Continuous Sliding Mode Control for Quadrotor Attitude Regulation: Method and Application. IEEE Access 2021, 9, 64153–64164, doi:10.1109/ACCESS.2021.3076076.
75. Ganga, G.; Dharmana, M.M. MPC Controller for Trajectory Tracking Control of Quadcopter. In Proceedings of the 2017 International Conference on Circuit ,Power and Computing Technologies (ICCPCT); IEEE: Kollam, India, April 2017; pp. 1–6.
76. Abdolhosseini, M.; Zhang, Y.M.; Rabbath, C.A. An Efficient Model Predictive Control Scheme for an Unmanned Quadrotor Helicopter. J Intell Robot Syst 2013, 70, 27–38, doi:10.1007/s10846-012-9724-3.
77. Alexis, K.; Nikolakopoulos, G.; Tzes, A. Model Predictive Control Scheme for the Autonomous Flight of an Unmanned Quadrotor. In Proceedings of the 2011 IEEE International Symposium on Industrial Electronics; June 2011; pp. 2243–2248.
78. Torrente, G.; Kaufmann, E.; Föhn, P.; Scaramuzza, D. Data-Driven MPC for Quadrotors. IEEE Robotics and Automation Letters 2021, 6, 3769–3776, doi:10.1109/LRA.2021.3061307.
79. Kumar, R.; Nemati, A.; Kumar, M.; Sharma, R.; Cohen, K.; Cazaurang, F. Tilting-Rotor Quadcopter for Aggressive Flight Maneuvers Using Differential Flatness Based Flight Controller.; American Society of Mechanical Engineers: Tysons, Virginia, USA, October 11 2017; p. V003T39A006.
80. Ahmed, A.M.; Katupitiya, J. Modeling and Control of a Novel Vectored-Thrust Quadcopter. Journal of Guidance, Control, and Dynamics 2021, 44, 1399–1409, doi:10.2514/1.G005467.

81. Xu, J.; D'Antonio, D.S.; Saldaña, D. H-ModQuad: Modular Multi-Rotors with 4, 5, and 6 Controllable DOF. In Proceedings of the 2021 IEEE International Conference on Robotics and Automation (ICRA); Xi'an, China, May 2021; pp. 190–196.
82. Convens, B.; Merckaert, K.; Nicotra, M.M.; Naldi, R.; Garone, E. Control of Fully Actuated Unmanned Aerial Vehicles with Actuator Saturation. IFAC-PapersOnLine 2017, 50, 12715–12720, doi:10.1016/j.ifacol.2017.08.1823.
83. Cotorruelo, A.; Nicotra, M.M.; Limon, D.; Garone, E. Explicit Reference Governor Toolbox (ERGT). In Proceedings of the 2018 IEEE 4th International Forum on Research and Technology for Society and Industry (RTSI); IEEE: Palermo, Italy, September 2018; pp. 1–6.
84. Dunham, W.; Petersen, C.; Kolmanovsky, I. Constrained Control for Soft Landing on an Asteroid with Gravity Model Uncertainty. In Proceedings of the 2016 American Control Conference (ACC); IEEE: Boston, MA, USA, July 2016; pp. 5842–5847.
85. Hosseinzadeh, M.; Garone, E. An Explicit Reference Governor for the Intersection of Concave Constraints. IEEE Trans. Automat. Contr. 2020, 65, 1–11, doi:10.1109/TAC.2019.2906467.
86. Shen, Z.; Ma, Y.; Tsuchiya, T. Stability Analysis of a Feedback-Linearization-Based Controller with Saturation: A Tilt Vehicle with the Penguin-Inspired Gait Plan. arXiv preprint 2021.
87. Shen, Z.; Tsuchiya, T. State Drift and Gait Plan in Feedback Linearization Control of A Tilt Vehicle. In Proceedings of the Computer Science & Information Technology (CS & IT); Academy & Industry Research Collaboration Center (AIRCC): Vienna, Austria, March 19 2022; Vol. 12, pp. 1–17.
88. Hamandi, M.; Usai, F.; Sablé, Q.; Staub, N.; Tognon, M.; Franchi, A. Design of Multirotor Aerial Vehicles: A Taxonomy Based on Input Allocation. The International Journal of Robotics Research 2021, 40, 1015–1044, doi:10.1177/02783649211025998.
89. Nemati, A.; Kumar, M. Modeling and Control of a Single Axis Tilting Quadcopter. In Proceedings of the 2014 American Control Conference; IEEE: Portland, OR, USA, June 2014; pp. 3077–3082.

# Chapter 2

# Singular Zone in Quadrotor Yaw–Position Feedback Linearization

**This Chapter has been published on *Drones* 2022, *6*, 20, doi:doi.org/10.3390/drones6040084 as the first author. The rest co-author(s) contribute nothing other than supervising and/or minor edition.**

**Abstract:** Feedback linearization-based controllers are widely exploited in stabilizing a tilt rotor (eight or twelve inputs); each degree of freedom (six degrees of freedom in total) is manipulated individually to track the desired trajectory, since no singular decoupling matrix is introduced while applying this method. The conventional quadrotor (four inputs), on the other hand, is an under-actuated MIMO system that can directly track four independent degrees of freedom at most. Common selections of these outputs can be yaw–position and attitude–altitude. It is reported that no singularity is found in the decoupling matrix while applying feedback linearization in the yaw–position-tracking problem. However, in this research, we argue the existence of the ignored singular zone within the range of interest, which can cause the failure in the controller design. This paper visualizes this noninvertible area and details the process of deduction for the first time. An attempt (switch controller) to avert the singular problem is later discussed with the verification by simulation in Simulink and MATLAB. All the results are sketched in the roll–pitch diagram.

## 1. Introduction

With the higher demands on the performance of the conventional quadrotor (four inputs) in trajectory tracking, the effects of the controllers based on the linearization at the equilibrium state are reduced. Several nonlinear controllers have been developed to stabilize the quadrotor. References [1–5] applied feedback linearization equipped with a PID controller. References [6–9] employed the backstepping method, which guarantees stability using the Lyapunov criteria. References [10–13] developed a sliding mode controller. References [14–17] utilized MPC.

Among these controllers, the feedback linearization method is relatively special, since it transfers the original nonlinear system into a linear one compatible with the linear controllers. With the development of the concept of the tilt rotor [18–20] in the last decade, feedback linearization [4,18,20–22] and sliding mode control [13,23] have reclaimed their high popularity.

Although feedback linearization yields convenience in designing the exterior loop controller, several problems may hinder its application. One requirement is that the control signal and the state variable should not activate the constraints; hitting the input saturation or non-negative constraint are strictly prohibited in the exchange to hold the relevant stability criteria. To meet this requirement, References [24–27] put forward Reference Governor.

Another potential issue is the singular decoupling matrix; a singular matrix can block the application of the feedback linearization, though some methods, e.g., PCH [28], dynamic approximation [29], etc., are put forward to avert this issue.

Interestingly, the singular decoupling matrix does not universally exist in the dynamics of a quadrotor or tilt rotor. The conclusion can differ from the choices of the independent outputs directly tracked, even for the same type of tilt rotor or quadrotor.

It was proved in Reference [19] that feedback linearization produces no singular decoupling matrix for tilt rotors (eight inputs) if all the degrees of freedom (six) are tracked independently. Notice that this solution can cause a state drift phenomenon [30].

For the same dynamics (eight-input tilt rotor), Reference [31] points out that feedback linearization is hindered by a singular decoupling matrix at several specific attitudes, if only four degrees of freedom (attitude and altitude) are controlled independently. These attitudes are calculated and visualized in Reference [31].

As for the conventional quadrotor, the number of inputs (six) is less than the number of degrees of freedom (four). Based on this, four degrees of freedom, at most, can be tracked independently for a quadrotor. The typical choices of these degrees of freedom can are yaw–position and attitude–altitude.

In feedback linearization, attitude–altitude ( $\phi, \theta, \psi, \text{and } z$ ) are the independent tracking variables, and the conventional quadrotor introduces no singular decoupling matrix; the delta matrix in Reference [32] is strictly a full rank within the range of interest.

The other choices of independent variables are yaw ($\psi$) and position (x, y, and z) combinations. By applying two integrators as the input signal, Reference [1] showed their results in a simulation with the conclusion that the singularity does not exist once the attitude is in the zone of interest. That is: $\phi \in (-\pi/2, \pi/2)$ and $\theta \in (-\pi/2, \pi/2)$.

On the contrary, we will report the extra singular zone ignored by previous research in feedback linearization by selecting yaw–position ($\psi, x, \text{y, and z}$) as the independent controlled variables in this paper.

It is worth mentioning that Reference [33] tried feedback linearization of the quadrotor dynamics based on yaw–position and seemed to successfully avert the noninvertible problem. While the fundamental reason for its success was an estimation of the dynamics, a linearization at the equilibrium–attitude was applied before the feedback linearization process. This approach, however, lost the essence of the dynamic inversion. As tested in Reference [34], this approximation is only applicable to the controllers within a specific range of attitude.

Notice that the number of selected independent controlled variables in a quadrotor can be less than three; Reference [35] used a cascade control structure with three controlled variables for each layer to control all six variables. The singularity problem is also avoided in this way. However, these choices are beyond the scope of this research.

There are sophisticated works focusing on singularity avoidance, escape, and penetration in gyro systems [36,37]. While designing a sophisticated controller to escape the singular zone for our system is not the primary objective of this research, the main focus of this research is to provide sound proof of the existence of the singular zone in quadrotor yaw–position feedback linearization. A potential solution, a preliminary attempt, specifically, to avert this reported singular zone is put forward after deducing the singular zone later in this paper; a switch control strategy attempts to cancel the effect of this zone, achieving stability within a particular attitude zone.

The remainder of this paper is organized as follows. In Section 2, previous works on the dynamics and feedback linearization applied to a quadrotor are introduced. Section 3 deduces the ignored singular area within the zone of interest. This noninvertible zone is visualized in Section 4. A potential approach to avoid this singular zone is put forward and simulated in Section 5. Finally, there are the conclusions and further discussions in Section 6.

## 2. Dynamics and Feedback Linearization in UAV Control

This section briefs the feedback linearization control in Reference [1]. The details in deducing the dynamics can be seen in Reference [1].

The input vector is given by

$$[u_1 \quad u_2 \quad u_3 \quad u_4]^T \tag{1}$$

where $u_1$ is the total thrust generated by four propellers, $u_2$ is the difference of thrust between the left rotor and the right rotor, $u_3$ is the difference of thrust between the front rotor and the back rotor, and $u_4$ is the difference of torque between the two clockwise-turning rotors and the two counterclockwise-turning rotors.

As introduced in Reference [1], the virtual inputs are

$$[\bar{u}_1 \quad \bar{u}_2 \quad \bar{u}_3 \quad \bar{u}_4]^T \tag{2}$$

where

$$\bar{u}_1 = \dot{\xi}, \tag{3}$$

$$\xi = \dot{\zeta}, \tag{4}$$

$$\zeta = u_1, \tag{5}$$

$$\bar{u}_2 = u_2, \tag{6}$$

$$\bar{u}_3 = u_3, \tag{7}$$

$$\bar{u}_4 = u_4. \tag{8}$$

The dynamics of the quadrotor are then written as

$$\dot{\bar{x}} = f(\bar{x}) + \sum_{i=1}^{4} g_i(\bar{x}) \cdot \bar{u}_i \tag{9}$$

where $\bar{x}$ is the augmented state vector,

$$\bar{x} = [x, y, z, \psi, \theta, \phi, \dot{x}, \dot{y}, \dot{z}, \zeta, \xi, p, q, r]^T, \tag{10}$$

where $x, y,$ and $z$ are the positions with respect to the earth frame; $\psi, \theta,$ and $\phi$ are the yaw angle, pitch angle, and roll angle, respectively, and $p, q,$ and $r$ are the angular velocities along the body-fixed x-axis, y-axis, and z-axis, respectively.

$g_i(\bar{x})$ is defined as

$$g_1(\bar{x}) = [0,0,0,0,0,0,0,0,0,0,1,0,0,0]^T, \tag{11}$$

$$g_2(\bar{x}) = \left[0,0,0,0,0,0,0,0,0,0,0,\frac{d}{I_x},0,0\right]^T, \tag{12}$$

$$g_3(\bar{x}) = \left[0,0,0,0,0,0,0,0,0,0,0,0,\frac{d}{I_y},0\right]^T, \tag{13}$$

$$g_4(\bar{x}) = \left[0,0,0,0,0,0,0,0,0,0,0,0,0,\frac{d}{I_z}\right]^T, \tag{14}$$

where $d$ is the length of each arm, and $I_x, I_y,$ and $I_z$ are the rotational inertia along the body-fixed x-axis, y-axis, and z-axis, respectively.

$f(\bar{x})$ is defined as

$$f(\bar{x}) = \begin{bmatrix} \dot{x} \\ \dot{y} \\ \dot{z} \\ \frac{s\phi}{c\theta} \cdot q + \frac{c\phi}{c\theta} \cdot r \\ c\phi \cdot q - s\phi \cdot r \\ p + s\phi \cdot t\theta \cdot q + c\phi \cdot t\theta \cdot r \\ g_1^7 \cdot \zeta \\ g_1^8 \cdot \zeta \\ g_1^9 \cdot \zeta + g \\ \xi \\ 0 \\ \frac{I_y - I_z}{I_x} \cdot q \cdot r \\ \frac{I_z - I_x}{I_y} \cdot p \cdot r \\ \frac{I_x - I_y}{I_z} \cdot p \cdot q \end{bmatrix} \tag{15}$$

where $s\Lambda, c\Lambda,$ and $t\Lambda$ represent $\sin(\Lambda)$, $\cos(\Lambda)$, and $\tan(\Lambda)$, respectively, g is the gravitational acceleration, and $g_1^7, g_1^8,$ and $g_1^9$ are defined as

$$\begin{aligned} g_1^7 &= -\frac{1}{m} \cdot (c\phi \cdot c\psi \cdot s\theta + s\phi \cdot s\psi) \\ &\triangleq -\frac{1}{m} \cdot A_1, \end{aligned} \tag{16}$$

$$\begin{aligned} g_1^8 &= -\frac{1}{m} \cdot (c\phi \cdot s\psi \cdot s\theta - s\phi \cdot c\psi) \\ &\triangleq -\frac{1}{m} \cdot A_2, \end{aligned} \tag{17}$$

$$\begin{aligned} g_1^9 &= -\frac{1}{m} \cdot c\theta \cdot c\phi \\ &\triangleq -\frac{1}{m} \cdot A_3 \end{aligned} \tag{18}$$

where m is the mass of the quadrotor, and $A_1 = c\phi \cdot c\psi \cdot s\theta + s\phi \cdot s\psi$, $A_2 = c\phi \cdot s\psi \cdot s\theta - s\phi \cdot c\psi$, and $A_3 = c\theta \cdot c\phi$.

The four independent controlled variables ($y_1, y_2, y_3,$ and $y_4$) are selected as

$$\begin{bmatrix} y_1 \\ y_2 \\ y_3 \\ y_4 \end{bmatrix} = \begin{bmatrix} x \\ y \\ z \\ \psi \end{bmatrix}. \tag{19}$$

To apply the feedback linearization, Reference [1] calculated the higher derivative of the controlled variables and received

$$\begin{aligned} \begin{bmatrix} {y_1}^{(4)} \\ {y_2}^{(4)} \\ {y_3}^{(4)} \\ \ddot{y}_4 \end{bmatrix} &= \begin{bmatrix} x^{(4)} \\ y^{(4)} \\ z^{(4)} \\ \ddot{\psi} \end{bmatrix} \\ &= Ma(\bar{x}) + \Delta(\bar{x}) \cdot \begin{bmatrix} \bar{u}_1 \\ \bar{u}_2 \\ \bar{u}_3 \\ \bar{u}_4 \end{bmatrix} \end{aligned} \tag{20}$$

where both $Ma(\bar{x})$ and $\Delta(\bar{x})$ are a 4 ×4 matrix of state (10).

Once $\Delta(\bar{x})$ is invertible, the feedback linearization can be further applied. It was concluded in Reference [1] that $\Delta(\bar{x})$ is always invertible given that $\phi \in (-\pi/2, \pi/2)$, $\theta \in (-\pi/2, \pi/2)$, $\zeta \neq 0$.

However, based on our deductions, this condition does not hold within the entire space above. The additional requirement for $\Delta(\bar{x})$ and its proof are given in the next section.

## 3. Invertibility Analysis

### *3.1. Necessary and Sufficient Condition to Be Invertible*

$\Delta(\bar{x})$ is invertible within the region $\phi \in (-\pi/2, \pi/2)$, $\theta \in (-\pi/2, \pi/2)$, $\zeta \neq 0$ if and only if

$$-1 + \cos^2\theta \cdot \cos^2\phi - \cos^2\theta \cdot \cos\phi \cdot \sin\phi \neq 0. \quad (21)$$

### *3.2. Proof of The Invertible Condition*

This section details the process of deduction for receiving Equation (21).

The necessary and sufficient condition for receiving an invertible $\Delta(\bar{x})$ is

$$|P^{4\times4} \cdot \Delta(\bar{x}) \cdot Q^{4\times4}| \neq 0, (P \text{ and } Q \text{ are invertible}). \quad (22)$$

Before finding P and Q, we start with calculating $\Delta(\bar{x})$ analytically in Equation (16). Firstly, ${y_1}^{(4)}$, ${y_2}^{(4)}$, ${y_3}^{(4)}$, and $\ddot{y}_4$ are to be calculated.

${y_1}^{(4)}$ is calculated as follows:

$$\ddot{y}_1 = -\frac{1}{m} \cdot A_1 \cdot \zeta, \quad (23)$$

$${y_1}^{(3)} = -\frac{1}{m} \cdot \dot{A}_1 \cdot \zeta - \frac{1}{m} \cdot A_1 \cdot \xi, \quad (24)$$

$${y_1}^{(4)} = -\frac{1}{m} \cdot \ddot{A}_1 \cdot \zeta - \frac{2}{m} \cdot \dot{A}_1 \cdot \xi - \frac{1}{m} \cdot A_1 \cdot \bar{u}_1. \quad (25)$$

${y_2}^{(4)}$ and ${y_3}^{(4)}$ are deduced in a similar way:

$${y_2}^{(4)} = -\frac{1}{m} \cdot \ddot{A}_2 \cdot \zeta - \frac{2}{m} \cdot \dot{A}_2 \cdot \xi - \frac{1}{m} \cdot A_2 \cdot \bar{u}_1, \quad (26)$$

$${y_3}^{(4)} = -\frac{1}{m} \cdot \ddot{A}_3 \cdot \zeta - \frac{2}{m} \cdot \dot{A}_3 \cdot \xi - \frac{1}{m} \cdot A_3 \cdot \bar{u}_1. \quad (27)$$

$\ddot{y}_4$ is deduced in Equation (28):

$$\begin{aligned} \ddot{\psi} &= \left(\frac{s\phi}{c\theta} \cdot q + \frac{c\phi}{c\theta} \cdot r\right)' \\ &= A_4(\phi, \theta, p, q, r) + \frac{s\phi}{c\theta} \cdot \frac{d}{I_y} \cdot \bar{u}_3 + \frac{c\phi}{c\theta} \cdot \frac{d}{I_z} \cdot \bar{u}_4 \end{aligned} \quad (28)$$

where $A_4(\phi, \theta, p, q, r)$ are the remaining terms without containing $\bar{u}_3$ or $\bar{u}_4$.

Notice that $A_1$, $A_2$, and $A_3$ are the functions of $(\psi, \theta, \text{and } \phi)$. Thus, $\dot{A}_1, \dot{A}_2, \text{and } \dot{A}_3$ are the functions of the state $\bar{x}$ in Equation (10), containing no $\bar{u}_1$, $\bar{u}_2$, or $\bar{u}_3$. Consequently, the terms $-2/m \cdot \dot{A}_i \cdot \xi$ in Equations (25)–(27) do not contribute to the coefficients of $\bar{u}_1, \bar{u}_2$, and $\bar{u}_3$.

On the other hand, $\ddot{A}_1$, $\ddot{A}_2$, and $\ddot{A}_3$ are calculated by differentiating the state $\bar{x}$ in Equation (10), which generates $\bar{u}_1, \bar{u}_2, \text{and } \bar{u}_3$. Notice the forms of Equations (25)–(28); calculating $\ddot{A}_1, \ddot{A}_2, \text{and } \ddot{A}_3$ is necessary to receive the coefficients of $\bar{u}_1, \bar{u}_2, \text{and } \bar{u}_3$. In the following, we found the terms containing $\bar{u}_1$, $\bar{u}_2$, and $\bar{u}_3$ in $\ddot{A}_1, \ddot{A}_2, \text{and } \ddot{A}_3$.

We start with finding $\ddot{A}_1$.

Firstly, $\dot{A}_1$ is calculated as

$$\begin{aligned} \dot{A}_1 = -s\psi \cdot s\theta \cdot c\phi \cdot \dot{\psi} + c\psi \cdot c\theta \cdot c\phi \cdot \dot{\theta} - c\psi \cdot s\theta \cdot s\phi \cdot \dot{\phi} + c\psi \cdot s\phi \cdot \dot{\psi} + s\psi \\ \cdot c\phi \cdot \dot{\phi}. \end{aligned} \quad (29)$$

Substituting $\dot{\psi}$, $\dot{\theta}$, and $\dot{\phi}$ in Equation (9), we receive

$$\dot{A}_1 = M_1 \cdot \left(\frac{s\phi}{c\theta}\cdot q+\frac{c\phi}{c\theta}\cdot r\right) + N_1 \cdot (p+s\phi\cdot t\theta\cdot q+c\phi\cdot t\theta\cdot r) + O_1 \cdot (c\phi\cdot q-s\phi\cdot r) \quad (30)$$

where $M_1, N_1,$ and $O_1$ are defined as

$$M_1 = c\psi \cdot s\phi - s\psi \cdot s\theta \cdot s\phi, \quad (31)$$

$$N_1 = s\psi \cdot c\phi - c\psi \cdot s\theta \cdot s\phi, \quad (32)$$

$$O_1 = c\psi \cdot c\theta \cdot c\phi. \quad (33)$$

Further, we calculated the terms containing $(\dot{p}, \dot{q}, \dot{r})$ in $\ddot{A}_1$ based on the result of $\dot{A}_1$ in Equation (30). The sum of the terms containing $\dot{p}$, $\dot{q}$, and $\dot{r}$, which generate $\bar{u}_2$, $\bar{u}_3$, and $\bar{u}_4$, respectively, in $\ddot{A}_1$ is illustrated as

$$R_{A_1} = N_1 \cdot \dot{p} + \left(M_1\cdot\frac{s\phi}{c\theta}+N_1\cdot s\phi\cdot t\theta+O_1\cdot c\phi\right) \cdot \dot{q} + \left(M_1\cdot\frac{c\phi}{c\theta}+N_1\cdot c\phi\cdot t\theta-O_1\cdot s\phi\right) \cdot \dot{r}. \quad (34)$$

As a result, the coefficient of $\bar{u}_1$ in Equation (25) is $-1/m \cdot A_1$. The coefficient of $\bar{u}_2$ in Equation (25) confirmed by Equation (34) is $-\zeta/m \cdot N_1 \cdot \frac{d}{I_x}$. The coefficient of $\bar{u}_3$ in Equation (25) confirmed by Equation (34) is $-\zeta/m \cdot (M_1\cdot s\phi/c\theta+N_1\cdot s\phi\cdot t\theta+O_1\cdot c\phi) \cdot d/I_y$. The coefficient of $\bar{u}_4$ in Equation (25) confirmed by Equation (34) is $-\zeta/m \cdot (M_1\cdot c\phi/c\theta+N_1\cdot c\phi\cdot t\theta-O_1\cdot s\phi) \cdot d/I_z$. So far, we have received the results of the coefficients of $\bar{u}_1, \bar{u}_2, \bar{u}_3,$ and $\bar{u}_4$ in ${y_1}^{(4)}$.

A similar procedure is conducted to calculate the coefficients of $\bar{u}_1, \bar{u}_2, \bar{u}_3,$ and $\bar{u}_4$ in ${y_2}^{(4)}$ in Equation (26). It starts with finding

$$\dot{A}_2 = M_2 \cdot \left(\frac{s\phi}{c\theta}\cdot q+\frac{c\phi}{c\theta}\cdot r\right) + N_2 \cdot (p+s\phi\cdot t\theta\cdot q+c\phi\cdot t\theta\cdot r) + O_2 \cdot (c\phi\cdot q-s\phi\cdot r) \quad (35)$$

where $M_2, N_2,$ and $O_2$ are defined as

$$M_2 = s\psi \cdot s\phi + c\psi \cdot s\theta \cdot s\phi, \quad (36)$$

$$N_2 = -c\psi \cdot c\phi - s\psi \cdot s\theta \cdot s\phi, \quad (37)$$

$$O_2 = s\psi \cdot c\theta \cdot c\phi. \quad (38)$$

Further, we calculated the terms containing $(\dot{p}, \dot{q}, \dot{r})$ in $\ddot{A}_2$ based on the result of $\dot{A}_2$ in Equation (35). The sum of the terms containing $\dot{p}$, $\dot{q}$, and $\dot{r}$, which generate $\bar{u}_2$, $\bar{u}_3$, and $\bar{u}_4$, respectively, in $\ddot{A}_2$ is

$$R_{A_2} = N_2 \cdot \dot{p} + \left(M_2\cdot\frac{s\phi}{c\theta}+N_2\cdot s\phi\cdot t\theta+O_2\cdot c\phi\right) \cdot \dot{q} + \left(M_2\cdot\frac{c\phi}{c\theta}+N_2\cdot c\phi\cdot t\theta-O_2\cdot s\phi\right) \cdot \dot{r}. \quad (39)$$

As a result, the coefficient of $\bar{u}_1$ in Equation (26) is $-1/m \cdot A_2$. The coefficient of $\bar{u}_2$ in Equation (26) confirmed by Equation (39) is $-\zeta/m \cdot N_2 \cdot d/I_x$. The coefficient of $\bar{u}_3$ in Equation (26) confirmed by Equation (39) is $-\zeta/m \cdot (M_2\cdot s\phi/c\theta+N_2\cdot s\phi\cdot t\theta+O_2\cdot c\phi) \cdot d/I_y$. The coefficient of $\bar{u}_4$ in Equation (26) confirmed by Equation (39) is $-\zeta/m \cdot (M_2\cdot c\phi/c\theta+N_2\cdot c\phi\cdot t\theta-O_2\cdot s\phi) \cdot d/I_z$.

A similar procedure is conducted to calculate the coefficients of $\bar{u}_1, \bar{u}_2, \bar{u}_3,$ and $\bar{u}_4$ in ${y_3}^{(4)}$ in Formula (27).

As a result, the coefficient of $\bar{u}_1$ in Equation (27) is $-1/m \cdot A_3$. The coefficient of $\bar{u}_2$ in Equation (27) is $-\zeta/m \cdot N_3 \cdot d/I_x$. The coefficient of $\bar{u}_3$ in Equation (27) is $-\zeta/m \cdot (N_3\cdot s\phi\cdot t\theta+O_3\cdot c\phi) \cdot d/I_y$. The coefficient of $\bar{u}_4$ in Equation (27) is $-\zeta/m \cdot (N_3\cdot c\phi\cdot t\theta-O_3\cdot s\phi) \cdot d/I_z$. $N_3$ and $O_3$ are defined as

$$N_3 = -c\theta \cdot s\phi, \quad (40)$$

$$O_3 = -s\theta \cdot c\phi. \quad (41)$$

As for the coefficients of $\bar{u}_1, \bar{u}_2, \bar{u}_3,$ and $\bar{u}_4$ in $\ddot{y}_4$, we can find them in Equation (28). As a result, the coefficient of $\bar{u}_1$ in Equation (28) is 0. The coefficient of $\bar{u}_2$ in Equation (28) is 0. The coefficient of $\bar{u}_3$ in Equation (28) is $s\phi/c\theta \cdot d/I_y$. The coefficient of $\bar{u}_4$ in Equation (28) is $c\phi/c\theta \cdot d/I_z$.

So far, we have found all the coefficients of $\bar{u}_1$, $\bar{u}_2$, $\bar{u}_3$, and $\bar{u}_4$ in ${y_1}^{(4)}, {y_2}^{(4)}, {y_3}^{(4)},$ and $\ddot{y}_4$. Thus, the decoupling matrix (delta matrix), $\Delta(\bar{x})$, in Equation (20) is found. That is

$$\Delta(\bar{x}) = \begin{bmatrix} \Delta_{11} & \Delta_{12} & \Delta_{13} & \Delta_{14} \\ \Delta_{21} & \Delta_{22} & \Delta_{23} & \Delta_{24} \\ \Delta_{31} & \Delta_{32} & \Delta_{33} & \Delta_{34} \\ \Delta_{41} & \Delta_{42} & \Delta_{43} & \Delta_{44} \end{bmatrix} \tag{42}$$

where

$$\Delta_{11} = -\frac{1}{m} \cdot A_1,$$

$$\Delta_{21} = -\frac{1}{m} \cdot A_2,$$

$$\Delta_{31} = -\frac{1}{m} \cdot A_3,$$

$$\Delta_{41} = 0,$$

$$\Delta_{12} = -\frac{\zeta}{m} \cdot N_1 \cdot \frac{d}{I_x},$$

$$\Delta_{22} = -\frac{\zeta}{m} \cdot N_2 \cdot \frac{d}{I_x},$$

$$\Delta_{32} = -\frac{\zeta}{m} \cdot N_3 \cdot \frac{d}{I_x},$$

$$\Delta_{42} = 0,$$

$$\Delta_{13} = -\frac{\zeta}{m} \cdot \left(M_1 \cdot \frac{s\phi}{c\theta} + N_1 \cdot s\phi \cdot t\theta + O_1 \cdot c\phi\right) \cdot \frac{d}{I_y},$$

$$\Delta_{23} = -\frac{\zeta}{m} \cdot \left(M_2 \cdot \frac{s\phi}{c\theta} + N_2 \cdot s\phi \cdot t\theta + O_2 \cdot c\phi\right) \cdot \frac{d}{I_y},$$

$$\Delta_{33} = -\frac{\zeta}{m} \cdot (N_3 \cdot s\phi \cdot t\theta + O_3 \cdot c\phi) \cdot \frac{d}{I_y},$$

$$\Delta_{43} = \frac{s\phi}{c\theta} \cdot \frac{d}{I_y},$$

$$\Delta_{14} = -\frac{\zeta}{m} \cdot \left(M_1 \cdot \frac{c\phi}{c\theta} + N_1 \cdot c\phi \cdot t\theta - O_1 \cdot s\phi\right) \cdot \frac{d}{I_z},$$

$$\Delta_{24} = -\frac{\zeta}{m} \cdot \left(M_2 \cdot \frac{c\phi}{c\theta} + N_2 \cdot c\phi \cdot t\theta - O_2 \cdot s\phi\right) \cdot \frac{d}{I_z},$$

$$\Delta_{34} = -\frac{\zeta}{m} \cdot (N_3 \cdot c\phi \cdot t\theta - O_3 \cdot s\phi) \cdot \frac{d}{I_z},$$

$$\Delta_{44} = \frac{c\phi}{c\theta} \cdot \frac{d}{I_z}.$$

It can be concluded from this that $\zeta$ should be nonzero if we expect the delta matrix to be a full rank.

The next step is to find a proper invertible matrix $P^{4\times4}$ and $Q^{4\times4}$ in Equation (22). This step is designed to simplify the delta matrix to the form that we can calculate the determinant based on easily.

Notice that

$$P^{4\times4} \cdot \Delta(\bar{x}) \cdot Q^{4\times4} = \begin{bmatrix} D_{11} & s\psi \cdot c\phi & c\psi & 0 \\ D_{21} & -c\psi \cdot c\phi & s\psi & 0 \\ 0 & s\phi & s\theta & 0 \\ 0 & 0 & 0 & 1 \end{bmatrix} \tag{43}$$

where

$$D_{11} = c\psi \cdot s\theta \cdot c\phi - c\psi \cdot s\theta \cdot s\phi + s\psi \cdot s\phi + s\psi \cdot c\phi,$$

$$D_{21} = s\psi \cdot s\theta \cdot c\phi - s\psi \cdot s\theta \cdot s\phi - c\psi \cdot s\phi - c\psi \cdot c\phi,$$

$$P = P_1 \cdot P_2 \cdot P_3,$$

$$P_1 = \begin{bmatrix} 1 & 0 & 0 & 0 \\ 0 & 1 & 0 & 0 \\ 0 & 0 & -c\theta & 0 \\ 0 & 0 & 0 & 1 \end{bmatrix},$$

$$P_2 = \begin{bmatrix} 1 & 0 & 0 & s\phi \cdot s\psi \cdot c\theta \\ 0 & 1 & 0 & s\phi \cdot s\psi \cdot c\theta \\ 0 & 0 & 1 & -s\phi \cdot s\theta \\ 0 & 0 & 0 & 1 \end{bmatrix},$$

$$P_3 = \begin{bmatrix} 1 & 0 & 0 & \frac{\zeta}{m} \cdot c\theta \\ 0 & 1 & 0 & \frac{\zeta}{m} \cdot c\theta \\ 0 & 0 & 1 & \frac{\zeta}{m} \cdot c\theta \\ 0 & 0 & 0 & -\frac{\zeta}{m} \cdot c\theta \end{bmatrix},$$

$$Q = Q_1 \cdot Q_2 \cdot Q_3,$$

$$Q_1 = \begin{bmatrix} -m & 0 & 0 & 0 \\ 0 & -\frac{m \cdot I_x}{\zeta \cdot d} & 0 & 0 \\ 0 & 0 & -\frac{m \cdot I_y}{\zeta \cdot d} & 0 \\ 0 & 0 & 0 & -\frac{m \cdot I_z}{\zeta \cdot d} \end{bmatrix},$$

$$Q_2 = \begin{bmatrix} 1 & 0 & 0 & 0 \\ 0 & 1 & 0 & 0 \\ 0 & 0 & 1 & 0 \\ 0 & 0 & -\frac{s\phi}{c\phi} & 1 \end{bmatrix},$$

$$Q_3 = \begin{bmatrix} 1 & 0 & 0 & 0 \\ 1 & 1 & 0 & 0 \\ 0 & \frac{s\theta \cdot s\phi}{c\theta} & \frac{1}{c\theta} & 0 \\ 0 & 0 & 0 & \frac{1}{c\phi} \end{bmatrix}.$$

Further,

$$|P^{4\times4} \cdot \Delta(\bar{x}) \cdot Q^{4\times4}| = -1 + \cos^2\theta \cdot \cos^2\phi - \cos^2\theta \cdot \cos\phi \cdot \sin\phi. \quad (44)$$

Thus, $\Delta(\bar{x})$ is invertible if and only if the right side of Equation (44) is nonzero, which is exactly Equation (21). This completes the proof.

## 4. Visualize The Singular Zone

The singular space is determined by

$$|P^{4\times4} \cdot \Delta(\bar{x}) \cdot Q^{4\times4}| = -1 + \cos^2\theta \cdot \cos^2\phi - \cos^2\theta \cdot \cos\phi \cdot \sin\phi. \quad (45)$$

This space is defined by $\phi$ and $\theta$, which are not controlled independently; the independently controlled variables selected in Equation (19) do not include the roll and pitch. Consequently, avoiding the singular space defined in Equation (45) can be not straightforward or even not feasible for some trajectories.

We visualize the singular space by defining the "determinant surface":

$$S(\theta, \phi) = -1 + \cos^2\theta \cdot \cos^2\phi - \cos^2\theta \cdot \cos\phi \cdot \sin\phi. \tag{46}$$

We plot $S(\theta, \phi)$ in Figure 1 (3D) and Figure 2 (contour plot).

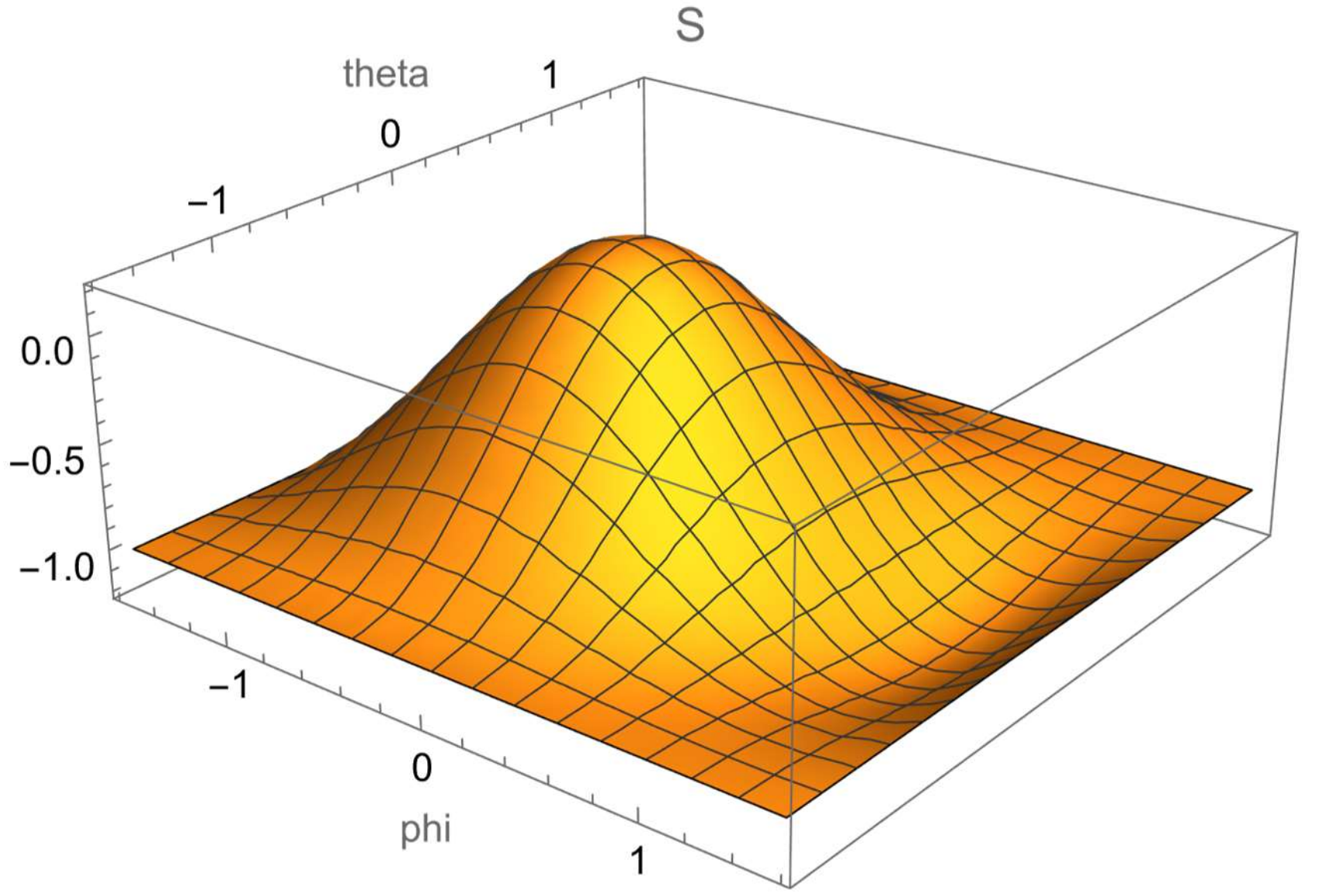

**Figure 1.** Three-dimensional plot of the determinant surface, $S(\theta, \phi)$.

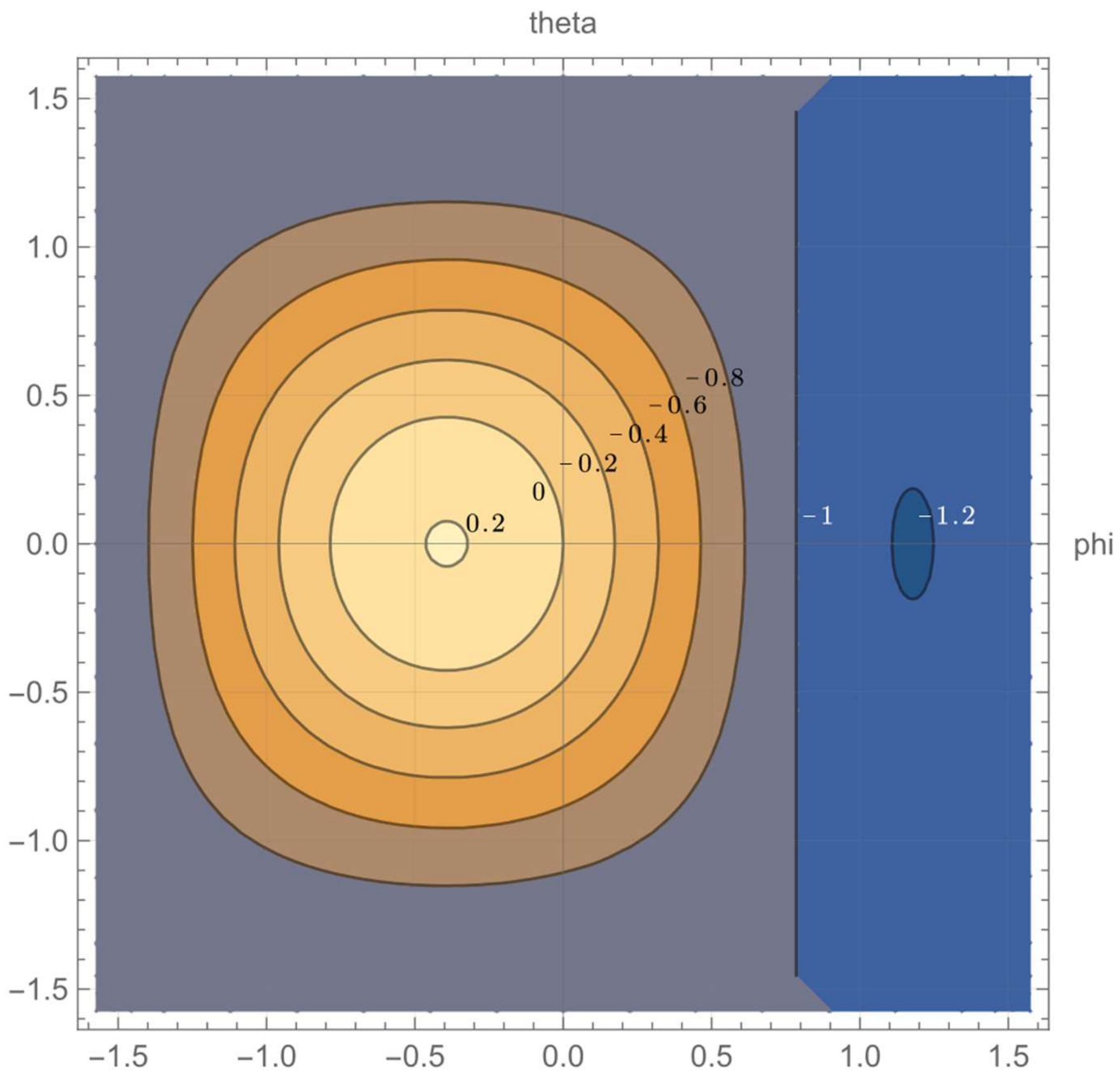

**Figure 2.** Contour plot of the determinant surface, $S(\theta, \phi)$.

Figure 3 plots the singular curve, given $S(\theta, \phi) = 0$, where the attitudes on the curve introduce the singular decoupling matrix. Clearly, there exists a singular region within the space of interest, $\phi \in (-\pi/2, \pi/2), \theta \in (-\pi/2, \pi/2)$.

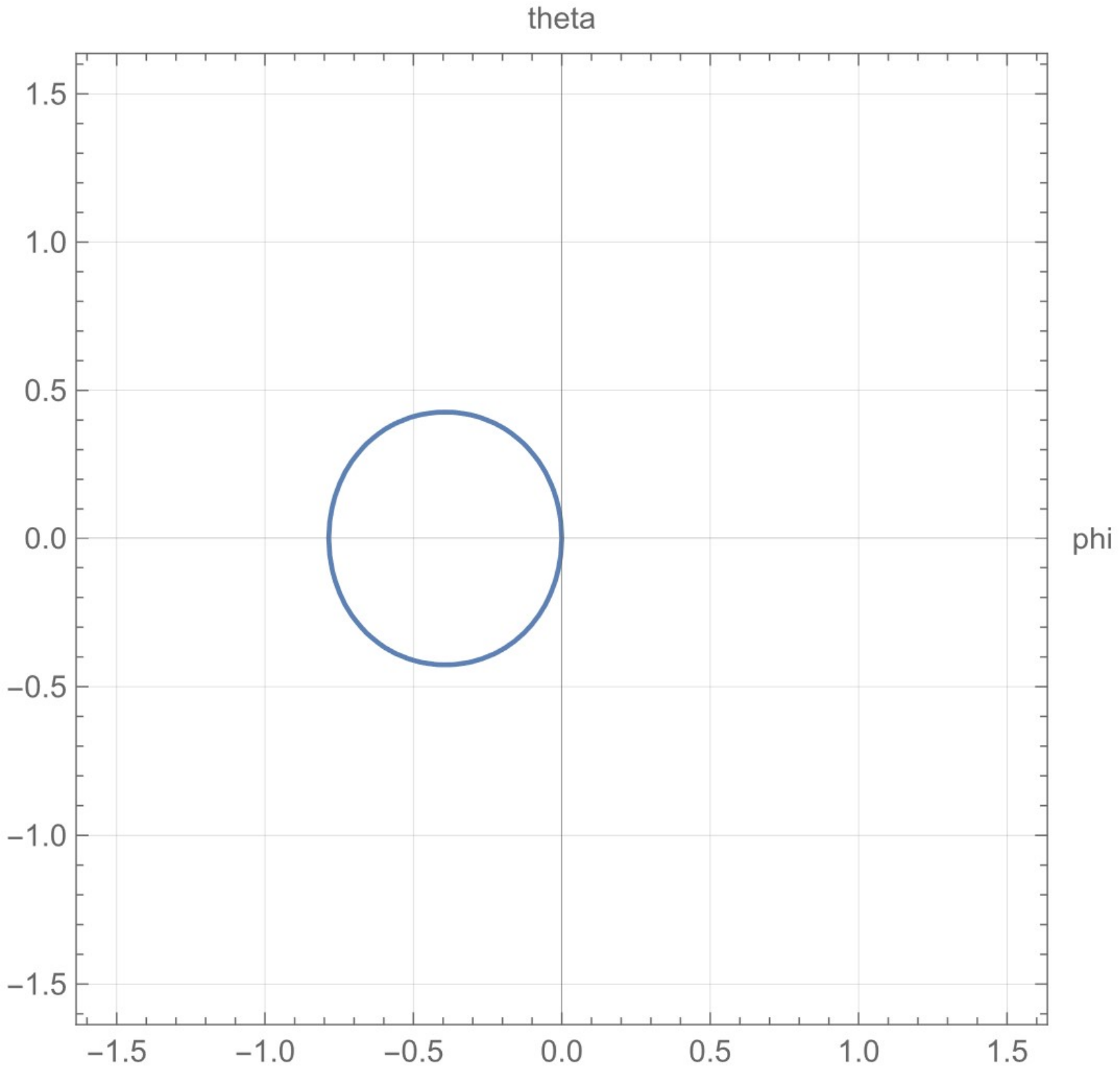

**Figure 3.** The singular curve satisfying $S(\theta, \phi) = 0$.

To avoid touching the singular curve, the quadrotor is either outside the "circular" area in Figure 3 or inside it throughout the entire flight. Crossing the "circular" area is strictly prohibited, since it causes the noninvertible problem in the decoupling matrix.

However, this requirement can be too strict, since we can find in Figure 3 that $(\theta, \phi) = (0,0)$ is on the singular curve, while the quadrotor is near this state during most flights or even on it, e.g., while hovering.

It is worth mentioning that this result is not influenced by changing the definition of the positive direction of the body-fixed frame; changing the positive direction of the body-fixed frame (e.g., Reference [38]) changes the signs in several rows in the delta matrix, which has no influence on the rank of a matrix.

## 5. Switch Controller

This section puts forward an approach attempting to avoid the above singular zone by switching to another controller when the state is near the singular curve.

Though the previous analysis showed that yaw–position output-based feedback linearization introduces a singular zone, attitude–altitude output-based ($\phi,\theta,\psi,z$) feedback linearization does not (see Reference [32]). Inspired by this, we replace the control rule near the singular area.

As illustrated in Figure 4, the attitude lies either within the purple zone or the yellow zone during the entire flight based on the independent output choice $(x, y, z, \psi)$. Crossing the singular curve from the yellow zone to the purple zone is prohibited and vice versa.

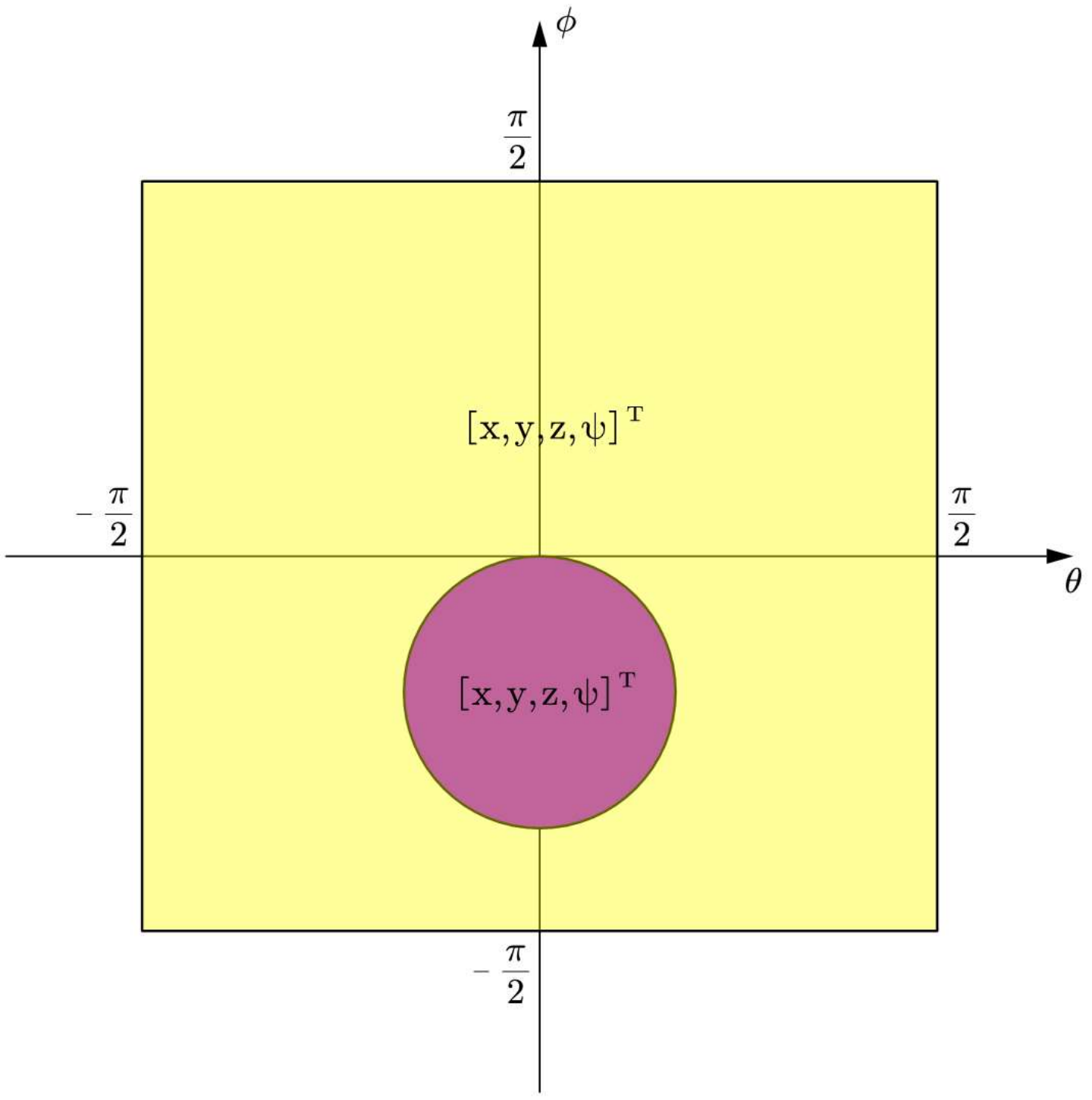

**Figure 4.** The independent output choice $(x, y, z, \psi)$ introduces the singular curve.

To avoid touching the singular curve, we change the controller in some attitudes (Figure 5). In the orange zone, we use another independent output choice $(\phi, \theta, \psi, z)$ in the feedback linearization, while the independent output choice for the rest of the attitude area (yellow zone) remains unchanged $(x, y, z, \psi)$. Based on this, the singular curve in Figure 4 is eliminated in the entire attitude area. The attitude is allowed to change within the yellow or orange areas or to cross from one to the other.

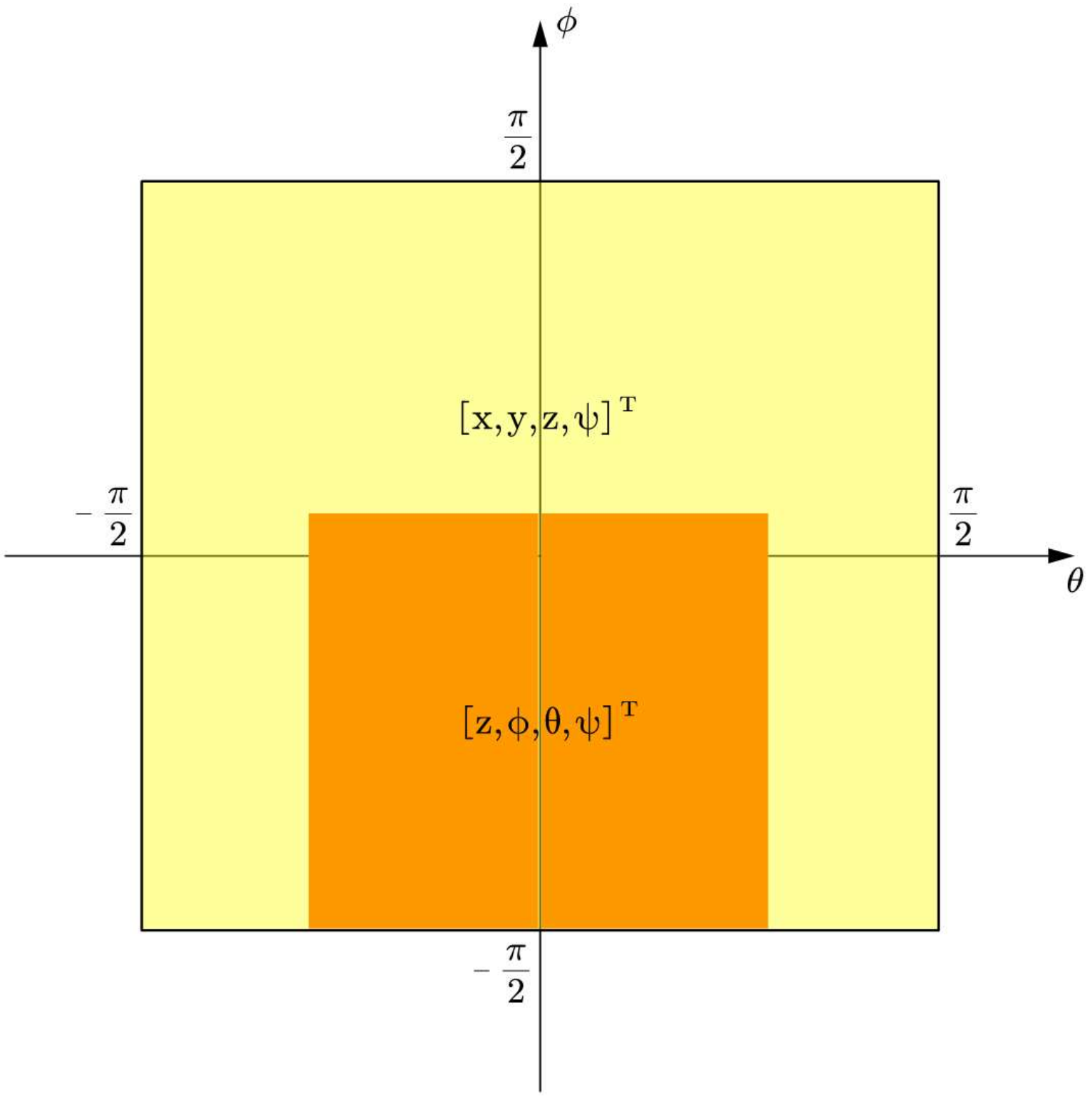

**Figure 5.** The independent output choice in the orange area is $(\phi, \theta, \psi, z)$. The independent output choice in the yellow area is $(x, y, z, \psi)$.

The attitude zone where the output $(\phi, \theta, \psi, z)$ is chosen, the orange zone in Figure 5, is determined by

$$\begin{cases} -0.5 \leqslant \theta \leqslant 0.5 \\ -\frac{\pi}{2} \leqslant \phi \leqslant 0.2 \end{cases}. \tag{47}$$

The Simulink block is pictured in Figure 6, where the relevant parts are noted. The parameters of the quadrotor in the simulator are specified in Appendix A (Table A1). The adopted controllers are PD controllers for both the altitude–attitude-linearized system and the position–yaw-linearized system. The design and the parameters of the PD controller are demonstrated in Appendix B.

The initial attitude (roll, pitch, and yaw) is set as: $(\phi_0, \theta_0, \psi_0) = (0.5, 0.5, 0)$.

In the first experiment, the position reference and the attitude reference for two controllers are marked in Figure 7. The altitude–attitude reference is set as $(z_r, \phi_r, \theta_r, \psi_r) = (0, 0.01, 0, 0)$. The altitude reference and the yaw reference in the independent choice $(x_r, y_r, z_r, \psi_r)$ are set as $z_r = \psi_r = 0$, while $x_r$ and $y_r$ are specified in Appendix B.

Figure 8 sketches the results of the attitude trajectory. It can be seen that attitude $(\phi, \theta)$ starts from (0.5,0.5) under the dominance of the yaw–position controller (yellow zone). After some time, it enters the orange zone governed by the altitude–attitude controller and is captured by this zone before being stabilized at $(\phi, \theta) = (0.01, 0)$.

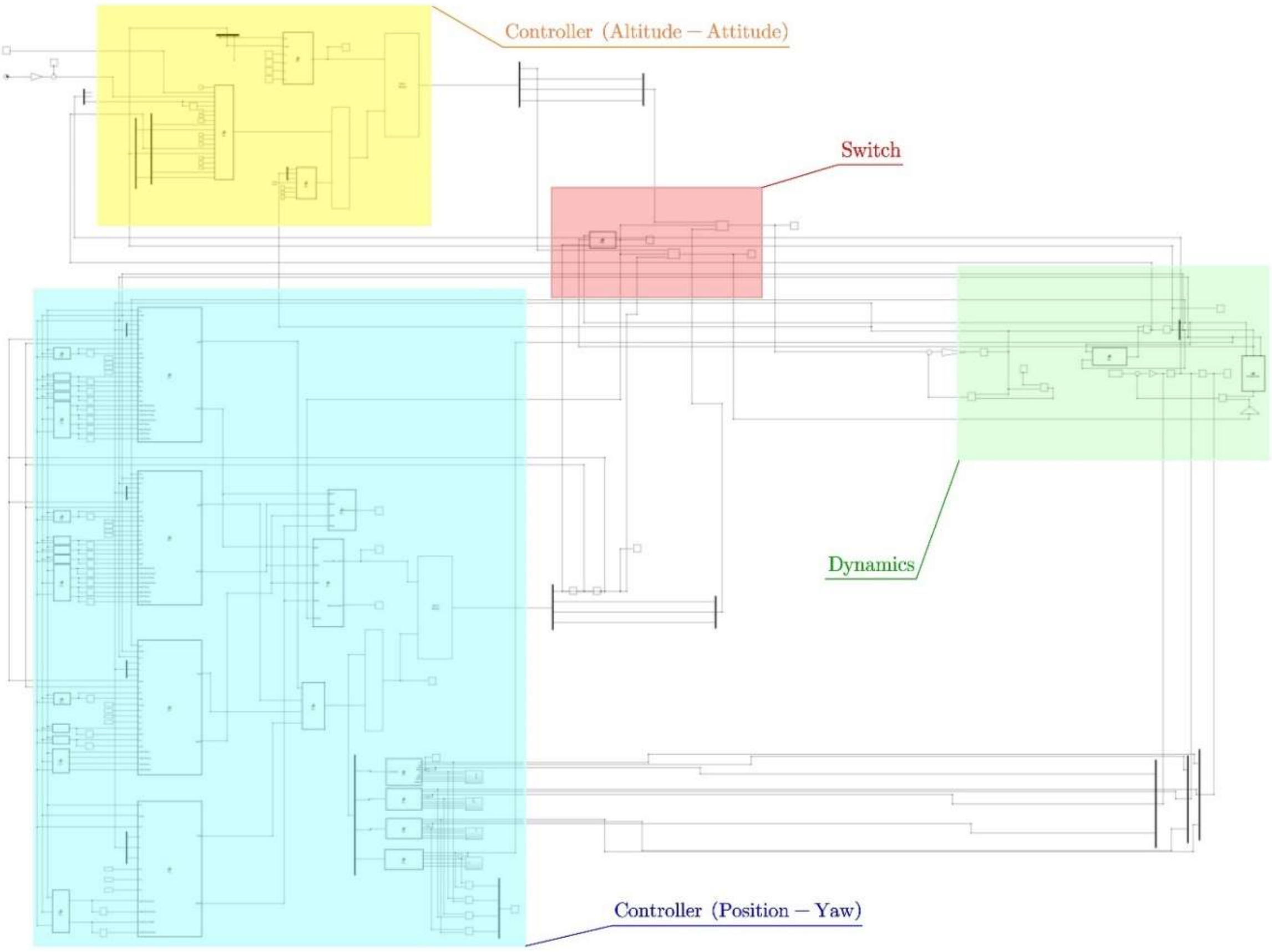

**Figure 6.** The Simulink diagram for the modified controller equipped system.

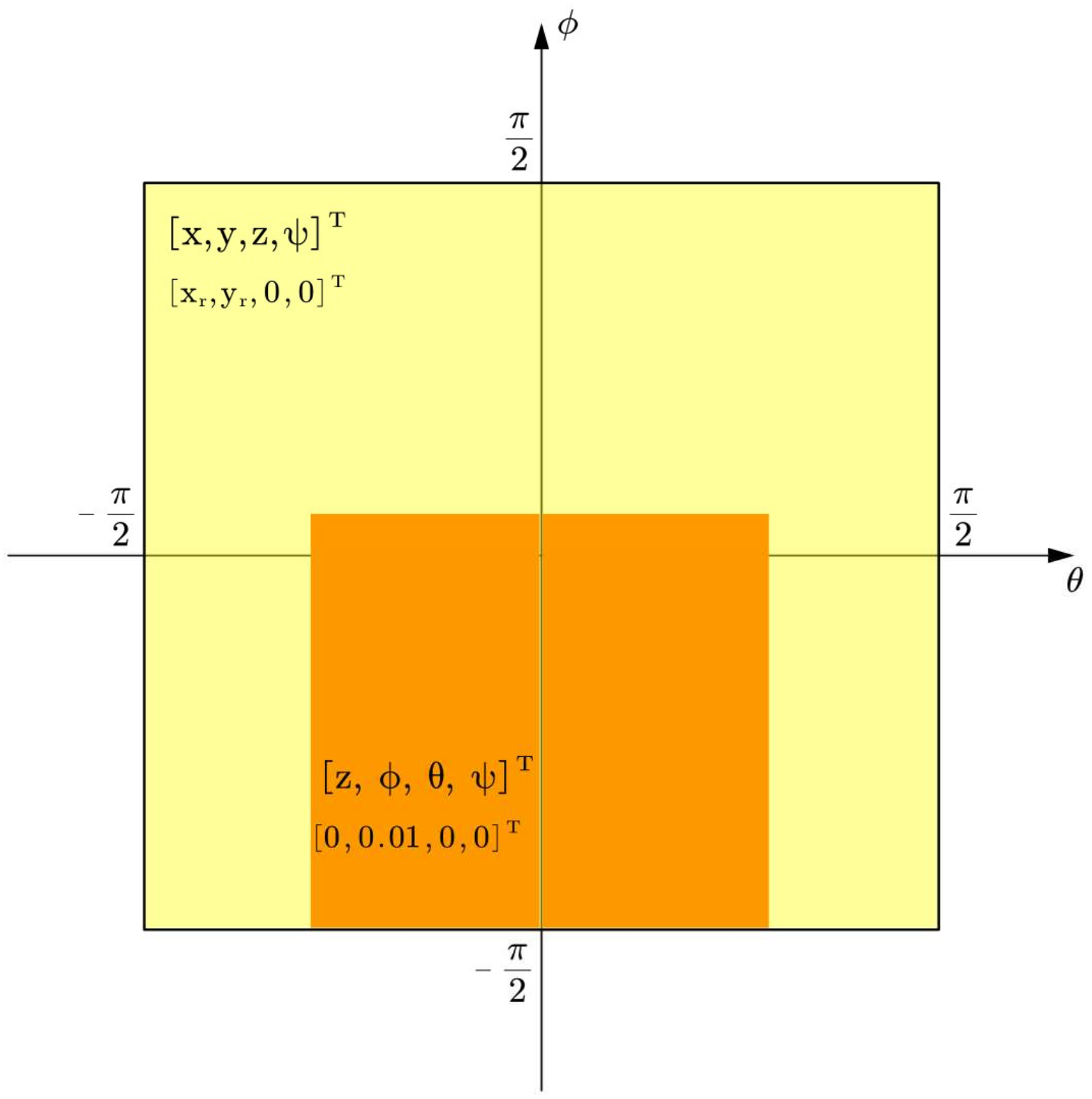

**Figure 7.** The altitude–attitude reference is set as $(z_r,\phi_r,\theta_r,\psi_r) = (0{,}0.01{,}0{,}0)$. The altitude reference and the yaw reference in the independent choice $(x_r,y_r,z_r,\psi_r)$ are set as $z_r = \psi_r = 0$.

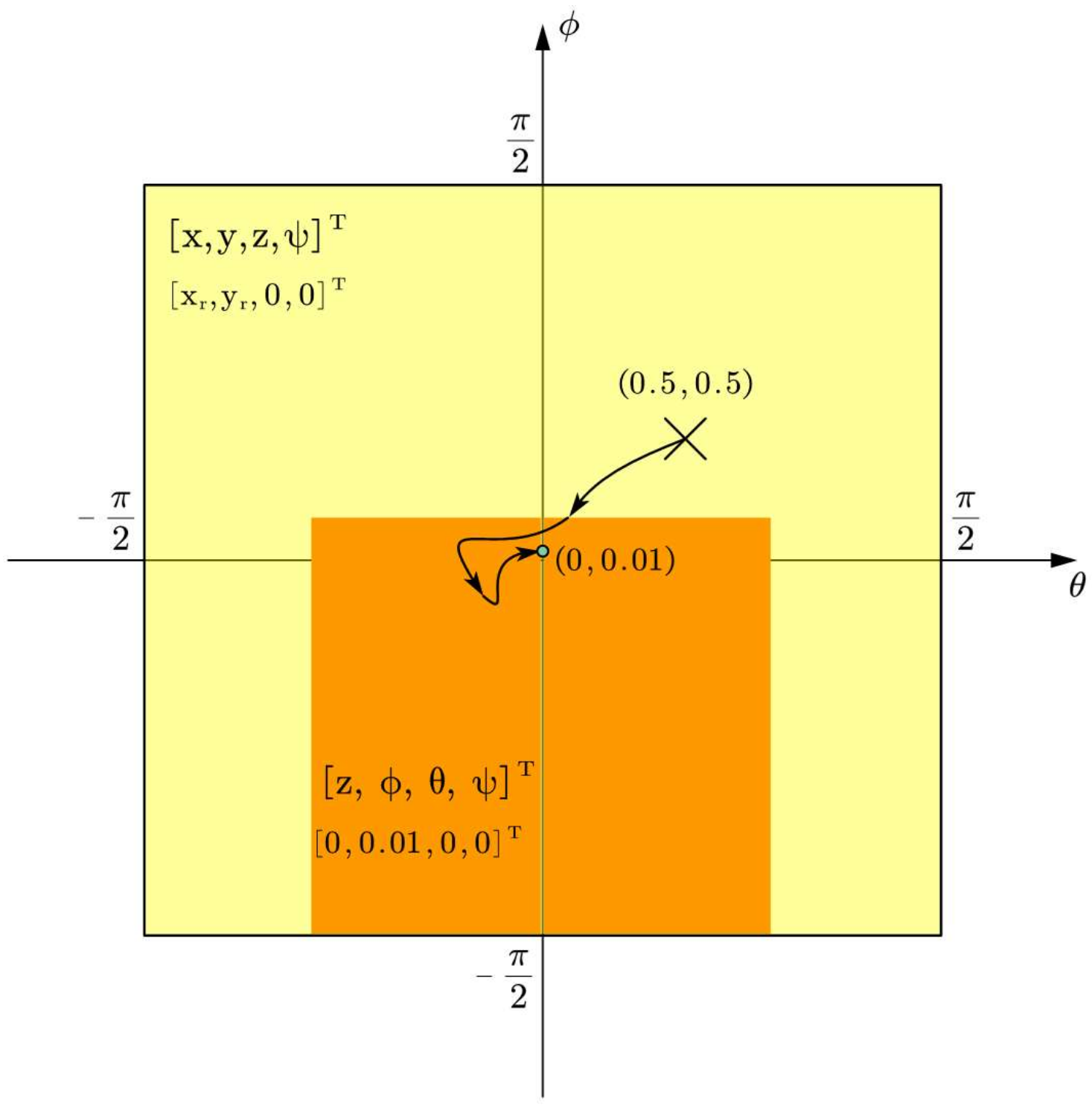

**Figure 8.** The attitude history of the experiment (results).

In the second experiment, the attitude is expected to be stabilized at $(\phi_f,\theta_f) = (0.5,-0.5)$. Thus, we set the references for both controllers in Figure 9. The altitude–attitude reference is set as $(z_r,\phi_r,\theta_r,\psi_r) = (0,-0.5,0.5,0)$. The altitude reference and the yaw reference in the independent choice $(x_r,y_r,z_r,\psi_r)$ are set as $z_r = \psi_r = 0$.

Figure 10 sketches the results of the attitude trajectory.

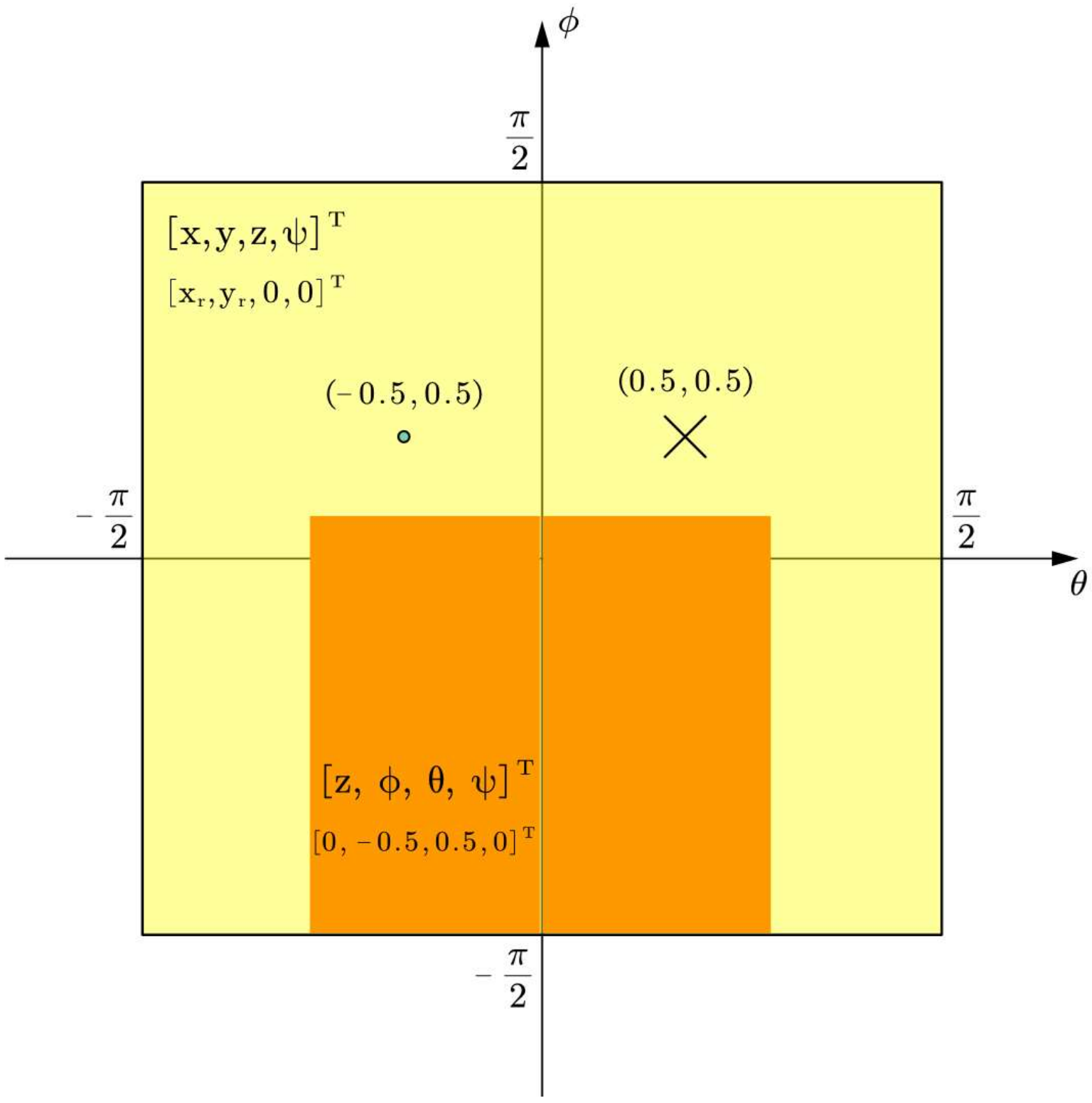

**Figure 9.** The Simulink diagram for the modified controller equipped system.

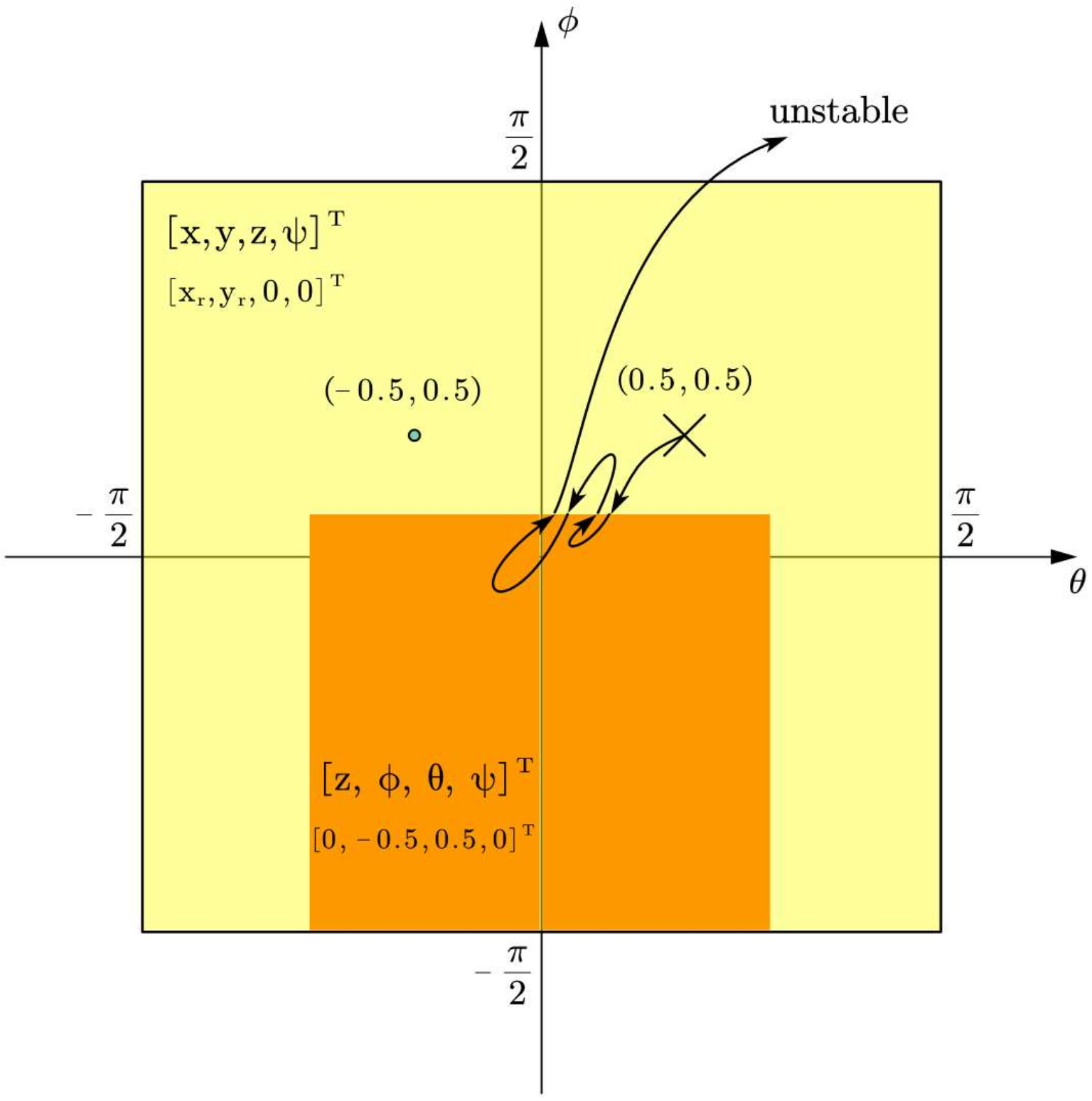

**Figure 10.** The Simulink diagram for the modified controller equipped system.

The attitude witnesses several switches in controllers before finally becoming unstable. Although the altitude–attitude control zone (orange zone) tries to let the attitude reach $(\phi_f,\theta_f)$ =

(0.5,−0.5), the altitude–attitude controller is switched to the yaw–position controller once the attitude escapes the orange zone. The yaw–position controller has no effect on driving the attitude roll and pitch, $\phi$ and $\theta$. Thus, the attitude in the yellow zone can escape the area of interest, $\phi \in (-\pi/2, \pi/2), \theta \in (-\pi/2, \pi/2)$, becoming unstable. The relevant attitude signal history is plotted in Figure 11.

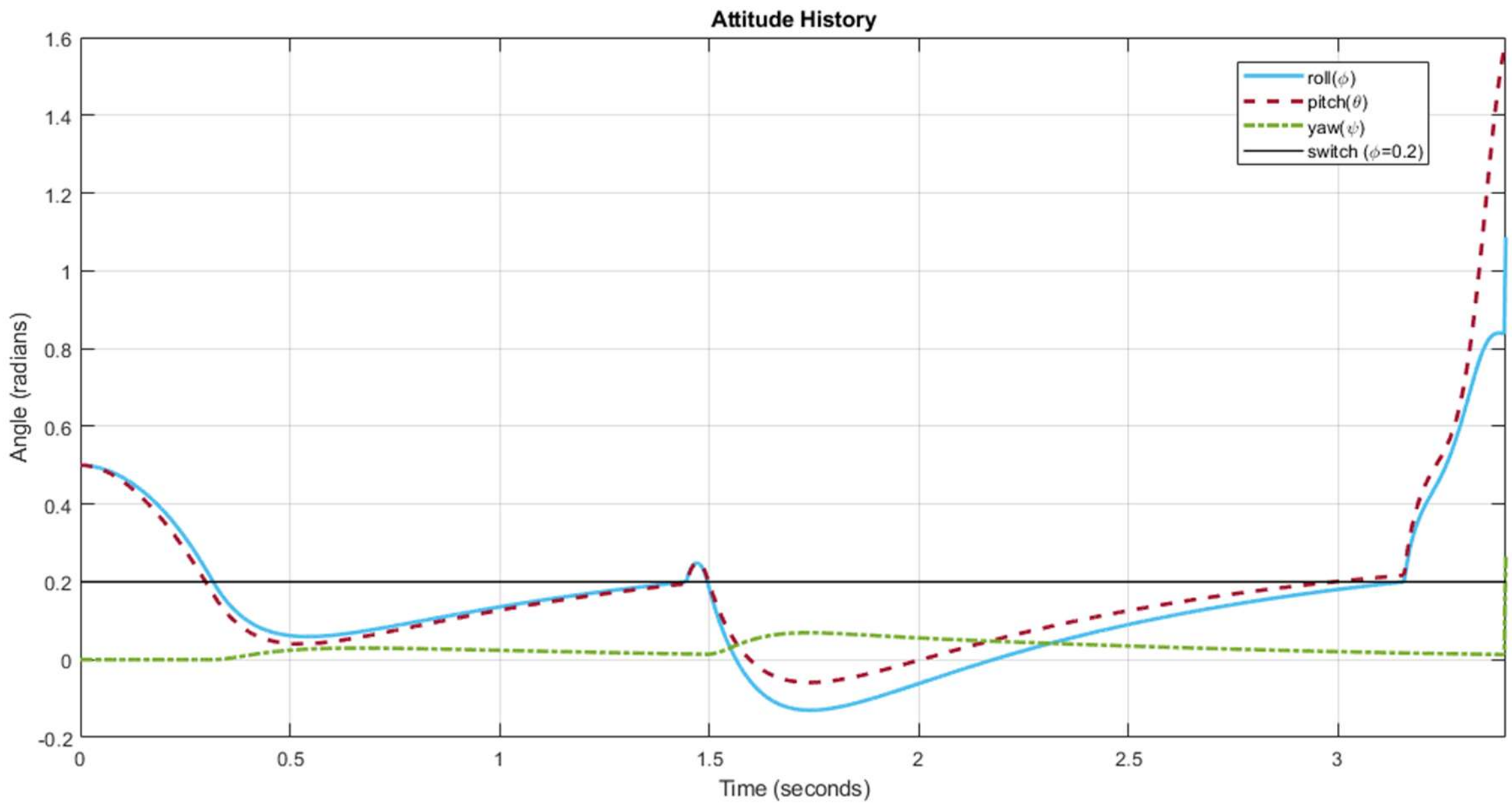

**Figure 11.** The attitude history of the quadrotor before being unstable.

## 6. Conclusions and Discussions

The decoupling matrix in feedback linearization is not always invertible in the entire attitude zone for the yaw–position output combination. Thus, the relevant controller risks encountering the singular decoupling matrix that causes the failure in controlling.

The attitude causing the singularity in the decoupling matrix was presented analytically and visualized in this research.

Substituting part of the yaw–position-based feedback linearization with altitude–attitude-based feedback linearization can avoid the noninvertible problem. The attitude is finally stabilized within the region governed by the altitude–attitude controller. Attempts in driving the attitude to a reference inside the region dominated by the yaw–position controller can be unstable and was not realized in this research.

Figure 12 demonstrates the statue of this work compared with other research discussing the property of the decoupling matrix in the quadrotor while implementing the exact linearization.

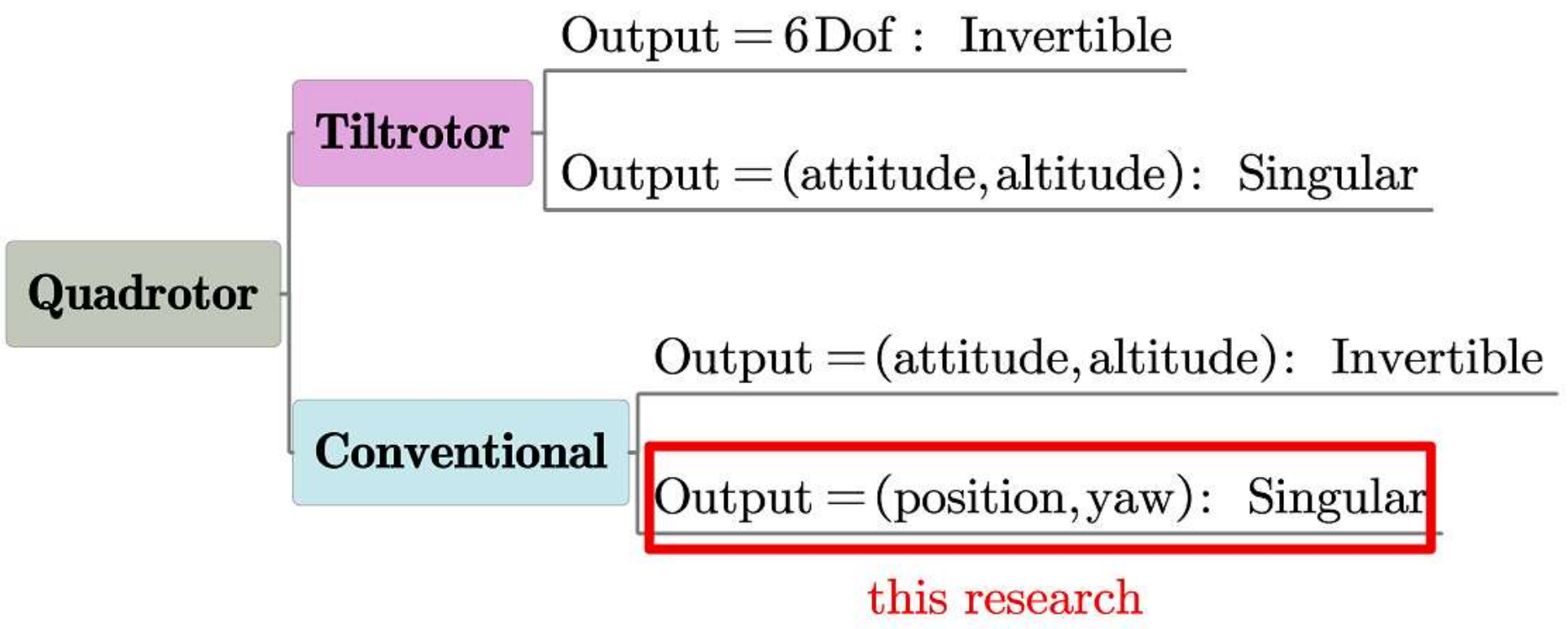

**Figure 12.** The statue of this work.

There are still several questions worth exploring.

Although we report the singular problem in the decoupling matrix in the yaw–position choice in the outputs, the properties of the decoupling matrices in several other combinations of independent outputs are still waiting to be uncovered by research. Actually, there are 15 ($C_6^4$) combinations in total if we pick four outputs as the independent controlled variables. The current research only explored two of them.

Moreover, the number of the combinations increase if we only pick three outputs [35] as the independent controlled variables. There could be 20 ($C_6^3$) cases in total, most of which have not been given attention yet.

Discussing the whole picture of the property of the decoupling matrix for each case is helpful for other researchers working on feedback linearization.

Further, we did not successfully stabilize the attitude in the attitude zone governed by the yaw–position controller while applying the exact feedback linearization. Designing a control method that directly avoids this singular space is also a further step.

## Appendix A

The parameters of the quadrotor in our simulator are specified in the following table (Table A1).

**Table A1.** The parameters of the quadrotor in the simulator.

| Parameter | Value |
|---|---|
| mass | 0.429 kg |
| gravitational acceleration | $9.8\ \mathrm{m/s^2}$ |
| length of the arm | 0.1785 m |
| moment of inertia of the body | $\begin{bmatrix} 2.237568 & & \\ & 2.985236 & \\ & & 4.80374 \end{bmatrix} \cdot 10^{-3}\ \mathrm{kg \cdot m^2}$ |
| coefficient of the thrust | $8.048 \times 10^{-4}\ \mathrm{N \cdot s^2/rad^2}$ |
| coefficient of the drag moment | $8.048 \times 10^{-4}\ \mathrm{N \cdot m \cdot s^2/rad^2}$ |

## Appendix B

The controllers designed to control the two linearized systems are detailed here.

The PD controllers are developed for the position–yaw-linearized system as

$$\begin{bmatrix} \bar{u}_1 \\ \bar{u}_2 \\ \bar{u}_3 \\ \bar{u}_4 \end{bmatrix} = \Delta(\bar{x})^{-1} \cdot \left( \begin{bmatrix} x_d^{(4)} \\ y_d^{(4)} \\ z_d^{(4)} \\ \ddot{\psi}_d \end{bmatrix} - Ma(\bar{x}) \right) \tag{A1}$$

where $x_d^{(4)}, y_d^{(4)}, z_d^{(4)}$, and $\ddot{\psi}_d$ are defined as

$$\begin{aligned} \begin{bmatrix} x_d^{(4)} \\ y_d^{(4)} \\ z_d^{(4)} \\ \ddot{\psi}_d \end{bmatrix} = \begin{bmatrix} x_r^{(4)} \\ y_r^{(4)} \\ z_r^{(4)} \\ \ddot{\psi}_r \end{bmatrix} + k_1 \cdot \left( \begin{bmatrix} \dddot{x}_r \\ \dddot{y}_r \\ \dddot{z}_r \\ 0 \end{bmatrix} - \begin{bmatrix} \dddot{x} \\ \dddot{y} \\ \dddot{z} \\ 0 \end{bmatrix} \right) + k_2 \cdot \left( \begin{bmatrix} \ddot{x}_r \\ \ddot{y}_r \\ \ddot{z}_r \\ 0 \end{bmatrix} - \begin{bmatrix} \ddot{x} \\ \ddot{y} \\ \ddot{z} \\ 0 \end{bmatrix} \right) + k_3 \cdot \left( \begin{bmatrix} \dot{x}_r \\ \dot{y}_r \\ \dot{z}_r \\ \dot{\psi}_r \end{bmatrix} - \begin{bmatrix} \dot{x} \\ \dot{y} \\ \dot{z} \\ \dot{\psi} \end{bmatrix} \right) \\ + k_4 \cdot \left( \begin{bmatrix} x_r \\ y_r \\ z_r \\ \psi_r \end{bmatrix} - \begin{bmatrix} x \\ y \\ z \\ \psi \end{bmatrix} \right) \end{aligned} \tag{A2}$$

where

$$k_1 = \begin{bmatrix} 10 & & & \\ & 10 & & \\ & & 10 & \\ & & & 0 \end{bmatrix}, \quad \text{(A3)}$$

$$k_2 = \begin{bmatrix} 20 & & & \\ & 20 & & \\ & & 20 & \\ & & & 0 \end{bmatrix}, \quad \text{(A4)}$$

$$k_3 = \begin{bmatrix} 50 & & & \\ & 50 & & \\ & & 50 & \\ & & & 5 \end{bmatrix}, \quad \text{(A5)}$$

$$k_4 = \begin{bmatrix} 500 & & & \\ & 500 & & \\ & & 500 & \\ & & & 5 \end{bmatrix}, \quad \text{(A6)}$$

$x_r$, $y_r$, $z_r$, and $\psi_r$ are the references.

$x_r$ and $y_r$ in the position references are set as $1 - \cos(t/20)$ and $-\sin(t/20)$, respectively. This indicates that the quadrotor will follow a circle if it is stabilized by the dominance of the position–yaw-linearized controller.

The remaining references ($z_r$ and $\psi_r$) are set according to the experiment and are explained in Section 5.

The stability proof for linear systems applies here. It can be found if one substitutes (A1)–(A6) into Equation (20). Being stable also requires the decoupling matrix to be invertible, which is the main focus of this research.

As for the attitude–altitude-linearized controller, PD controllers also work with slight changes (lower order). The PD controllers for the attitude–altitude-linearized system are

$$\begin{bmatrix} \ddot{z}_d \\ \ddot{\phi}_d \\ \ddot{\theta}_d \\ \ddot{\psi}_d \end{bmatrix} = \begin{bmatrix} \ddot{z}_r \\ \ddot{\phi}_r \\ \ddot{\theta}_r \\ \ddot{\psi}_r \end{bmatrix} + k_{p1} \cdot \left( \begin{bmatrix} \dot{z}_r \\ \dot{\phi}_r \\ \dot{\theta}_r \\ \dot{\psi}_r \end{bmatrix} - \begin{bmatrix} \dot{z} \\ \dot{\phi} \\ \dot{\theta} \\ \dot{\psi} \end{bmatrix} \right) + k_{p2} \cdot \left( \begin{bmatrix} z_r \\ \phi_r \\ \theta_r \\ \psi_r \end{bmatrix} - \begin{bmatrix} z \\ \phi \\ \theta \\ \psi \end{bmatrix} \right) \quad \text{(A7)}$$

where

$$k_{p1} = \begin{bmatrix} 10 & & & \\ & 10 & & \\ & & 10 & \\ & & & 10 \end{bmatrix}, \quad \text{(A8)}$$

$$k_{p2} = \begin{bmatrix} 10 & & & \\ & 10 & & \\ & & 10 & \\ & & & 10 \end{bmatrix}, \quad \text{(A9)}$$

$z_r$, $\phi_r$, $\theta_r$, and $\psi_r$ are the references. They are set according to the experiment and are specified in Section 5.

## References

1. Mistler, V.; Benallegue, A.; M'Sirdi, N.K. Exact Linearization and Noninteracting Control of a 4 Rotors Helicopter via Dynamic Feedback. In Proceedings of the 10th IEEE International Workshop on Robot and Human Interactive Communication, Paris, France, 18–21 September 2001; ROMAN 2001 (Cat. No.01TH8591); pp. 586–593.
2. Bolandi, H.; Rezaei, M.; Mohsenipour, R.; Nemati, H.; Smailzadeh, S.M. Attitude Control of a Quadrotor with Optimized PID Controller. *ICA* **2013**, *4*, 335–342. https://doi.org/10.4236/ica.2013.43039.

3. Bouabdallah, S.; Noth, A.; Siegwart, R. PID vs LQ Control Techniques Applied to an Indoor Micro Quadrotor. In Proceedings of the 2004 IEEE/RSJ International Conference on Intelligent Robots and Systems (IROS) (IEEE Cat. No.04CH37566), Sendai, Japan, 28 September–2 October 2004; Volume 3, pp. 2451–2456.
4. Martins, L.; Cardeira, C.; Oliveira, P. Feedback Linearization with Zero Dynamics Stabilization for Quadrotor Control. *J. Intell. Robot. Syst.* **2021**, *101*, 7. https://doi.org/10.1007/s10846-020-01265-2.
5. Wang, S.; Polyakov, A.; Zheng, G. Quadrotor Stabilization under Time and Space Constraints Using Implicit PID Controller. *J. Frankl. Inst.* **2022**, *359*, 1505–1530. https://doi.org/10.1016/j.jfranklin.2022.01.002.
6. Bouabdallah, S.; Siegwart, R. Backstepping and Sliding-Mode Techniques Applied to an Indoor Micro Quadrotor. In Proceedings of the 2005 IEEE International Conference on Robotics and Automation, Barcelona, Spain, 18–22 April 2005; pp. 2247–2252.
7. Madani, T.; Benallegue, A. Backstepping Control for a Quadrotor Helicopter. In Proceedings of the 2006 IEEE/RSJ International Conference on Intelligent Robots and Systems, Beijing, China, 9–15 October 2006; pp. 3255–3260.
8. Chen, F.; Lei, W.; Zhang, K.; Tao, G.; Jiang, B. A Novel Nonlinear Resilient Control for a Quadrotor UAV via Backstepping Control and Nonlinear Disturbance Observer. *Nonlinear Dyn.* **2016**, *85*, 1281–1295. https://doi.org/10.1007/s11071-016-2760-y.
9. Liu, P.; Ye, R.; Shi, K.; Yan, B. Full Backstepping Control in Dynamic Systems With Air Disturbances Optimal Estimation of a Quadrotor. *IEEE Access* **2021**, *9*, 34206–34220. https://doi.org/10.1109/ACCESS.2021.3061598.
10. Xu, R.; Ozguner, U. Sliding Mode Control of a Quadrotor Helicopter. In Proceedings of the 45th IEEE Conference on Decision and Control, San Diego, CA, USA, 13–15 December 2006; pp. 4957–4962.
11. Runcharoon, K.; Srichatrapimuk, V. Sliding Mode Control of Quadrotor. In Proceedings of the 2013 The International Conference on Technological Advances in Electrical, Electronics and Computer Engineering (TAEECE), Konya, Turkey, 9–11 May 2013; pp. 552–557.
12. Luque-Vega, L.; Castillo-Toledo, B.; Loukianov, A.G. Robust Block Second Order Sliding Mode Control for a Quadrotor. *J. Frankl. Inst.* **2012**, *349*, 719–739. https://doi.org/10.1016/j.jfranklin.2011.10.017.
13. Xu, L.; Shao, X.; Zhang, W. USDE-Based Continuous Sliding Mode Control for Quadrotor Attitude Regulation: Method and Application. *IEEE Access* **2021**, *9*, 64153–64164. https://doi.org/10.1109/ACCESS.2021.3076076.
14. Ganga, G.; Dharmana, M.M. MPC Controller for Trajectory Tracking Control of Quadcopter. In Proceedings of the 2017 International Conference on Circuit, Power and Computing Technologies (ICCPCT), Kollam, India, 20–21 April 2017; pp. 1–6.
15. Abdolhosseini, M.; Zhang, Y.M.; Rabbath, C.A. An Efficient Model Predictive Control Scheme for an Unmanned Quadrotor Helicopter. *J. Intell. Robot. Syst.* **2013**, *70*, 27–38. https://doi.org/10.1007/s10846-012-9724-3.
16. Alexis, K.; Nikolakopoulos, G.; Tzes, A. Model Predictive Control Scheme for the Autonomous Flight of an Unmanned Quadrotor. In Proceedings of the 2011 IEEE International Symposium on Industrial Electronics, 27–30 June 2011; pp. 2243–2248.
17. Torrente, G.; Kaufmann, E.; Föhn, P.; Scaramuzza, D. Data-Driven MPC for Quadrotors. *IEEE Robot. Autom. Lett.* **2021**, *6*, 3769–3776. https://doi.org/10.1109/LRA.2021.3061307.
18. Ryll, M.; Bulthoff, H.H.; Giordano, P.R. A Novel Overactuated Quadrotor Unmanned Aerial Vehicle: Modeling, Control, and Experimental Validation. *IEEE Trans. Contr. Syst. Technol.* **2015**, *23*, 540–556. https://doi.org/10.1109/TCST.2014.2330999.
19. Ryll, M.; Bulthoff, H.H.; Giordano, P.R. Modeling and Control of a Quadrotor UAV with Tilting Propellers. In Proceedings of the 2012 IEEE International Conference on Robotics and Automation, St. Paul, MN, USA, 14–18 May 2012; pp. 4606–4613.
20. Kumar, R.; Nemati, A.; Kumar, M.; Sharma, R.; Cohen, K.; Cazaurang, F. *Tilting-Rotor Quadcopter for Aggressive Flight Maneuvers Using Differential Flatness Based Flight Controller*; American Society of Mechanical Engineers: Tysons, VA, USA, 2017; p. V003T39A006.
21. Ahmed, A.M.; Katupitiya, J. Modeling and Control of a Novel Vectored-Thrust Quadcopter. *J. Guid. Control. Dyn.* **2021**, *44*, 1399–1409. https://doi.org/10.2514/1.G005467.
22. Xu, J.; D'Antonio, D.S.; Saldaña, D. H-ModQuad: Modular Multi-Rotors with 4, 5, and 6 Controllable DOF. In Proceedings of the 2021 IEEE International Conference on Robotics and Automation (ICRA), Xi'an, China, 30 May–5 June 2021; pp. 190–196.
23. Phong Nguyen, N.; Kim, W.; Moon, J. Observer-Based Super-Twisting Sliding Mode Control with Fuzzy Variable Gains and Its Application to Overactuated Quadrotors. In Proceedings of the 2018 IEEE Conference on Decision and Control (CDC), Miami Beach, FL, USA, 17–19 December 2018; pp. 5993–5998.
24. Convens, B.; Merckaert, K.; Nicotra, M.M.; Naldi, R.; Garone, E. Control of Fully Actuated Unmanned Aerial Vehicles with Actuator Saturation. *IFAC-PapersOnLine* **2017**, *50*, 12715–12720. https://doi.org/10.1016/j.ifacol.2017.08.1823.

25. Cotorruelo, A.; Nicotra, M.M.; Limon, D.; Garone, E. Explicit Reference Governor Toolbox (ERGT). In Proceedings of the 2018 IEEE 4th International Forum on Research and Technology for Society and Industry (RTSI), Palermo, Italy, 10–13 September 2018; pp. 1–6.
26. Dunham, W.; Petersen, C.; Kolmanovsky, I. Constrained Control for Soft Landing on an Asteroid with Gravity Model Uncertainty. In Proceedings of the 2016 American Control Conference (ACC), Boston, MA, USA, 6–8 July 2016; pp. 5842–5847.
27. Hosseinzadeh, M.; Garone, E. An Explicit Reference Governor for the Intersection of Concave Constraints. *IEEE Trans. Automat. Contr.* **2020**, *65*, 1–11. https://doi.org/10.1109/TAC.2019.2906467.
28. Scholz, G.; Trommer, G.F. Model Based Control of a Quadrotor with Tiltable Rotors. *Gyroscopy Navig.* **2016**, *7*, 72–81. https://doi.org/10.1134/S2075108716010120.
29. Taniguchi, T.; Sugeno, M. Trajectory Tracking Controls for Non-Holonomic Systems Using Dynamic Feedback Linearization Based on Piecewise Multi-Linear Models. *IAENG Int. J. Appl. Math.* **2017**, *47*, 339–351.
30. Shen, Z.; Tsuchiya, T. State Drift and Gait Plan in Feedback Linearization Control of A Tilt Vehicle. In Proceedings of the Computer Science & Information Technology (CS & IT); Academy & Industry Research Collaboration Center (AIRCC): Vienna, Austria, 19 March 2022; Volume 12, pp. 1–17.
31. Shen, Z.; Tsuchiya, T. Gait Analysis for a Tiltrotor: The Dynamic Invertible Gait. *Robotics* **2022**, *11*, 33. https://doi.org/10.3390/robotics11020033.
32. Das, A.; Subbarao, K.; Lewis, F. Dynamic Inversion of Quadrotor with Zero-Dynamics Stabilization. In Proceedings of the 2008 IEEE International Conference on Control Applications, San Antonio, TX, USA, 3–5 Sptember 2008; pp. 1189–1194.
33. Ghandour, J.; Aberkane, S.; Ponsart, J.-C. Feedback Linearization Approach for Standard and Fault Tolerant Control: Application to a Quadrotor UAV Testbed. *J. Phys. Conf. Ser.* **2014**, *570*, 082003. https://doi.org/10.1088/1742-6596/570/8/082003.
34. Cowling, I.; Yakimenko, O.; Whidborne, J.; Cooke, A. Direct Method Based Control System for an Autonomous Quadrotor. *J. Intell. Robot. Syst.* **2010**, *60*, 285–316. https://doi.org/10.1007/s10846-010-9416-9.
35. Voos, H. Nonlinear Control of a Quadrotor Micro-UAV Using Feedback-Linearization. In Proceedings of the 2009 IEEE International Conference on Mechatronics, Malaga, Spain, 14–17 April 2009; pp. 1–6.
36. Wie, B. New Singularity Escape/Avoidance Steering Logic for Control Moment Gyro Systems. In Proceedings of the AIAA Guidance, Navigation, and Control Conference and Exhibit; American Institute of Aeronautics and Astronautics: Austin, TX, USA, 11 August 2003.
37. Sands, T.; Kim, J.; Agrawal, B. Singularity Penetration with Unit Delay (SPUD). *Mathematics* **2018**, *6*, 23. https://doi.org/10.3390/math6020023.
38. Luukkonen, T. Modelling and Control of Quadcopter. *Indep. Res. Proj. Appl. Math. Espoo* **2011**, *22*, 22.

# Chapter 3

# Over-intensive Control Signals and Singular Zone in Tiltrotor Feedback Linearization (Gait Analysis for a Tiltrotor: The Dynamic Invertible Gait)

**This Chapter has been published on Gait Analysis for a Tiltrotor: The Dynamic Invertible Gait. *Robotics* 2022, *11*, 33, doi:10.3390/robotics11020033 as the first author. The rest co-author(s) contribute nothing other than supervising and/or minor edition.**

**Abstract:** A conventional feedback-linearization-based controller, when applied to a tiltrotor (eight inputs), results in extensive changes in tilting angles, which are not expected in practice. To solve this problem, we introduce the novel concept of "UAV gait" to restrict the tilting angles. The gait plan was initially used to solve the control problems in quadruped (four-legged) robots. Applying this approach, accompanied by feedback linearization, to a tiltrotor may give rise to the well-known non-invertible problem in the decoupling matrix. In this study, we explored invertible gait in a tiltrotor, and applied feedback linearization to stabilize the attitude and the altitude. The conditions necessary to achieve a full-rank decoupling matrix were deduced and simplified to near-zero roll and zero pitch. This paper proposes several invertible gaits to conduct an attitude–altitude control test. The accepted gaits within the region of interest were visualized. The simulation was conducted in Simulink, MATLAB. The results show promising responses in stabilizing attitude and altitude.

## 1. Introduction

In the past decade, tiltrotors have attracted great interest. Tiltrotors are a novel type of quadrotor [1–8], wherein the axes of the propellers tilt, imparting the ability to change the direction of each thrust. Typical control methods to stabilize a tiltrotor include LQR and PID [9–11], backstepping and sliding mode [12–16], feedback linearization [17–23], optimal control [24–26], adaptive control [27,28], etc. Feedback linearization [29–31] explicitly decouples the nonlinear parts, and enables one to utilize the over-actuated properties. This approach is not only effective for a standard tiltrotor, but also for a tiltrotor with predetermined tilting angles [32].

Given the benefits of feedback linearization, we have achieved capabilities such as tracking with a rapid response for sophisticated references [17,18,22]. However, several potential risks may hinder the applications of this technique. One of these is saturation restriction. This limit refers to the upper bound of a given motor and the non-negative bound of the thrust; negative thrust is usually not acceptable in applications. Some research has focused on avoiding either of the above bounds [33–36]. Although hitting a boundary does not necessarily result in instability, the corresponding stability criteria for these cases can be hard to trace or generalize [37–39].

Another typical issue of feedback linearization when applied in over-actuated systems is the so-called "State Drift" phenomenon, as defined in [40]; the state or input may drift as the simulation continues. In [1,22], the authors defined the optimal control conditions to avert this problem. The extra requirements restrict the freedom of the inputs to some extent. Unfortunately, relevant stability proof has not been provided. An alternative method to avoid state drift is to reduce the number of inputs by predefining some of them [40]. This method is inspired by gait plan, which is widely adopted in quadruped (four-legged) robots [41–45].

The number of inputs (the magnitude of each thrust and the direction of each thrust) of the tiltrotor in this research was eight at most. The number of degrees of freedom was six (attitude and position). The current feedback-linearization-based control strategies used to stabilize this tiltrotor are categorized into two groups.

One seeks to manipulate all eight inputs so that the tiltrotor becomes an over-actuated system with the potential to simultaneously maneuver in all degrees of freedom. The other sacrifices the number of inputs to make it equal to the number of degrees of freedom; a typical means of achieving this is to make the tilting angles of the thrusts at the opposite arms equal.

However, both strategies lead to unrealistic changes in tilting angles. The tilting angles undergo large changes during flight [1,22]; some of these are unexpectedly fast. These requirements are hardly practical; a reasonable tilting angle should be within a small range and not extend beyond a certain period ($2\pi$).

In this research, we prohibited changes in tilting angles during our flight and analyzed each combination of four tilting angles to find those wherein feedback linearization was applicable (i.e., the decoupling matrix was invertible). This means that the number of inputs was four in this research. Furthermore, the accepted region for designing the gait was explored in an attitude–altitude stabilization test.

We compared the control result by completing the same task but with the other controller, which fully utilized all the inputs (over-actuated) of the same tiltrotor with identical parameters.

The rest of this paper is structured as follows. Section 2 introduces the dynamics of the tiltrotor. Then, the developed feedback-linearization-based controller is presented in Section 3. The necessary conditions for an applicable gait are analyzed in Section 4. In the simulation presented in Section 5, the attitude and the altitude of the tiltrotor were stabilized by the controller proposed in Section 3. Section 6 shows the results and the sufficient conditions for an applicable gait in the region of interest. The conclusions and discussions are addressed in Section 7.

## 2. Dynamics of the Tiltrotor

A tiltrotor is a type of modified quadrotor whose directions of thrust are able to change [1–7]. Figure 1 illustrates a tiltrotor. Here, the direction of each thrust is shown in the corresponding yellow plane. For details of the kinematics, [1,2,4,5,7] are recommended.

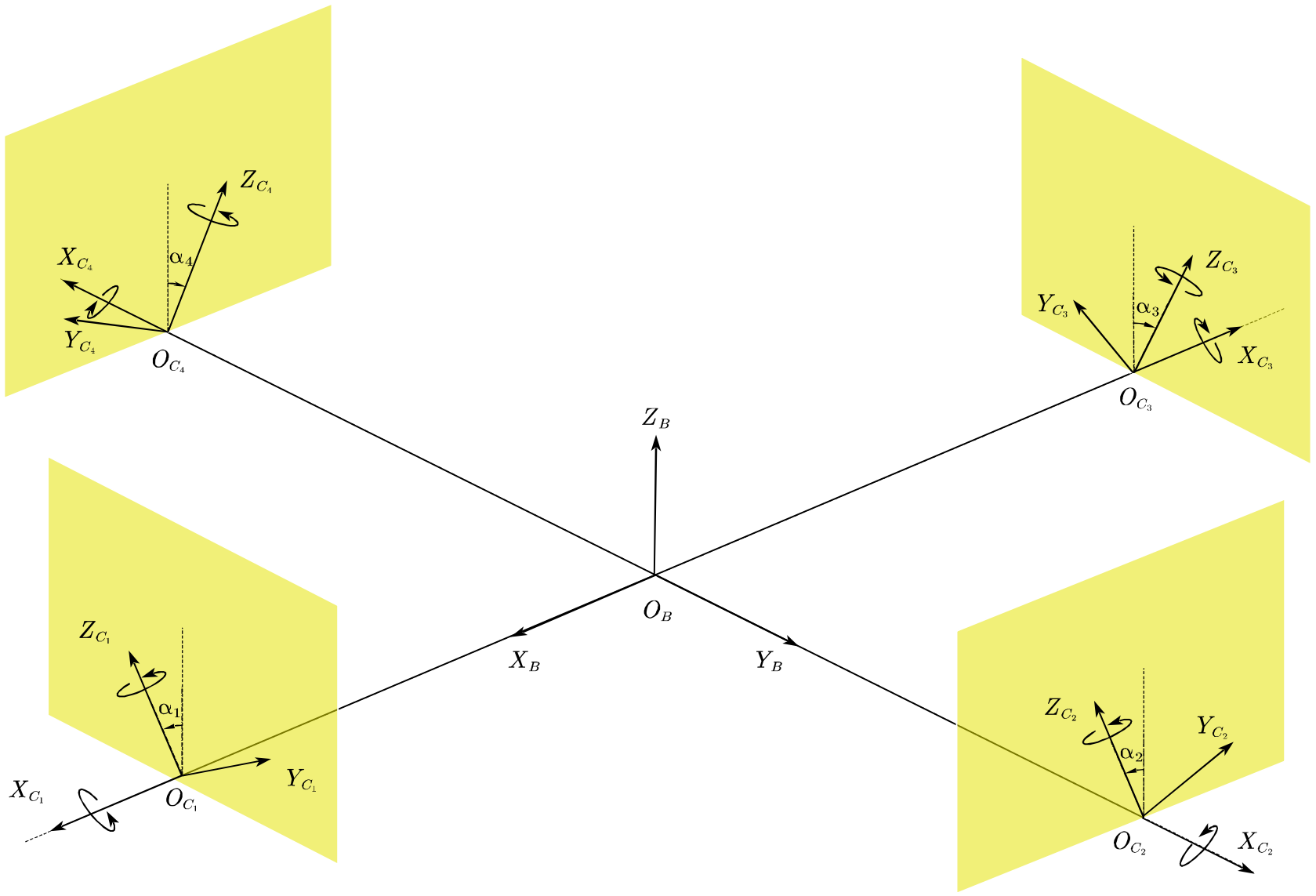

**Figure 1.** Sketch of the tiltrotor.

Among the quadrotor dynamics discussed in previous research, the widely accepted model from [1,2,7,22] is constructed by separately analyzing each part (body frame, tilting motor frames, and propeller frames) using Newton's Law. The controller was developed based on simplified dynamics via Equations (1)–(5).

The position $P = [X \quad Y \quad Z]^T$ is given by:

$$\ddot{P} = \begin{bmatrix} 0 \\ 0 \\ -g \end{bmatrix} + \frac{1}{m} \cdot {}^{W}R \cdot F(\alpha) \cdot \begin{bmatrix} \varpi_1 \cdot |\varpi_1| \\ \varpi_2 \cdot |\varpi_2| \\ \varpi_3 \cdot |\varpi_3| \\ \varpi_4 \cdot |\varpi_4| \end{bmatrix} \triangleq \begin{bmatrix} 0 \\ 0 \\ -g \end{bmatrix} + \frac{1}{m} \cdot {}^{W}R \cdot F(\alpha) \cdot w \quad (1)$$

where $m$ is the total mass, $g$ is the gravitational acceleration, and $\varpi_i$, $(i = 1,2,3,4)$ is the angular velocity of the propeller ( $\varpi_{1,3} < 0$ , $\varpi_{2,4} > 0$ ) with respect to the propeller-fixed frame, $w = [w_1 \quad w_2 \quad w_3 \quad w_4]^T$. $w_i = \varpi_i \cdot |\varpi_i|$, $m$ is the mass of the tilt-rotor. Rotor 1 and Rotor 3 were assumed to rotate clockwise. Rotor 2 and Rotor 4 were assumed to rotate counter-clockwise. ${}^{W}R$ is the rotational matrix [46] between the inertial frame and the body-fixed frame (Equation (2)):

$${}^{W}R = \begin{bmatrix} c\theta \cdot c\psi & s\phi \cdot s\theta \cdot c\psi - c\phi \cdot s\psi & c\phi \cdot s\theta \cdot c\psi + s\phi \cdot s\psi \\ c\theta \cdot s\psi & s\phi \cdot s\theta \cdot s\psi + c\phi \cdot c\psi & c\phi \cdot s\theta \cdot s\psi - s\phi \cdot c\psi \\ -s\theta & s\phi \cdot c\theta & c\phi \cdot c\theta \end{bmatrix}, \quad (2)$$

where $s\Lambda = \sin(\Lambda)$ and $c\Lambda = \cos(\Lambda)$. $\phi$, $\theta$, and $\psi$ are the roll angle, pitch angle, and yaw angle, respectively. The tilting angles $\alpha = [\alpha_1 \quad \alpha_2 \quad \alpha_3 \quad \alpha_4]$. The positive directions of the tilting angles are defined in Figure 1. $F(\alpha)$ is defined as:

$$F(\alpha) = \begin{bmatrix} 0 & K_f \cdot s2 & 0 & -K_f \cdot s4 \\ K_f \cdot s1 & 0 & -K_f \cdot s3 & 0 \\ -K_f \cdot c1 & K_f \cdot c2 & -K_f \cdot c3 & K_f \cdot c4 \end{bmatrix} \quad (3)$$

where $si = \sin(\alpha_i)$, $ci = \cos(\alpha_i)$, and $(i = 1,2,3,4)$. $K_f$ $(8.048 \times 10^{-6} N \cdot s^2/rad^2)$ is the coefficient of the thrust.

The angular velocity of the body with respect to its own frame, $\omega_B = [p \quad q \quad r]^T$, is governed by Newton–Euler Formula as:

$$\dot{\omega}_B = {I_B}^{-1} \cdot \tau(\alpha) \cdot w \quad (4)$$

where $I_B$ is the matrix of moments of inertia, $K_m$ $(2.423 \times 10^{-7} N \cdot m \cdot s^2/rad^2)$ is the coefficient of the drag, and $L$ is the length of the arm:

$$\tau(\alpha) = \begin{bmatrix} 0 & L \cdot K_f \cdot c2 - K_m \cdot s2 & 0 & -L \cdot K_f \cdot c4 + K_m \cdot s4 \\ L \cdot K_f \cdot c1 + K_m \cdot s1 & 0 & -L \cdot K_f \cdot c3 - K_m \cdot s3 & 0 \\ L \cdot K_f \cdot s1 - K_m \cdot c1 & -L \cdot K_f \cdot s2 - K_m \cdot c2 & L \cdot K_f \cdot s3 - K_m \cdot c3 & -L \cdot K_f \cdot s4 - K_m \cdot c4 \end{bmatrix}. \quad (5)$$

Thus far, we have determined the dynamics of a tiltrotor. There are several remarks we can make.

Firstly, the tilting angles $(\alpha_1, \alpha_2, \alpha_3, \alpha_4)$ are predetermined before controlling. This indicates that they are constant during flight, i.e.:

$$\dot{\alpha}_i \equiv 0, i = 1, 2, 3, 4. \quad (6)$$

In this research, we will determine the proper tilting angles required to support our flight.

Secondly, the relationship [47–49] between the angular velocity of the body, $\omega_B$, and the attitude rotation matrix $({}^{W}R)$ is given by:

$${}^{W}\dot{R} = {}^{W}R \cdot \widehat{\omega}_B \quad (7)$$

where " $\widehat{\ }$ " is the hat operation used to produce the skew matrix, and ${}^{W}\dot{R}$ represents the derivative of rotation matrix.

Our simulator was built based on Equations (1)–(7).

In the controller design process, we approximated the relationship between the angular velocity of the body, $\omega_B$, and the attitude angle, $(\phi, \theta, \psi)$, by:

$$\begin{bmatrix}\dot{\phi}\\ \dot{\theta}\\ \dot{\psi}\end{bmatrix} = \omega_B. \tag{8}$$

Instead of further exploiting Equation (7), the controller was designed based on Equations (1)–(6) and (8).

The parameters of this tiltrotor are as follows: $m = 0.429$ kg, $L = 0.1785$ m, $g = 9.8$ N/kg, and $I_B = \text{diag}([2.24 \times 10^{-3}, 2.99 \times 10^{-3}, 4.80 \times 10^{-3}])\ \text{kg} \cdot \text{m}^2$.

## 3. Feedback Linearization and Control

The control scenario consisted of two sections. Firstly, the nonlinear dynamics were dynamically inverted by feedback linearization. Secondly, the linearized system was stabilized based on a third-order PD controller. The rest of this section introduces these strategies.

### *3.1. Feedback Linearization*

The first step in feedback linearization is to select the output. In this research, we chose attitude–altitude, described in Equation (9), as our output, because the choice of position–yaw may introduce a further singular zone [50,51]:

$$\begin{bmatrix}y_1\\ y_2\\ y_3\\ y_4\end{bmatrix} = \begin{bmatrix}\phi\\ \theta\\ \psi\\ Z\end{bmatrix}. \tag{9}$$

Calculating the second derivative of Equation (9) yields:

$$\begin{bmatrix}\ddot{y}_1\\ \ddot{y}_2\\ \ddot{y}_3\\ \ddot{y}_4\end{bmatrix} = \begin{bmatrix}0\\ 0\\ 0\\ -g\end{bmatrix} + \begin{bmatrix} {I_B}^{-1} \cdot \tau(\alpha) \\ [0 \quad 0 \quad 1] \cdot \frac{K_f}{m} \cdot {}^{W}R \cdot F(\alpha)\end{bmatrix}^{4\times 4} \cdot w. \tag{10}$$

Notably, we derive Equation (11) if $\varpi_{1,3} < 0$, $\varpi_{2,4} > 0$:

$$(\varpi_i \cdot |\varpi_i|)' = 2 \cdot \dot{\varpi}_i \cdot |\varpi_i|. \tag{11}$$

Differentiating Equation (10) yields:

$$\begin{aligned}\begin{bmatrix}\dddot{y}_1\\ \dddot{y}_2\\ \dddot{y}_3\\ \dddot{y}_4\end{bmatrix} &= \begin{bmatrix} {I_B}^{-1} \cdot \tau(\alpha) \\ [0 \quad 0 \quad 1] \cdot \dfrac{K_f}{m} \cdot {}^{W}R \cdot F(\alpha) \cdot 2 \cdot \begin{bmatrix}|\varpi_1| & & & \\ & |\varpi_2| & & \\ & & |\varpi_3| & \\ & & & |\varpi_4|\end{bmatrix}\end{bmatrix}^{4\times 4} \cdot \begin{bmatrix}\dot{\varpi}_1\\ \dot{\varpi}_2\\ \dot{\varpi}_3\\ \dot{\varpi}_4\end{bmatrix} \\ &\quad + [0 \quad 0 \quad 1] \cdot \frac{K_f}{m} \cdot {}^{W}R \cdot \widehat{\omega}_B \cdot F(\alpha) \cdot w \cdot \begin{bmatrix}0\\ 0\\ 0\\ 1\end{bmatrix} \\ &\triangleq \bar{\Delta} \cdot \begin{bmatrix}\dot{\varpi}_1\\ \dot{\varpi}_2\\ \dot{\varpi}_3\\ \dot{\varpi}_4\end{bmatrix} + Ma\end{aligned} \tag{12}$$

where $\bar{\Delta}$ is called the decoupling matrix [32], and $[\dot{\varpi}_1 \quad \dot{\varpi}_2 \quad \dot{\varpi}_3 \quad \dot{\varpi}_4]^T \triangleq U$ is the new input vector.

From Equation (12), we may derive the decoupled relationship in Equation (13), which is compatible with the controller design process:

$$\begin{bmatrix}\dot{\varpi}_1\\ \dot{\varpi}_2\\ \dot{\varpi}_3\\ \dot{\varpi}_4\end{bmatrix} = \bar{\Delta}^{-1} \cdot \left(\begin{bmatrix}\dddot{y}_{1d}\\ \dddot{y}_{2d}\\ \dddot{y}_{3d}\\ \dddot{y}_{4d}\end{bmatrix} - Ma\right). \tag{13}$$

Obviously, the necessary condition for receiving Equation (13) is that the decoupling matrix ($\bar{\Delta}$) is invertible. Section 4 deepens this discussion.

Once solving Equation (13), the controller may be applied to this linearized system. In this research, we deployed third-order PD controllers.

*3.2. Third-Order PD Controllers*

We designed the third-order PD controllers as:

$$\begin{bmatrix} \dddot{y}_{1d} \\ \dddot{y}_{2d} \\ \dddot{y}_{3d} \end{bmatrix} = \begin{bmatrix} \dddot{y}_{1r} \\ \dddot{y}_{2r} \\ \dddot{y}_{3r} \end{bmatrix} + {K_{P1}}^{3\times3} \cdot \left( \begin{bmatrix} \ddot{y}_{1r} \\ \ddot{y}_{2r} \\ \ddot{y}_{3r} \end{bmatrix} - \begin{bmatrix} \ddot{y}_{1} \\ \ddot{y}_{2} \\ \ddot{y}_{3} \end{bmatrix} \right) + {K_{P2}}^{3\times3} \cdot \left( \begin{bmatrix} \dot{y}_{1r} \\ \dot{y}_{2r} \\ \dot{y}_{3r} \end{bmatrix} - \begin{bmatrix} \dot{y}_{1} \\ \dot{y}_{2} \\ \dot{y}_{3} \end{bmatrix} \right) + {K_{P3}}^{3\times3} \cdot \left( \begin{bmatrix} y_{1r} \\ y_{2r} \\ y_{3r} \end{bmatrix} - \begin{bmatrix} y_{1} \\ y_{2} \\ y_{3} \end{bmatrix} \right), \quad (14)$$

$$\dddot{y}_{4d} = \dddot{y}_{4r} + K_{PZ_1} \cdot (\ddot{y}_{4r} - \ddot{y}_4) + K_{PZ_2} \cdot (\dot{y}_{4r} - \dot{y}_4) + K_{PZ_3} \cdot (y_{4r} - y_4), \quad (15)$$

where $K_{Pi}$ ($i = 1,2,3$) is the three-by-three diagonal control coefficient matrix, $K_{PZ_i}$ ($i = 1,2,3$) is the control coefficient (scalar), $y_j$ ($j = 1,2,3,4$) is the state, and $y_{jr}$ ($j = 1,2,3,4$) is the reference.

The control parameters in this section are specified as follows: $K_{P1} = K_{P2} = K_{P3} = \text{diag}([1,1,1])$, $K_{PZ_1} = 10$, $K_{PZ_2} = 5$, and $K_{PZ_3} = 10$.

## 4. Applicable Gait (Necessary Conditions)

As discussed, the necessary condition for applying this control is that the decoupling matrix ($\bar{\Delta}$) is invertible. In this section, we find the equivalent conditions to derive an invertible decoupling matrix.

Notice the relationship:

$$\bar{\Delta} \sim \begin{bmatrix} \tau(\alpha) \\ [0 \quad 0 \quad 1] \cdot {}^{W}R \cdot F(\alpha) \end{bmatrix} \quad (16)$$

where "$A \sim B$" indicates that Matrix $A$ is equivalent to Matrix $B$. Two matrices are called equivalent if, and only if, there exist invertible matrices $P$ and $Q$, so that $A = P \cdot B \cdot Q$.

The following propositions and proofs are specifically applicable only to the tiltrotor with our chosen control parameters and coefficients.

**Proposition 1.** *The decoupling matrix is invertible if, and only if:*

$$\begin{aligned}
&1.000 \cdot c1 \cdot c2 \cdot c3 \cdot s4 \cdot s\theta - 1.000 \cdot c1 \cdot s2 \cdot c3 \cdot c4 \cdot s\theta \\
&- 2.880 \cdot c1 \cdot c2 \cdot s3 \cdot s4 \cdot s\theta + 2.880 \cdot c1 \cdot s2 \cdot s3 \cdot c4 \cdot s\theta \\
&- 2.880 \cdot s1 \cdot c2 \cdot c3 \cdot s4 \cdot s\theta + 2.880 \cdot s1 \cdot s2 \cdot c3 \cdot c4 \cdot s\theta \\
&- 1.000 \cdot s1 \cdot c2 \cdot s3 \cdot s4 \cdot s\theta + 1.000 \cdot s1 \cdot s2 \cdot s3 \cdot c4 \cdot s\theta \\
&+ 4.000 \cdot c1 \cdot c2 \cdot c3 \cdot c4 \cdot c\phi \cdot c\theta + 5.592 \cdot c1 \cdot c2 \cdot c3 \cdot s4 \\
&\cdot c\phi \cdot c\theta - 5.592 \cdot c1 \cdot c2 \cdot s3 \cdot c4 \cdot c\phi \cdot c\theta + 5.592 \cdot c1 \cdot s2 \\
&\cdot c3 \cdot c4 \cdot c\phi \cdot c\theta - 5.592 \cdot s1 \cdot c2 \cdot c3 \cdot c4 \cdot c\phi \cdot c\theta + 1.000 \\
&\cdot c1 \cdot c2 \cdot s3 \cdot c4 \cdot s\phi \cdot c\theta + 0.9716 \cdot c1 \cdot c2 \cdot s3 \cdot s4 \cdot c\phi \cdot c\theta \\
&- 2.000 \cdot c1 \cdot s2 \cdot c3 \cdot s4 \cdot c\phi \cdot c\theta + 0.9716 \cdot c1 \cdot s2 \cdot s3 \cdot c4 \\
&\cdot c\phi \cdot c\theta - 1.000 \cdot s1 \cdot c2 \cdot c3 \cdot c4 \cdot s\phi \cdot c\theta + 0.9716 \cdot s1 \cdot c2 \\
&\cdot c3 \cdot s4 \cdot c\phi \cdot c\theta - 2.000 \cdot s1 \cdot c2 \cdot s3 \cdot c4 \cdot c\phi \cdot c\theta + 0.9716 \\
&\cdot s1 \cdot s2 \cdot c3 \cdot c4 \cdot c\phi \cdot c\theta + 2.880 \cdot c1 \cdot c2 \cdot s3 \cdot s4 \cdot s\phi \cdot c\theta \\
&+ 2.880 \cdot c1 \cdot s2 \cdot s3 \cdot c4 \cdot s\phi \cdot c\theta - 0.1687 \cdot c1 \cdot s2 \cdot s3 \cdot s4 \\
&\cdot c\phi \cdot c\theta - 2.880 \cdot s1 \cdot c2 \cdot c3 \cdot s4 \cdot s\phi \cdot c\theta + 0.1687 \cdot s1 \cdot c2 \\
&\cdot s3 \cdot s4 \cdot c\phi \cdot c\theta - 2.880 \cdot s1 \cdot s2 \cdot c3 \cdot c4 \cdot s\phi \cdot c\theta - 0.1687 \\
&\cdot s1 \cdot s2 \cdot c3 \cdot s4 \cdot c\phi \cdot c\theta + 0.1687 \cdot s1 \cdot s2 \cdot s3 \cdot c4 \cdot c\phi \cdot c\theta \\
&- 1.000 \cdot c1 \cdot s2 \cdot s3 \cdot s4 \cdot s\phi \cdot c\theta + 1.000 \cdot s1 \cdot s2 \cdot c3 \cdot s4 \\
&\cdot s\phi \cdot c\theta \\
&\neq 0.
\end{aligned} \quad (17)$$

**Proof of Proposition 1.** *Expanding the second matrix in Equation (16) yields:*

$$\bar{\Delta}\sim\begin{bmatrix} 0 & L\cdot K_f\cdot c2-K_m\cdot s2 & 0 & -L\cdot K_f\cdot c4+K_m\cdot s4 \\ L\cdot K_f\cdot c1+K_m\cdot s1 & 0 & -L\cdot K_f\cdot c3-K_m\cdot s3 & 0 \\ L\cdot K_f\cdot s1-K_m\cdot c1 & -L\cdot K_f\cdot s2-K_m\cdot c2 & L\cdot K_f\cdot s3-K_m\cdot c3 & -L\cdot K_f\cdot s4-K_m\cdot c4 \\ c\theta\cdot s\phi\cdot s1-c\theta\cdot c\phi\cdot c1 & -s\theta\cdot s2+c\theta\cdot c\phi\cdot c2 & -c\theta\cdot s\phi\cdot s3-c\theta\cdot c\phi\cdot c3 & s\theta\cdot s4+c\theta\cdot c\phi\cdot c4 \end{bmatrix}. \quad (18)$$

*Calculating the determinant of the second matrix in Equation (18) yields Condition (17).*

**Proposition 2.** *When the roll angle and pitch angle of the tiltrotor are close to zero, the decoupling matrix is invertible if, and only if:*

$$\begin{aligned} &4.000\cdot c1\cdot c2\cdot c3\cdot c4+5.592 \\ &\cdot(+c1{\cdot}c2{\cdot}c3{\cdot}s4-c1{\cdot}c2{\cdot}s3{\cdot}c4+c1{\cdot}s2{\cdot}c3{\cdot}c4-s1{\cdot}c2{\cdot}c3{\cdot}c4) \\ &+0.9716 \\ &\cdot(+c1{\cdot}c2{\cdot}s3{\cdot}s4+c1{\cdot}s2{\cdot}s3{\cdot}c4+s1{\cdot}c2{\cdot}c3{\cdot}s4+s1{\cdot}s2{\cdot}c3{\cdot}c4) \\ &+2.000\cdot(-c1{\cdot}s2{\cdot}c3{\cdot}s4-s1{\cdot}c2{\cdot}s3{\cdot}c4)+0.1687 \\ &\cdot(-c1{\cdot}s2{\cdot}s3{\cdot}s4+s1{\cdot}c2{\cdot}s3{\cdot}s4-s1{\cdot}s2{\cdot}c3{\cdot}s4+s1{\cdot}s2{\cdot}s3{\cdot}c4) \\ \neq\ &0. \end{aligned} \quad (19)$$

**Proof of Proposition 2.** *Make the assumptions in Equations (20) and (21):*

$$\theta=0. \quad (20)$$

$$\phi=0. \quad (21)$$

*Substituting Equations (20) and (21) into Condition (17) yields Condition (19).* □

**Remark 1.** *One may believe that Equation (18) would be an applicable gait-selecting zone, where any combination of* $(\alpha_1,\alpha_2,\alpha_3,\alpha_4)$ *satisfying Proposition 1 would lead to a stable result. Unfortunately, this is not always true, because the proven Proposition only guarantees the invertibility of the decoupling matrix. Hitting the non-negative angular velocity bound for* $\varpi_i$ *can result in instability. Obviously, neither Proposition 1 nor Proposition 2 rules out this situation. This is why we call the Propositions the "necessary" conditions for an applicable gait. The sufficient condition within the range of interest is presented in the simulation in Sections 5 and 6.*

**Remark 2.** *Notably, Condition (17) not only restricts the gait* $(\alpha_1,\alpha_2,\alpha_3,\alpha_4)$*, but also rules out some attitudes* $(\phi,\theta)$*; a specific gait can be driven to violate Condition (17) while steering to some specific attitudes. In actuality, a similar attitude-based condition means the failure makes the decoupling matrix invertible in [50], hindering further applications in feedback linearization (position–yaw output) before modification. This is because roll and pitch are not directly controlled via the position–yaw output choice. However, the adverse effect of this property is weakened in our research. The output choice (attitude–altitude) enables the tiltrotor to directly steer the attitude. Thus, roll angle and pitch angle can be relatively arbitrarily assigned. Steering the attitude away from the region violating Condition (17) is consequently made possible by this controller.*

## 5. Attitude–Altitude Stabilization Test

The conditions in Section 4 are necessary in achieving the applicable gait. The sufficient conditions are explored empirically in this section.

To reduce the complexity, the preliminary step was to restrict the gait to some extent and find the gait region of interest. The next step was to determine the exploration direction of the candidate gaits. The final step was to conduct the simulation for each gait along the exploring direction.

### *5.1. Restricted Gait Region*

The simulation was conducted at near-zero roll angle and pitch angle ($\phi = 0, \theta = 0$). Thus, the following analysis is based on Condition (19) rather than on Condition (17).

Although Condition (19) includes no information about the roll angle and pitch angle, there are still four parameters in the gait ($\alpha_1, \alpha_2, \alpha_3, \alpha_4$) to be determined, which is a complicated issue. Here, we simplify the gait first.

Instead of exploring the entire space of ($\alpha_1, \alpha_2, \alpha_3, \alpha_4$), we explore four gaits with one of the following restrictions in each gait: $\alpha_1 = \alpha_3$, $\alpha_1 = \frac{1}{2} \cdot \alpha_3$, $\alpha_1 = -\alpha_3$, and $\alpha_1 = -\frac{1}{2} \cdot \alpha_3$.

5.1.1. Case 1 (Equal)

This section demonstrates the result for some cases satisfying $\alpha_1 = \alpha_3$. Specifically, $\alpha_1 = \alpha_3 = -0.15, -0.075, 0, 0.075, 0.15$. Figure 2 plots the left side of Condition (19). The result is five surfaces about ($\alpha_2$,$\alpha_4$). Intercepting the surfaces in Figure 2 by the zero plane (Determinant $= 0$) yields Figure 3. Figure 3 plots ($\alpha_2$,$\alpha_4$), violating Condition (19).

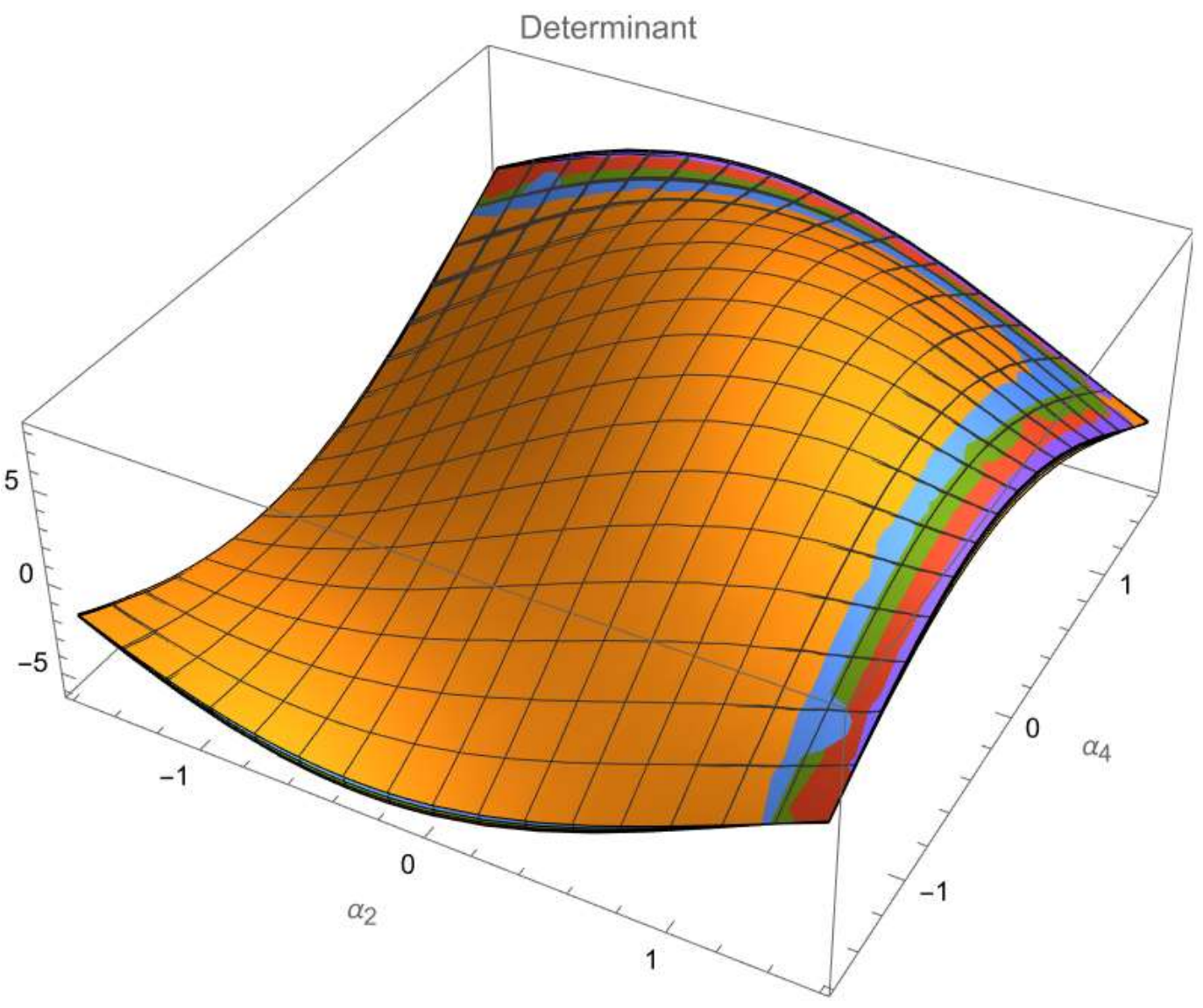

**Figure 2.** Surfaces of the determinant ($\alpha_1 = \alpha_3$).

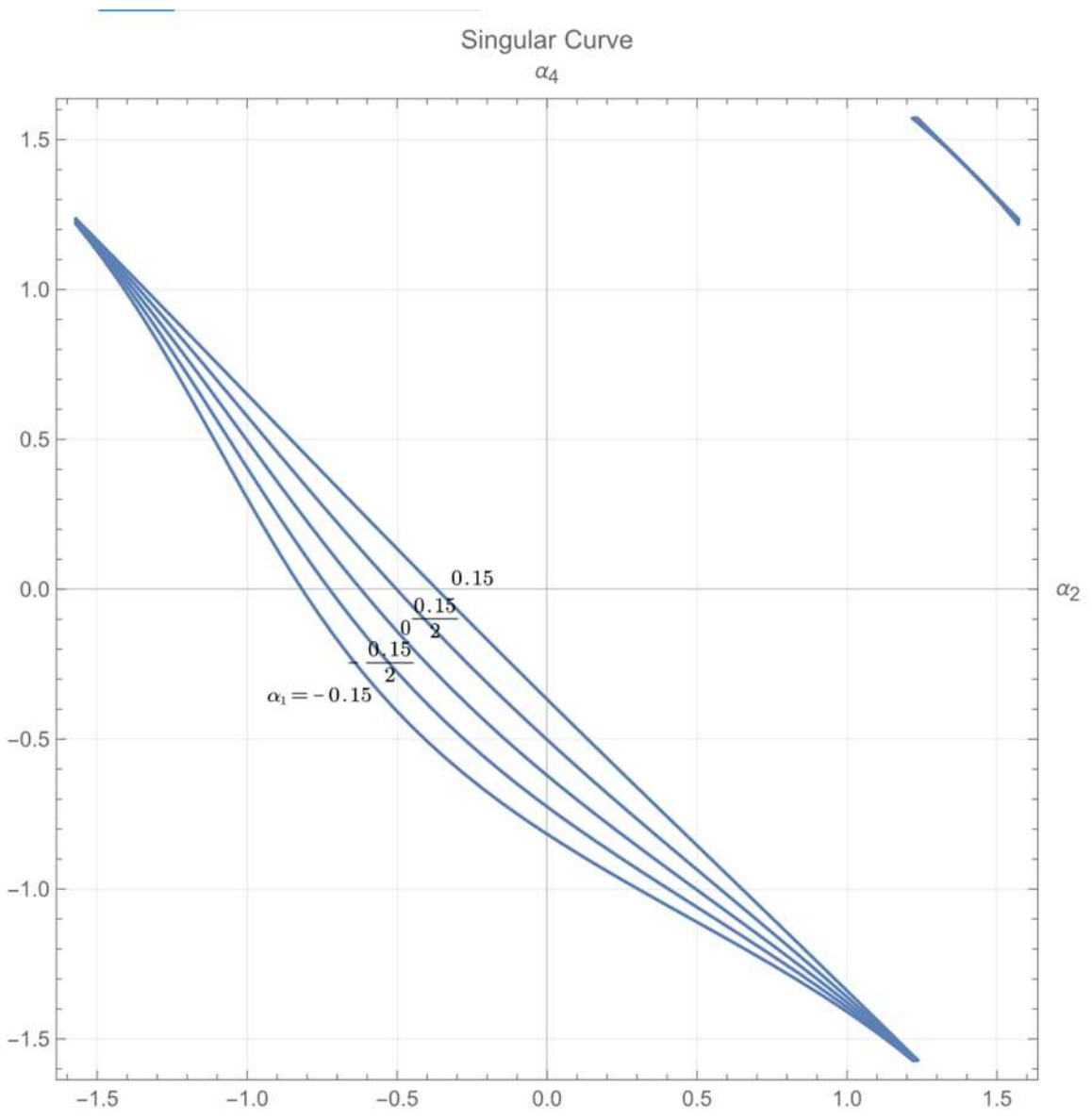

**Figure 3.** ($\alpha_2$,$\alpha_4$) violating Condition (19) ($\alpha_1 = \alpha_3$).

5.1.2. Case 2 (Half)

This section presents the results for some cases satisfying $\alpha_1 = \frac{1}{2} \cdot \alpha_3$. Specifically, $\alpha_1 = \frac{1}{2} \cdot \alpha_3 = -0.2, -0.1, 0, 0.1, 0.2$. Figure 4 plots the left side of Condition (19). The result is five surfaces about $(\alpha_2, \alpha_4)$. Intercepting the surfaces in Figure 4 by the zero plane (Determinant = 0) yields Figure 5. Figure 5 plots the $(\alpha_2, \alpha_4)$ violating Condition (19).

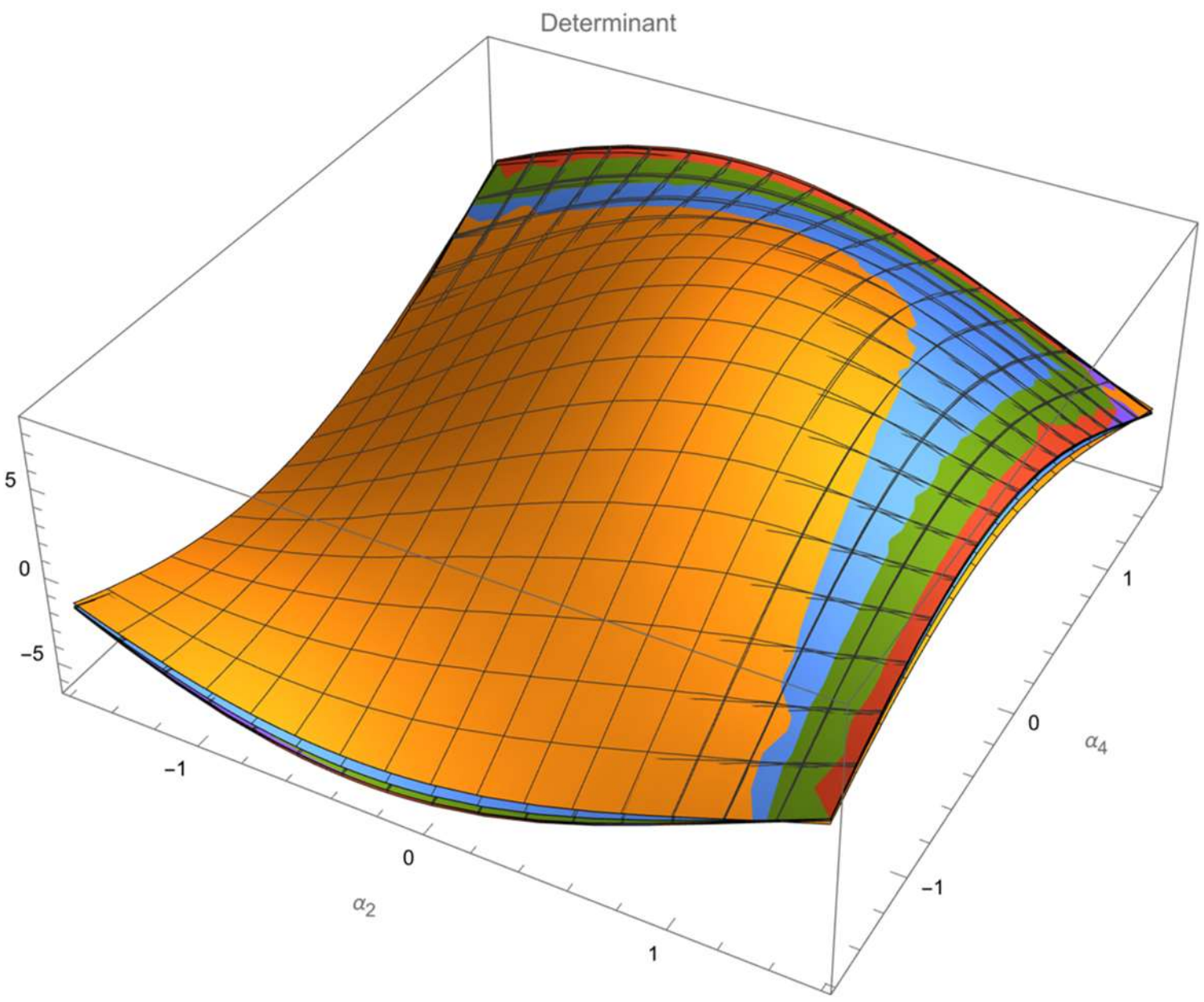

**Figure 4.** Surfaces of the determinant ($\alpha_1 = \frac{1}{2} \cdot \alpha_3$).

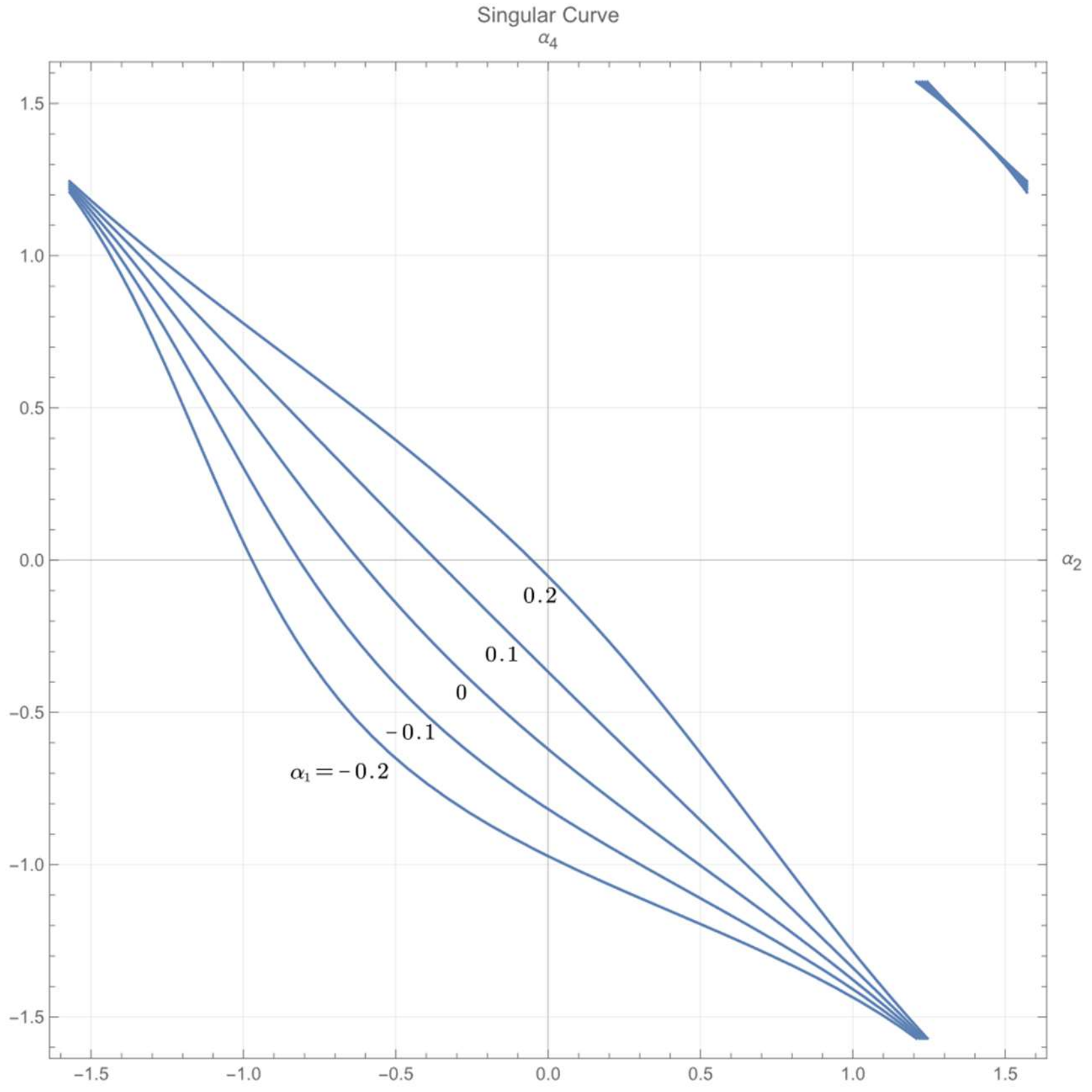

**Figure 5.** $(\alpha_2, \alpha_4)$ violating Condition (19) ($\alpha_1 = \frac{1}{2} \cdot \alpha_3$).

#### 5.1.3. Case 3 (Negative)

This section presents the results for some cases satisfying $\alpha_1 = -\alpha_3$. Specifically, $\alpha_1 = -\alpha_3 = -1.4, -0.7, 0, 0.7, 1.4$. Figure 6 plots the left side of Condition (19). The result is five surfaces about ($\alpha_2$,$\alpha_4$). Intercepting the surfaces in Figure 6 by the zero plane (Determinant = 0) yields Figure 7. Figure 7 plots ($\alpha_2$,$\alpha_4$), violating Condition (19).

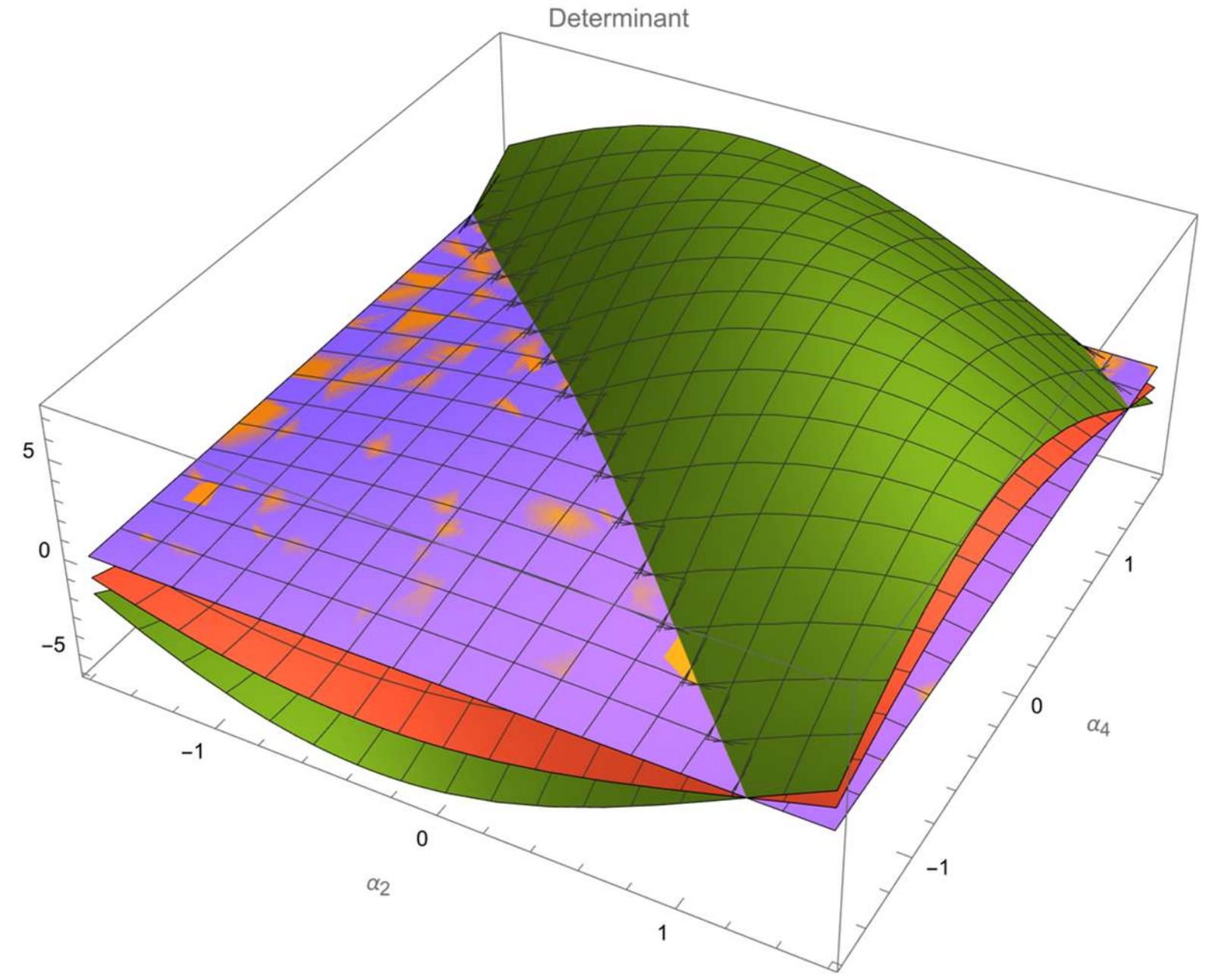

**Figure 6.** Surfaces of the determinant ($\alpha_1 = -\alpha_3$).

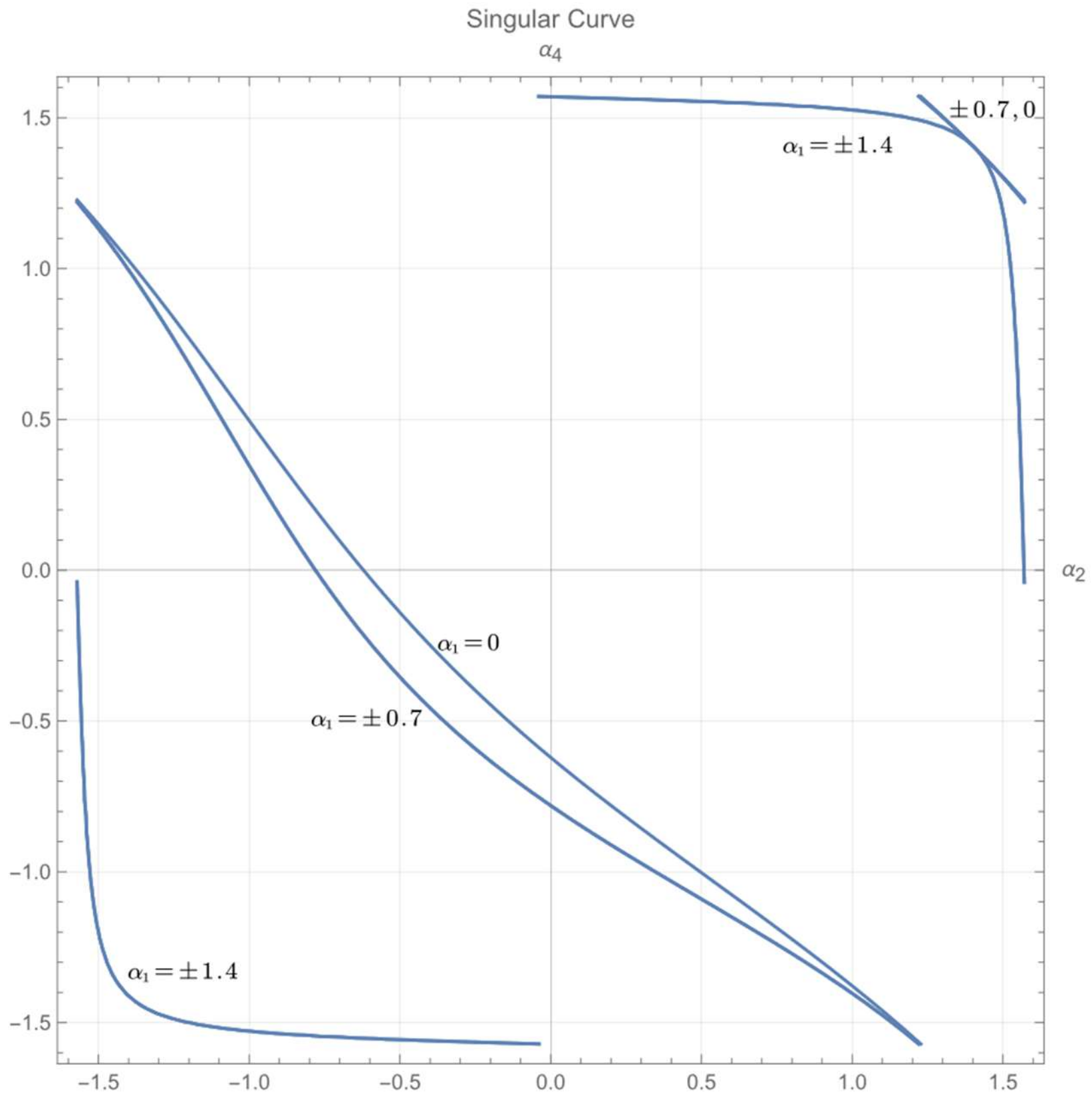

**Figure 7.** ($\alpha_2$,$\alpha_4$) violating Condition (19) ($\alpha_1 = -\alpha_3$).

5.1.4. Case 4 (Negative Half)

This section presents the result for some cases satisfying $\alpha_1 = -\frac{1}{2} \cdot \alpha_3$. Specifically, $\alpha_1 = -\frac{1}{2} \cdot \alpha_3 = -0.3, -0.15, 0, 0.15, 0.3$. Figure 8 plots the left side of Condition (19). The result is five surfaces about ($\alpha_2$,$\alpha_4$). Intercepting the surfaces in Figure 8 by the zero plane (Determinant = 0) yields Figure 9. Figure 9 plots ($\alpha_2$,$\alpha_4$), violating Condition (19).

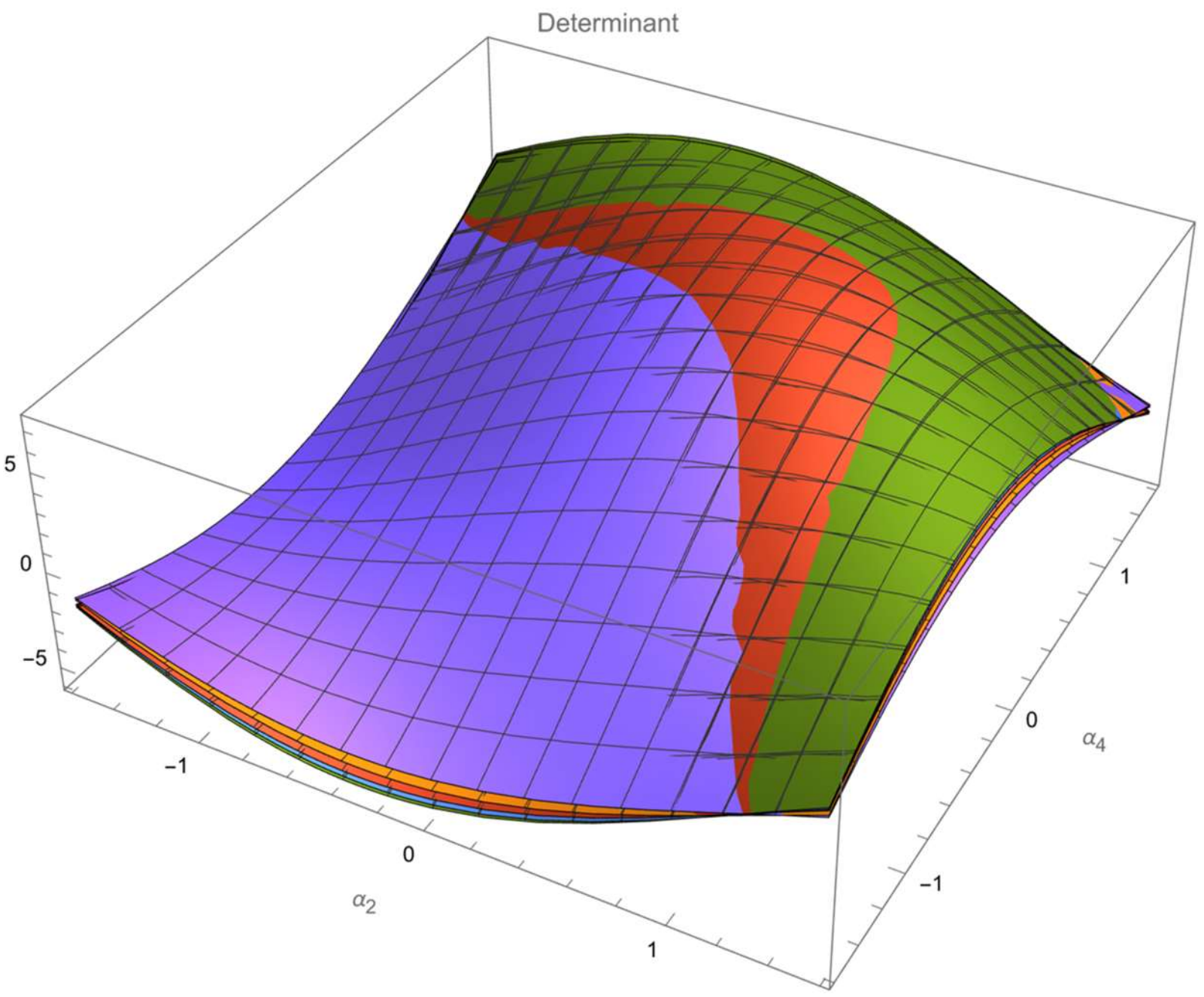

**Figure 8.** Surfaces of the determinant ($\alpha_1 = -\frac{1}{2} \cdot \alpha_3$).

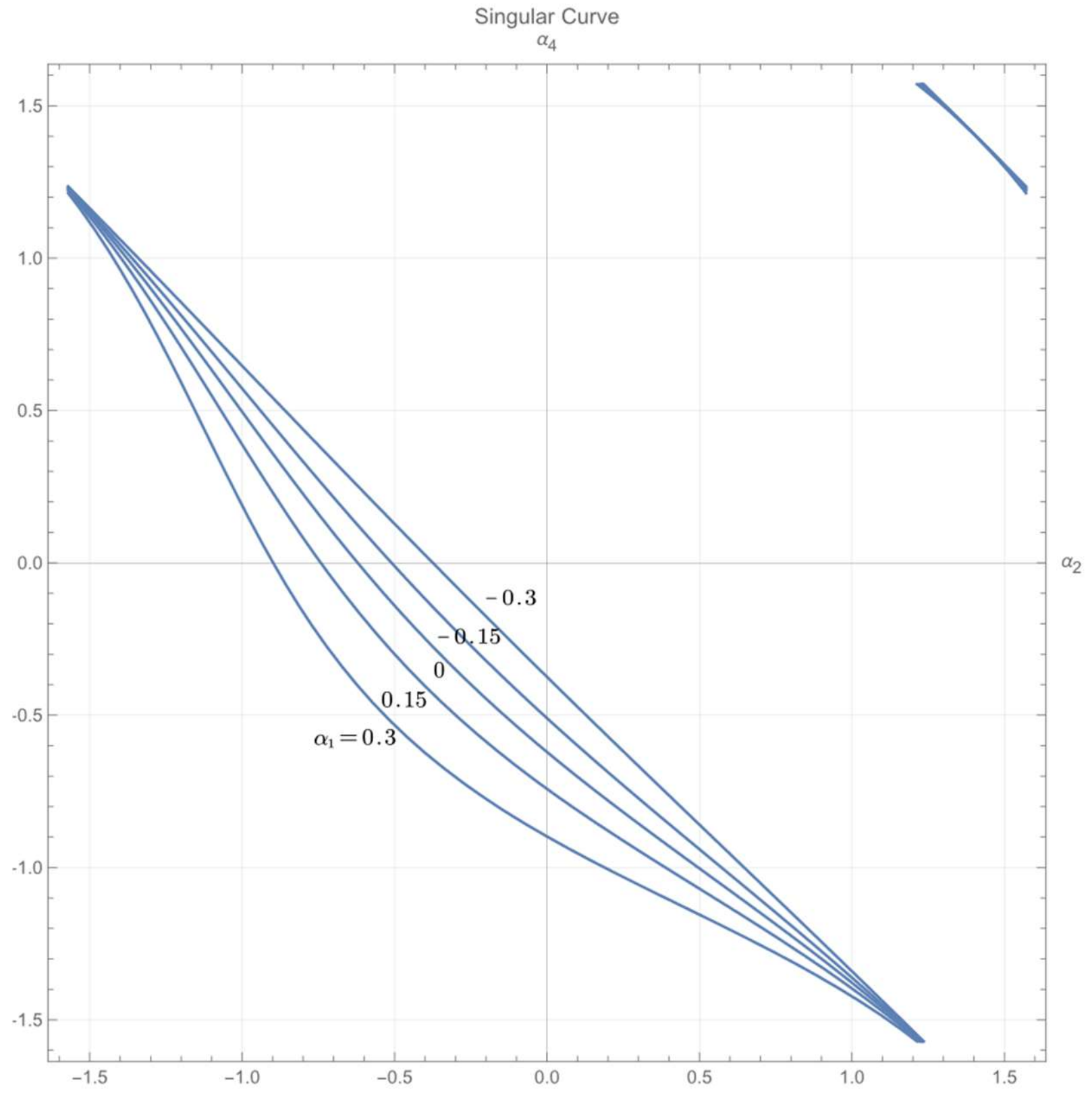

**Figure 9.** ($\alpha_2$,$\alpha_4$) violating Condition (19) ($\alpha_1 = -\frac{1}{2} \cdot \alpha_3$).

*5.2. Interested Gait Region and Direction of Exploration*

Notably, choosing the gait of $(\alpha_2,\alpha_4)$ on the relevant curve in Figures 3, 5, 7 and 9 is strictly prohibited. The decline in the relevant curve indicates the violation of Condition (19). It is also worth mentioning that the tiltrotor becomes the conventional quadrotor in terms of gait, satisfying $(\alpha_1,\alpha_2,\alpha_3,\alpha_4) = (0,0,0,0)$. Our gait analysis includes this special case.

Another point worth mentioning is the continuous requirement of gait switching. For example, switching from $(\alpha_2,\alpha_4) = (1,1)$ to $(\alpha_2,\alpha_4) = (-1,-1)$ is determined to violate Condition (19) for any of the cases shown in Figures 3, 5, 7 and 9. This is because the relevant switching process will cross the curve, which should be prohibited.

Given these concerns, we only choose part of the region containing $(\alpha_2,\alpha_4) = (0,0)$ as the gait region of interest.

5.2.1. Interested Gait Region

It can be concluded from the green region in Figures 10–13 that any $(\alpha_2,\alpha_4)$ located within or on the edge of the triangular zone defined in (22) for each case in Section 5.1 will not violate Condition (19).

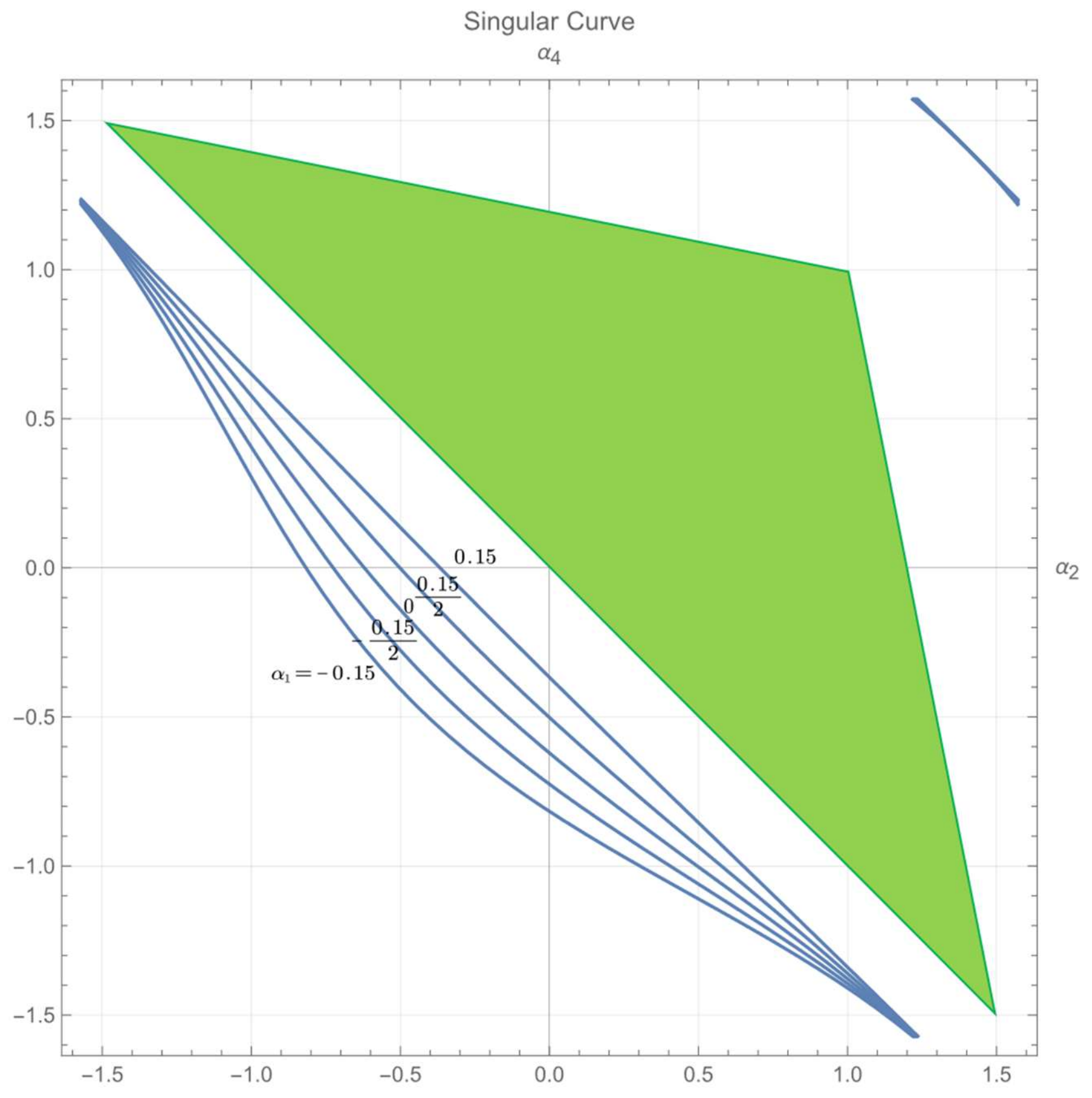

**Figure 10.** The invertible triangular region of interest ($\alpha_1 = \alpha_3$).

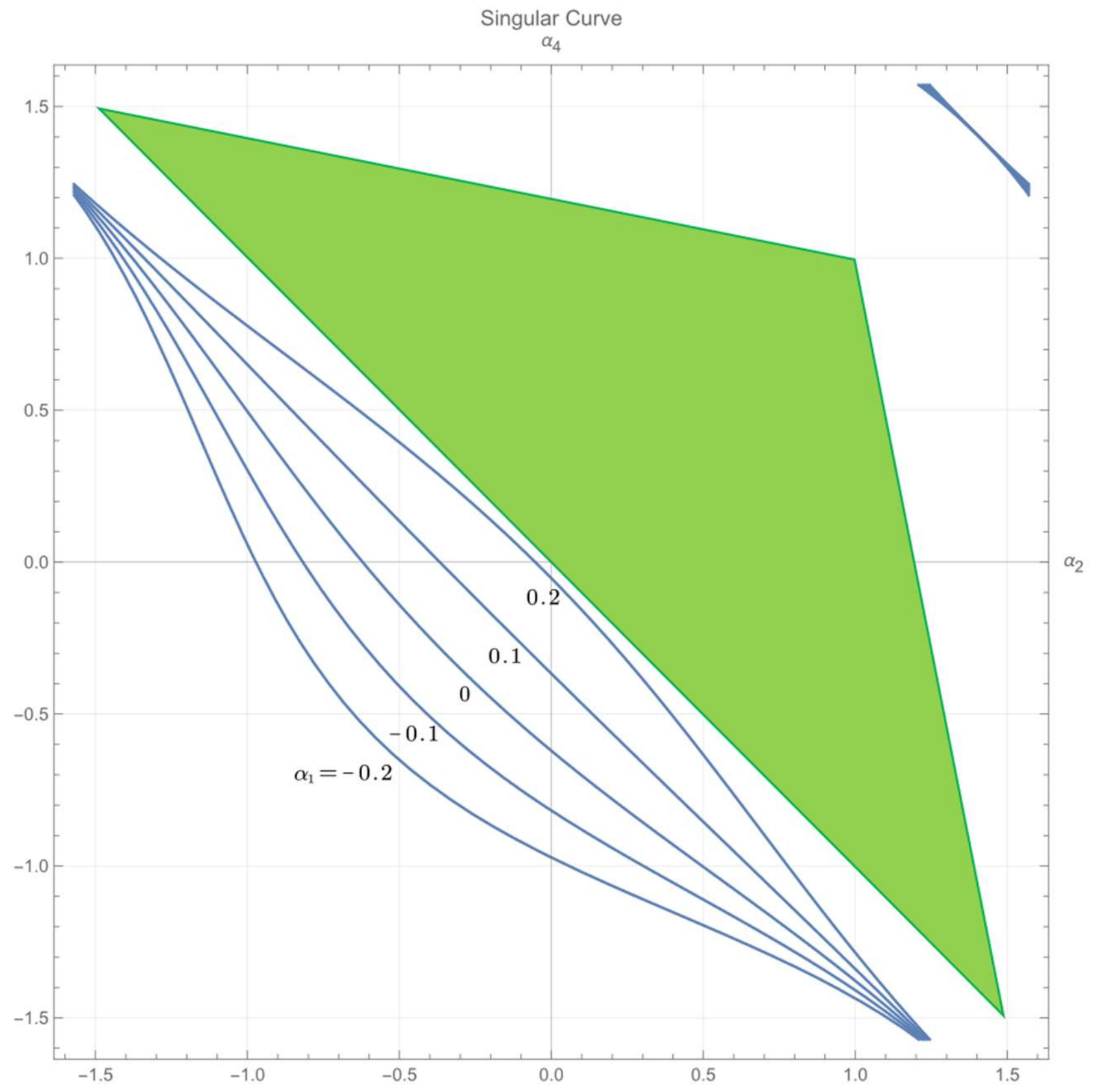

**Figure 11.** The invertible triangular region of interest ($\alpha_1 = \frac{1}{2} \cdot \alpha_3$).

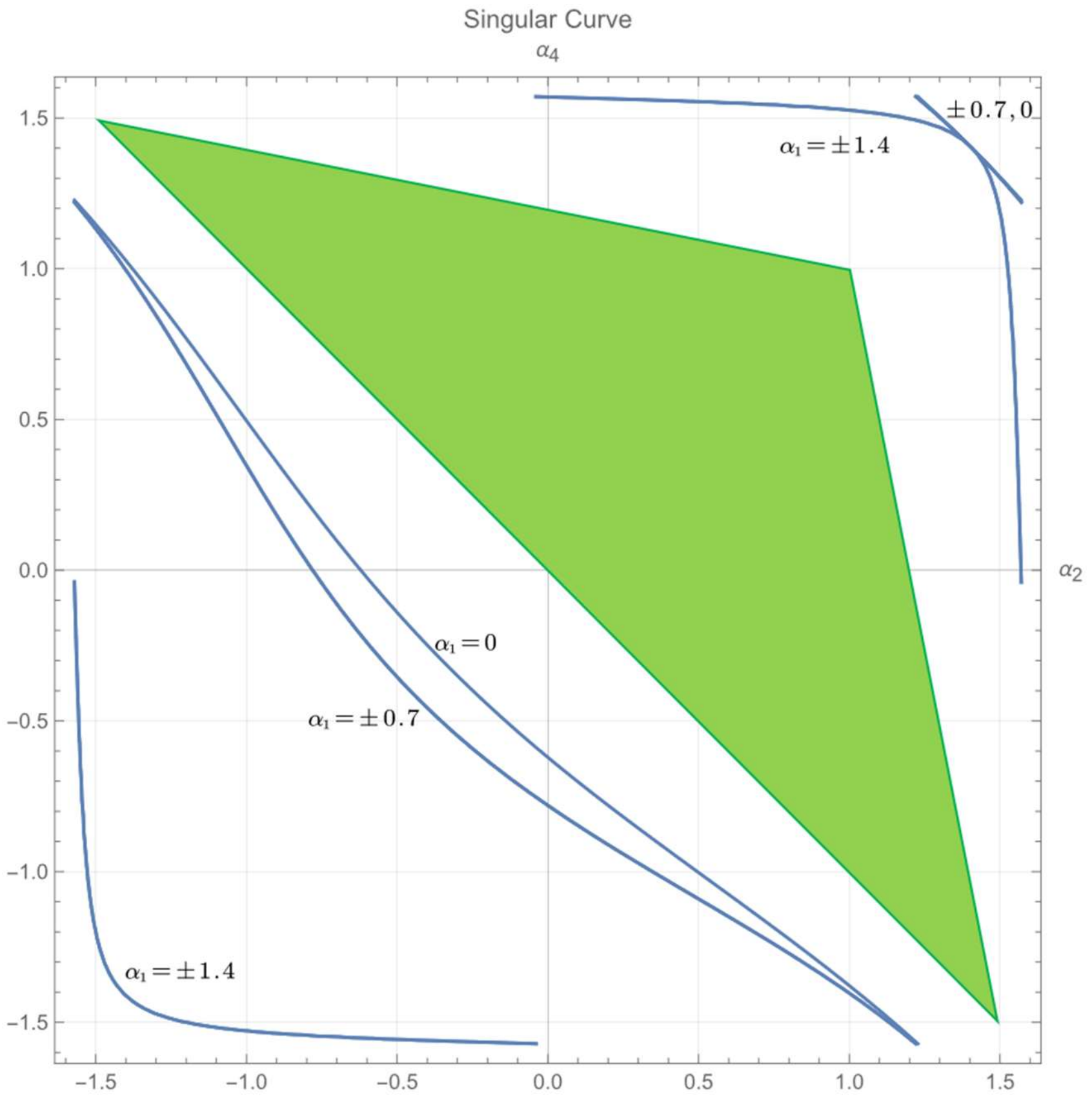

**Figure 12.** The invertible triangular region of interest ($\alpha_1 = -\alpha_3$).

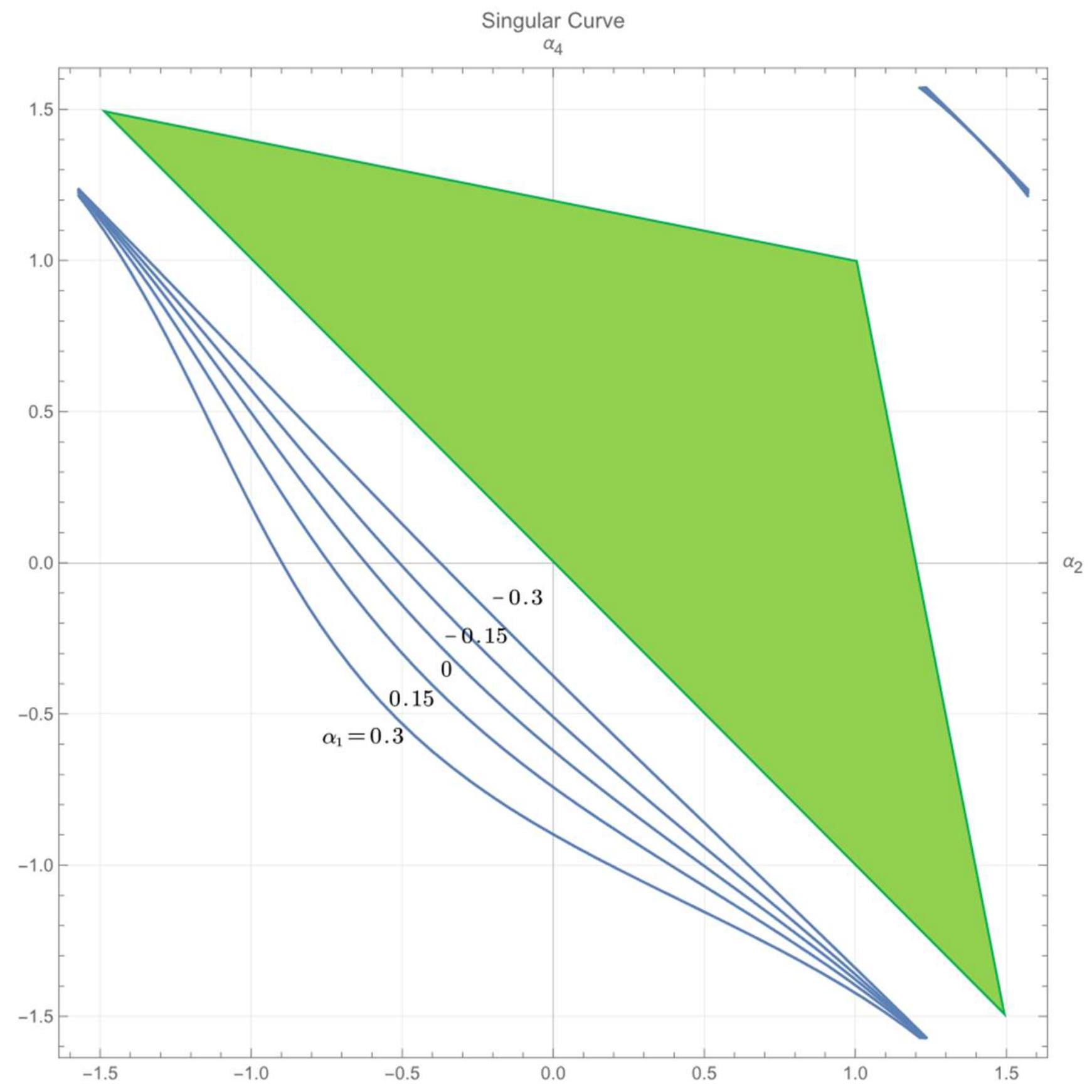

**Figure 13.** The invertible triangular region of interest ($\alpha_1 = -\frac{1}{2} \cdot \alpha_3$).

The rest of the paper focuses on the gait $(\alpha_1, \alpha_2, \alpha_3, \alpha_4)$, satisfying:

$$\text{Region within or on } \Delta MUV \text{ governed by } U(-1.5, 1.5), V(1.5, -1.5), M(1,1). \quad (22)$$

These gaits naturally satisfy the necessary condition given in Proposition 2. Section 5.2.2 outlines the direction of exploration for the simulation identifying the sufficient conditions of stability within the zone in (22).

5.2.2. Direction of Exploration

Exploring the entire space defined in (22) is not necessary or achievable. Thus, we explore parts of this region based on the definition of the directions of exploration in (23)–(25):

$$\alpha_4 = -\alpha_2, \alpha_2 \in \left[-\frac{3}{2}, 0\right]. \quad (23)$$

$$\alpha_4 = -\alpha_2, \alpha_2 \in \left[0, \frac{3}{2}\right]. \quad (24)$$

$$\alpha_4 = \alpha_2, \alpha_2 \in [0, 1]. \quad (25)$$

The exploration in each direction given in (23)–(25) starts from $|\alpha_2| = 0$. With the predetermined $\alpha_1$ and $\alpha_3$, the exploration ends at a critical $\alpha_{2M}$, defined in (26):

$$\begin{gathered} \forall \alpha_2 \; satisfying \; |\alpha_2| \leqslant |\alpha_{2C}|, (\alpha_1, \alpha_2, \alpha_3, \alpha_4) \Rightarrow stable, \\ \exists \alpha_2 \; satisfying \; |\alpha_2| > |\alpha_{2C}|, (\alpha_1, \alpha_2, \alpha_3, \alpha_4) \Rightarrow unstable. \\ \alpha_{2M} = \text{sign}(\alpha_2) \cdot \max(|\alpha_{2C}|). \end{gathered} \quad (26)$$

The expected output is the three gaits $(\alpha_1, \alpha_2, \alpha_3, \alpha_4)$ corresponding to the three $\alpha_{2M}$ along the directions given in (23)–(25).

*5.3. Attitude–Altitude Control*

Feedback linearization relies on the choice of output. In this research, the selected output was the attitude–altitude vector; we focus on only controlling the attitude and altitude. Notably, this control scheme can also be used to reach a desired position with a conventional quadrotor [52], which can be regarded as the special case in our tiltrotor, where $(\alpha_1,\alpha_2,\alpha_3,\alpha_4) = (0,0,0,0)$. However, position control in a tiltrotor was not the primary interest of this study.

In this simulation, the tiltrotor was expected to achieve attitude–altitude self-adjustment.

The initial attitude angles of the tiltrotor were assigned as $\phi_i = 0$, $\theta_i = 0$, and $\phi_i = 0$. The initial angular velocity vector of the tiltrotor with respect to the body-fixed frame was $\omega_B = [0,0,0]^T$. The initial position vector was $[0,0,0]^T$. The initial velocity vector was $[0,0,0]^T$.

The input vector is the derivative of each value of angular velocity of the propeller $[\dot{\varpi}_1,\dot{\varpi}_2,\dot{\varpi}_3,\dot{\varpi}_4]^T$; therefore, assigning an initial velocity for each propeller was necessary. The absolute value of each initial angular velocity was 300 (rad/s). Notably, these angular velocities are not sufficient to compensate for the effect of gravity, even for the case $(\alpha_1,\alpha_2,\alpha_3,\alpha_4) = (0,0,0,0)$.

The reference was a four-dimensional attitude–altitude vector, $[\phi_r,\theta_r,\psi_r,Z_r]^T$. To maintain zero attitude and zero height, the designed reference was $[0,0,0,0]^T$. Based on the control parameters set in Section 3 and each gait designed in Section 5, we recorded the gaits giving stable results.

## 6. Results

This section presents the results of the tests in the previous section. Section 6.1 shows the histories of the attitude, altitude, and angular velocities during flights of two typical gaits. Section 6.2 displays the admissible gaits. These are the results of the work described in Section 5.2.2. Section 6.3 presents the results for the same task completed with another control strategy, which utilized all the inputs (over-actuated) on the same tiltrotor with identical parameters.

### *6.1. Flight History*

Section 6.1 presents the results from two gaits. Gait 1: $(\alpha_1,\alpha_2,\alpha_3,\alpha_4) = (-0.1, 0.1, -0.2, 0.1)$. Gait 2: $(\alpha_1,\alpha_2,\alpha_3,\alpha_4) = (-0.15, -0.1, 0.3, -0.1)$.

#### 6.1.1. Gait 1

For the gait $(\alpha_1,\alpha_2,\alpha_3,\alpha_4) = (-0.1, 0.1, -0.2, 0.1)$, the attitude history, altitude history and angular velocities history of the propellers are plotted in Figures 14–16, respectively.

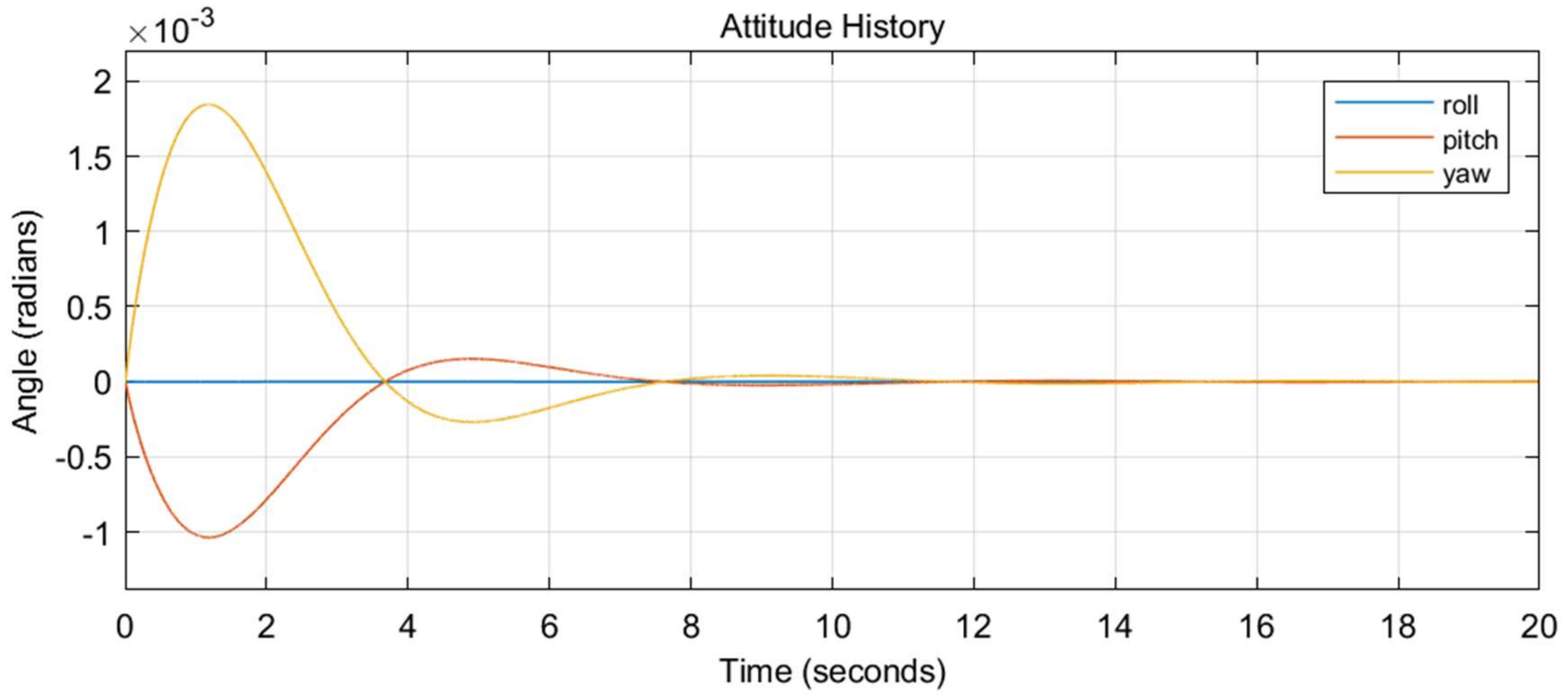

**Figure 14.** Attitude history of Gait 1.

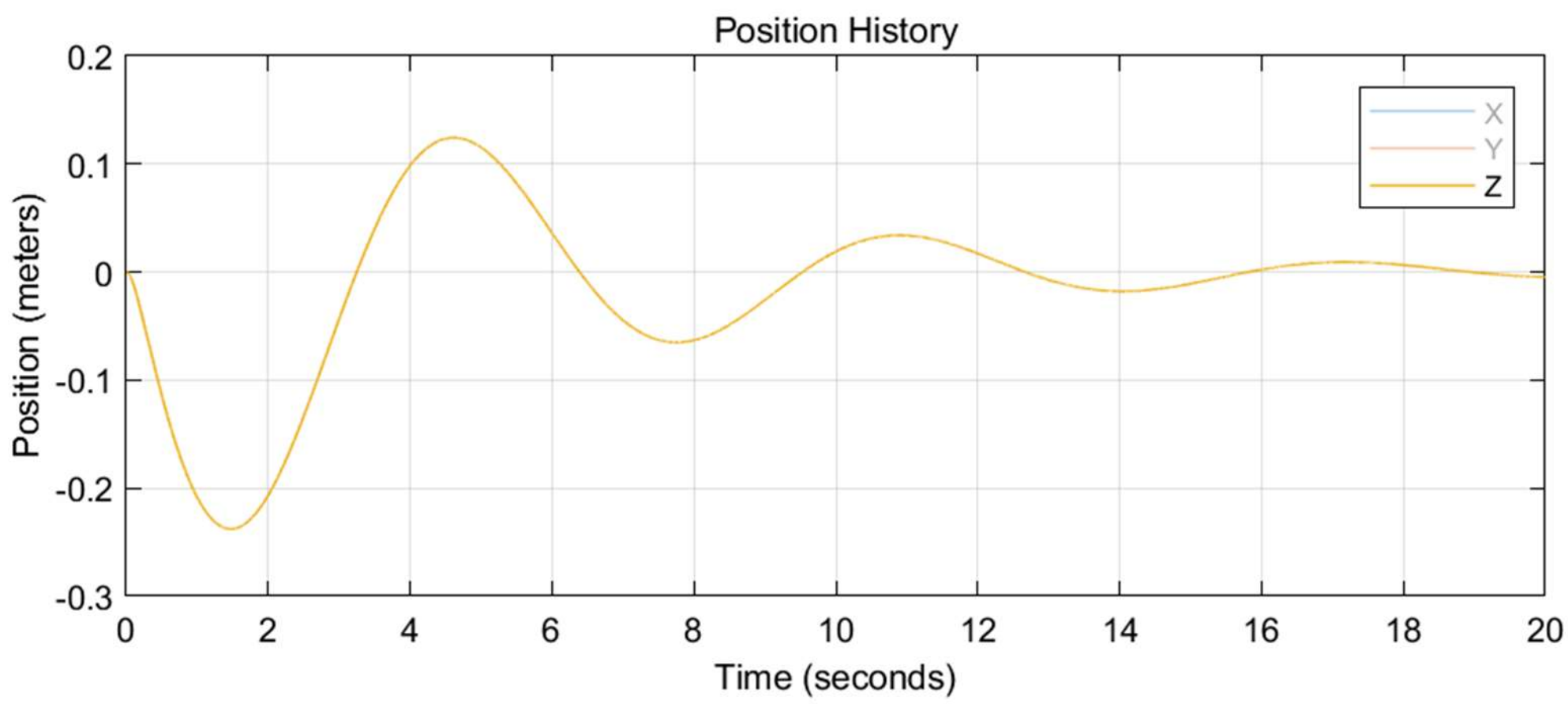

**Figure 15.** Altitude history of Gait 1.

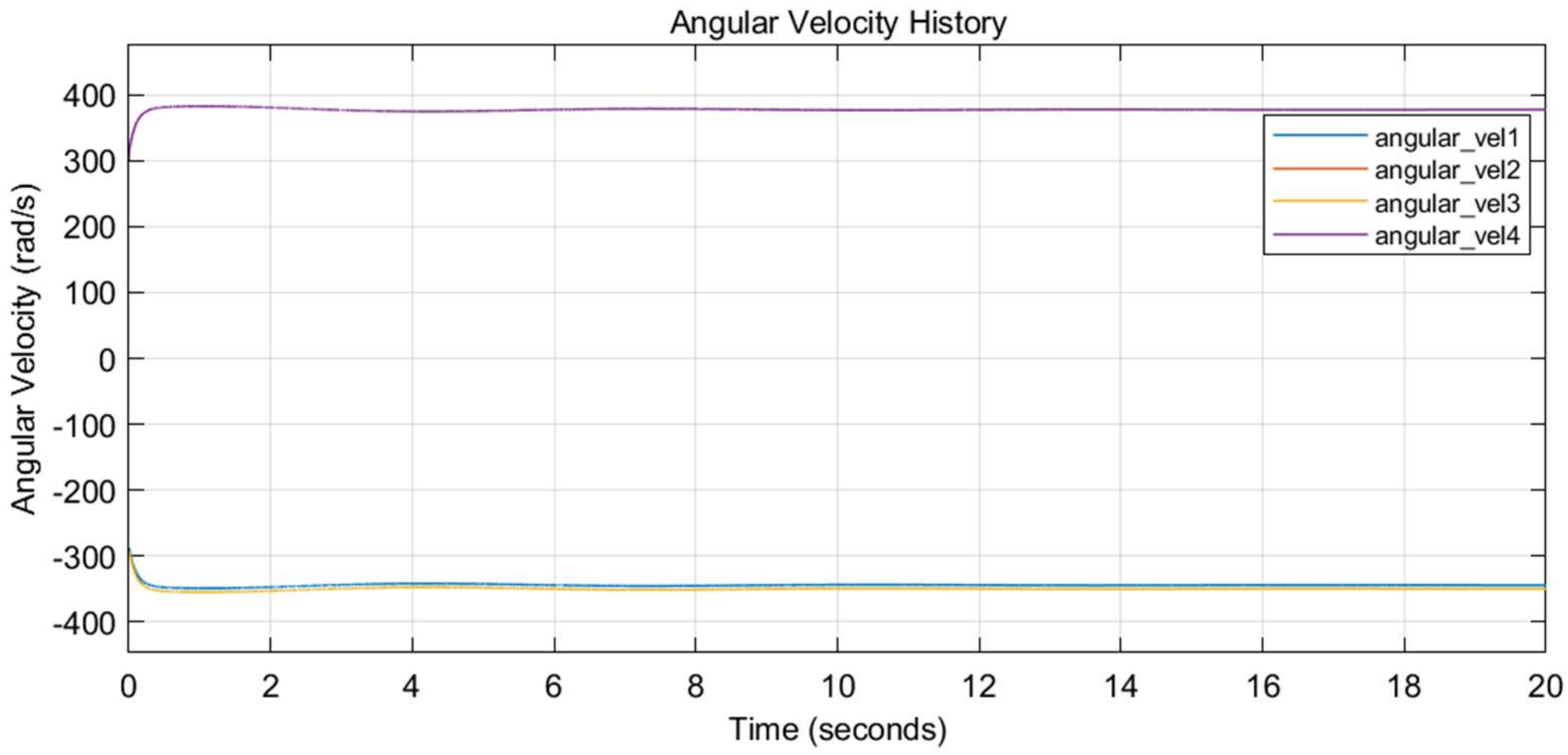

**Figure 16.** Angular velocity history of Gait 1.

Notably, none of the angular velocities touched 0, which guarantees the invertibility of the decoupling matrix in (12).

6.1.2. Gait 2

For the gait $(\alpha_1, \alpha_2, \alpha_3, \alpha_4) = (-0.15, -0.1, 0.3, -0.1)$, the attitude history, altitude history and angular velocities history of the propellers are plotted in Figures 17–19, respectively.

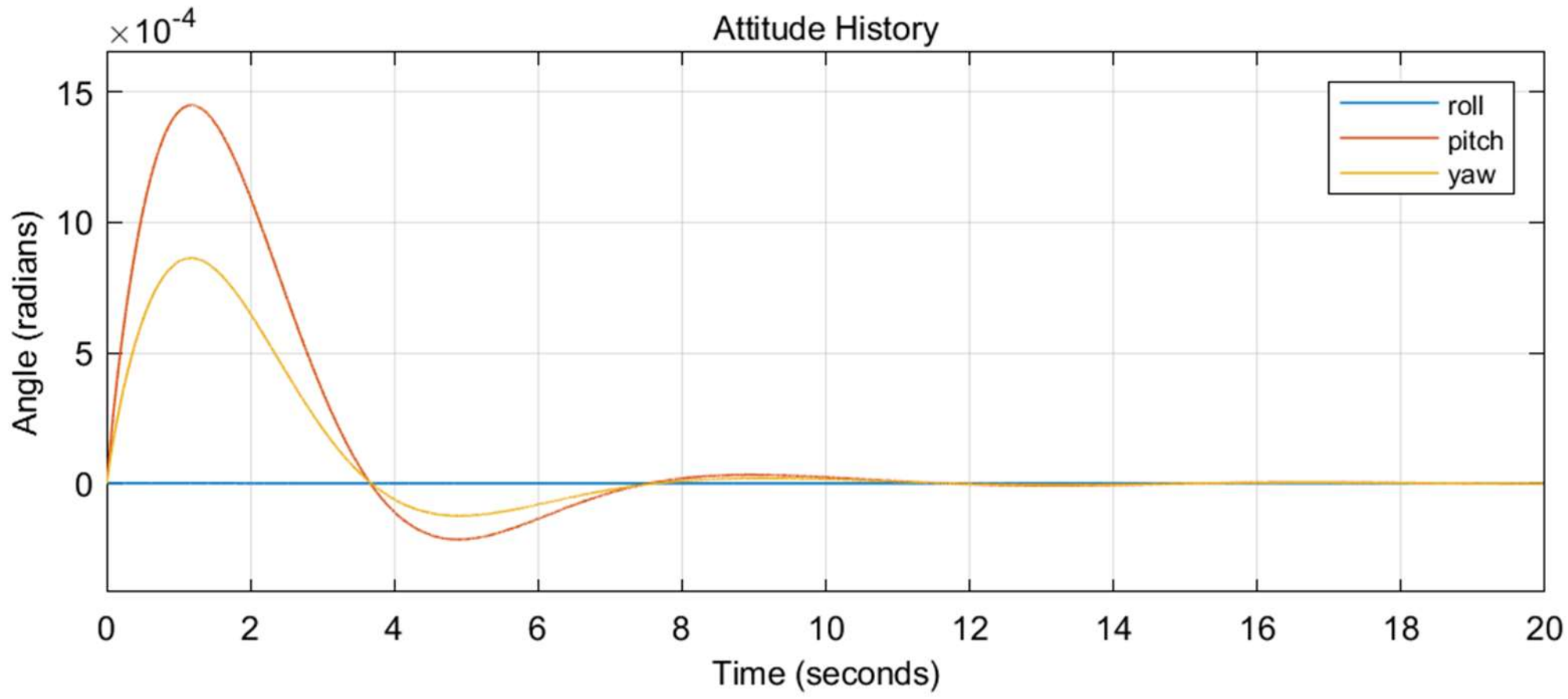

**Figure 17.** Attitude history of Gait 2.

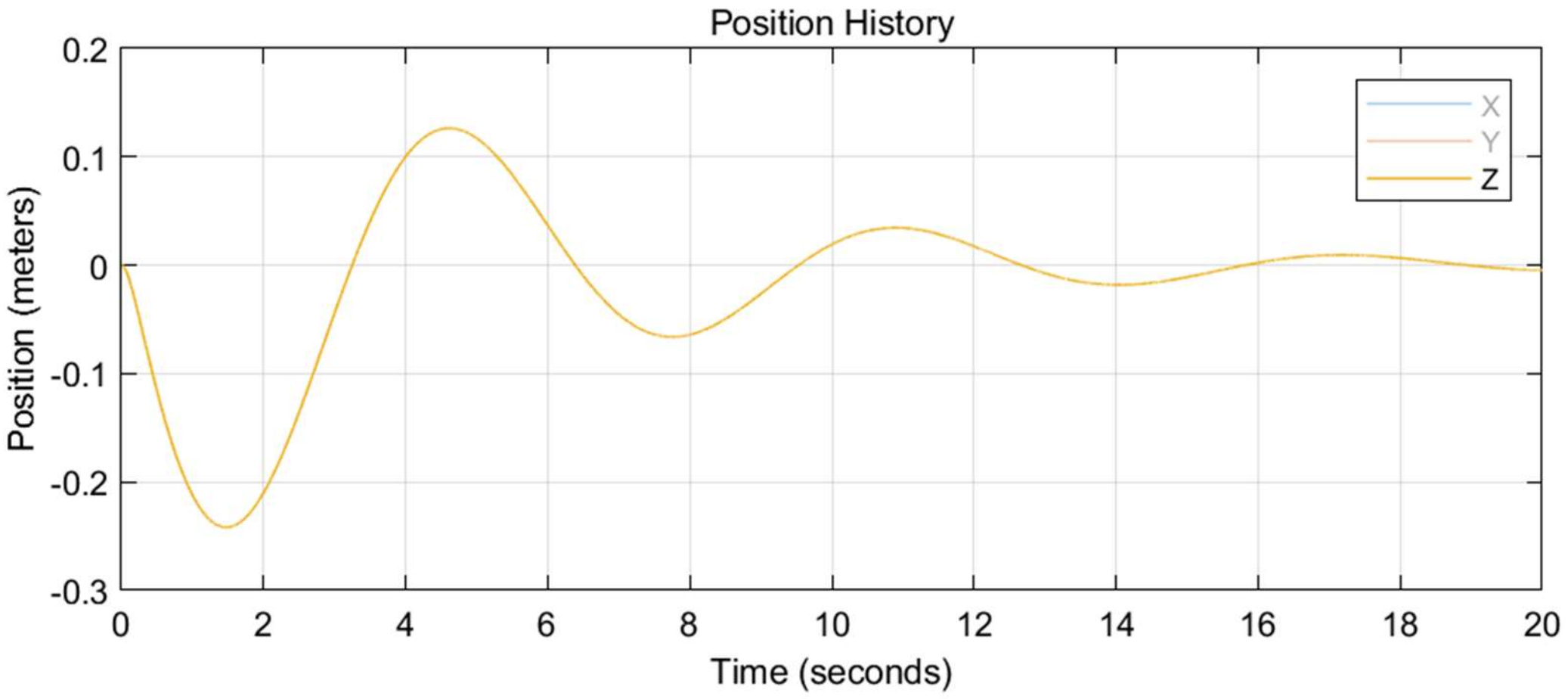

**Figure 18.** Altitude history of Gait 2.

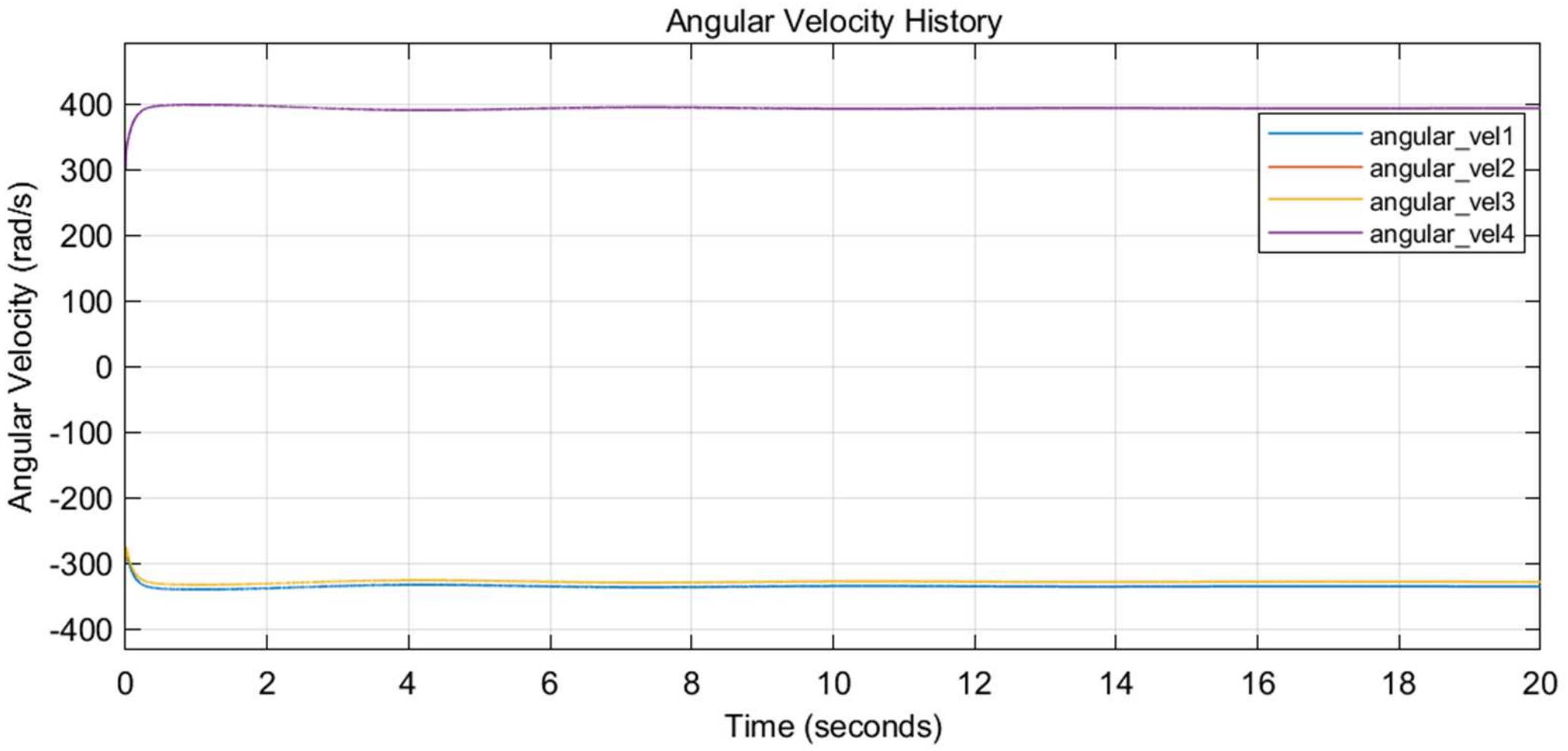

**Figure 19.** Angular velocity history of Gait 2.

Similarly, none of the angular velocities touched 0.

## *6.2. Applicable Gait (Sufficient Conditions)*

This section displays the applicability of each gait from Section 5.1.1–Section 5.1.4 (with five different ($\alpha_1$,$\alpha_3$) pairs in each section), determined via the method in (26).

Surprisingly, the resulting applicable values of ($\alpha_2$,$\alpha_4$) leading to stable results in each ($\alpha_1$,$\alpha_3$) in 20 (5 × 4) restrictions were highly similar. Except for one gait (($\alpha_1$,$\alpha_3$) = (0.2,0.4)), which led to instability, the regions of applicable ($\alpha_2$,$\alpha_4$) for the remaining 19 cases contained the same triangular region governed by (−1.3,1.3), (1.3,−1.3), and (1,1).

The specific gaits leading to stable control results for different ($\alpha_1$,$\alpha_3$) are detailed in Figures 20–23.

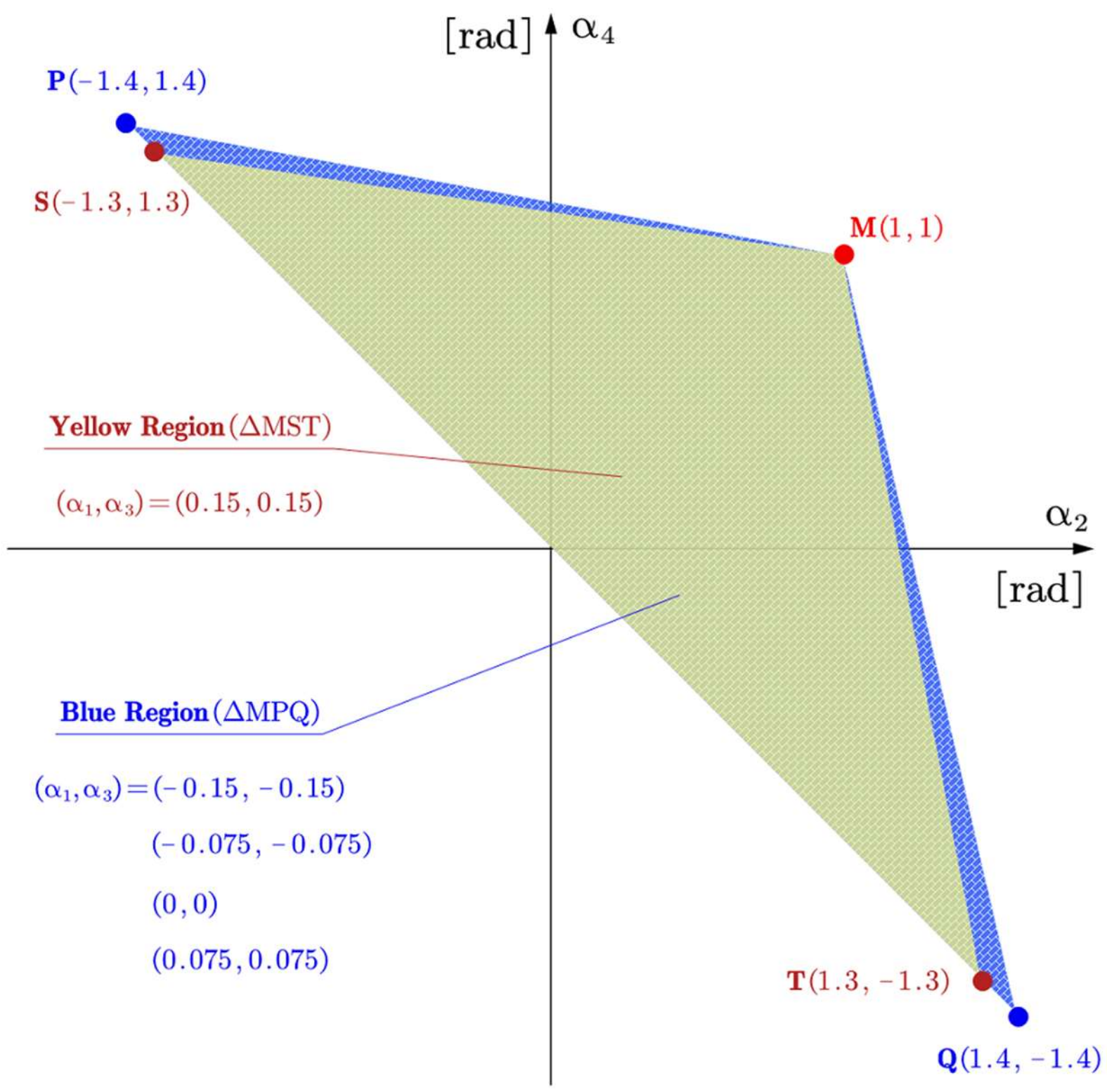

**Figure 20.** Admissible ($\alpha_2$,$\alpha_4$) when $\alpha_1 = \alpha_3$. M represents the gaits satisfying $\alpha_2 = \alpha_4 = 1$. P, S, T, Q represents the gaits satisfying ($\alpha_2$,$\alpha_4$) = (−1.4,1.4), (−1.3,1.3), (1.3,−1.3), (1.4,−1.4), respectively.

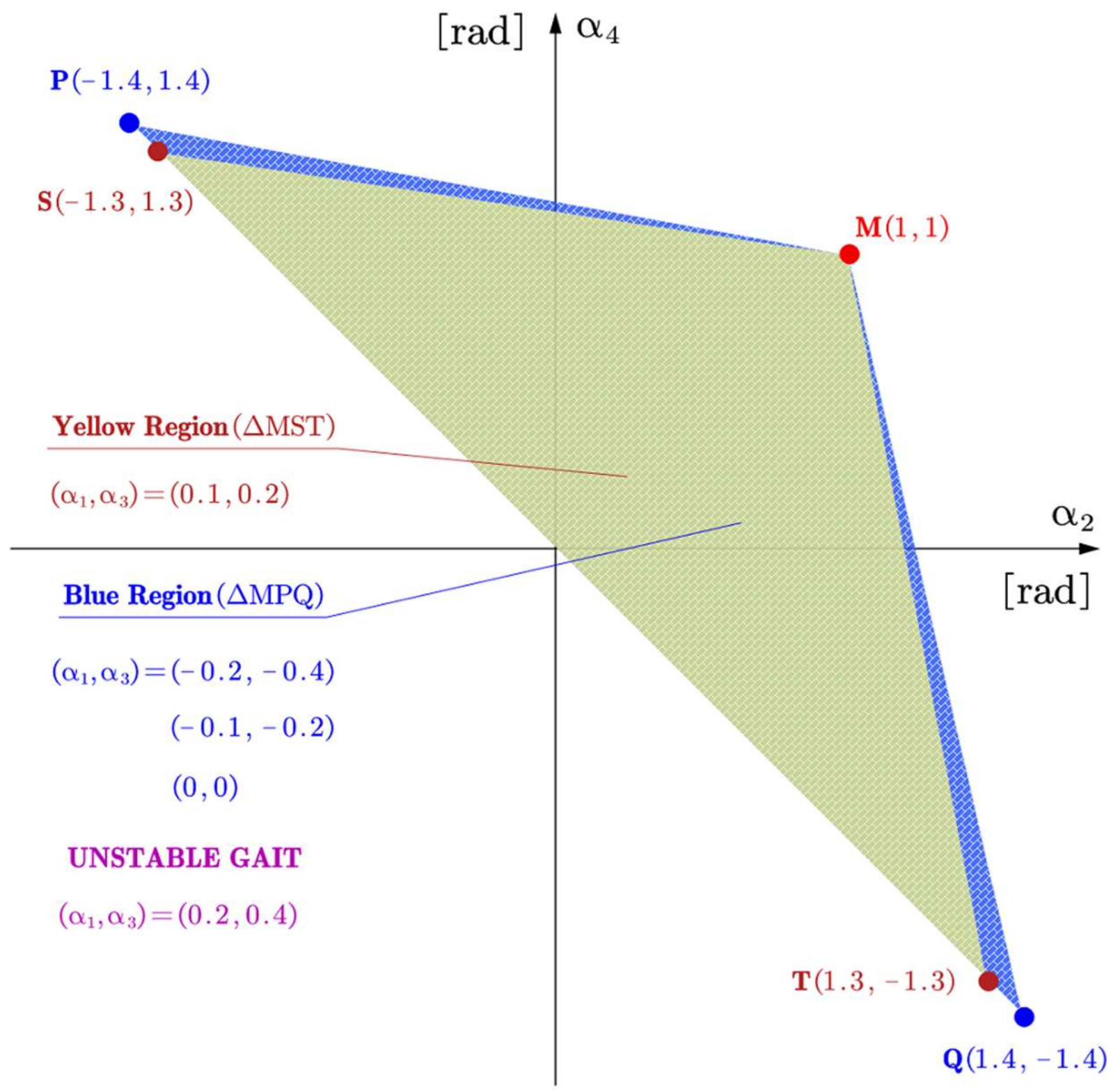

**Figure 21.** Admissible ($\alpha_2$,$\alpha_4$) when $\alpha_1 = \frac{1}{2}\alpha_3$. M represents the gaits satisfying $\alpha_2 = \alpha_4 = 1$. P, S, T, Q represent the gaits satisfying ($\alpha_2$,$\alpha_4$) = (−1.4,1.4), (−1.3,1.3), (1.3,−1.3), (1.4,−1.4), respectively.

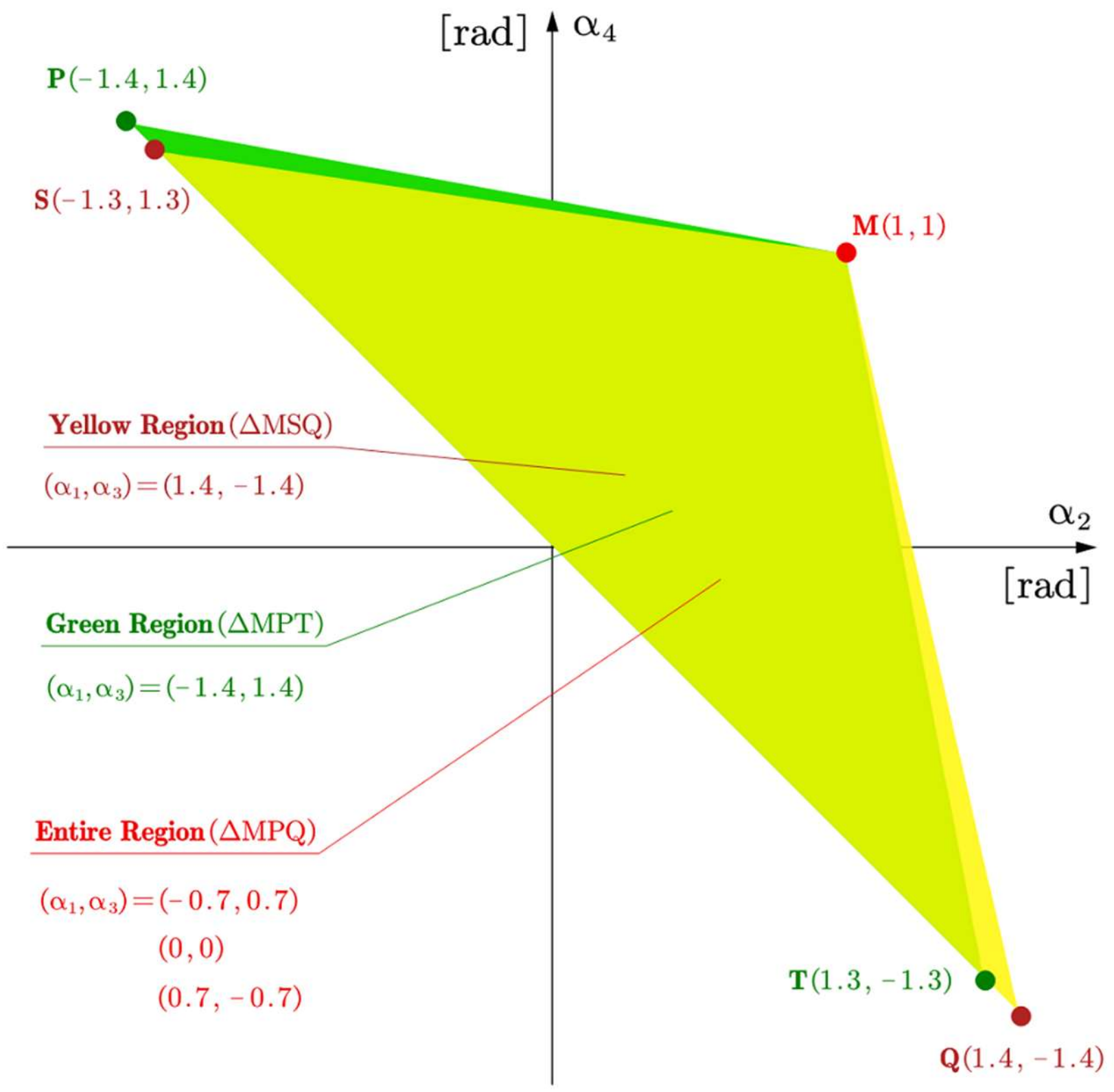

**Figure 22.** Admissible $(\alpha_2,\alpha_4)$ when $\alpha_1 = -\alpha_3$. M represents the gaits satisfying $\alpha_2 = \alpha_4 = 1$. P, S, T, Q represent the gaits satisfying $(\alpha_2,\alpha_4) = (-1.4,1.4), (-1.3,1.3), (1.3,-1.3), (1.4,-1.4)$, respectively.

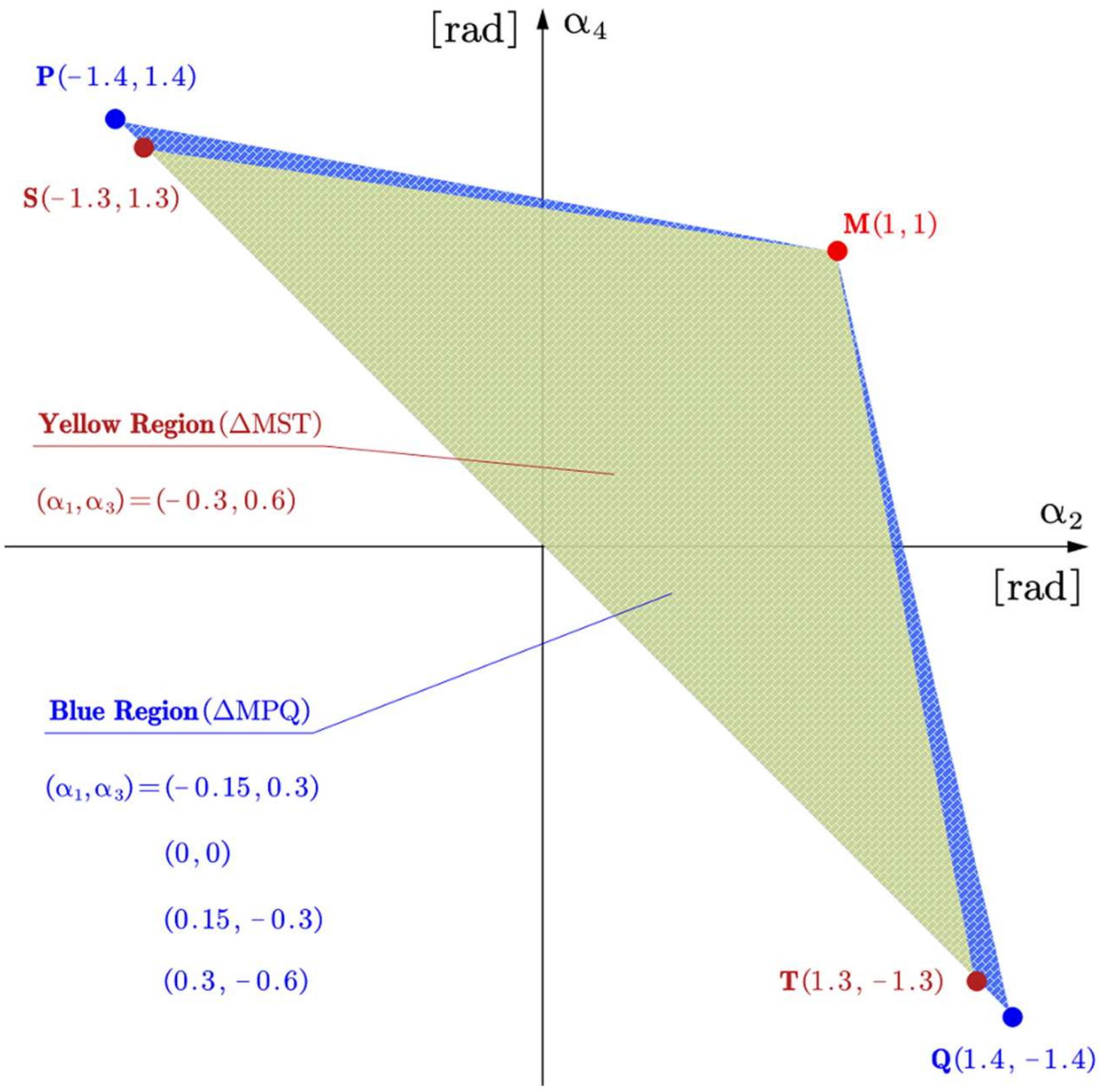

**Figure 23.** Admissible $(\alpha_2,\alpha_4)$ when $\alpha_1 = -\frac{1}{2}\alpha_3$. M represents the gaits satisfying $\alpha_2 = \alpha_4 = 1$. P, S, T, Q represent the gaits satisfying $(\alpha_2,\alpha_4) = (-1.4,1.4), (-1.3,1.3), (1.3,-1.3), (1.4,-1.4)$, respectively.

### 6.3. Over-Actuated Control

This section presents the control result for an over-actuated controller with eight control inputs, used to complete the same task with identical parameters and initial conditions. The attitude history, the altitude history, the angular velocities history of the propellers, and the tilting angles history are plotted in Figures 24–27, respectively.

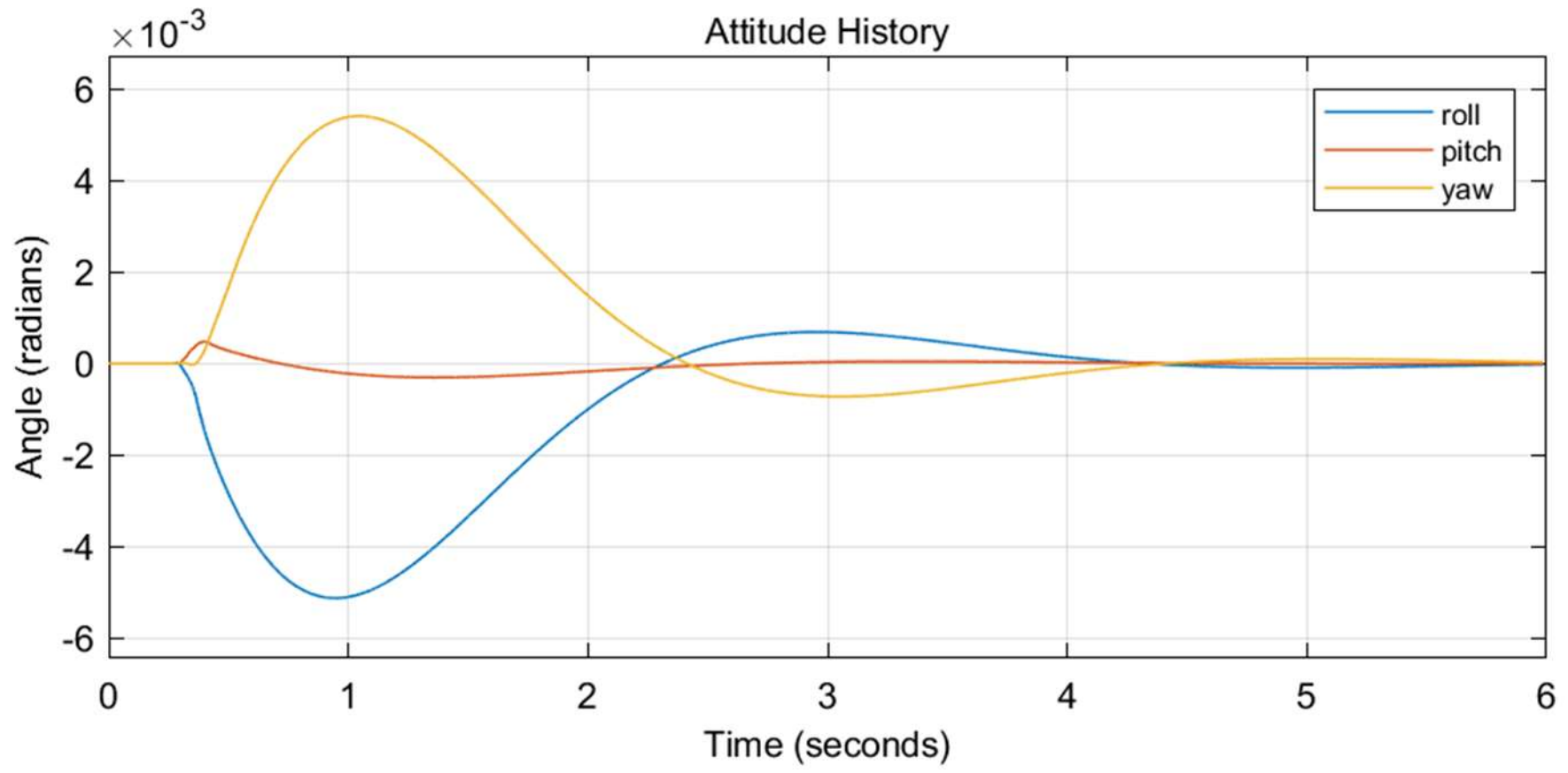

**Figure 24.** Attitude history of over-actuated control.

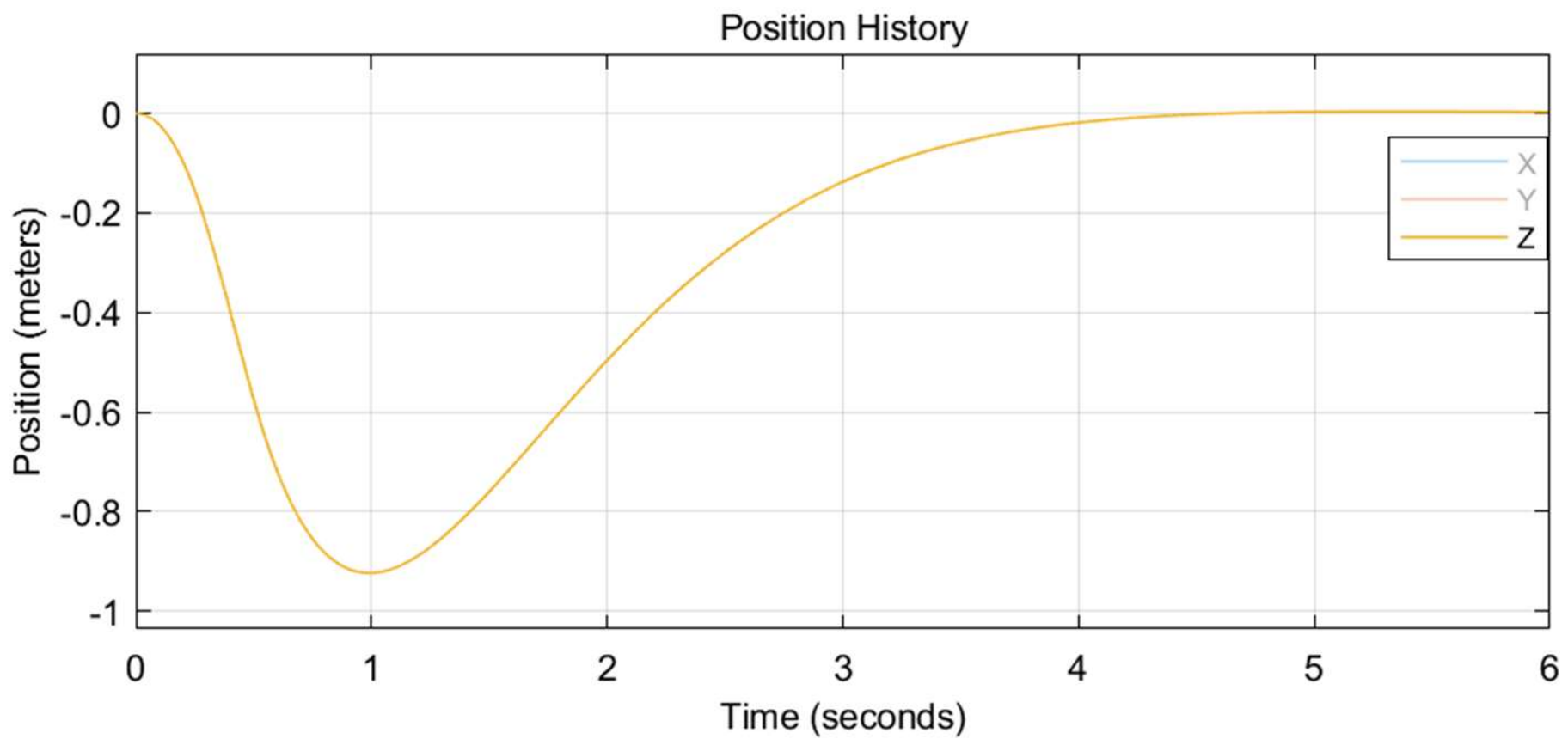

**Figure 25.** Attitude history of over-actuated control.

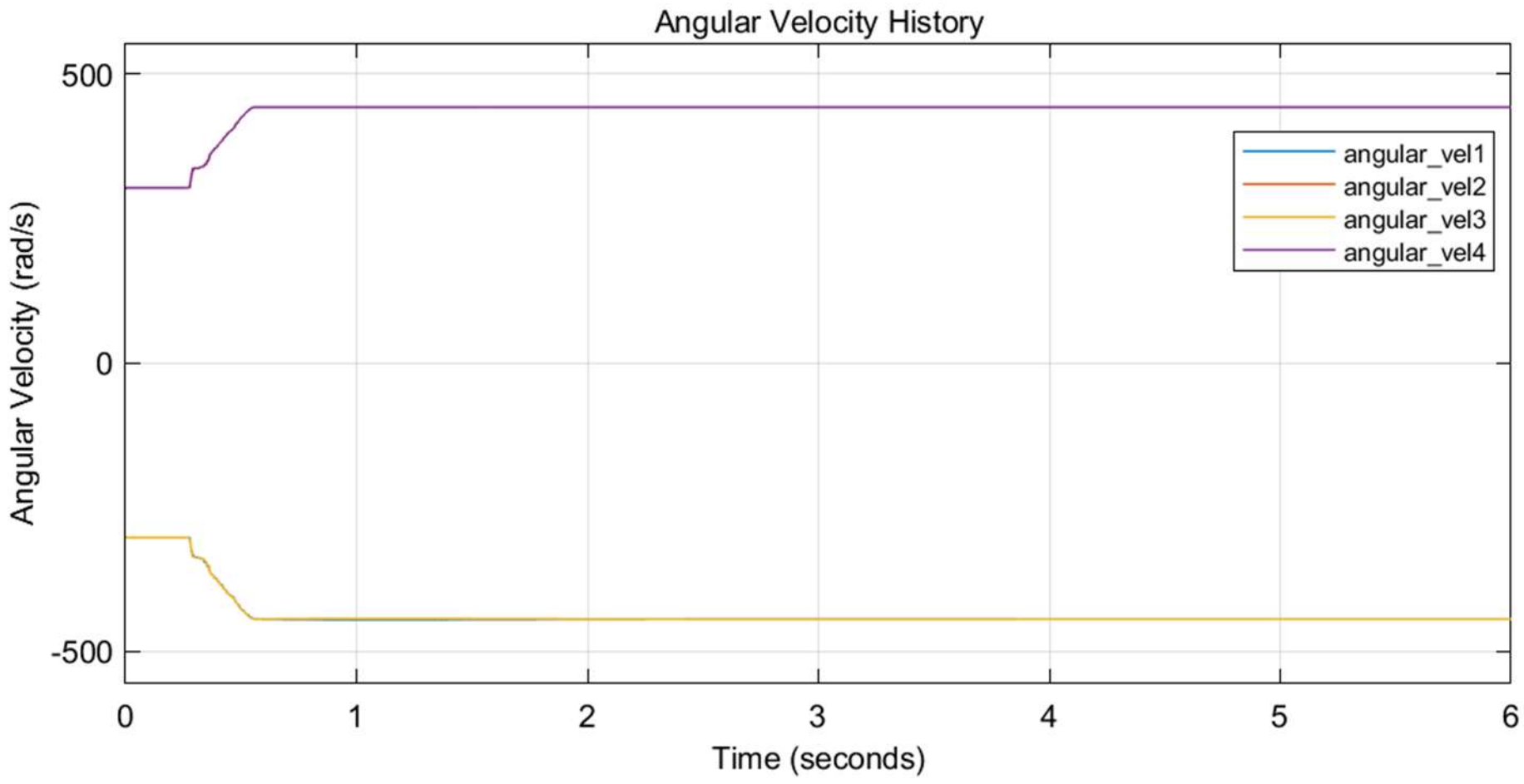

**Figure 26.** Angular velocity history of over-actuated control.

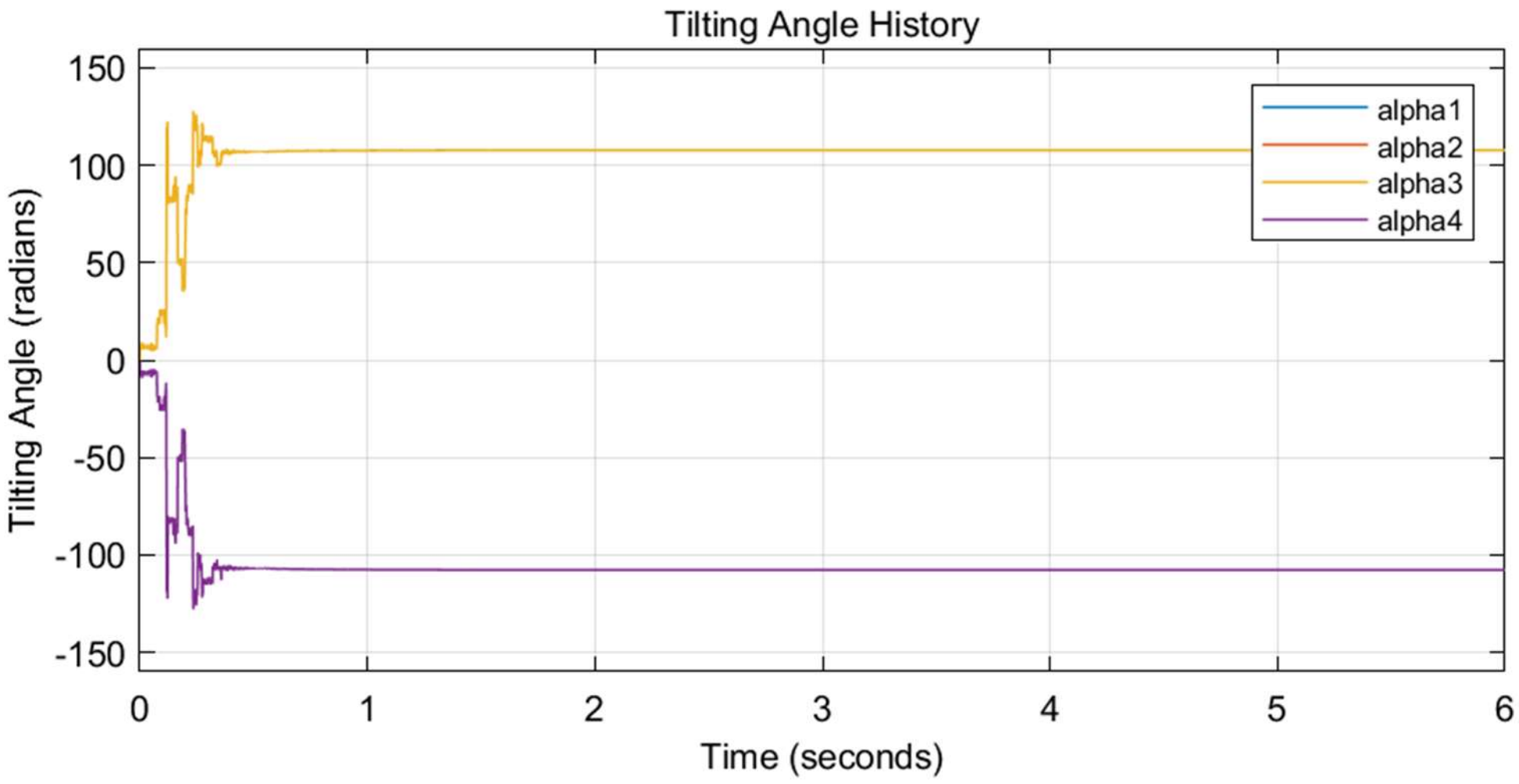

**Figure 27.** Tilting angle history of over-actuated control.

Although the response was faster than in the control method proposed in this research, the tilting angles changed over-rapidly and intensively, which is commonly believed to be unrealistic.

## 7. Conclusions and Discussions

The necessary conditions to develop an invertible decoupling matrix for the attitude–altitude-based feedback linearization method have been determined and visualized for a tiltrotor.

For the gait restricted by $\alpha_1 = \alpha_3, \alpha_1 = \frac{1}{2} \cdot \alpha_3, \alpha_1 = -\alpha_3, \alpha_1 = -\frac{1}{2} \cdot \alpha_3$ within the specific region of interest given in (22), the feedback linearization had great success in terms of stabilization if, and only if, ($\alpha_2$,$\alpha_4$) lay inside the relevant region defined in Figures 20–23.

The angular velocities in each simulation that manifested a stable result avoided touching the non-negative constraints.

When the over-actuated controller was used for completing the same task, the tilting angle was maneuvered over-rapidly and intensively. In return, the tiltrotor received a faster response and less overshoot.

For the defined ($\alpha_1$,$\alpha_3$) of interest, the control result is very likely to be unstable if ($\alpha_2$,$\alpha_4$) lies outside the relevant colored areas in Figures 20–23.

These conclusions can offer effective guidance for designing the gait in a safe (invertible) region. For example, a different kind of feedback-linearization-based gait may allow ($\alpha_2$,$\alpha_4$) to lie within the triangular region governed by (−1.3,1.3), (1.3,−1.3), and (1,1), because this region is most closely related to achieving a stable result.

The $\alpha_2 - \alpha_4$ planes (Figures 3, 5, 7 and 9) in Section 5.2 were divided into several regions by the invertibility-violating curves. We have argued that the applicable gait should lie in the same region. Furthermore, we asserted that crossing these curves is inevitable when switching from the gait in one region to the gait in another region. However, this is not always true.

It can be proven that the decoupling matrix for the gait ($\alpha_1$,$\alpha_2$,$\alpha_3$,$\alpha_4$) = (0,0,0,0) is always invertible, introducing no invertibility-violating curves. Thus, one may avoid crossing the invertibility-violating curves while switching the gait by adjusting ($\alpha_1$,$\alpha_3$) at the same time, e.g., the tiltrotor may return to the safe gait ($\alpha_1$,$\alpha_2$,$\alpha_3$,$\alpha_4$) = (0,0,0,0) as the middle step before switching to the desired gait. Further discussions are beyond the scope of this research.

Our next step is to develop a periodical gait for the tiltrotor in order to track a more complicated reference.

## References

1. Ryll, M.; Bülthoff, H.H.; Giordano, P.R. Modeling and Control of a Quadrotor UAV with Tilting Propellers. In Proceedings of the IEEE International Conference on Robotics and Automation, Saint Paul, MN, USA, 14–18 May 2012.

2. Senkul, F.; Altug, E. Adaptive Control of a Tilt-Roll Rotor Quadrotor UAV. In Proceedings of the 2014 International Conference on Unmanned Aircraft Systems, ICUAS 2014, Orlando, FL, USA, 27–30 May 2014.
3. Senkul, F.; Altug, E. Modeling and Control of a Novel Tilt-Roll Rotor Quadrotor UAV. In Proceedings of the 2013 International Conference on Unmanned Aircraft Systems, ICUAS 2013, Atlanta, GA, USA, 28–31 May 2013.
4. Nemati, A.; Kumar, M. Modeling and Control of a Single Axis Tilting Quadcopter. In Proceedings of American Control Conference, Portland, OR, USA, 4–6 June 2014.
5. Bin Junaid, A.; De Cerio Sanchez, A.D.; Bosch, J.B.; Vitzilaios, N.; Zweiri, Y. Design and Implementation of a Dual-Axis Tilting Quadcopter. *Robotics* **2018**, *7*, 65. https://doi.org/10.3390/robotics7040065.
6. Andrade, R.; Raffo, G.V.; Normey-Rico, J.E. Model Predictive Control of a Tilt-Rotor UAV for Load Transportation. In Proceedings of the 2016 European Control Conference, ECC 2016, Aalborg, Denmark, 29 June–1 July 2016.
7. Nemati, A.; Kumar, R.; Kumar, M. Stabilizing and Control of Tilting-Rotor Quadcopter in Case of a Propeller Failure. In Proceedings of the ASME 2016 Dynamic Systems and Control Conference, DSCC 2016, Minneapolis, MN, USA, 12–14 October 2016; Volume 1.
8. Anderson, R.B.; Marshall, J.A.; L'Afflitto, A. Constrained Robust Model Reference Adaptive Control of a Tilt-Rotor Quadcopter Pulling an Unmodeled Cart. *IEEE Trans. Aerosp. Electron. Syst.* **2021**, *57*, 39–54. https://doi.org/10.1109/TAES.2020.3008575.
9. Badr, S.; Mehrez, O.; Kabeel, A.E. A Design Modification for a Quadrotor UAV: Modeling, Control and Implementation. *Adv. Robot.* **2019**, *33*, 13–32. https://doi.org/10.1080/01691864.2018.1556116.
10. Bhargavapuri, M.; Patrikar, J.; Sahoo, S.R.; Kothari, M. A Low-Cost Tilt-Augmented Quadrotor Helicopter: Modeling and Control. In Proceedings of the 2018 International Conference on Unmanned Aircraft Systems, ICUAS 2018, Dallas, TX, USA, 12–15 June 2018.
11. Badr, S.; Mehrez, O.; Kabeel, A.E. A Novel Modification for a Quadrotor Design. In Proceedings of the 2016 International Conference on Unmanned Aircraft Systems, ICUAS 2016, Arlington, VA, USA, 7–10 June 2016.
12. Jiang, X.-Y.; Su, C.-L.; Xu, Y.-P.; Liu, K.; Shi, H.-Y.; Li, P. An Adaptive Backstepping Sliding Mode Method for Flight Attitude of Quadrotor UAVs. *J. Cent. South Univ.* **2018**, *25*, 616–631. https://doi.org/10.1007/s11771-018-3765-0.
13. Jin, S.; Kim, J.; Kim, J.W.; Bae, J.H.; Bak, J.; Kim, J.; Seo, T.W. Back-Stepping Control Design for an Underwater Robot with Tilting Thrusters. In Proceedings of the 17th International Conference on Advanced Robotics, ICAR 2015, Istanbul, Turkey, 27–31 July 2015.
14. Kadiyam, J.; Santhakumar, M.; Deshmukh, D.; Seo, T.W. Design and Implementation of Backstepping Controller for Tilting Thruster Underwater Robot. In Proceedings of the International Conference on Control, Automation and Systems, PyeongChang, Korea, 17–20 October 2018; Volume 2018.
15. Scholz, G.; Popp, M.; Ruppelt, J.; Trommer, G.F. Model Independent Control of a Quadrotor with Tiltable Rotors. In Proceedings of the IEEE/ION Position, Location and Navigation Symposium, PLANS 2016, Savannah, GA, USA, 11–14 April 2016.
16. Phong Nguyen, N.; Kim, W.; Moon, J. Observer-Based Super-Twisting Sliding Mode Control with Fuzzy Variable Gains and Its Application to Overactuated Quadrotors. In Proceedings of the IEEE Conference on Decision and Control, Miami, FL, USA, 17–19 December 2018; Volume 2018.
17. Kumar, R.; Nemati, A.; Kumar, M.; Sharma, R.; Cohen, K.; Cazaurang, F. Tilting-Rotor Quadcopter for Aggressive Flight Maneuvers Using Differential Flatness Based Flight Controller. In Proceedings of the ASME 2017 Dynamic Systems and Control Conference, DSCC 2017, Tysons, VA, USA, 11–13 October 2017; Volume 3.
18. Saif, A.-W.A. Feedback Linearization Control of Quadrotor with Tiltable Rotors under Wind Gusts. *Int. J. Adv. Appl. Sci.* **2017**, *4*, 150–159. https://doi.org/10.21833/ijaas.2017.010.021.
19. Offermann, A.; Castillo, P.; De Miras, J.D. Control of a PVTOL* with Tilting Rotors*. In Proceedings of the 2019 International Conference on Unmanned Aircraft Systems, ICUAS 2019, Atlanta, GA, USA, 11–14 June 2019.
20. Rajappa, S.; Bulthoff, H.H.; Odelga, M.; Stegagno, P. A Control Architecture for Physical Human-UAV Interaction with a Fully Actuated Hexarotor. In Proceedings of the IEEE International Conference on Intelligent Robots and Systems, Vancouver, BC, Canada, 24–28 September 2017; Volume 2017.
21. Scholz, G.; Trommer, G.F. Model Based Control of a Quadrotor with Tiltable Rotors. *Gyroscopy Navig.* **2016**, *7*, 72–81. https://doi.org/10.1134/S2075108716010120.
22. Ryll, M.; Bülthoff, H.H.; Giordano, P.R. A Novel Overactuated Quadrotor Unmanned Aerial Vehicle: Modeling, Control, and Experimental Validation. *IEEE Trans. Control Syst. Technol.* **2015**, *23*, 540–556. https://doi.org/10.1109/TCST.2014.2330999.
23. Elfeky, M.; Elshafei, M.; Saif, A.W.A.; Al-Malki, M.F. Quadrotor Helicopter with Tilting Rotors: Modeling and Simulation. In Proceedings of the 2013 World Congress on Computer and Information Technology, WCCIT 2013, Sousse, Tunisia, 22–24 June 2013.

24. Park, S.; Lee, J.; Ahn, J.; Kim, M.; Her, J.; Yang, G.H.; Lee, D. ODAR: Aerial Manipulation Platform Enabling Omnidirectional Wrench Generation. *IEEE/ASME Trans. Mechatron.* **2018**, *23*, 1907–1918. https://doi.org/10.1109/TMECH.2018.2848255.
25. Magariyama, T.; Abiko, S. Seamless 90-Degree Attitude Transition Flight of a Quad Tilt-Rotor UAV under Full Position Control. In Proceedings of the IEEE/ASME International Conference on Advanced Intelligent Mechatronics, AIM, Boston, MA, USA, 6–9 July 2020; Volume 2020.
26. Falanga, D.; Kleber, K.; Mintchev, S.; Floreano, D.; Scaramuzza, D. The Foldable Drone: A Morphing Quadrotor That Can Squeeze and Fly. *IEEE Robot. Autom. Lett.* **2019**, *4*, 209–216. https://doi.org/10.1109/LRA.2018.2885575.
27. Lu, D.; Xiong, C.; Zeng, Z.; Lian, L. Adaptive Dynamic Surface Control for a Hybrid Aerial Underwater Vehicle with Parametric Dynamics and Uncertainties. *IEEE J. Ocean. Eng.* **2020**, *45*, 740–758. https://doi.org/10.1109/JOE.2019.2903742.
28. Antonelli, G.; Cataldi, E.; Arrichiello, F.; Giordano, P.R.; Chiaverini, S.; Franchi, A. Adaptive Trajectory Tracking for Quadrotor MAVs in Presence of Parameter Uncertainties and External Disturbances. *IEEE Trans. Control Syst. Technol.* **2018**, *26*, 248–254. https://doi.org/10.1109/TCST.2017.2650679.
29. Lee, D.; Jin Kim, H.; Sastry, S. Feedback Linearization vs. Adaptive Sliding Mode Control for a Quadrotor Helicopter. *Int. J. Control Autom. Syst.* **2009**, *7*, 419–428. https://doi.org/10.1007/s12555-009-0311-8.
30. Al-Hiddabi, S.A. Quadrotor Control Using Feedback Linearization with Dynamic Extension. In Proceedings of the 2009 6th International Symposium on Mechatronics and its Applications, Sharjah, United Arab Emirates, 23–26 March 2009; pp. 1–3.
31. Mukherjee, P.; Waslander, S. Direct Adaptive Feedback Linearization for Quadrotor Control. In Proceedings of the AIAA Guidance, Navigation, and Control Conference, Minneapolis, MN, USA, 13–16 August 2012.
32. Rajappa, S.; Ryll, M.; Bulthoff, H.H.; Franchi, A. Modeling, Control and Design Optimization for a Fully-Actuated Hexarotor Aerial Vehicle with Tilted Propellers. In Proceedings of the IEEE International Conference on Robotics and Automation, Seattle, WA, USA, 26–30 May 2015; Volume 2015.
33. Dunham, W.; Petersen, C.; Kolmanovsky, I. Constrained Control for Soft Landing on an Asteroid with Gravity Model Uncertainty. In Proceedings of the Proceedings of the American Control Conference, Boston, MA, USA, 6–8 July 2016; Volume 2016.
34. McDonough, K.; Kolmanovsky, I. Controller State and Reference Governors for Discrete-Time Linear Systems with Pointwise-in-Time State and Control Constraints. In Proceedings of the American Control Conference, Chicago, IL, USA, 1–3 July 2015; Volume 2015.
35. Kolmanovsky, I.; Kalabić, U.; Gilbert, E. Developments in Constrained Control Using Reference Governors. *IFAC Proc.* **2012**, *4*, 282–290.
36. Bemporad, A. Reference Governor for Constrained Nonlinear Systems. *IEEE Trans. Automat. Contr.* **1998**, *43*, 415–419. https://doi.org/10.1109/9.661611.
37. Shen, Z.; Ma, Y.; Tsuchiya, T. Stability Analysis of a Feedback-Linearization-Based Controller with Saturation: A Tilt Vehicle with the Penguin-Inspired Gait Plan. *arXiv* **2021**, arXiv:2111.14456.
38. Franchi, A.; Carli, R.; Bicego, D.; Ryll, M. Full-Pose Tracking Control for Aerial Robotic Systems with Laterally Bounded Input Force. *IEEE Trans. Robot.* **2018**, *34*, 534–541. https://doi.org/10.1109/TRO.2017.2786734.
39. Horla, D.; Hamandi, M.; Giernacki, W.; Franchi, A. Optimal Tuning of the Lateral-Dynamics Parameters for Aerial Vehicles with Bounded Lateral Force. *IEEE Robot. Autom. Lett.* **2021**, *6*, 3949–3955. https://doi.org/10.1109/LRA.2021.3067229.
40. Shen, Z.; Tsuchiya, T. State Drift and Gait Plan in Feedback Linearization Control of A Tilt Vehicle. *arXiv* **2021**, arXiv: 2111.04307.
41. Chernova, S.; Veloso, M. An Evolutionary Approach to Gait Learning for Four-Legged Robots. In Proceedings of the 2004 IEEE/RSJ International Conference on Intelligent Robots and Systems (IROS), Sendai, Japan, 28 September–2 October 2004; Volume 3.
42. Bennani, M.; Giri, F. Dynamic Modelling of a Four-Legged Robot. *J. Intell. Robot. Syst. Theory Appl.* **1996**, *17*, 419–428. https://doi.org/10.1007/BF00571701.
43. Talebi, S.; Poulakakis, I.; Papadopoulos, E.; Buehler, M. Quadruped Robot Running with a Bounding Gait. In *Experimental Robotics VII.*; Rus, D., Singh, S., Eds.; Springer: Berlin/Heidelberg, Germany, 2007.
44. Lewis, M.A.; Bekey, G.A. Gait Adaptation in a Quadruped Robot. *Auton. Robot.* **2002**, *12*, 301–312.
45. Hirose, S. A Study of Design and Control of a Quadruped Walking Vehicle. *Int. J. Robot. Res.* **1984**, *3*, 113–133. https://doi.org/10.1177/027836498400300210.
46. Luukkonen, T. Modelling and Control of Quadcopter. *Indep. Res. Proj. Appl. Math.* **2011**, *22*, 22.
47. Goodarzi, F.A.; Lee, D.; Lee, T. Geometric Adaptive Tracking Control of a Quadrotor Unmanned Aerial Vehicle on SE(3) for Agile Maneuvers. *J. Dyn. Syst. Measur. Control* **2015**, *137*, 091007. https://doi.org/10.1115/1.4030419.
48. Shi, X.-N.; Zhang, Y.-A.; Zhou, D. A Geometric Approach for Quadrotor Trajectory Tracking Control. *Int. J. Control* **2015**, *88*, 2217–2227. https://doi.org/10.1080/00207179.2015.1039593.

49. Lee, T.; Leok, M.; McClamroch, N.H. Nonlinear Robust Tracking Control of a Quadrotor UAV on SE(3): Nonlinear Robust Tracking Control of a Quadrotor UAV. *Asian J. Control* **2013**, *15*, 391–408. https://doi.org/10.1002/asjc.567.
50. Shen, Z.; Tsuchiya, T. Singular Zone in Quadrotor Yaw-Position Feedback Linearization. *arXiv* **2021**, arXiv:2110.07179.
51. Mistier, V.; Benallegue, A.; M'Sirdi, N.K. Exact Linearization and Noninteracting Control of a 4 Rotors Helicopter via Dynamic Feedback. In Proceedings of the IEEE International Workshop on Robot and Human Interactive Communication, Paris, France, September 2001; pp. 586–593.
52. Das, A.; Subbarao, K.; Lewis, F. Dynamic Inversion with Zero-Dynamics Stabilisation for Quadrotor Control. *IET Control Theory Appl.* **2009**, *3*, 303–314. https://doi.org/10.1049/iet-cta:20080002.

# Chapter 4

# Modified Coupling Relationship between Attitude and Position (Feedback Linearization-Based Tracking Control of a Tilt-Rotor with Cat-Trot Gait Plan)

**This Chapter has been published on Feedback Linearization-Based Tracking Control of a Tilt-Rotor with Cat-Trot Gait Plan. *International Journal of Advanced Robotic Systems* 2022, *19*, 17298806221109360, doi:10.1177/17298806221109360 as the first author. The rest co-author(s) contribute nothing other than supervising and/or minor edition.**

**Abstract:** With the introduction of the laterally bounded forces, the tilt-rotor gains more flexibility in the controller design. Typical feedback linearization methods utilize all the inputs in controlling this vehicle; the magnitudes as well as the directions of the thrusts are maneuvered simultaneously based on a unified control rule. Although several promising results indicate that these controllers may track the desired complicated trajectories, the tilting angles are required to change relatively fast or in large scale during the flight, which turns to be a challenge in application. The recent gait plan for a tilt-rotor may solve this problem; the tilting angles are fixed or vary in a predetermined pattern without being maneuvered by the control algorithm. Carefully avoiding the singular decoupling matrix, several attitudes can be tracked without changing the tilting angles frequently. While the position was not directly regulated in that research, which left the position-tracking still an open question. In this research, we elucidate the coupling relationship between the position and the attitude. Based on this, we design the position-tracking controller, adopting feedback linearization. A cat-trot gait is further designed for a tilt-rotor to track the reference; three types of references are designed for our tracking experiments: setpoint, uniform rectilinear motion, and uniform circular motion. The significant improvement with less steady state error is witnessed after equipping with our modified attitude-position decoupler. It is also found that the frequency of the cat-trot gait highly influenced the steady state error.

## 1. Introduction

Comparing with the conventional quadrotor [1–4], the tilt-rotors [5–9] provide the lateral forces, which are not applicable to the collinear/coplanar platforms (e.g., conventional quadrotors). The additional mechanical structures (usually tilting motors) mounted on the arms of the tilt-rotor provide the possibility of changing the direction of each thrust or 'tilting'. As a consequence, the number of inputs increases to eight (four magnitudes of the thrust and four directions of the thrusts).

One of the systematic methods to solve tracking problem for a conventional quadrotor is feedback linearization (dynamic inversion), which transfers the nonlinear system to a linear one applicable to implement the linear controllers. Hereon, each output of interest is manipulated individually. Since the number of the inputs in conventional quadrotors (four) is less than the number of degrees of freedom (six), a controller can independently stabilize four outputs at most. Typical choices of these four outputs can be attitude-altitude [10–14] and position-yaw [6,15–18].

On the contrary, since the number of inputs in Ryll's tilt-rotors (eight) is larger than the number of degrees of freedom, the vehicle becomes an over-actuated system. This property intrigues the research on fully tracking the entire degrees of freedom; the tilt-rotor [7] not only tracks the desired time-specified position but also the desired attitude relying on the eight inputs [19,20]. Although the promising results yield acceptable tracking result, the tilting angles vary greatly or over-rapidly, which sharpens the feasibility of implementing the relevant controllers.

To solve this problem, our previous research [21] plans the gait of the tilt-rotor before applying feedback linearization. The tilting angles are assigned beforehand, rather than manipulated by a unified control rule. Hereby, the magnitudes of the thrusts are the only inputs.

However, several challenges may hinder the application of this method. One of them is the singular decoupling matrix [22] of feedback linearization. Considerable attention has been paid to this issue for the conventional quadrotor [10,23]. It has been proved that the decoupling matrix is always invertible for a tilt-rotor with an over-actuated control scenario [7]. However, this matrix may be singular if we take the magnitudes of the thrusts as the only inputs after the gait plan for a tilt-rotor. Further discussions on avoiding the singular decoupling matrix can refer to our previous work [21].

On the other hand, though the previous research [21] witnessed promising result in tracking attitude and altitude, the position $(X, Y)$ is not successfully stabilized. One may notice that the relationship between the acceleration $(\ddot{X}, \ddot{Y})$ and the attitude has been elucidated by an attitude-position decoupler in a conventional quadrotor [12,14,20,24–26]. While this decoupler does not hold for a tilt-rotor. A modified attitude-position decoupler applicable to a tilt-rotor is deduced in this article.

Similar to the gait plan problem in four-legged robots [27–29], we [21] proposed several gaits for a tilt-rotor, averting the singular decoupling matrix. However, the gait in that research [21] did not consider the gait patterns with varying tilting angles; all the gaits analyzed were the combinations of the fixed tilting angles. Another contribution of this research is to adopt the time-specified varying gait, which is inspired by cat trot [30–32].

Three types of references are designed to track for a tilt-rotor in simulation (Simulink). Notable improvements in tracking result are witnessed after the application with our modified attitude-position decoupler.

The rest of the article is structured as follows: Section 2 briefs the dynamics of the tilt-rotor. Section 3 explores the relationship coupled in attitude and position in a tilt-rotor before proposing the modified attitude-position decoupler, which is further used to design a controller. Section 4 introduces a gait for a tilt-rotor which is inspired by a cat-trot gait. The discussions upon the stability of the relevant controller are addressed in Section 5. Section 6 introduces the settings of the simulation tests, the results of which are displayed in Section 7. Finally, we make conclusions and discussions in Section 8.

## 2. Dynamics of the Tilt-Rotor

Figure 1 [21] sketches Ryll's tilt-rotor. This tilt-rotor [7,21] can adjust the direction of each thrust during the flight; tilting each arm changes the direction of the thrust, which is restricted in the relevant yellow plane in Figure 1. $\alpha_i$ ($i$=1,2,3,4) represents the tilting angles.

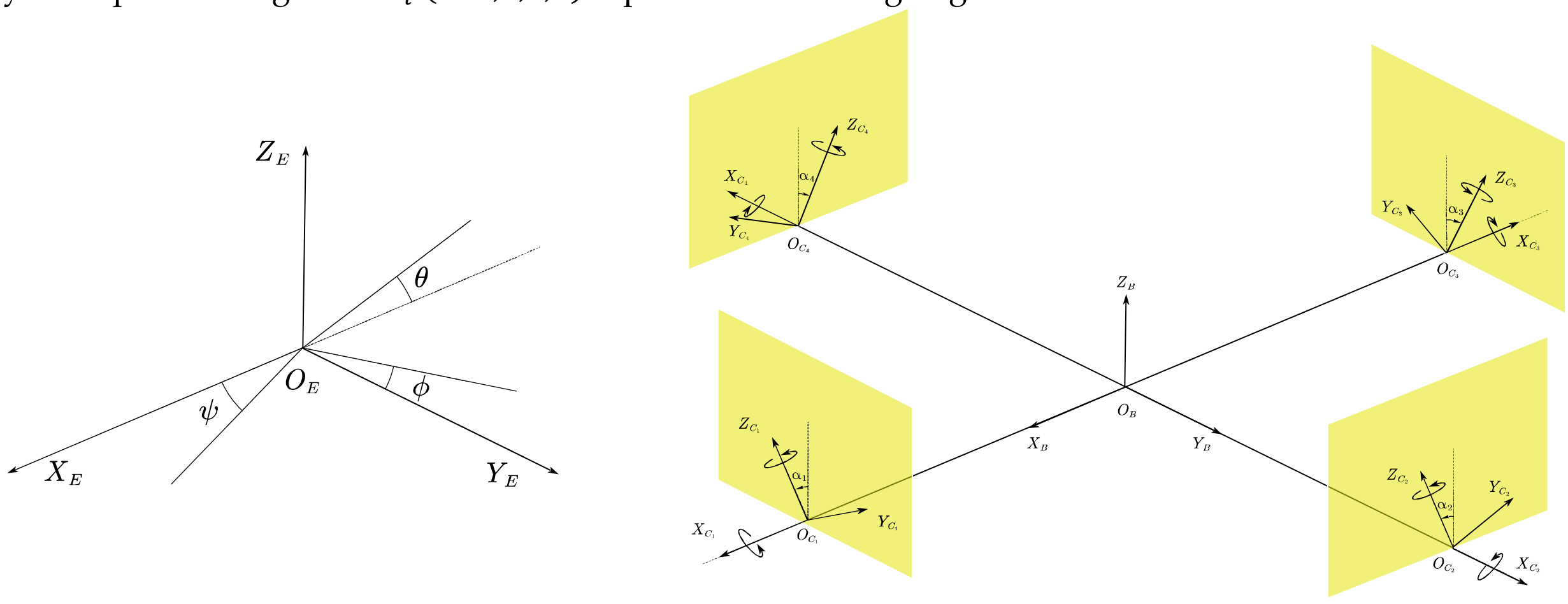

**Figure 1.** The sketch of the tilt-rotor.

The position $P = [X \quad Y \quad Z]^T$ is ruled by

$$\ddot{P} = \begin{bmatrix} 0 \\ 0 \\ -g \end{bmatrix} + \frac{1}{m} \cdot {}^{W}R \cdot F(\alpha) \cdot \begin{bmatrix} \varpi_1 \cdot |\varpi_1| \\ \varpi_2 \cdot |\varpi_2| \\ \varpi_3 \cdot |\varpi_3| \\ \varpi_4 \cdot |\varpi_4| \end{bmatrix} \triangleq \begin{bmatrix} 0 \\ 0 \\ -g \end{bmatrix} + \frac{1}{m} \cdot {}^{W}R \cdot F(\alpha) \cdot w \quad (1)$$

where $m$ is the total mass, $g$ is the gravitational acceleration, $\varpi_i$, $(i = 1,2,3,4)$ is the angular velocity of the propeller ($\varpi_{1,3} < 0$, $\varpi_{2,4} > 0$) with respect to the propeller-fixed frame, ${}^{W}R$ is the rotational matrix [4] from the inertial frame, $O_E X_E Y_E Z_E$, to the body-fixed frame, $O_B X_B Y_B Z_B$,

$${}^{W}R = \begin{bmatrix} c\theta \cdot c\psi & s\phi \cdot s\theta \cdot c\psi - c\phi \cdot s\psi & c\phi \cdot s\theta \cdot c\psi + s\phi \cdot s\psi \\ c\theta \cdot s\psi & s\phi \cdot s\theta \cdot s\psi + c\phi \cdot c\psi & c\phi \cdot s\theta \cdot s\psi - s\phi \cdot c\psi \\ -s\theta & s\phi \cdot c\theta & c\phi \cdot c\theta \end{bmatrix} \quad (2)$$

where $s\Lambda = \sin(\Lambda)$, $c\Lambda = \cos(\Lambda)$, $\phi$, $\theta$, and $\psi$ are roll angle, pitch angle, and yaw angle, respectively (see Figure 1), tilting angles $\alpha = [\alpha_1 \quad \alpha_2 \quad \alpha_3 \quad \alpha_4]$, the positive directions of which are defined in Figure 1, $F(\alpha)$ is defined as

$$F(\alpha) = \begin{bmatrix} 0 & K_f \cdot s2 & 0 & -K_f \cdot s4 \\ K_f \cdot s1 & 0 & -K_f \cdot s3 & 0 \\ -K_f \cdot c1 & K_f \cdot c2 & -K_f \cdot c3 & K_f \cdot c4 \end{bmatrix} \quad (3)$$

where $si = \sin(\alpha_i)$, $ci = \cos(\alpha_i)$, $(i = 1,2,3,4)$. $K_f$ ($8.048 \times 10^{-6} N \cdot s^2/rad^2$) is the coefficient of the thrust. $F(\alpha)$ transfers the angular velocity vector $w$ to the thrusts expressed in the body-fixed frame.

The angular velocity of the body with respect to its own frame, $\omega_B = [p \quad q \quad r]^T$, is governed by

$$\dot{\omega}_B = {I_B}^{-1} \cdot \tau(\alpha) \cdot w \quad (4)$$

where $I_B$ is the matrix of moments of inertia, $K_m$ ($2.423 \times 10^{-7} N \cdot m \cdot s^2/rad^2$) is the coefficient of the drag moment, $L$ is the length of the arm, $\tau(\alpha)$ is defined by

$$\tau(\alpha) = \begin{bmatrix} 0 & L \cdot K_f \cdot c2 - K_m \cdot s2 & 0 & -L \cdot K_f \cdot c4 + K_m \cdot s4 \\ L \cdot K_f \cdot c1 + K_m \cdot s1 & 0 & -L \cdot K_f \cdot c3 - K_m \cdot s3 & 0 \\ L \cdot K_f \cdot s1 - K_m \cdot c1 & -L \cdot K_f \cdot s2 - K_m \cdot c2 & L \cdot K_f \cdot s3 - K_m \cdot c3 & -L \cdot K_f \cdot s4 - K_m \cdot c4 \end{bmatrix}. \quad (5)$$

The relationship between the angular velocity of the body, $\omega_B$, and the attitude rotation matrix (${}^{W}R$) is given [33–35] by

$${}^{W}\dot{R} = {}^{W}R \cdot \widehat{\omega}_B \quad (6)$$

where "$\widehat{\ }$" is the hat operation to produce the skew matrix.

We refer the readers to [36,37] for the detail in modeling the coefficients of the thrust, drag moment, and inertia moment, $I_B$. They are assumed to be constant in this research.

The parameters of this tilt-rotor are specified as follows: $m = 0.429kg$, $L = 0.1785m$, $g = 9.8N/kg$, $I_B = \mathrm{diag}([2.24{\times}10^{-3}, 2.99{\times}10^{-3}, 4.80{\times}10^{-3}]) kg \cdot m^2$.

## 3. Controller Design and Gait Plan

The controller for the tilt-rotor comprises three parts: modified attitude-position decoupler, feedback linearization, and high-order PD controller.

### *3.1. Scenario of The Controller*

As mentioned, this controller can only control four degrees of freedom independently at most. Noticing that picking position and yaw may introduce the singular decoupling matrix [10,23], the independently controlled variables decided in this research are attitude and altitude [11,12,20,25].

The rest of the degrees of freedom (position ($X$,$Y$)) are tracked by adjusting the attitude, relying on the coupled relationship between the position and attitude (Section 3.2). This nested structure can be found in Figure 2 (green part).

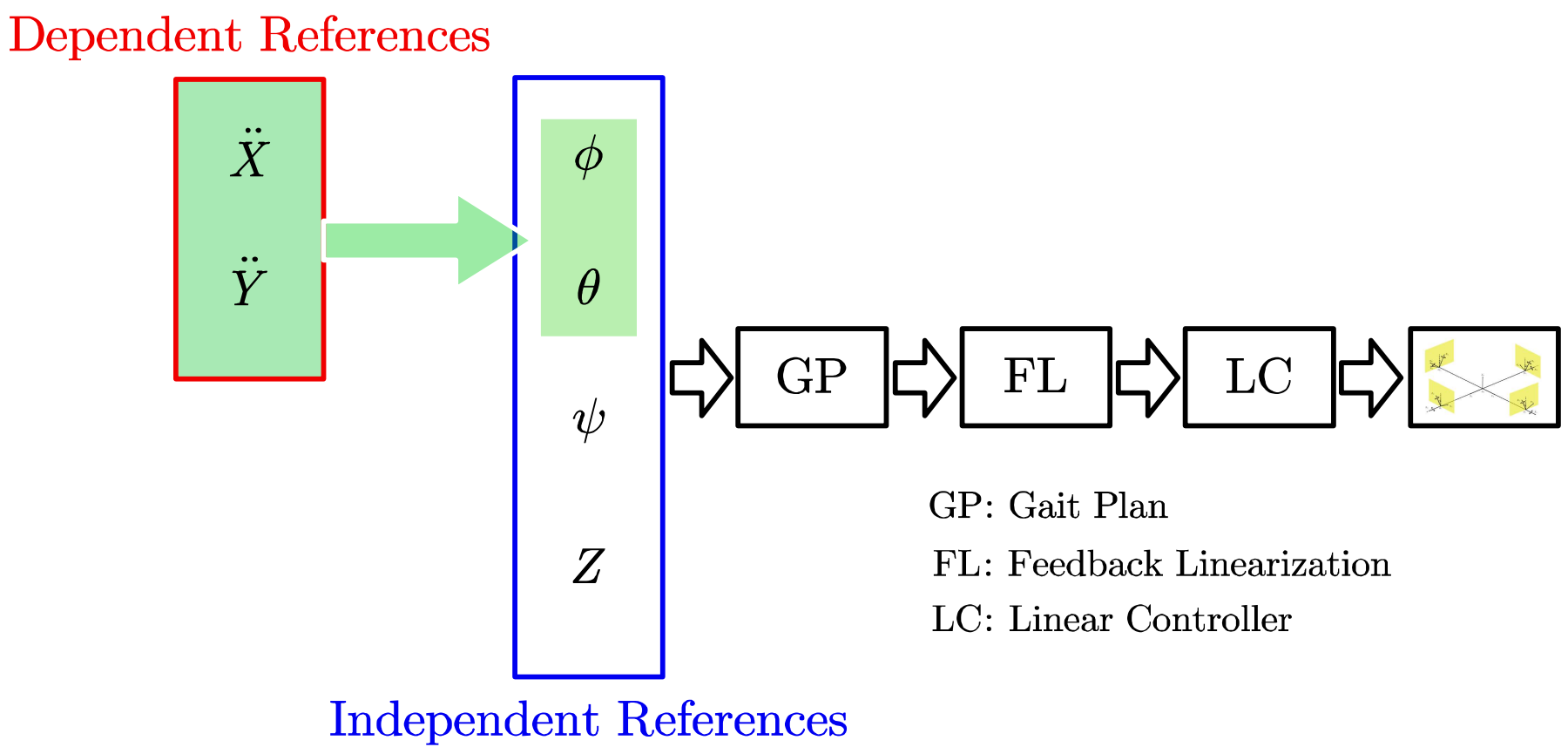

**Figure 2.** The Control diagram.

After dealing with the coupling relationship in position and attitude, an animal-inspired gait is designed for the tilt-rotor, which is detailed in Section 4.

Finally, feedback linearization (Section 3.3) is applied to accommodate a linear controller (Section 3.4).

*3.2. Modified Attitude-position Decoupler*

The relationship (conventional attitude-position decoupler) between attitude and position in a conventional quadrotor [10–14,17,20,25,26,38–40] is

$$\phi = \frac{1}{g} \cdot \left(\ddot{X} \cdot s\psi - \ddot{Y} \cdot c\psi\right), \quad (7)$$

$$\theta = \frac{1}{g} \cdot \left(\ddot{X} \cdot c\psi + \ddot{Y} \cdot s\psi\right). \quad (8)$$

This relationship is deduced by linearizing the dynamics of a conventional quadrotor near hovering. However, this attitude-position decoupler does not hold for the tilt-rotor if the tilting angles are not all zero. Proposition 1 gives the modified attitude-position decoupler for a tilt-rotor.

**Proposition 1. (Modified Attitude-position Decoupler)** *The attitude and the position of a tilt-rotor can be approximately decoupled as*

$$\phi = \frac{1}{g} \cdot \left(\ddot{X} \cdot s\psi - \ddot{Y} \cdot c\psi\right) + \frac{F_Y}{mg}, \quad (9)$$

$$\theta = \frac{1}{g} \cdot \left(\ddot{X} \cdot c\psi + \ddot{Y} \cdot s\psi\right) - \frac{F_X}{mg}, \quad (10)$$

where $F_X$ and $F_Y$ are defined by

$$\begin{bmatrix} F_X \\ F_Y \end{bmatrix} = K_f \cdot \begin{bmatrix} 0 & s2 & 0 & -s4 \\ s1 & 0 & -s3 & 0 \end{bmatrix} \cdot \begin{bmatrix} I_B{}^{-1} \cdot \tau(\alpha) \\ \frac{K_f}{m} \cdot [0 \quad 0 \quad 1] \cdot F(\alpha) \end{bmatrix}^{-1} \cdot \begin{bmatrix} 0 \\ 0 \\ 0 \\ g \end{bmatrix}. \tag{11}$$

**Proof.** Linearize Equation (1) at the equilibrium states,

$$\dot{\omega}_B = I_B{}^{-1} \cdot \tau(\alpha) \cdot w = 0, \tag{12}$$

$$\ddot{Z} = 0, \tag{13}$$

$$\phi, \theta = 0. \tag{14}$$

Ignoring the high-order infinitesimal terms (e.g., $\phi \cdot \theta, \phi^2, \theta^2$) yields Equation (11).

Formula (12) – (13) guarantee the angular accelerations and the vertical acceleration of the tilt-rotor are zero, given the assumption of the attitudes in Formula (14). The components of the thrust along $X$ and $Y$, with respect to the body-fixed frame, are then calculated in Formula (11) as $F_X$ and $F_Y$.

**Remark 2.** Given $\alpha_1 = \alpha_2 = \alpha_3 = \alpha_4 = 0$, we receive $F_X = F_Y = 0$. Subsequently, Equation (9), (10) degrade to Equation (7), (8) in this special case. In other words, the attitude-position decoupler for the conventional quadrotor is a special case of the modified attitude-position decoupler for the tilt-rotor.

*3.3. Feedback Linearization*

The degrees of freedoms for the controller to track independently are attitude and altitude. Define

$$\begin{bmatrix} y_1 \\ y_2 \\ y_3 \\ y_4 \end{bmatrix} = \begin{bmatrix} \phi \\ \theta \\ \psi \\ Z \end{bmatrix}. \tag{15}$$

Since $\varpi_{1,3} < 0, \varpi_{2,4} > 0$, we have

$$(\varpi_i \cdot |\varpi_i|)' = 2 \cdot \dot{\varpi}_i \cdot |\varpi_i|. \tag{16}$$

Assuming

$$\dot{\alpha}_i \equiv 0, i = 1,2,3,4, \tag{17}$$

calculating the third derivative of Equation (18) [21] yields

$$\begin{bmatrix} \dddot{y}_1 \\ \dddot{y}_2 \\ \dddot{y}_3 \\ \dddot{y}_4 \end{bmatrix} = \begin{bmatrix} I_B{}^{-1} \cdot \tau(\alpha) \\ [0 \quad 0 \quad 1] \cdot \frac{K_f}{m} \cdot {}^W R \cdot F(\alpha) \cdot 2 \cdot \begin{bmatrix} |\varpi_1| & & & \\ & |\varpi_2| & & \\ & & |\varpi_3| & \\ & & & |\varpi_4| \end{bmatrix} \end{bmatrix}^{4\times4} \cdot \begin{bmatrix} \dot{\varpi}_1 \\ \dot{\varpi}_2 \\ \dot{\varpi}_3 \\ \dot{\varpi}_4 \end{bmatrix} + [0 \quad 0 \quad 1] \cdot \frac{K_f}{m} \cdot {}^W R \cdot \widehat{\omega}_B \cdot F(\alpha) \cdot w \cdot \begin{bmatrix} 0 \\ 0 \\ 0 \\ 1 \end{bmatrix} \triangleq \bar{\Delta} \cdot \begin{bmatrix} \dot{\varpi}_1 \\ \dot{\varpi}_2 \\ \dot{\varpi}_3 \\ \dot{\varpi}_4 \end{bmatrix} + Ma \tag{18}$$

where $\bar{\Delta}$ is called decoupling matrix [22]. $[\dot{\varpi}_1 \quad \dot{\varpi}_2 \quad \dot{\varpi}_3 \quad \dot{\varpi}_4]^T \triangleq U$ is the updated input vector.

Observing Equation (18), one receives the decoupled relationship in Equation (19), compatible with the controller design process, which will be detailed in Section 3.4.

$$\begin{bmatrix} \dot{\varpi}_1 \\ \dot{\varpi}_2 \\ \dot{\varpi}_3 \\ \dot{\varpi}_4 \end{bmatrix} = \bar{\Delta}^{-1} \cdot \left( \begin{bmatrix} \dddot{y}_{1d} \\ \dddot{y}_{2d} \\ \dddot{y}_{3d} \\ \dddot{y}_{4d} \end{bmatrix} - Ma \right). \tag{19}$$

Obviously, it should be guaranteed that the decoupling matrix ($\bar{\Delta}$) in Equation (19) is invertible. Our previous work [21] details the equivalent requirements for receiving an invertible decoupling matrix in Proposition 2.

**Proposition 2.** *When the roll angle and pitch angle of the tilt-rotor are close to zero, the decoupling matrix is invertible if and only if Condition (20) holds.*

$$4.000 \cdot c1 \cdot c2 \cdot c3 \cdot c4 + 5.592 \cdot (+c1{\cdot}c2{\cdot}c3{\cdot}s4 - c1{\cdot}c2{\cdot}s3{\cdot}c4 + c1{\cdot}s2{\cdot}c3{\cdot}c4 - s1{\cdot}c2{\cdot}c3{\cdot}c4) + 0.9716 \cdot (+c1{\cdot}c2{\cdot}s3{\cdot}s4 + c1{\cdot}s2{\cdot}s3{\cdot}c4 + s1{\cdot}c2{\cdot}c3{\cdot}s4 + s1{\cdot}s2{\cdot}c3{\cdot}c4) + 2.000 \cdot (-c1{\cdot}s2{\cdot}c3{\cdot}s4 - s1{\cdot}c2{\cdot}s3{\cdot}c4) + 0.1687 \cdot (-c1{\cdot}s2{\cdot}s3{\cdot}s4 + s1{\cdot}c2{\cdot}s3{\cdot}s4 - s1{\cdot}s2{\cdot}c3{\cdot}s4 + s1{\cdot}s2{\cdot}s3{\cdot}c4) \neq 0, \tag{20}$$

where $s\Pi = \sin\alpha_\Pi$, $c\Pi = \cos\alpha_\Pi$, $(\Pi = \overline{1,4})$, $s\Lambda = \sin\Lambda$, $c\Lambda = \cos\Lambda$, $(\Lambda = \phi, \theta)$.

**Proof.** See our previous research [21].

In this research, we adopt the cat-trot gait defined by

$$\alpha_1 = -\rho, \tag{21}$$

$$\alpha_2 = -\rho, \tag{22}$$

$$\alpha_3 = \rho, \tag{23}$$

$$\alpha_4 = \rho, \tag{24}$$

where $\rho$ is a time-specified gait, which will be specified in Section 4.

**Proposition 3.** *When the roll angle and pitch angle of the tilt-rotor are close to zero, the cat-trot gait introduces no singular problem.*

**Proof.** Substituting (21) – (24) into the left side of (20) yields

$$4\cos^2(\rho). \tag{25}$$

Since $\rho$ lies in $[-0.65, 0.65]$, (25) is non-zero, satisfying Condition (20).

*3.4. Third-order PD Controller*

Design the third-order PD controllers as

$$\begin{bmatrix} \dddot{y}_{1d} \\ \dddot{y}_{2d} \\ \dddot{y}_{3d} \end{bmatrix} = \begin{bmatrix} \dddot{y}_{1r} \\ \dddot{y}_{2r} \\ \dddot{y}_{3r} \end{bmatrix} + K_{P1}{}^{3\times3} \cdot \left( \begin{bmatrix} \ddot{y}_{1r} \\ \ddot{y}_{2r} \\ \ddot{y}_{3r} \end{bmatrix} - \begin{bmatrix} \ddot{y}_1 \\ \ddot{y}_2 \\ \ddot{y}_3 \end{bmatrix} \right) + K_{P2}{}^{3\times3} \cdot \left( \begin{bmatrix} \dot{y}_{1r} \\ \dot{y}_{2r} \\ \dot{y}_{3r} \end{bmatrix} - \begin{bmatrix} \dot{y}_1 \\ \dot{y}_2 \\ \dot{y}_3 \end{bmatrix} \right) + K_{P3}{}^{3\times3} \cdot \left( \begin{bmatrix} y_{1r} \\ y_{2r} \\ y_{3r} \end{bmatrix} - \begin{bmatrix} y_1 \\ y_2 \\ y_3 \end{bmatrix} \right), \tag{26}$$

$$\ddot{y}_{4d} = \ddot{y}_{4r} + K_{PZ_1} \cdot (\ddot{y}_{4r} - \ddot{y}_4) + K_{PZ_2} \cdot (\dot{y}_{4r} - \dot{y}_4) + K_{PZ_3} \cdot (y_{4r} - y_4), \tag{27}$$

where $K_{Pi}$ $(i = 1,2,3)$ is the 3-by-3 diagonal control coefficient matrix, $K_{PZ_i}$ $(i = 1,2,3)$ is the control coefficient (scalar), $y_j$ $(j = 1,2,3,4)$ is the state, $y_{jr}$ $(j = 1,2,3,4)$ is the reference. $y_{3r}$ represents the reference of the yaw angle, which is kept zero in this research; that is $\dddot{y}_{3r} = \ddot{y}_{3r} = \dot{y}_{3r} = y_{3r} = 0$. $y_{4r}$ represents the reference of the altitude, which is also kept zero in this research; that is $\dddot{y}_{4r} = \ddot{y}_{4r} = \dot{y}_{4r} = y_{4r} = 0$. The derivatives of the references of the roll angle and pitch angle are set zero; that are $\dddot{y}_{1r} = \ddot{y}_{1r} = \dot{y}_{1r} = 0$ and $\dddot{y}_{2r} = \ddot{y}_{2r} = \dot{y}_{2r} = 0$. As for the reference of the roll angle and pitch angle, $y_{1r}$ $(\phi_r)$ and $y_{2r}$ $(\theta_r)$ are defined in Formula (28) – (30) to track the position (X and Y).

The control parameters in the above controller are specified as: $K_{P1} = \mathrm{diag}([1,1,1])$, $K_{P2} = \mathrm{diag}([10,10,1])$, $K_{P3} = \mathrm{diag}([50,50,1])$, $K_{PZ_1} = 10$, $K_{PZ_2} = 5$, $K_{PZ_3} = 10$.

As for tracking the rest degrees of freedoms ($X$ *and* $Y$), firstly design the PD controller

$$\begin{bmatrix} \ddot{X}_d \\ \ddot{Y}_d \end{bmatrix} = \begin{bmatrix} \ddot{X}_r \\ \ddot{Y}_r \end{bmatrix} + \begin{bmatrix} K_{X_1} & 0 \\ 0 & K_{Y_1} \end{bmatrix} \cdot \left( \begin{bmatrix} \dot{X}_r \\ \dot{Y}_r \end{bmatrix} - \begin{bmatrix} \dot{X} \\ \dot{Y} \end{bmatrix} \right) + \begin{bmatrix} K_{X_2} & 0 \\ 0 & K_{Y_2} \end{bmatrix} \cdot \left( \begin{bmatrix} X_r \\ Y_r \end{bmatrix} - \begin{bmatrix} X \\ Y \end{bmatrix} \right) \tag{28}$$

where $K_{X_1} = K_{Y_1} = K_{X_2} = K_{Y_2} = 1$. $X_r$, $\dot{X}_r$, $\ddot{X}_r$, $Y_r$, $\dot{Y}_r$, $\ddot{Y}_r$ are the references of the position and their higher derivatives. They are different in each simulation depending on the type of the reference and will be detailed in Section 6.

The output (left side) of this PD controller is the desired position, which will be transferred to the desired attitude by modified attitude-position decoupler,

$$\phi_r = \frac{1}{g} \cdot \left(\ddot{X}_d \cdot s\psi_r - \ddot{Y}_d \cdot c\psi_r\right) + \frac{F_Y}{mg}, \tag{29}$$

$$\theta_r = \frac{1}{g} \cdot \left(\ddot{X}_d \cdot c\psi_r + \ddot{Y}_d \cdot s\psi_r\right) - \frac{F_X}{mg}. \tag{30}$$

where $\psi_r$ is the reference of yaw angle ($y_{3r}$), which is kept zero in this research.

Notice that the attitude-altitude controller in (26) is designed much faster than the position tracking controller in (28); the control coefficients, K, are designed much larger in (26) than the ones in (28).

## 4. Cat-trot-inspired Gait

This research adopts three different gaits (fixed tilting angles, cat trot with instant switch, and cat trot with continuous switch.

### *4.1. Fixed Tilting Angles*

The gait with the fixed tilting angles is defined by

$$\alpha_1 = \alpha_2 = -\rho, \tag{31}$$

$$\alpha_3 = \alpha_4 = \rho. \tag{32}$$

The tilting angles remains constant ($\rho\ remains\ constant$) during the entire flight. In the simulations, we compare the results with different $\rho$ ($\rho$ =0.65, $\rho$ =0.65/2, $\rho$ =0, $\rho$ =-0.65/2, $\rho$ =-0.65).

### *4.2. Cat Trot with Instant Switch*

Typical cat gaits can be categorized as walk gait [30], trot gait [31,32], and gallop gait [31,32].

Parallel to the trot gait of a cat, we plan the cat-trot inspired gait, which is specified as

$$\begin{cases} t - n \cdot T \in \left[0, \frac{T}{2}\right] : \rho = 0.65 \\ t - n \cdot T \in \left(\frac{T}{2}, T\right) : \rho = -0.65 \end{cases} \tag{33}$$

where $\rho$ determines the tilting angles in (21) – (24), $T$ is the period of the gait, $t$ represents the current time, $n$ is the floor of $t/T$. Note that the tilting angles change instantly in this gait plan. Thus, we call this cat-trot gait with instant switch.

We compare the result in the simulations with different periods, $T$.

### *4.3. Cat Trot with Continuous Switch*

In the cat-trot gait with instant switch, the tilting angles change discretely. This section provides a continuous cat-trot gait,

$$\begin{cases} t - n \cdot T \in \left[0, \frac{T}{2}\right] : \rho = 0.65 - 2 \times 0.65 \cdot \frac{2}{T} \cdot (t - n \cdot T) \\ t - n \cdot T \in \left(\frac{T}{2}, T\right) : \rho = -0.65 + 2 \times 0.65 \cdot \frac{2}{T} \cdot \left(t - n \cdot T - \frac{T}{2}\right) \end{cases} \tag{34}$$

where $\rho$ determines the tilting angles in (21) – (24), $T$ is the period of the gait, $t$ represents the current time, $n$ is the floor of $t/T$. We call this cat-trot gait with continuous switch.

Also, we compare the result in the simulations with different periods, $T$.

## 5. Stability Analysis

In this section, we address the discussion on the stability of our controller. Firstly, we provide the stability proof to our controller. Secondly, we make comments on the potential state errors.

*5.1. Stability Proof*

Noticing that the singular decoupling matrix is avoided (see Proposition 3), the original nonlinear system is stable if the linearized system is proved stable, given that the constraints (e.g., non-negative constraints of the thrusts, upper bound of the thrusts, etc.) are not activated; advanced stability analyses are demanded if the constraints are activated [41]. In this research, these constraints are not activated, which simplifies our discussion on the stability.

**Proposition 4. (Exponential Stability of Attitude and Altitude)** *The attitude and altitude are exponentially stable if*

$$\begin{bmatrix} K_{P1} & \\ & K_{PZ_1} \end{bmatrix} > 0, \tag{35}$$

$$\begin{bmatrix} K_{P2} & \\ & K_{PZ_2} \end{bmatrix} > 0, \tag{36}$$

$$\begin{bmatrix} K_{P3} & \\ & K_{PZ_3} \end{bmatrix} > 0, \tag{37}$$

$$\begin{bmatrix} K_{P1} & \\ & K_{PZ_1} \end{bmatrix} \cdot \begin{bmatrix} K_{P2} & \\ & K_{PZ_2} \end{bmatrix} - \begin{bmatrix} K_{P3} & \\ & K_{PZ_3} \end{bmatrix} > 0. \tag{38}$$

**Proof.** Applying Hurwitz Criterion or Routh Criterion to (26) and (27) yields (35) – (38).

**Proposition 5. (Exponential Stability of Position $(\boldsymbol{X},\boldsymbol{Y})$)** *The position $(X,Y)$ is exponentially stable if*

$$\begin{bmatrix} K_{X_1} & 0 \\ 0 & K_{Y_1} \end{bmatrix} > 0, \tag{39}$$

$$\begin{bmatrix} K_{X_2} & 0 \\ 0 & K_{Y_2} \end{bmatrix} > 0. \tag{40}$$

**Proof.** Applying Hurwitz Criterion or Routh Criterion to (28) yields (39) – (40).

Thus, all degrees of freedom have been proved exponentially stable.

*5.2. Remarks to The Stability Proof*

In Section 5.1, we proved that the attitude and altitude are exponentially stable applying this controller, which has been verified by simulations [21]. However, the references in that study [21] were attitude and altitude. The position $(X,Y)$ was not required to be stabilized.

Though the modified attitude-position decoupler in Formula (29) – (30) and the position tracking controller in (28) may be used to track a position, proved in Proposition 4 and 5, there can be steady state error in position using this controller.

**Remark 6. (Steady State Error in Position $(\boldsymbol{X},\boldsymbol{Y})$ for Tilt-rotors)** There can be steady state error with the application of the controllers in (26) – (30). This is because that the modified attitude-position decoupler was deduced by linearization at the state defined in (12) – (14).However, the states in (12) – (14) may not be satisfied simultaneously; generating zero angular accelerations and vertical acceleration in (12) – (13) may only be possible for non-zero roll angles and pitch angles, indicating that the conditions in (14) are violated.

As the consequence, the system may be stabilized at a state near the state for linearization rather than on it. This unprecise relationship between the position and the attitude can introduce unexpected steady state error.

Figure 3 helps demonstrate the above discussions on the source of the steady state error. We linearize (green dash line) the dynamics of a tilt-rotor at the equilibrium for a conventional quadrotor (red point), where both the roll angle and pitch angle are zero. However, the real equilibrium for a tilt-rotor is at the blue point. The bias introduced in the linearization may produce the steady state error.

**Remark 7. (Steady State Error in Position ($X$,$Y$) for Conventional Quadrotors)** There are no steady state errors in position ($X$,$Y$) for a conventional quadrotor applying PD controllers in a position-tracking problem [42,43]. The equilibrium state for linearization to decouple the conventional quadrotor's attitude and position is also at (12) – (14). While the conventional quadrotor stabilizes at this state, which is picked for linearization. Consequently, no bias is introduced in the linearization process.

Figure 3 also helps demonstrate the reason why there are no steady state error for a conventional quadrotor. We linearize (green dash line) the dynamics of a quadrotor at the equilibrium for a conventional quadrotor (red point), where both the roll angle and pitch angle are zero. On the other hand, this state (red point) is exactly the quadrotor is supposed to be stabilized at.

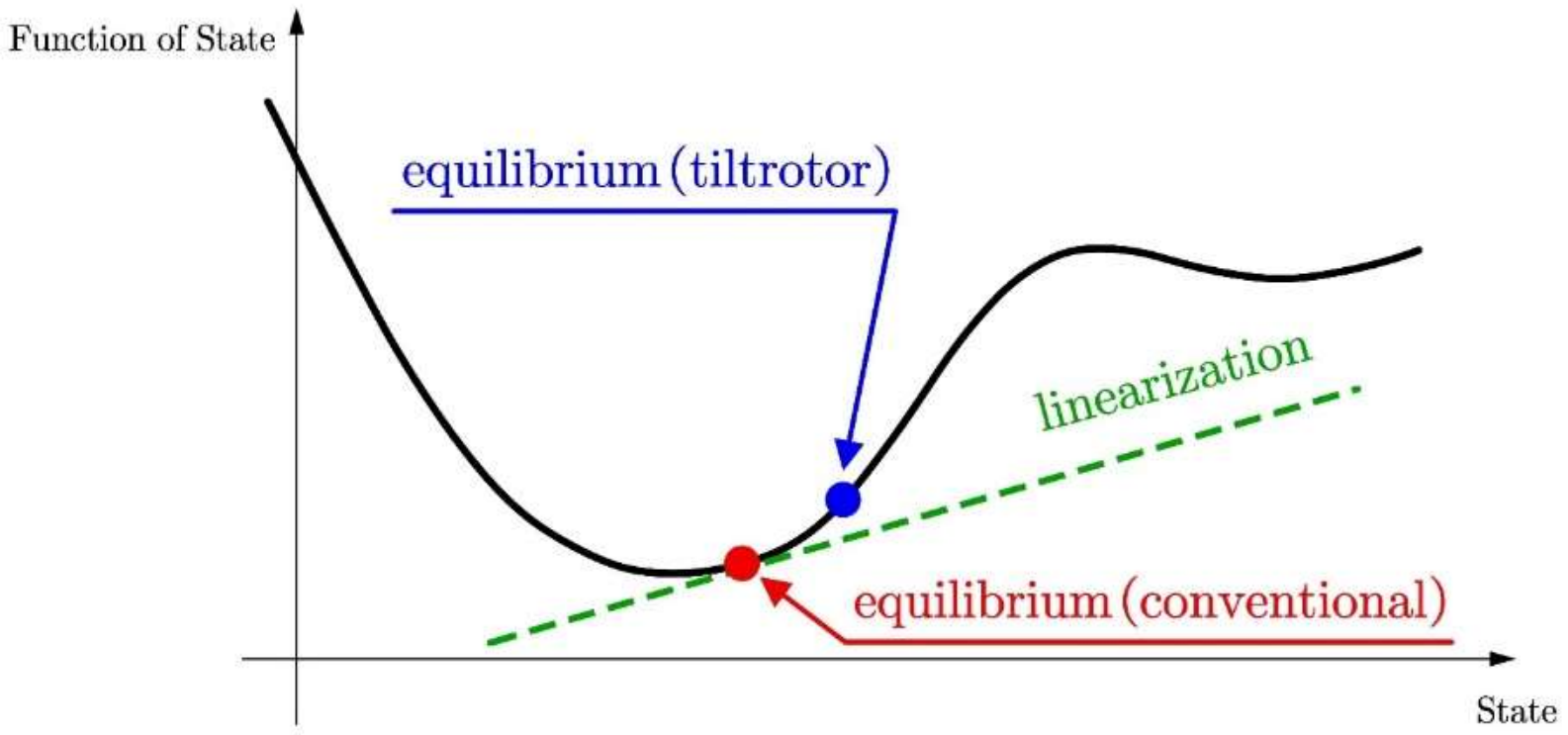

**Figure 3.** The linearization bias in the tilt-rotor.

## 6. Trajectory-tracking Experiments

The tilt-rotor is expected to track three types of trajectories in our experiments. They are setpoint, uniform rectilinear motion, and uniform circular motion.

### *6.1. Setpoint*

One may notice that our previous research [21] tracks the setpoint using the similar controller. However, the setpoint set in that work [21] is the attitude-altitude reference (e.g., ($\phi$,$\theta$,$\psi$,$Z$) = (0,0,0,0)). The position ($X$,$Y$) is not stabilized; stabilizing at ($\phi$,$\theta$,$\psi$,$Z$) = (0,0,0,0) produces accelerations along $X$ and $Y$ for the tilt-rotor.

In modified attitude-position decoupler (Proposition 1), we explicit the relationship between the position ($X$,$Y$) and attitude, providing the possibility of position-tracking for the tilt-rotor.

The setpoint reference designed in this experiment is position-yaw pair,

$$(X,Y,Z,\psi) = (0,0,0,0) \tag{41}$$

Thus, the reference settings in Formula (28) are $X_r = \dot{X}_r = \ddot{X}_r = 0$, $Y_r = \dot{Y}_r = \ddot{Y}_r = 0$.

We track this set point adopting the gaits with the different fixed tilting angles $\rho$ ($\rho$ =0.65, $\rho$ =0.65/2, $\rho$ =0, $\rho$ =-0.65/2, $\rho$ =-0.65) (Section 4.1).

As discussed in Remark 6, the steady state error is taken to evaluate the result of each experiment. We compare the resulting steady state errors between the controller adopting the conventional attitude-position decoupler in Formula (7), (8) and the controller adopting the modified attitude-position decoupler in Proposition 1 for each gait.

*6.2. Uniform Rectilinear Motion*

The thorough introductions on the cat gaits can be referred to specialized studies on cats [30–32], where cat-walk gait is typically adopted at the moving speed of $0.54 - 0.74\,m/s$ [30], cat-trot gait and cat-gallop gait are typically adopted at the moving speeds of $2.12\,m/s$ and $3.5\,m/s$, respectively [31].

Referring to this, we pair our reference in (42) with the speed around $2.12\,m/s$.

$$\begin{cases} X_r = 1.5 \cdot t \\ Y_r = 1.5 \cdot t \end{cases}. \tag{42}$$

Besides Formula (42), further reference settings in Formula (28) are $\dot{X}_r = 1.5$, $\ddot{X}_r = 0$, $\dot{Y}_r = 1.5$, $\ddot{Y}_r = 0$.

Three different kinds of the gaits are analyzed: the fixed tilting angle gaits ($\rho = 0.65$) in Section 4.1, the cat trot with instant switch with different periods in Section 4.2, the cat trot with continuous switch with different periods in Section 4.3.

Also, the steady state error is taken to evaluate the result of each experiment. We compare the resulting steady state errors between the controller adopting the conventional attitude-position decoupler in Formula (7), (8) and the controller adopting the modified attitude-position decoupler in Proposition 1 for each gait.

*6.3. Uniform Circular Motion*

A circular trajectory (uniform circular motion) is designed as

$$\begin{cases} X_r = 5 \cdot \cos(0.1 \cdot t) \\ Y_r = 5 \cdot \sin(0.1 \cdot t) \end{cases}, \tag{43}$$

$$\begin{cases} \dot{X}_r = -0.1 \times 5 \cdot \sin(0.1 \cdot t) \\ \dot{Y}_r = 0.1 \times 5 \cdot \cos(0.1 \cdot t) \end{cases}. \tag{44}$$

This indicate that the radius of the desired trajectory is 5 meters. The period is $20\pi$ seconds.

Specially, besides Formula (43) and (44), further reference settings in Formula (28) are $\ddot{X}_r = 0$ and $\ddot{Y}_r = 0$. Although the derivatives of Formula (44) are not zero, $\ddot{X}_r$ and $\ddot{Y}_r$ are set zero to avoid introducing the steady state error from high order.

Same as the previous reference, three different kinds of the gaits are analyzed in this reference: the fixed tilting angle gaits ($\rho = 0.65$) in Section 4.1, the cat trot with instant switch with different periods in Section 4.2, the cat trot with continuous switch with different periods in Section 4.3.

Similarly, the steady state error is taken to evaluate the result of each experiment. We compare the resulting steady state errors between the controller adopting the conventional attitude-position decoupler in Formula (7), (8) and the controller adopting the modified attitude-position decoupler in Proposition 1 for each gait.

*6.4. Initial Condition*

The absolute value of each initial angular velocity of the propellers is 300 rad/s in all the simulations. This angular velocity is not sufficient to compensate the effect of the gravity and will cause unstable in attitude if maintaining this speed. The tilt-rotor is expected to track the desired trajectories (Section 6.1 – 6.3) relying on the controller designed in Section 3, starting from this initial condition.

## 7. Results

This section displays the results of the tracking experiments with different references.

*7.1. Setpoint*

The setpoint are tracked by the gaits with the conventional attitude-position decoupler (marked as $B$ in Figure 4) and with the modified attitude-position decoupler (marked as $A$ in Figure 4).

The steady state errors along X and Y ($e_X$,$e_Y$) are reported in Figure 4.

$A_1(\rho{=}0.65)$, $A_2\left(\rho{=}\frac{0.65}{2}\right)$, $A_3(\rho{=}0)$, $A_4\left(\rho{=}-\frac{0.65}{2}\right)$, $A_5(\rho{=}-0.65)$ in Figure 4 represent the resulting steady state errors adopting the relevant gaits equipped with the conventional attitude-position decoupler. Similarly, $B_1(\rho{=}0.65)$, $B_2\left(\rho{=}\frac{0.65}{2}\right)$, $B_3(\rho{=}0)$, $B_4\left(\rho{=}-\frac{0.65}{2}\right)$, $B_5(\rho{=}-0.65)$ in Figure 4 represent the resulting steady state errors adopting the relevant gaits equipped with the modified attitude-position decoupler.

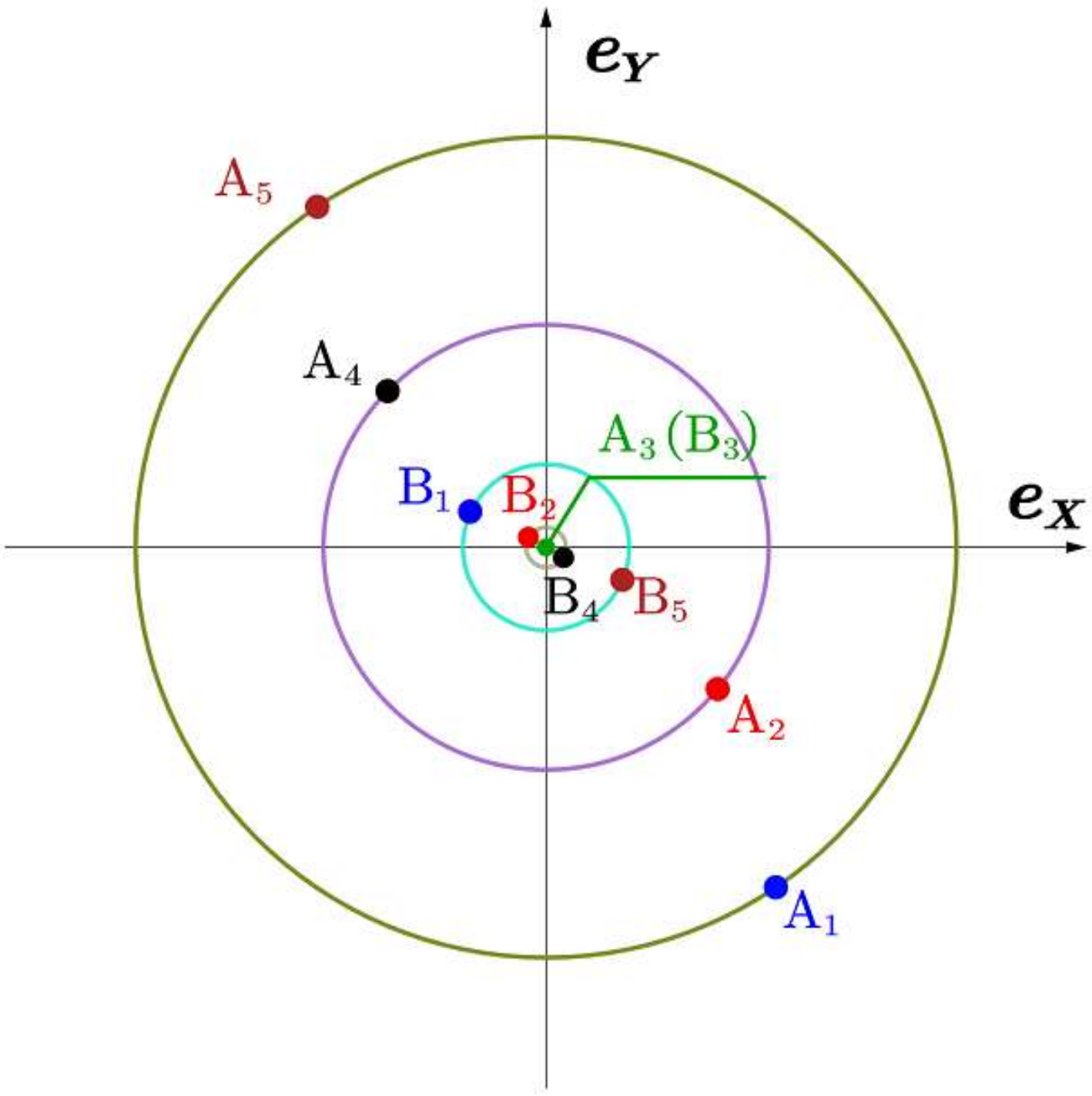

**Figure 4.** The steady state errors. $A_1(3.35,-3.56)$, $A_2(1.61,-1.64)$, $A_3(0,0)$, $A_4(-1.61,1.64)$, $A_5(-3.35,3.56)$, $B_1(-0.38,0.17)$, $B_2(-0.04,0.02)$, $B_3(0,0)$, $B_4(0.04,-0.02)$, $B_5(0.38,-0.17)$.

Notice that the gait $\rho = 0$ receives zero dynamic state error for both cases, with the conventional attitude-position decoupler (marked as $B_3$) and with the modified attitude-position decoupler (marked as $A_3$). This is because that the tilt-rotor degrades to the conventional quadrotor in this gait. For the rest gaits, the modified attitude-position decoupler notably reduces the steady state error in the same gait.

*7.2. Uniform Rectilinear Motion*

In tracking a uniform rectilinear motional reference defined in Section 6.2, Figure 5 displays the resulting dynamic state errors, $e_x = x_r - x$ and $e_y = y_r - y$, adopting the gait $\rho = 0.65$ (fixed tilting angles), equipped with the controller with the conventional attitude-position decoupler, denoted by $e_x$ (without) and $e_y$ (without), and with the controller with the modified attitude-position decoupler, denoted by $e_x$ (with) and $e_y$ (with).

It can be clearly seen that the yellow curve, $e_x$ (with), and the purple curve, $e_y$ (with), are closer to zero comparing with the blue curve, $e_x$ (without), and the red curve, $e_y$ (without), respectively, after sufficient time. This indicates that the steady state errors are reduced after equipping with the modified attitude-position decoupler.

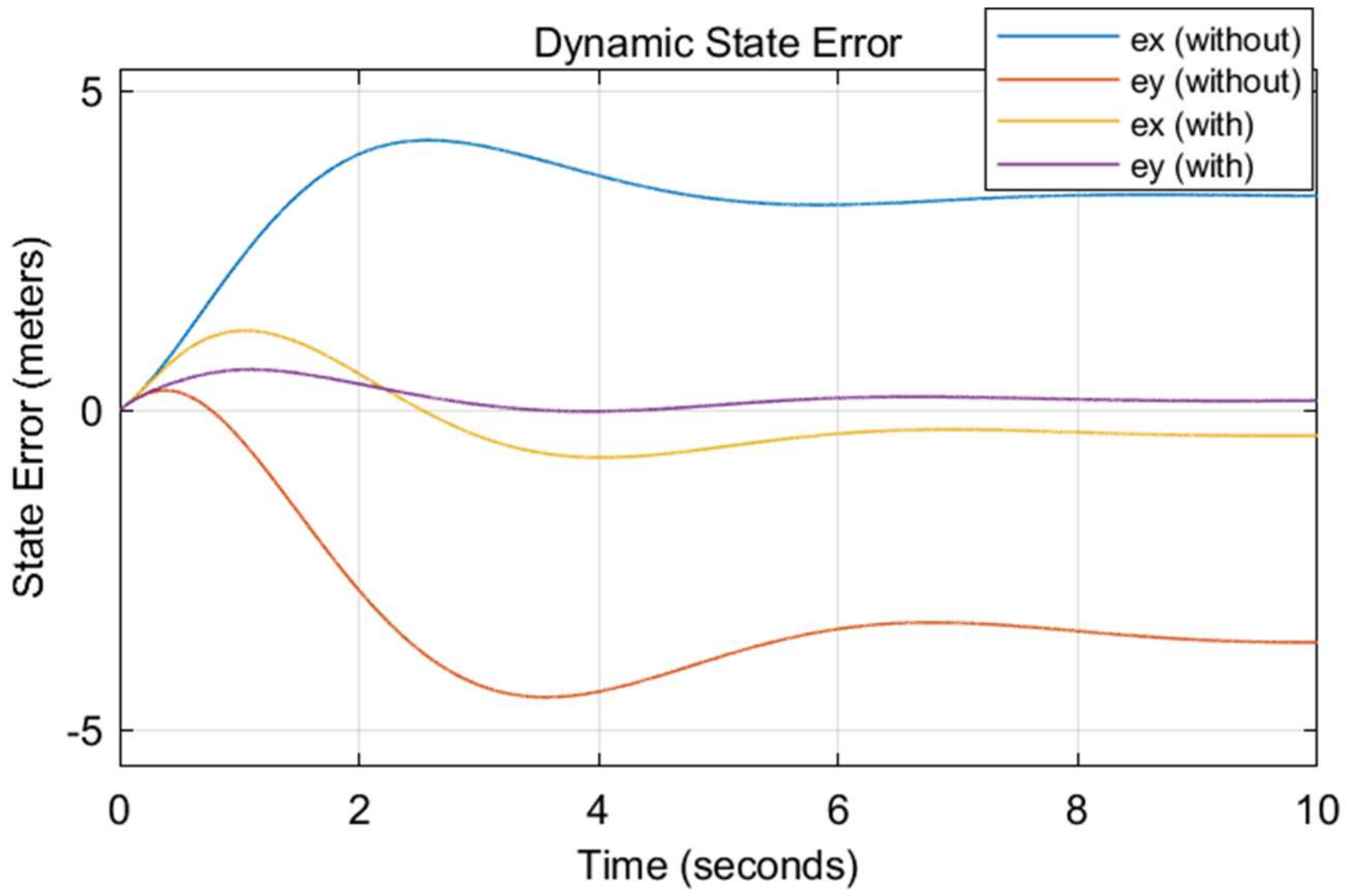

**Figure 5.** The resulting dynamic state errors (constant tilting angles, $\rho = 0.65$) with the controller equipped with the conventional attitude-position decoupler (blue curve: the dynamic state error along X, red curve: the dynamic state error along Y) and with the controller equipped with the modified attitude-position decoupler (yellow curve: the dynamic state error along X, purple curve: the dynamic state error along Y).

The tracking result (dynamic state errors) for the uniform rectilinear motional reference adopting the cat-trot gait with instant switch ($T = 2s$), defined in Section 4.2, is plotted in Figure 6. The blue and red curves represent the dynamic state errors along X and Y, respectively, equipped with the conventional attitude-position decoupler. The yellow and purple curves represent the dynamic state errors along X and Y, respectively, equipped with the modified attitude-position decoupler.

It can be concluded that the controller equipped with the modified attitude-position decoupler receives smaller supreme of the dynamic state errors, calculating from sufficient time; the maximums of the yellow and purple curves, calculating from sufficient time, are smaller than the maximums of the blue and red curves, respectively.

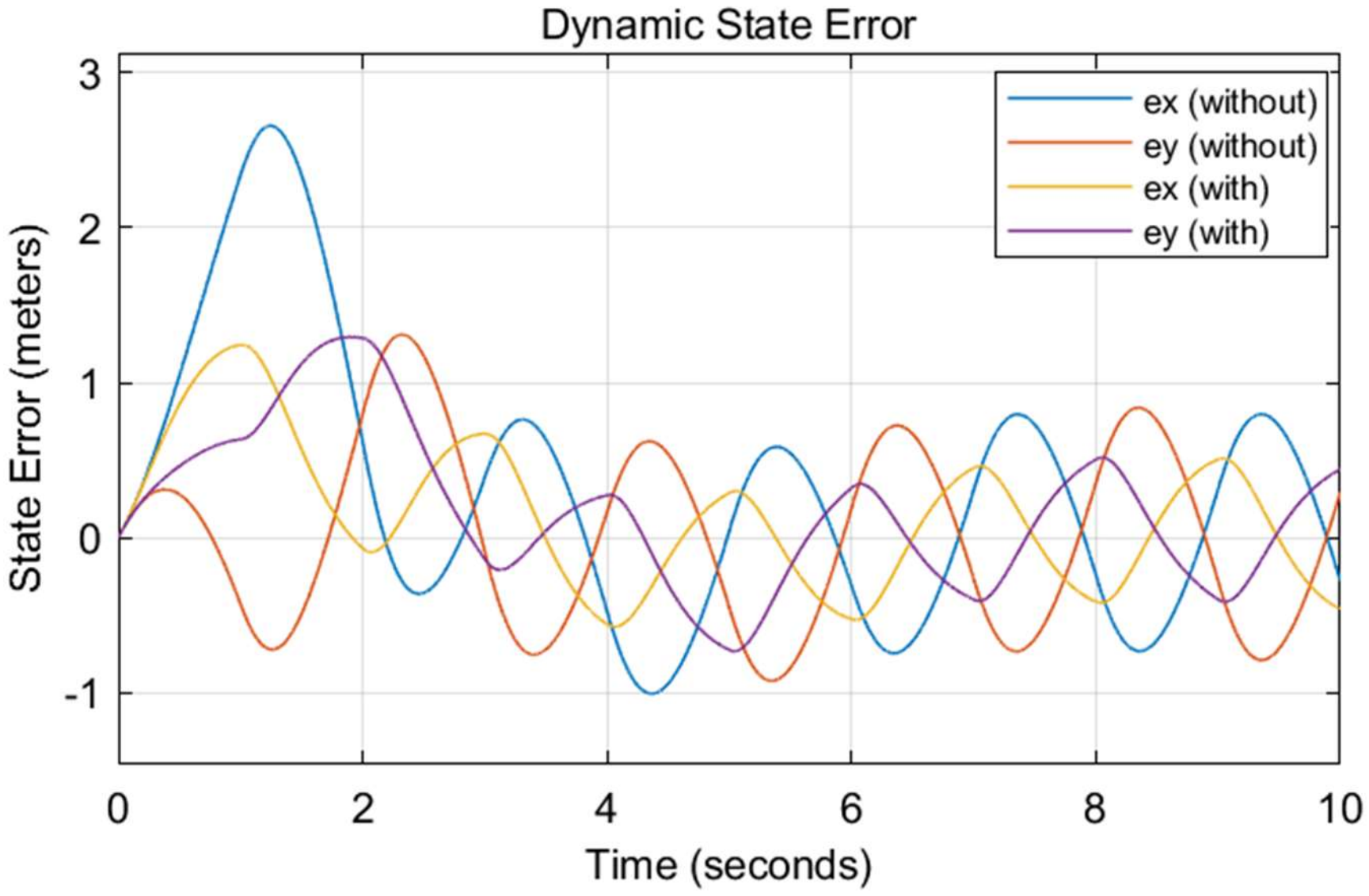

**Figure 6.** The resulting dynamic state errors, adopting the cat-trot gait with instant switch ($T = 2s$), equipped with the conventional attitude-position decoupler (blue curve: the dynamic state error along X, red curve: the dynamic state error along Y) and with the modified attitude-position decoupler (yellow curve: the dynamic state error along X, purple curve: the dynamic state error along Y).

Figure 7 demonstrates the dynamic state errors for the uniform rectilinear motional reference adopting the cat-trot gait with continuous switch ($T = 2s$), defined in Section 4.3. Identical to the previous denotations, the blue and red curves represent the dynamic state errors along X and Y, respectively, equipped with the conventional attitude-position decoupler. The yellow and purple curves represent the dynamic state errors along X and Y, respectively, equipped with the modified attitude-position decoupler.

The similar results can be concluded for the cat-trot gait with continuous switch ($T = 2s$) that the controller equipped with the modified attitude-position decoupler receives smaller supreme of the dynamic state errors, calculating from sufficient time; the maximums of the yellow and purple curves, calculating from sufficient time, are smaller than the maximums of the blue and red curves, respectively.

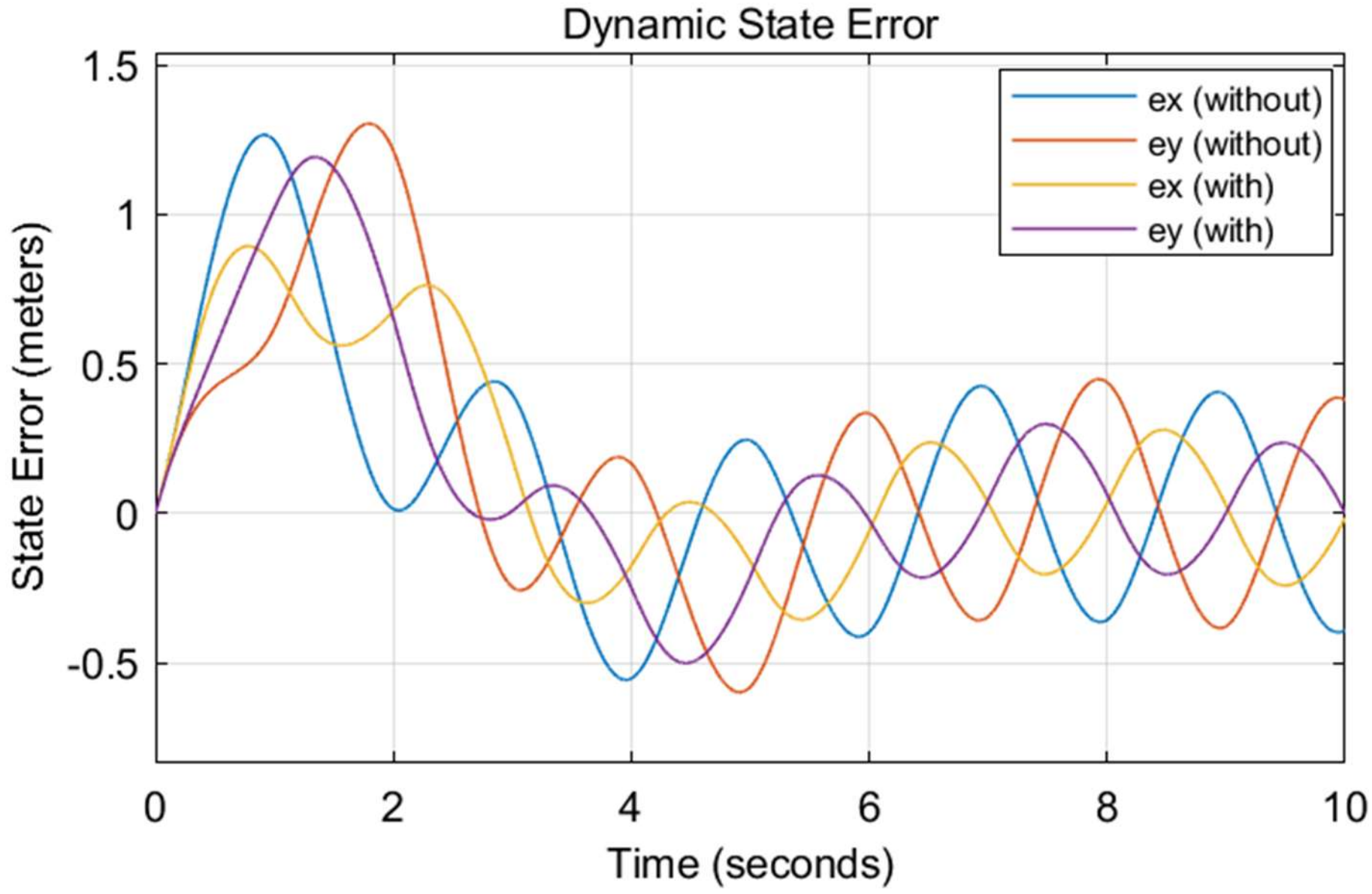

**Figure 7.** The resulting dynamic state errors, adopting the cat-trot gait with continuous switch ($T = 2s$), equipped with the conventional attitude-position decoupler (blue curve: the dynamic state error along X, red curve: the dynamic state error along Y) and with the modified attitude-position decoupler (yellow curve: the dynamic state error along X, purple curve: the dynamic state error along Y).

To compare the supremum of the dynamic state error, $\left({e_x}^2+{e_y}^2\right)^{\frac{1}{2}}$, calculating from sufficient time, for both controllers, with the conventional attitude-position decoupler and the modified attitude-position decoupler, with different periods $T$ ($T = 1, T = 2, T = 3$), we display the supremum of the relevant results in Figure 8 and 9, where Figure 8 shows the results adopting cat-trot gait with instant switch and Figure 9 shows the results adopting cat-trot gait with continuous switch.

The red curves in both figures represent the results of the supremum of the dynamic state error, $\left({e_x}^2+{e_y}^2\right)^{\frac{1}{2}}$, calculating from sufficient time, equipped with the conventional attitude-position decoupler. While the blue curves in both figures represent the results of the supremum of the dynamic state error, $\left({e_x}^2+{e_y}^2\right)^{\frac{1}{2}}$, calculating from sufficient time, equipped with the modified attitude-position decoupler.

Clearly, with the increase of the period, the gait receives larger dynamic state error (supremum) for both cat-trot gaits with instant switch and with continuous switch.

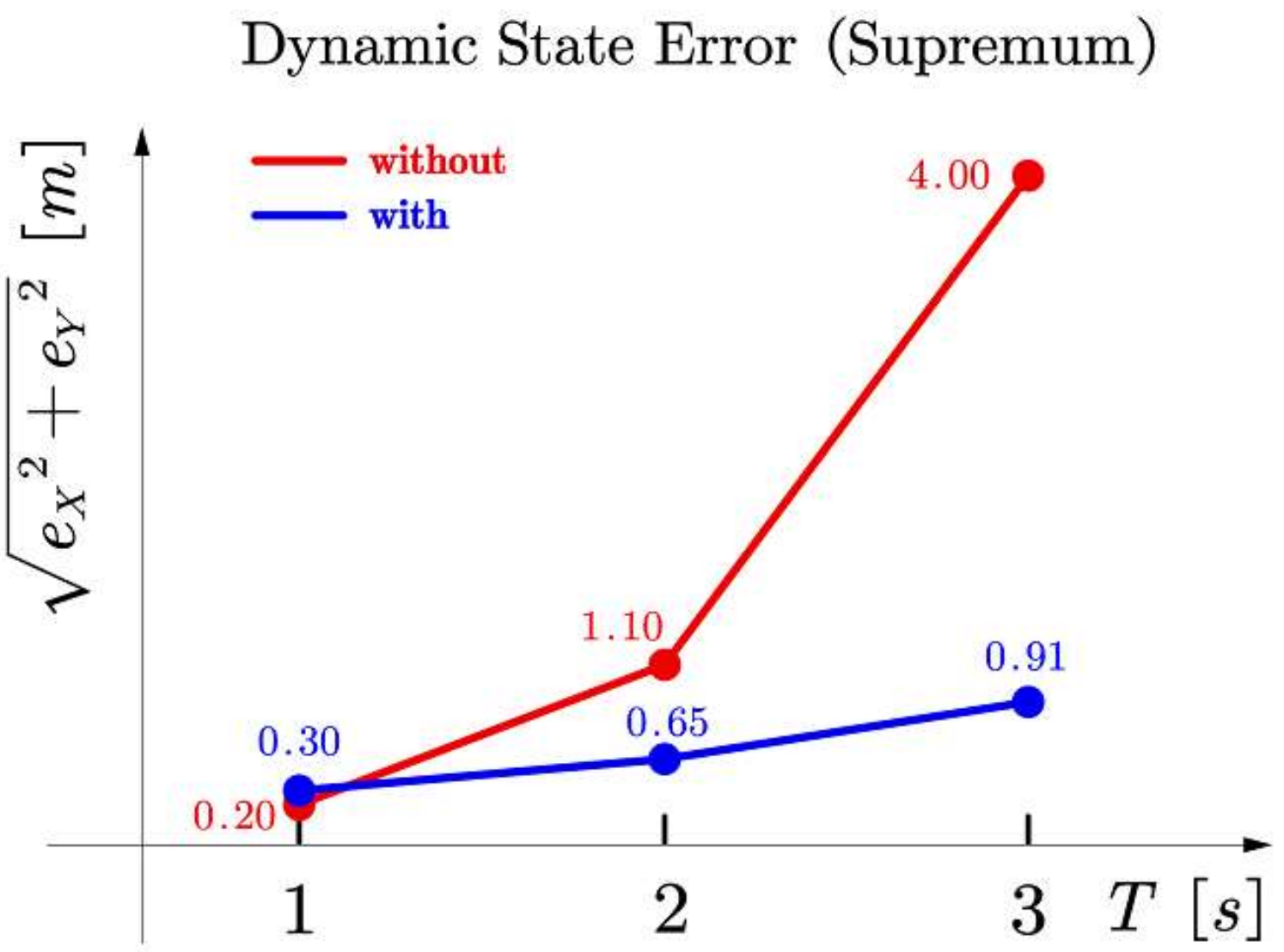

**Figure 8.** Supremum of the dynamic state error, $\left(e_x{}^2 + e_y{}^2\right)^{\frac{1}{2}}$, calculating from sufficient time, in cat-trot gait with instant switch with different periods ($T = 1, T = 2, T = 3$). The red curve represents the results equipped with the conventional attitude-position decoupler. The blue curve represents the results equipped with the modified attitude-position decoupler.

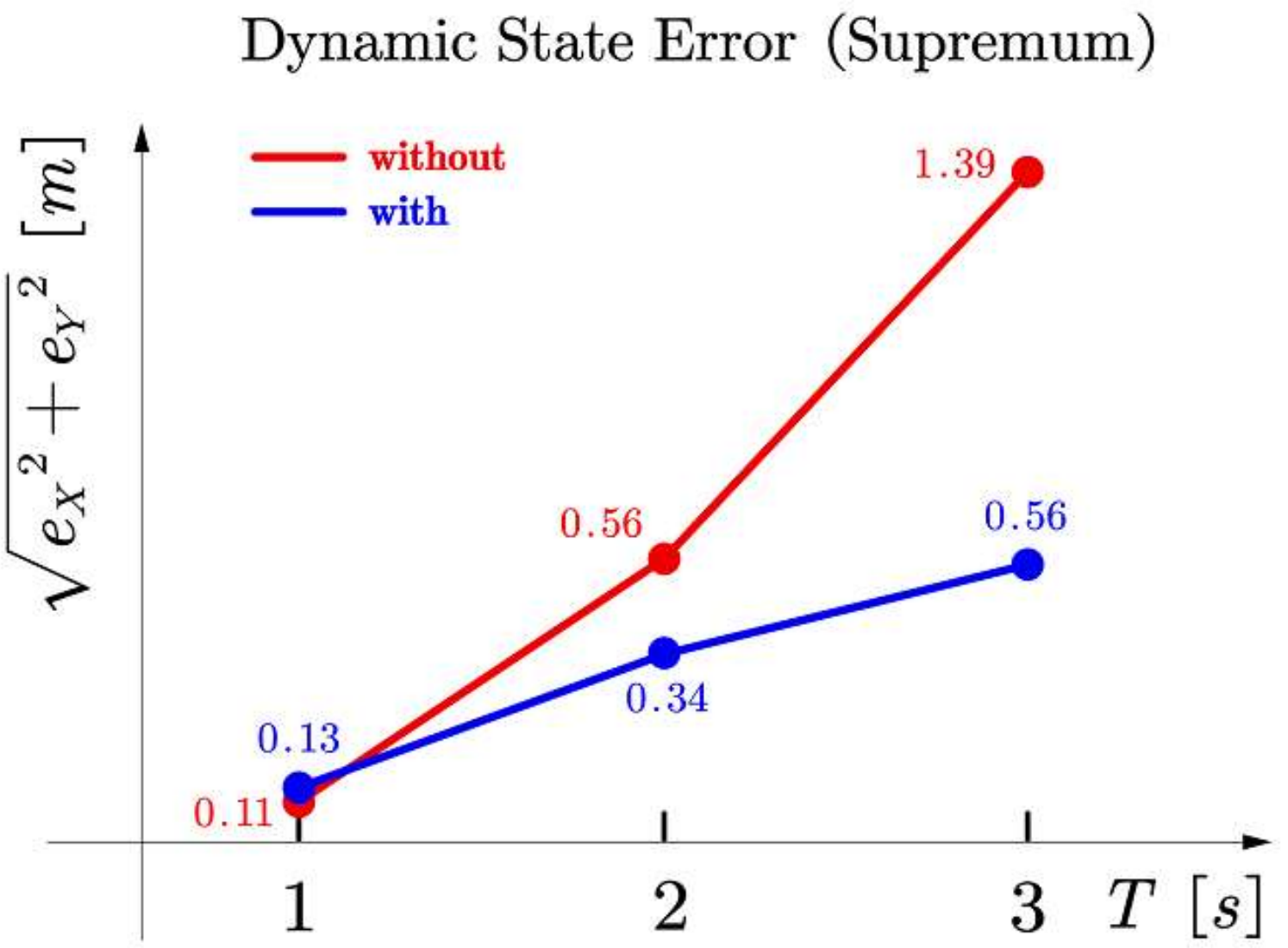

**Figure 9.** Supremum of the dynamic state error, $\left(e_x{}^2 + e_y{}^2\right)^{\frac{1}{2}}$, calculating from sufficient time, in cat-trot gait with continuous switch with different periods ($T = 1, T = 2, T = 3$). The red curve represents the results equipped with the conventional attitude-position decoupler. The blue curve represents the results equipped with the modified attitude-position decoupler.

The modified attitude-position decoupler significantly reduces the dynamic state error (supremum), especially for the gait with long period (e.g., $T = 3$) for both cat-trot gaits with instant switch and with continuous switch. While no significant differences in the dynamic state error (supremum) are reported for the gaits whose period is short for both cat-trot gaits with instant switch and with continuous switch.

In addition, the dynamic state error (supremum) received by the cat-trot gait with continuous switch (Figure 9) is smaller than the corresponding dynamic state error (supremum) received by the cat-trot gait with instant switch (Figure 8), given the s period and type of the attitude-position decoupler.

*7.3. Uniform Circular Motion*

In tracking a uniform circular motional reference defined in Section 6.3, Figure 10 displays the resulting dynamic state errors, $e_x = x_r - x$ and $e_y = y_r - y$, adopting the gait $\rho = 0.65$ (fixed tilting angles), equipped with the controller with the conventional attitude-position decoupler, denoted by $e_x$ (without) and $e_y$ (without), and with the controller with the modified attitude-position decoupler, denoted by $e_x$ (with) and $e_y$ (with).

Similar results can be found that the yellow curve, $e_x$ (with), and the purple curve, $e_y$ (with), are closer to zero comparing with the blue curve, $e_x$ (without), and the red curve, $e_y$ (without), respectively, after sufficient time. This indicates that the steady state errors are reduced after equipping with the modified attitude-position decoupler.

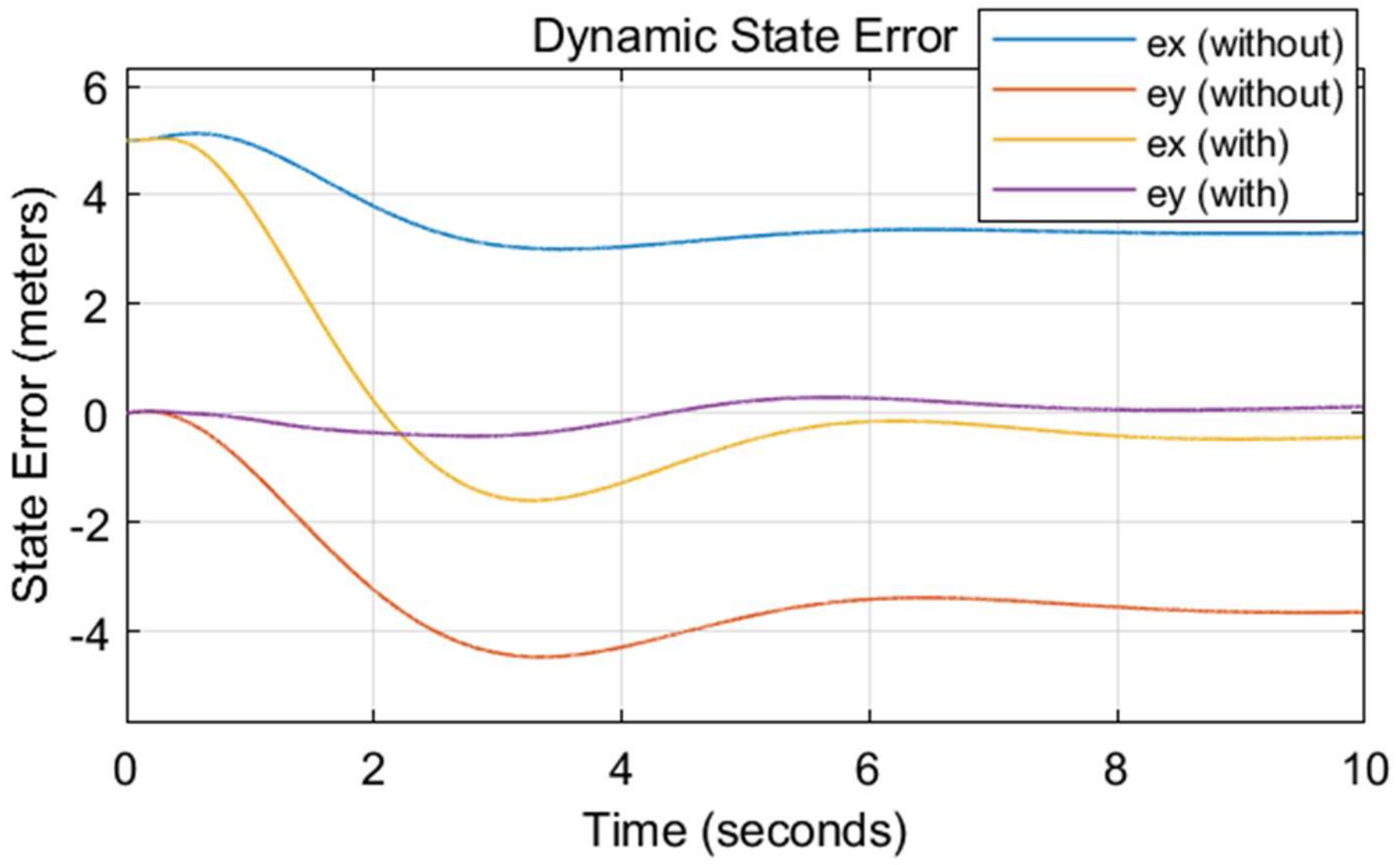

**Figure 10.** The resulting dynamic state errors (constant tilting angles, $\rho = 0.65$) with the controller equipped with the conventional attitude-position decoupler (blue curve: the dynamic state error along X, red curve: the dynamic state error along Y) and with the controller equipped with the modified attitude-position decoupler (yellow curve: the dynamic state error along X, purple curve: the dynamic state error along Y).

The tracking result (dynamic state errors) for the uniform circular motional reference adopting the cat-trot gait with instant switch ($T = 2s$), defined in Section 4.2, is plotted in Figure 11. The blue and red curves represent the dynamic state errors along X and Y, respectively, equipped with the conventional attitude-position decoupler. The yellow and purple curves represent the dynamic state errors along X and Y, respectively, equipped with the modified attitude-position decoupler.

Similarly, the results show that the controller equipped with the modified attitude-position decoupler receives smaller supreme of the dynamic state errors, calculating from sufficient time; the maximums of the yellow and purple curves, calculating from sufficient time, are smaller than the maximums of the blue and red curves, respectively.

Figure 12 demonstrates the dynamic state errors for the uniform rectilinear motional reference adopting the cat-trot gait with continuous switch ($T = 2s$), defined in Section 4.3. Identical to the previous denotations, the blue and red curves represent the dynamic state errors along X and Y, respectively, equipped with the conventional attitude-position decoupler. The yellow and purple curves represent the dynamic state errors along X and Y, respectively, equipped with the modified attitude-position decoupler.

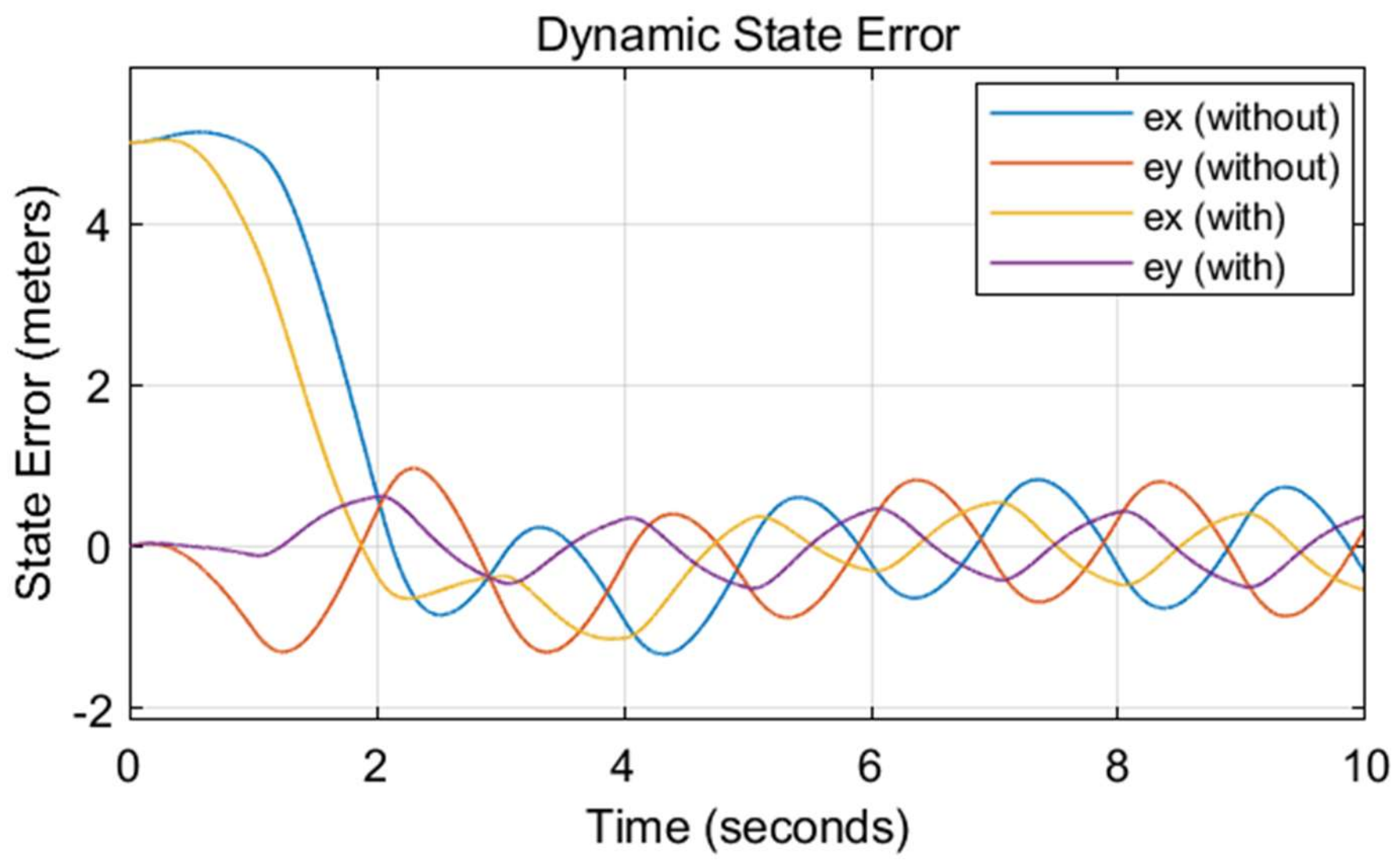

**Figure 11.** The resulting dynamic state errors, adopting the cat-trot gait with instant switch ($T = 2s$), equipped with the conventional attitude-position decoupler (blue curve: the dynamic state error along X, red curve: the dynamic state error along Y) and with the modified attitude-position decoupler (yellow curve: the dynamic state error along X, purple curve: the dynamic state error along Y).

Also, the similar results are concluded for the cat-trot gait with continuous switch ($T = 2s$) in this case that the controller equipped with the modified attitude-position decoupler receives smaller supreme of the dynamic state errors, calculating from sufficient time; the maximums of the yellow and purple curves, calculating from sufficient time, are smaller than the maximums of the blue and red curves, respectively.

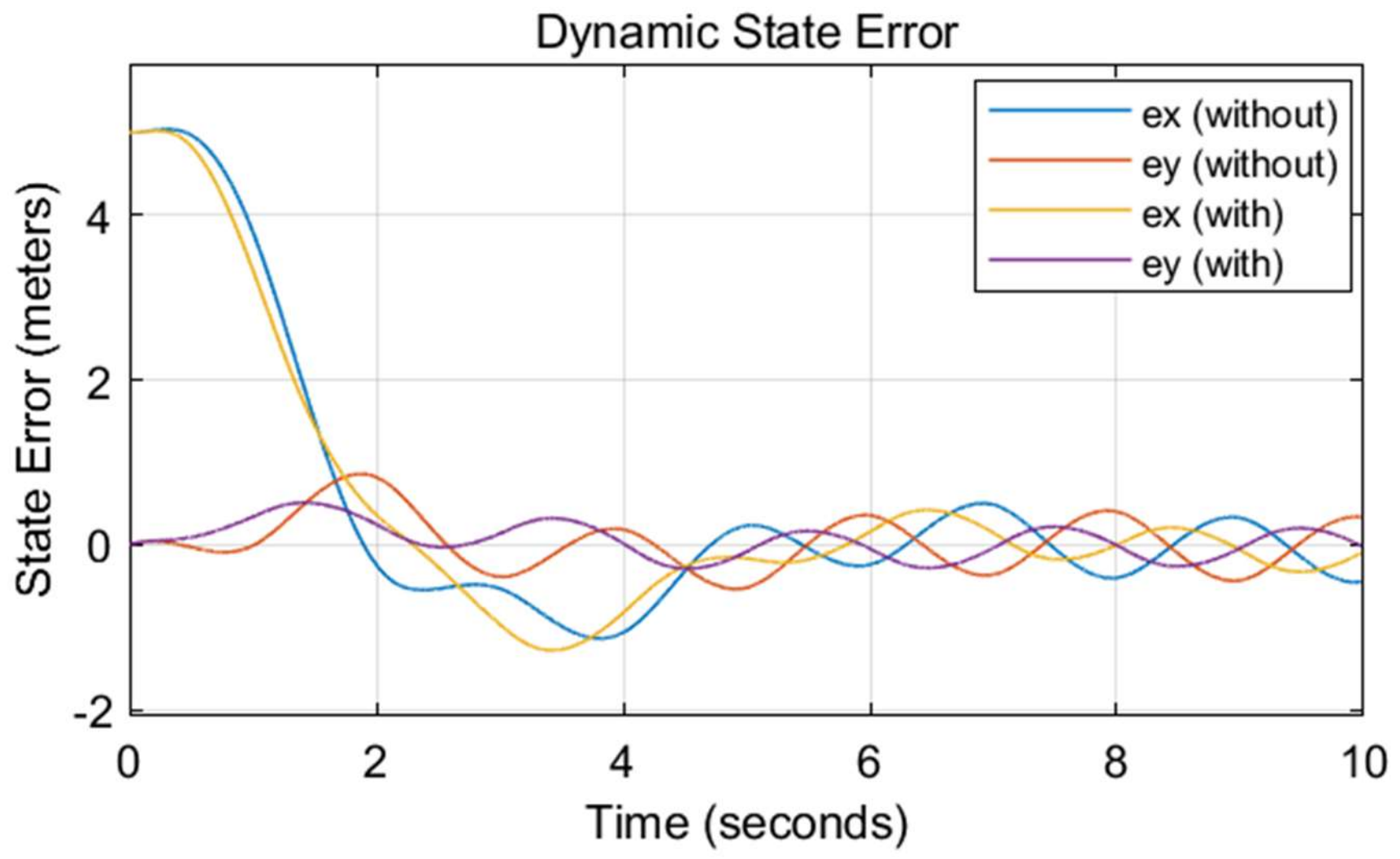

**Figure 12.** The resulting dynamic state errors, adopting the cat-trot gait with continuous switch ($T = 2s$), equipped with the conventional attitude-position decoupler (blue curve: the dynamic state error along X, red curve: the dynamic state error along Y) and with the modified attitude-position decoupler (yellow curve: the dynamic state error along X, purple curve: the dynamic state error along Y).

Likewise, to compare the supremum of the dynamic state error, $({e_x}^2 + {e_y}^2)^{\frac{1}{2}}$, calculating from sufficient time, for both controllers, with the conventional attitude-position decoupler and the modified attitude-position decoupler, with different periods $T$ ($T = 1, T = 2, T = 3$), we display the supremum of the relevant results in Figure 13 and 14, where Figure 13 shows the results adopting cat-trot gait with instant switch and Figure 14 shows the results adopting cat-trot gait with continuous switch.

The red curves in both figures represent the results of the supremum of the dynamic state error, $({e_x}^2 + {e_y}^2)^{\frac{1}{2}}$, calculating from sufficient time, equipped with the conventional attitude-position decoupler. While the blue curves in both figures represent the results of the supremum of the dynamic state error, $(e_x{}^2+e_y{}^2)^{\frac{1}{2}}$, calculating from sufficient time, equipped with the modified attitude-position decoupler.

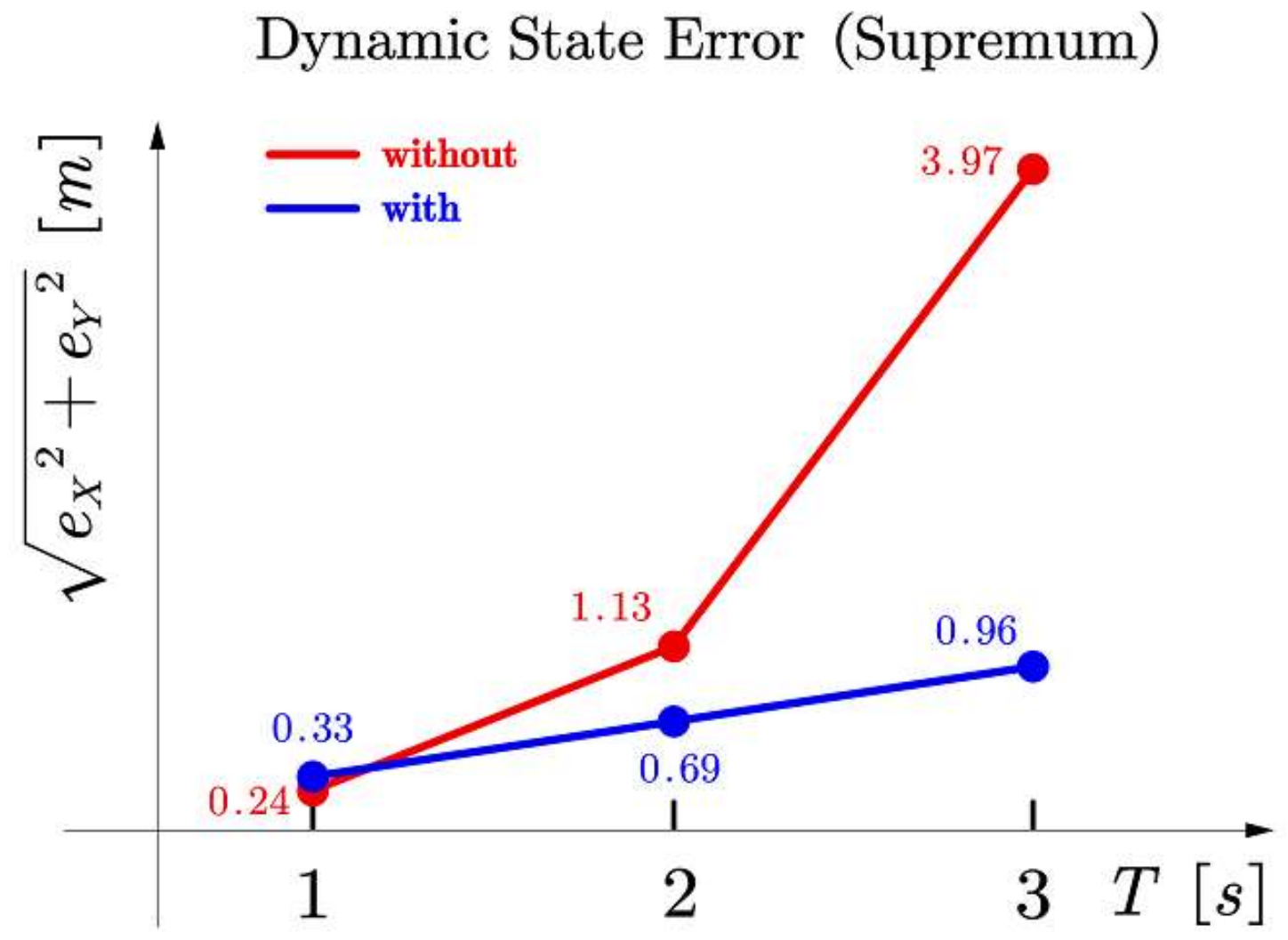

**Figure 13.** Supremum of the dynamic state error, $(e_x{}^2+e_y{}^2)^{\frac{1}{2}}$, calculating from sufficient time, in cat-trot gait with instant switch with different periods ($T = 1, T = 2, T = 3$). The red curve represents the results equipped with the conventional attitude-position decoupler. The blue curve represents the results equipped with the modified attitude-position decoupler.

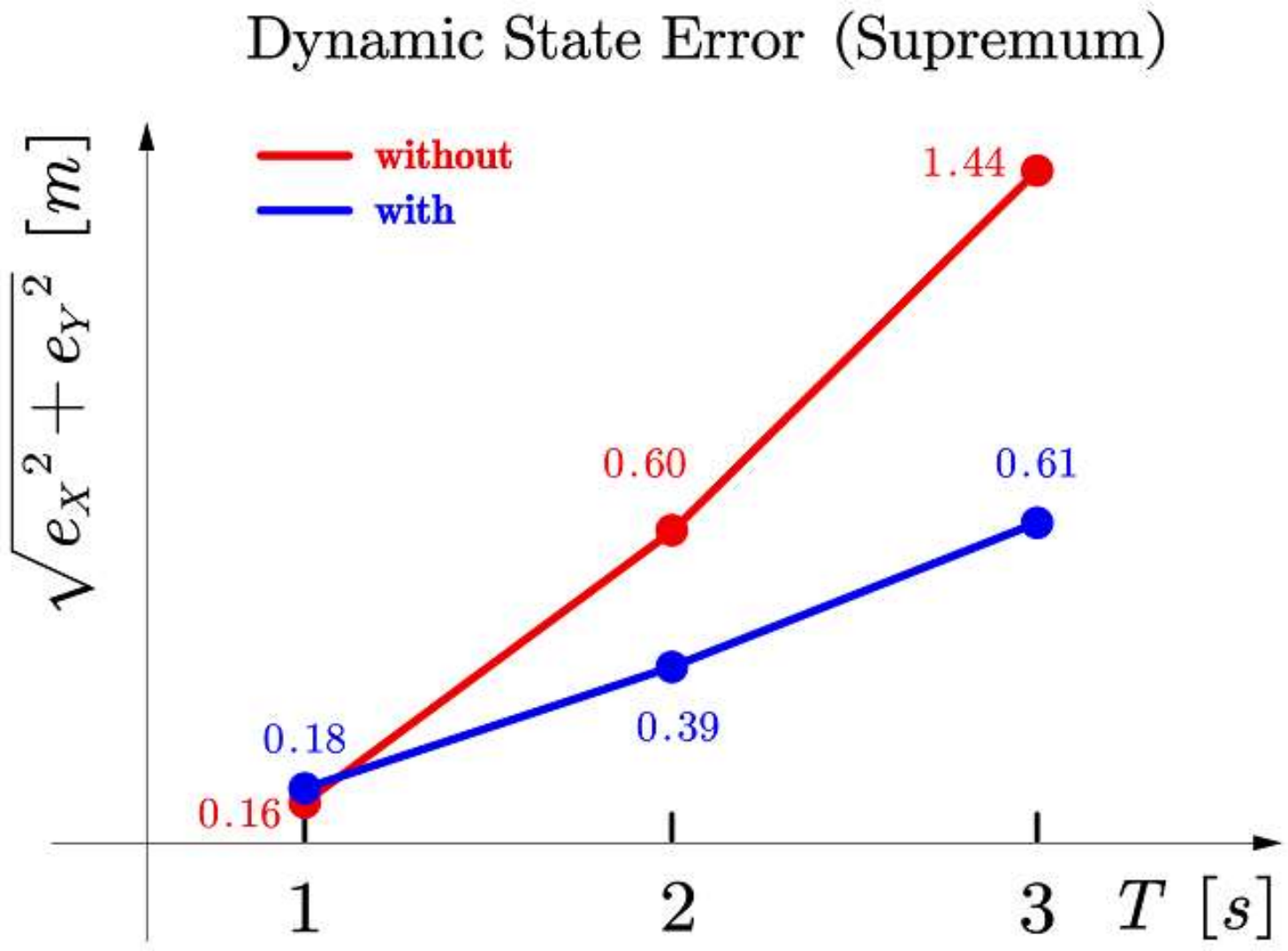

**Figure 14.** Supremum of the dynamic state error, $(e_x{}^2+e_y{}^2)^{\frac{1}{2}}$, calculating from sufficient time, in cat-trot gait with continuous switch with different periods ($T = 1, T = 2, T = 3$). The red curve represents the results equipped with the conventional attitude-position decoupler. The blue curve represents the results equipped with the modified attitude-position decoupler.

The similar results here show that, with the increase of the period, the gait receives larger dynamic state error (supremum) for both cat-trot gaits with instant switch and with continuous switch.

Also, the modified attitude-position decoupler significantly reduces the dynamic state error (supremum), especially for the gait with long period (e.g., $T = 3$) for both cat-trot gaits with instant switch and with continuous switch. While no significant differences in the dynamic state error (supremum) are reported for the gaits whose period is short for both cat-trot gaits with instant switch and with continuous switch.

In addition, the dynamic state error (supremum) received by the cat-trot gait with continuous switch (Figure 14) is smaller than the corresponding dynamic state error (supremum) received by the cat-trot gait with instant switch (Figure 13), given the s period and type of the attitude-position decoupler.

## 8. Conclusions and Discussions

The decoupling matrix is invertible in feedback linearization (attitude-altitude) for a tilt-rotor, adopting cat-trot gait. All the references considered in this research (setpoint, uniform rectilinear motion, uniform circular motion) are successfully tracked with acceptable steady state error by cat-trot gait.

The relationship between position and attitude of the tilt-rotor is elucidated. The modified attitude-position decoupler is invented for position-tracking problem for a tilt-rotor. It significantly reduces the dynamic state error (or steady state error for the point reference) comparing with the conventional attitude-position decoupler.

The length of the period of the cat-trot gait highly influences the dynamic state error in tracking a uniform rectilinear reference and a uniform circular reference. Specifically, a gait with a shorter period tends to receive a smaller supremum of the dynamic state error, calculating after sufficient time. On the other hand, the modified attitude-position decoupler notably reduces the dynamic state error for the cat-trot gait with a large period.

In general, the continuous cat-trot gait receives smaller supremum of the dynamic state error, calculating after sufficient time, than the discrete cat-trot gait, given same period and the type of the attitude-position decoupler.

There are several questions remaining to be answered further.

Firstly, Equation (17) takes the tilting angles constant; they are assumed to be constant during the entire flight. While the switching process in the discrete gait and the entire period in the continuous gait violate this condition. Addressing the discussions on the underlying robustness is beyond the scope of this research, which can be a further step.

Secondly, though the steady state errors and the supremum of the dynamic state errors, calculating after sufficient time, are found reduced after applying the modified attitude-position decoupler, further reduction of the dynamic state error may be possible by not ignoring the high-order infinitesimal terms while deducing the modified attitude-position decoupler.

Also, analysis on other different cat gaits (e.g., gallop and walk) can be another further step for the gait plan for the tilt-rotor.

## References

1. Ícaro Bezerra Viana; Luiz Manoel Santos Santana; Raphael Ballet; Davi Antônio dos Santos; Luiz Carlos Sandoval Góes Experimental Validation of a Trajectory Tracking Control Using the AR.Drone Quadrotor.; Fortaleza, Ceará, Brasil, 2016.
2. Merheb, A.-R.; Noura, H.; Bateman, F. Emergency Control of AR Drone Quadrotor UAV Suffering a Total Loss of One Rotor. *IEEE/ASME Trans. Mechatron.* **2017**, *22*, 961–971, doi:10.1109/TMECH.2017.2652399.
3. Lee, T.; Leok, M.; McClamroch, N.H. Geometric Tracking Control of a Quadrotor UAV on SE(3). In Proceedings of the 49th IEEE Conference on Decision and Control (CDC); IEEE: Atlanta, GA, December 2010; pp. 5420–5425.
4. Luukkonen, T. Modelling and Control of Quadcopter. *Independent research project in applied mathematics, Espoo* **2011**, *22*, 22.
5. Horla, D.; Hamandi, M.; Giernacki, W.; Franchi, A. Optimal Tuning of the Lateral-Dynamics Parameters for Aerial Vehicles With Bounded Lateral Force. *IEEE Robot. Autom. Lett.* **2021**, *6*, 3949–3955, doi:10.1109/LRA.2021.3067229.
6. Franchi, A.; Carli, R.; Bicego, D.; Ryll, M. Full-Pose Tracking Control for Aerial Robotic Systems With Laterally Bounded Input Force. *IEEE Trans. Robot.* **2018**, *34*, 534–541, doi:10.1109/TRO.2017.2786734.
7. Ryll, M.; Bulthoff, H.H.; Giordano, P.R. Modeling and Control of a Quadrotor UAV with Tilting Propellers. In Proceedings of the 2012 IEEE International Conference on Robotics and Automation; IEEE: St Paul, MN, USA, May 2012; pp. 4606–4613.
8. Nemati, A.; Kumar, R.; Kumar, M. Stabilizing and Control of Tilting-Rotor Quadcopter in Case of a Propeller Failure.; American Society of Mechanical Engineers: Minneapolis, Minnesota, USA, October 12 2016; p. V001T05A005.

9. Junaid, A.; Sanchez, A.; Bosch, J.; Vitzilaios, N.; Zweiri, Y. Design and Implementation of a Dual-Axis Tilting Quadcopter. *Robotics* **2018**, *7*, 65, doi:10.3390/robotics7040065.
10. Mistler, V.; Benallegue, A.; M'Sirdi, N.K. Exact Linearization and Noninteracting Control of a 4 Rotors Helicopter via Dynamic Feedback. In Proceedings of the Proceedings 10th IEEE International Workshop on Robot and Human Interactive Communication. ROMAN 2001 (Cat. No.01TH8591); IEEE: Paris, France, September 2001; pp. 586–593.
11. Mokhtari, A.; Benallegue, A. Dynamic Feedback Controller of Euler Angles and Wind Parameters Estimation for a Quadrotor Unmanned Aerial Vehicle. In Proceedings of the IEEE International Conference on Robotics and Automation, 2004. Proceedings. ICRA '04. 2004; IEEE: New Orleans, LA, USA, 2004; pp. 2359-2366 Vol.3.
12. Das, A.; Subbarao, K.; Lewis, F. Dynamic Inversion of Quadrotor with Zero-Dynamics Stabilization. In Proceedings of the 2008 IEEE International Conference on Control Applications; IEEE: San Antonio, TX, USA, 2008; pp. 1189–1194.
13. Lee, D.; Jin Kim, H.; Sastry, S. Feedback Linearization vs. Adaptive Sliding Mode Control for a Quadrotor Helicopter. *Int. J. Control Autom. Syst.* **2009**, *7*, 419–428, doi:10.1007/s12555-009-0311-8.
14. Martins, L.; Cardeira, C.; Oliveira, P. Feedback Linearization with Zero Dynamics Stabilization for Quadrotor Control. *J Intell Robot Syst* **2021**, *101*, 7, doi:10.1007/s10846-020-01265-2.
15. Mutoh, Y.; Kuribara, S. Control of Quadrotor Unmanned Aerial Vehicles Using Exact Linearization Technique with the Static State Feedback. *J. Autom. Control Eng* **2016**, *4*, 340–346, doi:10.18178/joace.4.5.340-346.
16. Nemati, A.; Kumar, M. Non-Linear Control of Tilting-Quadcopter Using Feedback Linearization Based Motion Control.; American Society of Mechanical Engineers: San Antonio, Texas, USA, October 22 2014; p. V003T48A005.
17. Al-Hiddabi, S.A. Quadrotor Control Using Feedback Linearization with Dynamic Extension. In Proceedings of the 2009 6th International Symposium on Mechatronics and its Applications; Sharjah, United Arab Emirates, March 2009; pp. 1–3.
18. Taniguchi, T.; Sugeno, M. Trajectory Tracking Controls for Non-Holonomic Systems Using Dynamic Feedback Linearization Based on Piecewise Multi-Linear Models. *IAENG Int. J. Appl. Math.* **2017**, *47*, 339–351.
19. Ryll, M.; Bulthoff, H.H.; Giordano, P.R. A Novel Overactuated Quadrotor Unmanned Aerial Vehicle: Modeling, Control, and Experimental Validation. *IEEE Trans. Contr. Syst. Technol.* **2015**, *23*, 540–556, doi:10.1109/TCST.2014.2330999.
20. Kumar, R.; Nemati, A.; Kumar, M.; Sharma, R.; Cohen, K.; Cazaurang, F. Tilting-Rotor Quadcopter for Aggressive Flight Maneuvers Using Differential Flatness Based Flight Controller.; American Society of Mechanical Engineers: Tysons, Virginia, USA, October 11 2017; p. V003T39A006.
21. Shen, Z.; Tsuchiya, T. Gait Analysis for a Tiltrotor: The Dynamic Invertible Gait. *Robotics* **2022**, *11*, 33, doi:10.3390/robotics11020033.
22. Rajappa, S.; Ryll, M.; Bulthoff, H.H.; Franchi, A. Modeling, Control and Design Optimization for a Fully-Actuated Hexarotor Aerial Vehicle with Tilted Propellers. In Proceedings of the 2015 IEEE International Conference on Robotics and Automation (ICRA); IEEE: Seattle, WA, USA, May 2015; pp. 4006–4013.
23. Shen, Z.; Tsuchiya, T. Singular Zone in Quadrotor Yaw–Position Feedback Linearization. *Drones* **2022**, *6*, 20, doi:doi.org/10.3390/drones6040084.
24. Mokhtari, A.; M'Sirdi, N.K.; Meghriche, K.; Belaidi, A. Feedback Linearization and Linear Observer for a Quadrotor Unmanned Aerial Vehicle. *Advanced Robotics* **2006**, *20*, 71–91, doi:10.1163/156855306775275495.
25. Ansari, U.; Bajodah, A.H.; Hamayun, M.T. Quadrotor Control Via Robust Generalized Dynamic Inversion and Adaptive Non-Singular Terminal Sliding Mode. *Asian Journal of Control* **2019**, *21*, 1237–1249, doi:10.1002/asjc.1800.
26. Kuantama, E.; Tarca, I.; Tarca, R. Feedback Linearization LQR Control for Quadcopter Position Tracking. In Proceedings of the 2018 5th International Conference on Control, Decision and Information Technologies (CoDIT); IEEE: Thessaloniki, April 2018; pp. 204–209.
27. Hirose, S. A Study of Design and Control of a Quadruped Walking Vehicle. *The International Journal of Robotics Research* **1984**, *3*, 113–133, doi:10.1177/027836498400300210.
28. Bennani, M.; Giri, F. Dynamic Modelling of a Four-Legged Robot. *Journal of Intelligent and Robotic Systems* **1996**, *17*, 419–428.
29. Lewis, M.A.; Bekey, G.A. Gait Adaptation in a Quadruped Robot. *Autonomous Robots* **2002**, 301–312.
30. Verdugo, M.R.; Rahal, S.C.; Agostinho, F.S.; Govoni, V.M.; Mamprim, M.J.; Monteiro, F.O. Kinetic and Temporospatial Parameters in Male and Female Cats Walking over a Pressure Sensing Walkway. *BMC Veterinary Research* **2013**, *9*, 129, doi:10.1186/1746-6148-9-129.
31. Wisleder, D.; Zernicke, R.F.; Smith, J.L. Speed-Related Changes in Hindlimb Intersegmental Dynamics during the Swing Phase of Cat Locomotion. *Exp Brain Res* **1990**, *79*, doi:10.1007/BF00229333.
32. Vilensky, J.A.; Njock Libii, J.; Moore, A.M. Trot-Gallop Gait Transitions in Quadrupeds. *Physiology & Behavior* **1991**, *50*, 835–842, doi:10.1016/0031-9384(91)90026-K.

33. Goodarzi, F.A.; Lee, D.; Lee, T. Geometric Adaptive Tracking Control of a Quadrotor Unmanned Aerial Vehicle on SE(3) for Agile Maneuvers. *Journal of Dynamic Systems, Measurement, and Control* **2015**, *137*, 091007, doi:10.1115/1.4030419.
34. Shi, X.-N.; Zhang, Y.-A.; Zhou, D. A Geometric Approach for Quadrotor Trajectory Tracking Control. *International Journal of Control* **2015**, *88*, 2217–2227, doi:10.1080/00207179.2015.1039593.
35. Lee, T.; Leok, M.; McClamroch, N.H. Nonlinear Robust Tracking Control of a Quadrotor UAV on SE(3): Nonlinear Robust Tracking Control of a Quadrotor UAV. *Asian J Control* **2013**, *15*, 391–408, doi:10.1002/asjc.567.
36. Chovancová, A.; Fico, T.; Chovanec, Ľ.; Hubinsk, P. Mathematical Modelling and Parameter Identification of Quadrotor (a Survey). *Procedia Engineering* **2014**, *96*, 172–181, doi:10.1016/j.proeng.2014.12.139.
37. Abas, N.; Legowo, A.; Akmeliawati, R. Parameter Identification of an Autonomous Quadrotor. In Proceedings of the 2011 4th International Conference on Mechatronics (ICOM); May 2011; pp. 1–8.
38. Altug, E.; Ostrowski, J.P.; Mahony, R. Control of a Quadrotor Helicopter Using Visual Feedback. In Proceedings of the Proceedings 2002 IEEE International Conference on Robotics and Automation (Cat. No.02CH37292); Washington, DC, USA, May 2002; Vol. 1, pp. 72–77.
39. Voos, H. Nonlinear Control of a Quadrotor Micro-UAV Using Feedback-Linearization. In Proceedings of the 2009 IEEE International Conference on Mechatronics; IEEE: Malaga, Spain, 2009; pp. 1–6.
40. Zhou, Q.-L.; Zhang, Y.; Rabbath, C.-A.; Theilliol, D. Design of Feedback Linearization Control and Reconfigurable Control Allocation with Application to a Quadrotor UAV. In Proceedings of the 2010 Conference on Control and Fault-Tolerant Systems (SysTol); IEEE: Nice, France, October 2010; pp. 371–376.
41. Shen, Z.; Ma, Y.; Tsuchiya, T. Stability Analysis of a Feedback-Linearization-Based Controller with Saturation: A Tilt Vehicle with the Penguin-Inspired Gait Plan. *arXiv preprint arXiv:2111.14456* **2021**.
42. Shen, Z.; Tsuchiya, T. The Pareto-Frontier-Based Stiffness of A Controller: Trade-off between Trajectory Plan and Controller Design. *arXiv:2108.08667 [cs, eess]* **2021**.
43. Michael, N.; Mellinger, D.; Lindsey, Q.; Kumar, V. The GRASP Multiple Micro-UAV Testbed. *IEEE Robotics Automation Magazine* **2010**, *17*, 56–65, doi:10.1109/MRA.2010.937855.

# Chapter 5

# Singularity and Unacceptable Attitude Curve (Cat-Inspired Gaits for a Tilt-Rotor -- From Symmetrical to Asymmetrical)

**This Chapter has been published on Cat-Inspired Gaits for a Tilt-Rotor—From Symmetrical to Asymmetrical. *Robotics* 2022, *11*, 60, doi:10.3390/robotics11030060 as the first author. The rest co-author(s) contribute nothing other than supervising and/or minor edition.**

**Abstract:** Among the tilt-rotors (quadrotors) developed in recent decades, Ryll's model with eight inputs (four magnitudes of thrusts and four tilting angles) attracted great attention. Typical feedback linearization maneuvers all of the eight inputs with a united control rule to stabilize this tilt-rotor. Instead of assigning the tilting angles by the control rule, the recent research predetermines the tilting angles and leaves the magnitudes of thrusts with the only control signals. These tilting angles are designed to mimic the cat-trot gait while avoiding the singular decoupling matrix in feedback linearization. To complete the discussions of the cat-gait inspired tilt-rotor gaits, this research addresses the analyses on the rest of the common cat gaits, walk, run, transverse gallop, and rotary gallop. It is found that the singular decoupling matrix exists in walk gait, transverse gallop gait, and rotary gallop gait; the decoupling matrix can hardly be guaranteed to be invertible analytically. Further modifications (scaling) are conducted to these three gaits to accommodate the application of feedback linearization; the acceptable attitudes, leading to invertible decoupling matrix, for each scaled gait are evaluated in the roll-pitch diagram. The modified gaits with different periods are then applied to the tilt-rotor in tracking experiments, in which the references are uniform rectilinear motion and uniform circular motion with or without the equipment of the modified attitude-position decoupler. All the experiments are simulated in Simulink, MATLAB. The result shows that these gaits, after modifications, are feasible in tracking references, especially for the cases equipped with the modified attitude-position decoupler.

## 1. Introduction

The tilt-rotor quadrotor [1–4], which is also referred to as the thrust vectoring quadrotor [5,6], is a novel type of quadrotor. Augmented with an additional mechanical structure [7,8], it is able to provide lateral force. Among the designs of the tilt-rotor, Ryll's model, the tilt-rotor with eight inputs, received great attention in the last decade.

Various control methods have been analyzed in stabilizing Ryll's model. These methods include feedback linearization [1], geometric control [9], PID (proportional integral derivative) control [10], backstepping and sliding mode control [11], etc. Among them, the feedback linearization is the first approach [12] proposed in controlling Ryll's model; this method decouples the original nonlinear system to generate the scenario compatible with the linear controller.

However, several unique properties of feedback linearization can hinder the application of this method. One of them is the saturation in feedback linearization [13], which is parallel to the saturation in the geometric control [14]. Also, the state drift phenomenon can be a problem [15]. Another issue is the intensive change in the tilting angles while applying feedback linearization; the resulting changes in the tilting angles can be too large or too intensive. Notice that this intensive change in the tilting angles is not unique in the feedback linearization, e.g., PID [16].

Generally [17], the eight inputs are fully assigned by a united control rule, which makes the number of degrees of freedom less than [12] or equal to [10] the number of inputs. Indeed, these approaches avoid the under-actuated system. Further, the decoupling matrix in this scenario is invertible within the interested attitude region while applying feedback linearization. However, the adverse effect is the intensive change in the tilting angles mentioned beforehand, which may not be desired in application.

Our previous research [18] averts this problem by decreasing the number of inputs of the tilt-rotor. Instead of assigning the tilting angles by the united control rule, a gait plan is applied to the tilting angles beforehand, leaving the magnitudes of thrusts the only actual control inputs. It produces an under-actuated control scenario with the attitude region introducing the singularity in the decoupling matrix. The tilting angles mimicked the cat-trot gait in other research [19] about the tracking problems.

The cat-trot-inspired gait plan for the tilt-rotor guarantees that the decoupling matrix is always invertible during the entire flight, given that the attitude is close to zero, e.g., roll angle and pitch angle are close to zero [19]. The determinant of the relevant decoupling matrix is proved non-zero analytically, assuming that the roll angle and pitch angle are zero.

Unfortunately, the determinant of the decoupling matrix of the gaits analyzed in this paper are not guaranteed to be non-zero without the restrictions on the attitude.

This paper provides novel gaits inspired by the rest of the typical cat gaits, both symmetrical (walk and run) and asymmetrical (transverse gallop and rotary gallop) [20], for the tilt-rotor. The singularity of the decoupling matrix for each of these gaits is analyzed numerically; some gaits are modified by scaling to receive an invertible decoupling matrix before applying feedback linearization. Note that this scaling method is proved to be an effective approach to modify the unqualified gaits, e.g., the gait introducing invertible decoupling matrix only in a small region of the attitude, for a tilt-rotor for the first time.

The degrees of freedom tracked directly are attitude and altitude, e.g., roll, pitch, yaw, and altitude. The rest positions are influenced by manipulating the attitude properly; the modified position-attitude decoupler for the tilt-rotor advanced in the previous research [19] is adopted to track the position.

Two references (uniform rectilinear motion and uniform circular motion) are designed for the tilt-rotor to track in the experiments. Each of the four gaits is applied and analyzed with different periods in the tracking experiment. The result of the position-tracking problem is also compared in the cases with or without advancing the attitude-altitude decoupler. The experiment is simulated in Simulink, MATLAB.

The rest of this paper is structured as follows. Section 2 introduces the dynamics of the tilt-rotor. The controller and the gaits are designed in Section 3. Section 4 analyzes the singularity of the decoupling matrix in each gait and puts forward a gait modification method. Section 5 sets up the references in the tracking problem. The result is demonstrated in Section 6. The conclusions and discussions are addressed in Section 7.

## 2. Dynamics of the Tilt-Rotor

This section details the dynamics of the tilt-rotor. A comprehensive discussion on it can be referred to in previous studies adopting the same dynamics model [12,18,19].

The model of the tilt-rotor investigated in this study, Figure 1 [18], was initially put forward by Ryll [12].

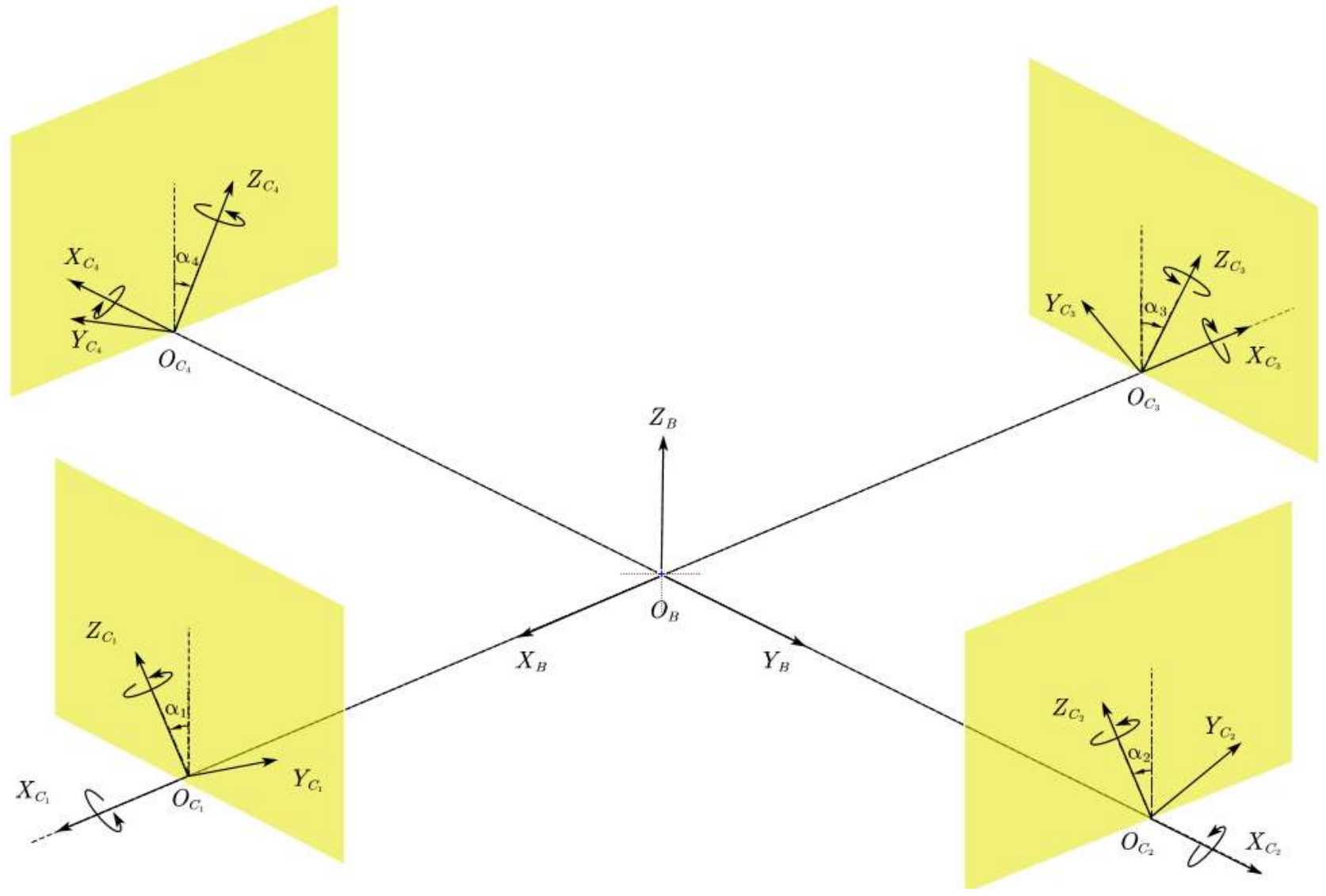

**Figure 1.** The sketch of a tilt-rotor.

The frames introduced in the dynamics of this tilt-rotor are the earth frame $\mathcal{F}_E$, body-fixed frame $\mathcal{F}_B$, and four rotor frames $\mathcal{F}_{C_i}(i$=1,2,3,4), each of which is fixed on the tilt motor mounted on the end of each arm. Rotor 1 and 3 are assumed to rotate clockwise along $Z_{C_1}$ and $Z_{C_3}$. Rotor 2 and 4 are assumed to rotate counter-clockwise along $Z_{C_2}$ and $Z_{C_4}$.

Based on the Newton-Euler formula, the position of the tilt-rotor [18] is given by

$$\ddot{P} = \begin{bmatrix} 0 \\ 0 \\ -g \end{bmatrix} + \frac{1}{m} \cdot {}^{W}R \cdot F(\alpha) \cdot \begin{bmatrix} \varpi_1 \cdot |\varpi_1| \\ \varpi_2 \cdot |\varpi_2| \\ \varpi_3 \cdot |\varpi_3| \\ \varpi_4 \cdot |\varpi_4| \end{bmatrix} \triangleq \begin{bmatrix} 0 \\ 0 \\ -g \end{bmatrix} + \frac{1}{m} \cdot {}^{W}R \cdot F(\alpha) \cdot w \tag{1}$$

where $P = [X \quad Y \quad Z]^T$ represents the position with respect to $\mathcal{F}_E$, $m$ represents the total mass, $g$ represents the gravitational acceleration, $\varpi_i$, ($i = 1,2,3,4$) represents the angular velocity of the propeller ($\varpi_{1,3} < 0$, $\varpi_{2,4} > 0$) with respect to $\mathcal{F}_{C_i}(i = 1,2,3,4)$, $w = [\varpi_1 \cdot |\varpi_1|, \varpi_2 \cdot |\varpi_2|, \varpi_3 \cdot |\varpi_3|, \varpi_4 \cdot |\varpi_4|]^T$, ${}^{W}R$ represents the rotational matrix [21],

$${}^{W}R = \begin{bmatrix} c\theta \cdot c\psi & s\phi \cdot s\theta \cdot c\psi - c\phi \cdot s\psi & c\phi \cdot s\theta \cdot c\psi + s\phi \cdot s\psi \\ c\theta \cdot s\psi & s\phi \cdot s\theta \cdot s\psi + c\phi \cdot c\psi & c\phi \cdot s\theta \cdot s\psi - s\phi \cdot c\psi \\ -s\theta & s\phi \cdot c\theta & c\phi \cdot c\theta \end{bmatrix} \tag{2}$$

where $s\Lambda = \sin(\Lambda)$ and $c\Lambda = \cos(\Lambda)$. $\phi$, $\theta$, and $\psi$ are roll angle, pitch angle, and yaw angle, respectively, the tilting angles $\alpha = [\alpha_1 \quad \alpha_2 \quad \alpha_3 \quad \alpha_4]$. The positive directions of the tilting angles are defined in Figure 1. $F(\alpha)$ is given by

$$F(\alpha) = \begin{bmatrix} 0 & K_f \cdot s2 & 0 & -K_f \cdot s4 \\ K_f \cdot s1 & 0 & -K_f \cdot s3 & 0 \\ -K_f \cdot c1 & K_f \cdot c2 & -K_f \cdot c3 & K_f \cdot c4 \end{bmatrix} \tag{3}$$

where $si = \sin(\alpha_i)$, $ci = \cos(\alpha_i)$, and ($i = 1,2,3,4$). $K_f$ ($8.048 \times 10^{-6} N \cdot s^2/rad^2$) is the coefficient of the thrust.

The angular velocity of the body with respect to $\mathcal{F}_B$, $\omega_B = [p \quad q \quad r]^T$, is governed (Newton-Euler formula) by

$$\dot{\omega}_B = {I_B}^{-1} \cdot \tau(\alpha) \cdot w \tag{4}$$

where $I_B$ is the matrix of moments of inertia, $K_m$ ($2.423 \times 10^{-7} N \cdot m \cdot s^2/rad^2$) is the coefficient of the drag, and $L$ is the length of the arm,

$$\tau(\alpha)=\begin{bmatrix}0 & L\cdot K_f\cdot c2-K_m\cdot s2 & 0 & -L\cdot K_f\cdot c4+K_m\cdot s4\\ L\cdot K_f\cdot c1+K_m\cdot s1 & 0 & -L\cdot K_f\cdot c3-K_m\cdot s3 & 0\\ L\cdot K_f\cdot s1-K_m\cdot c1 & -L\cdot K_f\cdot s2-K_m\cdot c2 & L\cdot K_f\cdot s3-K_m\cdot c3 & -L\cdot K_f\cdot s4-K_m\cdot c4\end{bmatrix}. \quad (5)$$

The relationship [22–24] between the angular velocity of the body, $\omega_B$, and the attitude rotation matrix (${}^{W}R$) is given by

$${}^{W}\dot{R}={}^{W}R\cdot\widehat{\omega}_B \quad (6)$$

where " $\widehat{\ }$ " is the hat operation used to produce the skew matrix, ${}^{W}\dot{R}$ represents the derivative of rotation matrix.

The parameters of this tilt-rotor are specified as follows: $m=0.429kg,\ L=0.1785m,\ g=9.8N/kg,\ I_B=\mathrm{diag}([2.24\times10^{-3},2.99\times10^{-3},4.80\times10^{-3}])kg\cdot m^2$.

## 3. Controller Design and Gait Plan

The same control method as in our previous research [18,19] is adopted. This section briefs this controller and introduces animal-inspired gaits analyzed in this study.

### *3.1. Feedback Linearization and Modified Attitude-Position Decoupler*

The degrees of freedom controlled independently in this research are selected as attitude ($\phi$, $\theta$, and $\psi$) and altitude ($Z$).

Define

$$\begin{bmatrix}y_1\\y_2\\y_3\\y_4\end{bmatrix}=\begin{bmatrix}\phi\\\theta\\\psi\\Z\end{bmatrix}. \quad (7)$$

Assuming

$$\dot{\alpha}_i\equiv 0, i=1,2,3,4. \quad (8)$$

we receive

$$\begin{aligned}\begin{bmatrix}\dddot{y}_1\\\dddot{y}_2\\\dddot{y}_3\\\dddot{y}_4\end{bmatrix}&=\begin{bmatrix}I_B{}^{-1}\cdot\tau(\alpha)\\ [0\quad 0\quad 1]\cdot\frac{K_f}{m}\cdot{}^{W}R\cdot F(\alpha)\cdot 2\cdot\begin{bmatrix}|\varpi_1| & & & \\ & |\varpi_2| & & \\ & & |\varpi_3| & \\ & & & |\varpi_4|\end{bmatrix}\end{bmatrix}^{4\times4}\cdot\begin{bmatrix}\dot{\varpi}_1\\\dot{\varpi}_2\\\dot{\varpi}_3\\\dot{\varpi}_4\end{bmatrix}\\ &+[0\quad 0\quad 1]\cdot\frac{K_f}{m}\cdot{}^{W}R\cdot\widehat{\omega}_B\cdot F(\alpha)\cdot w\cdot\begin{bmatrix}0\\0\\0\\1\end{bmatrix}\\ &\triangleq\bar{\Delta}\cdot\begin{bmatrix}\dot{\varpi}_1\\\dot{\varpi}_2\\\dot{\varpi}_3\\\dot{\varpi}_4\end{bmatrix}+Ma.\end{aligned} \quad (9)$$

where $\bar{\Delta}$ is called the decoupling matrix [25], $[\dot{\varpi}_1\quad\dot{\varpi}_2\quad\dot{\varpi}_3\quad\dot{\varpi}_4]^T\triangleq U$ is the new input vector, and $Ma$ are the remaining terms not containing $\dot{\varpi}_1$, $\dot{\varpi}_2$, $\dot{\varpi}_3$, or $\dot{\varpi}_4$, which is $[0\quad 0\quad 1]\cdot K_f/m\cdot{}^{W}R\cdot\widehat{\omega}_B\cdot F(\alpha)\cdot w\cdot[0\quad 0\quad 0\quad 1]^T$.

Finally, design the PD (proportional derivative) controller based on

$$\begin{bmatrix}\dot{\varpi}_1\\\dot{\varpi}_2\\\dot{\varpi}_3\\\dot{\varpi}_4\end{bmatrix}=\bar{\Delta}^{-1}\cdot\left(\begin{bmatrix}\dddot{y}_{1d}\\\dddot{y}_{2d}\\\dddot{y}_{3d}\\\dddot{y}_{4d}\end{bmatrix}-Ma\right). \quad (10)$$

where $\dddot{y}_{id}$ ($i$=1,2,3,4) represents the PD controller, which is detailed in Appendix A.

Our previous research [18] approximates the necessary and sufficient condition to receive an

$$\begin{aligned}&4.000 \cdot c1 \cdot c2 \cdot c3 \cdot c4 + 5.592 \cdot (+c1{\cdot}c2{\cdot}c3{\cdot}s4 - c1{\cdot}c2{\cdot}s3{\cdot}c4 + c1{\cdot}s2{\cdot}c3{\cdot}c4 - s1{\cdot}c2{\cdot}c3{\cdot}c4)\\ &+ 0.9716 \cdot (+c1{\cdot}c2{\cdot}s3{\cdot}s4 + c1{\cdot}s2{\cdot}s3{\cdot}c4 + s1{\cdot}c2{\cdot}c3{\cdot}s4 + s1{\cdot}s2{\cdot}c3{\cdot}c4) + 2.000\\ &\cdot (-c1{\cdot}s2{\cdot}c3{\cdot}s4 - s1{\cdot}c2{\cdot}s3{\cdot}c4) + 0.1687\\ &\cdot (-c1{\cdot}s2{\cdot}s3{\cdot}s4 + s1{\cdot}c2{\cdot}s3{\cdot}s4 - s1{\cdot}s2{\cdot}c3{\cdot}s4 + s1{\cdot}s2{\cdot}s3{\cdot}c4)\\ &\neq 0.\end{aligned} \tag{11}$$

invertible decoupling matrix, given non-zero angular velocities in the propellers. That is

Once the gait (combination of the tilting angles) satisfies (11), the decoupling matrix in feedback linearization is asserted to be invertible in our case, given $\phi \approx 0, \theta \approx 0, \psi \approx 0$.

The next question is how to control the remaining degrees of freedom, $X$ and $Y$ in position.

The conventional quadrotor tracks the position by adjusting its attitude based on the conventional attitude-position decoupler [16,24,26]. This decoupler, however, may not be valid for a tilt-rotor. Our previous research [19] deduced the modified attitude-position decoupler for the tilt-rotor,

$$\phi = \frac{1}{g} \cdot \left(\ddot{X}{\cdot}s\psi - \ddot{Y}{\cdot}c\psi\right) + \frac{F_Y}{mg}. \tag{12}$$

$$\theta = \frac{1}{g} \cdot \left(\ddot{X}{\cdot}c\psi + \ddot{Y}{\cdot}s\psi\right) - \frac{F_X}{mg}. \tag{13}$$

where $F_X$ and $F_Y$ are defined by

$$\begin{bmatrix} F_X \\ F_Y \end{bmatrix} = K_f \cdot \begin{bmatrix} 0 & s2 & 0 & -s4 \\ s1 & 0 & -s3 & 0 \end{bmatrix} \cdot \begin{bmatrix} I_B^{-1} \cdot \tau(\alpha) \\ \frac{K_f}{m} \cdot \begin{bmatrix} 0 & 0 & 1 \end{bmatrix} \cdot F(\alpha) \end{bmatrix}^{-1} \cdot \begin{bmatrix} 0 \\ 0 \\ 0 \\ g \end{bmatrix}. \tag{14}$$

One of the comparisons we will make in the tracking result is the simulations equipped with the conventional attitude-position decoupler and with the modified attitude-position decoupler.

*3.2. Symmetrical and Asymmetrical Cat-Inspired Gait*

The gait for a tilt-rotor is defined as the combination of time-specified tilting angles. Our previous research [19] deployed the cat-trot inspired gait, which leads to the invertible decoupling matrix, satisfying (11).

In this research, several other common cat gaits are discussed before being modified to accommodate (11) and being applied to the tilt-rotor-gait plan.

The cat gaits can be classified as symmetrical gaits and asymmetrical gaits. The footfalls in the former gaits touches the ground at evenly spaced interval of time, which is not the case for the latter gaits [20]. Typical symmetrical gaits include walk, run, and trot. While transverse gallop and rotary gallop are asymmetrical gaits. The target gaits within the scope of this research are walk, run, transverse gallop, and rotary gallop.

These four gaits, given that the period is 1 s, can be approximated (interpolation) in Figures 2–5, respectively. The abscissa represents the time in a period (0 to 1 s). The ordinate represents the relevant tilting angles, $\alpha_1$, $\alpha_2$, $\alpha_3$, and $\alpha_4$, at the corresponding given time point.

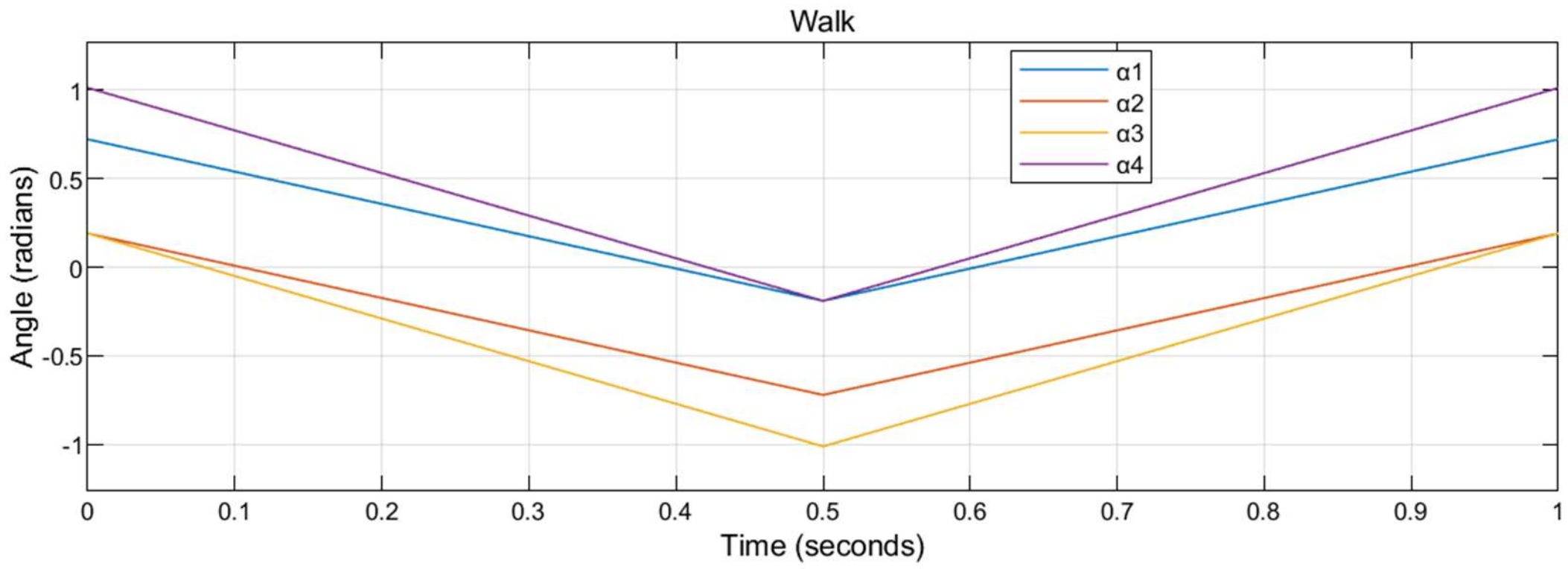

**Figure 2.** The walk gait of a cat. The period is set as 1 s.

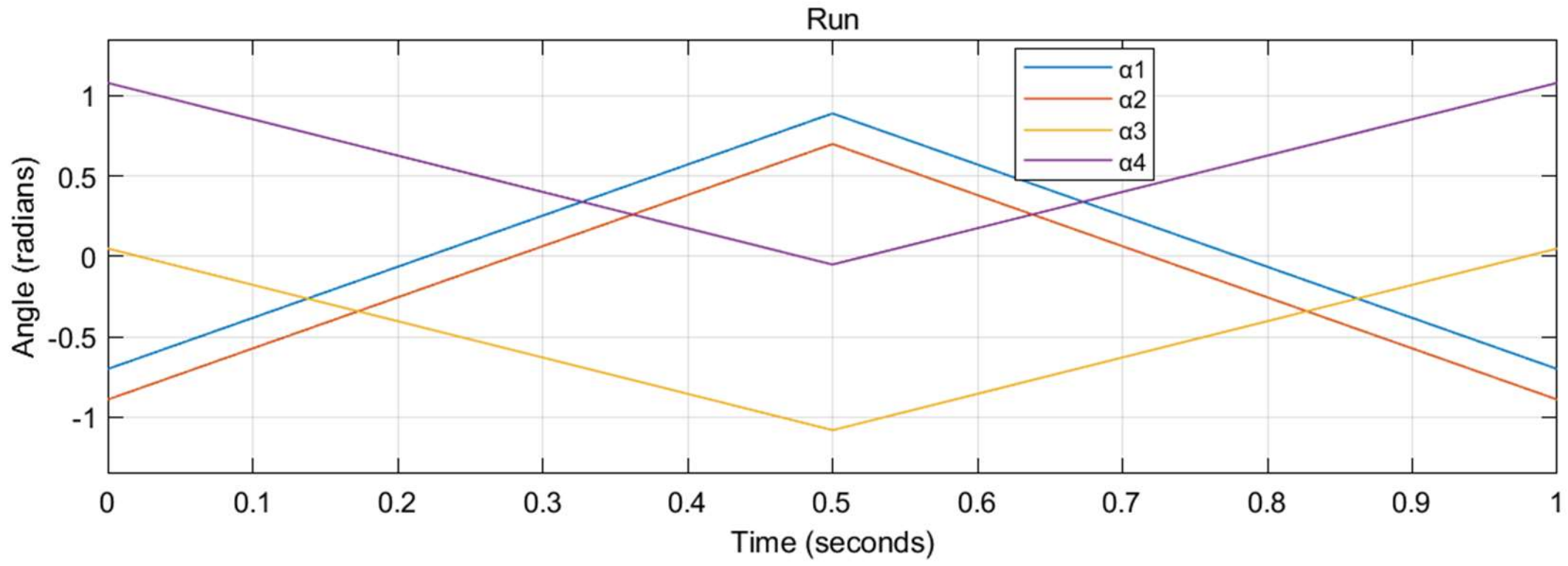

**Figure 3.** The run gait of a cat. The period is set as 1 s.

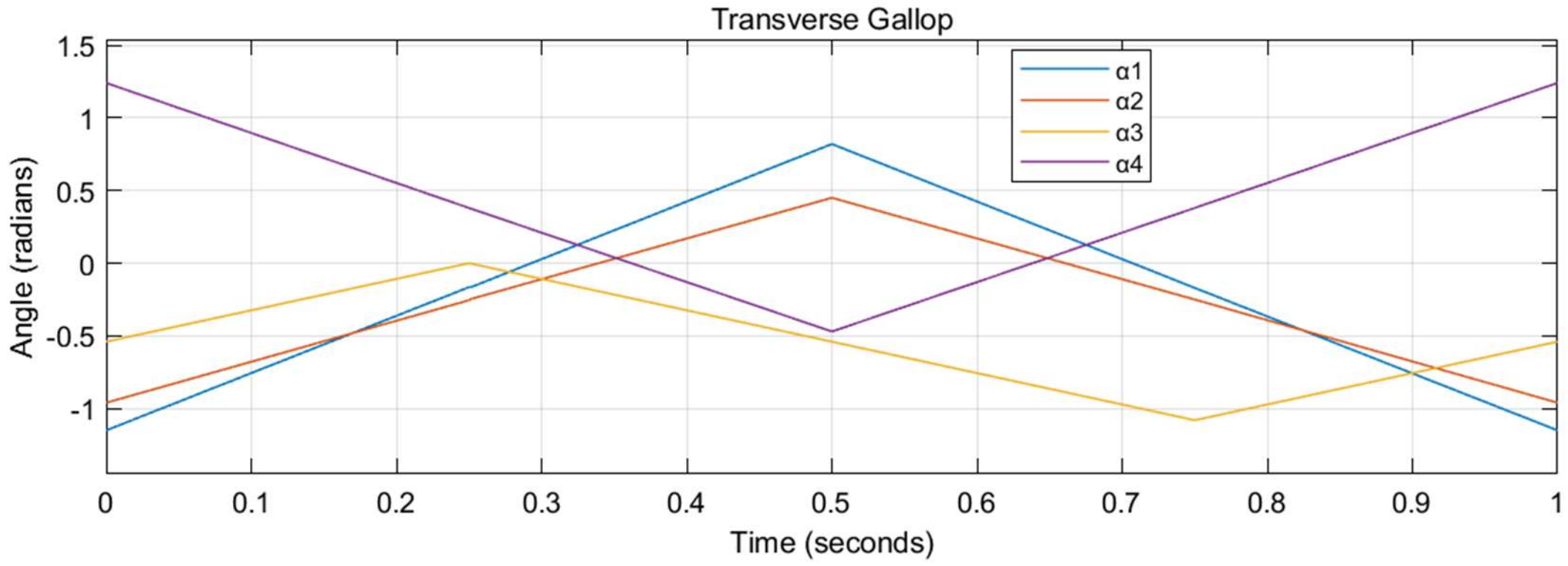

**Figure 4.** The transverse gallop gait of a cat. The period is set as 1 s.

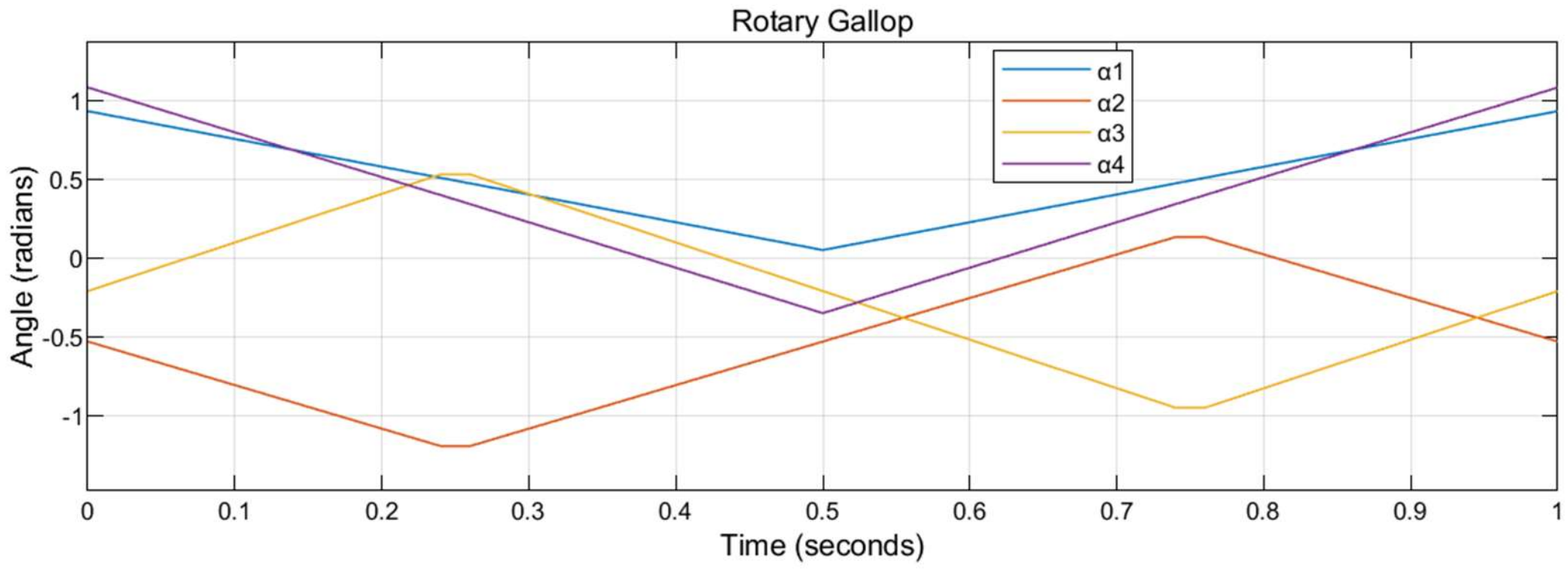

**Figure 5.** The rotary gallop gait of a cat. The period is set as 1 s.

Obviously, the condition (11) may not hold for the gaits designed, leading to a singular decoupling matrix. The problems referring to singularities are discussed in the next section.

## 4. Singular Decoupling Matrix and Gait Modification

The decoupling matrix is required to be invertible while applying feedback linearization. This section discusses the singularity of the decoupling matrix.

The preliminary condition to receive an invertible decoupling matrix is given in (11). However, satisfying (11) does not necessarily mean that the decoupling matrix is invertible.

One may notice that satisfying (11) can also encounter zero angular velocities of the propellers, leading to a singular decoupling matrix. There is other research focusing on the bound-avoidance of the inputs/states [27], which is beyond the scope of this study.

Also, notice that the necessary and sufficient condition in (11) is an approximation at $\phi = 0, \theta = 0$. On the other hand, as explained in our previous research [19], the state $\phi = 0, \theta = 0$ is not a typical equilibrium state. This causes us to visualize the attitudes with the risk of introducing a singular decoupling matrix.

The exact necessary and sufficient condition [18] to receive an invertible decoupling matrix is

$$\begin{aligned}
&1.000 \cdot c1 \cdot c2 \cdot c3 \cdot s4 \cdot s\theta - 1.000 \cdot c1 \cdot s2 \cdot c3 \cdot c4 \cdot s\theta - 2.880 \cdot c1 \cdot c2 \cdot s3 \cdot s4 \cdot s\theta + 2.880 \cdot c1 \cdot s2 \\
&\cdot s3 \cdot c4 \cdot s\theta - 2.880 \cdot s1 \cdot c2 \cdot c3 \cdot s4 \cdot s\theta + 2.880 \cdot s1 \cdot s2 \cdot c3 \cdot c4 \cdot s\theta - 1.000 \cdot s1 \cdot c2 \cdot s3 \cdot s4 \cdot s\theta \\
&+ 1.000 \cdot s1 \cdot s2 \cdot s3 \cdot c4 \cdot s\theta + 4.000 \cdot c1 \cdot c2 \cdot c3 \cdot c4 \cdot c\phi \cdot c\theta + 5.592 \cdot c1 \cdot c2 \cdot c3 \cdot s4 \cdot c\phi \cdot c\theta \\
&- 5.592 \cdot c1 \cdot c2 \cdot s3 \cdot c4 \cdot c\phi \cdot c\theta + 5.592 \cdot c1 \cdot s2 \cdot c3 \cdot c4 \cdot c\phi \cdot c\theta - 5.592 \cdot s1 \cdot c2 \cdot c3 \cdot c4 \cdot c\phi \cdot c\theta \\
&+ 1.000 \cdot c1 \cdot c2 \cdot s3 \cdot c4 \cdot s\phi \cdot c\theta + 0.9716 \cdot c1 \cdot c2 \cdot s3 \cdot s4 \cdot c\phi \cdot c\theta - 2.000 \cdot c1 \cdot s2 \cdot c3 \cdot s4 \cdot c\phi \cdot c\theta \\
&+ 0.9716 \cdot c1 \cdot s2 \cdot s3 \cdot c4 \cdot c\phi \cdot c\theta - 1.000 \cdot s1 \cdot c2 \cdot c3 \cdot c4 \cdot s\phi \cdot c\theta + 0.9716 \cdot s1 \cdot c2 \cdot c3 \cdot s4 \cdot c\phi \cdot c\theta \\
&- 2.000 \cdot s1 \cdot c2 \cdot s3 \cdot c4 \cdot c\phi \cdot c\theta + 0.9716 \cdot s1 \cdot s2 \cdot c3 \cdot c4 \cdot c\phi \cdot c\theta + 2.880 \cdot c1 \cdot c2 \cdot s3 \cdot s4 \cdot s\phi \cdot c\theta \\
&+ 2.880 \cdot c1 \cdot s2 \cdot s3 \cdot c4 \cdot s\phi \cdot c\theta - 0.1687 \cdot c1 \cdot s2 \cdot s3 \cdot s4 \cdot c\phi \cdot c\theta - 2.880 \cdot s1 \cdot c2 \cdot c3 \cdot s4 \cdot s\phi \cdot c\theta \\
&+ 0.1687 \cdot s1 \cdot c2 \cdot s3 \cdot s4 \cdot c\phi \cdot c\theta - 2.880 \cdot s1 \cdot s2 \cdot c3 \cdot c4 \cdot s\phi \cdot c\theta - 0.1687 \cdot s1 \cdot s2 \cdot c3 \cdot s4 \cdot c\phi \cdot c\theta \\
&+ 0.1687 \cdot s1 \cdot s2 \cdot s3 \cdot c4 \cdot c\phi \cdot c\theta - 1.000 \cdot c1 \cdot s2 \cdot s3 \cdot s4 \cdot s\phi \cdot c\theta + 1.000 \cdot s1 \cdot s2 \cdot c3 \cdot s4 \cdot s\phi \cdot c\theta \\
&\neq 0.
\end{aligned} \tag{15}$$

Once $\alpha_i (i=1,2,3,4)$ is determined, the unacceptable attitude can be found in $\phi - \theta$ plane by equaling the left side of (15) to 0. The unqualified gaits, leading to the invertible decoupling matrix only in a small attitude region, can be modified by scaling,

$$\alpha_i \leftarrow \frac{\alpha_i}{n}, i = 1,2,3,4, n \geqslant 2, n \in \mathbb{N}. \tag{16}$$

The tilting angles are then scaled by $1/n$ in this modification.

**Proposition 1.** *There will always be a positive integer n such that the modified gait by scaling by* $1/n$ *produces an invertible decoupling matrix.*

**Proof of Proposition s1.** For a sufficiently large $n$, we have

$$\lim_{n \to +\infty} \alpha_i = 0. \tag{17}$$

Substituting (17) into the left side of (15) yields

$$4.000 \cdot c\phi \cdot c\theta. \tag{18}$$

which is non-zero, given $\phi \in (-\pi/2, \pi/2), \theta \in (-\pi/2, \pi/2)$, satisfying (15). □

The walk gait, transverse gallop, and rotary gallop gait are scaled by 1/3, respectively, into Figures 6–8. Identical to the previous notations, the abscissa represents the time in a period (0 to 1 s). The ordinate represents the relevant tilting angles, $\alpha_1$, $\alpha_2$, $\alpha_3$, and $\alpha_4$, at the corresponding given time point.

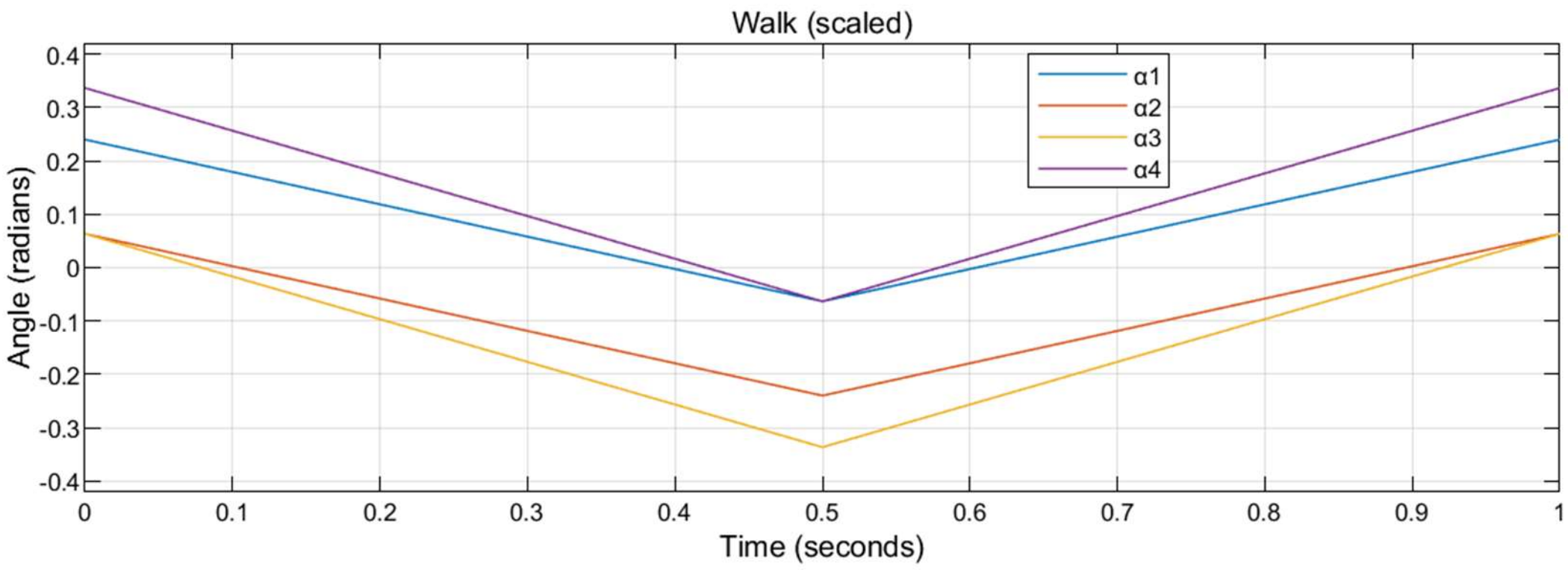

**Figure 6.** The scaled (1/3) walk gait of a cat. The period is set as 1 s.

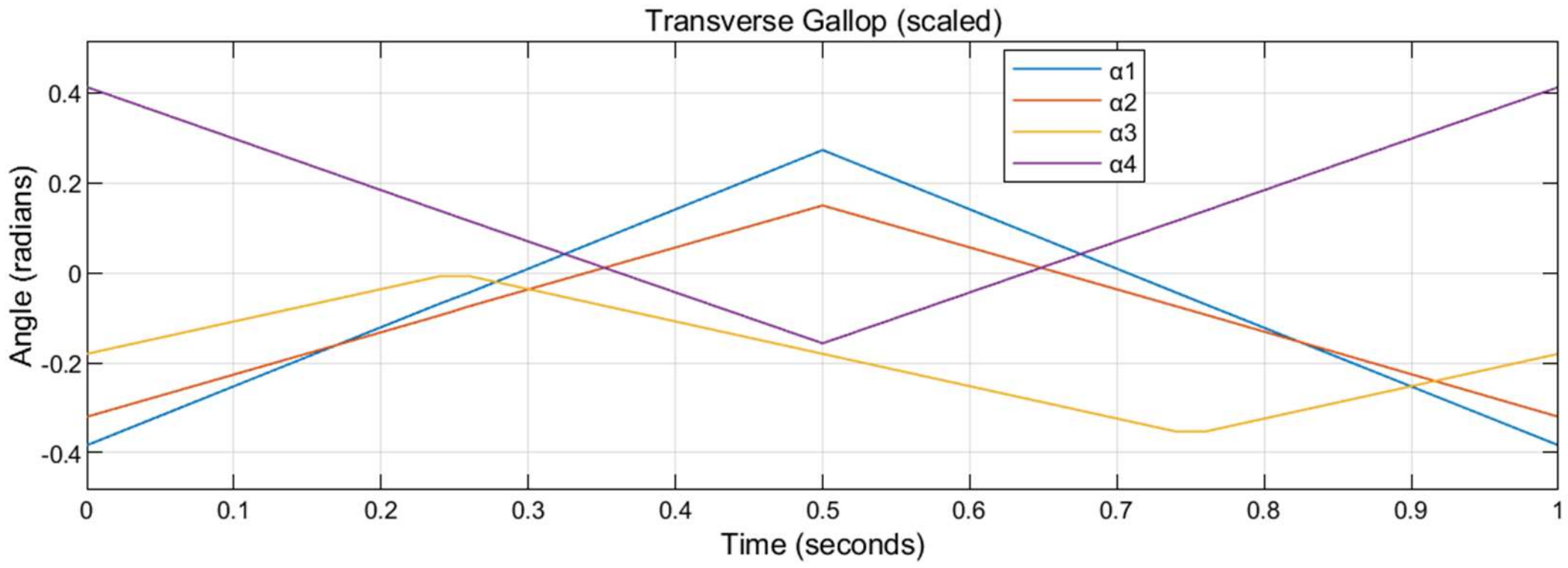

**Figure 7.** The scaled (1/3) transverse gallop gait of a cat. The period is set as 1 s.

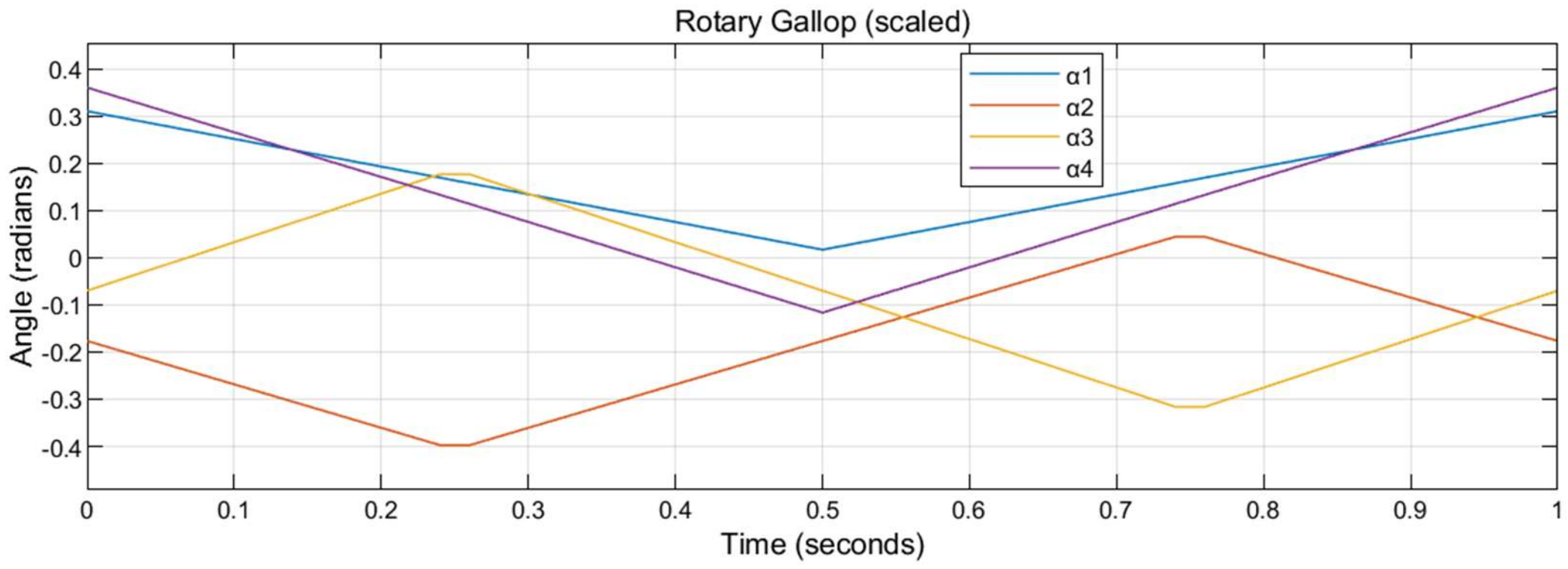

**Figure 8.** The scaled (1/3) rotary gallop gait of a cat. The period is set as 1 s.

For a given gait, four time-specified tilting angles define the unacceptable attitudes as the attitudes that violate Formula (15); these unacceptable attitudes lead the left side of Formula (15) to zero.

Equaling the left side of Formula (15) to zero, the unacceptable attitudes, $\phi$ and $\theta$, can be tracked by finding the roots, given a determined gait. The unacceptable attitudes in $\phi - \theta$ plane for different scaled walk gaits are plotted in Figure 9.

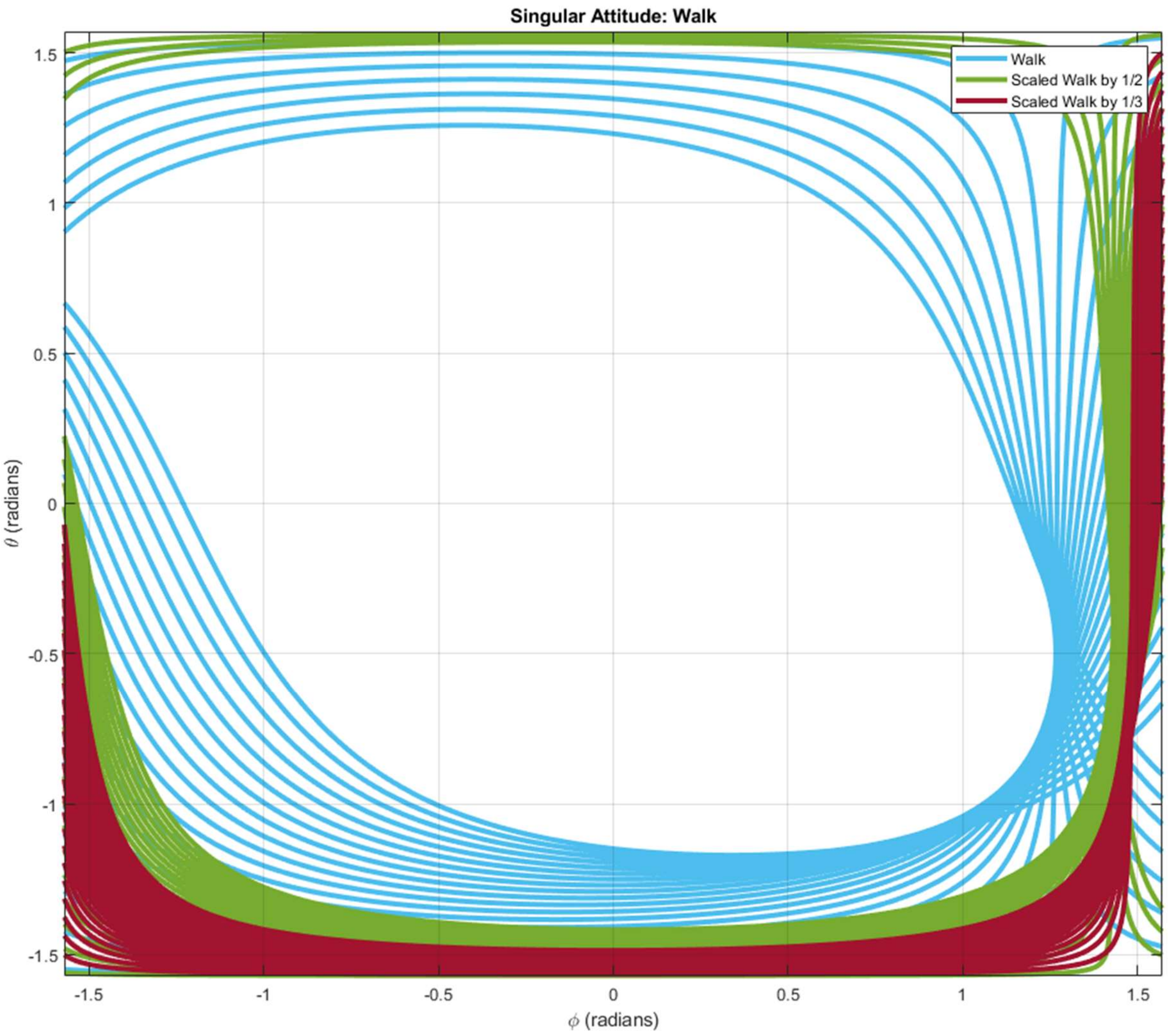

**Figure 9.** The attitudes introducing the singular decoupling matrix. The blue curve represents the attitudes adopting the walk gait. The green and brown curves represent the attitudes adopting the scaled (1/2 and 1/3, respectively) walk gait.

Since the attitudes of most flights are near $(\phi,\theta) = (0,0)$, we evaluate the quality of a gait by finding the distance between $(\phi,\theta) = (0,0)$ and the closest curve of the unacceptable attitudes. For a gait receiving a long distance between (0,0) and the closest curve of the unacceptable attitudes, the tilt-rotor has a large acceptable attitude region. A gait receiving a short distance between (0,0) and the closest curve of the unacceptable attitudes in the roll-pitch diagram provides a small acceptable attitude region.

Thus, a gait with a larger distance between (0,0) and the curve of the unacceptable attitudes is more robust to the attitude; the acceptable attitude region for the tilt-rotor adopting this gait is wider.

It can be found that the acceptable attitude region is enlarging while evenly shrinking the scale of the walk gait. A similar result can also be notably observed in scaled transverse gallop gaits (Figure 10) and scaled rotary gallop gaits (Figure 11).

The unreferred gait, run gait, remains identical to the original cat-run gait.

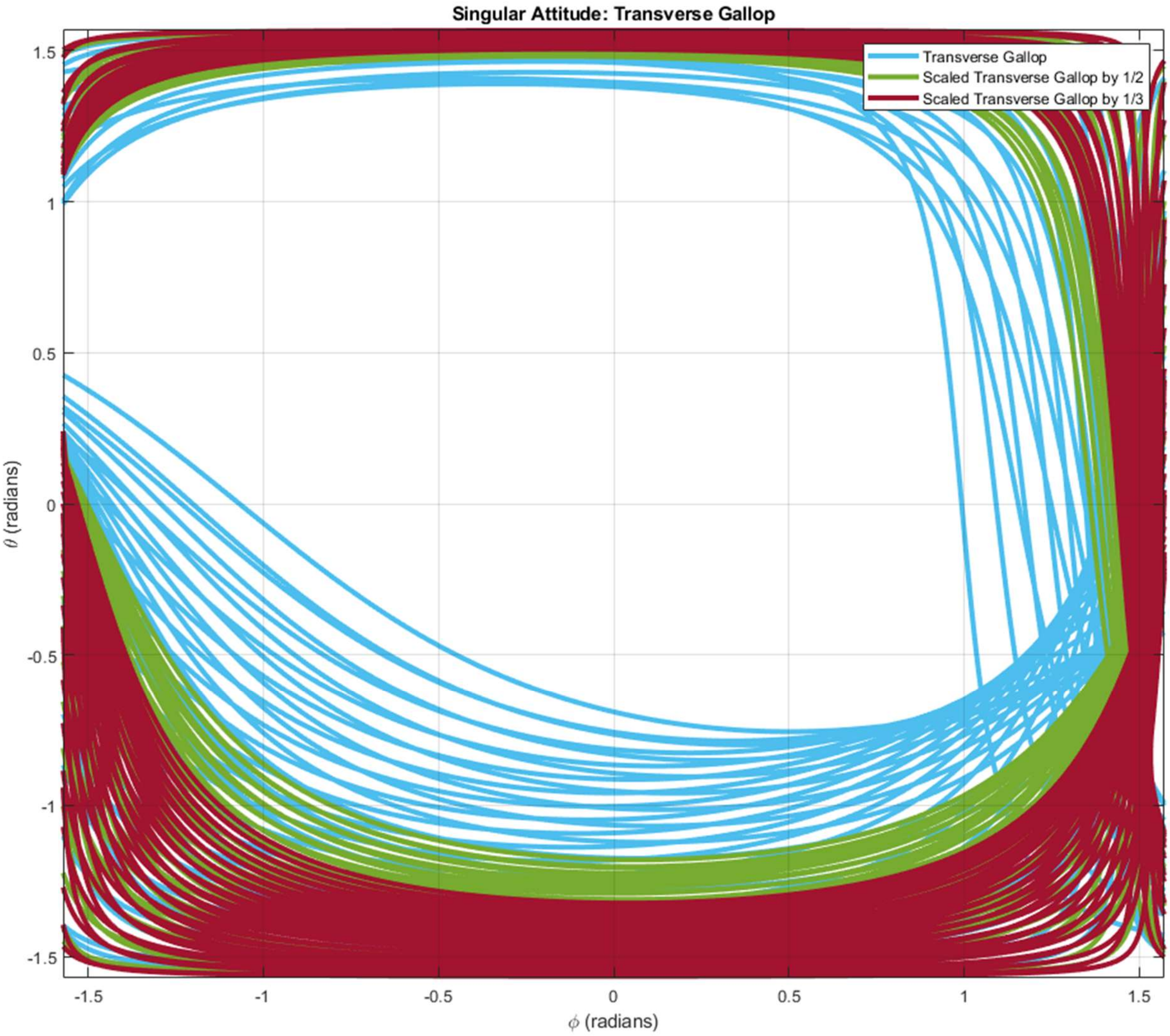

**Figure 10.** The attitudes introducing the singular decoupling matrix. The blue curve represents the attitudes adopting the transverse gallop gait. The green and brown curves represent the attitudes adopting the scaled (1/2 and 1/3, respectively) transverse gallop gait.

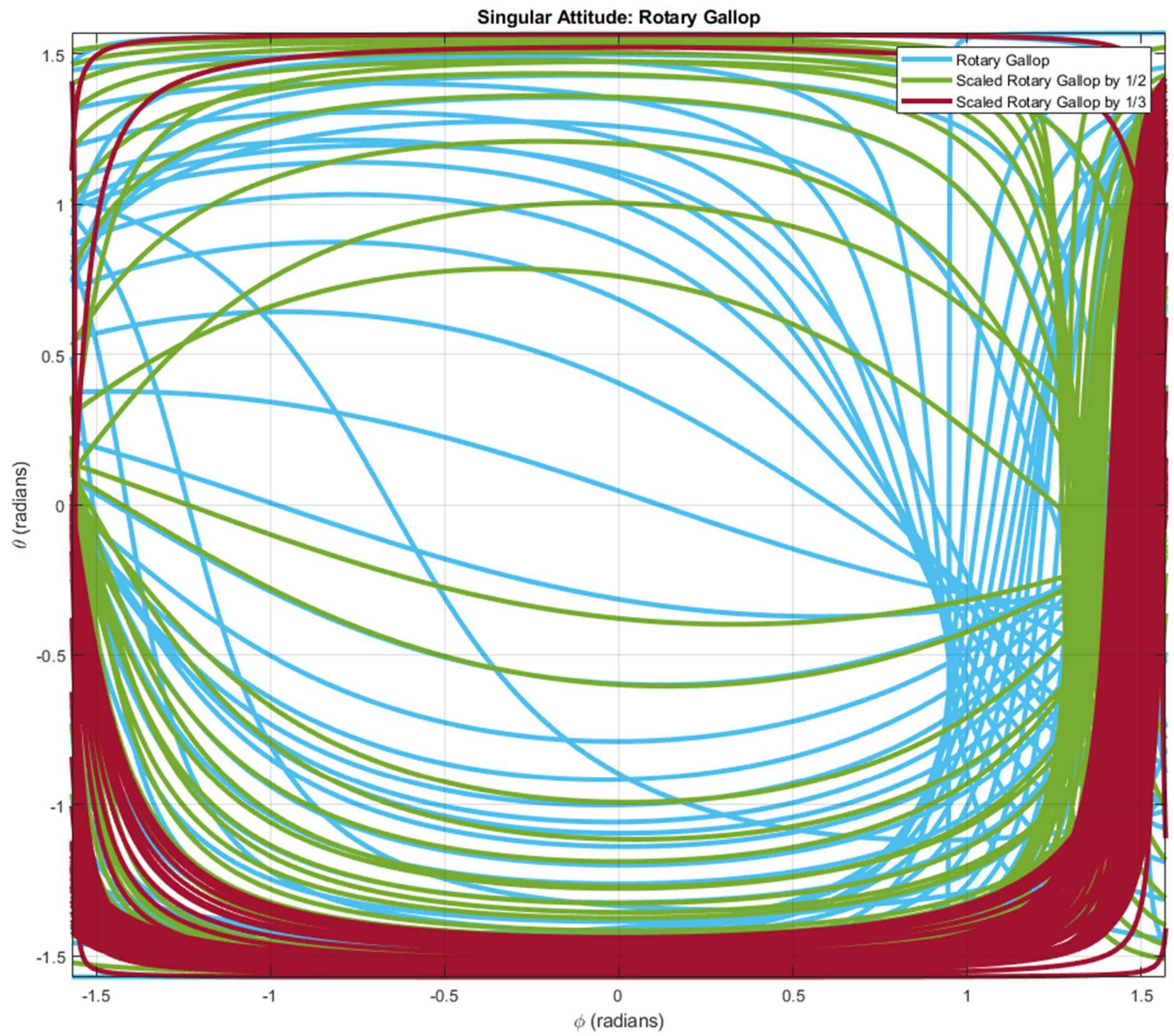

**Figure 11.** The attitudes introducing the singular decoupling matrix. The blue curve represents the attitudes adopting the rotary gallop gait. The green and brown curves represent the attitudes adopting the scaled (1/2 and 1/3, respectively) rotary gallop gait.

## 5. Simulation Settings

Similar to our previous research [19], the references set in this research are the uniform rectilinear motion and the uniform circular motion with zero yaw and zero altitude ($\psi_r = 0, Z_r = 0$), which are specified as

$$\begin{cases} X_r = 1.5 \cdot t \\ Y_r = 1.5 \cdot t \end{cases}. \tag{19}$$

$$\begin{cases} X_r = 5 \cdot \cos(0.1 \cdot t) \\ Y_r = 5 \cdot \sin(0.1 \cdot t) \end{cases}. \tag{20}$$

We adopt this reference since this speed accommodates the cat-trot gait only, which is not preferable to any of the gaits analyzed in this study. The tilt-rotor is then required to track these unbiased references with the relevant gaits.

The absolute value of each initial angular velocity was 300 (rad/s). Each gait is adopted with three different periods (1 (s), 2 (s), and 3 (s)) to track these two references. In addition, the conventional attitude-position decoupler and the modified attitude-position decoupler for a tilt-rotor [19] are both tested and compared.

The supremum of the dynamic state error in position, after sufficient time, is defined as

$$\max_{t \geqslant 20s} \|e(t)\| \tag{21}$$

where $e(t)$ is the dynamic state error in position defined as

$$e = \sqrt{{e_X}^2 + {e_Y}^2} \tag{22}$$

where $e_X$ and $e_Y$ represent the dynamic state errors along $X$ and $Y$ directions, respectively.

The supremum of the dynamic state error in position, after sufficient time, is recorded and compared in each test.

The simulation is conducted in Simulink, MATLAB. Ode3 with sampling time 0.001 (s) is adopted in our solver.

**6. Results**

For the first reference, uniform rectilinear motion, the suprema of the dynamic state error (unit: meter), after sufficient time, for different gaits and different periods are concluded in Figure 12.

The results are classified into 3 sections based on the period of the gaits. They are 1 s (gray), 2 s (red), and 3 s (yellow), respectively. The axis W, R, TG, and RG represents walk gait, run gait, transverse gallop gait, and rotary gallop gait, respectively. The vertexes of the outer quadrilateral (blue) represent the result equipped with the conventional attitude-position decoupler. The vertexes of the inner quadrilateral (purple) represent the result equipped with the modified attitude-position decoupler, (12) and (13).

For example, the "2.47" on the "W" axis on the outer quadrilateral (blue) in "Period: 1 (s)" in Figure 12 means that the supremum of the dynamic state error, after sufficient time, is 2.47 m, adopting the walk-inspired gait with a period of 1 s equipped with the conventional attitude-position decoupler. The "0.19" on the "W" axis on the inner quadrilateral (purple) in "Period: 1 (s)" in Figure 12 means that the supremum of the dynamic state error, after sufficient time, is 0.19 m, adopting the walk-inspired gait with a period of 1 s equipped with the modified attitude-position decoupler.

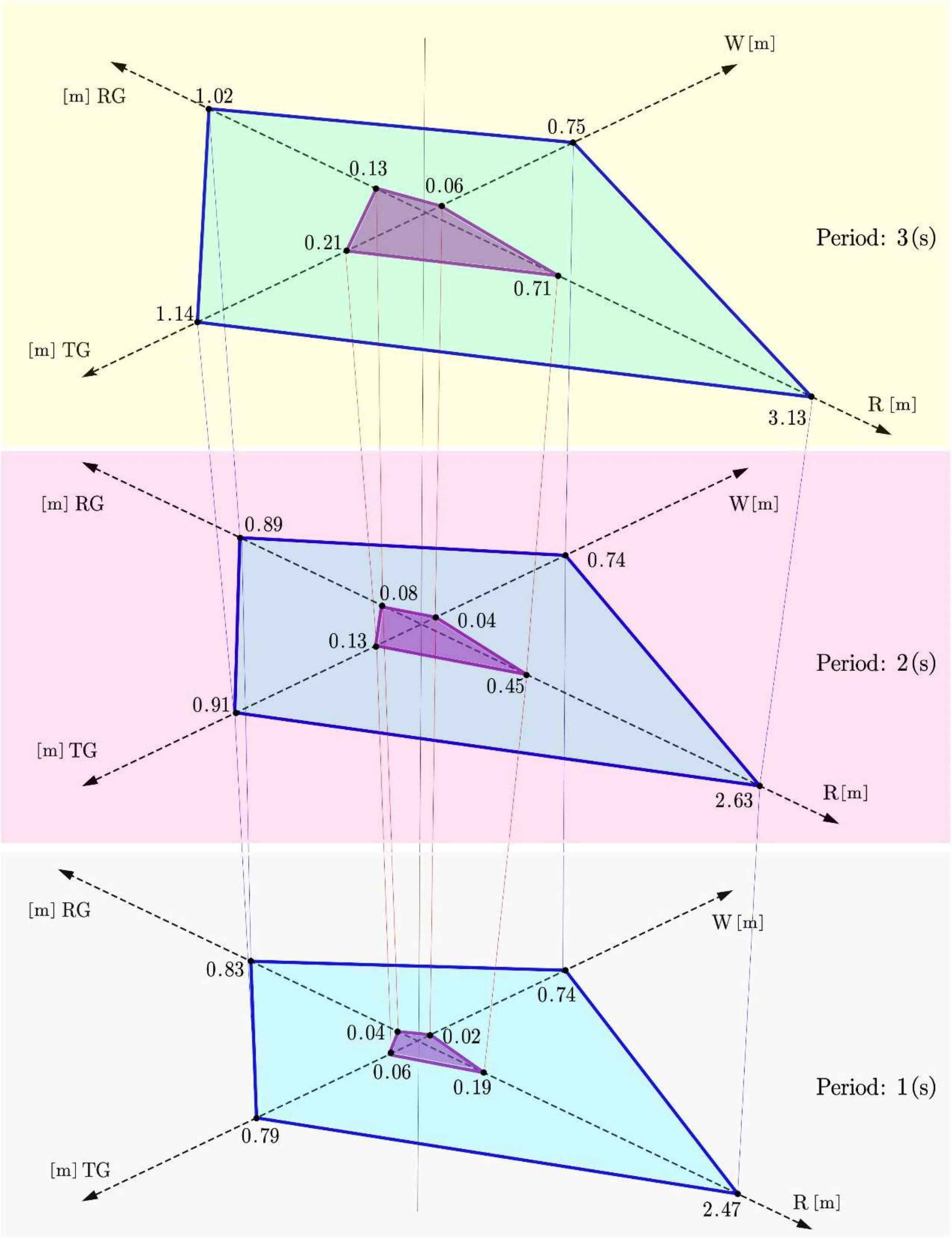

**Figure 12.** The result of the simulation with the uniform-rectilinear-motion reference: supremum of the dynamic state error (unit: m) after sufficient time for three different time periods. W: walk gait. R: run gait. TG: transverse gallop gait. RG: rotary gallop gait. Outer quadrilateral (blue): equip with the conventional attitude-position decoupler. Inner quadrilateral (purple): equip with the modified attitude-position decoupler.

In addition, the result of the simulation with the uniform circular motion is plotted in Figure 13.

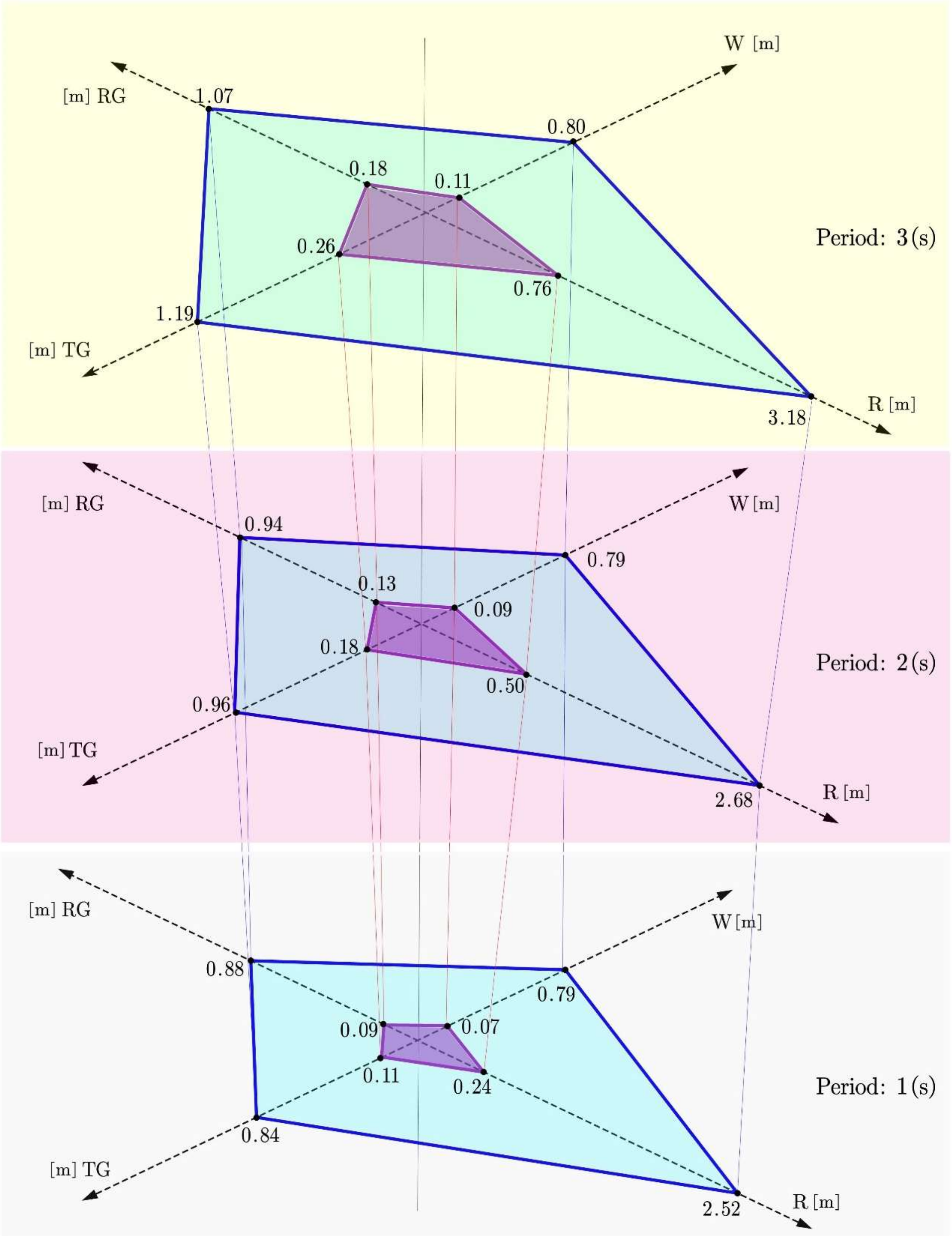

**Figure 13.** The result of the simulation with the uniform-circular-motion reference: supremum of the dynamic state error (unit: m) after sufficient time for three different time periods. W: walk gait. R: run gait. TG: transverse gallop gait. RG: rotary gallop gait. Outer quadrilateral (blue): equip with the conventional attitude-position decoupler. Inner quadrilateral (purple): equip with the modified attitude-position decoupler.

The same notation rule is adopted. Specifically, the results are classified into 3 sections based on the period of the gaits. They are 1 s (gray), 2 s (red), and 3 s (yellow), respectively. The axis W, R, TG, and RG represents walk gait, run gait, transverse gallop gait, and rotary gallop gait, respectively. The vertexes of the outer quadrilateral (blue) represent the result equipped with the conventional attitude-position decoupler. The vertexes of the inner quadrilateral (purple) represents the result equipped with the modified attitude-position decoupler, (12) and (13).

It can be clearly seen that the inner purple quadrilateral is much smaller than the outer blue quadrilateral, which demonstrates that our modified attitude-position decoupler significantly decreases the dynamic state error for all the gaits discussed in this research. This novel attitude-position decoupler is particularly effective for the case suffering from a large dynamic state error equipped with the conventional attitude-position decoupler.

Another interesting result is that the choice of the period influences the dynamic state error. The longer the period is, the larger dynamic state error there tends to be in walk gait, run gait, transverse gallop gait, and rotary gallop gait. The result of the walk gaits is insignificantly influenced by the period settings, while the result in the run gait highly relies on the period settings.

**7. Conclusions and Discussions**

The four cat gaits, walk gait, run gait, transverse gallop gait, and rotary gallop gait are feasible to be transplanted to solve the gait plan problem for a tilt-rotor. However, walk gait, transverse gallop gait, and rotary gallop gait are required to be modified, e.g., scaling, before being adopted.

The previous research proved that the cat-trot-gait planned tilt-rotor receives the invertible decoupling matrix near zero attitude region analytically. The parallel analytical necessary and sufficient condition to receive a regular decoupling matrix is not straightforward for the four cat gaits, walk gait, run gait, transverse gallop gait, and rotary gallop gait. Thus, this research proposed the singular curves in the roll-pitch diagram to analyze the property of the decoupling matrix for the first time.

Before this article, no systematic methods were put forward to modify a gait, which is liable to introduce a singular decoupling matrix in a tilt-rotor. The scaling method in the gait modification is proven to be feasible in finding a valid gait, liable to lead to the invertible decoupling matrix for the first time; this scaling method tends to enlarge the acceptable attitude zone in the roll-pitch diagram, indicating that this method strengthens the relevant gait.

The modified gaits in this simulation show promising tracking results, where the dynamic state error is acceptable in the tracking problem. Further, the modified walk-inspired gait in this research receives the least suprema of the dynamic state error after sufficient time, while the run-inspired gait receives the largest suprema of the dynamic state error after sufficient time. As for the rest of the gaits analyzed in his research, the modified transverse gallop gait and rotary gallop gait receive similar suprema of the dynamic state error after sufficient time. Beware that the dynamic state error may also be influenced by the maximum tilting angle of each gait, which is not identical in different gaits in this research.

It is not surprising that the modified attitude-position decoupler significantly reduces the dynamic state error for the tilt-rotor for each gait analyzed in this research; we have witnessed similar results in previous research [19]. This research further verifies the effectiveness of the modified attitude-position decoupler.

The length of the period of the gaits influences the dynamic state error. In general, the longer the period is, the larger dynamic state error there tends to be. Elucidating the underlying mechanism is beyond the scope of this research.

The unique contributions of this research can be concluded as follows: 1. A numerical method (singular curves in the roll-pitch diagram) of analyzing the robustness of the gaits is put forward for the first time. 2. A novel method for modifying the unqualified gaits is created and proven feasible theoretically. 3. The four typical cat-inspired gaits (walk gait, run gait, transverse gallop gait, and rotary gallop gait) are modified to accommodate the tracking problem based on feedback linearization.

Table 1 compares the methods in analyzing the property of the decoupling matrix of the animal-inspired gaits (cat walk/run/transverse gallop/rotary gallop) in this research and of the cat-trot inspired gait in the previous research.

**Table 1.** Methods in analyzing the property of the decoupling matrix.

| Cat Trot | Cat Walk/Run/Transverse Gallop/Rotary Gallop |
|---|---|
| analytical conditions | numerical conditions |
| attitude is assumed zero | effects of the attitude are considered |

| - | put forward gait modification methods |
|---|---|

There are several points worth exploring further. Firstly, since the scaling method is the only approach in modifying the gaits in this research, there might be other effective methods with sound mathematical grounds.

Figures 14 and 15 display the angular velocity histories while tracking the rectilinear reference defined in Formula (19) with the walk gait scaled by 1/3 and 1/6, respectively. Though the periods of both gaits are identical (3 s), the scale of the tilting angles influences the resulting angular velocities of the propellers of the tilt-rotor. It can be asserted that the desired magnitude of the thrust of each propeller is influenced by the adopted gait.

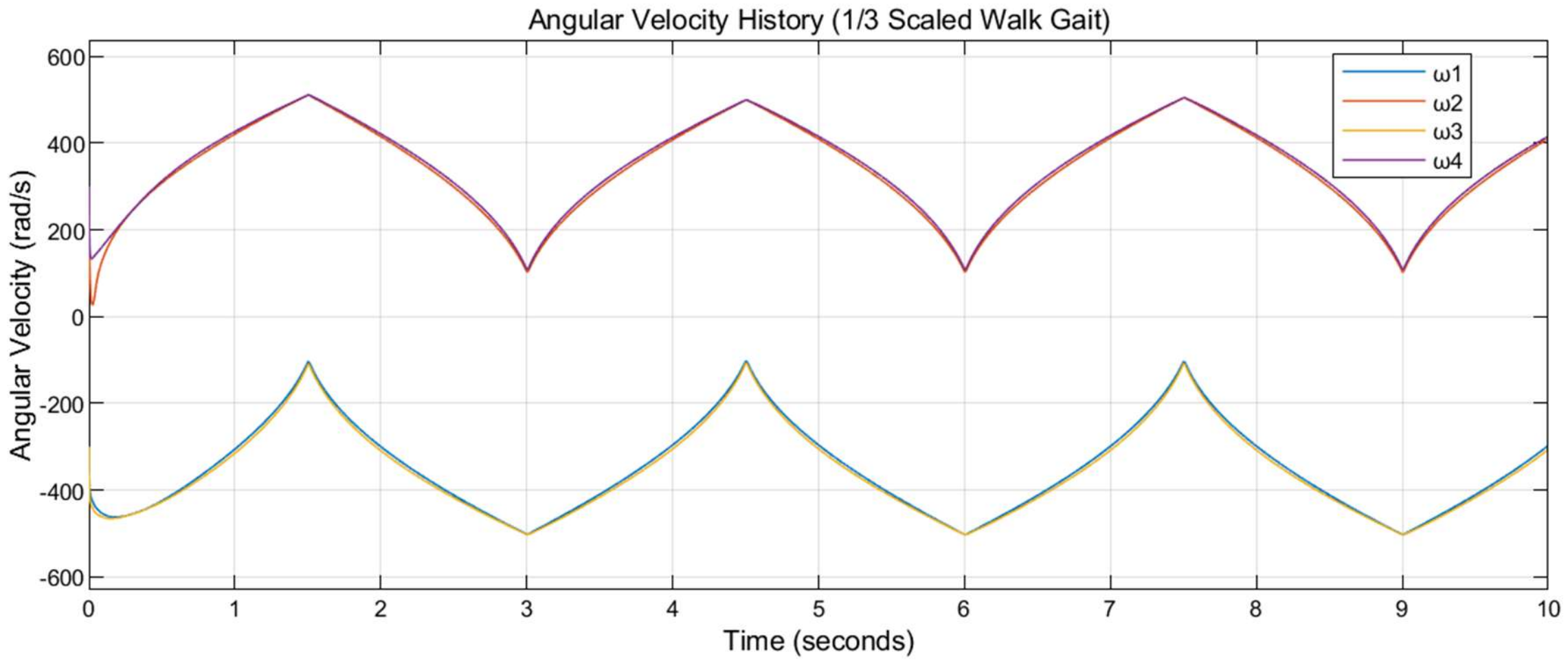

**Figure 14.** The angular velocities, $\varpi_1$, $\varpi_2$, $\varpi_3$, and $\varpi_4$, during the flight tracking the rectilinear reference. The scaled (1/3) walk gait is adopted with the period 3 s.

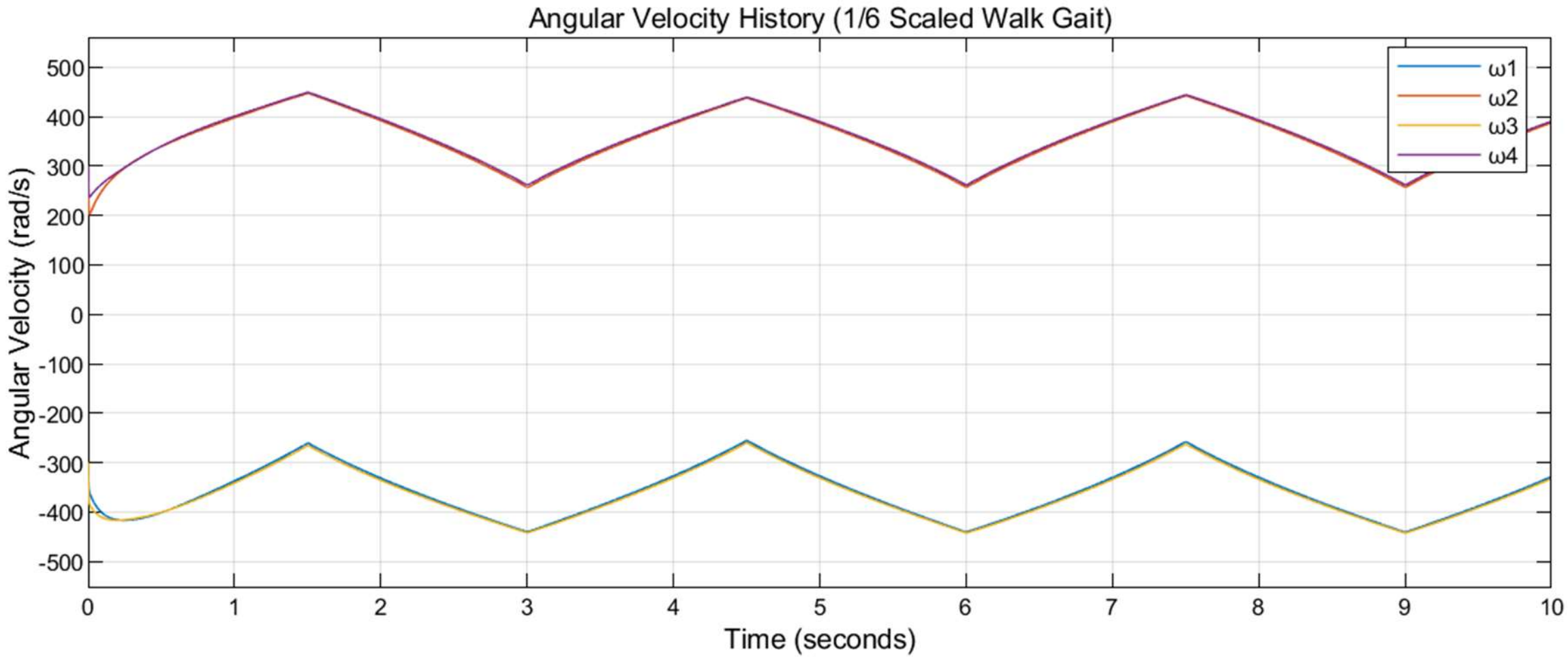

**Figure 15.** The angular velocities, $\varpi_1$, $\varpi_2$, $\varpi_3$, and $\varpi_4$, during the flight tracking the rectilinear reference. The scaled (1/6) walk gait is adopted with the period of 3 s.

It can also be found that the robustness of the tilt-rotor increases while scaling. This process, however, sacrifices the lateral force generated by the tilt-rotor. The trade-off between the scaling of the magnitude of the lateral force would be interesting to investigate during further research, as well as a real model to test this rule.

## Appendix A

This appendix details the PD controllers, $[\ddot{y}_{1d} \quad \ddot{y}_{2d} \quad \ddot{y}_{3d} \quad \ddot{y}_{4d}]^T$, in (10) as follows

$$\begin{bmatrix} \dddot{y}_{1d} \\ \dddot{y}_{2d} \\ \dddot{y}_{3d} \\ \dddot{y}_{4d} \end{bmatrix} = \begin{bmatrix} \dddot{y}_{1r} \\ \dddot{y}_{2r} \\ \dddot{y}_{3r} \\ \dddot{y}_{4r} \end{bmatrix} + K_1 \cdot \left( \begin{bmatrix} \ddot{y}_{1r} \\ \ddot{y}_{2r} \\ \ddot{y}_{3r} \\ \ddot{y}_{4r} \end{bmatrix} - \begin{bmatrix} \ddot{y}_1 \\ \ddot{y}_2 \\ \ddot{y}_3 \\ \ddot{y}_4 \end{bmatrix} \right) + K_2 \cdot \left( \begin{bmatrix} \dot{y}_{1r} \\ \dot{y}_{2r} \\ \dot{y}_{3r} \\ \dot{y}_{4r} \end{bmatrix} - \begin{bmatrix} \dot{y}_1 \\ \dot{y}_2 \\ \dot{y}_3 \\ \dot{y}_4 \end{bmatrix} \right) + K_3 \cdot \left( \begin{bmatrix} y_{1r} \\ y_{2r} \\ y_{3r} \\ y_{4r} \end{bmatrix} - \begin{bmatrix} y_1 \\ y_2 \\ y_3 \\ y_4 \end{bmatrix} \right) \tag{A1}$$

where $y_{ir}$ ($i$=1,2,3,4) represents the reference which is defined in Section 5, $K_1$, $K_2$, and $K_3$ are the control coefficients,

$$K_1 = \begin{bmatrix} 1 & & & \\ & 1 & & \\ & & 1 & \\ & & & 10 \end{bmatrix}, \tag{A2}$$

$$K_2 = \begin{bmatrix} 10 & & & \\ & 10 & & \\ & & 1 & \\ & & & 5 \end{bmatrix}, \tag{A3}$$

$$K_3 = \begin{bmatrix} 50 & & & \\ & 50 & & \\ & & 1 & \\ & & & 10 \end{bmatrix}. \tag{A4}$$

The stability proof can be found by substituting (A1)–(A4) into (10).

## References

1. Ryll, M.; Bulthoff, H.H.; Giordano, P.R. A novel overactuated quadrotor unmanned aerial vehicle: Modeling, control, and experimental validation. *IEEE Trans. Control Syst. Technol.* **2015**, *23*, 540–556. https://doi.org/10.1109/TCST.2014.2330999.
2. Jin, S.; Bak, J.; Kim, J.; Seo, T.; Kim, H.S. Switching PD-based sliding mode control for hovering of a tilting-thruster underwater robot. *PLoS ONE* **2018**, *13*, e0194427. https://doi.org/10.1371/journal.pone.0194427.
3. Kumar, R.; Sridhar, S.; Cazaurang, F.; Cohen, K.; Kumar, M. *Reconfigurable Fault-Tolerant Tilt-Rotor Quadcopter System.*; American Society of Mechanical Engineers: Atlanta, GA, USA, 2018; p. V003T37A008.
4. Anderson, R.B.; Marshall, J.A.; L'Afflitto, A. constrained robust model reference adaptive control of a tilt-rotor quadcopter pulling an unmodeled cart. *IEEE Trans. Aerosp. Electron. Syst.* **2021**, *57*, 39–54. https://doi.org/10.1109/TAES.2020.3008575.
5. Invernizzi, D.; Giurato, M.; Gattazzo, P.; Lovera, M. Comparison of control methods for trajectory tracking in fully actuated unmanned aerial vehicles. *IEEE Trans. Control Syst. Technol.* **2021**, *29*, 1147–1160. https://doi.org/10.1109/TCST.2020.2992389.
6. Invernizzi, D.; Lovera, M. Trajectory tracking control of thrust-vectoring UAVs. *Automatica* **2018**, *95*, 180–186. https://doi.org/10.1016/j.automatica.2018.05.024.
7. Imamura, A.; Miwa, M.; Hino, J. Flight characteristics of quad rotor helicopter with thrust vectoring equipment. *J. Robot. Mechatron.* **2016**, *28*, 334–342. https://doi.org/10.20965/jrm.2016.p0334.
8. Ryll, M.; Bulthoff, H.H.; Giordano, P.R. First flight tests for a quadrotor UAV with tilting propellers. In Proceedings of the 2013 IEEE International Conference on Robotics and Automation, Karlsruhe, Germany, 6–8 May 2013; IEEE: Karlsruhe, Germany ,2013; pp. 295–302.
9. Michieletto, G.; Cenedese, A.; Zaccarian, L.; Franchi, A. Hierarchical nonlinear control for multi-rotor asymptotic stabilization based on zero-moment direction. *Automatica* **2020**, *117*, 108991. https://doi.org/10.1016/j.automatica.2020.108991.
10. Nemati, A.; Kumar, M. Modeling and control of a single axis tilting quadcopter. In Proceedings of the 2014 American Control Conference, Portland, OR, USA, 4–6 June 2014; IEEE: Portland, OR, USA; pp. 3077–3082.
11. Jin, S.; Kim, J.; Kim, J.-W.; Bae, J.; Bak, J.; Kim, J.; Seo, T. Back-stepping control design for an underwater robot with tilting thrusters. In Proceedings of the 2015 International Conference on Advanced Robotics (ICAR), Istanbul, Turkey, 27–31 July 2015; IEEE: Istanbul, Turkey; pp. 1–8.
12. Ryll, M.; Bulthoff, H.H.; Giordano, P.R. Modeling and control of a quadrotor UAV with tilting propellers. In Proceedings of the 2012 IEEE International Conference on Robotics and Automation, St Paul, MN, USA, 14–18 May 2012; IEEE: St Paul, MN, USA; pp. 4606–4613.
13. Shen, Z.; Ma, Y.; Tsuchiya, T. Stability analysis of a feedback-linearization-based controller with saturation: A tilt vehicle with the penguin-inspired gait plan. *ArXiv Prepr* **2021**, ArXiv*:* 211114456.

14. Franchi, A.; Carli, R.; Bicego, D.; Ryll, M. Full-pose tracking control for aerial robotic systems with laterally bounded input force. *IEEE Trans. Robot.* **2018**, *34*, 534–541. https://doi.org/10.1109/TRO.2017.2786734.
15. Shen, Z.; Tsuchiya, T. State drift and gait plan in feedback linearization control of a tilt vehicle. In Proceedings of the Computer Science & Information Technology (CS & IT), Vienna, Austria, 2022; Academy & Industry Research Collaboration Center (AIRCC): Vienna, Austria; pp. 1–17.
16. Kumar, R.; Nemati, A.; Kumar, M.; Sharma, R.; Cohen, K.; Cazaurang, F. *Tilting-Rotor Quadcopter for Aggressive Flight Maneuvers Using Differential Flatness Based Flight Controller*; American Society of Mechanical Engineers: Tysons, VA, USA, 2017; p. V003T39A006.
17. Hamandi, M.; Usai, F.; Sablé, Q.; Staub, N.; Tognon, M.; Franchi, A. design of multirotor aerial vehicles: A taxonomy based on input allocation. *Int. J. Robot. Res.* **2021**, *40*, 1015–1044. https://doi.org/10.1177/02783649211025998.
18. Shen, Z.; Tsuchiya, T. Gait analysis for a tiltrotor: The dynamic invertible gait. *Robotics* **2022**, *11*, 33. https://doi.org/10.3390/robotics11020033.
19. Shen, Z.; Ma, Y.; Tsuchiya, T. Feedback linearization based tracking control of a tilt-rotor with cat-trot gait plan. *ArXiv* **2022**, ArXiv*:* 2202.02926 *CsRO*.
20. Vilensky, J.A.; Njock Libii, J.; Moore, A.M. Trot-gallop gait transitions in quadrupeds. *Physiol. Behav.* **1991**, *50*, 835–842. https://doi.org/10.1016/0031-9384(91)90026-K.
21. Luukkonen, T. Modelling and control of quadcopter. *Indep. Res. Proj. Appl. Math. Espoo* **2011**, *22*, 22.
22. Goodarzi, F.A.; Lee, D.; Lee, T. Geometric adaptive tracking control of a quadrotor unmanned aerial vehicle on SE(3) for agile maneuvers. *J. Dyn. Syst. Meas. Control* **2015**, *137*, 091007. https://doi.org/10.1115/1.4030419.
23. Shi, X.-N.; Zhang, Y.-A.; Zhou, D. A geometric approach for quadrotor trajectory tracking control. *Int. J. Control* **2015**, *88*, 2217–2227. https://doi.org/10.1080/00207179.2015.1039593.
24. Lee, T.; Leok, M.; McClamroch, N.H. Nonlinear robust tracking control of a quadrotor UAV on SE(3): Nonlinear robust tracking control of a quadrotor UAV. *Asian J. Control* **2013**, *15*, 391–408. https://doi.org/10.1002/asjc.567.
25. Rajappa, S.; Ryll, M.; Bulthoff, H.H.; Franchi, A. Modeling, control and design optimization for a fully-actuated hexarotor aerial vehicle with tilted propellers. In Proceedings of the 2015 IEEE International Conference on Robotics and Automation (ICRA), Seattle, WA, USA, 26–30 May 2015; IEEE: Seattle, WA, USA; pp. 4006–4013.
26. Ansari, U.; Bajodah, A.H.; Hamayun, M.T. Quadrotor control via robust generalized dynamic inversion and adaptive non-Singular terminal sliding mode. *Asian J. Control* **2019**, *21*, 1237–1249. https://doi.org/10.1002/asjc.1800.
27. Kolmanovsky, I.; Kalabić, U.; Gilbert, E. Developments in constrained control using reference governors. *IFAC Proc. Vol.* **2012**, *45*, 282–290. https://doi.org/10.3182/20120823-5-NL-3013.00042.

# Chapter 6

## Robust Gait Plan for the Tiltrotor (Four-Dimensional Gait Surfaces for a Tilt-Rotor—Two Color Map Theorem)

**This Chapter has been published on Four-Dimensional Gait Surfaces for a Tilt-Rotor—Two Color Map Theorem. *Drones* 2022, *6*, 103, doi:10.3390/drones6050103. as the first author. The rest co-author(s) contribute nothing other than supervising and/or minor edition.**

**Abstract:** This article presents the four-dimensional surfaces that guide the gait plan for a tilt-rotor. The previous gaits analyzed in the tilt-rotor research are inspired by animals; no theoretical base backs the robustness of these gaits. This research deduces the gaits by diminishing the adverse effect of the attitude of the tilt-rotor for the first time. Four-dimensional gait surfaces are subsequently found on which the gaits are expected to be robust to the attitude. These surfaces provide the region where the gait is suggested to be planned. However, a discontinuous region may hinder the gait plan process while utilizing the proposed gait surfaces. The 'Two Color Map Theorem' is then established to guarantee the continuity of each gait designed. The robustness of the typical gaits on the gait surface, obeying the Two Color Map Theorem, is demonstrated by comparing the singular curves in attitude with the gaits not on the gait surface. The result shows that the gaits on the gait surface receive wider regions of the acceptable attitudes.

### 1. Introduction

In stabilizing a quadrotor, feedback linearization is favored for its unique character in linearizing the nonlinear system [1–5]; several degrees of freedom can be subsequently controlled independently by this method. These degrees of freedom can be selected as attitude and altitude [4,6,7], position and yaw angle [8–11], attitude only [12], etc. The reason for not assigning all six degrees of freedom as independent controlled variables is that the number of inputs in a conventional quadrotor is four, marking the maximum number of degrees of freedom to be independently controlled [13] less than the entire degrees of freedom.

Typical feedback linearization requires an invertible decoupling matrix [14,15], which is always satisfied for a quadrotor with attitude–altitude independent output choice, while the position-yaw independent output choice witnesses a singular decoupling matrix for some attitudes [11,16].

On the other hand, some studies stabilize Ryll's tilt-rotor [17–19], a novel UAV with eight inputs, by controlling all degrees of freedom independently and simultaneously. The relevant decoupling matrices with six inputs and eight inputs have been proven invertible, marking the feasibility of the application of feedback linearization.

These controllers, however, can command the tilting angles of the tilt-rotor to change over-intensively [17,20,21], not expected in application.

With this concern, our previous research sets the magnitudes of the thrusts the only four inputs assigned by the subsequent united control rule [22]. The four tilting angles are defined in gait plan beforehand, which is totally not influenced by the control rule.

Undeniably, the maximum number of degrees of freedom that can be independently controlled decreases to four, less than the number of the entire degrees of freedom (six); attitude and altitude are selected as the only degrees of freedom to be directly stabilized. The remaining degrees of freedom, X and Y in position, are tracked by adjusting the attitude properly based on a modified attitude-position decoupler [23]. While the conventional attitude-position decoupler [7,24–26] for a quadrotor no longer works for a tilt-rotor.

Note that the relevant decoupling matrix can be singular for some attitudes while adopting some gaits [22]. Various animal-inspired gaits are subsequently analyzed and modified to avert a non-invertible decoupling matrix by scaling [27–29], which has been proven to be a valid approach to modify a gait. The modified gaits show wider regions of the acceptable attitudes in the roll-pitch diagram, indicating the enhanced robustness.

However, no research guides the gait plan, considering the robustness, thus far, without adopting an existing gait (e.g., animal-inspired gait [23,27]). Indeed, deducing the explicit relationship between the attitude and the singularity of the decoupling matrix can be cumbersome since they are tangled in a highly nonlinear way.

This article articulates the influence of the attitude to the singularity of the decoupling matrix by little attitude approximation. The deduced gaits are robust to the change of the attitude; restricted disturbances in attitude will not change the singularity of the decoupling matrix while adopting the deduced gaits. The resulting acceptable gaits lie on the four-dimensional gait surface.

To avoid the discontinuous change in the tilting angles, the gait is required to move continuously along all the four dimensions of the four-dimensional gait surface. The Two Color Map Theorem is subsequently developed to further guide the gait plan to design a continuous gait, which is inspired by the well-known Four Color Map Theorem [30,31].

The acceptable attitudes of four typical gaits on the different parts of the four-dimensional gait surface are analyzed in the roll-pitch diagram. The results are compared with the relevant biased gaits, locating outside the four-dimensional gait surfaces. It shows that the robustness of the gait is improved if the gait is planned on the four-dimensional gait surface.

The rest of this article is organized as follows. Section 2 reviews the necessary condition to receive the invertible decoupling matrix for our tilt-rotor. Section 3 thoroughly deduces the acceptable gaits considering the robustness. The resulting four-dimensional gait surfaces are also visualized in the same section. Section 4 proposes the Two Color Map Theorem to guide a continuous gait plan. The verification of the robustness of the proposed gaits is analyzed in the roll-pitch diagram in Section 5. Finally, Section 6 addresses the conclusions and discussions.

## 2. Invertible Decoupling Matrix: Necessary Condition

The target tilt-rotor analyzed in this research is sketched in Figure 1 [22]. This structure was initially put forward by Ryll [13].

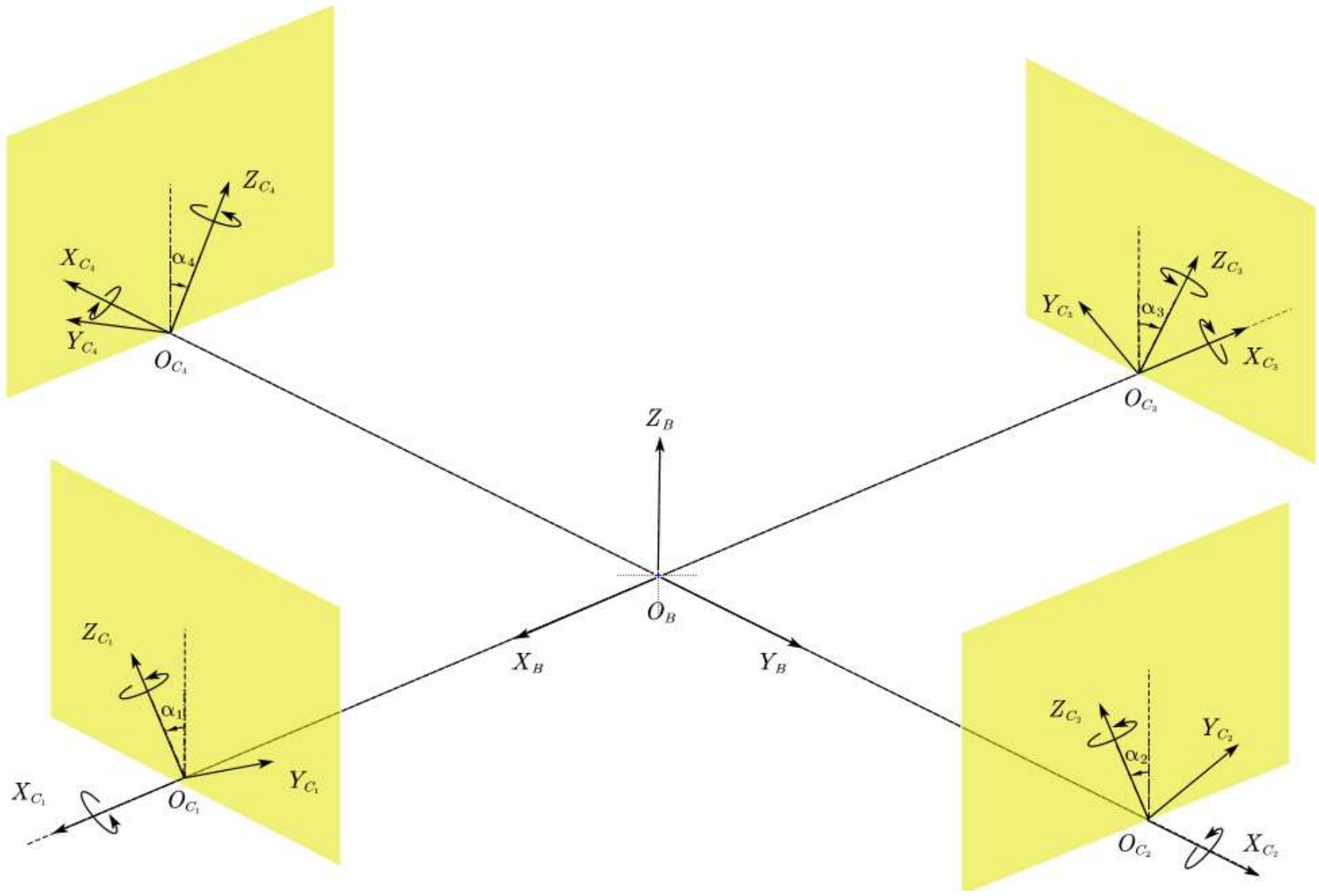

**Figure 1.** The sketch of Ryll's tilt-rotor.

The frames introduced in the dynamics of this tilt-rotor are earth frame $\mathcal{F}_E$, body-fixed frame $\mathcal{F}_B$, and four rotor frames $\mathcal{F}_{C_i}(i=1,2,3,4)$, each of which is fixed on the tilt motor mounted at the end of each arm. Rotor 1 and 3 are assumed to rotate clockwise along $Z_{C_1}$ and $Z_{C_3}$. While rotor 2 and 4 are assumed to rotate counterclockwise along $Z_{C_2}$ and $Z_{C_4}$.

As can be seen in Figure 1, the thrusts can be assigned to the vector along the directions on the highlighted yellow planes (tilting). This is completed by changing the tilting angles marked by $\alpha_1, \alpha_2, \alpha_3, \alpha_4$ in Figure 1.

Ryll proved that the decoupling matrix in feedback linearization, programmed to independently stabilize each degree of freedom, is invertible [13]. This result guarantees that the dynamic inversion can be conducted throughout the entire flight for any attitude. This approach, however, introduces over-intensive changes in the tilting angles, a typical result of which can be found in Figure 2 [22]. Given the initial thrusts insufficient to compensate the gravity, the tilt-rotor successfully hovers relying on the united control rule with eight inputs (over-actuated control). While the tilting angles changes over-intensively.

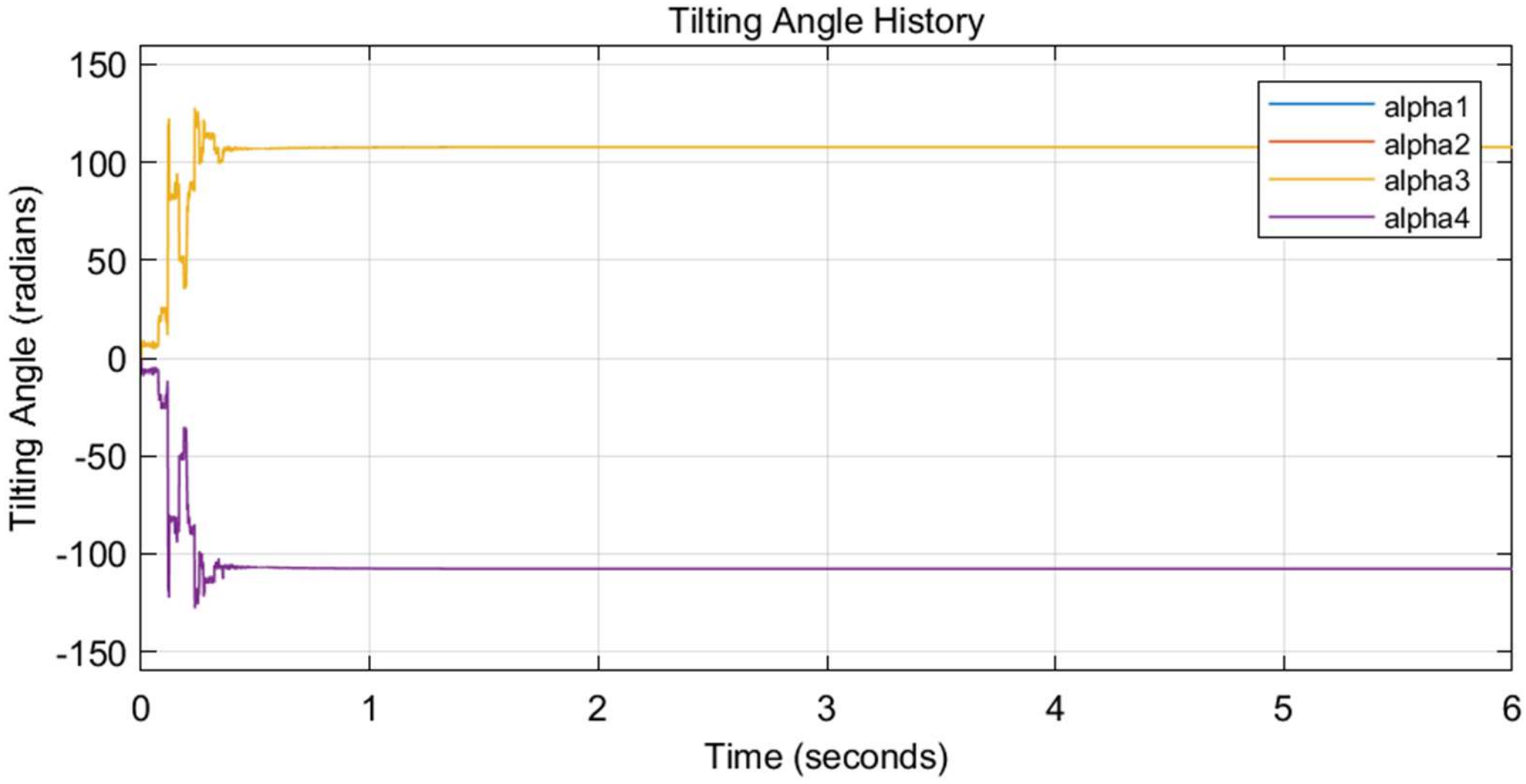

**Figure 2.** Tilting angle history of over-actuated control.

To overcome this obstacle, an alternative selection of the independent controlled degrees of freedom, attitude and altitude, is given [22] while utilizing the magnitudes as the only inputs. Define the combination of the tilting angle $(\alpha_1, \alpha_2, \alpha_3, \alpha_4)$ as gait. The tilting angles $(\alpha_1, \alpha_2, \alpha_3, \alpha_4)$ are specified with time beforehand as a separate process (gait plan), not influenced by the subsequent feedback linearization. The only inputs resulted from a united control rule are the magnitudes of the four thrusts.

For example, our previous research [27] planned an animal-inspired gait (inspired by cat walk), Figure 3, where the period is set as 1 s. The tilting angles changes periodically during the whole flight.

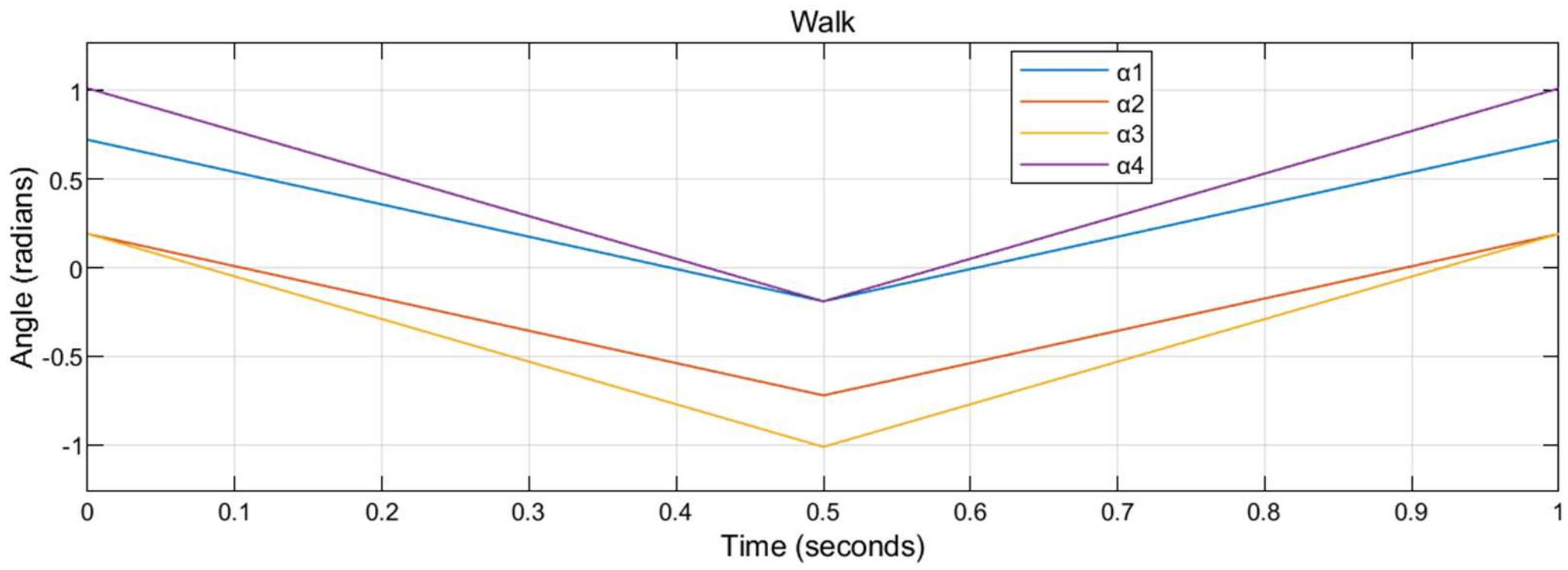

**Figure 3.** The walk gait of a cat. The period is set as 1 s.

Interestingly, this selection of independent controlled variables (attitude and altitude) introduces no singular decoupling matrix in a conventional quadrotor, which, actually, can be regarded as a special case for a tilt-rotor, e.g., $\alpha_1 = 0, \alpha_2 = 0, \alpha_3 = 0, \alpha_4 = 0$. However, it introduces the singular decoupling matrix in a tilt-rotor for some tilting angles given some attitudes.

Specifically, the necessary condition [22] to receive an invertible decoupling matrix in our tilt-rotor is given by

$$\begin{aligned}
&1.000 \cdot c1 \cdot c2 \cdot c3 \cdot s4 \cdot s\theta - 1.000 \cdot c1 \cdot s2 \cdot c3 \cdot c4 \cdot s\theta - 2.880 \cdot c1 \\
&\cdot c2 \cdot s3 \cdot s4 \cdot s\theta + 2.880 \cdot c1 \cdot s2 \cdot s3 \cdot c4 \cdot s\theta - 2.880 \cdot s1 \cdot c2 \cdot c3 \\
&\cdot s4 \cdot s\theta + 2.880 \cdot s1 \cdot s2 \cdot c3 \cdot c4 \cdot s\theta - 1.000 \cdot s1 \cdot c2 \cdot s3 \cdot s4 \cdot s\theta \\
&+ 1.000 \cdot s1 \cdot s2 \cdot s3 \cdot c4 \cdot s\theta + 4.000 \cdot c1 \cdot c2 \cdot c3 \cdot c4 \cdot c\phi \cdot c\theta \\
&+ 5.592 \cdot c1 \cdot c2 \cdot c3 \cdot s4 \cdot c\phi \cdot c\theta - 5.592 \cdot c1 \cdot c2 \cdot s3 \cdot c4 \cdot c\phi \cdot c\theta \\
&+ 5.592 \cdot c1 \cdot s2 \cdot c3 \cdot c4 \cdot c\phi \cdot c\theta - 5.592 \cdot s1 \cdot c2 \cdot c3 \cdot c4 \cdot c\phi \cdot c\theta \\
&+ 1.000 \cdot c1 \cdot c2 \cdot s3 \cdot c4 \cdot s\phi \cdot c\theta + 0.9716 \cdot c1 \cdot c2 \cdot s3 \cdot s4 \cdot c\phi \cdot c\theta \\
&- 2.000 \cdot c1 \cdot s2 \cdot c3 \cdot s4 \cdot c\phi \cdot c\theta + 0.9716 \cdot c1 \cdot s2 \cdot s3 \cdot c4 \cdot c\phi \cdot c\theta \\
&- 1.000 \cdot s1 \cdot c2 \cdot c3 \cdot c4 \cdot s\phi \cdot c\theta + 0.9716 \cdot s1 \cdot c2 \cdot c3 \cdot s4 \cdot c\phi \cdot c\theta \\
&- 2.000 \cdot s1 \cdot c2 \cdot s3 \cdot c4 \cdot c\phi \cdot c\theta + 0.9716 \cdot s1 \cdot s2 \cdot c3 \cdot c4 \cdot c\phi \cdot c\theta \\
&+ 2.880 \cdot c1 \cdot c2 \cdot s3 \cdot s4 \cdot s\phi \cdot c\theta + 2.880 \cdot c1 \cdot s2 \cdot s3 \cdot c4 \cdot s\phi \cdot c\theta \\
&- 0.1687 \cdot c1 \cdot s2 \cdot s3 \cdot s4 \cdot c\phi \cdot c\theta - 2.880 \cdot s1 \cdot c2 \cdot c3 \cdot s4 \cdot s\phi \cdot c\theta \\
&+ 0.1687 \cdot s1 \cdot c2 \cdot s3 \cdot s4 \cdot c\phi \cdot c\theta - 2.880 \cdot s1 \cdot s2 \cdot c3 \cdot c4 \cdot s\phi \cdot c\theta \\
&- 0.1687 \cdot s1 \cdot s2 \cdot c3 \cdot s4 \cdot c\phi \cdot c\theta + 0.1687 \cdot s1 \cdot s2 \cdot s3 \cdot c4 \cdot c\phi \\
&\cdot c\theta - 1.000 \cdot c1 \cdot s2 \cdot s3 \cdot s4 \cdot s\phi \cdot c\theta + 1.000 \cdot s1 \cdot s2 \cdot c3 \cdot s4 \cdot s\phi \\
&\cdot c\theta \\
&\neq 0
\end{aligned} \tag{1}$$

where $s\Lambda = \sin(\Lambda)$ and $c\Lambda = \cos(\Lambda)$, $si = \sin(\alpha_i)$, $ci = \cos(\alpha_i)$, $(i = 1,2,3,4)$, $\phi$ and $\theta$ are roll angle and pitch angle, respectively.

Since the attitude (roll and pitch) is not predictable in the gait plan process, the left side of Formula (1) cannot be achieved in the gait plan. The unknown $\phi$ and $\theta$ on the left side of Formula 1 make it difficult to plan a gait while satisfying this condition. Considering that roll angle and pitch angle are small in typical flights, our previous research [23] makes zero attitude approximation to Formula (1): substituting $\phi = 0$ and $\theta = 0$ into Formula (1) yields

$$
\begin{aligned}
&4.000 \cdot c1 \cdot c2 \cdot c3 \cdot c4 + 5.592 \\
&\cdot (+c1{\cdot}c2{\cdot}c3{\cdot}s4{-}c1{\cdot}c2{\cdot}s3{\cdot}c4{+}c1{\cdot}s2{\cdot}c3{\cdot}c4{-}s1{\cdot}c2{\cdot}c3{\cdot}c4) \\
&+ 0.9716 \\
&\cdot (+c1{\cdot}c2{\cdot}s3{\cdot}s4{+}c1{\cdot}s2{\cdot}s3{\cdot}c4{+}s1{\cdot}c2{\cdot}c3{\cdot}s4{+}s1{\cdot}s2{\cdot}c3{\cdot}c4) \\
&+ 2.000 \cdot (-c1{\cdot}s2{\cdot}c3{\cdot}s4{-}s1{\cdot}c2{\cdot}s3{\cdot}c4) + 0.1687 \\
&\cdot (-c1{\cdot}s2{\cdot}s3{\cdot}s4{+}s1{\cdot}c2{\cdot}s3{\cdot}s4{-}s1{\cdot}s2{\cdot}c3{\cdot}s4{+}s1{\cdot}s2{\cdot}s3{\cdot}c4) \\
&\neq 0.
\end{aligned} \tag{2}
$$

Formula (2) contains the tilting angles only, making the verification in the gait plan process possible. Several animal-inspired gaits were then evaluated by Formula (2), and received satisfying tracking result [27].

This result (Formula (2)) from zero attitude approximation, however, discards the influence of the attitude. The effect of the disturbance of roll angle and pitch angle, which may violate Formula (1) cannot be traced from Formula (2).

From this point, though Formula (2) may be useful to determine whether a gait is feasible, it is not suitable to design a "robust gait" from nothing. A robust gait here is a time-specified $(\alpha_1,\alpha_2,\alpha_3,\alpha_4)$ that guarantees the left side of Formula (1) is nonzero, even with the introduction of small disturbances (away from zero) in attitude (roll angle and pitch angle).

A novel method of planning robust gaits is specified in the next section.

### 3. Four-Dimensional Gait Surface

Instead of utilizing zero attitude approximation, make the following near zero attituded approximation: $s\Lambda = \Lambda$, $c\Lambda = 1$. Substituting this near zero attitude approximation into Formula (1) yields

$$
R_\phi(\alpha_1,\alpha_2,\alpha_3,\alpha_4) \cdot \phi + R_\theta(\alpha_1,\alpha_2,\alpha_3,\alpha_4) \cdot \theta + R(\alpha_1,\alpha_2,\alpha_3,\alpha_4) \neq 0 \tag{3}
$$

where

$$
\begin{aligned}
&R_\phi(\alpha_1,\alpha_2,\alpha_3,\alpha_4) \\
&= 1.000 \cdot c1 \cdot c2 \cdot s3 \cdot c4 - 1.000 \cdot s1 \cdot c2 \cdot c3 \cdot c4 - 1.000 \cdot c1 \cdot s2 \cdot s3 \cdot s4 + \\
&1.000 \cdot s1 \cdot s2 \cdot c3 \cdot s4 + 2.880 \cdot c1 \cdot c2 \cdot s3 \cdot s4 + 2.880 \cdot c1 \cdot s2 \cdot s3 \cdot c4 - \\
&2.880 \cdot s1 \cdot c2 \cdot c3 \cdot s4 - 2.880 \cdot s1 \cdot s2 \cdot c3 \cdot c4,
\end{aligned} \tag{4}
$$

$$
\begin{aligned}
&R_\theta(\alpha_1,\alpha_2,\alpha_3,\alpha_4) \\
&= 1.000 \cdot c1 \cdot c2 \cdot c3 \cdot s4 - 1.000 \cdot c1 \cdot s2 \cdot c3 \cdot c4 - 1.000 \cdot s1 \cdot c2 \cdot s3 \cdot s4 + \\
&1.000 \cdot s1 \cdot s2 \cdot s3 \cdot c4 - 2.880 \cdot c1 \cdot c2 \cdot s3 \cdot s4 + 2.880 \cdot c1 \cdot s2 \cdot s3 \cdot c4 - \\
&2.880 \cdot s1 \cdot c2 \cdot c3 \cdot s4 + 2.880 \cdot s1 \cdot s2 \cdot c3 \cdot c4,
\end{aligned} \tag{5}
$$

$$
\begin{aligned}
&R(\alpha_1,\alpha_2,\alpha_3,\alpha_4) \\
&= 4.000 \cdot c1 \cdot c2 \cdot c3 \cdot c4 + 5.592 \cdot c1 \cdot c2 \cdot c3 \cdot s4 - 5.592 \cdot c1 \cdot c2 \cdot s3 \cdot c4 + \\
&5.592 \cdot c1 \cdot s2 \cdot c3 \cdot c4 - 5.592 \cdot s1 \cdot c2 \cdot c3 \cdot c4 + 0.9716 \cdot c1 \cdot c2 \cdot s3 \cdot s4 + \\
&0.9716 \cdot c1 \cdot s2 \cdot s3 \cdot c4 + 0.9716 \cdot s1 \cdot c2 \cdot c3 \cdot s4 + 0.9716 \cdot s1 \cdot s2 \cdot c3 \cdot c4 - \\
&2.000 \cdot c1 \cdot s2 \cdot c3 \cdot s4 - 2.000 \cdot s1 \cdot c2 \cdot s3 \cdot c4 - 0.1687 \cdot c1 \cdot s2 \cdot s3 \cdot s4 + \\
&0.1687 \cdot s1 \cdot c2 \cdot s3 \cdot s4 - 0.1687 \cdot s1 \cdot s2 \cdot c3 \cdot s4 + 0.1687 \cdot s1 \cdot s2 \cdot s3 \cdot c4.
\end{aligned} \tag{6}
$$

To diminish the disturbance of the attitude, roll angle and pitch angle on the left side of Formula (3), both $R_\phi(\alpha_1,\alpha_2,\alpha_3,\alpha_4)$ and $R_\theta(\alpha_1,\alpha_2,\alpha_3,\alpha_4)$ are set zero.

Further, to satisfy Formula (3), $R(\alpha_1,\alpha_2,\alpha_3,\alpha_4)$ is required to be non-zero.

In conclusion, the necessary condition to receive an invertible gait considering robustness is

$$
R_\phi(\alpha_1,\alpha_2,\alpha_3,\alpha_4) = 0, \tag{7}
$$

$$
R_\theta(\alpha_1,\alpha_2,\alpha_3,\alpha_4) = 0, \tag{8}
$$

$$R(\alpha_1,\alpha_2,\alpha_3,\alpha_4) \neq 0. \quad (9)$$

Since $R_\phi(\alpha_1,\alpha_2,\alpha_3,\alpha_4)$, $R_\theta(\alpha_1,\alpha_2,\alpha_3,\alpha_4)$, and $R(\alpha_1,\alpha_2,\alpha_3,\alpha_4)$ are highly nonlinear, the numerical results will be calculated only.

Evenly dividing the range $\alpha_1 \in [-\pi/2,\pi/2]$ into 16 pieces generates 17 grids, $-\pi/2 + (i-1) \cdot \pi/16, (i=\overline{1,17})$. Only these $\alpha_1$ on the grids are to be calculated. Similarly, only $\alpha_2 = -\pi/2 + (i-1) \cdot \pi/16, (i=\overline{1,17})$ are to be calculated. With this configuration, there are altogether 17 × 17 $(\alpha_1,\alpha_2)$ on the grids.

For each $(\alpha_1,\alpha_2)$, the corresponding $(\alpha_3,\alpha_4)$ within the range $\alpha_3 \in [-\pi/2,\pi/2]$, $\alpha_4 \in [-\pi/2,\pi/2]$ are found based on Equation (7) and (8) with the aid of Mathematica. The solver used is NSolve (Precision Set = 10).

Figure 4 plots the number of the roots of $(\alpha_3,\alpha_4)$ for each given $(\alpha_1,\alpha_2)$ on the grids.

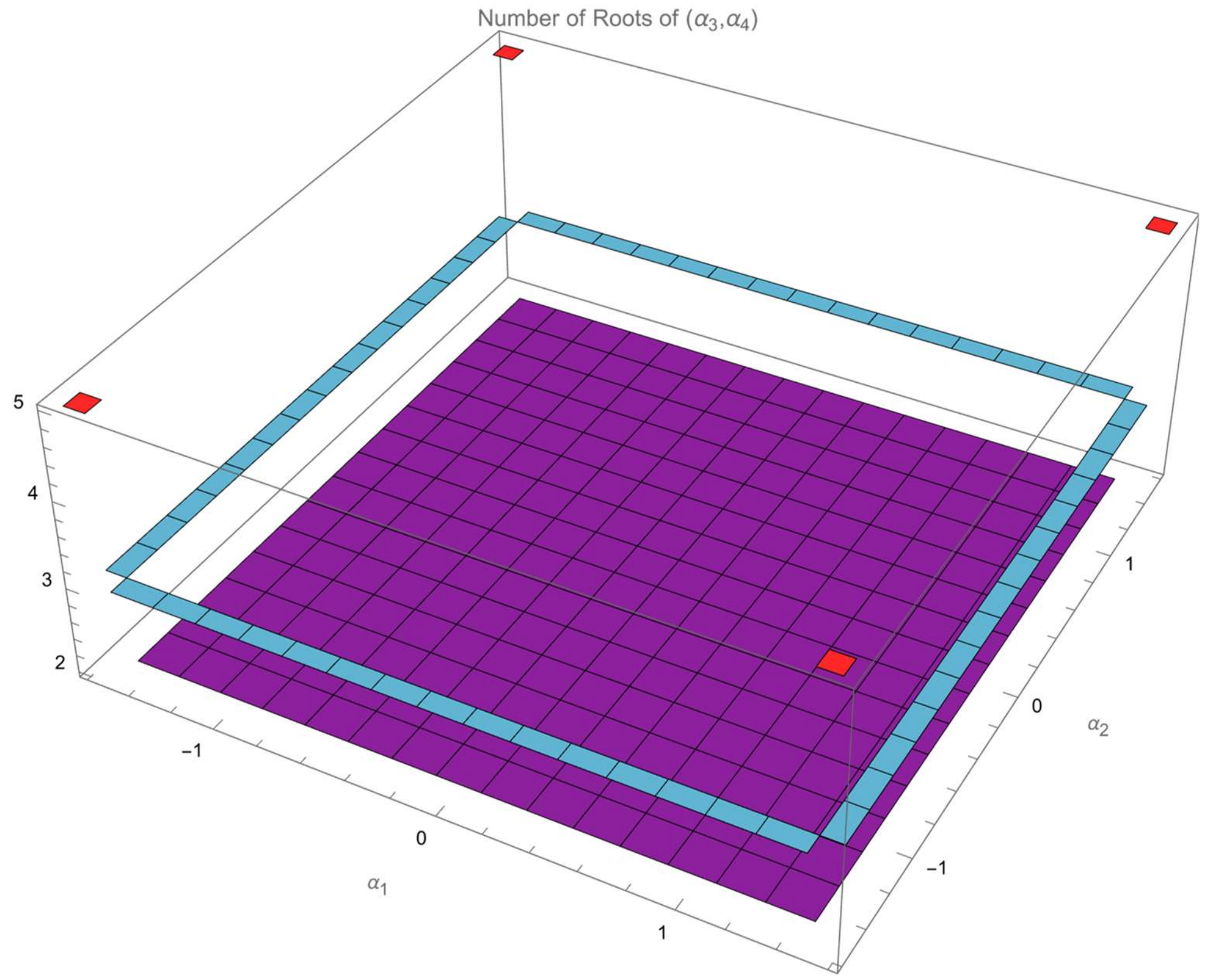

**Figure 4.** Distribution of the number of roots of $(\alpha_3,\alpha_4)$ for different $(\alpha_1,\alpha_2)$.

It can be seen that the number of roots of $(\alpha_3,\alpha_4)$ is 2 in the middle, $\alpha_1 \in (-\pi/2,\pi/2) \cap \alpha_2 \in (-\pi/2,\pi/2)$. Each corner, $\alpha_1 \in \{-\pi/2,\pi/2\} \cap \alpha_2 \in \{-\pi/2,\pi/2\}$, returns 5 roots to the corresponding $(\alpha_3,\alpha_4)$. Each of the rest $(\alpha_1,\alpha_2)$ on the side returns 3 roots of $(\alpha_3,\alpha_4)$.

Discarding $(\alpha_1,\alpha_2)$ on the side and the corner, we only discuss the $(\alpha_1,\alpha_2)$ in the middle, where there are 2 roots for each given $(\alpha_1,\alpha_2)$.

Figures 5 and 6 plots $\alpha_3$ and $\alpha_4$, respectively, in one of two groups of the roots of $(\alpha_3,\alpha_4)$.

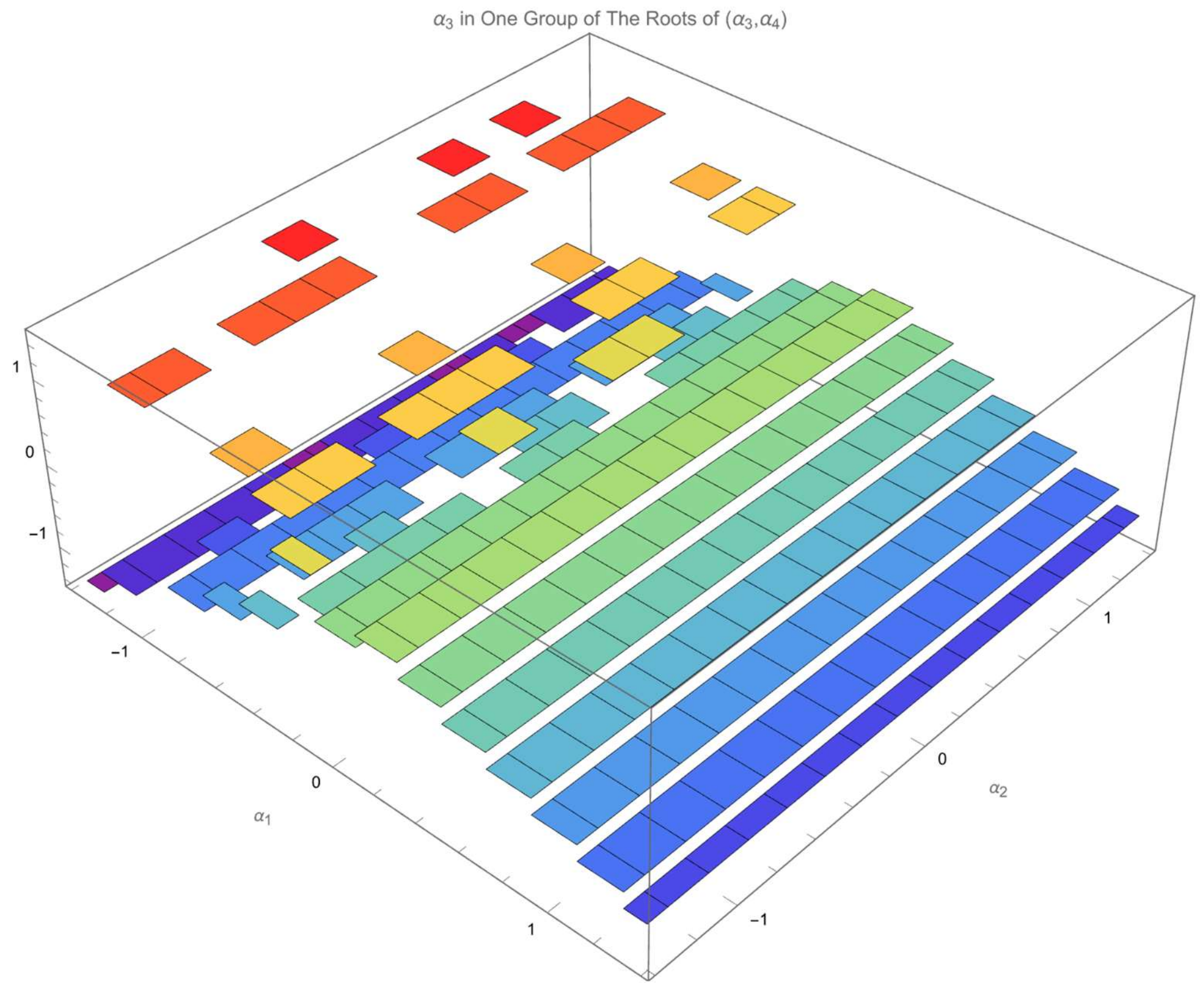

**Figure 5.** The value of $\alpha_3$ in one group of the roots of ($\alpha_3$,$\alpha_4$).

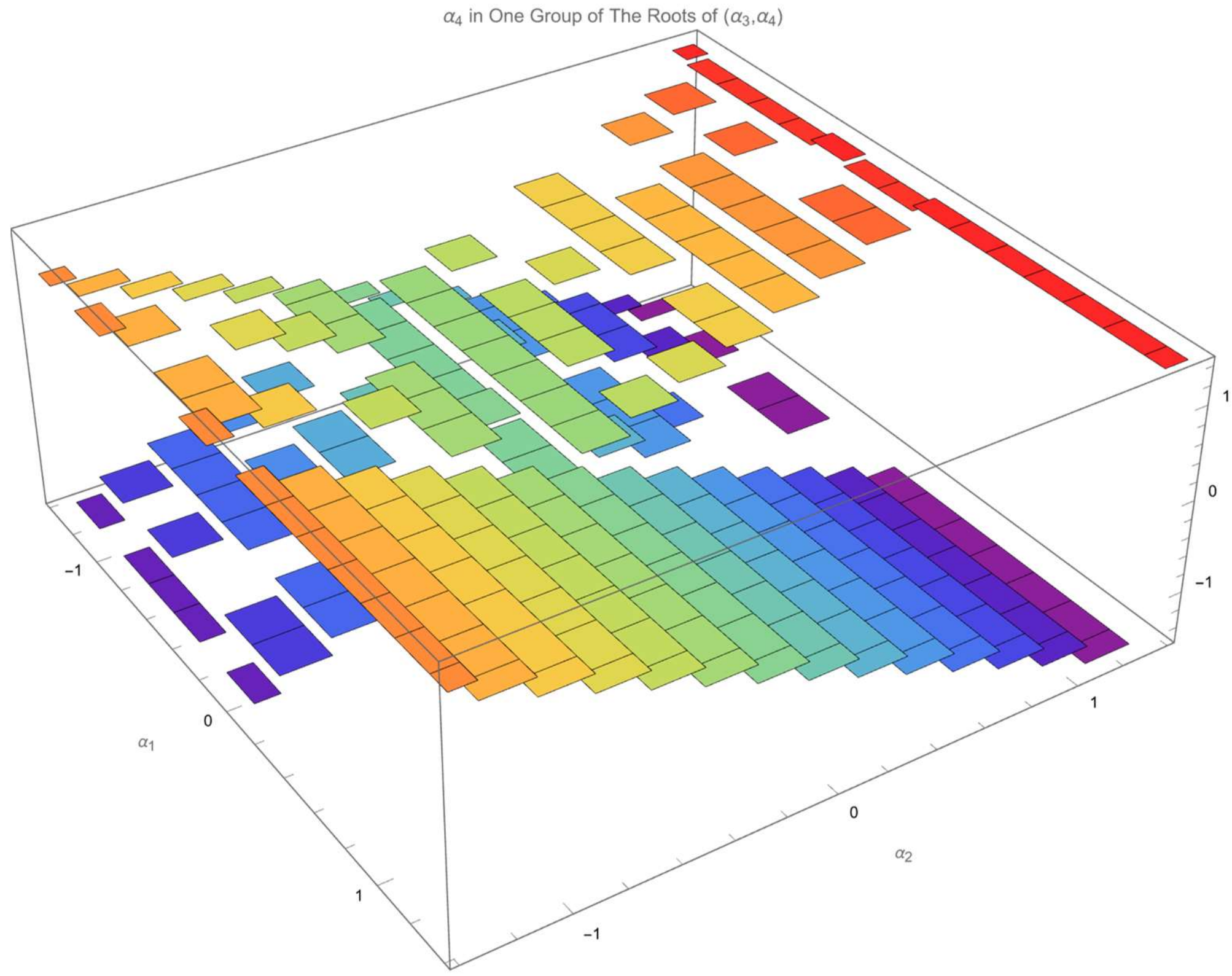

**Figure 6.** The value of $\alpha_4$ in one group of the roots of ($\alpha_3$,$\alpha_4$).

Figures 7 and 8 plots $\alpha_3$ and $\alpha_4$, respectively, in the other root of ($\alpha_3$,$\alpha_4$).

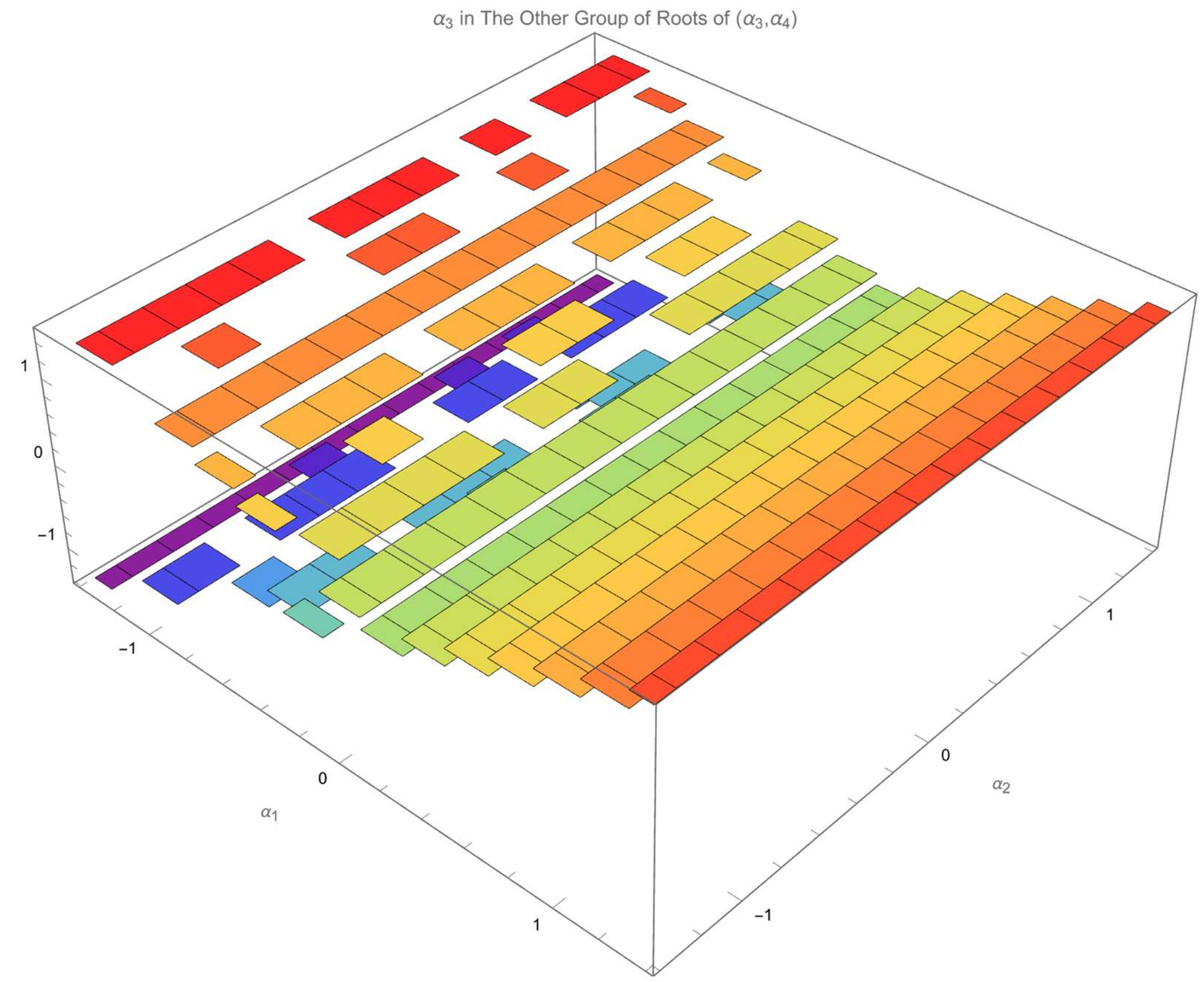

**Figure 7.** The value of $\alpha_3$ in the other group of roots of ($\alpha_3$,$\alpha_4$).

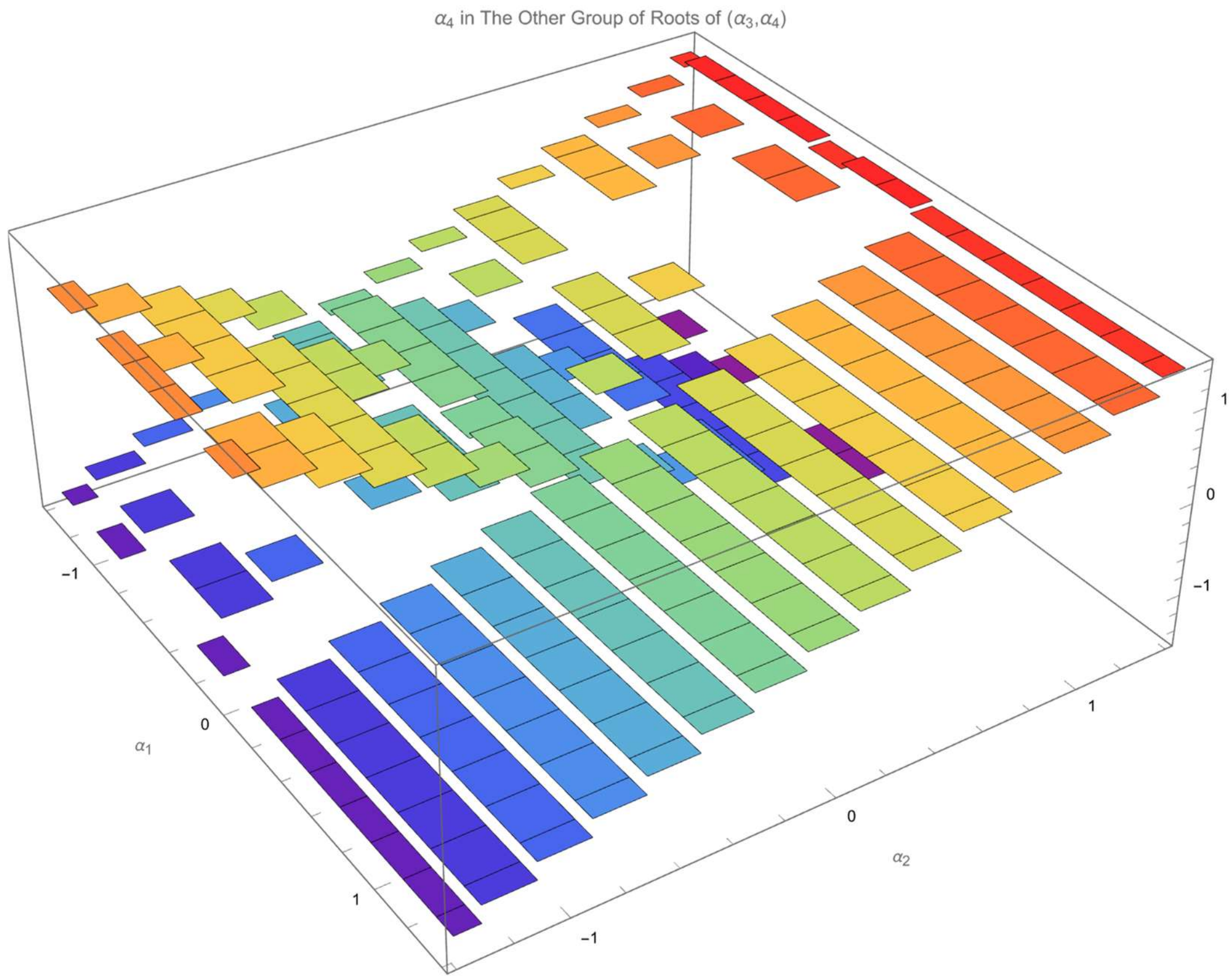

**Figure 8.** The value of $\alpha_4$ in the other group of roots of ($\alpha_3$,$\alpha_4$).

Show two values of $\alpha_3$ in the same figure (Figure 9). Show two values of $\alpha_4$ in the same figure (Figure 10).

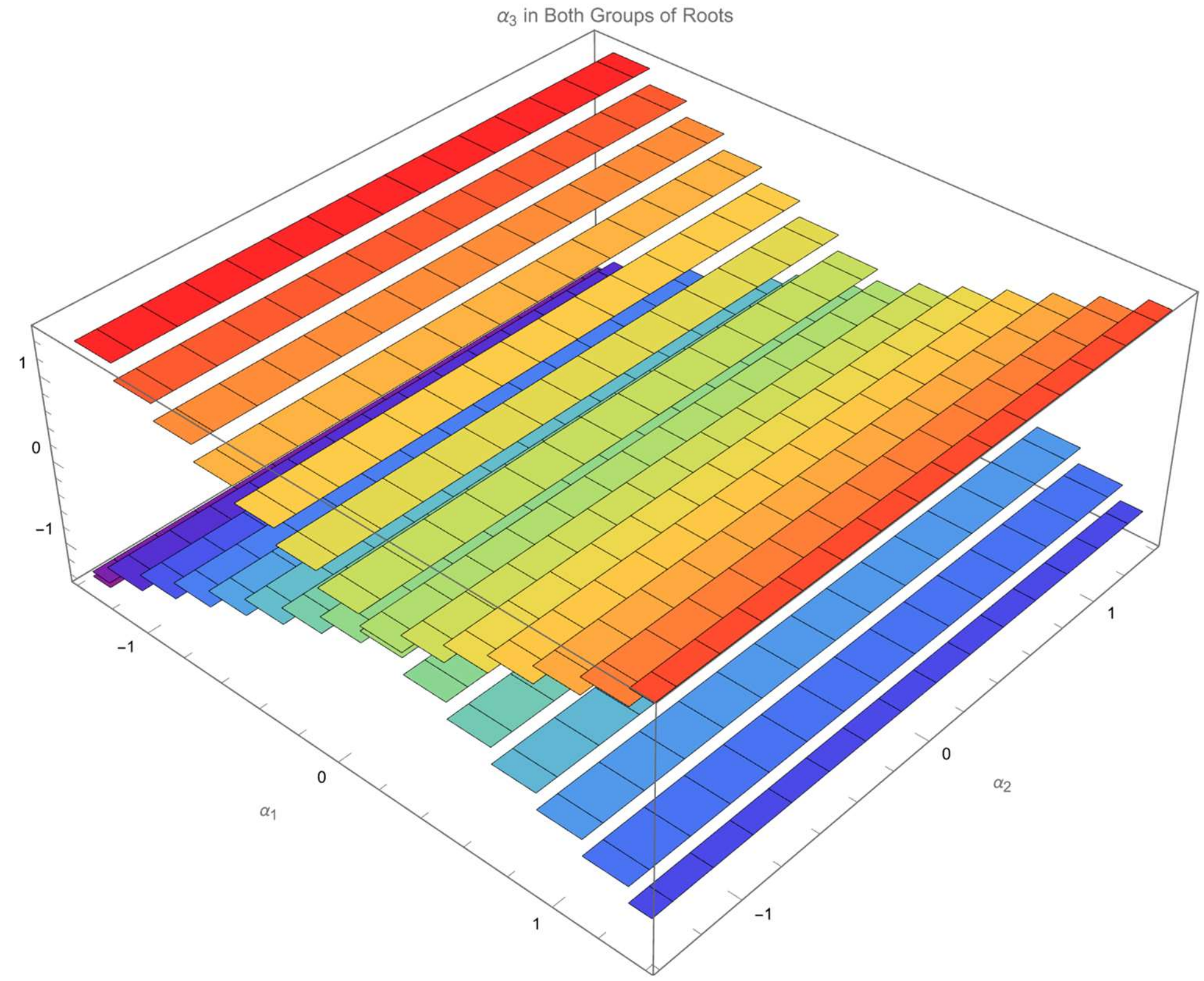

**Figure 9.** The value of $\alpha_3$ in both groups of roots of ($\alpha_3$,$\alpha_4$).

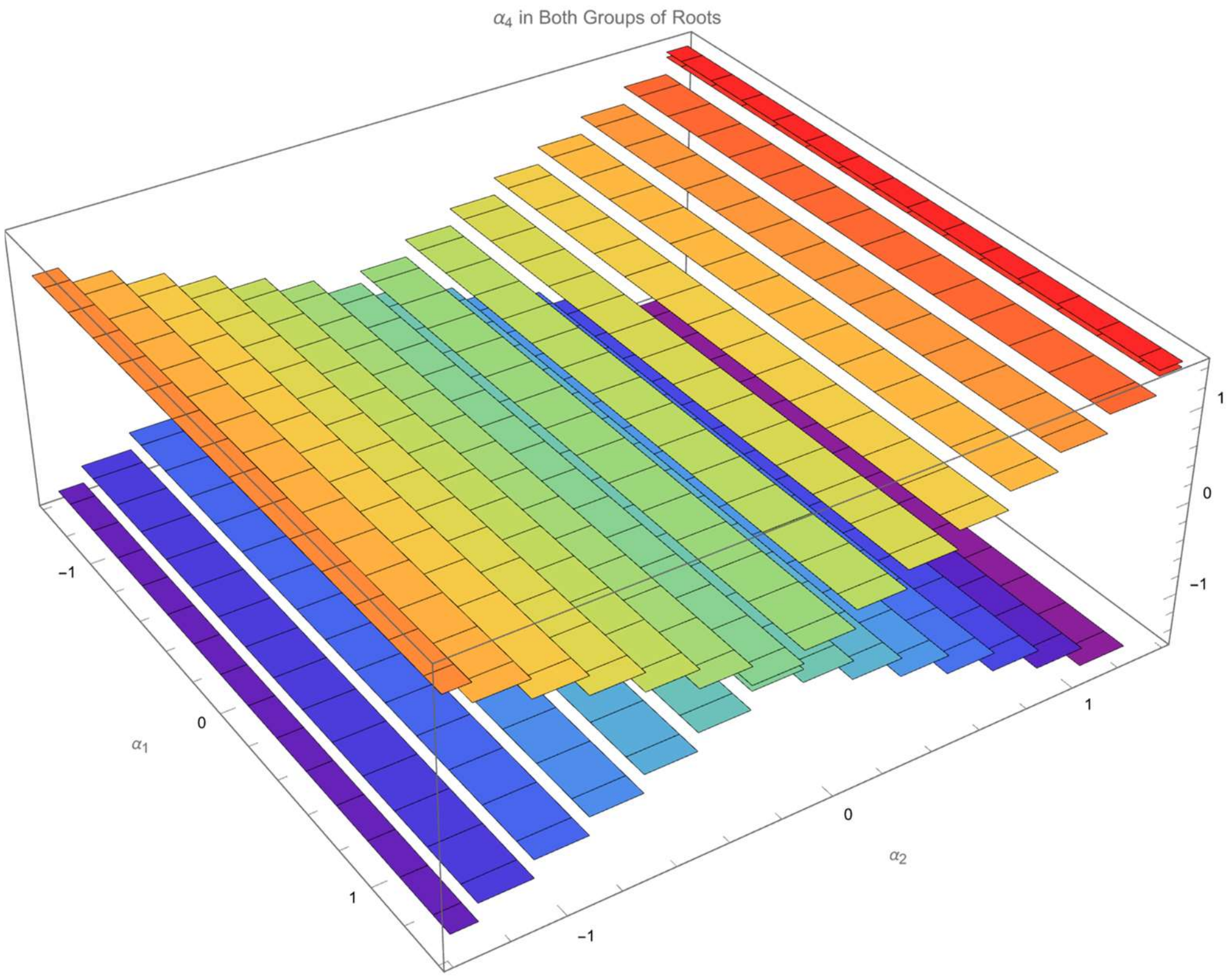

**Figure 10.** The value of $\alpha_4$ in both groups of roots of ($\alpha_3$,$\alpha_4$).

Although the regularity of the distribution of $\alpha_3$ in either root of $(\alpha_3,\alpha_4)$, Figures 5 and 7, can hardly be tracked, Figure 9 shows that both $\alpha_3$ lie on the two planes. One plane increases $\alpha_3$ when $\alpha_1$ increases. Define this plane "+" plane. The other plane decreases $\alpha_3$ when $\alpha_1$ increases. Define this plane "−" plane.

Similarly, Figure 10 shows that both $\alpha_4$ also lie on another two planes. One plane increases $\alpha_4$ when $\alpha_2$ increases. Define this plane "+" plane. The other plane decreases $\alpha_4$ when $\alpha_2$ increases. Define this plane "−" plane.

With these definitions, the roots of $(\alpha_3,\alpha_4)$ can be classified by the mark (±,±). For example, (+,−) represents the root of $(\alpha_3,\alpha_4)$, whose $\alpha_3$ lies on "+" plane and $\alpha_4$ lies on "−" plane.

Observing Figures 3–6, there are only 7 categories of the roots of $(\alpha_3,\alpha_4)$. They are (+,+), (−,−), (+,?), (?,+), (−,?), (?,−), (?,?). "?" in either of planes represents that the intersection of the relevant "+" plane and "−" plane.

Note that no roots of $(\alpha_3,\alpha_4)$ belong to (+,−) or (−,+).

These definitions will facilitate the discussions on the Two Color Map Theorem, detailed in the next section.

Observing Formula (6), we assert that $R(\alpha_1,\alpha_2,\alpha_3,\alpha_4)$ is continuous, given $\alpha_1$, $\alpha_2$, $\alpha_3$, and $\alpha_4$ are continuous.

**Proposition 1.** *Given $R_\phi(\alpha_1,\alpha_2,\alpha_3,\alpha_4) = 0$ and $R_\theta(\alpha_1,\alpha_2,\alpha_3,\alpha_4) = 0$, the necessary condition to receive an invertible decoupling matrix is*

$$\forall\alpha_1, \alpha_2, \alpha_3, \alpha_4, R(\alpha_1,\alpha_2,\alpha_3,\alpha_4) > 0 \tag{10}$$

*or*

$$\forall\alpha_1, \alpha_2, \alpha_3, \alpha_4, R(\alpha_1,\alpha_2,\alpha_3,\alpha_4) < 0. \tag{11}$$

**Proof of Proposition 1.** Considering that $R(\alpha_1,\alpha_2,\alpha_3,\alpha_4)$ is continuous, the rebuttal method yields this result. □

Based on Proposition 1, the roots of $(\alpha_3,\alpha_4)$ previously found are further classified into the ones receiving positive $R(\alpha_1,\alpha_2,\alpha_3,\alpha_4)$ and the ones receiving negative $R(\alpha_1,\alpha_2,\alpha_3,\alpha_4)$.

Figure 11 plots the roots of $(\alpha_3,\alpha_4)$ in Figures 5 and 6 receiving positive $R(\alpha_1,\alpha_2,\alpha_3,\alpha_4)$ with $\alpha_3$ in red and $\alpha_4$ in blue. The value of the relevant $R(\alpha_1,\alpha_2,\alpha_3,\alpha_4)$ is illustrated in Figure 12.

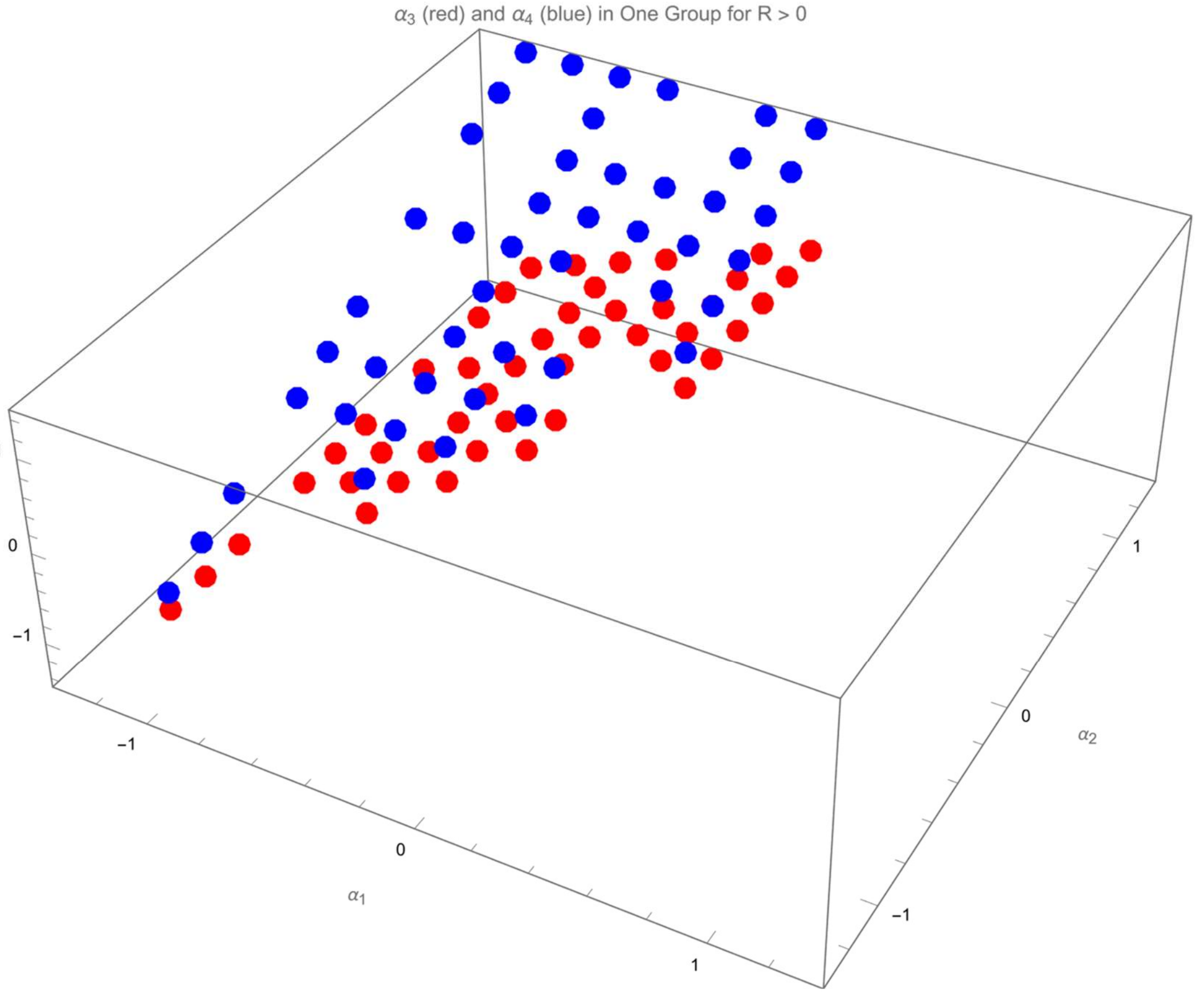

**Figure 11.** The value of $\alpha_3$ and $\alpha_4$ in one group of the roots of ($\alpha_3$,$\alpha_4$) receiving positive $R(\alpha_1,\alpha_2,\alpha_3,\alpha_4)$. The red points represent $\alpha_3$. The blue points represent $\alpha_4$.

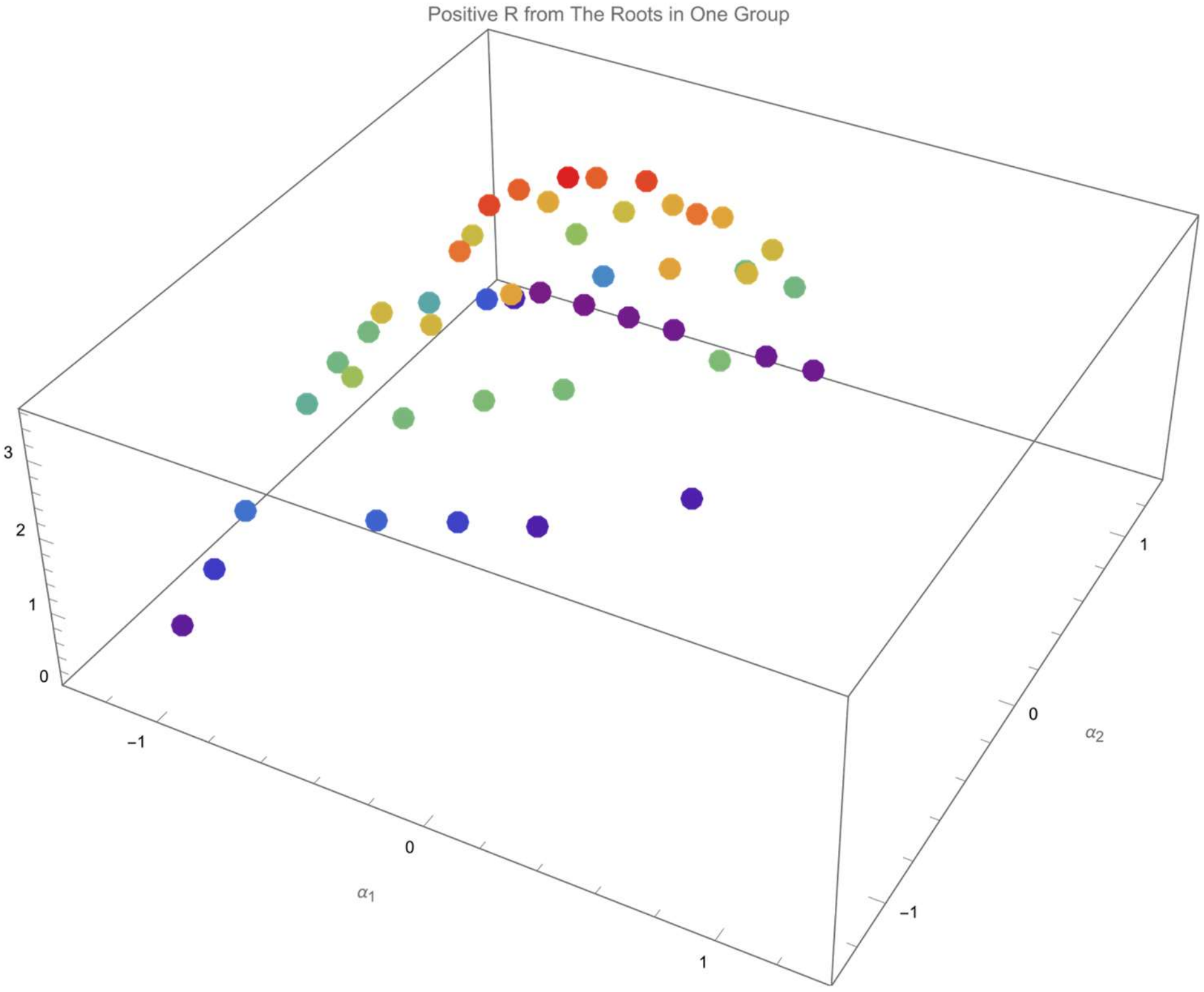

**Figure 12.** The value of the positive $R(\alpha_1,\alpha_2,\alpha_3,\alpha_4)$ received by one group of the roots of ($\alpha_3$,$\alpha_4$).

Figure 13 plots the roots of $(\alpha_3,\alpha_4)$ in Figures 7 and 8 receiving positive $R(\alpha_1,\alpha_2,\alpha_3,\alpha_4)$ with $\alpha_3$ in red and $\alpha_4$ in blue. The value of the relevant $R(\alpha_1,\alpha_2,\alpha_3,\alpha_4)$ is illustrated in Figure 14.

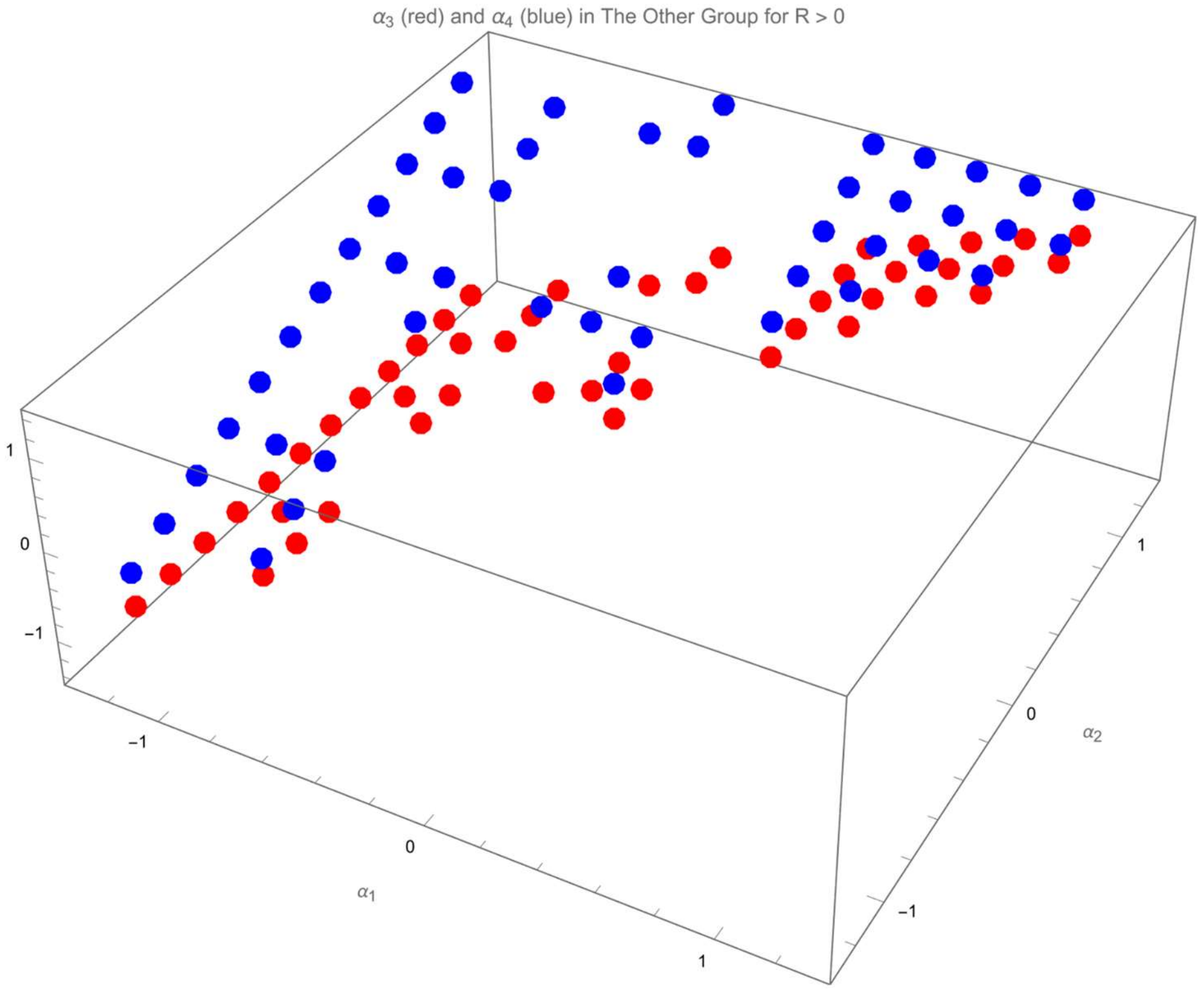

**Figure 13.** The value of $\alpha_3$ and $\alpha_4$ in the other group of the roots of $(\alpha_3,\alpha_4)$ receiving positive $R(\alpha_1,\alpha_2,\alpha_3,\alpha_4)$. The red points represent $\alpha_3$. The blue points represent $\alpha_4$.

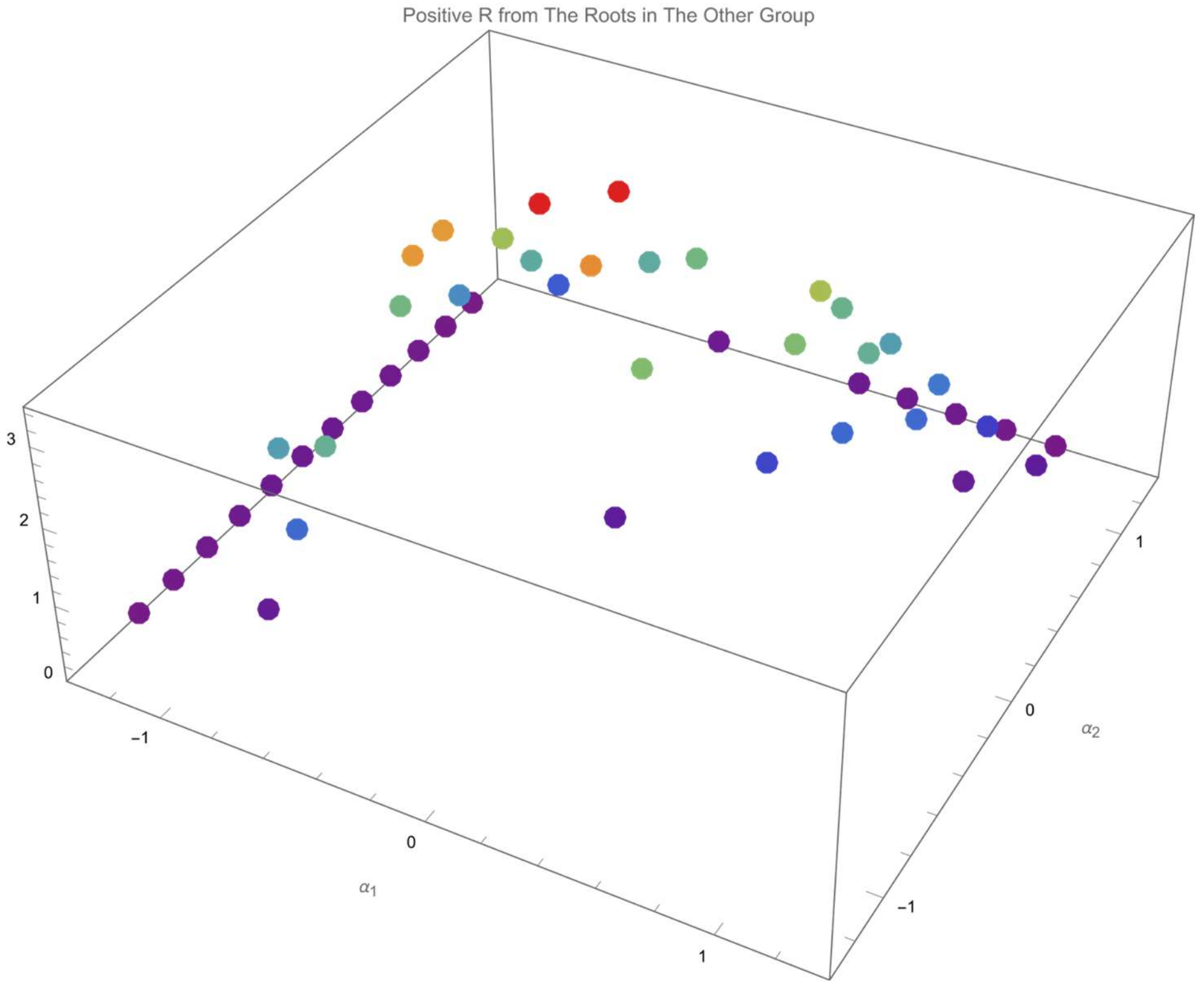

**Figure 14.** The value of the positive $R(\alpha_1,\alpha_2,\alpha_3,\alpha_4)$ received by the other group of the roots of $(\alpha_3,\alpha_4)$.

Note that both roots of $(\alpha_3,\alpha_4)$ receiving positive $R(\alpha_1,\alpha_2,\alpha_3,\alpha_4)$, Figures 11 and 13, belong to (+,+) (+,?), and (?,+) types.

Further, check the corresponding $(\alpha_1,\alpha_2)$ of each blue/red points in Figures 11 and 13. It can be found that a triangular area in the $\alpha_1 - \alpha_2$ plane is occupied by the corresponding $(\alpha_1,\alpha_2)$ in both figures: the projection of these applicable $(\alpha_3,\alpha_4)$ in both roots fully occupy the triangular area of $(\alpha_1,\alpha_2)$ without overlapping, indicating that there is only one root of $(\alpha_3,\alpha_4)$ meeting $R(\alpha_1,\alpha_2,\alpha_3,\alpha_4) > 0$ for each $(\alpha_1,\alpha_2)$ in the triangular area. This triangular area is governed by three vertices: $(\alpha_1,\alpha_2) = (-7\pi/16, 7\pi/16)$, $(\alpha_1,\alpha_2) = (-7\pi/16, -5\pi/16)$, $(\alpha_1,\alpha_2) = (5\pi/16, 7\pi/16)$.

Similarly, the distributions of both roots of $(\alpha_3,\alpha_4)$ receiving negative $R(\alpha_1,\alpha_2,\alpha_3,\alpha_4)$ are plotted in Figures 15 and 16 with $\alpha_3$ in red and $\alpha_4$ in blue.

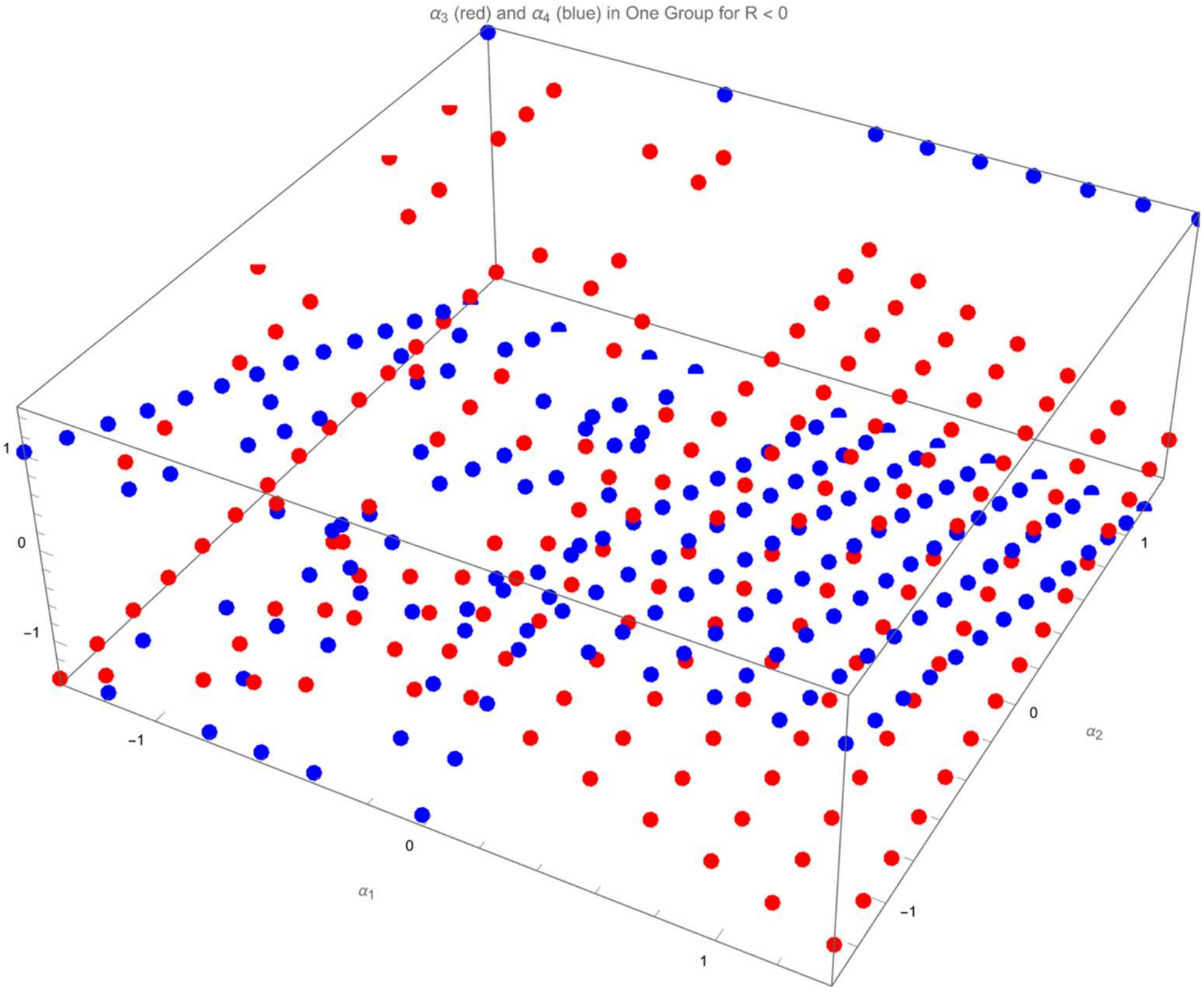

**Figure 15.** The value of $\alpha_3$ and $\alpha_4$ in one group of the roots of $(\alpha_3,\alpha_4)$ receiving negative $R(\alpha_1,\alpha_2,\alpha_3,\alpha_4)$. The red points represent $\alpha_3$. The blue points represent $\alpha_4$.

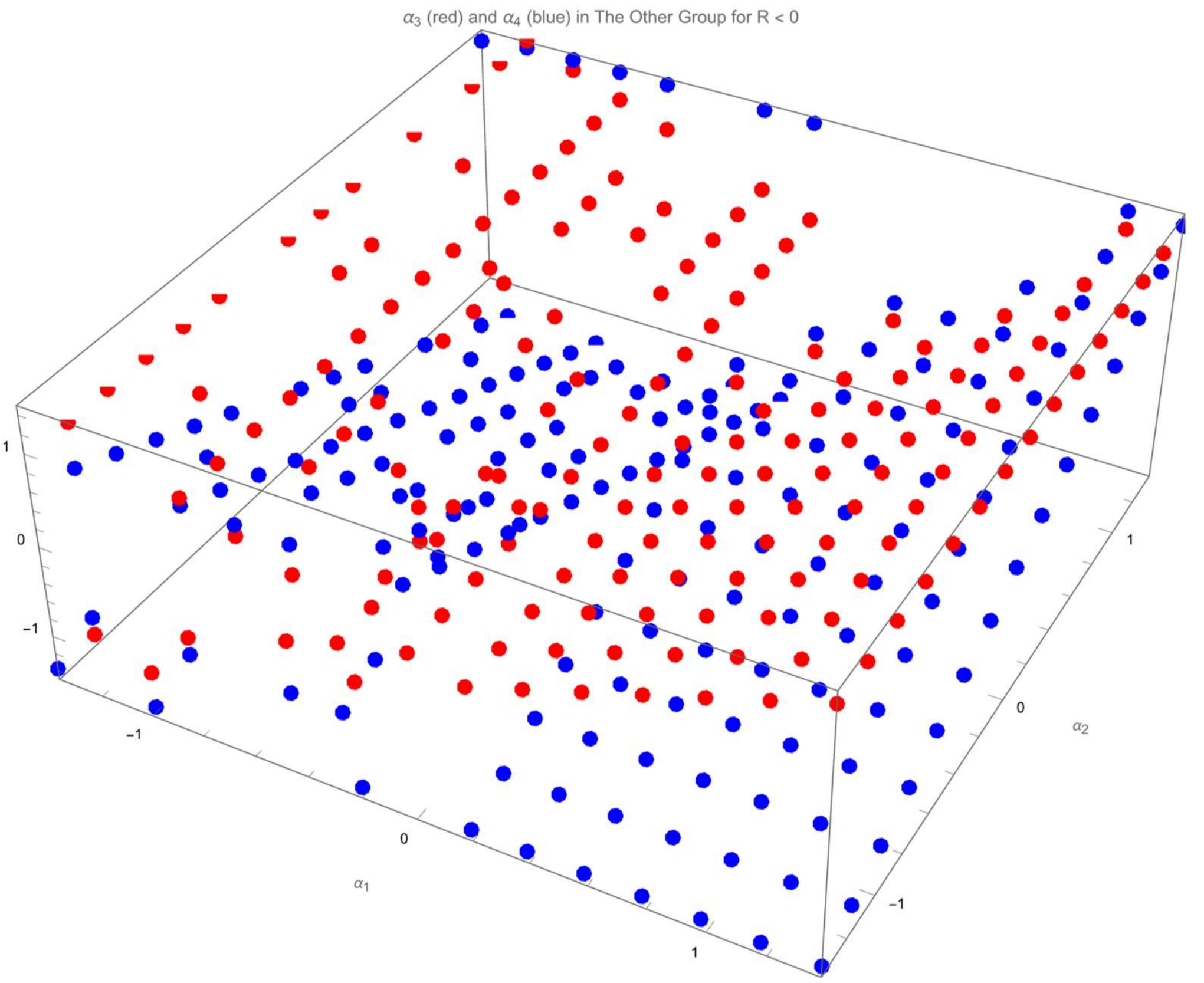

**Figure 16.** The value of $\alpha_3$ and $\alpha_4$ in the other group of the roots of $(\alpha_3,\alpha_4)$ receiving negative $R(\alpha_1,\alpha_2,\alpha_3,\alpha_4)$. The red points represent $\alpha_3$. The blue points represent $\alpha_4$.

Both roots of $(\alpha_3,\alpha_4)$ receiving negative $R(\alpha_1,\alpha_2,\alpha_3,\alpha_4)$, Figures 15 and 16, covers all seven types: (+,+), (−,−), (+,?), (?,+), (−,?), (?,−), (?,?).

It can be found that for a $(\alpha_1,\alpha_2)$, there can be one or two roots of $(\alpha_3,\alpha_4)$ in Figures 15 and 16. In other words, the number of the roots of $(\alpha_3,\alpha_4)$ receiving negative $R(\alpha_1,\alpha_2,\alpha_3,\alpha_4)$ is 1 or 2. This research will focus only on parts of these gaits, receiving negative $R(\alpha_1,\alpha_2,\alpha_3,\alpha_4)$, $(\alpha_1,\alpha_2)$ of which will be detailed later in Section 4.

Thus far, all the $(\alpha_1,\alpha_2,\alpha_3,\alpha_4)$ satisfying the necessary condition to receive an invertible gait, Formula (7–9) on the four-dimensional gait surface has been classified.

One may plan a gait by varying $(\alpha_1,\alpha_2)$ continuously and recording the corresponding $(\alpha_3,\alpha_4)$. The resulting $(\alpha_1,\alpha_2,\alpha_3,\alpha_4)$ is then taken as a gait. This pattern works in the case $R(\alpha_1,\alpha_2,\alpha_3,\alpha_4) > 0$.

While two obstacles may hinder the application of this pattern when $R(\alpha_1,\alpha_2,\alpha_3,\alpha_4) < 0$. Firstly, there are the regions of $(\alpha_1,\alpha_2)$ returning two roots to $(\alpha_3,\alpha_4)$ when $R(\alpha_1,\alpha_2,\alpha_3,\alpha_4) < 0$; determining the proper root, $(\alpha_3,\alpha_4)$, for these $(\alpha_1,\alpha_2)$ is necessary. Secondly, varying $(\alpha_1,\alpha_2)$ continuously does not necessarily guarantee that $(\alpha_3,\alpha_4)$ is varying continuously. Ironically, one of the advantages of the gait plan is the continuous change in the tilting angles.

With these concerns, a theory helping select the proper $(\alpha_3,\alpha_4)$ for every given $(\alpha_1,\alpha_2)$ is demanded to receive a continuous gait (First Class Continuous Gait) when $R(\alpha_1,\alpha_2,\alpha_3,\alpha_4) < 0$.

**Definition 1.** (First Class Continuous Gait) A gait $\left(\alpha_1(t),\alpha_2(t),\alpha_3(t),\alpha_4(t)\right)$ is a First Class Continuous Gait if and only if

$$\forall t_1 \geqslant 0, \lim_{t\to t_1^{+}} \alpha_i(t) = \lim_{t\to t_1^{-}} \alpha_i(t) = \alpha_i(t_1), i = 1,2,3,4.$$

In other words, a gait belongs to First Class Continuous Gait if each tilting angle of this gait is continuous.

A complete exploration of all First Class Continuous Gait on the four-dimensional gait surface is beyond the scope of this research. Instead, we only analyze parts of the First Class Continuous Gait that meet the following condition (Second Class Continuous Gait).

**Definition 2.** (Second Class Continuous Gait) Given $\alpha_1$ and $\alpha_2$ at time $t$, denote $\alpha_3$ and $\alpha_4$ on the Four-dimensional Gait Surface as $\alpha_3(\alpha_1(t),\alpha_2(t))$ and $\alpha_4(\alpha_1(t),\alpha_2(t))$, respectively. A gait $(\alpha_1(t),\alpha_2(t),\alpha_3(t),\alpha_4(t))$ is a Second Class Continuous Gait if and only if $\forall t_1 \geqslant 0$, all the following conditions are satisfied.

1. $\alpha_1(t_1), \alpha_2(t_1), \alpha_3(\alpha_1(t_1),\alpha_2(t_1)), \alpha_4(\alpha_1(t_1),\alpha_2(t_1))$ exist.

2. $\begin{vmatrix} \dot{\alpha}_1(t_1) & \dot{\alpha}_2(t_1) \\ 1 & 0 \end{vmatrix} = 0$ or $\begin{vmatrix} \dot{\alpha}_1(t_1) & \dot{\alpha}_2(t_1) \\ 0 & 1 \end{vmatrix} = 0$.

3. $\|[\dot{\alpha}_1(t_1) \quad \dot{\alpha}_2(t_1)]\|_2 \neq 0$.

4. $\lim_{t\to t_1^+} \alpha_i(t) = \lim_{t\to t_1^-} \alpha_i(t) = \alpha_i(t_1), i = 1{,}2$.

5. $\lim_{t\to t_1^+} \alpha_j(\alpha_1(t),\alpha_2(t)) = \lim_{t\to t_1^-} \alpha_j(\alpha_1(t),\alpha_2(t)) = \alpha_j(\alpha_1(t_1),\alpha_2(t_1)), j = 3{,}4$.

In plain words, these conditions define the directions of the $[\alpha_1(t) \quad \alpha_2(t)]$ while the consequent $\alpha_3(\alpha_1(t),\alpha_2(t))$ and $\alpha_4(\alpha_1(t),\alpha_2(t))$ are required to be continuous. Without further specifications, the "continuous gait" in the rest of this article refers to second class continuous gait.

As mentioned, designing a gait by varying $\alpha_1$ and $\alpha_2$, obeying Point 1–4 in the definition of Second Class Continuous Gait, and recording the corresponding $\alpha_3$ and $\alpha_4$, on the Four-dimensional Gait Surface, without a rule may result in a gait violating Point 5.

The Two Color Map Theorem in the next section instructs the choice of these corresponding $\alpha_3$ and $\alpha_4$ to guarantee Point 5.

## 4. Two Color Map Theorem

The continuous gait is guaranteed by a sophisticated $(\alpha_3,\alpha_4)$-picking theorem, Two Color Map Theorem. It is an augmented rule for varying $(\alpha_1,\alpha_2,\alpha_3,\alpha_4)$ while $R(\alpha_1,\alpha_2,\alpha_3,\alpha_4) < 0$. Without specification, the region in the rest of this section represents the region of $(\alpha_1,\alpha_2)$ with negative $R(\alpha_1,\alpha_2,\alpha_3,\alpha_4)$.

Recalling $(\alpha_3,\alpha_4)$ receiving $R(\alpha_1,\alpha_2,\alpha_3,\alpha_4) < 0$, there are altogether seven types of $(\alpha_3,\alpha_4)$ in this case: (+,+), (−,−), (+,?), (?,+), (−,?), (?,−), (?,?).

Notice that "+" plane and "−" plane of $\alpha_3$ intersects at "?", which is a line. In addition, "+" plane and "−" plane of $\alpha_4$ intersects at a line marked by "?". $\alpha_3$ is not allowed to move from "+" to "−" or to move from "−" to "+". Both cases will introduce the discontinuous change of $\alpha_3$. Similarly, $\alpha_4$ is not allowed to move from "+" to "−" or to move from "−" to "+". Both cases will introduce the discontinuous change of $\alpha_4$.

To change the plane, the points on the intersecting lines "?" of the two planes, "+" and "−", is required to be an intermediate state. For example, the relevant changes from "+" to "?" to "−" and from "−" to "?" to "+" are both continuous.

The allowed types for two adjacent $\alpha_3$ or $\alpha_4$ are paired in Figure 17a. There are three types of $\alpha_3$: locating on "+" surface, locating on "−" surface, and locating on the intersection of two planes (marked as "?"). $\alpha_3$ is not allowed to change the plane while $(\alpha_1,\alpha_2)$ is varying continuously. Thus, $\alpha_3$, lying on "+" plane is supposed to remain on "+" plane or move to the intersection of two planes, which is "?". Similarly, $\alpha_3$, lying on "−" plane is supposed to remain on "−" plane or move to the intersection of two planes, which is "?". Specially, $\alpha_3$, lying on "?" is allowed to move to "+" plane, to "−" plane, or to another "?". Obviously, the rule for $\alpha_4$ is identical to the one for $\alpha_3$.

The allowed types for two adjacent $(\alpha_3,\alpha_4)$ are paired in Figure 17b where the case (?,?) is omitted since it accommodates all types of adjacent roots. The first element in the plane type of $(\alpha_3,\alpha_4)$ represents the type of the plane on which $\alpha_3$ lies. The pairing rule for the first element follows Figure 17a. The second element in the plane type of $(\alpha_3,\alpha_4)$ represents the type of the plane on which $\alpha_4$ lies. The pairing rule for the second element also follows Figure 17a.

The allowed adjacent roots and root pairs.

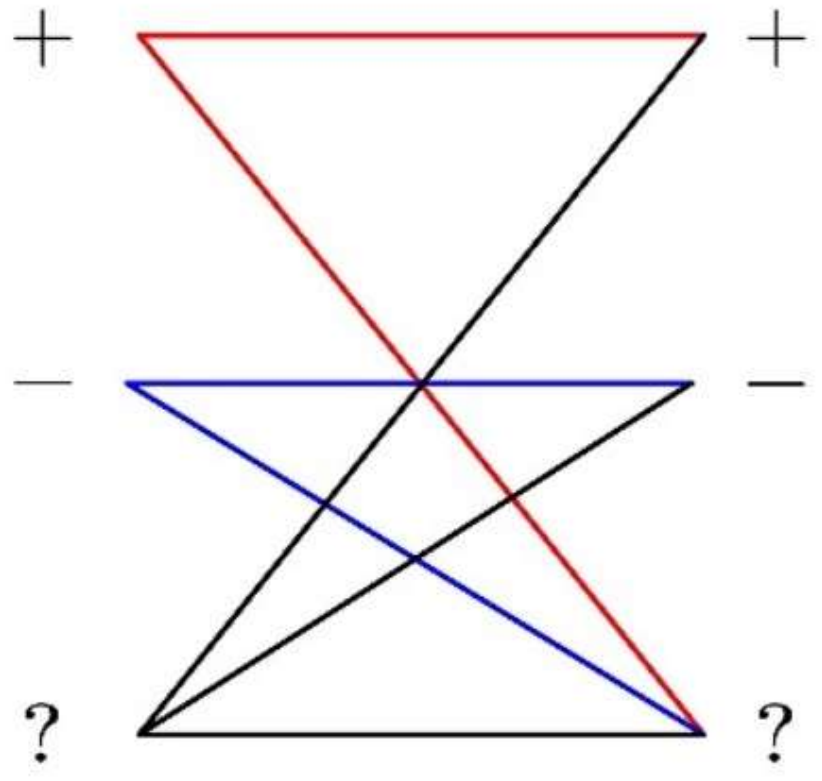

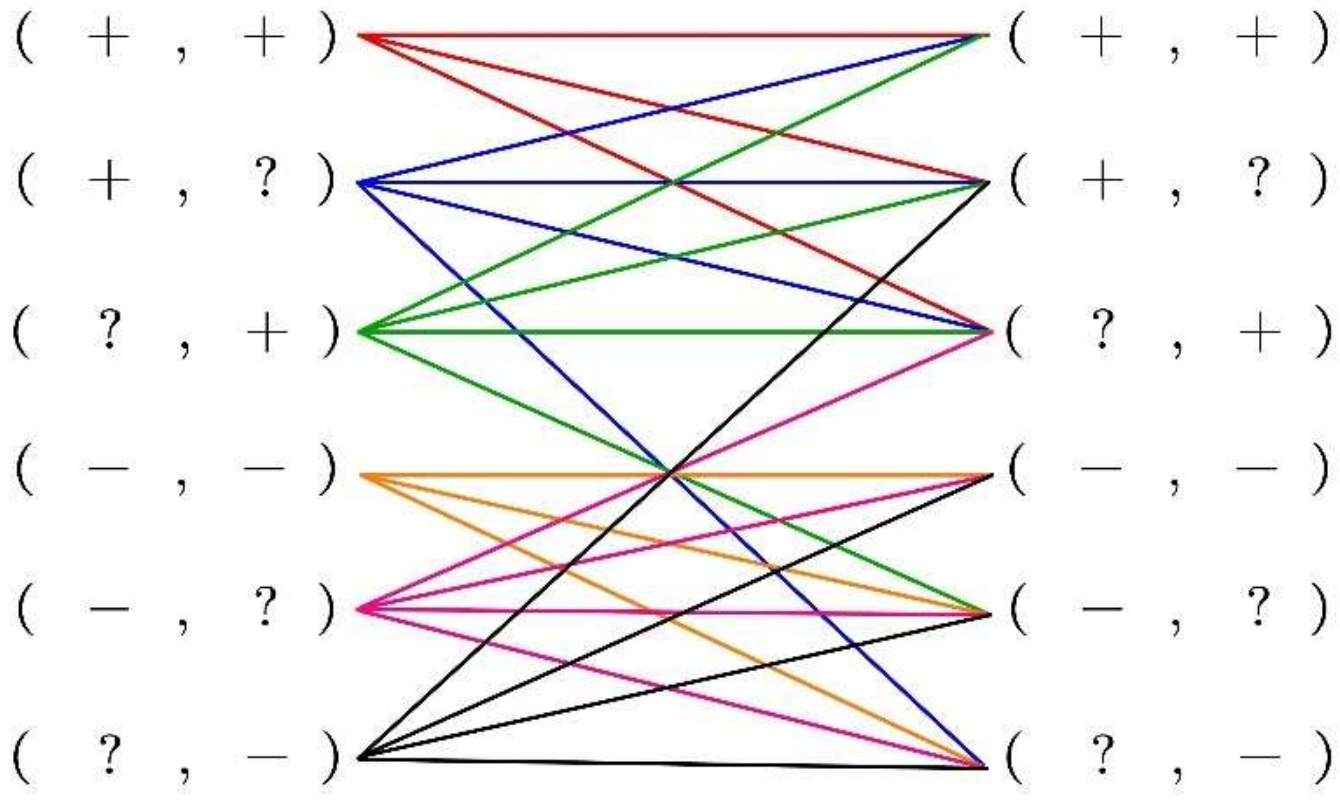

**Figure 17.** The allowed adjacent roots are lined: (**a**) The allowed types of the adjacent $\alpha_3$ or $\alpha_4$; (**b**) The allowed types of the adjacent ($\alpha_3$,$\alpha_4$). The case (?,?) is not plotted.

Paint ($\alpha_1$,$\alpha_2$) red if the corresponding ($\alpha_3$,$\alpha_4$) belongs to (+,+). Paint ($\alpha_1$,$\alpha_2$) blue if the corresponding ($\alpha_3$,$\alpha_4$) belongs to (−,−). Paint ($\alpha_1$,$\alpha_2$) half red half blue if there are two corresponding ($\alpha_3$,$\alpha_4$), which belong to (−,−) and (+,+), respectively. Marking the rest categories directly, Figure 18 paints the whole region of ($\alpha_1$,$\alpha_2$) of interest. It is parts of the regions satisfying $R(\alpha_1,\alpha_2,\alpha_3,\alpha_4) < 0$.

It can be seen that most ($\alpha_1$,$\alpha_2$) receive two applicable ($\alpha_3$,$\alpha_4$), belonging to (+,+) and (−,−), respectively. While ($\alpha_1$,$\alpha_2$) in the triangular region on the top, receive one applicable ($\alpha_3$,$\alpha_4$) only, the type of which is (−,−).

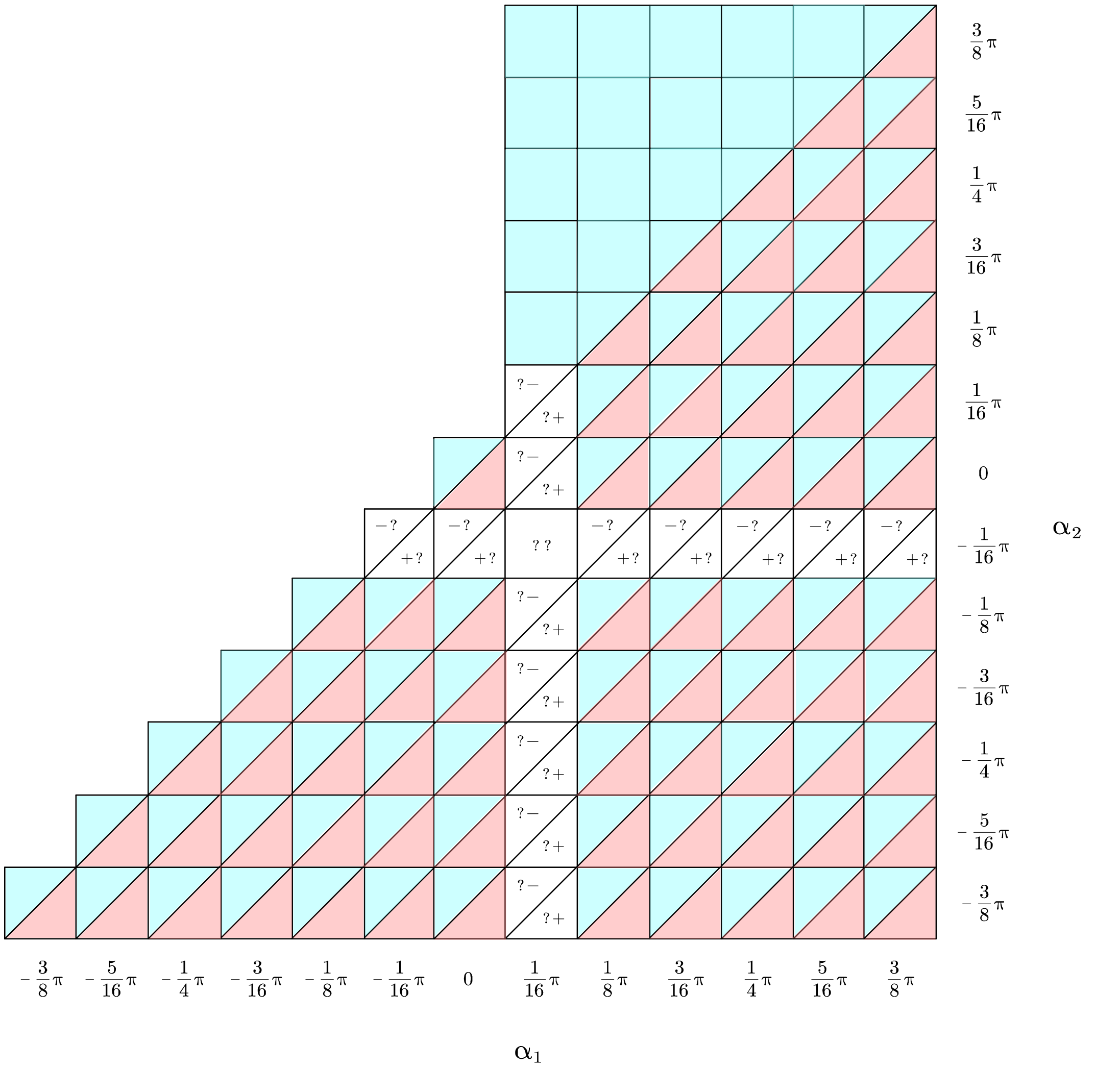

**Figure 18.** The types of the corresponding ($\alpha_3$,$\alpha_4$), resulting in $R(\alpha_1,\alpha_2,\alpha_3,\alpha_4) < 0$, on parts of the $\alpha_1 - \alpha_2$ plane. ($\alpha_1$,$\alpha_2$) receiving ($\alpha_3$,$\alpha_4$) of type (+,+) are painted in red. ($\alpha_1$,$\alpha_2$) receiving ($\alpha_3$,$\alpha_4$) of type (−,−) are painted in blue. ($\alpha_1$,$\alpha_2$) receiving two ($\alpha_3$,$\alpha_4$) of both types are painted in half red half blue. The rest uncolored ($\alpha_1$,$\alpha_2$) are marked based on the types of the corresponding ($\alpha_3$,$\alpha_4$), some of which have 2 roots.

As seen in Figure 18, some ($\alpha_1$,$\alpha_2$) receive 2 roots of ($\alpha_3$,$\alpha_4$). One can pick either root as wished, which can be interpreted as a color decision process: decide blue or red for the ($\alpha_1$,$\alpha_2$) painted in half red half blue in Figure 18.

The gait plan problem is then equivalent to the following processes: 1. Find an enclosed curve (including an overlapping curve with zero size) in ($\alpha_1$,$\alpha_2$) plane in Figure 18; 2. Decide the color for ($\alpha_1$,$\alpha_2$) (or decide ($\alpha_3$,$\alpha_4$)) traveled by the enclosed curve found; 3. Specify the decided four-dimensional curve with direction and time.

The following Two Color Map Theorem restricts the selection of ($\alpha_3$,$\alpha_4$), leading to a continuous gait.

**Theorem 1.** (Two Color Map Theorem) *The planned gait is continuous if the color selections of* ($\alpha_1$,$\alpha_2$) *on the enclosed curve meet the following requirement: The adjacent* ($\alpha_1$,$\alpha_2$) *are in the same color or do not violate the rule in Figure 17.*

**Theorem 2.** (Proof of Two Color Map Theorem) The same colored adjacent ($\alpha_1$,$\alpha_2$) indicate that $\alpha_3$ and $\alpha_4$ will not change the type of the plane when ($\alpha_1$,$\alpha_2$) varies. Further, since ($\alpha_1$,$\alpha_2$) varies continuously, ($\alpha_3$,$\alpha_4$) is consequently continuous. The planned gait ($\alpha_1$,$\alpha_2$,$\alpha_3$,$\alpha_4$) is a continuous gait. □

Thus far, the method of planning a continuous gait robust to the attitude change in a tilt-rotor has been elucidated. Several examples are given in the next section.

## 5. Analysis of Typical Acceptable Gaits

This section evaluates four different gaits. Three of them satisfy $R(\alpha_1,\alpha_2,\alpha_3,\alpha_4) < 0$ for any given time. The rest one satisfies $R(\alpha_1,\alpha_2,\alpha_3,\alpha_4) > 0$ for any given time. In comparison, the relevant biased gaits (by scaling) are also evaluated.

The gaits satisfying $R(\alpha_1,\alpha_2,\alpha_3,\alpha_4) < 0$, analyzed in this section are plotted in Figure 19. Based on this choice, the color of all $(\alpha_1,\alpha_2)$ traveled by the enclosed curve, Gait 1 in Figure 19, are painted blue based on the Two Color Map Theorem.

On the other hand, the color of $(\alpha_1,\alpha_2)$ traveled by the enclosed curve, Gait 2 and Gait 3 in Figure 19, can be painted either all in blue or all in red based on the Two Color Map Theorem.

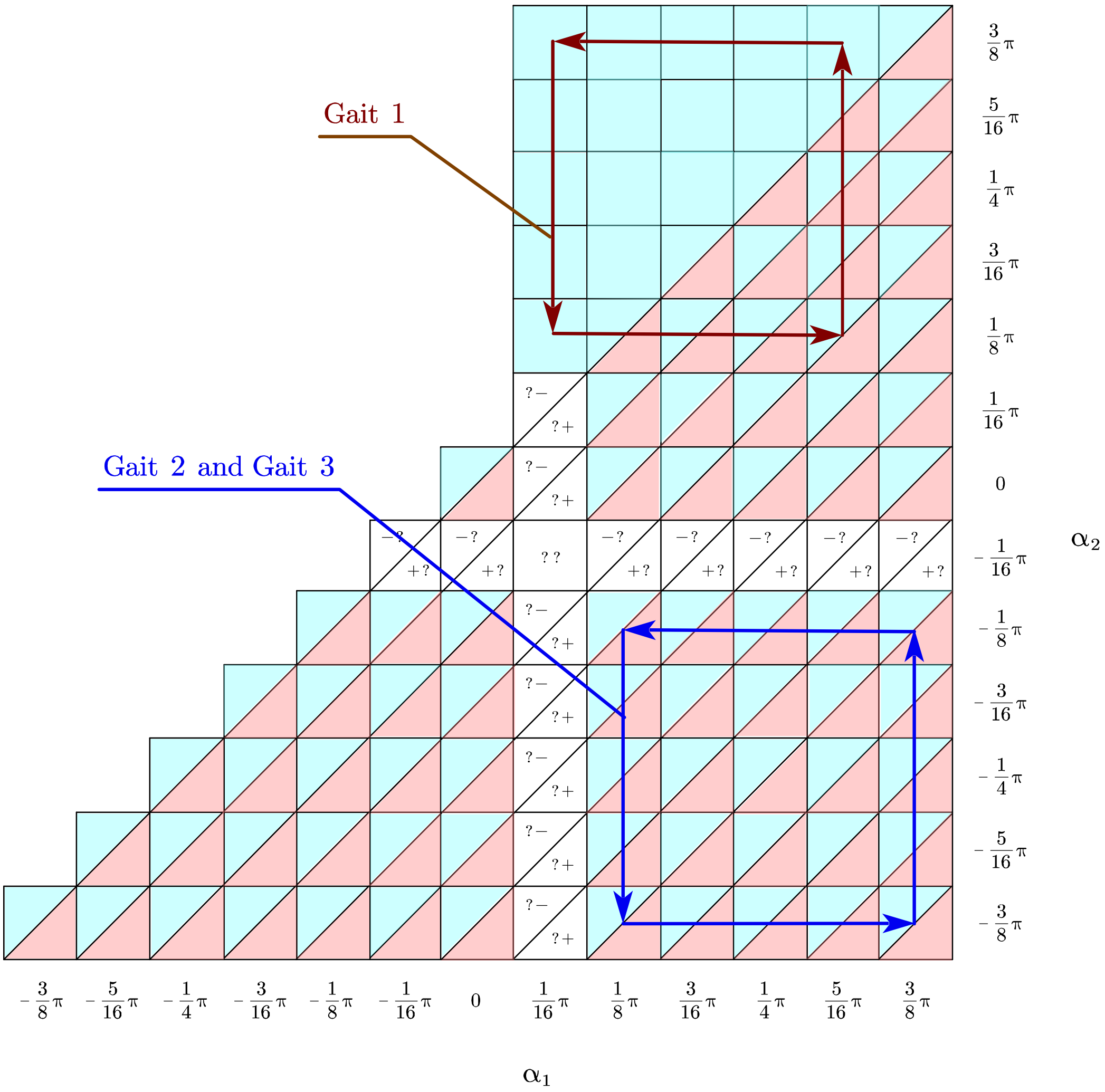

**Figure 19.** The projections of all three designed gaits satisfying $R(\alpha_1,\alpha_2,\alpha_3,\alpha_4) < 0$ are the enclosed curves. The colors of $(\alpha_1,\alpha_2)$ traveled by each gait should satisfy Two Color Map Theorem.

As for the gait (Gait 4, not sketched), satisfying $R(\alpha_1,\alpha_2,\alpha_3,\alpha_4) > 0$, there is only one root of $(\alpha_3,\alpha_4)$, belonging to (+,+), corresponding to a given $(\alpha_1,\alpha_2)$ (see Figures 11 and 13). Varying $(\alpha_1,\alpha_2)$ continuously naturally guarantees a continuous change in $(\alpha_1,\alpha_2,\alpha_3,\alpha_4)$.

The robustness to the disturbance in the roll angle and pitch angles of each proposed gait is evaluated. The unacceptable attitudes (roll and pitch) for each gait are found by violating Formula (1), e.g., equal the left side of Formula (1) zero.

The biased gaits not on the four-dimensional gait surface are coined to compare the robustness with the gaits on the four-dimensional gait surface. These biased gaits are generated by maintaining ($\alpha_1$,$\alpha_2$) identical to the acceptable gaits while scaling ($\alpha_3$,$\alpha_4$) throughout the entire gait:

$$\alpha_1 \leftarrow \alpha_1, \tag{12}$$

$$\alpha_2 \leftarrow \alpha_2, \tag{13}$$

$$\alpha_3 \leftarrow \eta \cdot \alpha_3, \tag{14}$$

$$\alpha_4 \leftarrow \eta \cdot \alpha_4, \tag{15}$$

where $\eta \in (0,1)$ is the scaling coefficient, which will be specified in each gait.

Note that though scaling evenly each tilting angle has been proven a valid approach in modifying a gait receiving a singular decoupling matrix [27], the discussions on scaling unevenly, e.g., scaling $\alpha_3$ and $\alpha_4$ only, have not been addressed yet.

The period of the gait ($T$) is set as 1 s.

Define the ($\alpha_1$,$\alpha_2$,$\alpha_3$,$\alpha_4$) as a vertex of a four-dimensional curve if and only if

$$\alpha_1 \in \left\{\min_{t\in T}(\alpha_1),\max_{t\in T}(\alpha_1)\right\}, \tag{16}$$

$$\alpha_2 \in \left\{\min_{t\in T}(\alpha_2),\max_{t\in T}(\alpha_2)\right\}, \tag{17}$$

$$\alpha_3 \in \left\{\min_{t\in T}(\alpha_3),\max_{t\in T}(\alpha_3)\right\}, \tag{18}$$

$$\alpha_4 \in \left\{\min_{t\in T}(\alpha_4),\max_{t\in T}(\alpha_4)\right\}. \tag{19}$$

The number of the vertices of the four-dimensional curve in each designed gait in this research is four.

### *5.1. Gait 1*

All ($\alpha_1$,$\alpha_2$) projected by the four-dimensional curve (Gait 1) are painted in blue to meet Two Color Map Theorem.

The vertices of the gait ($\alpha_1$,$\alpha_2$,$\alpha_3$,$\alpha_4$) are: $(5\pi/16,\pi/8,-0.648,-0.727)$ , $(5\pi/16,3\pi/8,-0.648,-1.512)$, $(\pi/16,3\pi/8,0.138,-1.512)$, $(\pi/16,\pi/8,0.138,-0.727)$. Gait 1 is plotted in Figure 20.

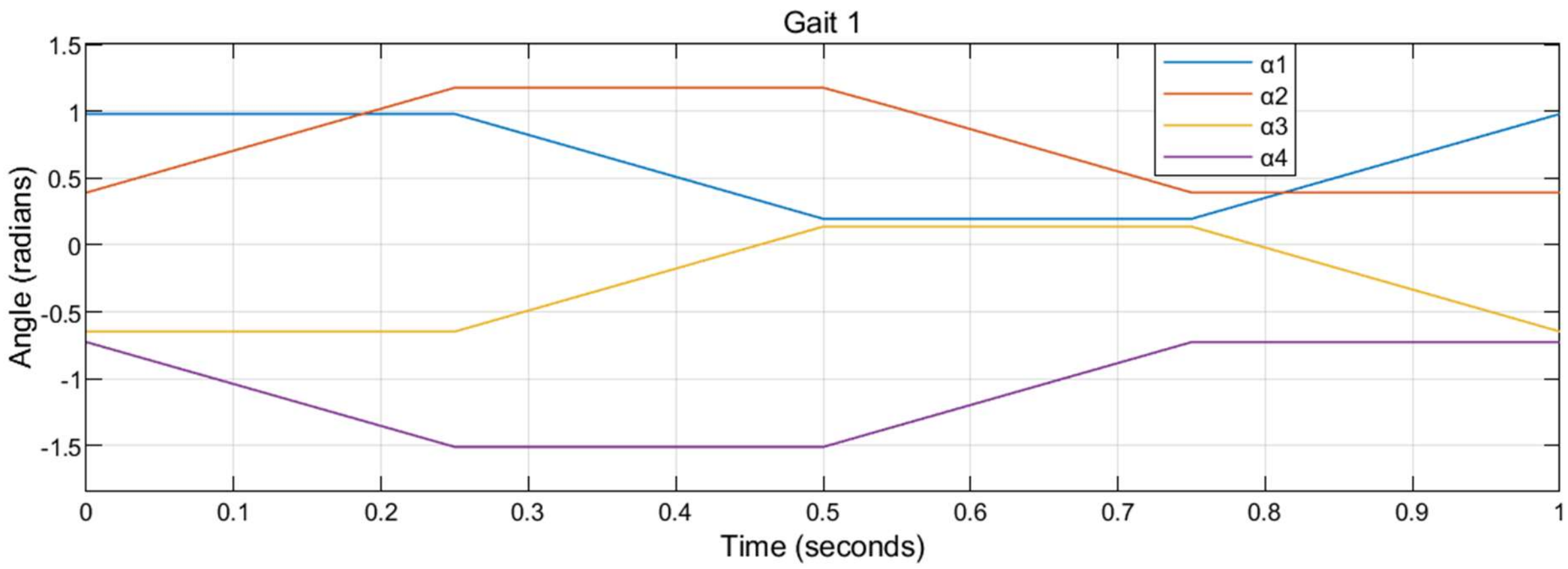

**Figure 20.** The tilting angle of each propeller in Gait 1.

In comparison, the corresponding biased gait for Gait 1 is coined by setting the scaling coefficient, $\eta$ in Formulas (14) and (15), as 80%.

Figure 21 plots the unacceptable attitudes for the gait on the four-dimensional gait surface (red curves) and for the biased gait (blue curves).

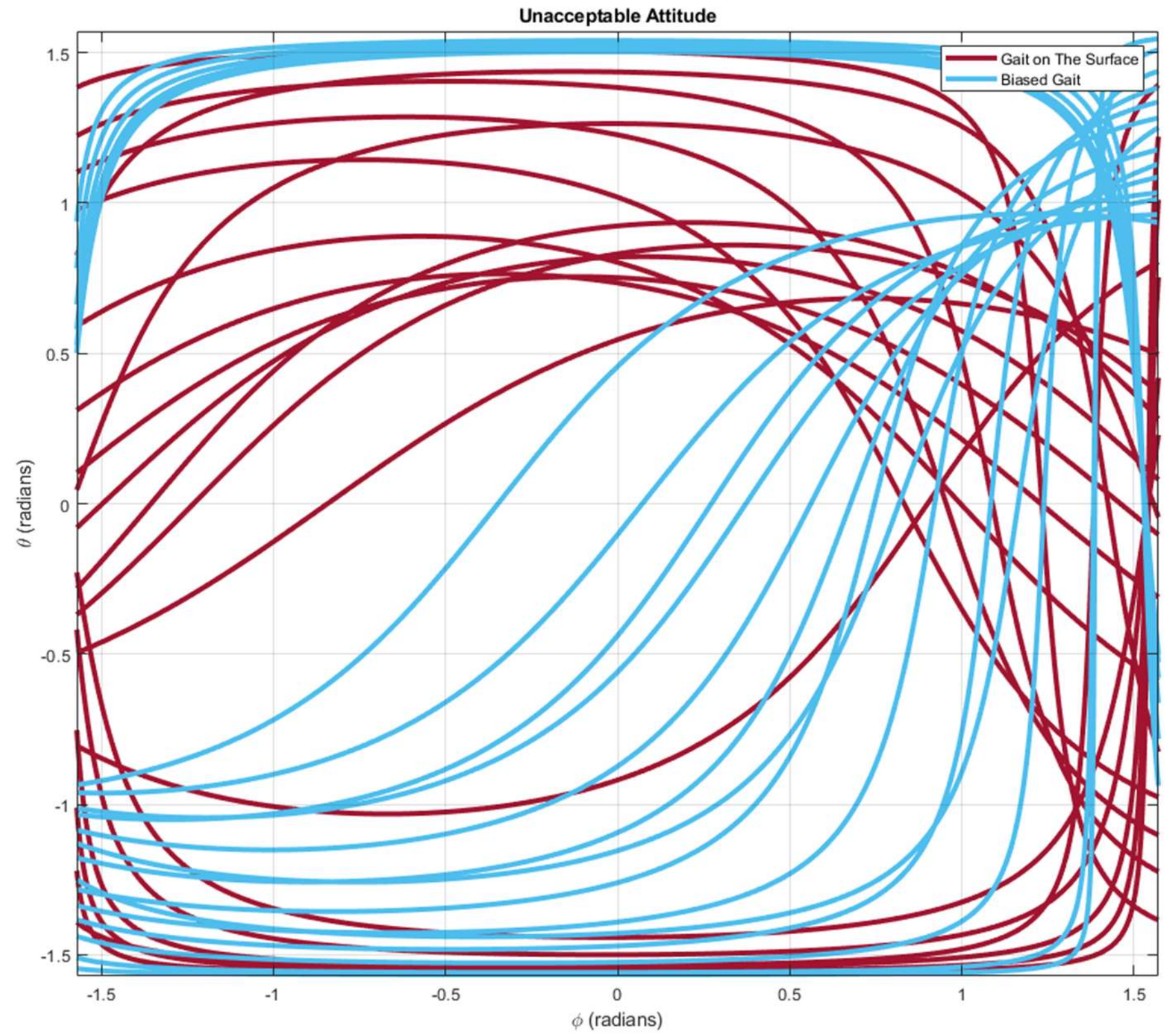

**Figure 21.** The unacceptable attitudes for Gait 1 and for biased Gait 1. The red curve represents the attitudes receiving the singular decoupling matrix in Gait 1. The blue curve represents the attitudes receiving the singular decoupling matrix in biased Gait 1.

It can be seen that the red curves are generally further to the origin, $(\phi,\theta) = (0,0)$, comparing with the blue curves, demonstrating that the gait on the four-dimensional gait surface is allowed to move in a wider region of attitude.

*5.2. Gait 2 and 3*

All $(\alpha_1,\alpha_2)$ projected by the four-dimensional curve (Gait 2 and Gait 3) can either be painted in blue or in red to meet Two Color Map Theorem. This leads to Gait 2 (all in blue) and Gait 3 (all in red).

The vertices of the tilting angles $(\alpha_1,\alpha_2,\alpha_3,\alpha_4)$ in Gait 2 are: $(3\pi/8,-3\pi/8,-0.844,0.844)$, $(3\pi/8,-\pi/8,-0.844,0.059)$, $(\pi/8,-\pi/8,-0.059,0.059)$, $(\pi/8,-3\pi/8,-0.059,0.844)$. Gait 2 is plotted in Figure 22. The corresponding biased gait for Gait 2 is coined by setting the scaling coefficient, $\eta$ in Formulas (14) and (15), as 99%.

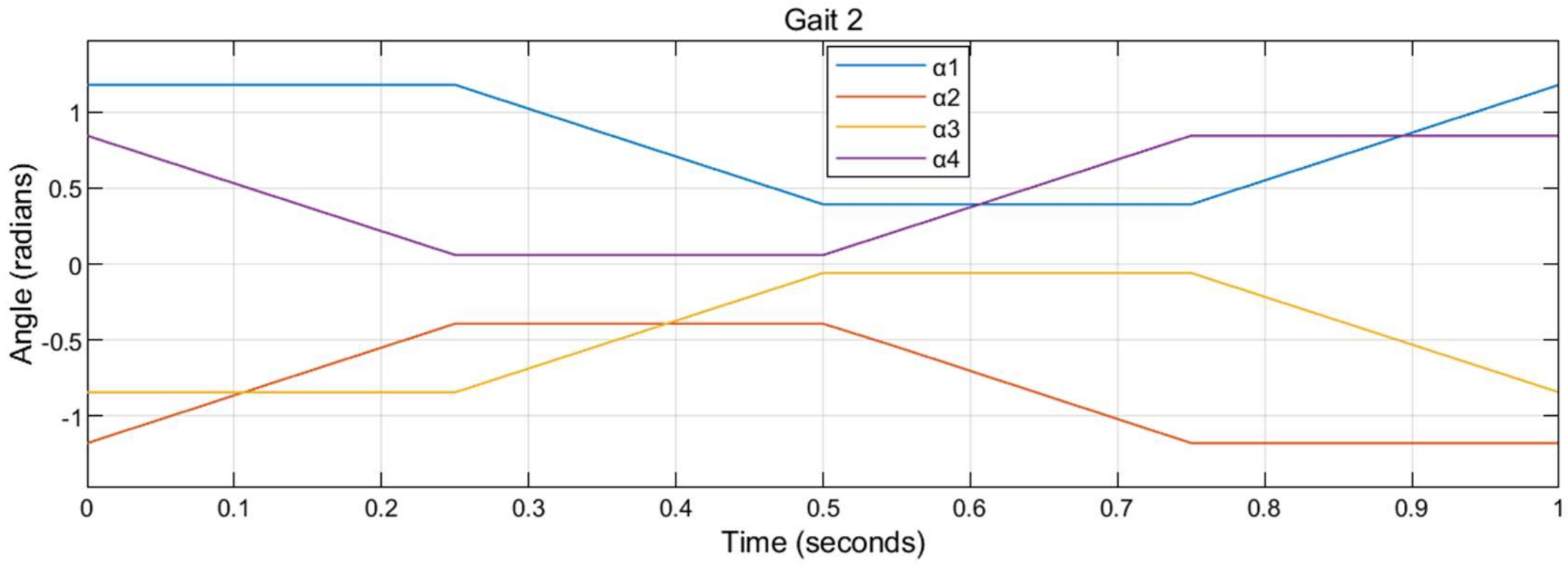

**Figure 22.** The tilting angle of each propeller in Gait 2.

Figure 23 plots the unacceptable attitudes for the gait on the four-dimensional gait surface (red curves) and for the biased gait (blue curves).

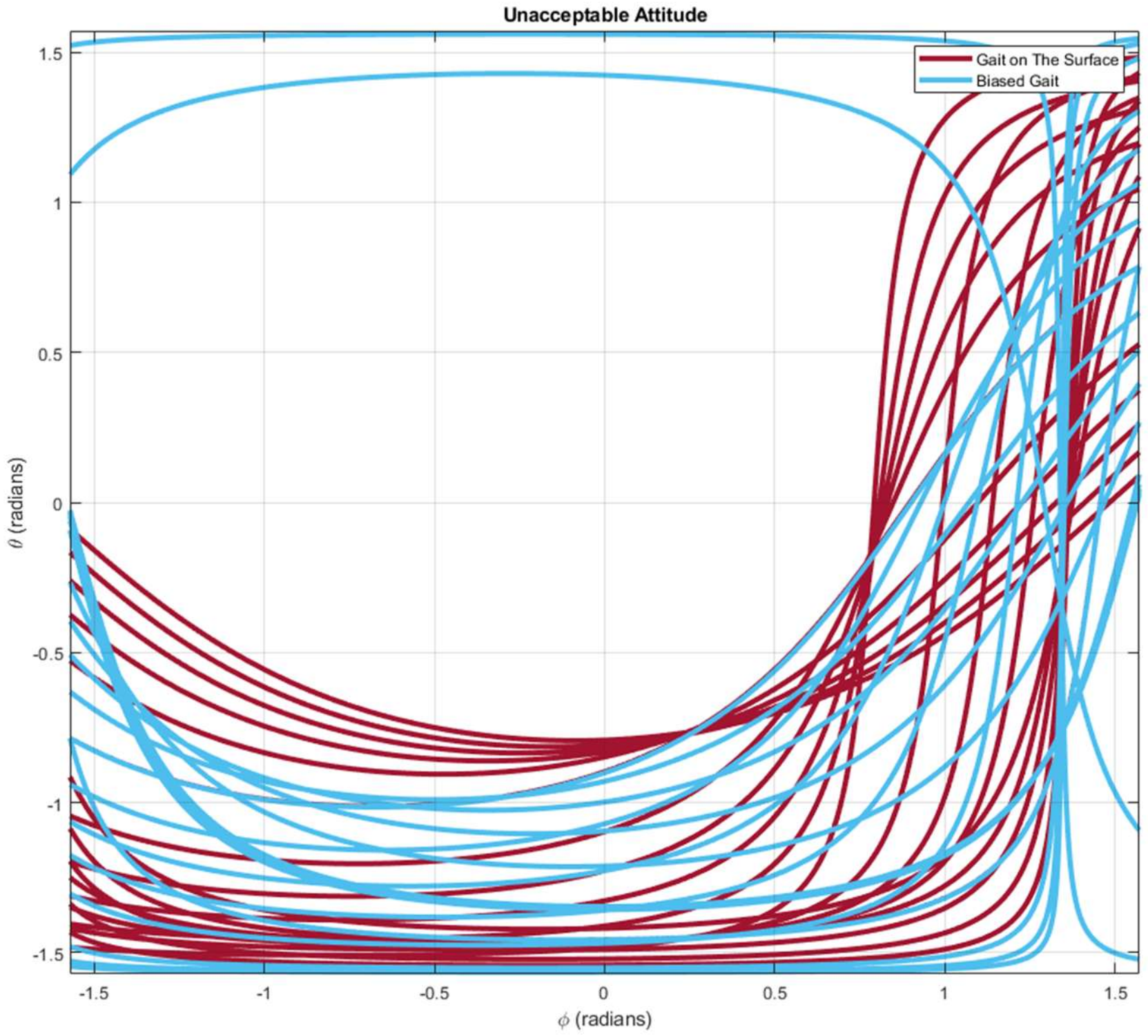

**Figure 23.** The unacceptable attitudes for Gait 2 and for biased Gait 2. The red curve represents the attitudes receiving the singular decoupling matrix in Gait 2. The blue curve represents the attitudes receiving the singular decoupling matrix in biased Gait 2.

Similarly, the red curves are generally further to the origin, $(\phi,\theta) = (0,0)$, comparing with the blue curves, demonstrating that the gait on the four-dimensional gait surface is allowed to move in a wider region of attitude.

The vertices of the tilting angles $(\alpha_1,\alpha_2,\alpha_3,\alpha_4)$ in Gait 3 are: $(3\pi/8,-3\pi/8,1.178,-1.178)$, $(3\pi/8,-\pi/8,1.178,-0.393)$, $(\pi/8,-\pi/8,0.393,-0.393)$, $(\pi/8,-3\pi/8,0.393,-1.178)$. Note that $\alpha_1 = \alpha_3$, $\alpha_2 = \alpha_4$ in this case. Gait 3 is plotted in Figure 24. The corresponding biased gait for Gait 3 is coined by setting the scaling coefficient, $\eta$ in Formulas (14) and (15), as 80%.

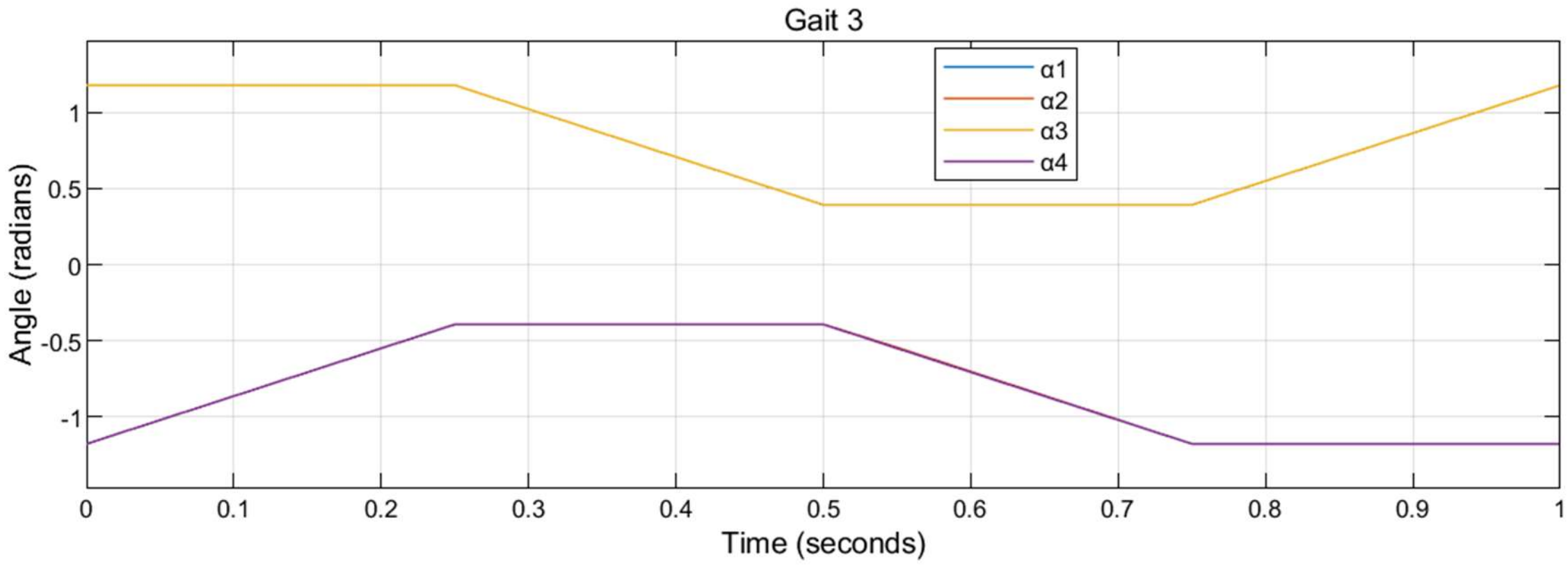

**Figure 24.** The tilting angle of each propeller in Gait 3. Note that $\alpha_1$ and $\alpha_3$ overlap. $\alpha_2$ and $\alpha_4$ overlap.

Figure 25 plots the unacceptable attitudes for the gait on the four-dimensional gait surface (red curves) and for the biased gait (blue curves).

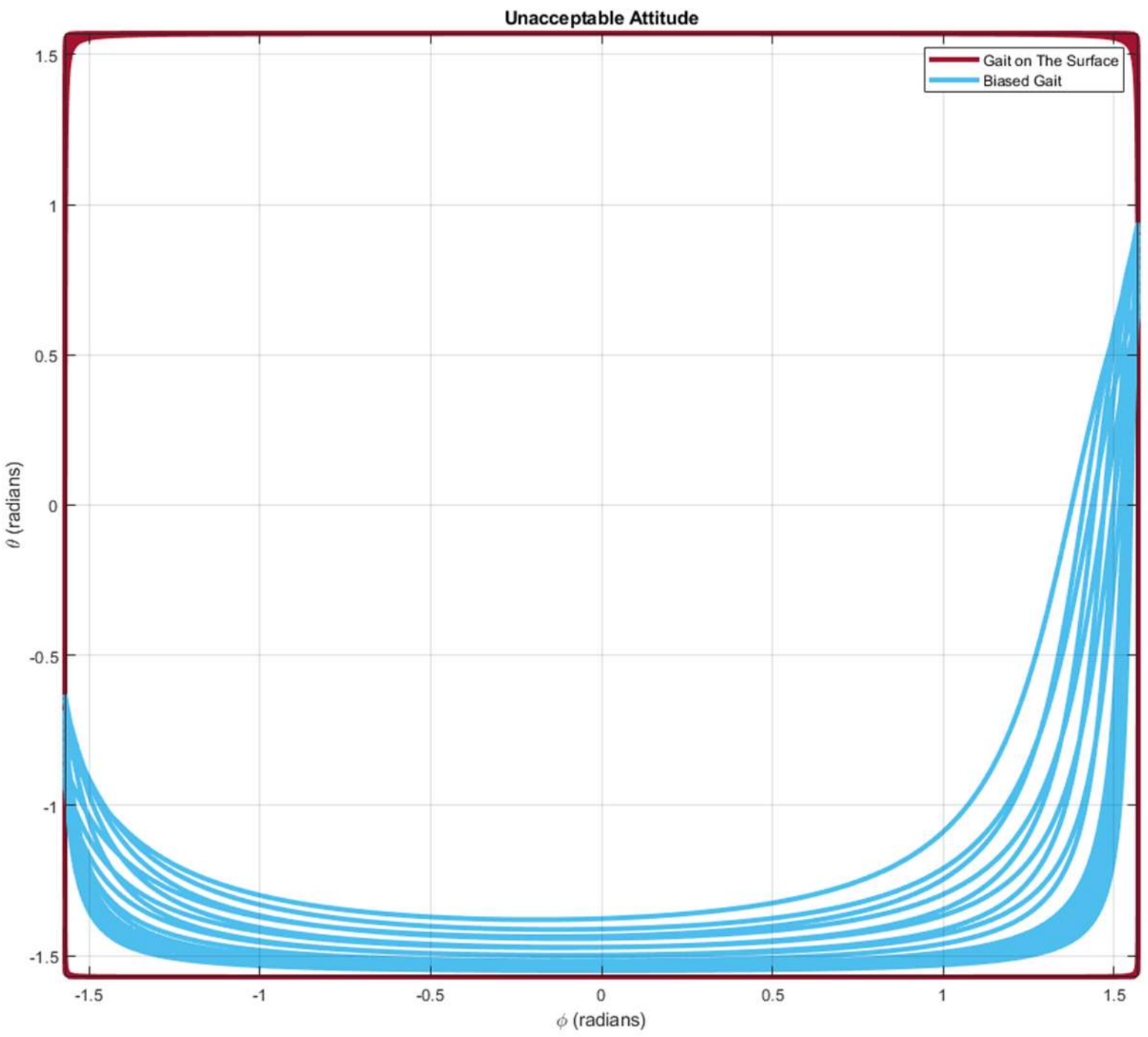

**Figure 25.** The unacceptable attitudes for Gait 3 and for biased Gait 3. The red curve represents the attitudes receiving the singular decoupling matrix in Gait 3. The blue curve represents the attitudes receiving the singular decoupling matrix in biased Gait 3.

Clearly, the red curves are generally further to the origin, $(\phi,\theta) = (0,0)$, comparing with the blue curves, demonstrating that the gait on the four-dimensional gait surface is allowed to move in a wider region of attitude. The region of admissible attitudes enlarges nearly to the whole region of $\phi \in (-\pi/2,\pi/2), \theta \in (-\pi/2,\pi/2)$.

### *5.3. Gait 4*

Gait 4 is designed on the four-dimensional gait surface satisfying $R(\alpha_1,\alpha_2,\alpha_3,\alpha_4) > 0$ (Figures 11 and 13). Only one root of $(\alpha_3,\alpha_4)$ exists for a given $(\alpha_1,\alpha_2)$ in this region, the triangular area of interest on $\alpha_1 - \alpha_2$ plane is governed by three vertices: $(\alpha_1,\alpha_2) = (-7\pi/16,7\pi/16)$, $(\alpha_1,\alpha_2) = (-7\pi/16,-5\pi/16)$, $(\alpha_1,\alpha_2) = (5\pi/16,7\pi/16)$.

Moreover, $(\alpha_3,\alpha_4)$ varies continuously if $(\alpha_1,\alpha_2)$ varies continuously since all $(\alpha_1,\alpha_2)$ in this region, satisfying $R(\alpha_1,\alpha_2,\alpha_3,\alpha_4) > 0$ belong to $(+,+)$.

The vertices of the tilting angles $(\alpha_1,\alpha_2,\alpha_3,\alpha_4)$ in Gait 4 are: $(-\pi/8,\pi/8,-0.393,0.393)$, $(-\pi/8,3\pi/8,-0.393,1.178)$, $(-3\pi/8,3\pi/8,-1.178,1.178)$, $(-3\pi/8,\pi/8,-1.178,0.393)$. Note that $\alpha_1 = \alpha_3$, $\alpha_2 = \alpha_4$ in this case. Gait 4 is plotted in Figure 26. The corresponding biased gait for Gait 4 is coined by setting the scaling coefficient, $\eta$ in Formulas (14) and (15), as 80%.

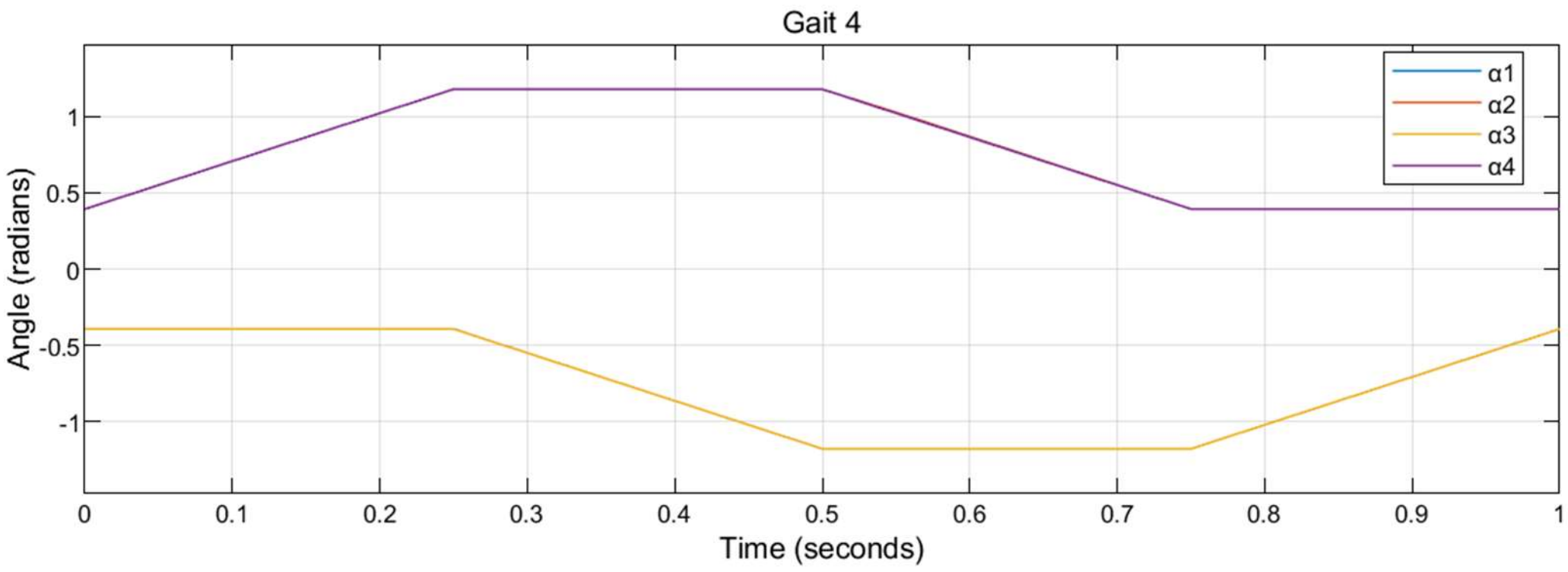

**Figure 26.** The tilting angle of each propeller in Gait 4. Note that $\alpha_1$ and $\alpha_3$ overlap. $\alpha_2$ and $\alpha_4$ overlap.

Figure 27 plots the unacceptable attitudes for the gait on the four-dimensional gait surface (red curves) and for the biased gait (blue curves).

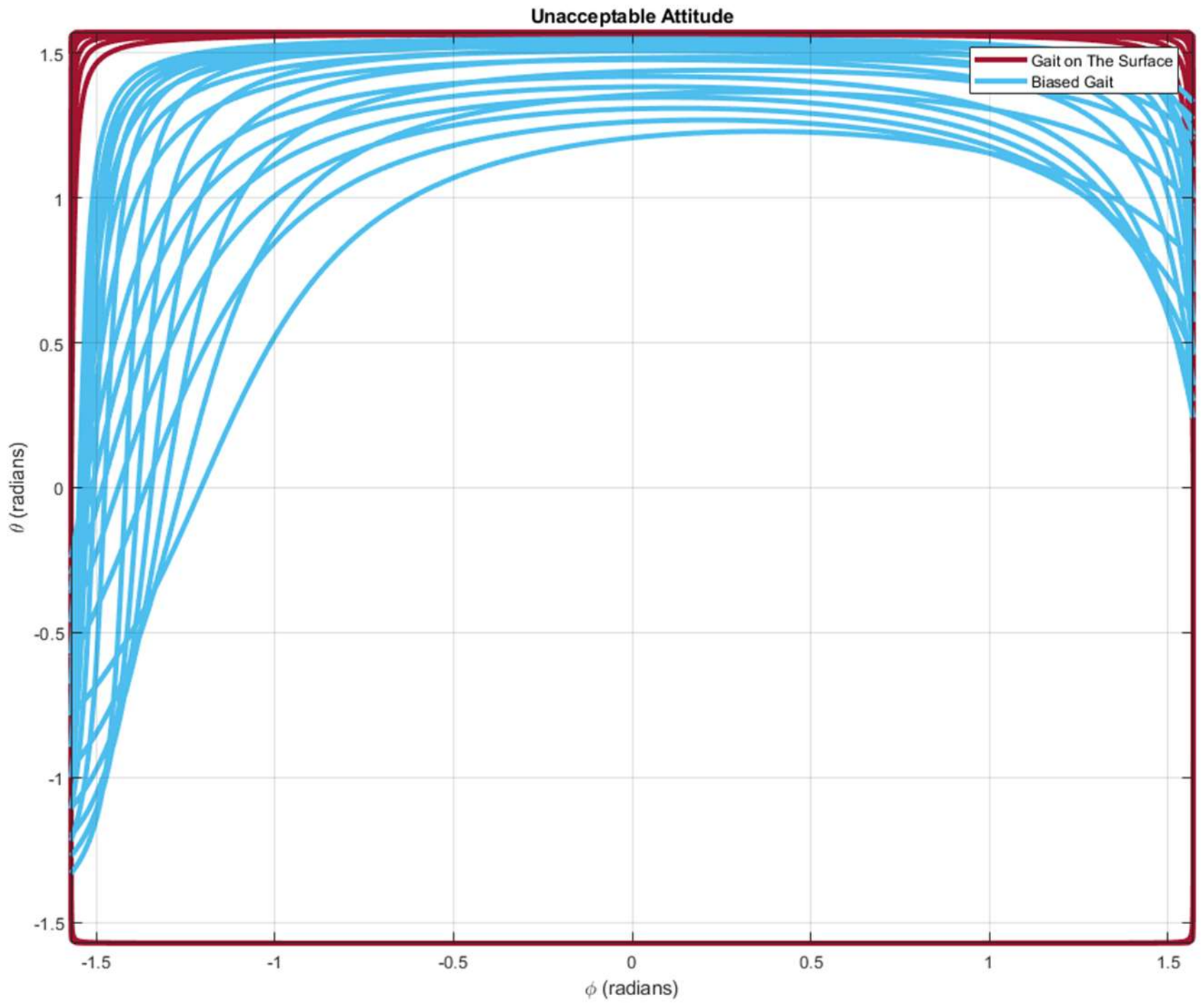

**Figure 27.** The unacceptable attitudes for Gait 4 and for biased Gait 4. The red curve represents the attitudes receiving the singular decoupling matrix in Gait 4. The blue curve represents the attitudes receiving the singular decoupling matrix in biased Gait 4.

Similar to the result in Figure 25, the red curves in Figure 27 are also generally further to the origin, $(\phi,\theta) = (0,0)$, comparing with the blue curves, demonstrating that the gait on the four-dimensional gait surface is allowed to move in a wider region of attitude. The region of admissible attitudes enlarges nearly to the whole region of $\phi \in (-\pi/2,\pi/2), \theta \in (-\pi/2,\pi/2)$.

## 6. Conclusions and Discussions

The proposed four-dimensional gait surface helps plan the gait robust to the attitude change. The gaits on the four-dimensional gait surface show a wider region of acceptable attitudes compared with the relevant gait biased by partially scaling.

Although scaling has been proven to be a valid approach to modifying an unacceptable gait, which results in a singular decoupling matrix, partially scaling may not increase the robustness and even may have the opposite effect for some gaits.

Two Color Map Theorem assists in finding the continuous gait on the four-dimensional gait surface. Multiple gaits can be found on the four-dimensional gait surface without violating this theorem.

The robust gaits distributed on the four-dimensional gait surface can be classified into (+,+) type and (−,−) type. The tilting angles on the (+,+) surface satisfy $\alpha_1 = \alpha_3$, $\alpha_2 = \alpha_4$.

The gait planned on the planes (+,+) generally receives a larger acceptable attitude zone than the gait planned on the planes (−,−).

An unusual result in Gait 2 is worth discussing.

Figure 28 plots the unacceptable attitudes for Gait 2 on the four-dimensional gait surface (red curves) and for the corresponding biased Gait 2 by scaling with $\eta = 80\%$ (blue curves).

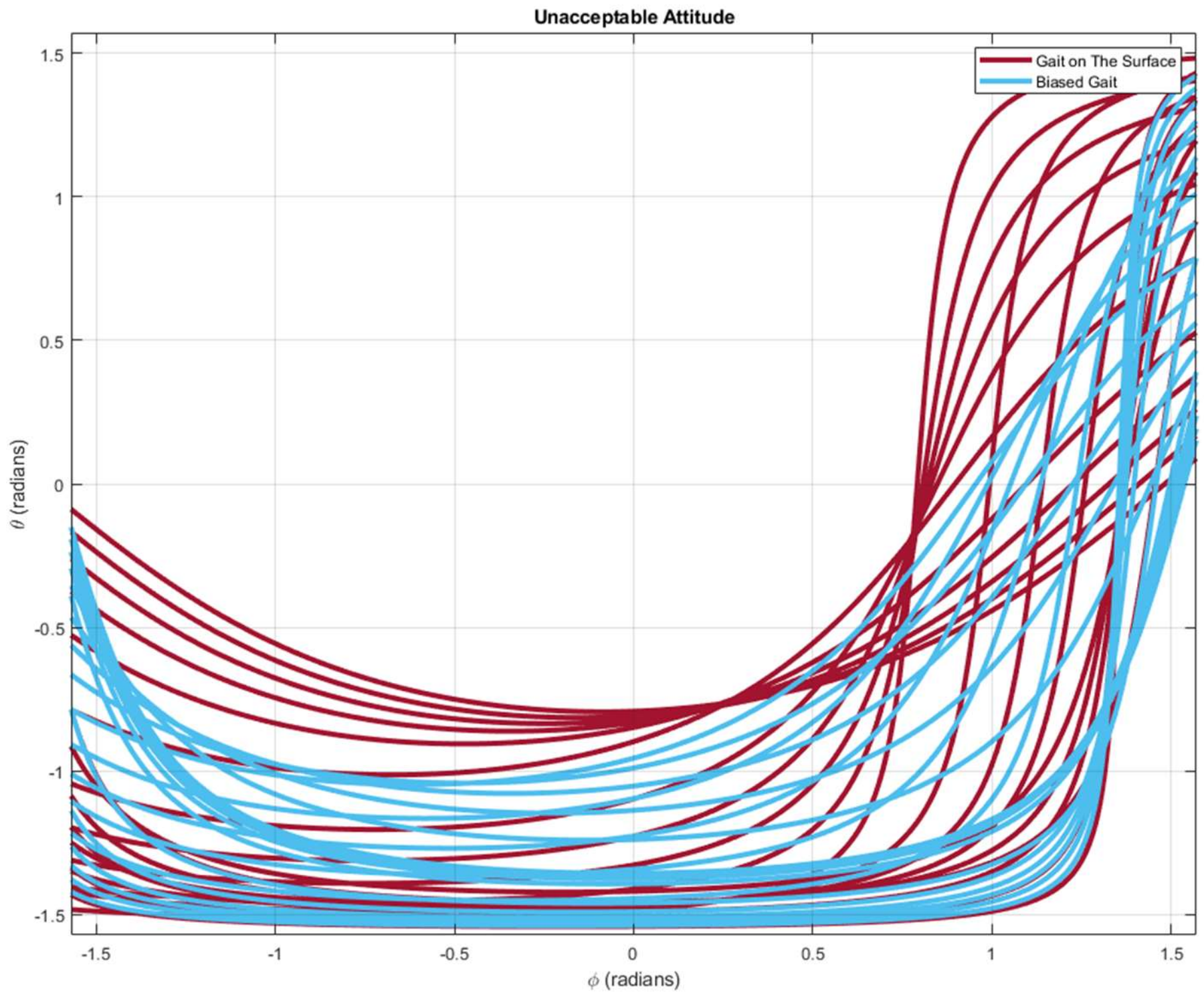

**Figure 28.** The unacceptable attitudes for Gait 2 and for biased Gait 2 ($\eta = 80\%$). The red curve represents the attitudes receiving the singular decoupling matrix in Gait 2. The blue curve represents the attitudes receiving the singular decoupling matrix in biased Gait 2 ($\eta = 80\%$).

It can be seen that the robustness of Gait 2 has little advantages compared with this biased gait by scaling ($\eta = 80\%$). At the same time, it shows a greater advantage compared with the other biased gait ($\eta = 99\%$) in Figure 23. This may be caused by the fact that Gait 2 is a locally most robust gait. Further deducing this unexpected phenomenon is beyond the scope of this research.

There are several further research. Firstly, applying the gait to the feedback linearization in a tilt-rotor control is desired, e.g., a tracking experiment with different gaits. Secondly, the underlying mechanism that the gaits on the planes (+,+) receives greater robustness compared with the gaits on the planes (−,−) is waiting to be elucidated.

## References

1. Chen, C.-C.; Chen, Y.-T. Feedback Linearized Optimal Control Design for Quadrotor With Multi-Performances. *IEEE Access* **2021**, *9*, 26674–26695. https://doi.org/10.1109/ACCESS.2021.3057378.
2. Lian, S.; Meng, W.; Lin, Z.; Shao, K.; Zheng, J.; Li, H.; Lu, R. Adaptive Attitude Control of a Quadrotor Using Fast Nonsingular Terminal Sliding Mode. *IEEE Trans. Ind. Electron.* **2022**, *69*, 1597–1607. https://doi.org/10.1109/TIE.2021.3057015.
3. Chang, D.E.; Eun, Y. Global Chartwise Feedback Linearization of the Quadcopter with a Thrust Positivity Preserving Dynamic Extension. *IEEE Trans. Autom. Control.* **2017**, *62*, 4747–4752. https://doi.org/10.1109/TAC.2017.2683265.
4. Martins, L.; Cardeira, C.; Oliveira, P. Feedback Linearization with Zero Dynamics Stabilization for Quadrotor Control. *J. Intell. Robot. Syst.* **2021**, *101*, 7. https://doi.org/10.1007/s10846-020-01265-2.
5. Kuantama, E.; Tarca, I.; Tarca, R. Feedback Linearization LQR Control for Quadcopter Position Tracking. In Proceedings of the 2018 5th International Conference on Control, Decision and Information Technologies (CoDIT), Thessaloniki, Greece, 10–13 April 2018; pp. 204–209.
6. Ansari, U.; Bajodah, A.H.; Hamayun, M.T. Quadrotor Control Via Robust Generalized Dynamic Inversion and Adaptive Non-Singular Terminal Sliding Mode. *Asian J. Control* **2019**, *21*, 1237–1249. https://doi.org/10.1002/asjc.1800.
7. Lewis, F.; Das, A.; Subbarao, K. Dynamic Inversion with Zero-Dynamics Stabilisation for Quadrotor Control. *IET Control Theory Appl.* **2009**, *3*, 303–314. https://doi.org/10.1049/iet-cta:20080002.
8. Taniguchi, T.; Sugeno, M. Trajectory Tracking Controls for Non-Holonomic Systems Using Dynamic Feedback Linearization Based on Piecewise Multi-Linear Models. *IAENG Int. J. Appl. Math.* **2017**, *47*, 339–351.
9. Lee, D.; Kim, H.J.; Sastry, S. Feedback Linearization vs. Adaptive Sliding Mode Control for a Quadrotor Helicopter. *Int. J. Control. Autom. Syst.* **2009**, *7*, 419–428. https://doi.org/10.1007/s12555-009-0311-8.
10. Mutoh, Y.; Kuribara, S. Control of Quadrotor Unmanned Aerial Vehicles Using Exact Linearization Technique with the Static State Feedback. *J. Autom. Control Eng.* **2016**, *4*, 340–346. https://doi.org/10.18178/joace.4.5.340-346.
11. Shen, Z.; Tsuchiya, T. Singular Zone in Quadrotor Yaw–Position Feedback Linearization. *Drones* **2022**, *6*, 84, doi:doi.org/10.3390/drones6040084.
12. Voos, H. Nonlinear Control of a Quadrotor Micro-UAV Using Feedback-Linearization. In Proceedings of the 2009 IEEE International Conference on Mechatronics, Malaga, Spain, 14–17 April 2009; pp. 1–6.
13. Ryll, M.; Bulthoff, H.H.; Giordano, P.R. Modeling and Control of a Quadrotor UAV with Tilting Propellers. In Proceedings of the 2012 IEEE International Conference on Robotics and Automation, St Paul, MN, USA, 14–18 May 2012; pp. 4606–4613.
14. Rajappa, S.; Ryll, M.; Bulthoff, H.H.; Franchi, A. Modeling, Control and Design Optimization for a Fully-Actuated Hexarotor Aerial Vehicle with Tilted Propellers. In Proceedings of the 2015 IEEE International Conference on Robotics and Automation (ICRA), Seattle, WA, USA, 25–30 May 2015; pp. 4006–4013.
15. Saif, A.-W.A. Feedback Linearization Control of Quadrotor with Tiltable Rotors under Wind Gusts. *Int. J. Adv. Appl. Sci.* **2017**, *4*, 150–159. https://doi.org/10.21833/ijaas.2017.010.021.
16. Mistler, V.; Benallegue, A.; M'Sirdi, N.K. Exact Linearization and Noninteracting Control of a 4 Rotors Helicopter via Dynamic Feedback. In Proceedings of the 10th IEEE International Workshop on Robot and Human Interactive Communication. ROMAN 2001 (Cat. No.01TH8591), Paris, France, 18–21 September 2001; pp. 586–593.
17. Ryll, M.; Bulthoff, H.H.; Giordano, P.R. A Novel Overactuated Quadrotor Unmanned Aerial Vehicle: Modeling, Control, and Experimental Validation. *IEEE Trans. Contr. Syst. Technol.* **2015**, *23*, 540–556. https://doi.org/10.1109/TCST.2014.2330999.
18. Scholz, G.; Trommer, G.F. Model Based Control of a Quadrotor with Tiltable Rotors. *Gyroscopy Navig.* **2016**, *7*, 72–81. https://doi.org/10.1134/S2075108716010120.
19. Imamura, A.; Miwa, M.; Hino, J. Flight Characteristics of Quad Rotor Helicopter with Thrust Vectoring Equipment. *J. Robot. Mechatron.* **2016**, *28*, 334–342. https://doi.org/10.20965/jrm.2016.p0334.
20. Ryll, M.; Bulthoff, H.H.; Giordano, P.R. First Flight Tests for a Quadrotor UAV with Tilting Propellers. In Proceedings of the 2013 IEEE International Conference on Robotics and Automation, Karlsruhe, Germany, 6–10 May 2013; pp. 295–302.
21. Kumar, R.; Nemati, A.; Kumar, M.; Sharma, R.; Cohen, K.; Cazaurang, F. *Tilting-Rotor Quadcopter for Aggressive Flight Maneuvers Using Differential Flatness Based Flight Controller*; American Society of Mechanical Engineers: Tysons, VA, USA, 2017; p. V003T39A006.
22. Shen, Z.; Tsuchiya, T. Gait Analysis for a Tiltrotor: The Dynamic Invertible Gait. *Robotics* **2022**, *11*, 33. https://doi.org/10.3390/robotics11020033.
23. Shen, Z.; Ma, Y.; Tsuchiya, T. Feedback Linearization Based Tracking Control of a Tilt-Rotor with Cat-Trot Gait Plan. *arXiv* **2022** arXiv:2202.02926 [cs.RO].

24. Altug, E.; Ostrowski, J.P.; Mahony, R. Control of a Quadrotor Helicopter Using Visual Feedback. In Proceedings of the 2002 IEEE International Conference on Robotics and Automation (Cat. No.02CH37292), Washington, DC, USA, 11–15 May 2002; Volume 1, pp. 72–77.
25. Zhou, Q.-L.; Zhang, Y.; Rabbath, C.-A.; Theilliol, D. Design of Feedback Linearization Control and Reconfigurable Control Allocation with Application to a Quadrotor UAV. In Proceedings of the 2010 Conference on Control and Fault-Tolerant Systems (SysTol), Nice, France, 6–8 October 2010; pp. 371–376.
26. Mokhtari, A.; Benallegue, A. Dynamic Feedback Controller of Euler Angles and Wind Parameters Estimation for a Quadrotor Unmanned Aerial Vehicle. In Proceedings of the 2004 IEEE International Conference on Robotics and Automation, ICRA'04, New Orleans, LA, USA, 26 April–1 May2004; Volume 3; pp. 2359–2366.
27. Shen, Z.; Tsuchiya, T. Cat-Inspired Gaits for A Tilt-Rotor—From Symmetrical to Asymmetrical. *arXiv* **2022** arXiv:2203.12057 [cs.RO] 12.
28. Wisleder, D.; Zernicke, R.F.; Smith, J.L. Speed-Related Changes in Hindlimb Intersegmental Dynamics during the Swing Phase of Cat Locomotion. *Exp. Brain Res.* **1990**, *79*, 651–660. https://doi.org/10.1007/BF00229333.
29. Verdugo, M.R.; Rahal, S.C.; Agostinho, F.S.; Govoni, V.M.; Mamprim, M.J.; Monteiro, F.O. Kinetic and Temporospatial Parameters in Male and Female Cats Walking over a Pressure Sensing Walkway. *BMC Vet. Res.* **2013**, *9*, 129. https://doi.org/10.1186/1746-6148-9-129.
30. Appel, K.; Haken, W. The Solution of the Four-Color-Map Problem. *Sci. Am.* **1977**, *237*, 108–121.
31. Franklin, P. The Four Color Problem. *Am. J. Math.* **1922**, *44*, 225–236. https://doi.org/10.2307/2370527.

# Chapter 7

# The Robust Gait of a Tilt-Rotor and Its Application to Tracking Control -- Application of Two Color Map Theorem

**This Chapter has been published on The Robust Gait of a Tilt-Rotor and Its Application to Tracking Control -- Application of Two Color Map Theorem. *International Conference on Control, Automation and Systems (ICCAS)* 2022. as the first author. The rest co-author(s) contribute nothing other than supervising and/or minor edition.**

**Abstract:** Ryll's tilt-rotor is a UAV with eight inputs; the four magnitudes of the thrusts as well as four tilting angles of the thrusts can be specified in need, e.g., based on a control rule. Despite of the success in simulation, conventional feedback linearization witnesses the over-intensive change in the inputs while applying to stabilize Ryll's tilt-rotor. Our previous research thus puts the extra procedure named gait plan forward to suppress the unexpected changes in the tilting angles. Accompanying with Two Color Map Theorem, the tilting angles are planned robustly and continuously. The designed gaits are robust to the change of the attitude. However, this is not a complete theory before applying to the tracking simulation test. This paper further discusses some gaits following Two Color Map Theorem and simulates a tracking problem for a tilt-rotor. A uniform circular moving reference is designed to be tracked by the tilt-rotor equipped with the designed robust gait and the feedback linearization controller. The planned gaits satisfying Two Color Map Theorem in this research show the robustness. The results from the simulation show the success in tracking of the tilt-rotor.

## 1. Introduction

With the merit of the lateral thrusts, Ryll's tilt-rotor [1] attracted great attention in the past decade. While controlling this system utilizing feedback linearization, the tilting angles can change in unexpected over-intensive ways, which intrigues the birth of the gait plan of the tilt-rotor [1–4].

The gait plan is a procedure to pre-define the tilting angles (time-specified functions) of the tilt-rotor. Since there are eight inputs in the tilt-rotor, there will be only four magnitudes of the thrusts left to be specified by the controller after the gait plan procedure. A feedback linearization method is subsequently adopted to assign these four magnitudes of the thrusts.

Despite the fact that this control scenario successfully avoids the over-intensive change in the tilting angles from the theorem level, the subsequent feedback linearization method may encounter the singular decoupling matrix [5–7], causing the failure in the control result. The decoupling matrix is not invertible for several combinations of roll angle and pitch angle [8–10].

With this concern, a rule, Two Color Map Theorem [11], was put forward to help design the gait. Following Two Color Map Theorem, the designed gaits are guaranteed continuous and robust to the change of the attitude. The unacceptable combinations of the roll angles and pitch angles, which introduce the singular decoupling matrix, decrease, enabling the tilt-rotor to maneuver in wider attitude regions.

So far, no discussion on the tracking control has been addressed using the robust gaits, obeying Two Color Map Theorem. Tracking (dependently) the 6 degrees of freedom relying on the 4 inputs of the tilt-rotor can be realized with the help of the modified attitude-position decoupler put forward in our previous research [7]. Position X and Y are tracked relying on the coupling relationship with the degrees of freedom, roll, pitch, yaw, and altitude (Z), which are controlled independently.

In this paper, the robustness of several gaits following the Two Color Map Theorem is analyzed. One gait is then adopted to track a uniform circular moving reference. The designed gaits show their robustness to the attitude. The reference is successfully tracked in the simulation.

## 2. Invertible Decoupling Matrix: Necessary Condition

Ryll's tilt-rotor is sketched in Fig. 1 [2].

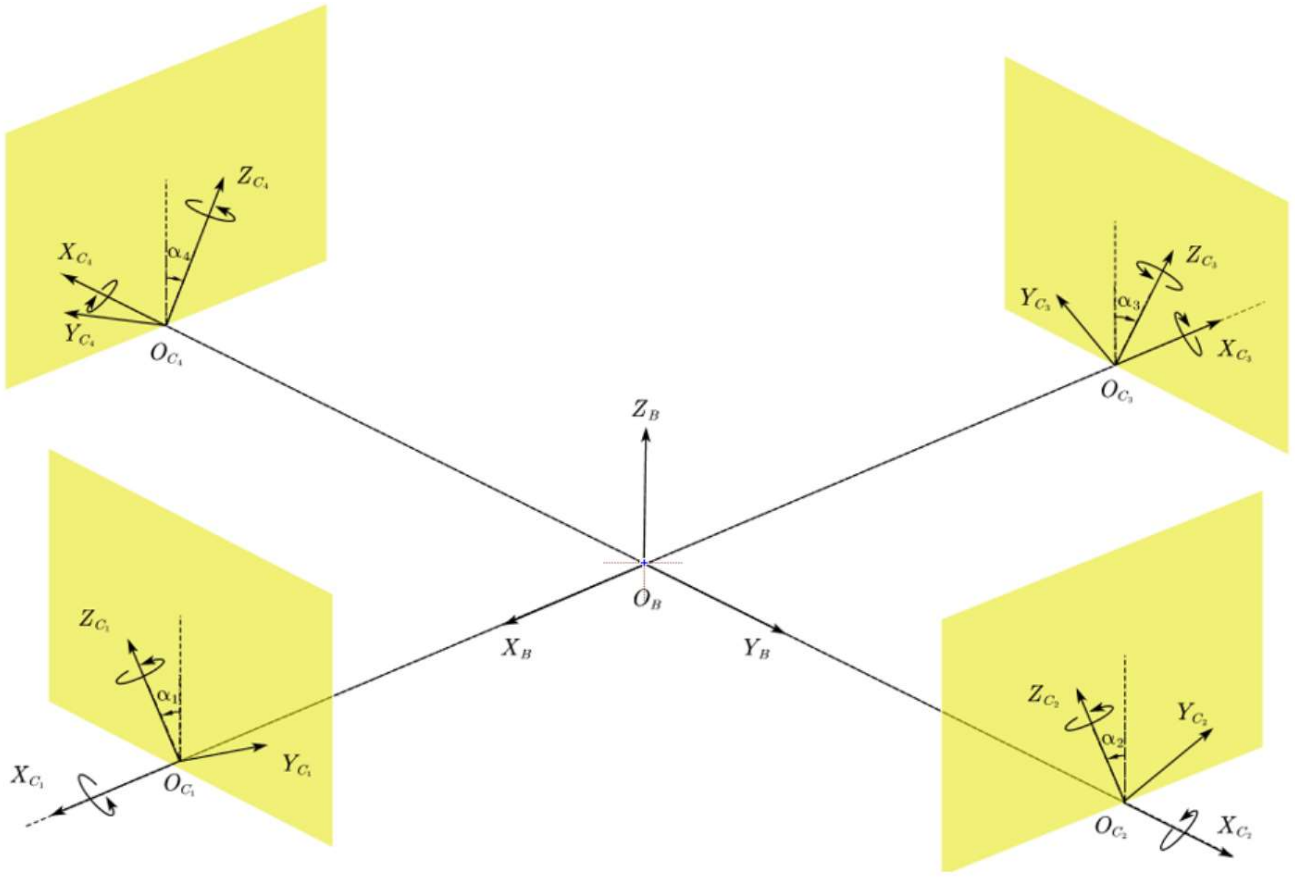

**Figure 1.** Ryll's tilt-rotor.

The position of the tilt-rotor [5,12] is given by

$$\ddot{P} = \begin{bmatrix} 0 \\ 0 \\ -g \end{bmatrix} + \frac{1}{m} \cdot {}^{W}R \cdot F(\alpha) \cdot \begin{bmatrix} \varpi_1 \cdot |\varpi_1| \\ \varpi_2 \cdot |\varpi_2| \\ \varpi_3 \cdot |\varpi_3| \\ \varpi_4 \cdot |\varpi_4| \end{bmatrix}$$
$$\triangleq \begin{bmatrix} 0 \\ 0 \\ -g \end{bmatrix} + \frac{1}{m} \cdot {}^{W}R \cdot F(\alpha) \cdot w, \quad (1)$$

where $P = [X \quad Y \quad Z]^T$ represents the position with respect to the earth frame, $m$ represents the total mass, $g$ represents the gravitational acceleration, $\varpi_i$, $(i = 1,2,3,4)$ represents the angular velocity of the propeller $w = [\varpi_1 \cdot |\varpi_1|, \varpi_2 \cdot |\varpi_2|, \varpi_3 \cdot |\varpi_3|, \varpi_4 \cdot |\varpi_4|]^T$, and the term ${}^{W}R$ represents the rotational matrix.

The tilting angles $\alpha = [\alpha_1 \quad \alpha_2 \quad \alpha_3 \quad \alpha_4]$. $F(\alpha)$ is given by

$$F(\alpha) = \begin{bmatrix} 0 & K_f \cdot s2 & 0 & -K_f \cdot s4 \\ K_f \cdot s1 & 0 & -K_f \cdot s3 & 0 \\ -K_f \cdot c1 & K_f \cdot c2 & -K_f \cdot c3 & K_f \cdot c4 \end{bmatrix}, \quad (2)$$

where $si = \sin(\alpha_i)$, $ci = \cos(\alpha_i)$, and $(i = 1,2,3,4)$. $K_f$ $(8.048 \times 10^{-6} N \cdot s^2/rad^2)$ is the coefficient of the thrust.

The angular velocity of the body [5,13] with respect to $\mathcal{F}_B$, $\omega_B = [p \quad q \quad r]^T$, is given by

$$\dot{\omega}_B = {I_B}^{-1} \cdot \tau(\alpha) \cdot w, \quad (3)$$

where $I_B$ is the matrix of moments of inertia, $K_m$ $(2.423 \times 10^{-7} N \cdot m \cdot s^2/rad^2)$ is the coefficient of the drag, and $L$ is the length of the arm,

$$\tau(\alpha) = \begin{bmatrix} 0 & L\cdot K_f\cdot c2 - K_m\cdot s2 & 0 & -L\cdot K_f\cdot c4 + K_m\cdot s4 \\ L\cdot K_f\cdot c1 + K_m\cdot s1 & 0 & -L\cdot K_f\cdot c3 - K_m\cdot s3 & 0 \\ L\cdot K_f\cdot s1 - K_m\cdot c1 & -L\cdot K_f\cdot s2 - K_m\cdot c2 & L\cdot K_f\cdot s3 - K_m\cdot c3 & -L\cdot K_f\cdot s4 - K_m\cdot c4 \end{bmatrix} \tag{4}$$

We refer the details of the dynamics and the feedback linearization control method to our previous research [5].

In our control scenario, the tilting angles, $\alpha$, are defined in advanced in a separated procedure called gait plan [2,14]. The tilt-rotor is controlled using the four magnitudes of the thrusts by feedback linearization. Note that the decoupling matrix is singular for some attitude. The gait plan is supposed to generate a robust gait, which enlarges the acceptable attitude region.

## 3. Gait Plan

The gait of a tilt-rotor is a time-specified tilting angles, $\alpha_1(t)$, $\alpha_2(t)$, $\alpha_3(t)$, and $\alpha_4(t)$. Our previous research [11] puts forward a theorem, Two Color Map Theorem, to design robust gaits. The robust gait is a gait that has a large region of the acceptable attitudes, which introduces the invertible decoupling matrix.

### *3.1. Two Color Map Theorem*

Firstly, design $\alpha_1(t)$ and $\alpha_2(t)$ on the $\alpha_1 - \alpha_2$ plane. For example, if we expect a periodic gait, then $(\alpha_1(t), \alpha_2(t))$ is an enclosed rectangular with direction.

For a given time point at $t_1$, $(\alpha_1(t_1), \alpha_2(t_1))$ is subsequently determined. Then, $(\alpha_3(t_1), \alpha_4(t_1))$ is to be determined to finish designing a gait.

It is proved [11] that there are two $(\alpha_3(t_1), \alpha_4(t_1))$ corresponding to a defined $(\alpha_1(t_1), \alpha_2(t_1))$, in general, to design a robust gait.

These two corresponding $(\alpha_3, \alpha_4)$, corresponding to $(\alpha_1, \alpha_2)$ on the entire $\alpha_1 - \alpha_2$ diagram, can be classified into two groups, 'red $(\alpha_3, \alpha_4)$' and 'blue $(\alpha_3, \alpha_4)$'.

Interestingly, $\alpha_3$ in the same color, all the red $\alpha_3$ or all the blue $\alpha_3$, lies on the same plane in the $0\alpha_1\alpha_2\alpha_3$ coordinate system. Similarly, $\alpha_4$ in the same color, all the red $\alpha_4$ or all the blue $\alpha_4$, lies on the same plane in the $0\alpha_1\alpha_2\alpha_4$ coordinate system.

Obviously, to design a robust gait, the adjacent $(\alpha_3, \alpha_4)$ must be in the same color for the interested $(\alpha_1, \alpha_2)$. This is the simplified Two Color Map Theorem without crossover. Note that the cases considering crossover is also considered in our previous research [11], which is beyond the scope of this paper. Fig. 2 and Fig. 3 display the entire map for the first time.

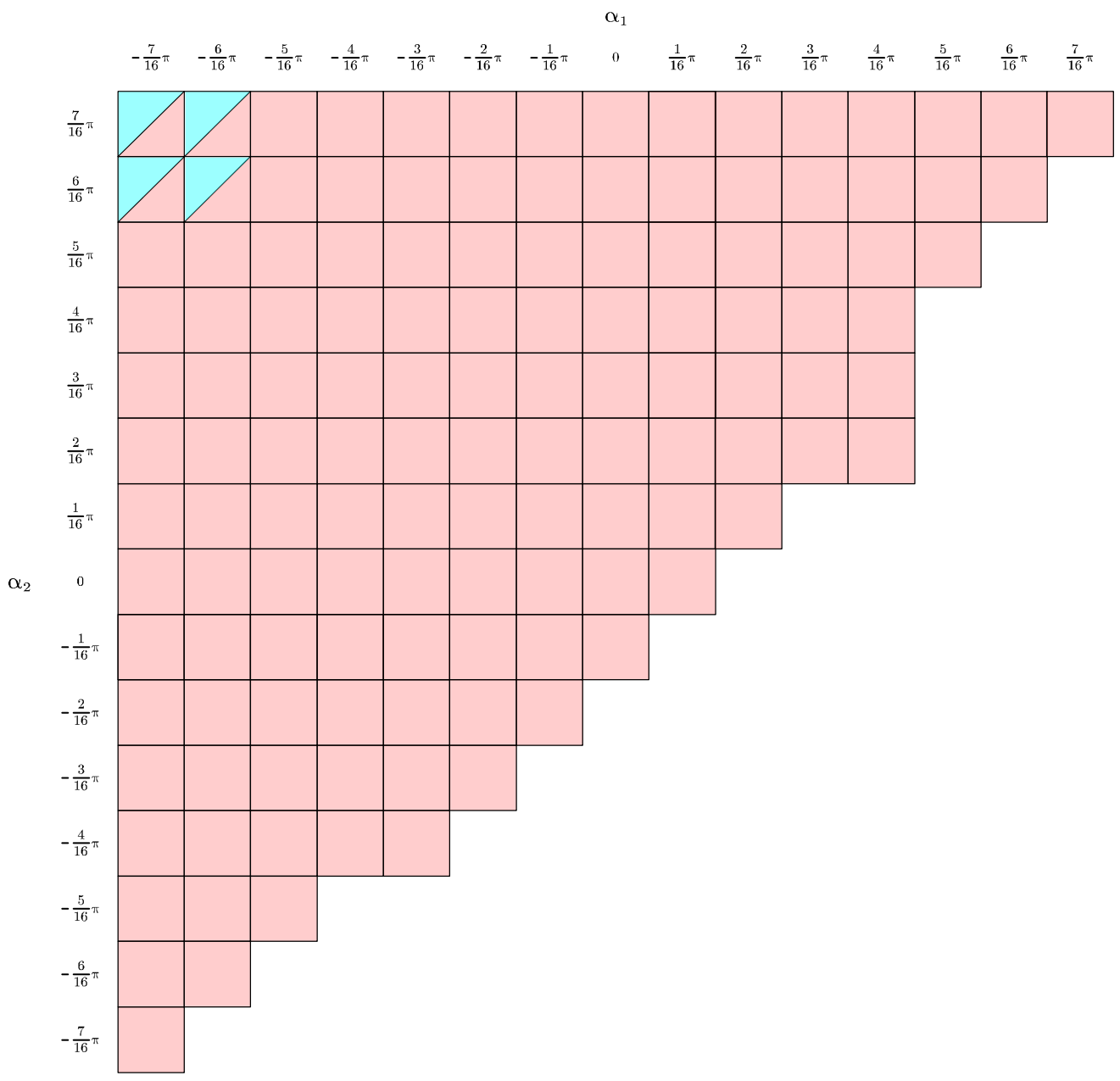

**Figure 2.** Two Color Map with positive residual.

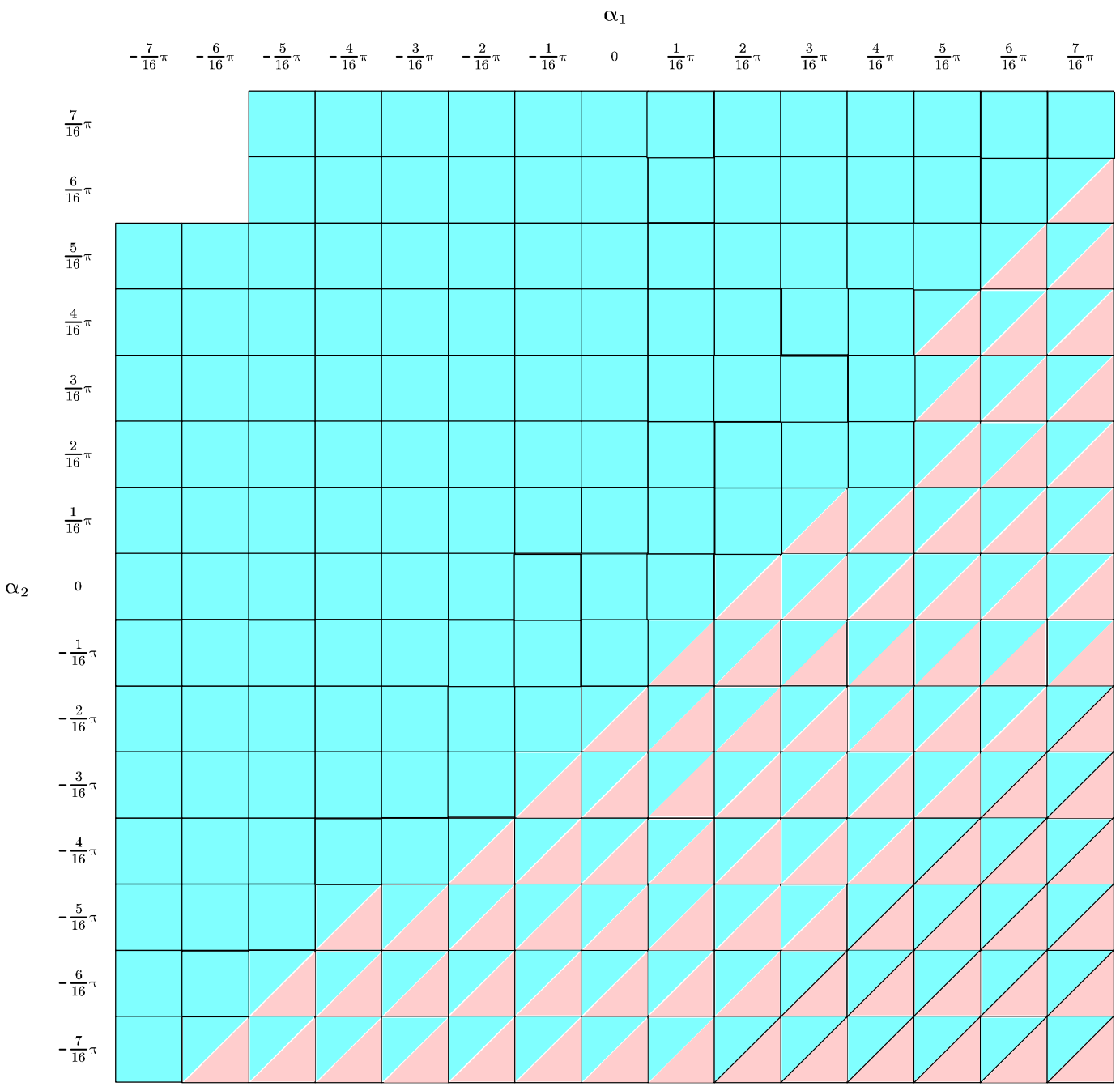

**Figure 3.** Two Color Map with negative residual.

## *3.2. Robustness analysis*

In this research, the robustness of four gaits is analyzed. They are Gait 1, Gait 2, and Gait 3, illustrated in Fig. 4, where $(\alpha_3, \alpha_4)$ of Gait 1 is blue $(\alpha_3, \alpha_4)$, $(\alpha_3, \alpha_4)$ of Gait 2 and Gait 3 are red $(\alpha_3, \alpha_4)$.

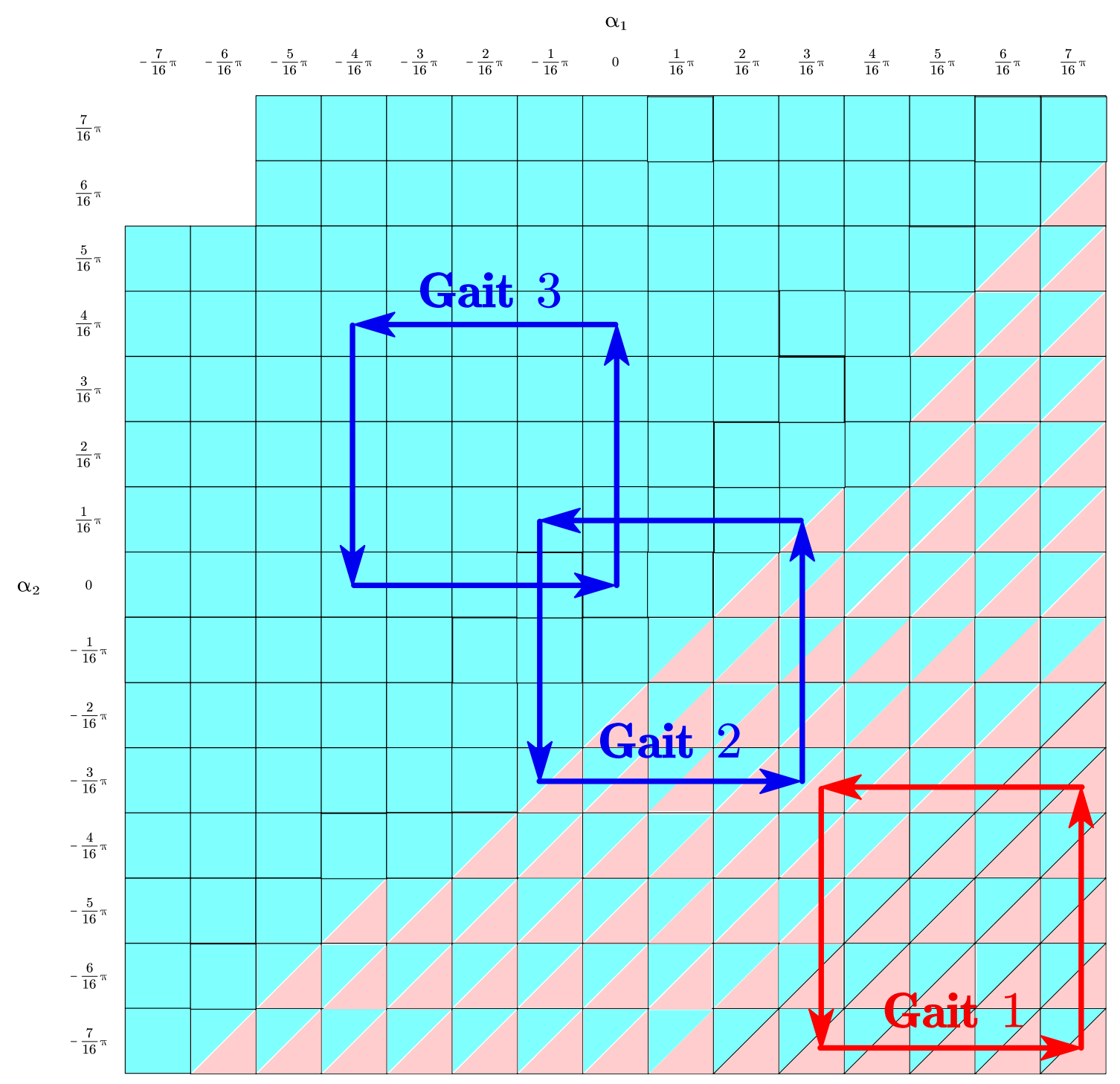

**Figure 4.** Three gaits analyzed in this paper. $(\alpha_3, \alpha_4)$ of Gait 1 is blue $(\alpha_3, \alpha_4)$. $(\alpha_3, \alpha_4)$ of Gait 2 and Gait 3 are red $(\alpha_3, \alpha_4)$.

Figs. 5 ~ 7 plot the unacceptable attitudes (red curves) for Gait 1 ~ Gait 3, respectively. The unacceptable attitudes will introduce the singular decoupling matrix in feedback linearization, which should be prohibited. The tilt-rotor can only maneuver in the attitude region, which is not occupied by these attitude curves.

In comparison, we create the biased gait for each gait; the biased gait is generated by scaling $\alpha_3$ and $\alpha_4$ to their 80%. The unacceptable attitudes (blue curves) for the biased Gait 1 ~ biased Gait 3 are also displayed in Figs 5 ~ 7, respectively.

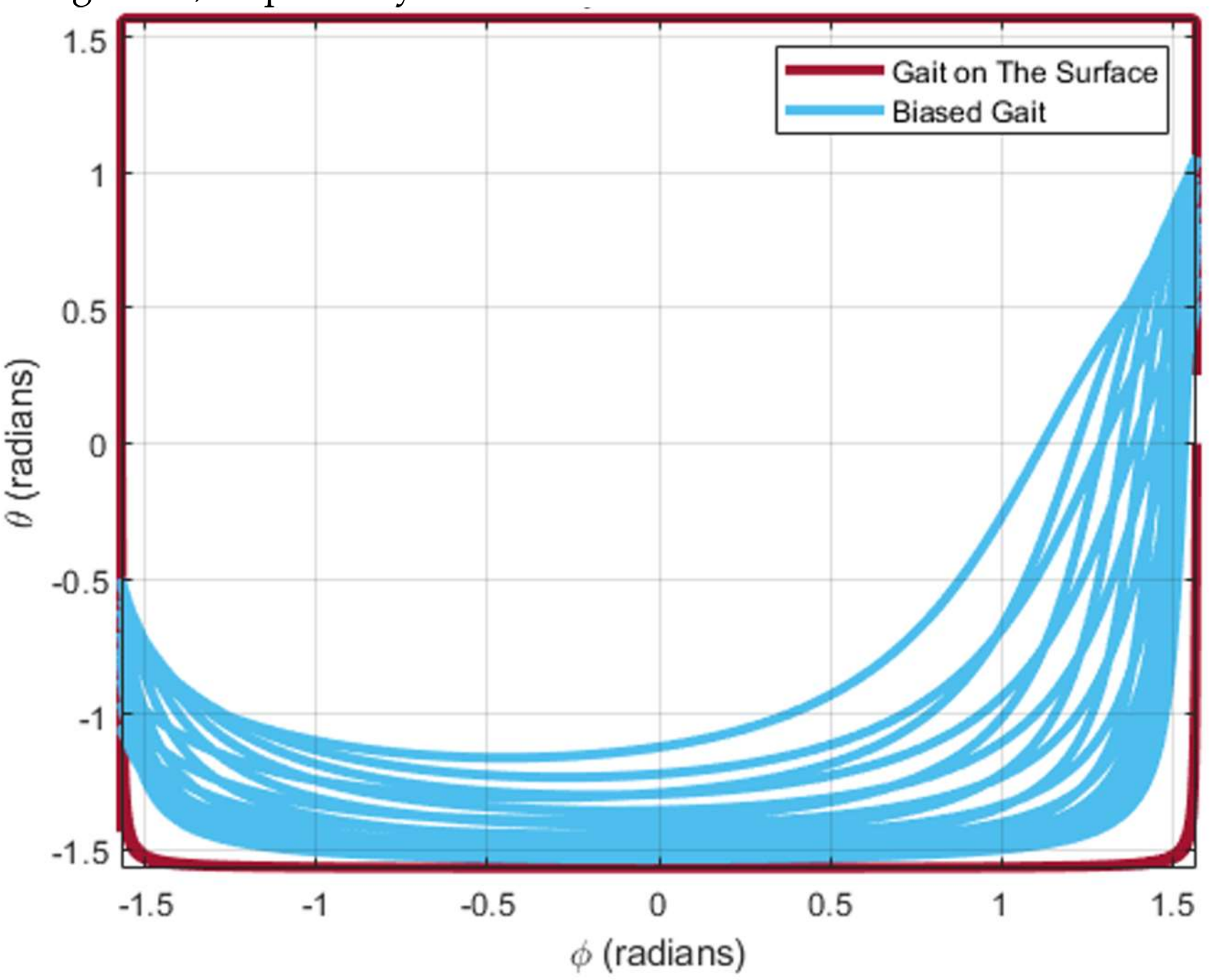

**Figure 5.** The red curves represent the unacceptable attitude of Gait 1. The blue curves represent the unacceptable attitude of the biased Gait 1.

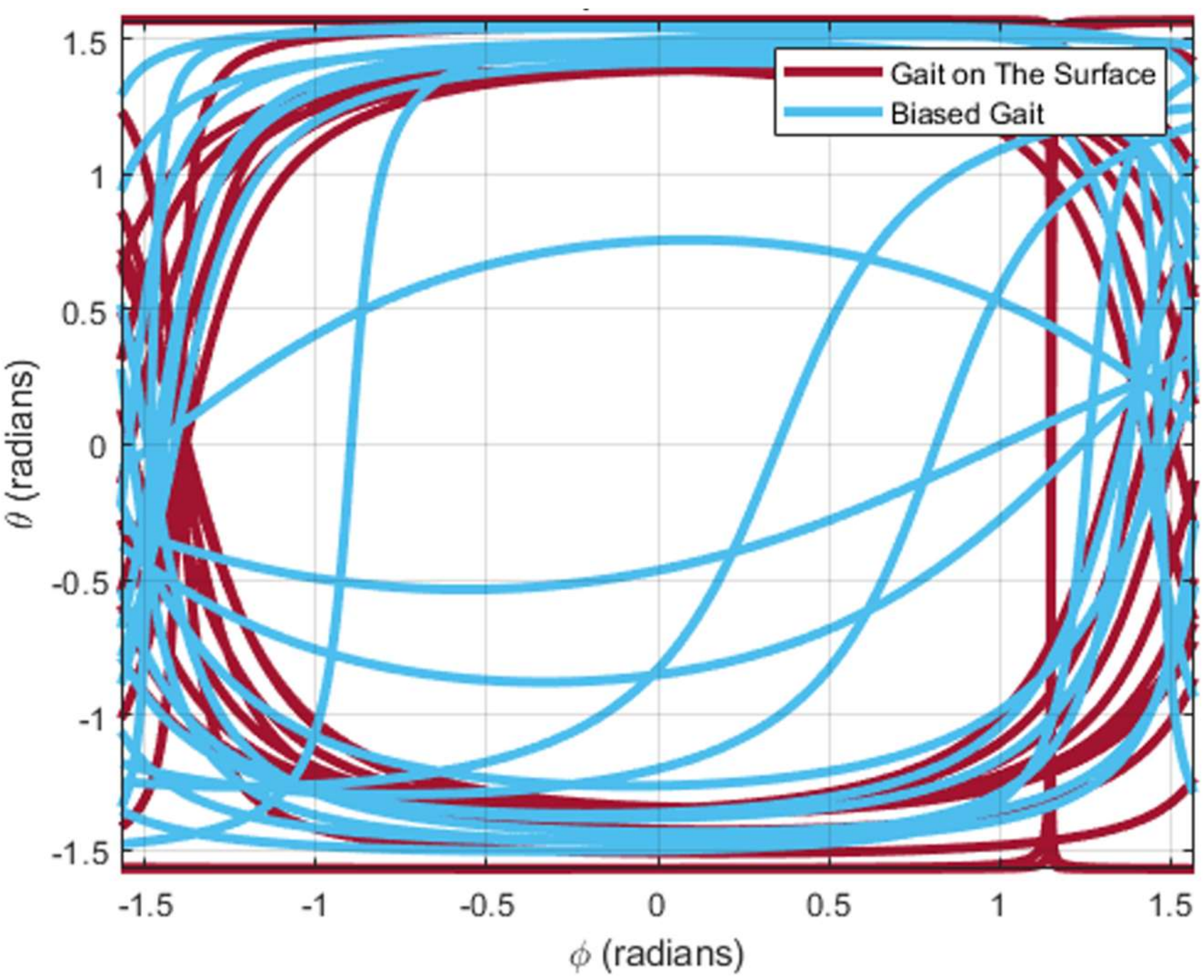

**Figure 6.** The red curves represent the unacceptable attitude of Gait 2. The blue curves represent the unacceptable attitude of the biased Gait 2.

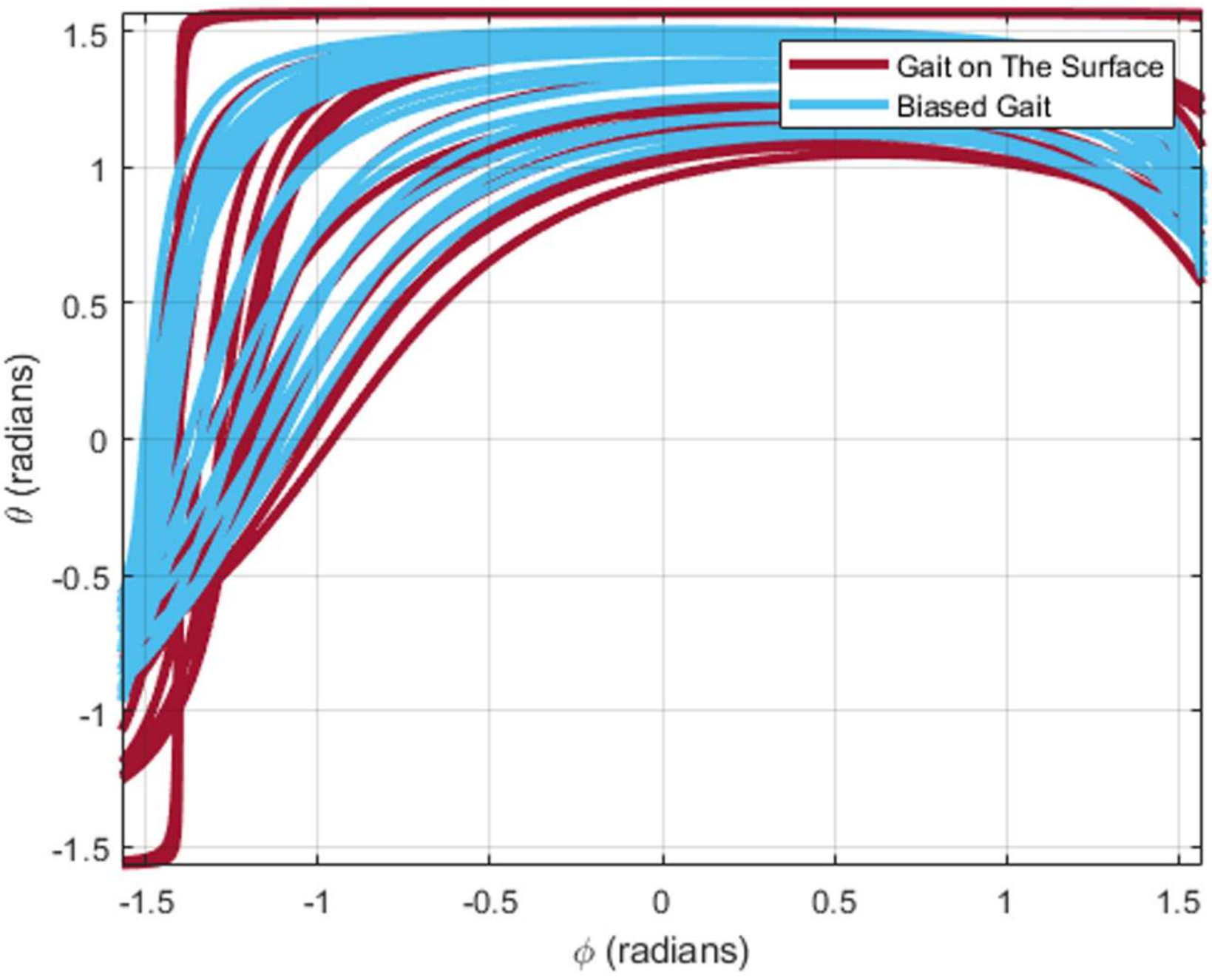

**Figure 7.** The red curves represent the unacceptable attitude of Gait 3. The blue curves represent the unacceptable attitude of the biased Gait 3.

In general, the acceptable attitude region enlarges if the gait following Two Color Map Theorem, especially for the gaits with the red $(\alpha_3, \alpha_4)$. The decoupling matrix introduced by the biased Gait 2 is very sensitive to the attitude. The tilt-rotor can be less likely to be stabilized by feedback linearization for these two biased gaits.

## 4. Tracking Simulation

### *4.1. Reference and feasible gait*

The reference set in the simulation is a uniform circular moving reference, the position of which is define in Eqs. (5) ~ (7).

$$x_r = 5 \cdot \cos(0.1 \cdot t). \tag{5}$$

$$y_r = 5 \cdot \sin(0.1 \cdot t). \tag{6}$$

$$z_r = 0. \tag{7}$$

The velocities of the reference along each dimension are received by calculating the derivatives of the velocities. Note that the acceleration of the reference is set as zero in the simulation test. We refer the detail of the controller as well as the modified attitude-position decoupler to our previous research [5,7].

Besides, although we planned 3 gaits, Gait 1, Gait 2, and Gait 3. The result shows that only Gait 1 works in tracking. The rest gaits receive singular decoupling matrix, which is caused by the saturation of the input (non-zero saturation). Further discussions on the saturations of the input and the stability of feedback linearization can be referred to [15].

The initial position of the tilt-rotor is at $(x_i, y_i, z_i)$ = (0,0,0). The initial angular velocities of the propellers are insufficient to compensate the gravity. The tilt-rotor is expected to stabilize its altitude as well as to track the reference.

### *4.2. Tracking simulation result*

Figure 8 displays the gait that we adopted (Gait 1) in one period. And, the angular velocity history of each propeller is in Figure 9.

The trajectory (red curve) of the tilt-rotor in the simulation is illustrated in Fig. 10, where the reference is represented by the blue curve.

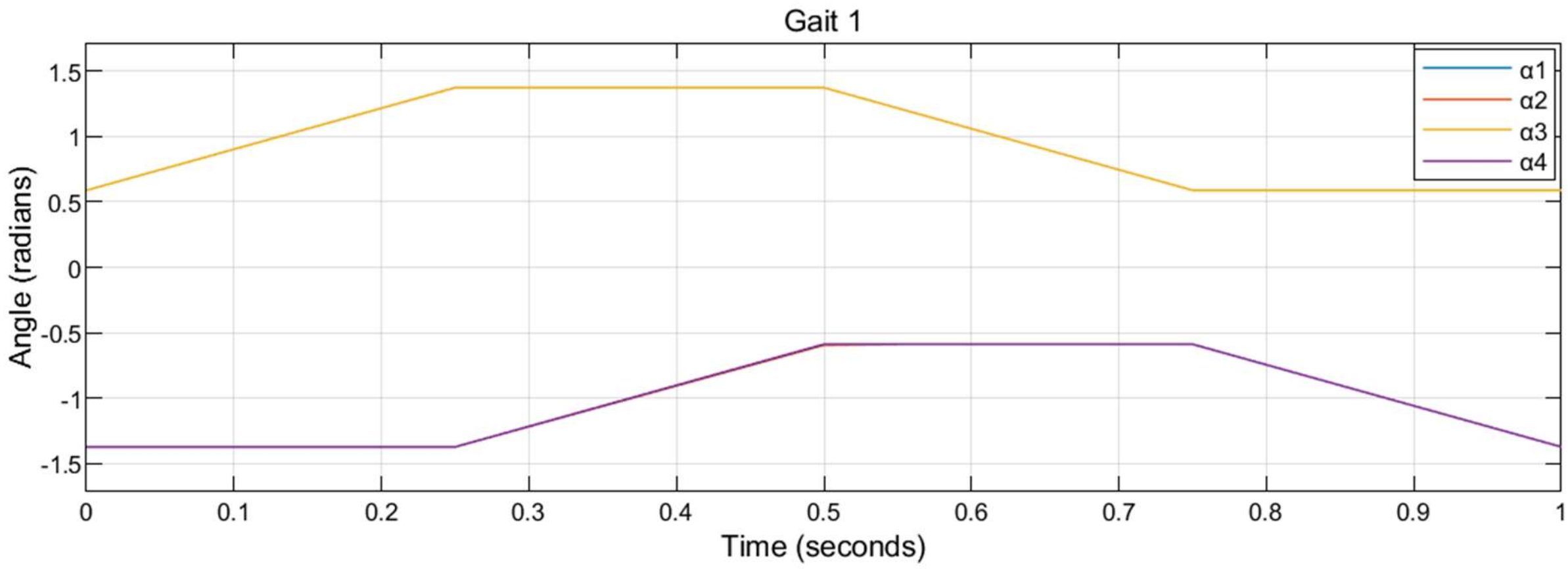

**Figure 8.** Gait 1 within the first period.

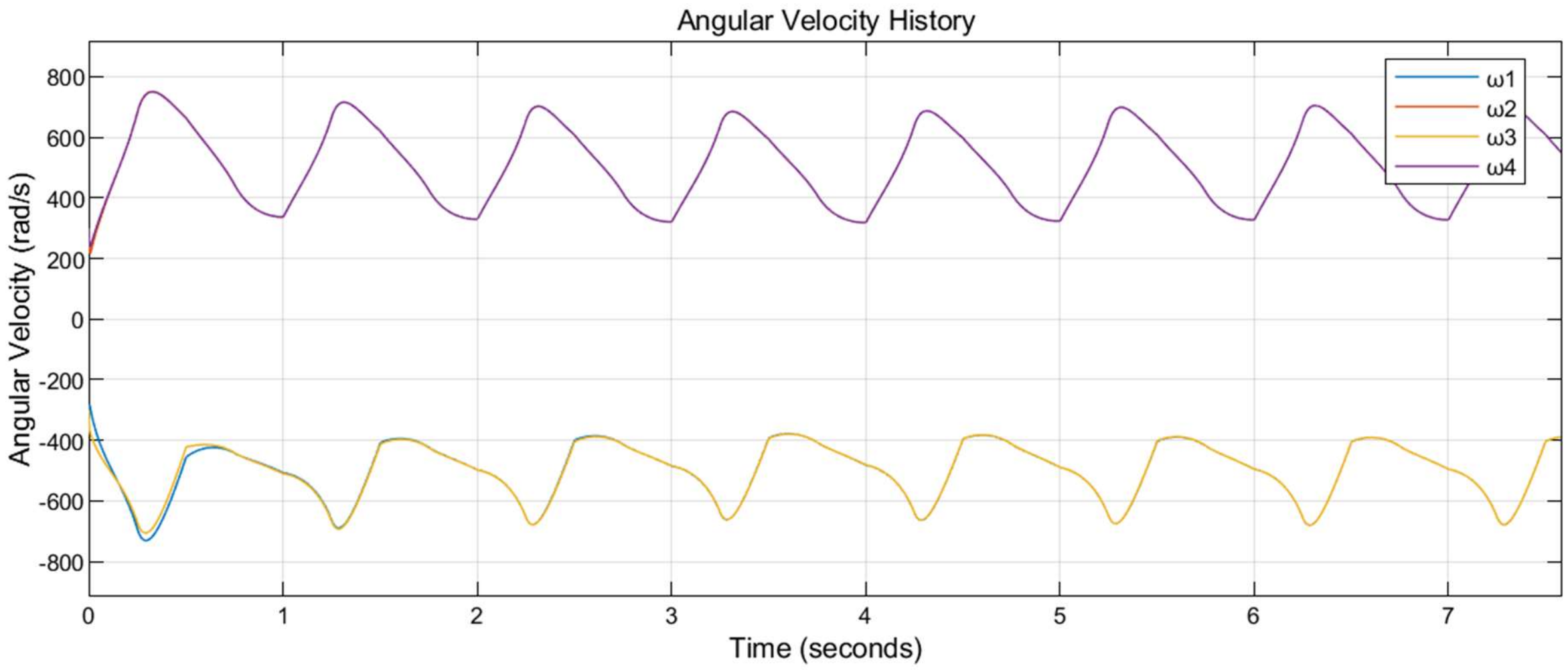

**Figure 9.** Angular velocity history.

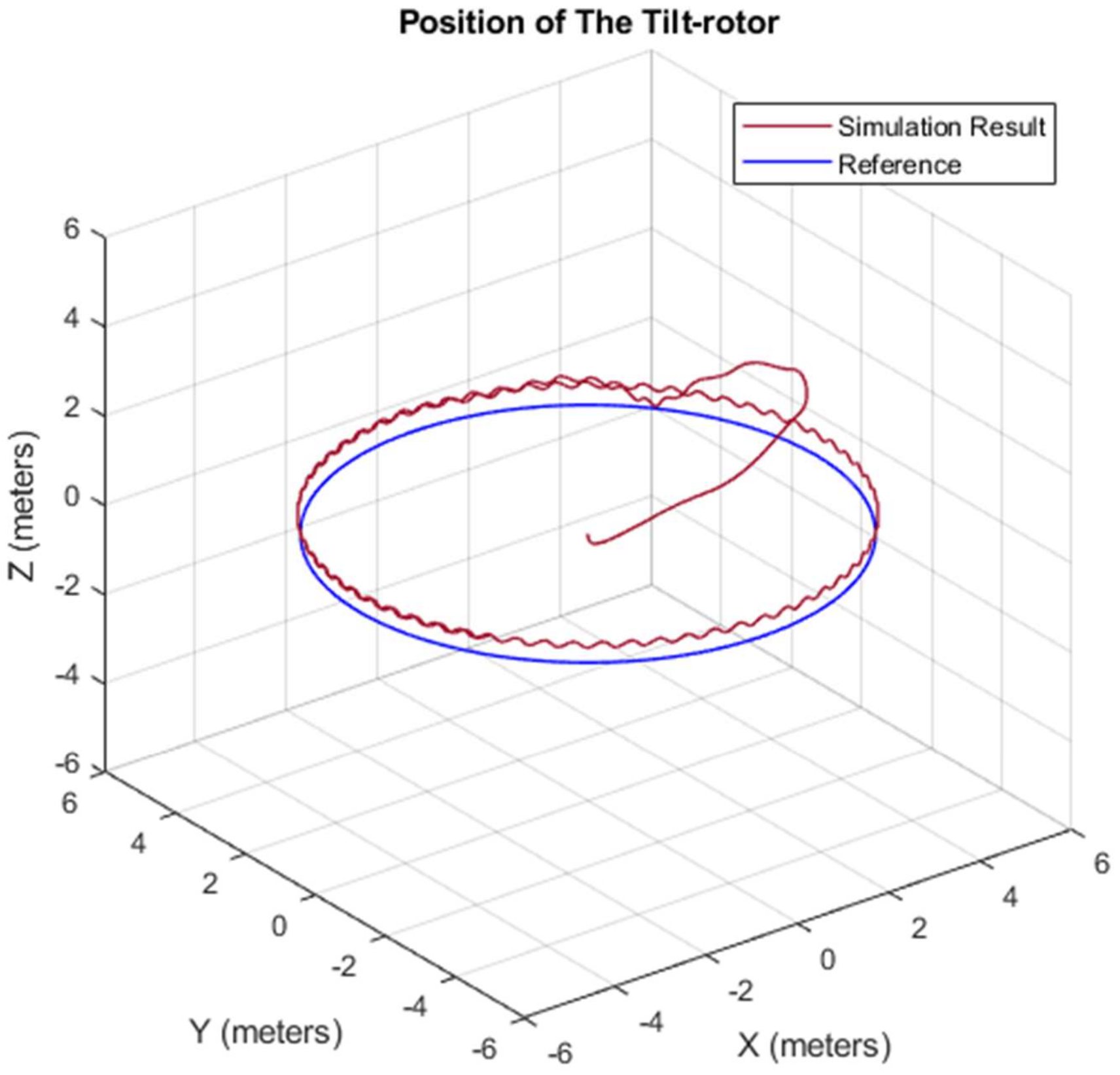

**Figure 10.** Tracking result of the tilt-rotor in the simulation. The blue circle is the reference. The red curve represents the actual trajectory of the tilt-rotor while tracking.

The dynamic state error is presented in Fig. 11. It is close to zero after sufficient time, indicating that the tilt-rotor approaches the reference.

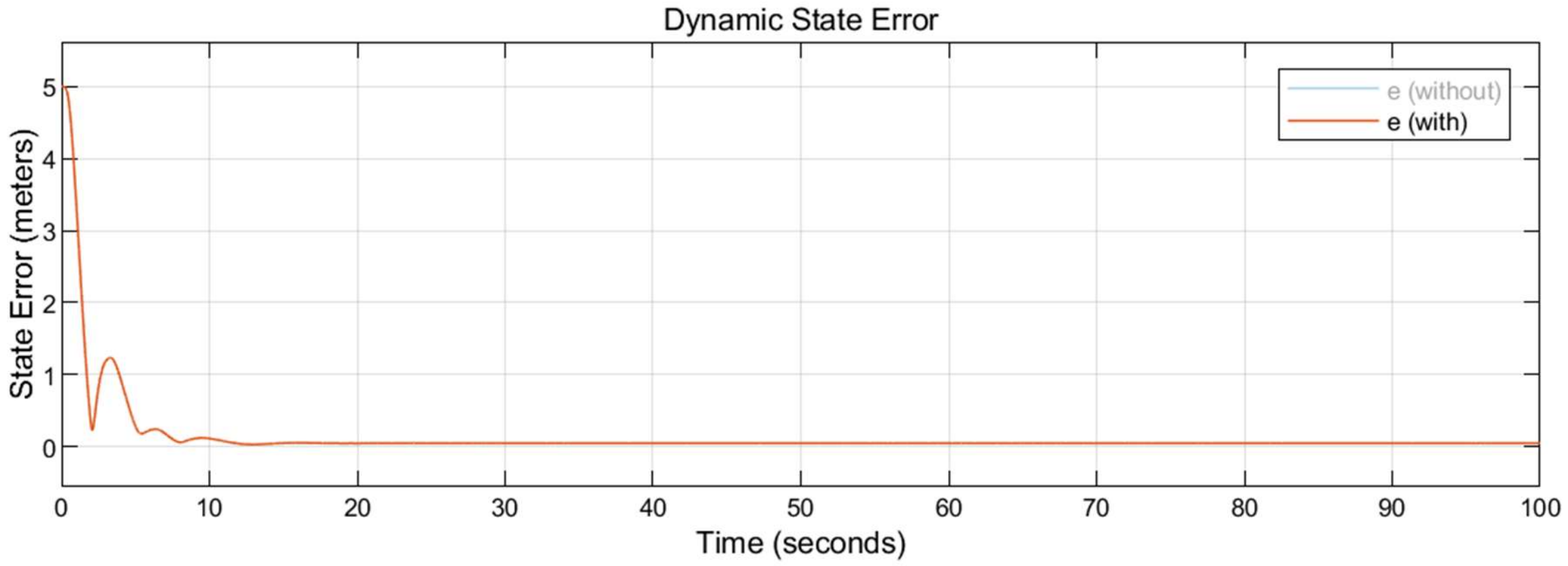

**Figure 11.** The dynamic state error of the tilt-rotor. It is defined by the difference between the reference and the actual position of the tilt-rotor.

## 5.Conclusion

In this paper, we further explore Two Color Map theorem. The complete map is displayed for the first time.

It is found that the gaits on the Two Color Map show greater robustness comparing with the adjacent biased gaits not on the map. In general, the roll-pitch diagrams demonstrate that robust gaits receive larger admissible region in attitudes, which result in invertible decoupling matrices, near $(\text{roll}, \text{pitch}) = (0,0)$. While the adjacent biased gaits receive the roll-pitch diagrams where the unacceptable attitude curves, which introduce the singular decoupling matrices, highly occupy the attitude region near $(\text{roll}, \text{pitch}) = (0,0)$.

The tilt-rotor, starting from an initial state of non-equilibrium, tracks the designed uniform circular reference successfully with little state error while adopting one robust gait on Two Color Map. No saturations or negative constraints of the angular velocities of the propeller are activated in the simulation. All the inputs, the angular velocities of the propellers and the gaits (tilting angles), are continuous. The decoupling matrix is invertible throughout the simulation with the adopted robust gait.

One of the further steps can be the exploration on the dynamics of the tilt-rotor and the gaits.

## References

1. Ryll, M.; Bulthoff, H.H.; Giordano, P.R. Modeling and Control of a Quadrotor UAV with Tilting Propellers. In Proceedings of the 2012 IEEE International Conference on Robotics and Automation; IEEE: St Paul, MN, USA, May 2012; pp. 4606–4613.
2. Shen, Z.; Tsuchiya, T. Gait Analysis for a Tiltrotor: The Dynamic Invertible Gait. *Robotics* **2022**, *11*, 33, doi:10.3390/robotics11020033.
3. Ryll, M.; Bulthoff, H.H.; Giordano, P.R. A Novel Overactuated Quadrotor Unmanned Aerial Vehicle: Modeling, Control, and Experimental Validation. *IEEE Trans. Contr. Syst. Technol.* **2015**, *23*, 540–556, doi:10.1109/TCST.2014.2330999.
4. Kumar, R.; Nemati, A.; Kumar, M.; Sharma, R.; Cohen, K.; Cazaurang, F. Tilting-Rotor Quadcopter for Aggressive Flight Maneuvers Using Differential Flatness Based Flight Controller.; American Society of Mechanical Engineers: Tysons, Virginia, USA, October 11 2017; p. V003T39A006.
5. Shen, Z.; Tsuchiya, T. Cat-Inspired Gaits for a Tilt-Rotor—From Symmetrical to Asymmetrical. *Robotics* **2022**, *11*, 60, doi:10.3390/robotics11030060.
6. Shen, Z.; Tsuchiya, T. Singular Zone in Quadrotor Yaw–Position Feedback Linearization. *Drones* **2022**, *6*, 20, doi:doi.org/10.3390/drones6040084.
7. Shen, Z.; Ma, Y.; Tsuchiya, T. Feedback Linearization-Based Tracking Control of a Tilt-Rotor with Cat-Trot Gait Plan. *International Journal of Advanced Robotic Systems* **2022**, *19*, 17298806221109360, doi:10.1177/17298806221109360.
8. Ghandour, J.; Aberkane, S.; Ponsart, J.-C. Feedback Linearization Approach for Standard and Fault Tolerant Control: Application to a Quadrotor UAV Testbed. *J. Phys.: Conf. Ser.* **2014**, *570*, 082003, doi:10.1088/1742-6596/570/8/082003.

9. Mistler, V.; Benallegue, A.; M'Sirdi, N.K. Exact Linearization and Noninteracting Control of a 4 Rotors Helicopter via Dynamic Feedback. In Proceedings of the Proceedings 10th IEEE International Workshop on Robot and Human Interactive Communication. ROMAN 2001 (Cat. No.01TH8591); IEEE: Paris, France, September 2001; pp. 586–593.
10. Mokhtari, A.; Benallegue, A. Dynamic Feedback Controller of Euler Angles and Wind Parameters Estimation for a Quadrotor Unmanned Aerial Vehicle. In Proceedings of the IEEE International Conference on Robotics and Automation, 2004. Proceedings. ICRA '04. 2004; IEEE: New Orleans, LA, USA, 2004; pp. 2359-2366 Vol.3.
11. Shen, Z.; Ma, Y.; Tsuchiya, T. Four-Dimensional Gait Surfaces for a Tilt-Rotor—Two Color Map Theorem. *Drones* **2022**, *6*, 103, doi:10.3390/drones6050103.
12. Luukkonen, T. Modelling and Control of Quadcopter. *Independent research project in applied mathematics, Espoo* **2011**, *22*, 22.
13. Lee, T.; Leok, M.; McClamroch, N.H. Geometric Tracking Control of a Quadrotor UAV on SE(3). In Proceedings of the 49th IEEE Conference on Decision and Control (CDC); IEEE: Atlanta, GA, December 2010; pp. 5420–5425.
14. Hirose, S. A Study of Design and Control of a Quadruped Walking Vehicle. *The International Journal of Robotics Research* **1984**, *3*, 113–133, doi:10.1177/027836498400300210.
15. Shen, Z.; Ma, Y.; Tsuchiya, T. Stability Analysis of a Feedback-Linearization-Based Controller with Saturation: A Tilt Vehicle with the Penguin-Inspired Gait Plan. *arXiv preprint arXiv:2111.14456* **2021**.

# Chapter 8

## Generalized Two Color Map Theorem (Generalized Two Color Map Theorem -- Complete Theorem of Robust Gait Plan for a Tilt-Rotor)

**Abstract:** Gait plan is a procedure that is typically applied on the ground robots, e.g., quadrupedal robots; the tilt-rotor, a novel type of quadrotor with eight inputs, is not one of them. While controlling the tilt-rotor relying on feedback linearization, the tilting angles (inputs) are expected to change over-intensively, which may not be expected in the application. To help suppress the intensive change in the tilting angles, a gait plan procedure is introduced to the tilt-rotor before feedback linearization. The tilting angles are specified with time in advance by users rather than be given by the control rule. However, based on this scenario, the decoupling matrix in feedback linearization can be singular for some attitudes, combinations of roll angle and pitch angle. It hinders the further application of the feedback linearization. With this concern, Two Color Map Theorem is established to maximize the acceptable attitude region, where the combinations of roll and pitch will give an invertible decoupling matrix. That theorem, however, over-restricts the choice of the tilting angles, which can rule out some feasible robust gaits. This paper gives the generalized Two Color Map Theorem; all the robust gaits can be found based on this generalized theorem. The robustness of three gaits that satisfy this Generalized Two Color Map Theorem (while violating Two Color Map Theorem) are analyzed. The results show that Generalized Two Color Map Theorem completes the search for the robust gaits for a tilt-rotor.

### 1. Introduction

It has been an exact decade since the first time Ryll's tilt-rotor was put forward and stabilized [1]. Comparing with the conventional quadrotor, this UAV attracts attentions since its unique capability of generating the lateral forces. As shown in Fig. 1 [2], the directions of the thrusts are able to be adjusted by changing the tilting angles, $\alpha_1, \alpha_2, \alpha_3, \alpha_4$, during the flight.

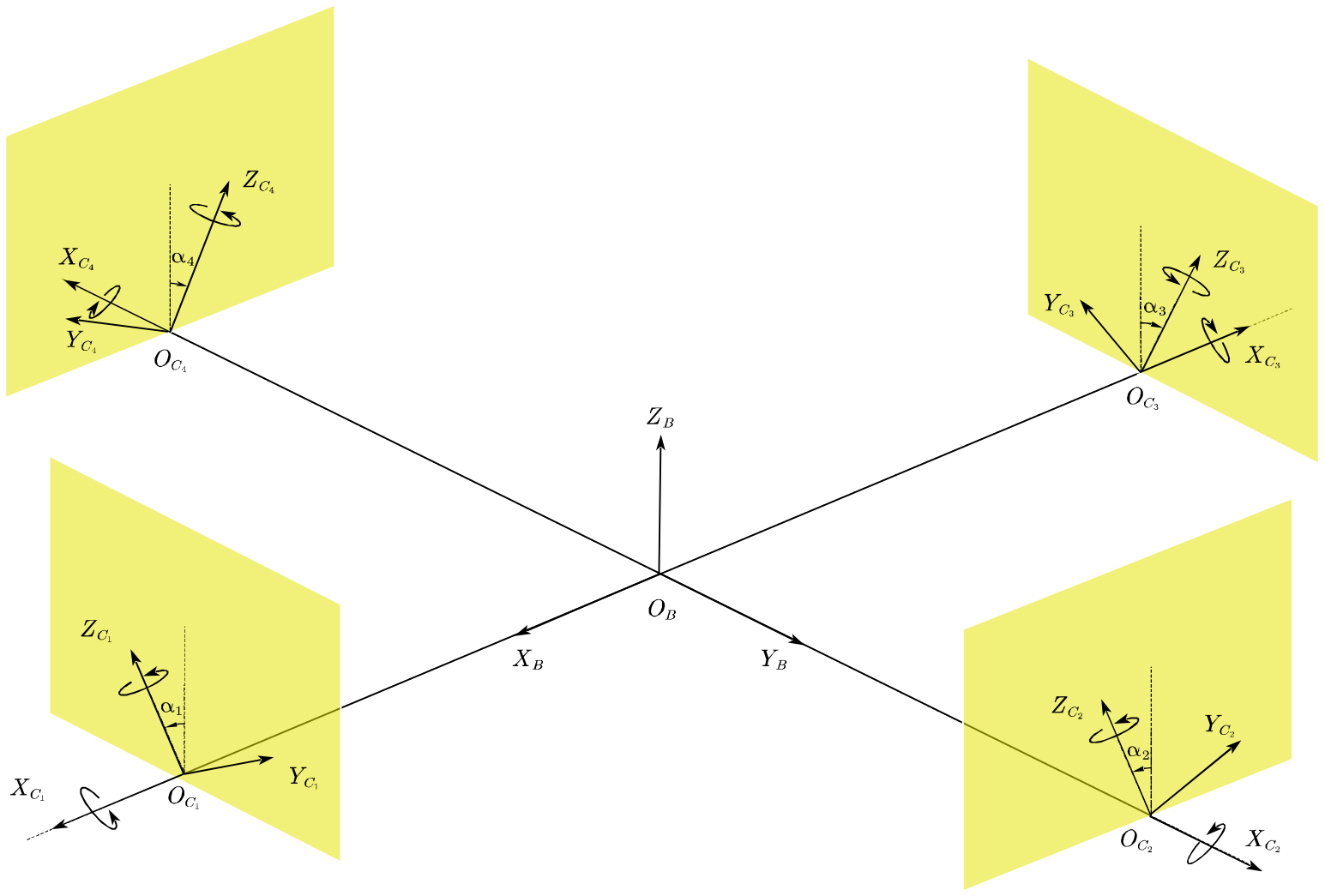

**Figure. 1.** The structure of Ryll's tilt-rotor. The directions of the thrusts are adjustable on the yellow planes during the flight.

Augmented with four tilting angles, the number of inputs in the tilt-rotor increases to eight from four, four magnitudes of the thrusts. Attempted by the over-actuated property, all the degrees of freedom (six) were initially stabilized by feedback linearization [3,4].

Besides feedback linearization, several linear and nonlinear control methods are also applied to stabilize the tilt-rotor. These controllers include LQR and PID [5,6], optimal control [7,8], backstepping and sliding mode control [9–11], adaptive control [12,13], etc. All these control methods can be classified according to the number of inputs calculated by the control rule [14]. Most of these controllers set eight inputs in a united control rule. Thus, the number of inputs for these controllers is eight. While some of them only calculate six inputs [15], the rest inputs are received by synchronizing the tilting angles. The number of inputs for these controllers is six.

Despite the success in some tracking problem for the tilt-rotor, the commonly used feedback linearization (dynamic inversion) technique may suffer from the so-called over-intensive change in the tilting angles. These tilting angles are required to change greatly within a short time step [3] or to change beyond one cycle [1]; the tilting angles are required to spin several rounds. For example, Fig. 2 is the tilting angle history in a hovering problem by feedback linearization for a tilt-rotor [2]. The tilting angles are supposed to change in high frequencies and on a large scale at the beginning.

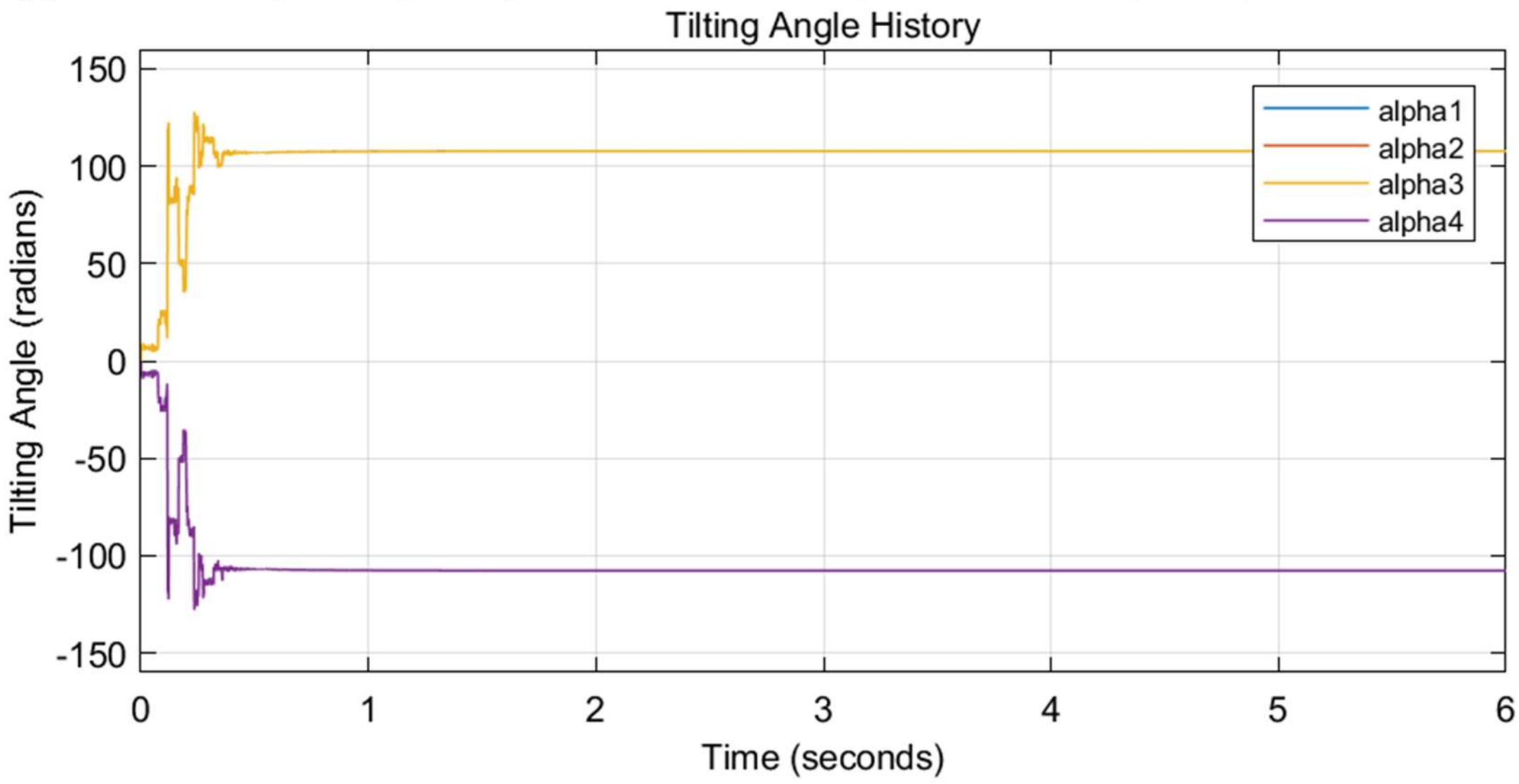

**Figure. 2.** The tilting angle history of a tilt-rotor while hovering.

These over-intensive changes in the tilting angles are resulted by the nature of feedback linearization. At each sampling time, the controller calculates the inverse of the dynamics, the result of which can be discontinuous, driving the tilting angles to change intensively. It is worth mentioning that this phenomenon is not unique in feedback linearization [16].

To stabilize the tilt-rotor by feedback linearization while avoiding the over-intensive changes in the tilting angles, our previous research put forward a scenario where the tilting angles are planned in a procedure called gait plan in advance before the subsequent feedback linearization [2]. In the gait plan procedure, four tilting angles are specified with time by the user, which is parallel to the gait plan procedure in a ground robot. The subsequent feedback linearization and controller only assign the four magnitudes of the thrusts. Then, the calculated magnitudes of the thrusts and the tilting angles (gait) planned are the eight inputs to stabilize the tilt-rotor.

It seems that the gait plan is totally independent of the subsequent feedback linearization; the tilting angles are specified by the users, which is not intervened by the controller. Several animal-inspired gaits [17–19] are adopted and show their success in the reference-tracking problems. Note that the number of inputs calculated by the controller is four, based on this scenario, marking the birth of the new branch of the tilt-rotor controllers.

As for the subsequent feedback linearization after the gait plan, the decoupling matrix is required to be invertible; a singular decoupling matrix will cause the failure in calculating finite control signal [20–22]. The singularity of the decoupling matrix, in this case, is highly influenced by the tilting angles as well as the attitude, roll and pitch in specific. The tilt-rotor will not be allowed to maneuver to several roll and pitch angles [2]. We call these roll and pitch angles as unacceptable attitudes.

Clearly, for roll-pitch diagram fully occupied by the unacceptable attitudes, given the specific gait, feedback linearization is not able to control the tilt-rotor. On the other hand, a roll-pitch diagram without the occupation of the unacceptable attitudes gives more freedom for the feedback linearization to control the tilt-rotor; it leaves more acceptable attitudes for the feedback linearization. Generally, the larger the acceptable region of attitudes is, the more robust this gait is.

Our previous research [17] proved that the robustness can be increased by the scaling method. Scaling the gait enlarges the acceptable region of attitudes. However, this method partially sacrifices the lateral force. In alternative, Two Color Map Theorem [23,24] is established to guide the gait plan in a general way, which gives more freedom to plan a desired gait. The limit of the Two Color Map Theorem is that the choice of $\alpha_1, \alpha_2$ is not arbitrary, which narrows the exploration of the robust gaits.

The main contribution of this paper is to advance Two Color Map Theorem. The generalized theorem, Generalized Two Color Map Theorem, is a complete theorem that includes all the continuous robust gaits. The robustness of several continuous robust gaits, which are ignored by Two Color Map Theorem but can be searched by Generalized Two Color Map Theorem, will be analyzed later in this paper.

The rest of this paper is organized as follows. Section 2 introduces the related works on Two Color Map Theorem. A generalized theorem, Generalized Two Color Map Theorem, is put forward in Section 3. The robustness of three robust gaits satisfying Generalized Two Color Map Theorem is analyzed in Section 4. Finally, Section 5 addresses the conclusions and discussions.

## 2. Related Work in Gait Plan for a Tiltrotor

### *2.1. Two Color Map Theorem*

A gait is the combination of four time-specified tilting angles, $\alpha_1(t), \alpha_2(t), \alpha_3(t), \alpha_4(t)$. Two Color Map Theorem guides the gait plan to create robust gaits, avoiding introducing the invertible decoupling matrix in large range in the subsequent feedback linearization in a large attitude region.

As the first step, users specify $\alpha_1(t), \alpha_2(t)$, continuously. In Two Color Map Theorem [23], $\alpha_1(t), \alpha_2(t)$ are required to vary while satisfying (1) ~ (2) for the time point $t_1$ where $\dot{\alpha}_1$ and $\dot{\alpha}_2$ exist.

$$\begin{vmatrix} \dot{\alpha}_1(t_1) & \dot{\alpha}_2(t_1) \\ 1 & 0 \end{vmatrix} = 0 \text{ or } \begin{vmatrix} \dot{\alpha}_1(t_1) & \dot{\alpha}_2(t_1) \\ 0 & 1 \end{vmatrix} = 0. \tag{1}$$

$$\|[\dot{\alpha}_1(t_1) \quad \dot{\alpha}_2(t_1)]\|_2 \neq 0. \tag{2}$$

In plain words, $(\alpha_1(t), \alpha_2(t))$ is required to move either horizontally or vertically on the $\alpha_1(t) - \alpha_2(t)$ diagram at any given time point. Users are allowed to specify a continuous $(\alpha_1(t), \alpha_2(t))$ freely based on this pattern.

After specifying $(\alpha_1(t), \alpha_2(t))$, $(\alpha_3(t), \alpha_4(t))$ are to be specified in the second step. Our previous research [23] proved that: To create a robust gait, there are only two candidates of $(\alpha_3, \alpha_4)$ corresponding to a given $(\alpha_1, \alpha_2)$, in general.

Further, observed from the $0\alpha_1\alpha_2\alpha_3$ coordinate system, these candidate $\alpha_3$ distribute on two planes. Also, observed from the $0\alpha_1\alpha_2\alpha_4$ coordinate system, these candidate $\alpha_4$ distribute on another two planes. We refer the distributions of these candidate $(\alpha_3, \alpha_4)$ to our previous publication [23].

Color the candidate $(\alpha_3, \alpha_4)$ on the first two corresponding planes in $0\alpha_1\alpha_2\alpha_3$ and $0\alpha_1\alpha_2\alpha_4$ red. Color the candidate $(\alpha_3, \alpha_4)$ on the other two corresponding planes in $0\alpha_1\alpha_2\alpha_3$ and $0\alpha_1\alpha_2\alpha_4$ blue. Then we receive the Two Color Map of $\alpha_1(t) - \alpha_2(t)$ diagram. A part of the Two Color Map is displayed in Fig. 3 [23], where the $(\alpha_1, \alpha_2)$ in half red and half blue represent that there are two candidate $(\alpha_3, \alpha_4)$. The user has to decide one. We refer the complete Two Color Map to the publication [24].

With this property, the selection of the candidate $(\alpha_3, \alpha_4)$ for each $(\alpha_1, \alpha_2)$ shall satisfy a rule to guarantee that the designed gait $(\alpha_1(t), \alpha_2(t), \alpha_3(t), \alpha_4(t))$ is continuous. This rule is called Two Color Map Theorem [23].

**Two Color Map Theorem:** The planned gait is continuous if the color selections of $(\alpha_1,\alpha_2)$ on the enclosed curve meet the following requirement: The adjacent $(\alpha_1,\alpha_2)$ are in the same color or do not violate the crossover rule.

**Proof:** See [23].

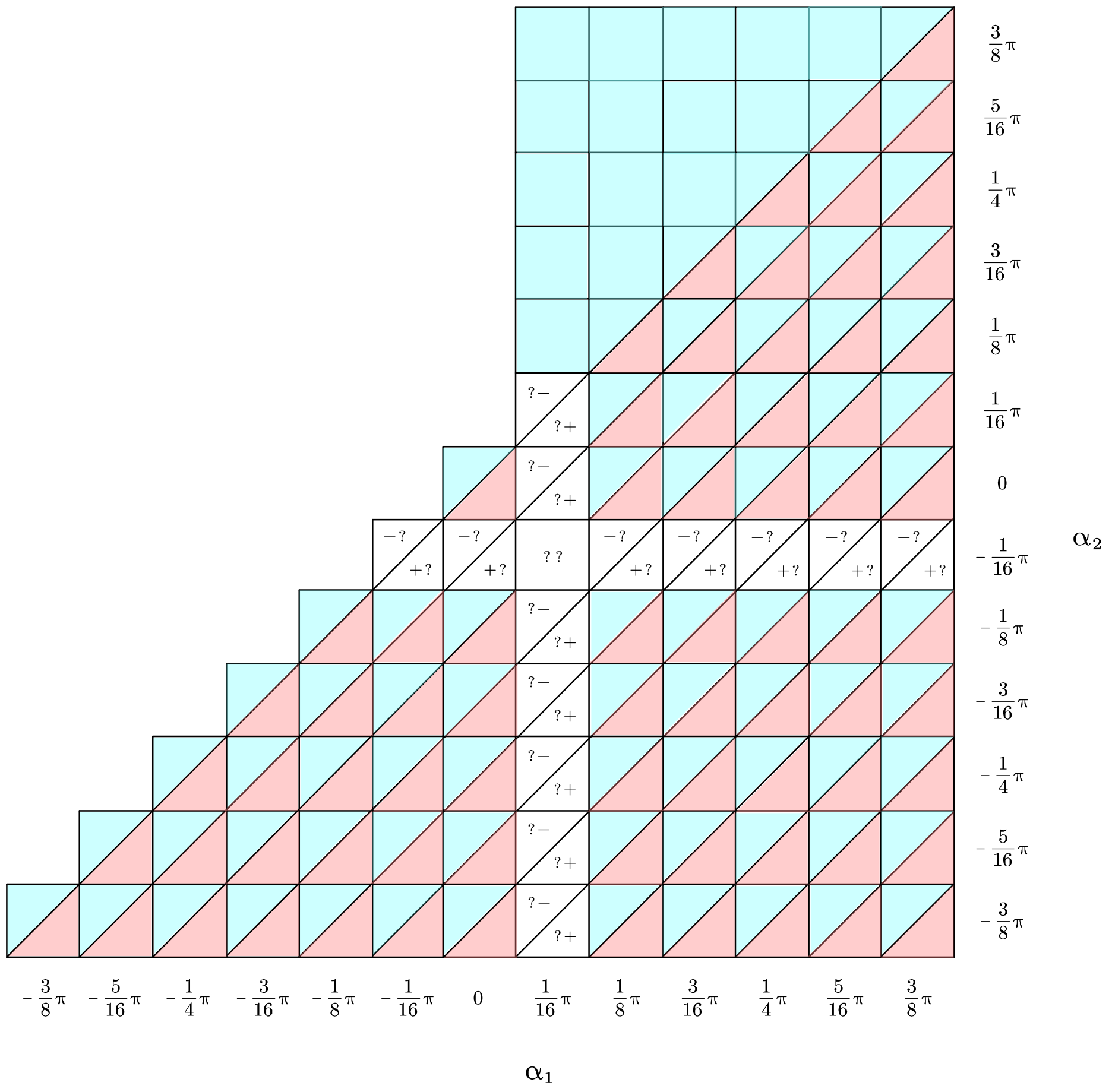

**Figure 3.** A part of Two Color Map. The red $(\alpha_1, \alpha_2)$ correspond to the $(\alpha_3, \alpha_4)$ where all $\alpha_3$ are on the same plane and all $\alpha_4$ are on the same plane. The $(\alpha_1, \alpha_2)$ in half red and half blue represent that there are two candidate $(\alpha_3, \alpha_4)$. The user has to decide one.

The gaits planned on the Two Color Map will be robust gaits. While the robust gaits which satisfies Two Color Map Theorem in the meanwhile are the continuous robust gaits, which are expected in application.

### *2.2. Surfaces in Two Color Map Theorem*

The robustness of each gait is evaluated by the region of the unacceptable attitudes on the roll-pitch diagram, which will introduce the singular decoupling matrix in the feedback linearization [20,25].

The determinant of the decoupling matrix is influenced by six variables, $\alpha_1$, $\alpha_2$, $\alpha_3$, $\alpha_4$, $\phi$ (roll), and $\theta$ (pitch). After specifying the gait with time, $(\alpha_1(t), \alpha_2(t), \alpha_3(t), \alpha_4(t))$, the combinations of $(\phi, \theta)$ which results zero determinants can be found on the $\phi - \theta$ diagram. These attitudes are not acceptable which hinder the application of feedback linearization.

In general, the further these curves are from the $(\phi, \theta) = (0,0)$, the more robust the gait is. For example, the curves in Fig. 4 and Fig. 5 are the unacceptable attitudes for the two gaits on the blue surface and the red surface, respectively. Obviously, the gait on the red surface (Fig. 5) is more robust. The details of these two gaits have been illustrated in our previous research [23].

One may notice that the robustness of these two gaits on two surfaces varies greatly although both gaits satisfy Two Color Map Theorem. This is because that Two Color Map Theorem only guarantees that the attitude region ***near*** $(\phi, \theta) = (0,0)$ introduces the invertible decoupling matrix [23]. Whereas the further region is not taken into consideration by this theorem.

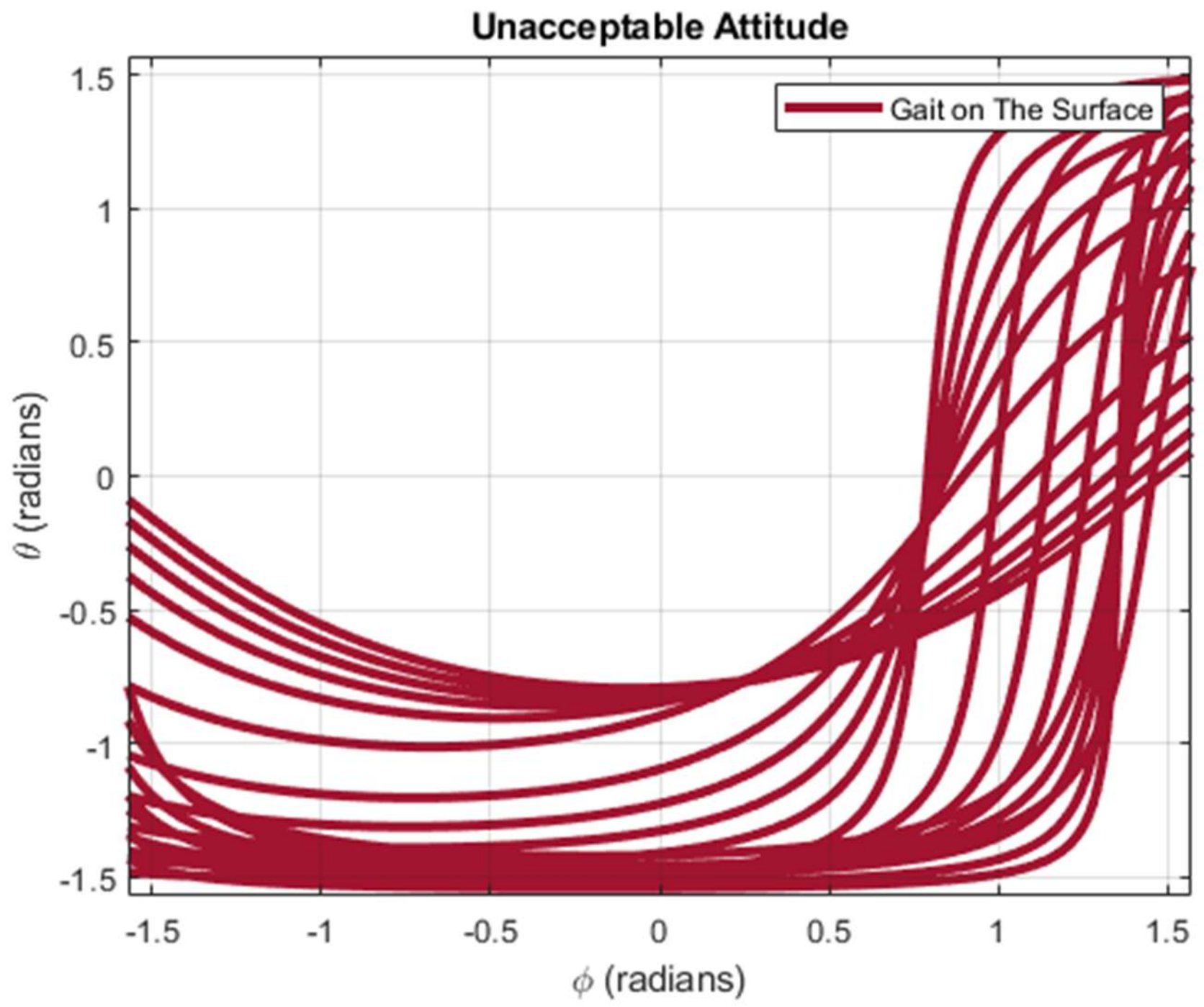

**Figure 4.** The unacceptable attitudes for one gait on the blue surface. All these curves represent the attitudes that will introduce the singular decoupling matrix in feedback linearization. Note that these curves are relatively close to $(\phi, \theta) = (0,0)$.

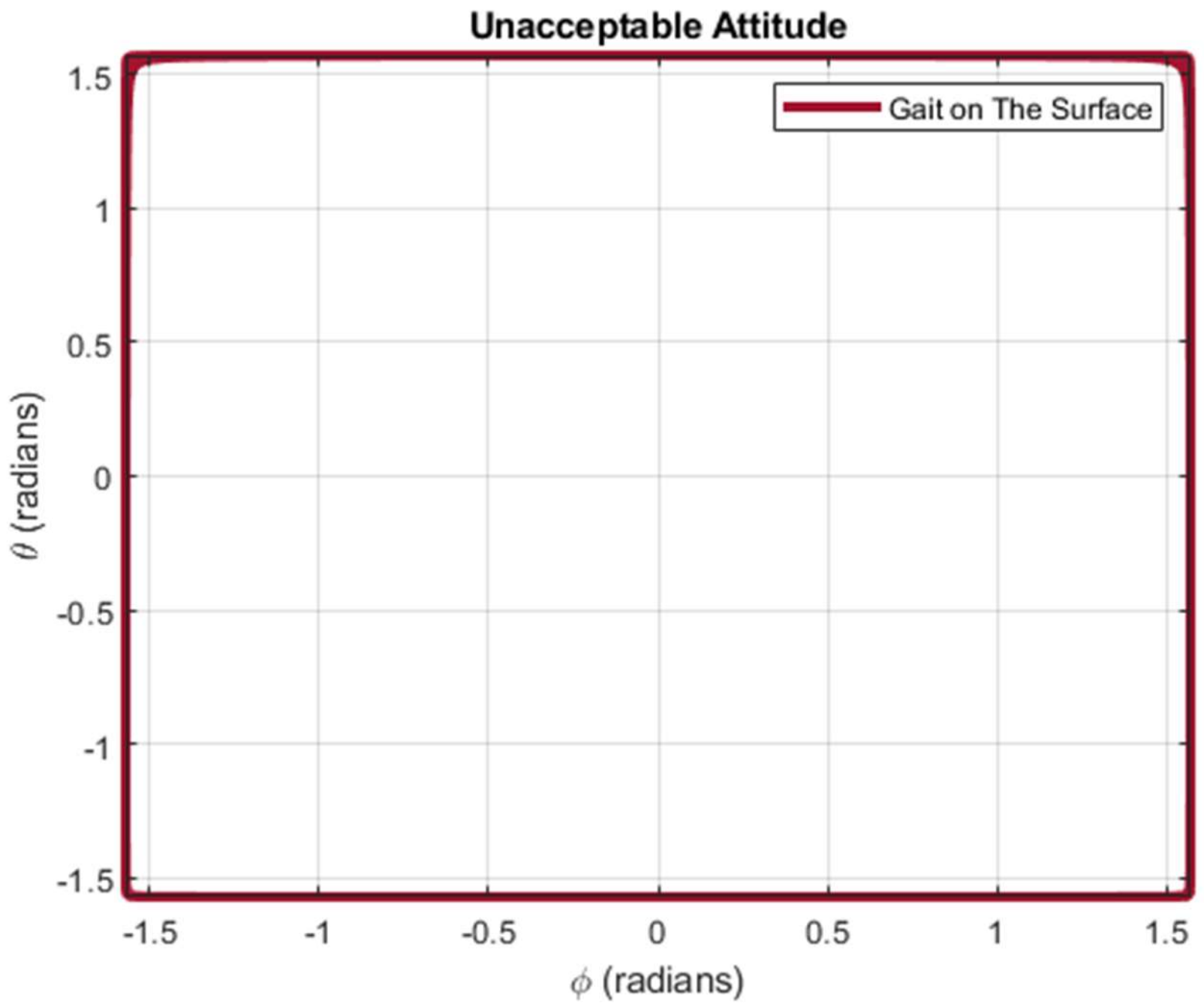

**Figure 5.** The unacceptable attitudes for one gait on the red surface. All these curves represent the attitudes that will introduce the singular decoupling matrix in feedback linearization. Note that these curves are far from $(\phi, \theta) = (0,0)$

The underlying mechanism is that the coefficient of $\phi$ and $\theta$ ( $R_\phi(\alpha_1,\alpha_2,\alpha_3,\alpha_4)$ and $R_\theta(\alpha_1,\alpha_2,\alpha_3,\alpha_4)$ in the Ref. [23]) are not exact zero in the blue surface after linearization. In other words, the effect of $\phi$ and $\theta$ is not totally suppressed. While the coefficient of $\phi$ and $\theta$ are exact zero in the red surface after linearization; the contribution of $\phi$ and $\theta$ is totally cancelled.

Further thorough discussions on this phenomenon are beyond the scope of this paper.

## 3. Generalized Two Color Map Theorem

Although Ref. [24] gives a complete Two Color Map, the designation of the robust gaits still follow the rectangular $(\alpha_1(t), \alpha_2(t))$ defined in (1) ~ (2). These requirements guarantee the continuity of the gaits. They may, on the other hand, rule out some robust gaits. E.g., the gait whose $\dot{\alpha}_1(t_1) \neq 0$ and $\dot{\alpha}_2(t_1) \neq 0$ at a time point $t_1$ is not included by Two Color Map Theorem since it violates (1).

With these concerns, we generalize Two Color Map Theorem in this section; Generalized Two Color Map Theorem gives more freedom to the gait plan on $\alpha_1 \in [-1,1] \cap \alpha_2 \in [-1,1]$.

### *3.1. Initial $(\alpha_1(t), \alpha_2(t))$ and Color*

The first step in planning a gait is determining the initial $(\alpha_1, \alpha_2)$ at $t = 0$ as well as the color of this initial $(\alpha_1, \alpha_2)$. Paint it either blue or red.

The first restriction of the initial $(\alpha_1, \alpha_2)$ is

$$\alpha_1 \in [-1,1] \cap \alpha_2 \in [-1,1]. \tag{3}$$

If the initial $(\alpha_1, \alpha_2)$ is painted in blue, the only restriction that it should obey is (3).

While if the initial $(\alpha_1, \alpha_2)$ is painted in red, it should further satisfy the following initial restrictions in (4), which is equivalent to (5).

$$(\alpha_1,\alpha_2) \notin \mathcal{L}, \tag{4}$$

$$(\alpha_1,\alpha_2) \in S_U \cup S_D, \tag{5}$$

where $\mathcal{L}$ is a curve within the region (3), defined by (6), space $S_U$ belongs to region (3), defined by (8), and space $S_D$ belongs to region (3), defined by (9), respectively.

$$\mathcal{L}: R(\alpha_1,\alpha_2) = 0, \tag{6}$$

where

$$R(\alpha_1,\alpha_2) = 4.000 \cdot c1 \cdot c2 \cdot c1 \cdot c2 + 5.592 \cdot c1 \cdot c2 \cdot c1 \cdot s2 - 5.592 \cdot c1 \cdot c2 \cdot s1 \cdot c2 + 5.592 \cdot c1 \cdot s2 \cdot c1 \cdot c2 - 5.592 \cdot s1 \cdot c2 \cdot c1 \cdot c2 + 0.9716 \cdot c1 \cdot c2 \cdot s1 \cdot s2 + 0.9716 \cdot c1 \cdot s2 \cdot s1 \cdot c2 + 0.9716 \cdot s1 \cdot c2 \cdot c1 \cdot s2 + 0.9716 \cdot s1 \cdot s2 \cdot c1 \cdot c2 - 2.000 \cdot c1 \cdot s2 \cdot c1 \cdot s2 - 2.000 \cdot s1 \cdot c2 \cdot s1 \cdot c2 - 0.1687 \cdot c1 \cdot s2 \cdot s1 \cdot s2 + 0.1687 \cdot s1 \cdot c2 \cdot s1 \cdot s2 - 0.1687 \cdot s1 \cdot s2 \cdot c1 \cdot s2 + 0.1687 \cdot s1 \cdot s2 \cdot s1 \cdot c2, \tag{7}$$

where $si = \sin(\alpha_i)$, $ci = \cos(\alpha_i)$, $(i = 1,2)$.

Fig. 6 visualizes $\mathcal{L}$ in $\alpha_1 - \alpha_2$ diagram. It is a black curve that divides the space (3) into 2 separate subspaces, one upper subspace above $\mathcal{L}$ and one lower space below $\mathcal{L}$.

It should be pointed out that a special point, Point $\mathcal{P}(\alpha_1, \alpha_2) = (0.167099, -0.167099)$, belongs to the lower subspace in Fig. 6. In other words, $\mathcal{P}$ lies below $\mathcal{L}$ rather than on it.

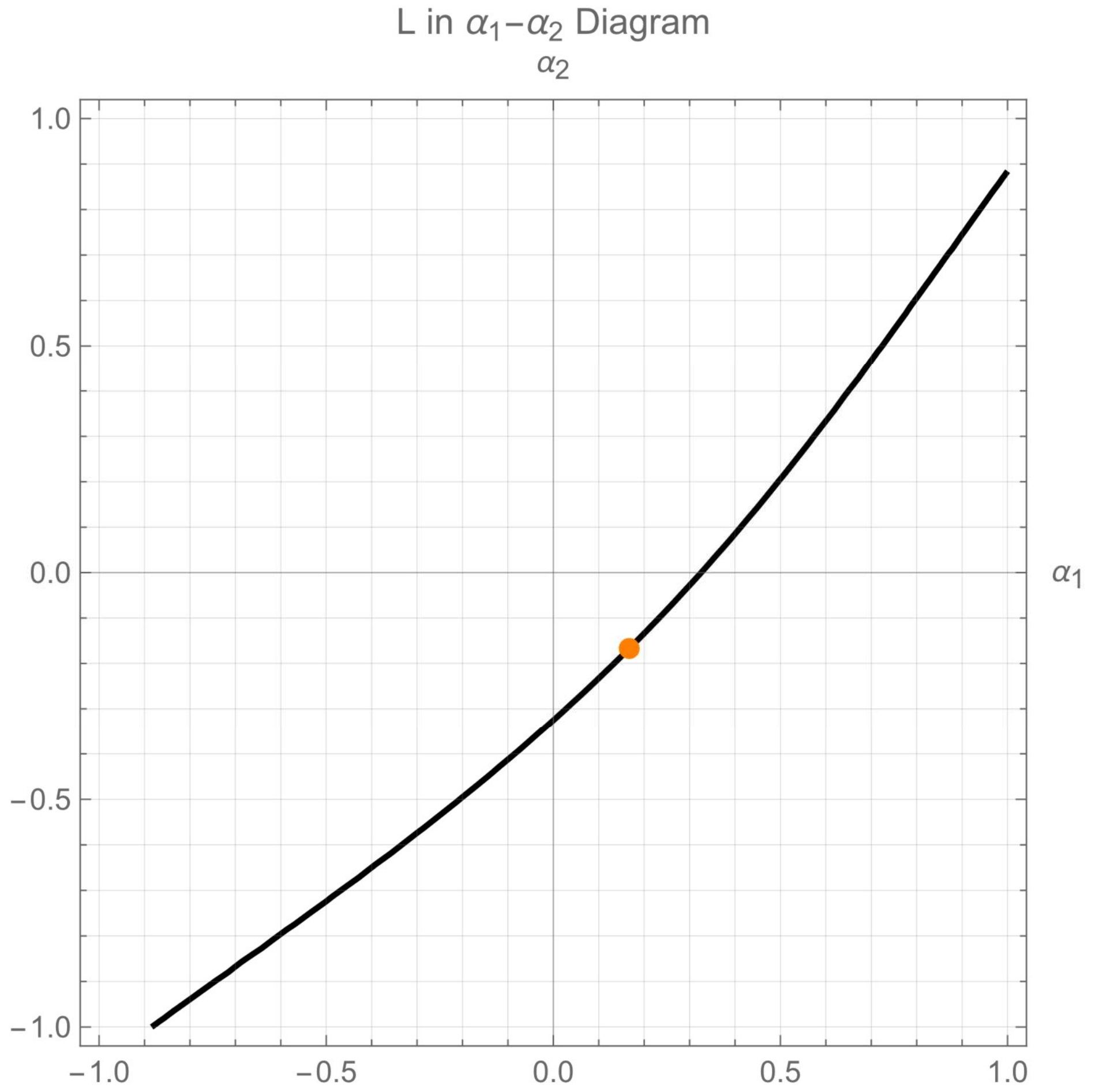

**Figure 6.** Curve $\mathcal{L}$ in $\alpha_1 - \alpha_2$ diagram. It divides the whole space into two subspaces, one upper space above $\mathcal{L}$ and one lower space below $\mathcal{L}$. The orange point is at $(\alpha_1, \alpha_2) = (0.167099, -0.167099)$. Note that this point is a little below $\mathcal{L}$ rather than on $\mathcal{L}$.

$S_U$ and $S_D$ satisfy

$$S_U: R(\alpha_1,\alpha_2) > 0, \tag{8}$$

$$S_D: R(\alpha_1,\alpha_2) < 0. \tag{9}$$

Actually, $S_U$ is exactly the upper subspace in Fig. 6. While $S_D$ is exactly the lower subspace in Fig. 6. Obviously, $\mathcal{P} \in S_D$.

### *3.2. Start $(\alpha_1(t), \alpha_2(t))$ in Red inside $S_U$*

If $(\alpha_1(t), \alpha_2(t))$ starts in red inside $S_U$, $(\alpha_1(t), \alpha_2(t))$ is not allowed to escape from $S_U$ for any later time, e.g., $t > 0$.

Further, $(\alpha_1(t), \alpha_2(t))$ should be always red for any given time point while being continuous, e.g., satisfying (10).

$$\forall t_1 \geqslant 0, \lim_{t \to {t_1}^+} \alpha_i(t) = \lim_{t \to {t_1}^-} \alpha_i(t) = \alpha_i(t_1), i = 1,2. \tag{10}$$

This completes the Generalized Two Color Map Theorem for the case where $(\alpha_1(t), \alpha_2(t))$ starts in Red inside $S_U$.

### *3.3. Start $(\alpha_1(t), \alpha_2(t))$ in Red inside $S_D$ or Start $(\alpha_1(t), \alpha_2(t))$ in Blue at Any Position*

This section discusses the case of starting $(\alpha_1(t), \alpha_2(t))$ in red inside $S_D$ and the case of starting $(\alpha_1(t), \alpha_2(t))$ in blue at any position.

If the current color of $(\alpha_1(t), \alpha_2(t))$ is red inside $S_D$, $(\alpha_1(t), \alpha_2(t))$ is not allowed to escape from $S_D$ for any later time, e.g., $t > 0$, given that $(\alpha_1(t), \alpha_2(t))$ is always red for any subsequent time points.

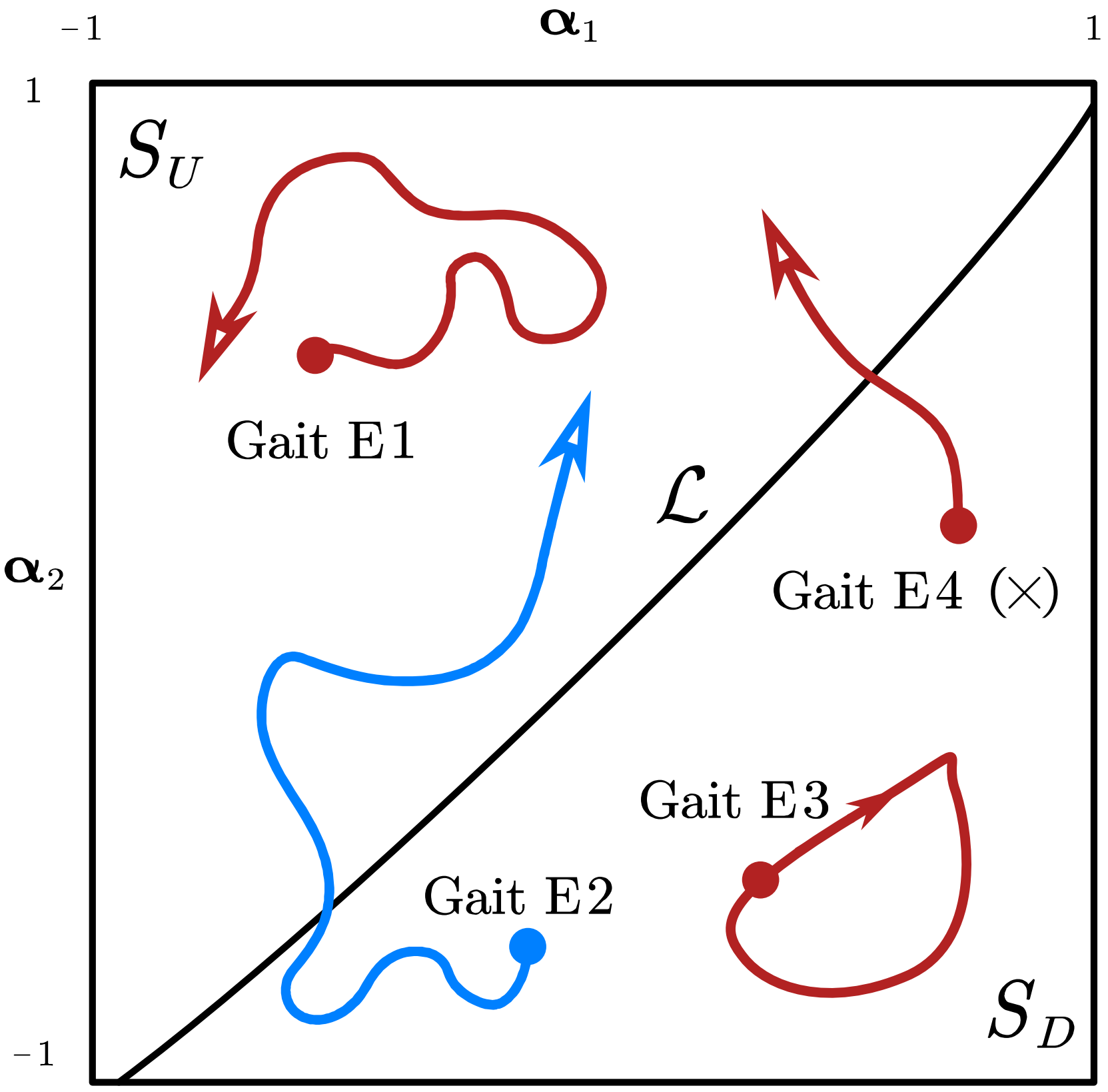

**Figure 7.** Four examples of gaits. Gait E1, Gait E3, and Gait E4 start with red $(\alpha_1, \alpha_2)$. Gait E2 starts with blue $(\alpha_1, \alpha_2)$. Gait E1 follows Generalized Two Color Map Theorem in III. B. Gait E2 and E3 follow Generalized Two Color Map Theorem in III. C. Gait E4 does not follow the Generalized Two Color Map Theorem in III. C; since it should not escape $S_D$ if its current color is red. So Gait E4 is not a robust gait.

If the current color of $(\alpha_1(t), \alpha_2(t))$ is blue at any position, $(\alpha_1(t), \alpha_2(t))$ is allowed to travel inside the entire space defined by (3) for any later time, e.g., $t > 0$, given that $(\alpha_1(t), \alpha_2(t))$ is always blue for any subsequent time points.

Still, $(\alpha_1(t), \alpha_2(t))$ is required to be continuous at any time, e.g., satisfying (10).

As an illustration, Fig. 7 gives four gaits, Gait E1, Gait E2, Gait E3, and Gait E4, on the $\alpha_1 - \alpha_2$ diagram. Gait E1 satisfies the requirements in Section 3. B. Gait E2 and Gait E3 satisfy the requirements in Section 3. C. So, Gait E1 ~ E3 are robust gaits satisfying Generalized Two Color Map Theorem. While Gait E4 violates the requirements in Section 3. C. So, Gait E4 is not a robust gait.

Note that it is possible to change the color of $(\alpha_1, \alpha_2)$ at the special point, $\mathcal{P}$, on $\alpha_1(t) - \alpha_2(t)$ diagram. This color switch is detailed in Section 3. 4.

### *3.4. Color Switch*

$(\alpha_1, \alpha_2)$ with the initial condition discussed in Section 3. C is in the same color for the time points later. While there is a special point, on $\alpha_1(t) - \alpha_2(t)$ diagram, where changing the color of the current $(\alpha_1, \alpha_2)$ is possible.

At point $\mathcal{P}(\alpha_1, \alpha_2) = (0.167099, -0.167099)$, $(\alpha_1, \alpha_2)$ is able to change the color, e.g., from red to blue or from blue to red. Note that $(\alpha_1, \alpha_2)$ can change the color if and only if it passes this point; changing the current color at any other position is not allowed. Note that this color change is not obligatory; the same color can be chosen for $(\alpha_1, \alpha_2)$ after passing $\mathcal{P}$.

Note that after changing the color, $(\alpha_1, \alpha_2)$ should follow the rule in Section 3. C based on the new current color. E.g., $(\alpha_1, \alpha_2)$ is allowed to travel in the whole space defined in (3) if its new current color is blue or is not allowed to escape $S_D$ if its new current color is red.

A gait with color change will be discussed in Section 4.

*3.5. Proof of Continuity*

Since we define $(\alpha_3, \alpha_4)$ in this research on the Two Color Map [23], the resulting gait $(\alpha_1, \alpha_2, \alpha_3, \alpha_4)$ is robust. This section proves the continuity of the resulting $(\alpha_3, \alpha_4)$.

Based on Ref. [23], the $(\alpha_3, \alpha_4)$ corresponding to red $(\alpha_1, \alpha_2)$ satisfy

$$\begin{cases} \alpha_3 = \alpha_1 \\ \alpha_4 = \alpha_2 \end{cases}. \tag{11}$$

And, the $(\alpha_3, \alpha_4)$ corresponding to blue $(\alpha_1, \alpha_2)$ satisfy

$$\begin{cases} \alpha_3 = -\alpha_1 + 0.334198 \\ \alpha_4 = -\alpha_2 - 0.334198 \end{cases}. \tag{12}$$

Obviously, $(\alpha_3, \alpha_4)$ is continuous corresponding to the $(\alpha_1, \alpha_2)$ of the same color, given that $(\alpha_1, \alpha_2)$ is changing continuously, e.g., satisfying (10). Thus, $(\alpha_1, \alpha_2, \alpha_3, \alpha_4)$ is a continuous gait for case with $(\alpha_1, \alpha_2)$ of same color.

Notice that Point $\mathcal{P}(\alpha_1, \alpha_2) = (0.167099, -0.167099)$ results the identical $(\alpha_3, \alpha_4)$ in both (11) and (12), enabling to change the color of $(\alpha_1, \alpha_2)$ while holding a continuous $(\alpha_3, \alpha_4)$, e.g., the left limit equals to the right limit, which equals to the value defined at Point $P$ for both $\alpha_1$ and $\alpha_2$. Thus, switching the color at $\mathcal{P}$ will not destroy the continuity of the gait. In conclusion, $(\alpha_3, \alpha_4)$ is always continuous.

Further, $(\alpha_1, \alpha_2)$ is continuous. Thus, the Generalized Two Color Map Theorem defines the continuous robust gaits, $(\alpha_1, \alpha_2, \alpha_3, \alpha_4)$.

The subspace-division and $\mathcal{L}$ guarantees that the resulting determinant of the decoupling matrix will be non-zero; it will either be always positive or be always negative. Crossing $\mathcal{L}$ while not following the requirements defined by Generalized Two Color Map Theorem will result a crossover at zero in that determinant of the decoupling matrix.

We refer the detail of this discussion to our previous publication [23].

## 4. Robustness Analysis

This section discusses the robustness of three periodic gaits, Gait 1, Gait 2, and Gait 3, that satisfy Generalized Two Color Map Theorem. Gait 1 undergoes color switch every half period. Gait 2 and 3 are changing in a circular pattern on $\alpha_1(t) - \alpha_2(t)$ diagram. Notice that all these three gaits does not satisfy Two Color Map Theorem since (1) ~ (2) are violated.

All these gaits are illustrated in Fig. 8.

Both Gait 2 and Gait 3 travels along a circle on $\alpha_1(t) - \alpha_2(t)$ diagram. While Gait 2 is red and Gait 3 is blue, which determine $(\alpha_3, \alpha_4)$ by (11) and (12), respectively. Both gaits maintain their won color throughout the time; no color change happens to these gaits.

Gait 1 starts with red color on the switching point, $\mathcal{P}$. After half period, it returns to point $\mathcal{P}$ and changes its color to blue before the next half period as blue. After a whole period, it returns to $\mathcal{P}$ and changes it color to red to begin a new cycle.

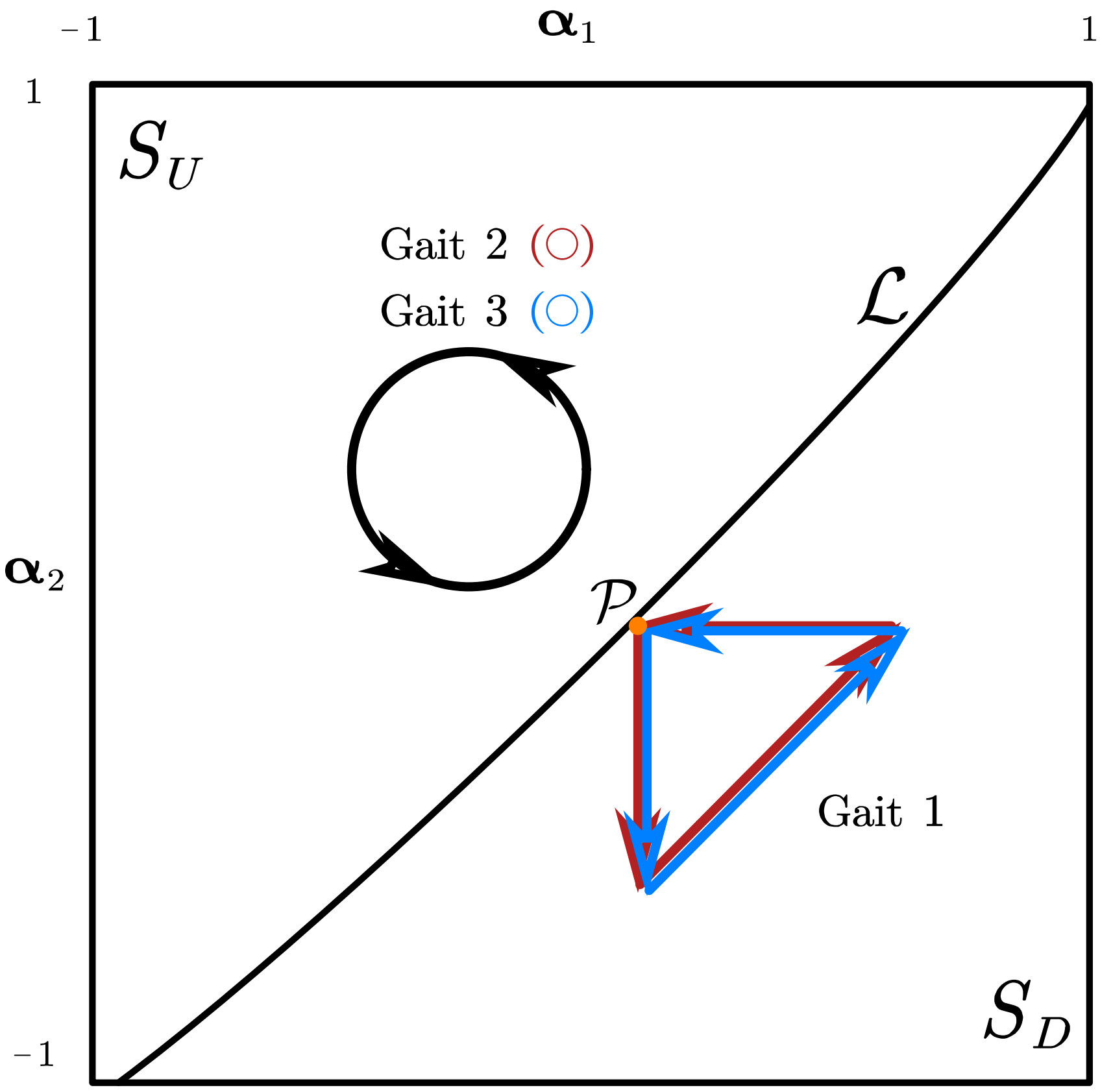

**Figure 8.** The robustness of three periodic gaits is to be analyzed. They are Gait 1, Gait 2, and Gait 3. Gait 1 travels along a triangular pattern, starting in red on point $\mathcal{P}$. It changes its color every time it passes $\mathcal{P}$. Both Gait 2 and Gait 3 travel along a circular pattern, starting in red and blue, respectively. No color change happens in Gait 2 or Gait 3.

In comparison, the corresponding adjacent biased gaits are created for each periodic gait by partially scaling, that is

$$\alpha_1 \leftarrow \alpha_1, \tag{13}$$

$$\alpha_2 \leftarrow \alpha_2, \tag{14}$$

$$\alpha_3 \leftarrow \eta \cdot \alpha_3, \tag{15}$$

$$\alpha_4 \leftarrow \eta \cdot \alpha_4, \tag{16}$$

where $\eta$ is the scaling coefficient. In this experiment, $\eta = 80\%$ for all the biased gaits.

*4.1. Gait 1*

The angle history of Gait 1 is plotted in Fig. 9. The unacceptable attitudes for Gait 1 and the biased Gait 1 are illustrated in Fig. 10 in red curves and blue curves, respectively.

Clearly, the biased Gait 1 leaves little acceptable attitudes for the tilt-rotor. While the unacceptable attitudes for Gait 1 are relatively far from $(\phi, \theta) = (0,0)$. This indicates that Gait 1 is more robust to the biased Gait 1.

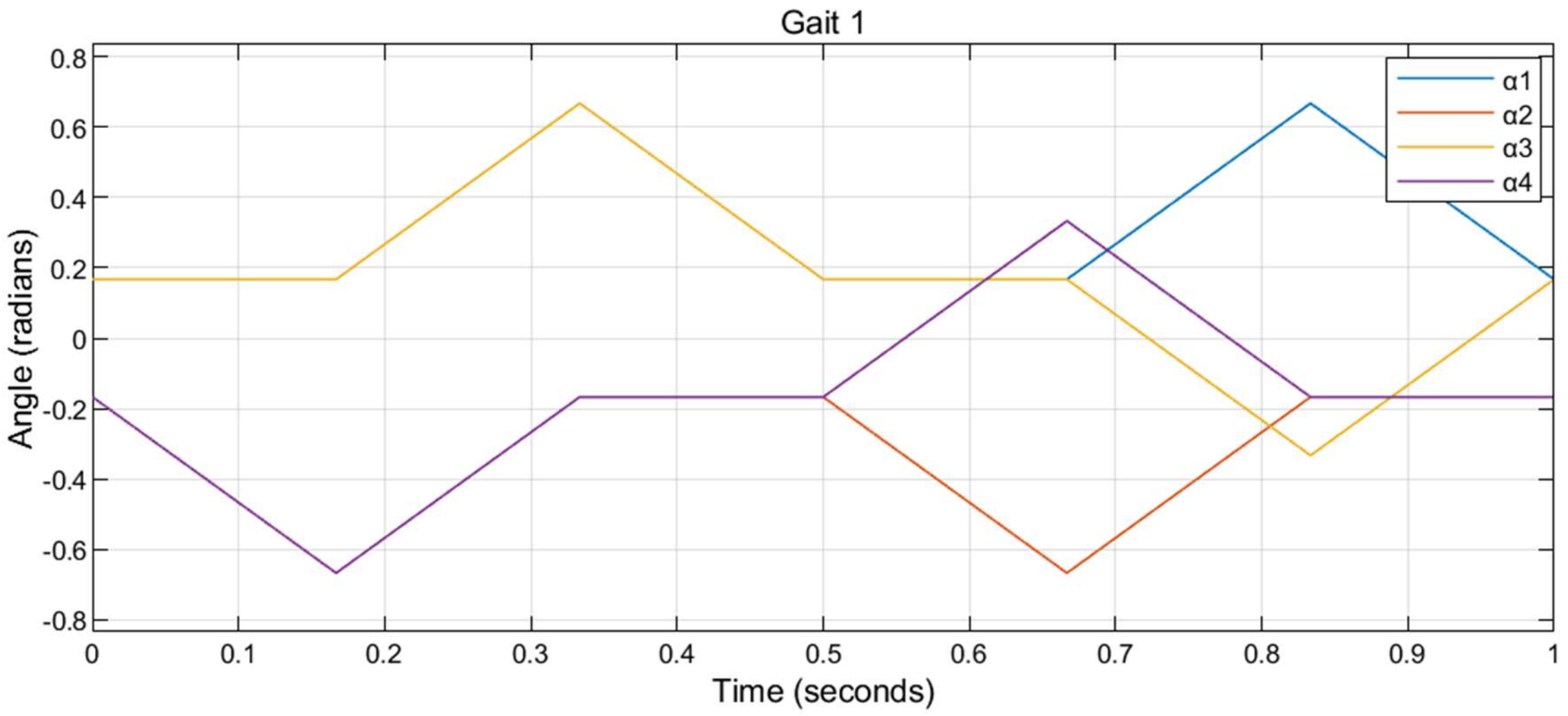

**Figure. 9.** The tilting angles in Gait 1 in the first period.

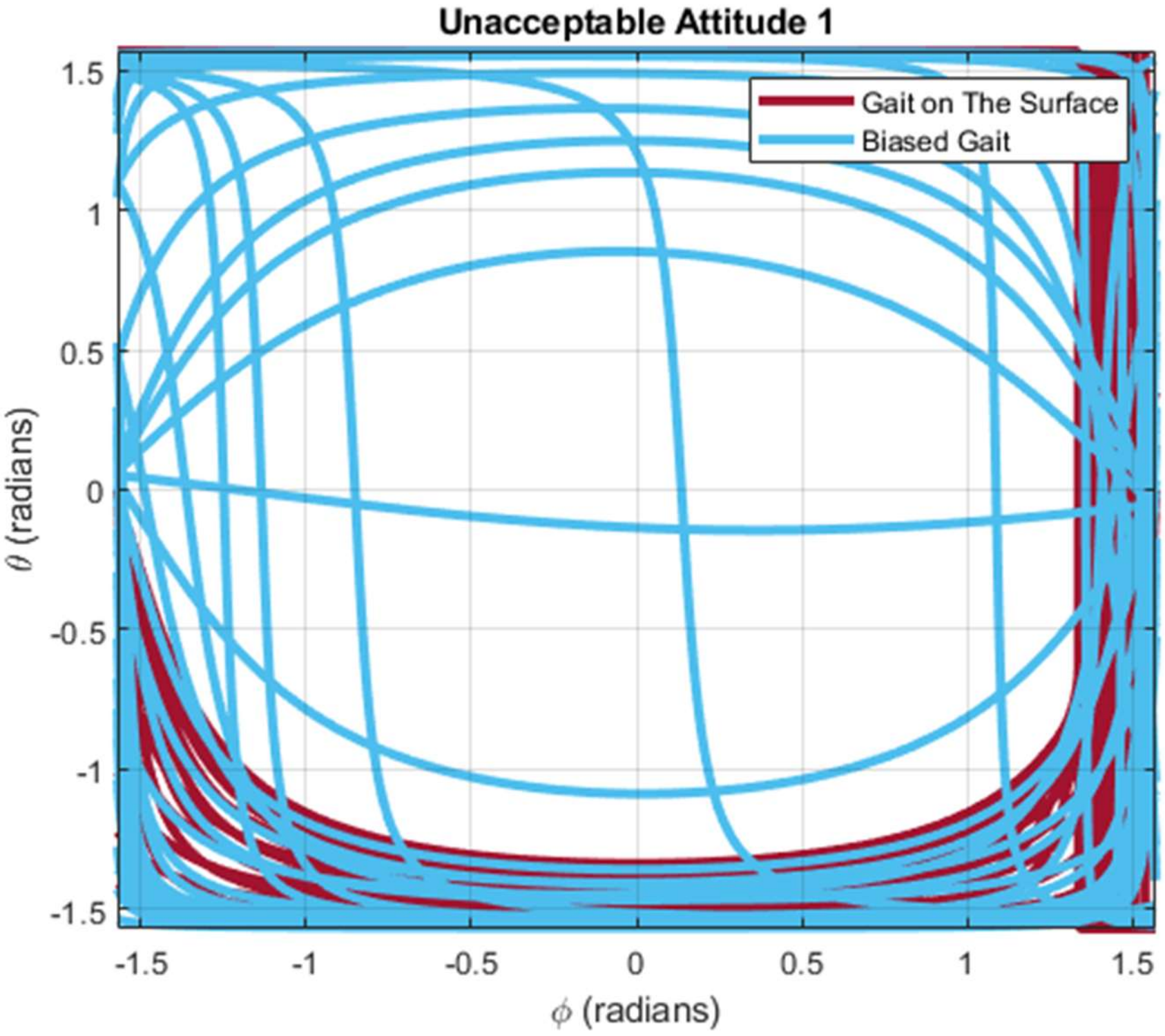

**Figure 10.** The unacceptable attitudes for Gait 1 (red curves) and its biased gait (blue curves). The red curves represent the attitudes that will introduce the singular decoupling matrix in feedback linearization for Gait 1. The blue curves represent the attitudes that will introduce the singular decoupling matrix in feedback linearization for the biased Gait 1.

### *4.2. Gait 2*

The angle history of Gait 2 is plotted in Fig. 11. The unacceptable attitudes for Gait 2 and the biased Gait 2 are illustrated in Fig. 12 in red curves and blue curves, respectively.

Interestingly, though the biased Gait 2 leaves relatively abundant acceptable attitudes for the tilt-rotor, there are no unacceptable attitudes for Gait 2; any attitude in this figure are acceptable that will introduce an invertible decoupling matrix, given that the input is not saturated.

Thus, Gait 2 is more robust to the biased Gait 2.

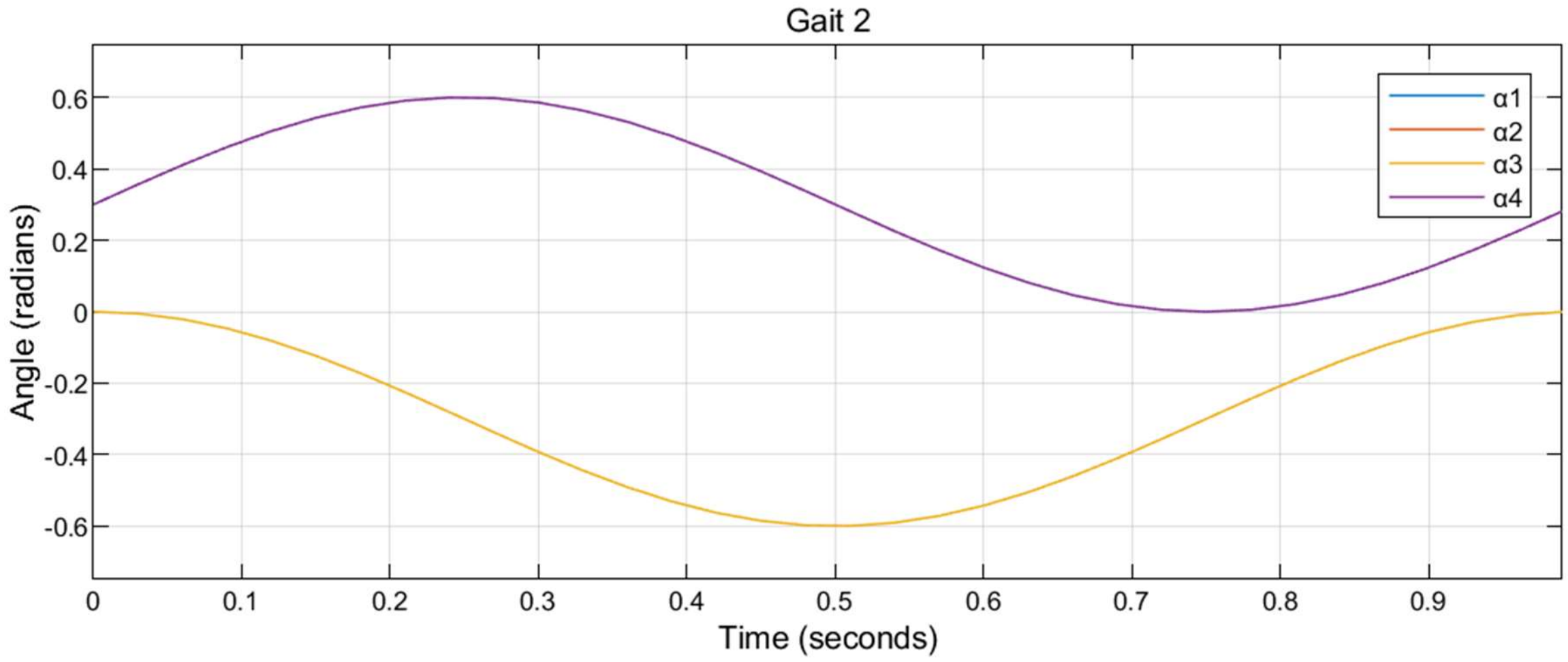

**Figure 11.** The tilting angles in Gait 2 in the first period.

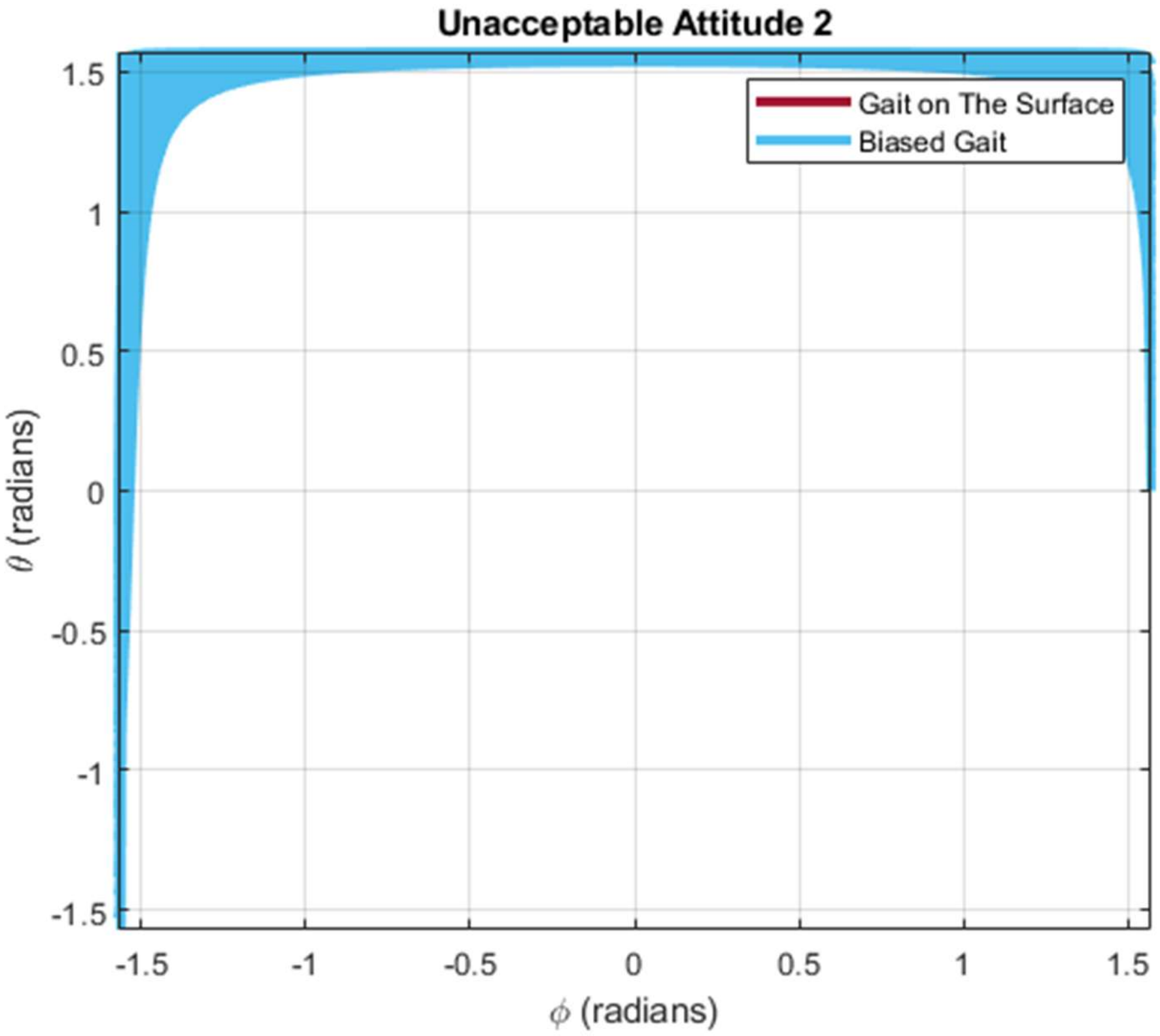

**Figure 12.** The unacceptable attitudes for Gait 2 (red curves) and its biased gait (blue curves). The red curves represent the attitudes that will introduce the singular decoupling matrix in feedback linearization for Gait 2. The blue curves represent the attitudes that will introduce the singular decoupling matrix in feedback linearization for the biased Gait 2. Notice that there are no red curves in this diagram at all.

*4.3. Gait 3*

The angle history of Gait 3 is plotted in Fig. 13. The unacceptable attitudes for Gait 3 and the biased Gait 3 are illustrated in Fig. 14 in red curves and blue curves, respectively.

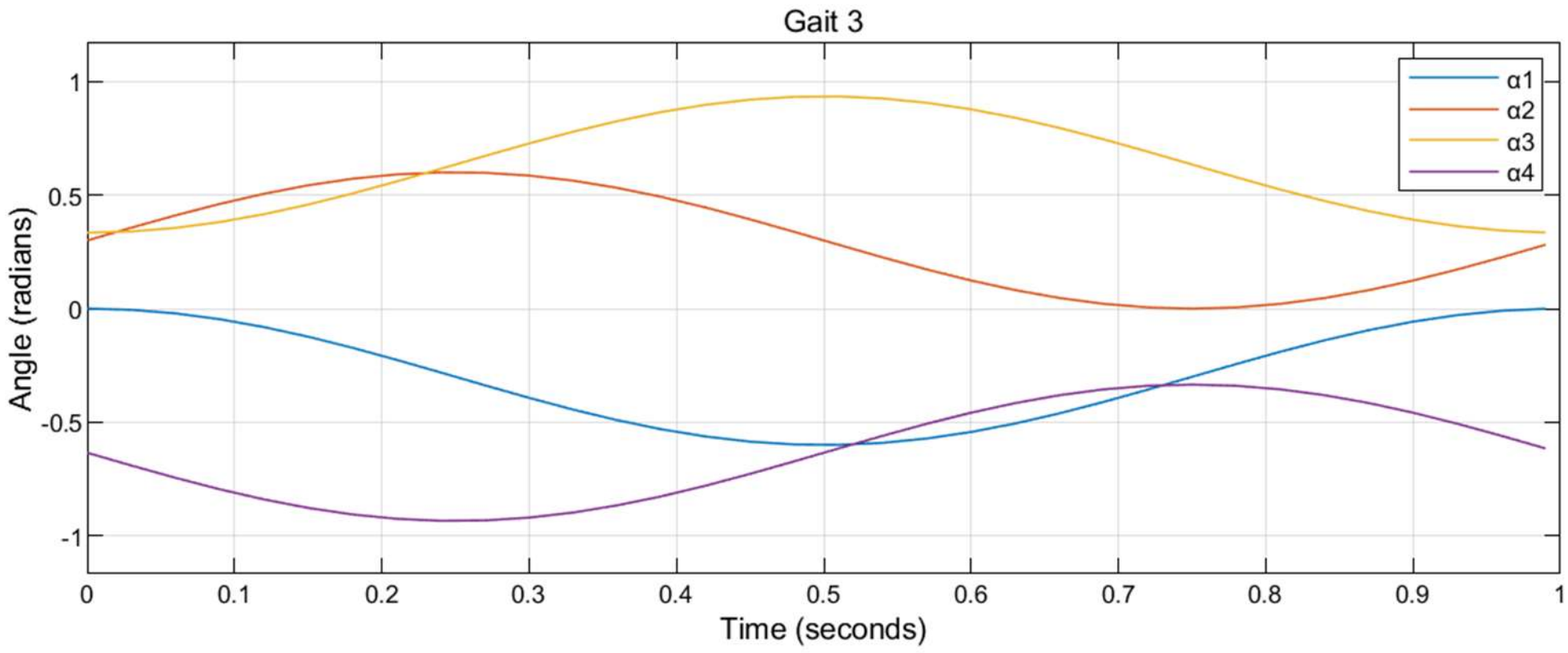

**Figure 13.** The tilting angles in Gait 3 in the first period.

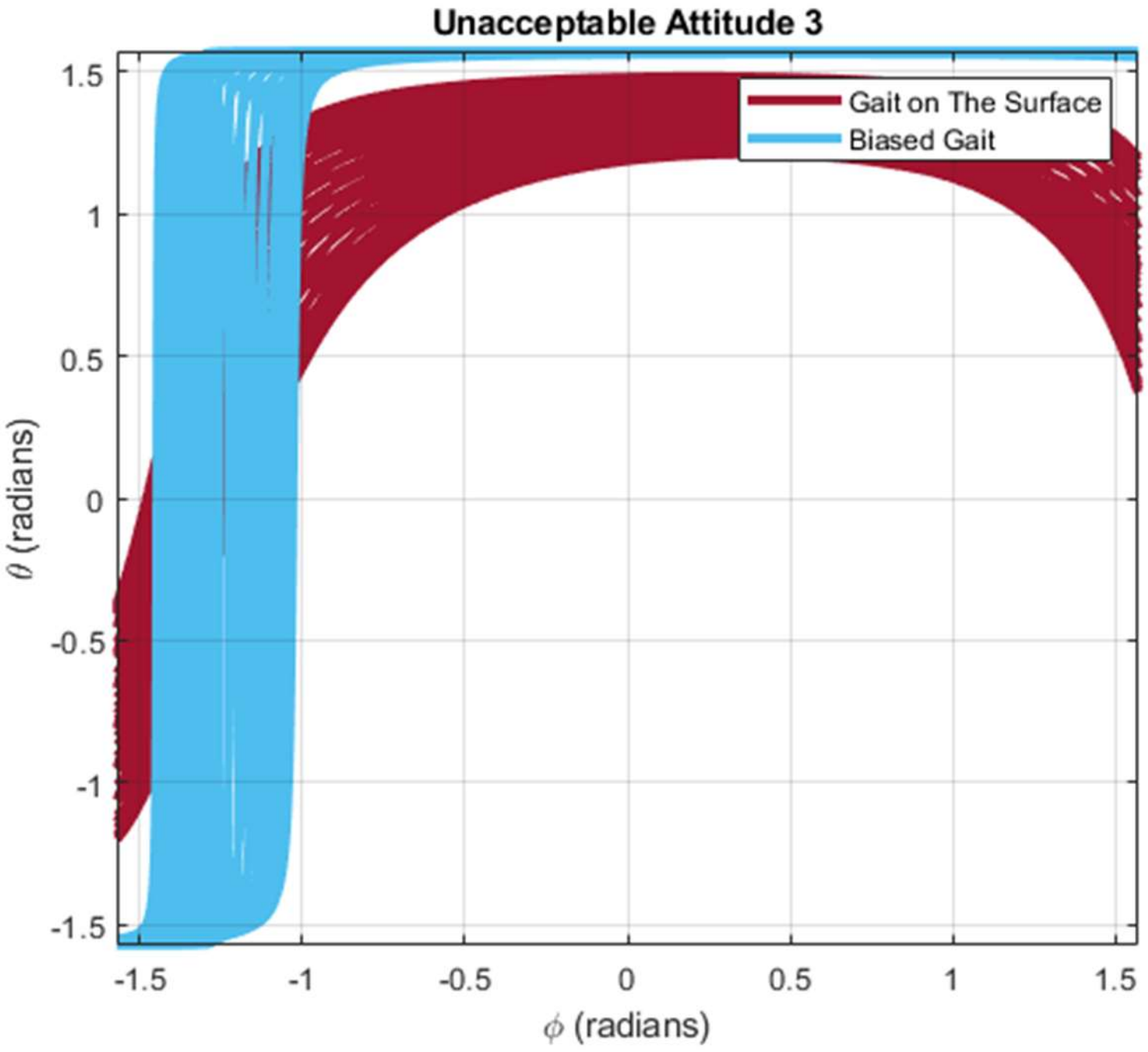

**Figure 14.** The unacceptable attitudes for Gait 3 (red curves) and its biased gait (blue curves). The red curves represent the attitudes that will introduce the singular decoupling matrix in feedback linearization for Gait 3. The blue curves represent the attitudes that will introduce the singular decoupling matrix in feedback linearization for the biased Gait 3.

It is hard to assert whether Gait 3 or the biased Gait 3 is more robust. Indeed, both gaits are relatively far from $(\phi, \theta) = (0{,}0)$, leaving sufficient attitude space for the tilt-rotor.

While it is not surprising to receive this result – gaits on the Two Color Map Theorem or Generalized Two Color Map Theorem generates robust gaits, ensuring the attitudes near $(\phi, \theta) = (0{,}0)$ are acceptable for the tilt-rotor. The theorem loses its effects for the attitudes far from the origin.

## 5. Conclusion

This research paper generalized the Two Color Map Theorem by relaxing the restrictions of the gait plan. With the sound proof deduced in this paper, the gaits planned following Generalized Two Color Map Theorem is always continuous and robust.

Comparing with the Two Color Map Theorem, the generalized theorem provides the robust gaits which are ignored by the previous former theorem. Three ignored gaits are evaluated and found their strong robustness to the attitude change. Besides, the robust gaits with color switch is evaluated for the first time in this paper.

In our future plan, we will develop tracking controller for the tilt-rotor adopting the robust gaits, relying on Generalized Two Color Map Theorem.

## References

1. Ryll, M.; Bulthoff, H.H.; Giordano, P.R. Modeling and Control of a Quadrotor UAV with Tilting Propellers. In Proceedings of the 2012 IEEE International Conference on Robotics and Automation; IEEE: St Paul, MN, USA, May 2012; pp. 4606–4613.
2. Shen, Z.; Tsuchiya, T. Gait Analysis for a Tiltrotor: The Dynamic Invertible Gait. *Robotics* **2022**, *11*, 33, doi:10.3390/robotics11020033.
3. Ryll, M.; Bulthoff, H.H.; Giordano, P.R. A Novel Overactuated Quadrotor Unmanned Aerial Vehicle: Modeling, Control, and Experimental Validation. *IEEE Trans. Contr. Syst. Technol.* **2015**, *23*, 540–556, doi:10.1109/TCST.2014.2330999.
4. Ryll, M.; Bulthoff, H.H.; Giordano, P.R. First Flight Tests for a Quadrotor UAV with Tilting Propellers. In Proceedings of the 2013 IEEE International Conference on Robotics and Automation; IEEE: Karlsruhe, Germany, May 2013; pp. 295–302.
5. Ahmed, A.M.; Katupitiya, J. Modeling and Control of a Novel Vectored-Thrust Quadcopter. *Journal of Guidance, Control, and Dynamics* **2021**, *44*, 1399–1409, doi:10.2514/1.G005467.
6. Oosedo, A.; Abiko, S.; Narasaki, S.; Kuno, A.; Konno, A.; Uchiyama, M. Flight Control Systems of a Quad Tilt Rotor Unmanned Aerial Vehicle for a Large Attitude Change. In Proceedings of the 2015 IEEE International Conference on Robotics and Automation (ICRA); IEEE: Seattle, WA, USA, May 2015; pp. 2326–2331.
7. Magariyama, T.; Abiko, S. Seamless 90-Degree Attitude Transition Flight of a Quad Tilt-Rotor UAV under Full Position Control. In Proceedings of the 2020 IEEE/ASME International Conference on Advanced Intelligent Mechatronics (AIM); IEEE: Boston, MA, USA, July 2020; pp. 839–844.
8. Park, S.; Lee, J.; Ahn, J.; Kim, M.; Her, J.; Yang, G.-H.; Lee, D. ODAR: Aerial Manipulation Platform Enabling Omnidirectional Wrench Generation. *IEEE/ASME Trans. Mechatron.* **2018**, *23*, 1907–1918, doi:10.1109/TMECH.2018.2848255.
9. Jin, S.; Kim, J.; Kim, J.-W.; Bae, J.; Bak, J.; Kim, J.; Seo, T. Back-Stepping Control Design for an Underwater Robot with Tilting Thrusters. In Proceedings of the 2015 International Conference on Advanced Robotics (ICAR); IEEE: Istanbul, Turkey, July 2015; pp. 1–8.
10. Kadiyam, J.; Santhakumar, M. Design and Implementation of Backstepping Controller for Tilting Thruster Underwater Robot. 6.
11. Phong Nguyen, N.; Kim, W.; Moon, J. Observer-Based Super-Twisting Sliding Mode Control with Fuzzy Variable Gains and Its Application to Overactuated Quadrotors. In Proceedings of the 2018 IEEE Conference on Decision and Control (CDC); IEEE: Miami Beach, FL, December 2018; pp. 5993–5998.
12. Blandino, T.; Leonessa, A.; Doyle, D.; Black, J. Position Control of an Omni-Directional Aerial Vehicle for Simulating Free-Flyer In-Space Assembly Operations. In Proceedings of the ASCEND 2021; American Institute of Aeronautics and Astronautics: Las Vegas, Nevada & Virtual, November 15 2021.
13. Lu, D.; Xiong, C.; Zeng, Z.; Lian, L. Adaptive Dynamic Surface Control for a Hybrid Aerial Underwater Vehicle With Parametric Dynamics and Uncertainties. *IEEE J. Oceanic Eng.* **2020**, *45*, 740–758, doi:10.1109/JOE.2019.2903742.
14. Hamandi, M.; Usai, F.; Sablé, Q.; Staub, N.; Tognon, M.; Franchi, A. Design of Multirotor Aerial Vehicles: A Taxonomy Based on Input Allocation. *The International Journal of Robotics Research* **2021**, *40*, 1015–1044, doi:10.1177/02783649211025998.
15. Nemati, A.; Kumar, M. Modeling and Control of a Single Axis Tilting Quadcopter. In Proceedings of the 2014 American Control Conference; IEEE: Portland, OR, USA, June 2014; pp. 3077–3082.
16. Kumar, R.; Nemati, A.; Kumar, M.; Sharma, R.; Cohen, K.; Cazaurang, F. Tilting-Rotor Quadcopter for Aggressive Flight Maneuvers Using Differential Flatness Based Flight Controller.; American Society of Mechanical Engineers: Tysons, Virginia, USA, October 11 2017; p. V003T39A006.
17. Shen, Z.; Tsuchiya, T. Cat-Inspired Gaits for a Tilt-Rotor—From Symmetrical to Asymmetrical. *Robotics* **2022**, *11*, 60, doi:10.3390/robotics11030060.
18. Shen, Z.; Ma, Y.; Tsuchiya, T. Feedback Linearization-Based Tracking Control of a Tilt-Rotor with Cat-Trot Gait Plan. *International Journal of Advanced Robotic Systems* **2022**, *19*, 17298806221109360, doi:10.1177/17298806221109360.
19. Vilensky, J.A.; Njock Libii, J.; Moore, A.M. Trot-Gallop Gait Transitions in Quadrupeds. *Physiology & Behavior* **1991**, *50*, 835–842, doi:10.1016/0031-9384(91)90026-K.
20. Shen, Z.; Tsuchiya, T. Singular Zone in Quadrotor Yaw–Position Feedback Linearization. *Drones* **2022**, *6*, 20, doi:doi.org/10.3390/drones6040084.
21. Mistler, V.; Benallegue, A.; M'Sirdi, N.K. Exact Linearization and Noninteracting Control of a 4 Rotors Helicopter via Dynamic Feedback. In Proceedings of the Proceedings 10th IEEE International Workshop on Robot and Human Interactive Communication. ROMAN 2001 (Cat. No.01TH8591); IEEE: Paris, France, September 2001; pp. 586–593.
22. Ghandour, J.; Aberkane, S.; Ponsart, J.-C. Feedback Linearization Approach for Standard and Fault Tolerant Control: Application to a Quadrotor UAV Testbed. *J. Phys.: Conf. Ser.* **2014**, *570*, 082003, doi:10.1088/1742-6596/570/8/082003.
23. Shen, Z.; Ma, Y.; Tsuchiya, T. Four-Dimensional Gait Surfaces for a Tilt-Rotor—Two Color Map Theorem. *Drones* **2022**, *6*, 103, doi:10.3390/drones6050103.
24. Shen, Z.; Tsuchiya, T. The Robust Gait of a Tilt-Rotor and Its Application to Tracking Control -- Application of Two Color Map Theorem 2022.
25. Hameduddin, I.; Bajodah, A.H. Nonlinear Generalised Dynamic Inversion for Aircraft Manoeuvring Control. *International Journal of Control* **2012**, *85*, 437–450, doi:10.1080/00207179.2012.656143.

# Chapter 9

# Relationship between The Structure and Dynamics (Tracking Control for a Tilt-rotor with Input Constraints by Robust Gaits)

**This Chapter has been published and presented on Tracking Control for a Tilt-rotor with Input Constraints by Robust Gaits, *IEEE Aerospace Conference* 2023, as the first author. The rest co-author(s) contribute nothing other than supervising and/or minor edition.**

**Abstract:** While the quantities of advanced control methods in the conventional quadrotor (with four inputs) surges to a plateau in the past two decades. Advanced with the lateral forces, controlling Ryll's tilt-rotor attracted increasing attentions recently. Ryll's tilt-rotor, mounted with the extra 4 motors to change the directions of the thrusts (with respect to its body-fixed frame), has eight inputs. The established control methods for stabilizing this vehicle are classified by the number of inputs determined by the controller; typical controllers calculate six or eight inputs to stabilize this vehicle. While these controllers can cause the over-intensive change in the tilting angles. The tilting angles can be expected to change in an over-wide range or over-rapidly, both of which are less expected in the application. With this concern, the controller in the most recent technique, augmented by the gait plan procedure, only determines four magnitudes of the thrusts. The four tilting angles are specified with time continuously during this gait plan procedure independently before the control procedure. It has been proved that if the gaits are planned on "Two Color Map" while following the relevant "Two Color Map Theorem", the feedback linearization-based controller in the latter of this cascade structure is robustified. Obviously, these admissible gaits on Two Color Map Theorem correspond to unique dynamics, showing different hoverabilities and trackable trajectories, which have not been elucidated yet. Despite the sound proof of the existence of the invertible decoupling matrix based on Two Color Map Theorem, the singular decoupling matrix can also be introduced by the activation of the saturation restrictions of the inputs. The activation of the non-zero constraints or maximum thrusts constraints can destroy the further application of the feedback linearization (dynamic inversion). This paper bridges the dynamics of the tilt-rotor to the robust gaits for the first time. Based on this, the relationship between the time-specified tilting angles and the trackable trajectories for the tilt-rotor is to be explored, concentrating on the gaits following Two Color Map Theorem. Then, the dynamics of the tilt-rotor adopting several typical gaits, satisfying Generalized Two Color Map Theorem, are analyzed by the exploration of the admissible (trackable) trajectories. The tracking simulations are developed to verify the relationship found. These simulations are carried out by Simulink, MATLAB.

## 1. Introduction

Ryll's tilt-rotor [1,2] is a novel type of quadrotor (Figure 1 [3]). This UAV has eight inputs; the four magnitudes of the thrusts as well as the four directions of the thrusts can be specified in completing the assigned tasks, e.g., reference-tracking [4,5].

A typical method for controlling this vehicle is feedback linearization [6–8], which inverse the dynamics to accommodate the subsequent linear controllers. This method, however, introduces the over-intensive control signals; the directions of the thrusts are expected to change in large scales and/or in high frequencies [1,2,4], which may not be expected in the application.

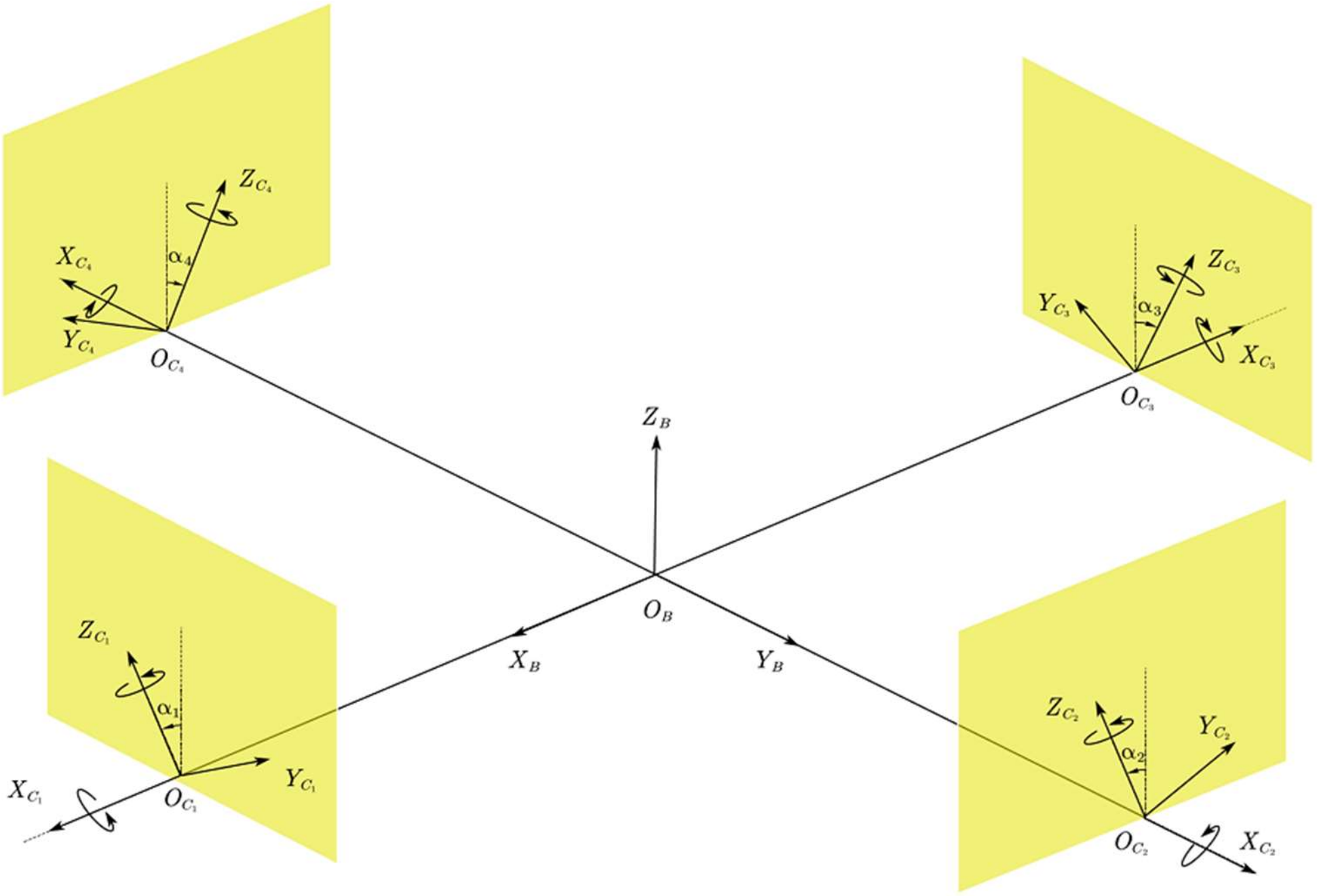

**Figure 1.** Ryll's tilt-rotor. The directions of the thrusts can be adjusted within the corresponding plane (yellow plane).

To avoid this unwanted changing speed in the tilting angles (directions of the thrusts), the research works [3,9,10] apply the feedback linearization on the four magnitudes of the thrusts only; the four tilting angles are not specified by the controller or the reference. Obviously, the tilting angles can be defined as a less aggressive time-specified function while completing the tracking control task. Our previous works [9–11] plan the tilting angles with the inspirations of animals (animal-inspired gait plan for the tilt-rotor).

A well-known obstacle to apply feedback linearization is the singularity of decoupling matrix [6]. An invertible decoupling matrix is typically required to receive infinite input signals. Unfortunately, this matrix is singular, given some specific combinations of attitude and tilting angles, hindering the application of this method. The robust gait plan method (Two Color Map Theorem) [12,13] is subsequently proposed. The gait planned following Two Color Map Theorem has a wider region of admissible attitude, which corresponding to invertible decoupling matrices.

Although some gaits following Two Color Map Theorem witnesses the success in reference tracking, the relationship between the robust gaits and the dynamics is not elucidated yet. Note that the parallel relationship in a quadrotor and hexacopter has been well established [14–16].

This paper explores the dynamical characters of the gaits obeying Two Color Map Theorem. The thrust-torque relationship is deduced for several special gaits in the different regions of Two Color Map. After that, we analyze the admissible acceleration of the tilt-rotor along X and Y directions for a gait obeying Two Color Map Theorem. In verifying our findings, several uniform circular references with a same radius and different accelerations are being tracked in a tilt-rotor simulator, written in Simulink, MATLAB.

The rest of this paper is organized as follows: Section 2 introduces the robust gaits and Two Color Map theorem. Section 3 gives the admissible region of torques and thrusts. The accessible accelerations are also given in this section. A tracking simulation is displayed in Section 4. Finally, Section 5 addresses the conclusions and discussions.

## 2. Robust Gaits of a Tiltrotor

This section briefs the control inputs and gaits of a tilt-rotor. As can be seen in Figure 1, the inputs of a tilt-rotor [17,18] are

$$\alpha_i, (i{=}1,2,3,4), \tag{1}$$

and

$$w_i, (i{=}1,2,3,4), \tag{2}$$

where $\alpha_i$ represents the tilting angles, whose positive direction is defined in Figure 1, $w_i = \omega_i \cdot |\omega_i|$, where $\omega_i$ is the angular velocity of the propeller, whose positive direction is also defined in Figure 1.

The generated thrust and torque [3] are specified as

$$\begin{bmatrix} F_X \\ F_Y \\ F_Z \end{bmatrix} = F_\alpha \cdot \begin{bmatrix} w_1 \\ w_2 \\ w_3 \\ w_4 \end{bmatrix}, \tag{3}$$

$$\begin{bmatrix} \tau_L \\ \tau_M \\ \tau_N \end{bmatrix} = \tau_\alpha \cdot \begin{bmatrix} w_1 \\ w_2 \\ w_3 \\ w_4 \end{bmatrix}, \tag{4}$$

where $F_X$, $F_Y$, $F_Z$ represent the thrusts along X-axis, Y-axis, Z-axis on the body-fixed frame, $\tau_L$, $\tau_M$, $\tau_N$ represent the torques along X-axis, Y-axis, Z-axis on the body-fixed frame, $F_\alpha$ and $\tau_\alpha$ are

$$F_\alpha = \begin{bmatrix} 0 & K_f \cdot s2 & 0 & -K_f \cdot s4 \\ K_f \cdot s1 & 0 & -K_f \cdot s3 & 0 \\ -K_f \cdot c1 & K_f \cdot c2 & -K_f \cdot c3 & K_f \cdot c4 \end{bmatrix}, \tag{5}$$

$$\tau_\alpha = \begin{bmatrix} 0 & L \cdot K_f \cdot c2 - K_m \cdot s2 & 0 & -L \cdot K_f \cdot c4 + K_m \cdot s4 \\ L \cdot K_f \cdot c1 + K_m \cdot s1 & 0 & -L \cdot K_f \cdot c3 - K_m \cdot s3 & 0 \\ L \cdot K_f \cdot s1 - K_m \cdot c1 & -L \cdot K_f \cdot s2 - K_m \cdot c2 & L \cdot K_f \cdot s3 - K_m \cdot c3 & -L \cdot K_f \cdot s4 - K_m \cdot c4 \end{bmatrix} \tag{6}$$

where $si = \sin\ (\alpha_i)$, $ci = \cos\ (\alpha_i)$, and $(i = 1,2,3,4)$. $K_f$ $(8.048 \times 10^{-6} N \cdot s^2/rad^2)$ is the coefficient of the thrust, $K_m$ $(2.423 \times 10^{-7} N \cdot m \cdot s^2/rad^2)$ is the coefficient of the drag.

In applying feedback linearization, the necessary and sufficient condition of receiving an invertible decoupling matrix with zero roll and pitch, $(\phi,\theta) = (0,0)$, is

$$R(\alpha_1,\alpha_2) \neq 0 \tag{7}$$

where

$$R(\alpha_1,\alpha_2) = 4.000 \cdot c1 \cdot c2 \cdot c1 \cdot c2 + 5.592 \cdot c1 \cdot c2 \cdot c1 \cdot s2 - 5.592 \cdot c1 \cdot c2 \cdot s1 \cdot c2 + 5.592 \cdot c1 \cdot s2 \cdot c1 \cdot c2 - 5.592 \cdot s1 \cdot c2 \cdot c1 \cdot c2 + 0.9716 \cdot c1 \cdot c2 \cdot s1 \cdot s2 + 0.9716 \cdot c1 \cdot s2 \cdot s1 \cdot c2 + 0.9716 \cdot s1 \cdot c2 \cdot c1 \cdot s2 + 0.9716 \cdot s1 \cdot s2 \cdot c1 \cdot c2 - 2.000 \cdot c1 \cdot s2 \cdot c1 \cdot s2 - 2.000 \cdot s1 \cdot c2 \cdot s1 \cdot c2 - 0.1687 \cdot c1 \cdot s2 \cdot s1 \cdot s2 + 0.1687 \cdot s1 \cdot c2 \cdot s1 \cdot s2 - 0.1687 \cdot s1 \cdot s2 \cdot c1 \cdot s2 + 0.1687 \cdot s1 \cdot s2 \cdot s1 \cdot c2. \tag{8}$$

Although the gait (a combination of $(\alpha_1,\alpha_2,\alpha_3,\alpha_4)$) satisfying Equation (8) results in an invertible decoupling matrix, this matrix at the attitude region near $(\phi,\theta) = (0,0)$ may be singular.

With this consideration, our previous work puts forward the concept of Two Color Map Theorem [12,13]. A gait lies on Two Color Map if it satisfies

$$\begin{cases}\alpha_3 = \alpha_1 \\ \alpha_4 = \alpha_2\end{cases}, \tag{9}$$

or

$$\begin{cases}\alpha_3 = -\alpha_1 + 0.334198 \\ \alpha_4 = -\alpha_2 - 0.334198\end{cases}. \tag{10}$$

A gait lying on Two Color Map is also called robust gait. The adjacent attitude region near $(\phi,\theta) = (0,0)$ also introduces an invertible decoupling matrix for a robust gait [12]. Some gaits lie on Two Color Map have been successfully deployed to complete the tracking control problems [13]. However, the dynamical property of the map, e.g., the relationship of the thrust and torque, has not been elucidated for the robust gaits, especially for the case with input constraints.

In this research, we analyze the properties of the thrusts, torques, and the admissible accelerations of the gaits satisfying Two Color Map Theorem considering the constraints in angular velocities of the propellers.

## 3. Admissible Torques, Thrusts, and Accelerations

### 3.1. Input Constraints

The angular velocities of the propellers are restricted by

$$\omega_{\min} \leqslant |\omega_i| \leqslant \omega_{\max}, (i=1,2,3,4) \tag{11}$$

where $\omega_{\min}$ and $\omega_{\max}$ represent the minimum angular velocity and maximum angular velocity of the propeller. Given that all the propeller rotates at $\omega_{\min}$ and that all the tilting angles are zero, the tilt-rotor generates the total vertical thrust slightly lower (3 N) than the gravity (around 4.2 N). Given that all the propeller rotates at $\omega_{\max}$ and that all the tilting angles are zero, the tilt-rotor generates the total vertical thrust slightly larger (5 N) than the gravity (around 4.2 N).

This section analyzes the relationship of vertical thrust ($F_Z$), torque along X-axis ($\tau_L$), and torque along Y-axis ($\tau_M$) under the input constraints. Also, the accelerations along X-axis and Y-axis when $F_Z = 4.2\ N$ are visualized.

### 3.2. Torques and Vertical Thrusts

Eight gaits are analyzed in this research. Four of them, Gait 1 – 4 in Equation (12) – (15), lie on the Two Color Map satisfying Equation (9).

$$Gait\ 1: (\alpha_1,\alpha_2,\alpha_3,\alpha_4) = (-0.5,-0.5,-0.5,-0.5) \tag{12}$$

$$Gait\ 2: (\alpha_1,\alpha_2,\alpha_3,\alpha_4) = (-0.5,0.5,-0.5,0.5) \tag{13}$$

$$Gait\ 3: (\alpha_1,\alpha_2,\alpha_3,\alpha_4) = (0.5,-0.5,0.5,-0.5) \tag{14}$$

$$Gait\ 4: (\alpha_1,\alpha_2,\alpha_3,\alpha_4) = (0.5,0.5,0.5,0.5) \tag{15}$$

The rest four gaits, Gait 5 – 8 in Equation (16) – (19), lie on the other surface of Two Color Map satisfying Equation (10).

$$Gait\ 5: (\alpha_1,\alpha_2,\alpha_3,\alpha_4) = (-0.5,-0.5,0.834,0.166) \tag{16}$$

$$Gait\ 6: (\alpha_1,\alpha_2,\alpha_3,\alpha_4) = (-0.5,0.5,0.834,-0.834) \tag{17}$$

$$Gait\ 7: (\alpha_1,\alpha_2,\alpha_3,\alpha_4) = (0.5,-0.5,-0.166,0.166) \tag{18}$$

$$Gait\ 8: (\alpha_1,\alpha_2,\alpha_3,\alpha_4) = (0.5,0.5,-0.166,-0.834) \tag{19}$$

Further, we make the following restriction [14] on the torque along Z-axis ($\tau_N$):

$$\tau_N = 0. \tag{20}$$

With the constraints in Equation (11) and (20), Gait 1 and Gait 4 fail to receive a solution, indicating that the adjacent region on Two Color Map can be unreasonable for hovering task. The rest two gaits (Gait 2 and Gait 3) receive the torque-vertical thrust relationship maps in Figure 2 and Figure 3.

We only display several slides (purple slides) of thrusts, where the relationship between the torques and thrusts are illustrated.

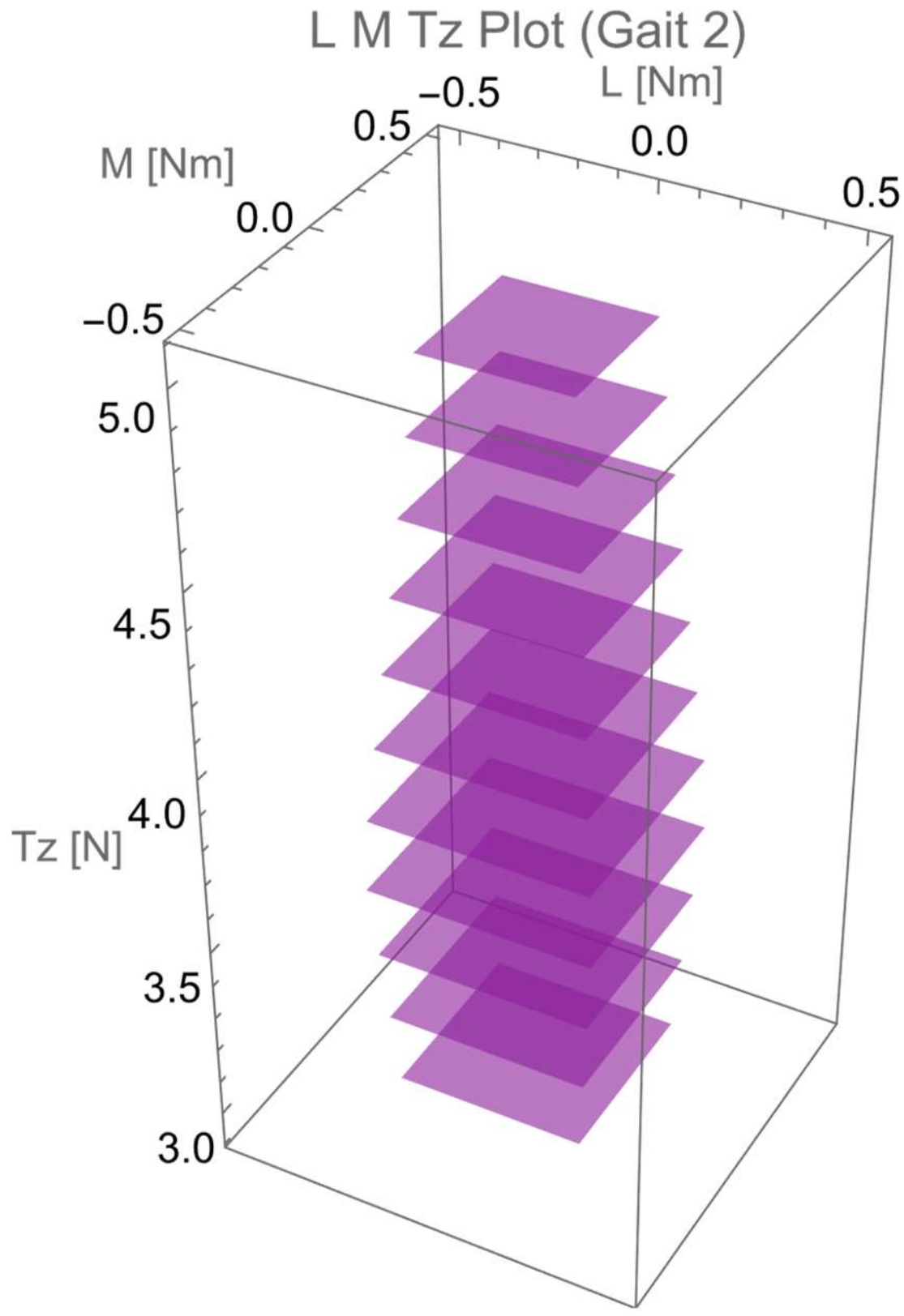

**Figure 2.** The torques and vertical thrusts of Gait 2. L, M, and Tz represent the torque along X-axis ($\boldsymbol{\tau_L}$), the torque along Y-axis ($\boldsymbol{\tau_M}$), and the vertical thrust. We display several slides of the vertical thrust.

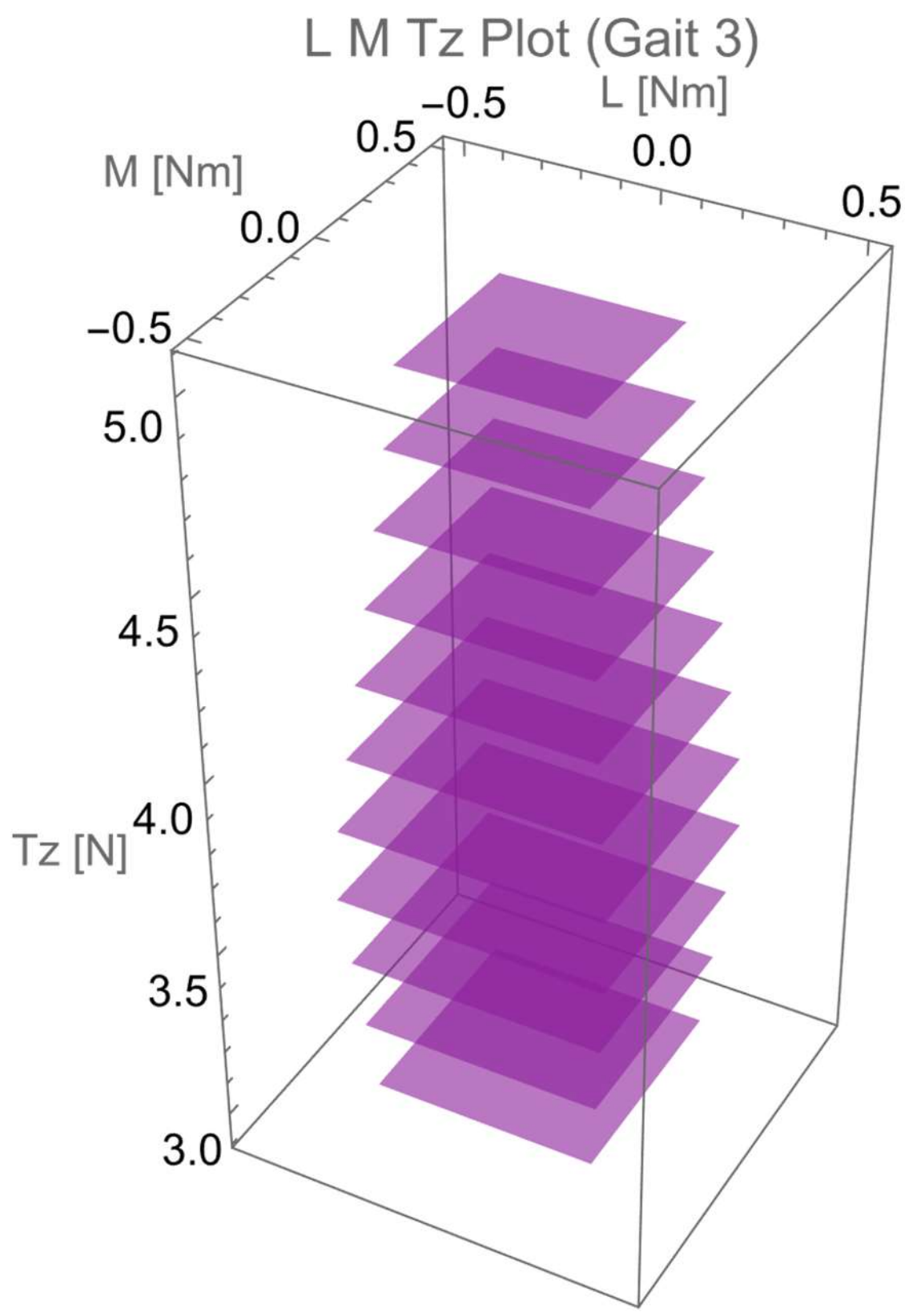

**Figure 3.** The torques and vertical thrusts of Gait 3. L, M, and Tz represent the torque along X-axis ($\boldsymbol{\tau_L}$), the torque along Y-axis ($\boldsymbol{\tau_M}$), and the vertical thrust. We display several slides of the vertical thrust.

The similar shapes of torques and thrusts in Figure 2 and 3 also appear in the conventional hexacopter [14,19].

Gait 5 – 8 satisfying the constraints in Equation (11) and (20) receive valid solutions. The results are demonstrated in Figure 4 – 7.

Similarly, we only display several slides of thrusts, where the relationship between the torques and thrusts are illustrated.

Notice that the relationship between torques for a given thrust is linearly coupled for these gaits rather than a plane. The selections of torques are narrowed comparing with the results from the gaits defined by Equation (9) (Figure 2 and 3).

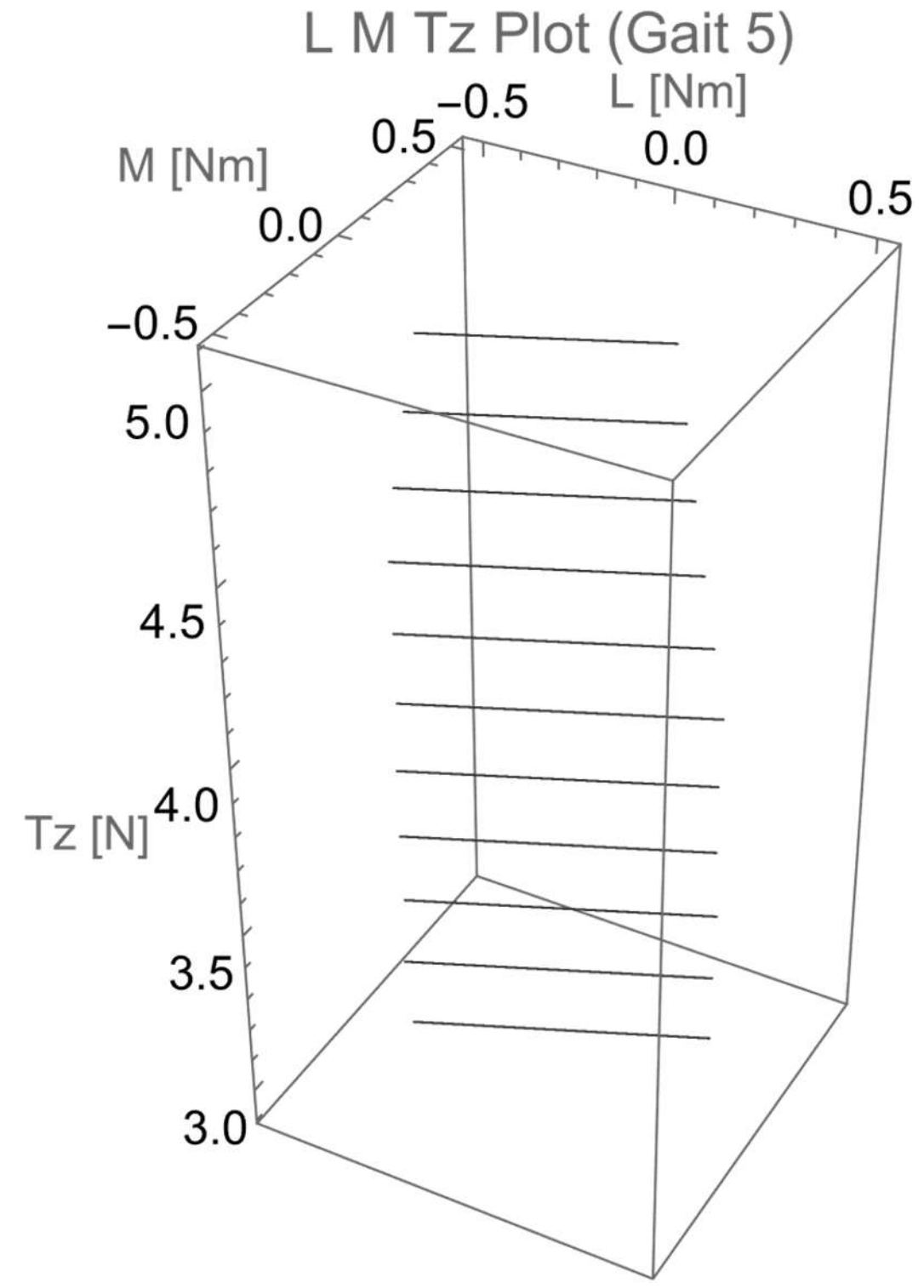

**Figure 4.** The torques and vertical thrusts of Gait 5. L, M, and Tz represent the torque along X-axis ($\boldsymbol{\tau_L}$), the torque along Y-axis ($\boldsymbol{\tau_M}$), and the vertical thrust. We display several slides of the vertical thrust.

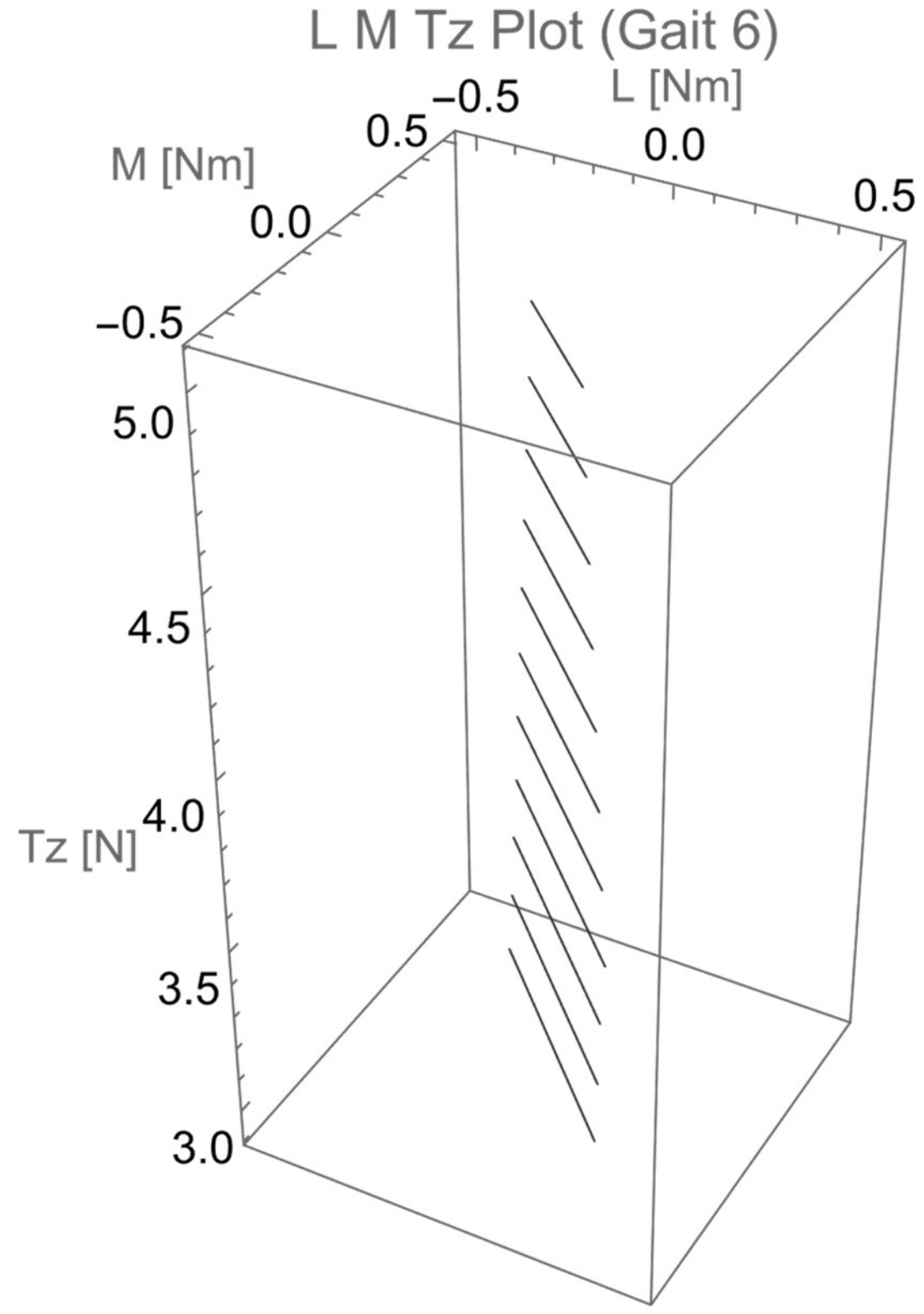

**Figure 5.** The torques and vertical thrusts of Gait 6. L, M, and Tz represent the torque along X-axis ($\boldsymbol{\tau_L}$), the torque along Y-axis ($\boldsymbol{\tau_M}$), and the vertical thrust. We display several slides of the vertical thrust.

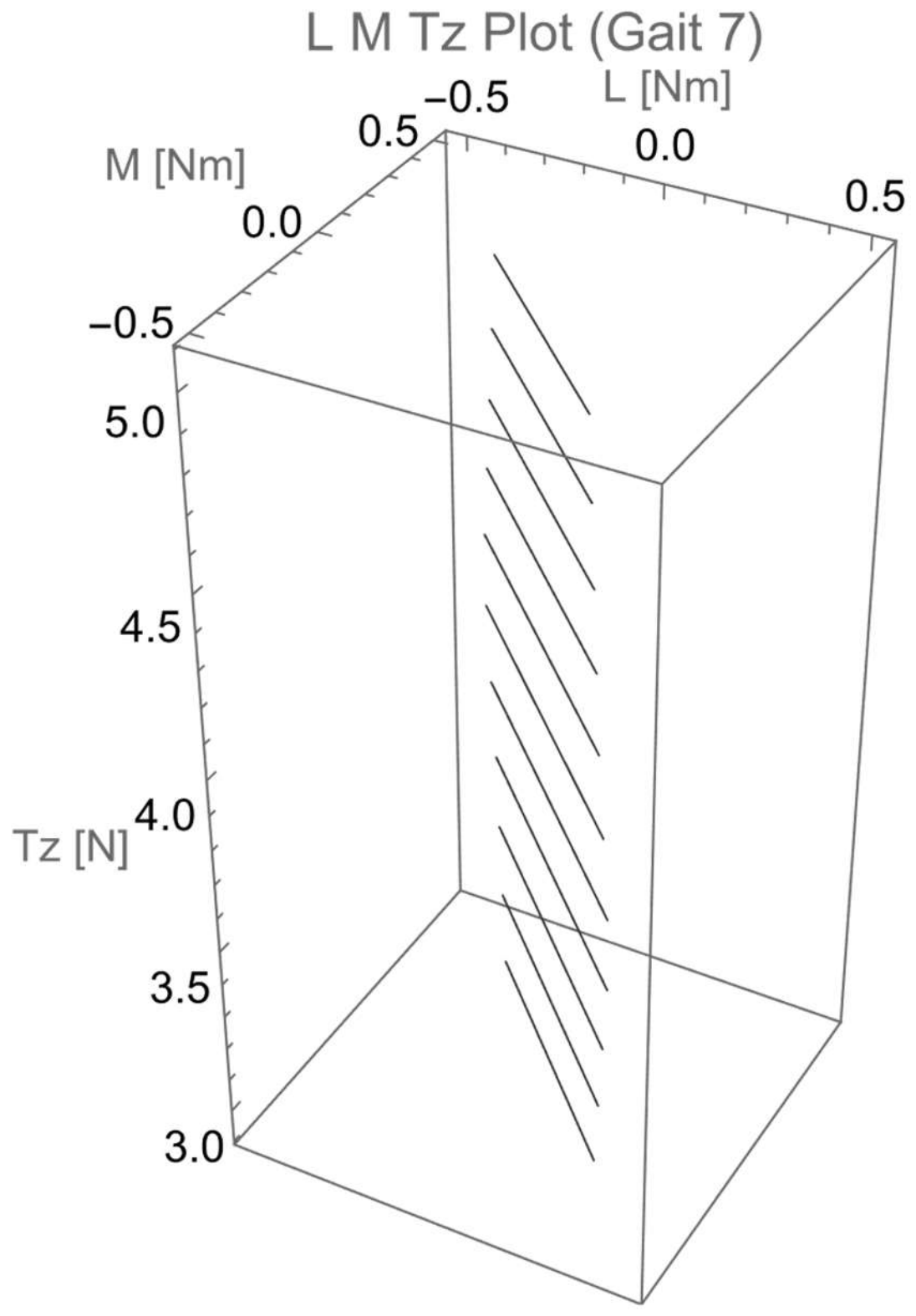

**Figure 6.** The torques and vertical thrusts of Gait 7. L, M, and Tz represent the torque along X-axis ($\boldsymbol{\tau_L}$), the torque along Y-axis ($\boldsymbol{\tau_M}$), and the vertical thrust. We display several slides of the vertical thrust.

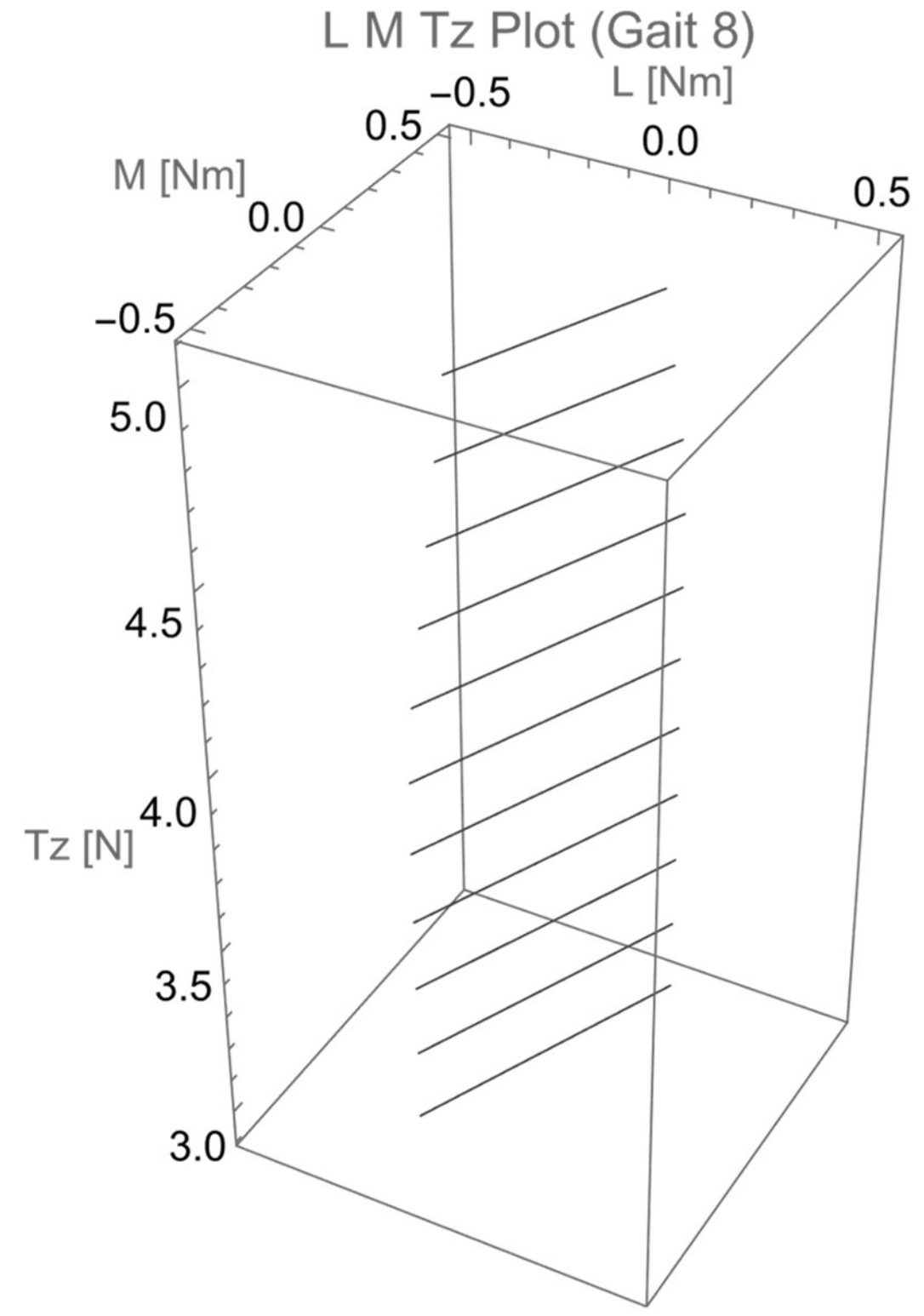

**Figure 7**. The torques and vertical thrusts of Gait 8. L, M, and Tz represent the torque along X-axis ($\boldsymbol{\tau_L}$), the torque along Y-axis ($\boldsymbol{\tau_M}$), and the vertical thrust. We display several slides of the vertical thrust.

*3.3. Admissible Accelerations in The Body-fixed Frame*

To investigate the trajectory-tracking properties of the tilt-rotor, the admissible accelerations are analyzed.

In this research, we only consider the case where the gravity is exactly compensated by the vertical thrust:

$$F_Z = 4.2\ N \tag{21}$$

The relevant torque can be found in one layer of the corresponding torque and thrust plots, e.g., Figure 2 – 7.

Further, we plot the admissible accelerations of the tilt-rotor with Gait 2 and Gait 3 in Figure 8 and 9.

It can be found that both gaits (Gait 2 and 3) receive the similar admissible accelerations ranging from $-2m/s^2$ to $2m/s^2$ along either axis. Note that these accelerations are with respect to the body-fixed frame and are resulted from Equation (3).

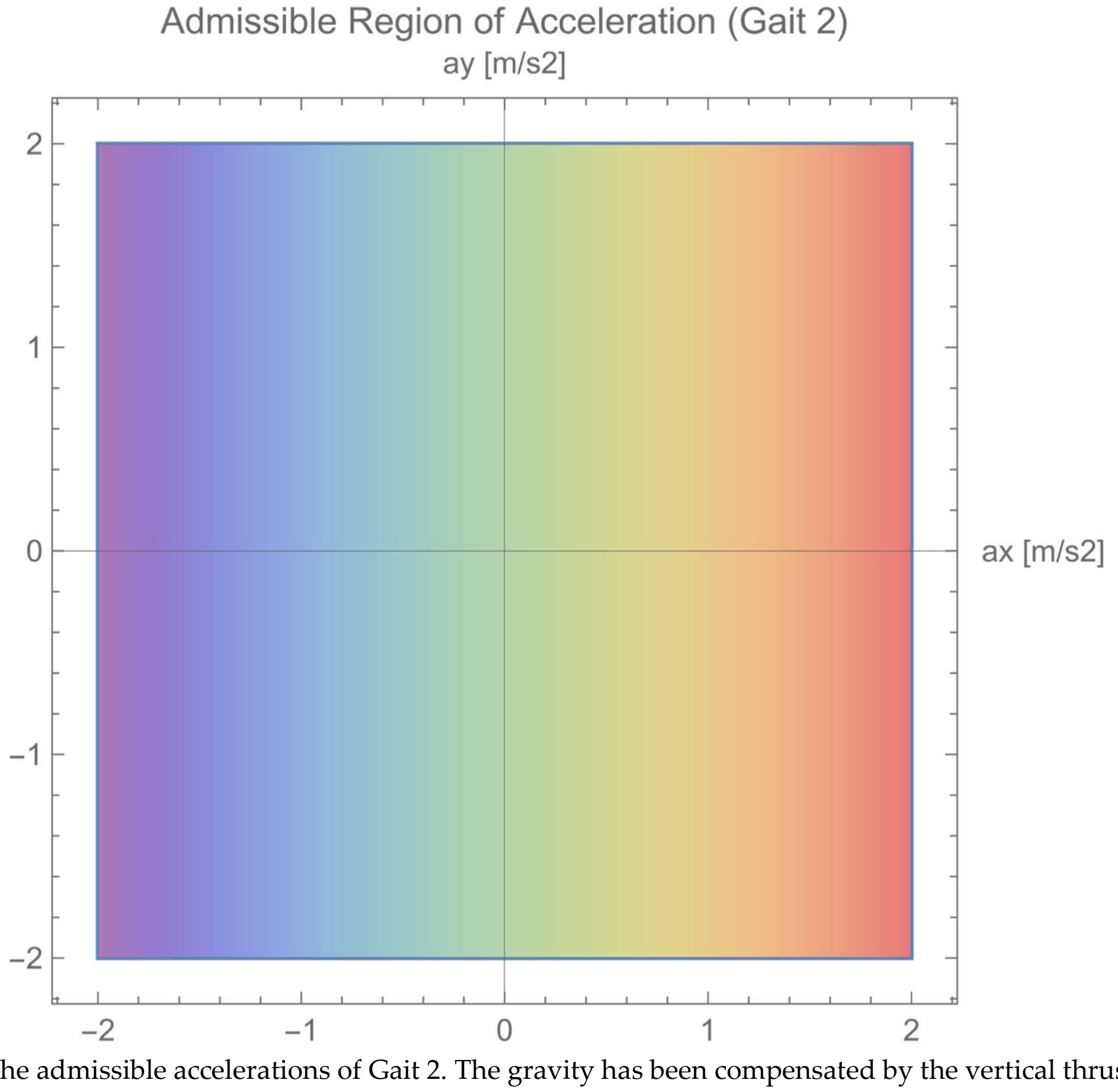

**Figure 8.** The admissible accelerations of Gait 2. The gravity has been compensated by the vertical thrust.

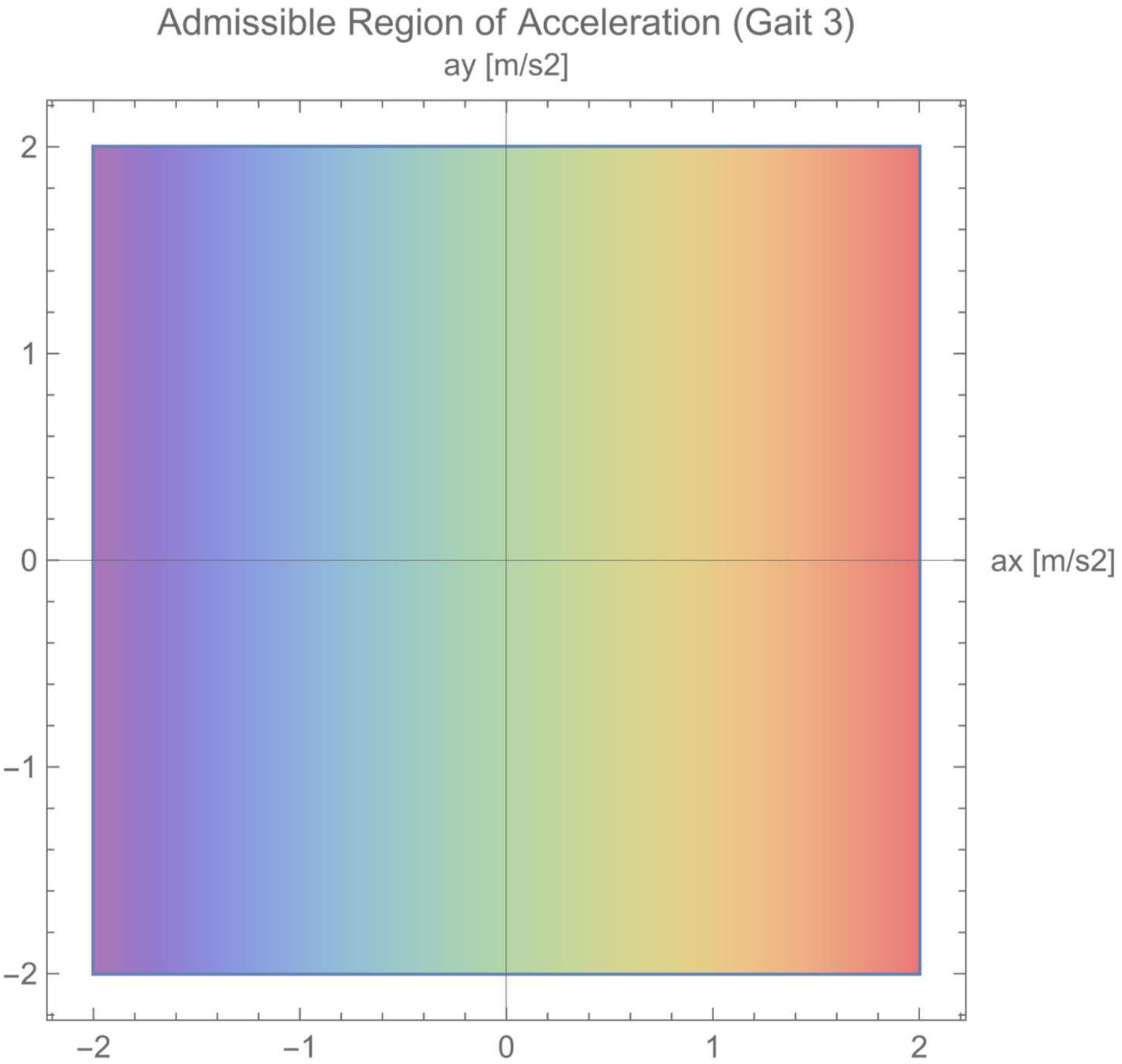

**Figure 9.** The admissible accelerations of Gait 3. The gravity has been compensated by the vertical thrust.

## 4. Tracking Simulation

### *4.1. Uniform Circular Reference*

To investigate the behaviors of the tilt-rotor while tracking the references with different accelerations, uniform circular references with identical radius and with different accelerations are designed as below:

$$Radius: R = 5\ meters \tag{22}$$

$$Angular\ Frequency: \Omega = 0.1, 0.5, 1, 1.5\ rad/s \tag{23}$$

The magnitudes of the corresponding accelerations of the references are $0.05, 1.25, 5, 11.25\ m/s^2$.

The tilt-rotor equipped with the feedback linearization and PD controller. For the detail of the controller, please refer to our previous work [9,10].

In this research the gait applied to the tilt-rotor is Gait 2. The initial position of the tilt-rotor is at the center of the circle. It is supposed to track these references while maintaining the height.

The simulation is conducted in a tilt-rotor simulator written in Simulink, MATLAB.

### *4.2. Tracking Result*

Figure 10 shows the tracking result in position. In general, all these four references result in periodical tracking trajectories.

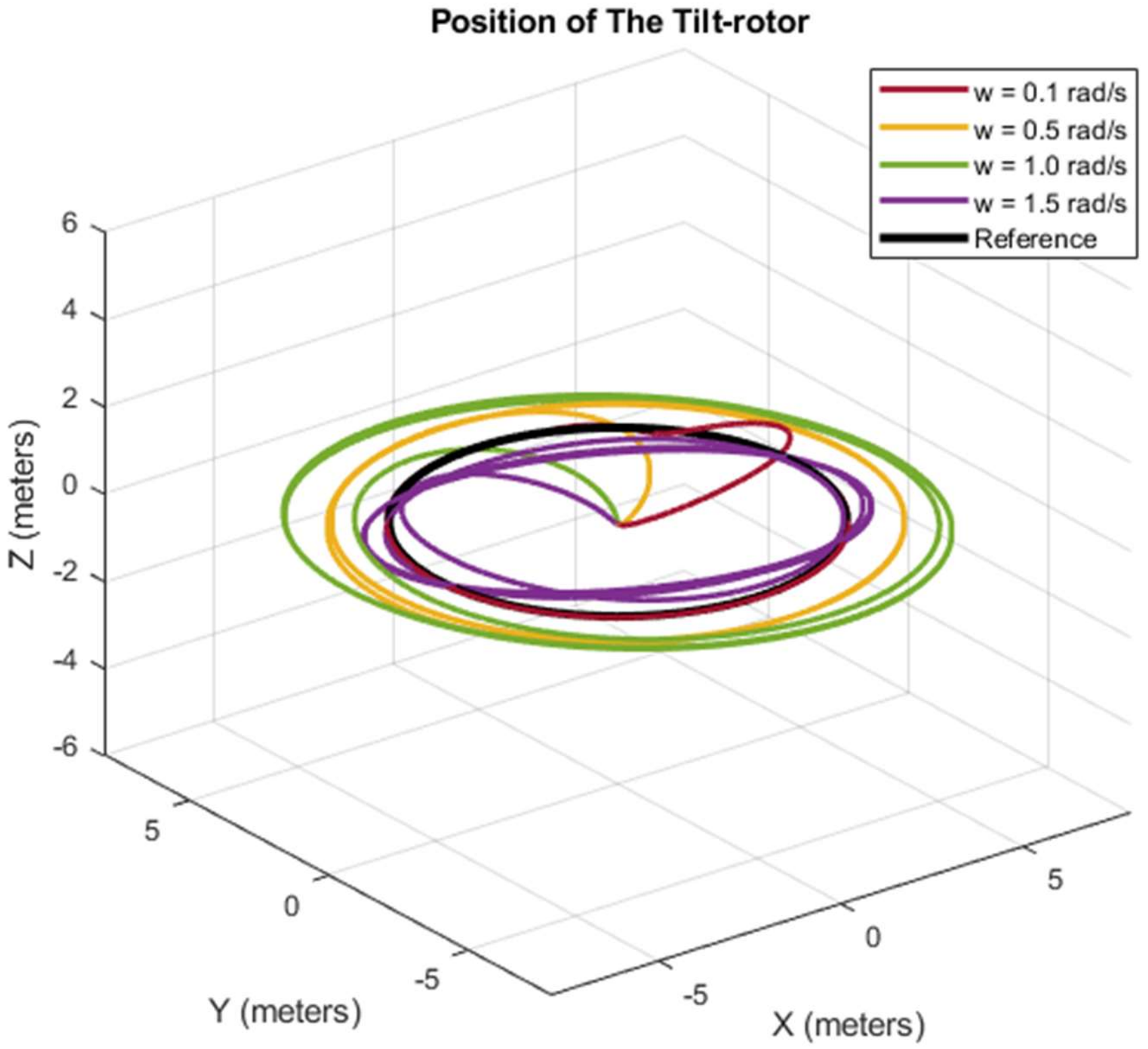

**Figure 10.** The tracking result of the tilt-rotor. The references (black circle) are uniform circles with directions and velocities. The red curve represents the position history while tracking the reference with $\Omega = 0.1\,rad/s$. The yellow curve represents the position history while tracking the reference with $\Omega = 0.5\,rad/s$. The green curve represents the position history while tracking the reference with $\Omega = 1.0\,rad/s$. The yellow curve represents the position history while tracking the reference with $\Omega = 1.5\,rad/s$.

With a reference with an acceleration much smaller than $2m/s^2$, e.g., red curve, the tilt-rotor tracks the reference with little state error.

With the references with the accelerations close to $2\ m/s^2$, e.g., yellow and green curves, the tilt-rotor tracks the references in a larger circular orbit with constant state error.

With the references with the accelerations much larger than $2m/s^2$, e.g., purple curves, the tilt-rotor tracks the references in an elliptical orbit with varying state error (periodic).

As for the references with even higher accelerations, e.g., $\Omega = 2.0\,rad/s$, the tilt-rotor fails to track them.

## 5. Conclusions and Discussions

This paper delivers the analysis of the dynamic properties of the robust gaits of the tilt-rotor. Several regions of the Two Color Map have been explored and found feasible in the trajectory-tracking tasks.

However, we show that some gaits on Two Color Map fail to maintain a zero torque along yaw with the constraints of inputs (constraints on angular velocities of the propeller).

The admissible region of the accelerations while compensating the gravity has been explored. The result demonstrates a rectangular region of these accelerations, satisfying the input constraints.

Uniform circular references with identical radius and different accelerations are designed for tracking. A reference with smaller accelerations results a better tracking result, e.g., less state error. The tilt-rotor fails to track an aggressive reference with an acceleration over-exceeding the acceleration constraints.

One might wonder why the tilt-rotor still tracks the references with the accelerations larger than the theoretical maximum of the accelerations. This is because that the analyzed acceleration constraints are with respect to the body-fixed frame. The actual effective accelerations to drive the tilt-rotor while tracking are with respect to the inertial frame. This means that a non-zero roll and/or pitch may influence this limit, broadening the admissible accelerations with respect to the inertial frame.

Another point worth mentioning is that the relationship between the torque and thrust for the gaits satisfying Equation (10) seems decreases the number of inputs of the tilt-rotor. For example, any relationship between $\tau_L$ and $\tau_M$ in Figure 4 –7 is linear for a given thrust. One may believe that there can be singularities in the decoupling matrix. However, this linear relationship strictly bases on the assumptions that thrust is constant and $\tau_N$ is zero. In fact, the decoupling matrix on Two Color Map has been proved invertible [12].

One of the further works can be the analysis of the cases where $\tau_N$ is varying. Also, more complicated references can be explored.

## References

1. Ryll, M.; Bulthoff, H.H.; Giordano, P.R. Modeling and Control of a Quadrotor UAV with Tilting Propellers. In Proceedings of the 2012 IEEE International Conference on Robotics and Automation; IEEE: St Paul, MN, USA, May 2012; pp. 4606–4613.
2. Ryll, M.; Bulthoff, H.H.; Giordano, P.R. A Novel Overactuated Quadrotor Unmanned Aerial Vehicle: Modeling, Control, and Experimental Validation. *IEEE Trans. Contr. Syst. Technol.* **2015**, *23*, 540–556, doi:10.1109/TCST.2014.2330999.
3. Shen, Z.; Tsuchiya, T. Gait Analysis for a Tiltrotor: The Dynamic Invertible Gait. *Robotics* **2022**, *11*, 33, doi:10.3390/robotics11020033.
4. Kumar, R.; Nemati, A.; Kumar, M.; Sharma, R.; Cohen, K.; Cazaurang, F. Tilting-Rotor Quadcopter for Aggressive Flight Maneuvers Using Differential Flatness Based Flight Controller.; American Society of Mechanical Engineers: Tysons, Virginia, USA, October 11 2017; p. V003T39A006.
5. Mellinger, D.; Kumar, V. Minimum Snap Trajectory Generation and Control for Quadrotors. In Proceedings of the 2011 IEEE International Conference on Robotics and Automation; IEEE: Shanghai, China, May 2011; pp. 2520–2525.
6. Shen, Z.; Tsuchiya, T. Singular Zone in Quadrotor Yaw–Position Feedback Linearization. *Drones* **2022**, *6*, 20, doi:doi.org/10.3390/drones6040084.
7. Ryll, M.; Muscio, G.; Pierri, F.; Cataldi, E.; Antonelli, G.; Caccavale, F.; Franchi, A. 6D Physical Interaction with a Fully Actuated Aerial Robot. In Proceedings of the 2017 IEEE International Conference on Robotics and Automation (ICRA); May 2017; pp. 5190–5195.
8. Martins, L.; Cardeira, C.; Oliveira, P. Feedback Linearization with Zero Dynamics Stabilization for Quadrotor Control. *J Intell Robot Syst* **2021**, *101*, 7, doi:10.1007/s10846-020-01265-2.
9. Shen, Z.; Tsuchiya, T. Cat-Inspired Gaits for a Tilt-Rotor—From Symmetrical to Asymmetrical. *Robotics* **2022**, *11*, 60, doi:10.3390/robotics11030060.
10. Shen, Z.; Ma, Y.; Tsuchiya, T. Feedback Linearization-Based Tracking Control of a Tilt-Rotor with Cat-Trot Gait Plan. *International Journal of Advanced Robotic Systems* **2022**, *19*, 17298806221109360, doi:10.1177/17298806221109360.
11. Shen, Z.; Ma, Y.; Tsuchiya, T. Stability Analysis of a Feedback-Linearization-Based Controller with Saturation: A Tilt Vehicle with the Penguin-Inspired Gait Plan. *arXiv preprint arXiv:2111.14456* **2021**.
12. Shen, Z.; Ma, Y.; Tsuchiya, T. Four-Dimensional Gait Surfaces for a Tilt-Rotor—Two Color Map Theorem. *Drones* **2022**, *6*, 103, doi:10.3390/drones6050103.
13. Shen, Z.; Ma, Y.; Tsuchiya, T. Generalized Two Color Map Theorem -- Complete Theorem of Robust Gait Plan for a Tilt-Rotor 2022.
14. Ducard, G.J.J.; Hua, M.-D. Discussion and Practical Aspects on Control Allocation for a Multi-Rotor Helicopter. *Int. Arch. Photogramm. Remote Sens. Spatial Inf. Sci.* **2012**, *XXXVIII-1/C22*, 95–100, doi:10.5194/isprsarchives-XXXVIII-1-C22-95-2011.
15. Horla, D.; Hamandi, M.; Giernacki, W.; Franchi, A. Optimal Tuning of the Lateral-Dynamics Parameters for Aerial Vehicles With Bounded Lateral Force. *IEEE Robot. Autom. Lett.* **2021**, *6*, 3949–3955, doi:10.1109/LRA.2021.3067229.
16. Mochida, S.; Matsuda, R.; Ibuki, T.; Sampei, M. A Geometric Method of Hoverability Analysis for Multirotor UAVs With Upward-Oriented Rotors. *IEEE Transactions on Robotics* **2021**, *37*, 1765–1779, doi:10.1109/TRO.2021.3064101.
17. Franchi, A.; Carli, R.; Bicego, D.; Ryll, M. Full-Pose Tracking Control for Aerial Robotic Systems With Laterally Bounded Input Force. *IEEE Trans. Robot.* **2018**, *34*, 534–541, doi:10.1109/TRO.2017.2786734.

18. Michieletto, G.; Ryll, M.; Franchi, A. Fundamental Actuation Properties of Multirotors: Force–Moment Decoupling and Fail–Safe Robustness. *IEEE Trans. Robot.* **2018**, *34*, 702–715, doi:10.1109/TRO.2018.2821155.
19. Grupo de Procesamiento de Señales, Identificación y Control (GPSIC), Departamento de Ingeniería Electrónica, Facultad de Ingeniería Universidad de Buenos Aires (FIUBA), Argentina; Giribet, J.I.; Pose, C.D.; Ghersin, A.S.; Mas, I. Experimental Validation of a Fault Tolerant Hexacopter with Tilted Rotors. *IJEETC* **2018**, 58–65, doi:10.18178/ijeetc.7.2.58-65.

# Chapter 10

# A Fictional Mobile Robot and Its Feedback-linearization Controller (State Drift and Gait Plan in Feedback Linearization Control of a Tilt Vehicle)

**This Chapter has been published and presented on State Drift and Gait Plan in Feedback Linearization Control of a Tilt Vehicle. In Proceedings of the Computer Science & Information Technology (CS & IT); Academy & Industry Research Collaboration Center (AIRCC): Vienna, Austria, March 19 2022; Vol. 12, pp. 1–17, as the first author. The rest co-author(s) contribute nothing other than supervising and/or minor edition.**

**Abstract:** To stabilize a conventional quadrotor, simplified equivalent vehicles (e.g., autonomous car) are developed to test the designed controller. Based on that, various controllers based on feedback linearization have been developed. With the recently developed concept of tilt-rotor, there lacks the simplified/equivalent model, however. Indeed, the tilt structure is relatively unusual in vehicles. In this research, we put forward a unique fictional vehicle with tilt structure, which is to help evaluate the property of the tilt-structure-aimed controllers. One phenomenon (state drift) in controlling an over-actuated tilt structure by feedback linearization is presented subsequently. State drift can be easily neglected and is not paid attention to in the current researches in tilt-rotor controllers' design so far. We report this phenomenon and provide a potential approach to avoid this behavior.

## 1. Introduction

Various controllers are analyzed to control the conventional quadrotors. Some of them are linear. [5] compares two linear controllers in stabilizing the attitude of a quadrotor model. Others are nonlinear. [6,7] apply geometric controller, which delicately defines the attitude error and provides the stability proof. [8] puts forward a sliding mode controller; it guarantees the stability using Lyapunov criteria. Backstepping controller with cascade structure can be found in [9].

Both linear and nonlinear controllers have their advantages and disadvantages. These pros and cons include the controlling time, stability region, dynamic bias, robustness, calculating cost, etc. Different from solely relying on linear or nonlinear controllers, feedback linearization [2,3] transfers a nonlinear system to a linear system. The linear controller can be subsequently applied to the generated linear system.

Unfortunately, the quadrotor is an under-actuated system where the number of inputs is less than the degrees of freedom. This nature prohibits us from building a full-state controller; the number of the states to be controlled directly is 4 at most, equal to the number of the inputs. Several different selections on these states can be found in [2,10,11]. Another problem hindering us from utilizing feedback linearization is the invertibility of the decouple matrix [2,12]. Some selections of the output combination introduce the singular decouple matrix. [12] details this problem and provides the potential approach to avoid hitting the singular zone.

The recent popular novel quadrotor (tilt-rotor) augments the number of inputs to 8 [4,13,14]; the quadrotor not only changes the magnitude of each thrust but also the direction. With this designation, the system becomes over-actuated. Subsequently, several researches focus on controlling this over-actuated system by feedback linearization [4], [15]-[17].

Controlling an over-actuated system by feedback linearization provides the possibility of full-state control. The input, however, might change even if the state has stabilized. This phenomenon is not paid attention to yet.

In this paper, we address the feedback-linearization-based controllers in a simplified tilt vehicle. Controlling a simplified vehicle is a common approach before applying this control method to the complicated system (e.g., UAV) [1]. We present the result with the state drift phenomenon. To avoid this unwanted drift, we decrease the number of the real inputs by gait plan, which is a gait schedule technique in quadruped robots [18].

This paper is structured as follows: Section 2 introduces the dynamics of the fictional vehicle. The feedback-linearization-based controller is designed in Section 3. Section 4 simulates the controller designed and presents the state drift phenomenon. Section 5 proposes gait plan and puts it into the controller design. Section 6 concludes and makes the discussion of the result.

## 2. Dynamics of The Tilt Vehicle

In this section, we introduce a two-dimensional fictional tilt vehicle (Figure 1). The motion of the vehicle is restricted to a level plane (two-dimensional movement). Two propellers are fixed on the vehicle with the angle 2θ between. The value of 2θ is $\frac{\pi}{3}$.

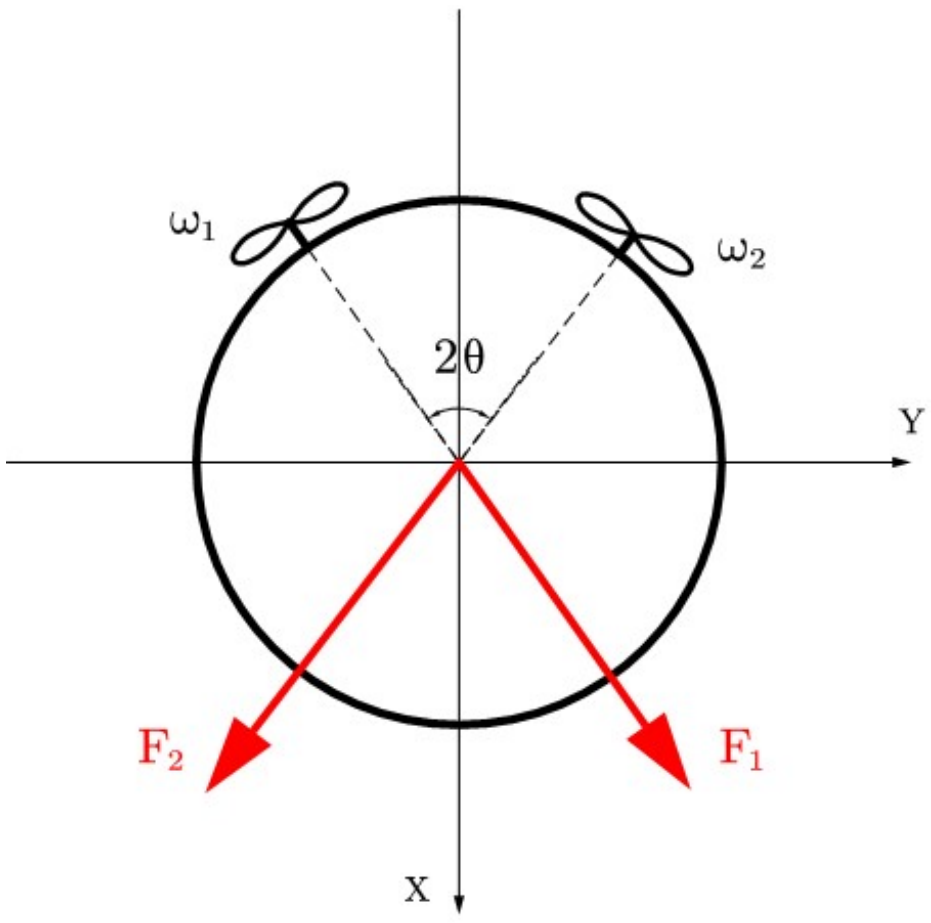

**Figure 1.** Two-dimensional Tilt Vehicle

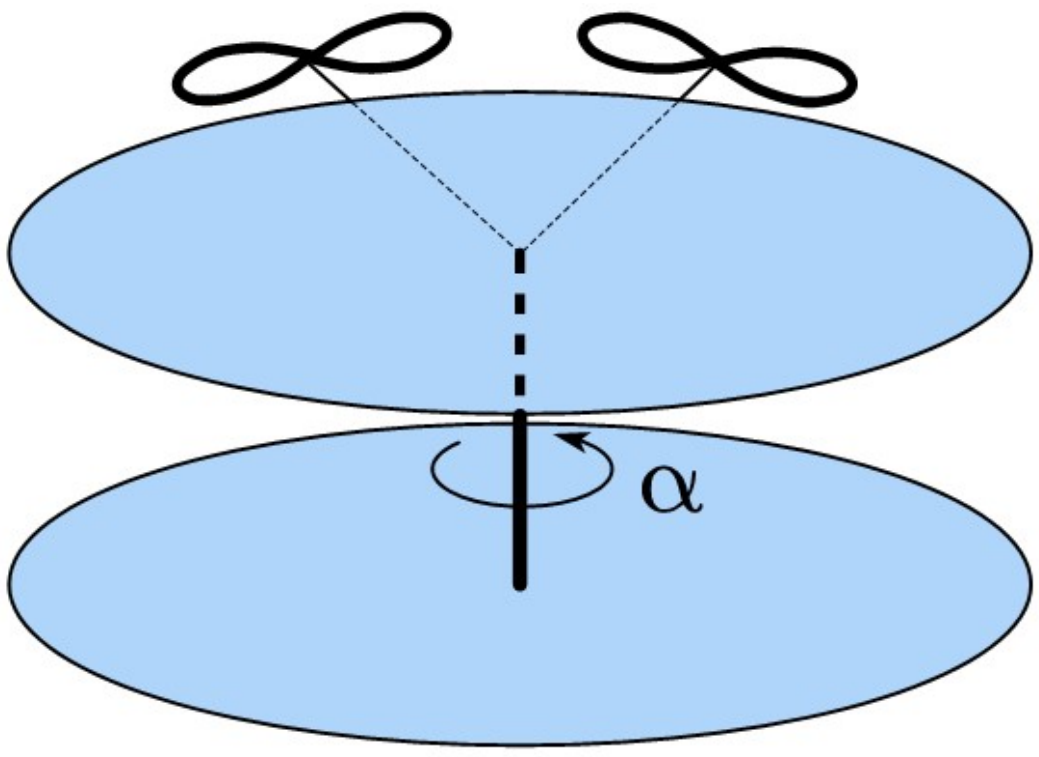

**Figure 2.** The Structure of the tilt vehicle

Each propeller can generate the thrust by rotating with angular velocity, $\omega_1$ and $\omega_2$. The trust generated follows Equation (1).

$$F_i = K_{F_i} \cdot {\omega_i}^2, i = 1{,}2(\omega_i \geqslant 0) \tag{1}$$

where $K_{F_i}$ is the thrust coefficient. Its value is $K_{F_i} = 0.001$. This proportional relationship in adopted in the dynamics of UAV control [1-3]. The thrust is always nonnegative. Also, the angular velocity is always nonnegative.

What makes this vehicle special is the tilt structure, as plotted in Figure 2. The propellers mentioned are fixed on the top disc. While the bottom disc contacts the $x - y$ plane directly. The frictional force from the $x - y$ plane is ignored. This structure provides the possibility of changing the directions of the thrusts by tilting the top disc (Figure 3—4) with an angle, $\alpha$.

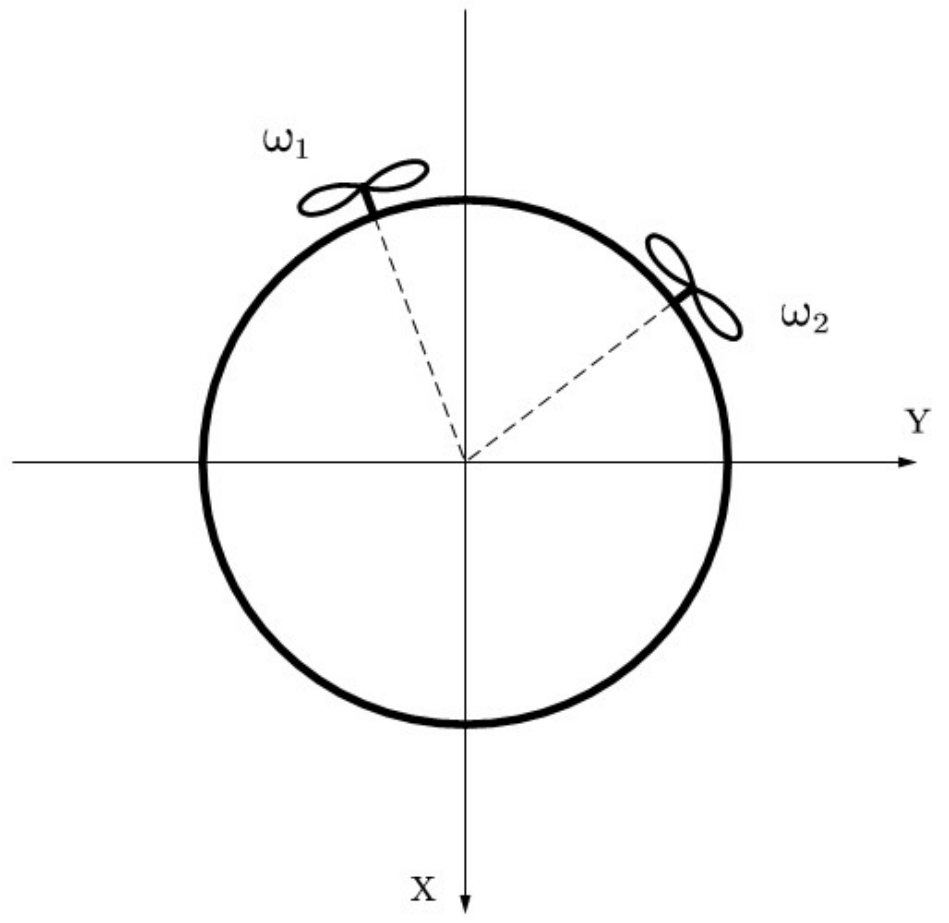

**Figure 3.** Tilt to right example

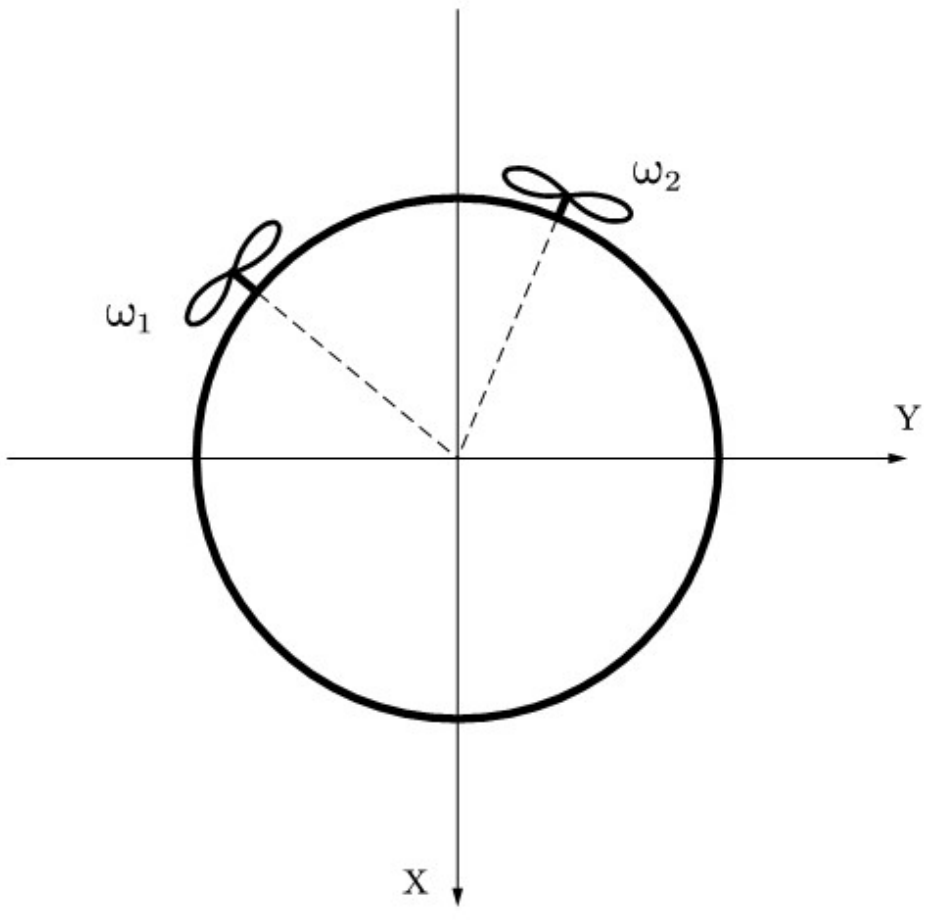

**Figure 4.** Tilt to left example

Based on Conservation of Angular Momentum, the relationship between the yaw, $\psi$, and the tilt angle, $\alpha$, is written in Equation (2).

$$\psi = -\frac{I_T}{I_B} \cdot \alpha \tag{2}$$

where $I_T$ and $I_B$ are the rotational inertias of the top disc and bottom disc, respectively. Their values are $I_T = 0.001$, $I_B = 0.002$.

Based on the Newtown's law, we write the position in Equation (3).

$$\begin{bmatrix} \ddot{x} \\ \ddot{y} \end{bmatrix} = \frac{1}{m} \cdot J_{\psi+\alpha} \cdot J_\theta \cdot \begin{bmatrix} {\omega_1}^2 \\ {\omega_2}^2 \end{bmatrix} \tag{3}$$

where $m$ is the mass of the entire vehicle. $J_{\psi+\alpha}$ and $J_\theta$ are defined in Equation (4)—(5).

$$J_{\psi+\alpha} = \begin{bmatrix} \cos(\psi+\alpha) & -\sin(\psi+\alpha) \\ \sin(\psi+\alpha) & \cos(\psi+\alpha) \end{bmatrix} \tag{4}$$

$$J_\theta = \begin{bmatrix} \cos(\theta) & 0 \\ 0 & \sin(\theta) \end{bmatrix} \cdot \begin{bmatrix} K_{F_1} & K_{F_2} \\ K_{F_1} & -K_{F_2} \end{bmatrix} \tag{5}$$

It is worth mentioning that both $J_{\psi+\alpha}$ and $J_\theta$ have full rank. Thus, $J_{\psi+\alpha} \cdot J_\theta$ is invertible.

$$|J_{\psi+\alpha} \cdot J_\theta| \neq 0 \tag{6}$$

So far, we have derived the dynamics of this vehicle in Equation (2)—(3). This is a MIMO system. There are three inputs, $\omega_1$, $\omega_2$, and $\alpha$. While there are two outputs, $x$, and $y$. So, it is an over-actuated system.

**3. Controller Design (Feedback Linearization)**

Since it is an over-actuated system, we may apply feedback linearization for this system while discarding some inputs to let the number of the outputs equal to the number of the inputs. Although Feedback Linearization is widely used the tilt-rotor control, one potential adverse effect (state drift) is easily neglected.

This section develops the feedback linearization controllers based on this over-actuated system model. The result and the state drift phenomenon are given in the next section.

*3.1. Third Derivative Feedback Linearization*

Since the input $\alpha$ is tangled in Equation (3), we calculate the higher derivative (third order here) of the position in Equation (7).

$$\begin{bmatrix} \dddot{x} \\ \dddot{y} \end{bmatrix} = \frac{1}{m} \cdot \dot{J}_{\psi+\alpha} \cdot J_\theta \cdot \begin{bmatrix} {\omega_1}^2 \\ {\omega_2}^2 \end{bmatrix} + \frac{1}{m} \cdot J_{\psi+\alpha} \cdot J_\theta \cdot \begin{bmatrix} 2\omega_1 \cdot \dot{\omega}_1 \\ 2\omega_2 \cdot \dot{\omega}_2 \end{bmatrix} \tag{7}$$

where $\dot{J}_{\psi+\alpha}$ is calculated in Equation (8).

$$\dot{J}_{\psi+\alpha} = \begin{bmatrix} -\sin(\psi+\alpha) & -\cos(\psi+\alpha) \\ \cos(\psi+\alpha) & -\sin(\psi+\alpha) \end{bmatrix} \cdot (\dot{\psi}+\dot{\alpha}) \tag{8}$$

Substituting Equation (2) into Equation (7) yields Equation (9).

$$\begin{bmatrix} \dddot{x} \\ \dddot{y} \end{bmatrix} = \left[ {\Delta_1}^{2\times2} \middle| {\Delta_2}^{2\times1} \right] \cdot \begin{bmatrix} \dot{\omega}_1 \\ \dot{\omega}_2 \\ \dot{\alpha} \end{bmatrix} \tag{9}$$

where ${\Delta_1}^{2\times2}$ and ${\Delta_2}^{2\times1}$ are defined in Equation (10)—(11).

$${\Delta_1}^{2\times2} = \frac{1}{m} \cdot J_{\psi+\alpha} \cdot J_\theta \cdot \begin{bmatrix} 2\omega_1 & 0 \\ 0 & 2\omega_2 \end{bmatrix} \tag{10}$$

$${\Delta_2}^{2\times1} = \frac{1}{m} \cdot \begin{bmatrix} -\sin(\psi+\alpha) & -\cos(\psi+\alpha) \\ \cos(\psi+\alpha) & -\sin(\psi+\alpha) \end{bmatrix} \cdot J_\theta \cdot \left(1 - \frac{I_T}{I_B}\right) \cdot \begin{bmatrix} {\omega_1}^2 \\ {\omega_2}^2 \end{bmatrix} \tag{11}$$

Since ${\Delta_1}^{2\times2}$ is invertible if and only if $\omega_1 \neq 0$ and $\omega_2 \neq 0$, $\left[{\Delta_1}^{2\times2} \middle| {\Delta_2}^{2\times1}\right]$ has full row rank if and only if $\omega_1 \neq 0$ and $\omega_2 \neq 0$. Thus, the pseudo-inverse of $\left[{\Delta_1}^{2\times2} \middle| {\Delta_2}^{2\times1}\right]$ has explicit form, Equation (12).

$$pinv(\Delta) = \Delta^T \cdot (\Delta \cdot \Delta^T)^{-1} \tag{12}$$

where $\Delta = \left[{\Delta_1}^{2\times2} \middle| {\Delta_2}^{2\times1}\right]$.

Separating the inputs in Equation (9), we receive Equation (13).

$$\begin{bmatrix} \dot{\omega}_1 \\ \dot{\omega}_2 \\ \dot{\alpha} \end{bmatrix} = pinv\left(\left[{\Delta_1}^{2\times2} \middle| {\Delta_2}^{2\times1}\right]\right) \cdot \begin{bmatrix} \dddot{x} \\ \dddot{y} \end{bmatrix} \tag{13}$$

Based on Equation (13), the controller is designed in Equation (14).

$$\begin{bmatrix} \dot{\omega}_1 \\ \dot{\omega}_2 \\ \dot{\alpha} \end{bmatrix} = pinv\left(\left[{\Delta_1}^{2\times2} \middle| {\Delta_2}^{2\times1}\right]\right) \cdot \begin{bmatrix} \dddot{x}_r + k_{x_1} \cdot (\ddot{x}_r - \ddot{x}) + k_{x_2} \cdot (\dot{x}_r - \dot{x}) + k_{x_3} \cdot (x_r - x) \\ \dddot{y}_r + k_{y_1} \cdot (\ddot{y}_r - \ddot{y}) + k_{y_2} \cdot (\dot{y}_r - \dot{y}) + k_{y_3} \cdot (y_r - y) \end{bmatrix} \quad (14)$$

where $x_r$ and $y_r$ are the position reference. $k_{x_i}$ and $k_{y_i}(i{=}1{,}2{,}3)$ are the coefficients for the controller.

## 4. Simulation and State Drift Phenomenon

This section provides the simulation results of the control rules in Section 3. An interesting phenomenon, which we call state drift, in the result is paid attention to.

The reference is a circular trajectory starting from (0,0) in Figure 5. The radius of the reference is 10. The velocity of the reference is 10. The initial position of the vehicle is (0,0). The initial velocity, is also zero. The initial angular velocities are $\omega_1 = 200$ and $\omega_2 = 200$.

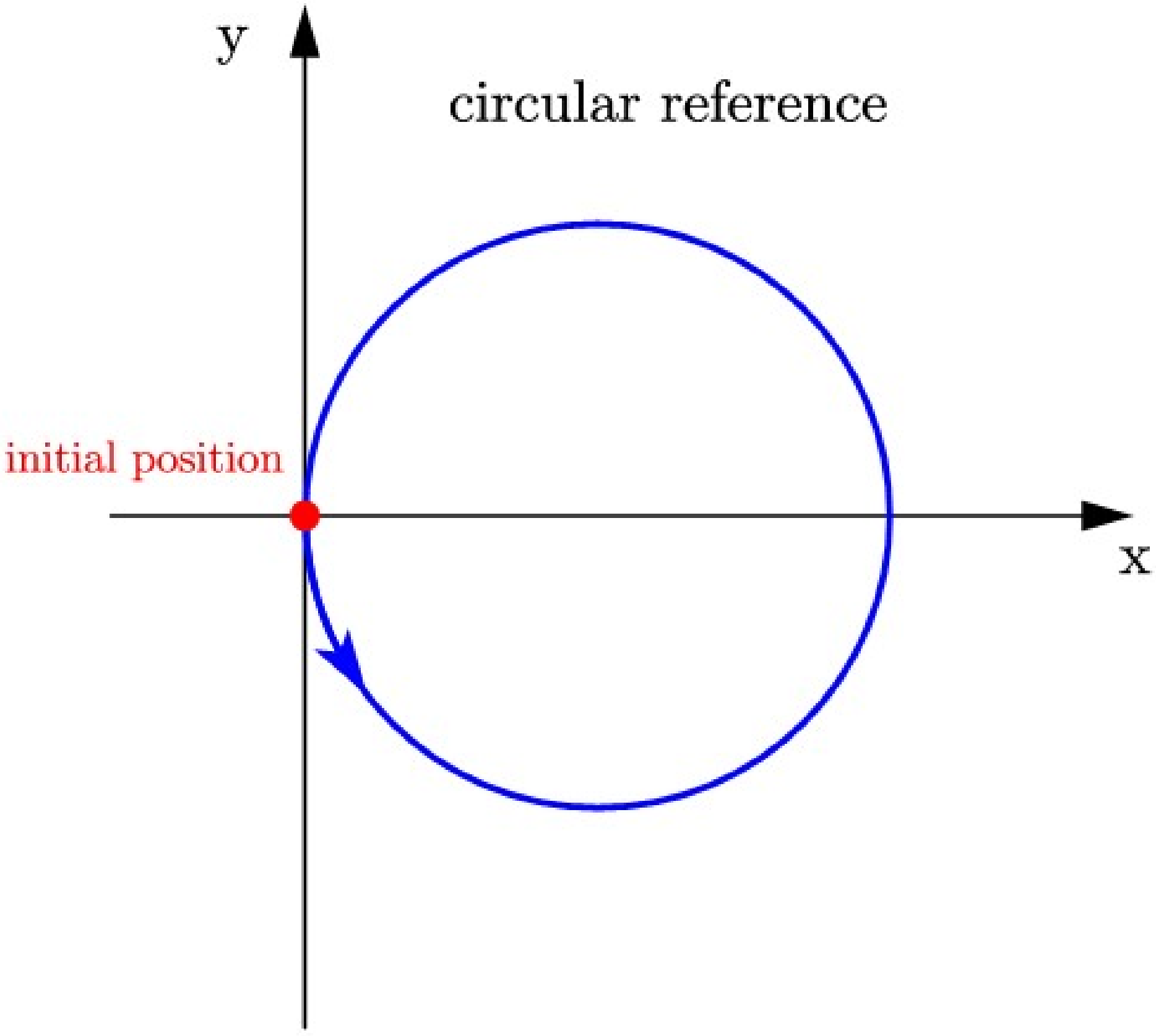

**Figure 5.** The circular reference.

The solver is a fixed-step ODE3 in Simulink, MATLAB. The sampling time is 0.01 second.

### *4.1. Results for Third Derivative Feedback Linearization*

The Simulink block diagram is plotted in Figure 6; the Dynamics part, Reference part, and Controller part are indicated.

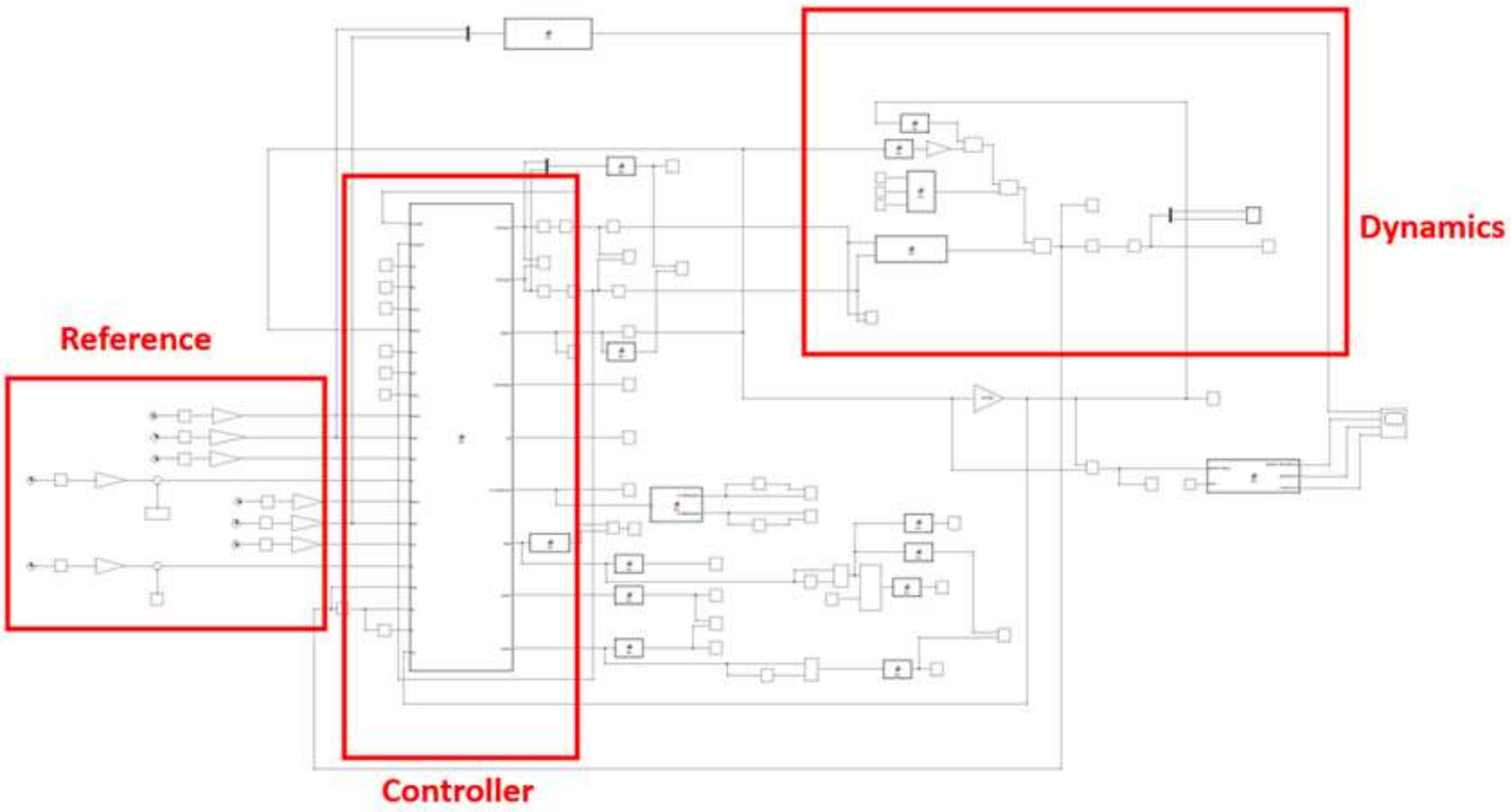

**Figure 6.** Simulink block in simulation

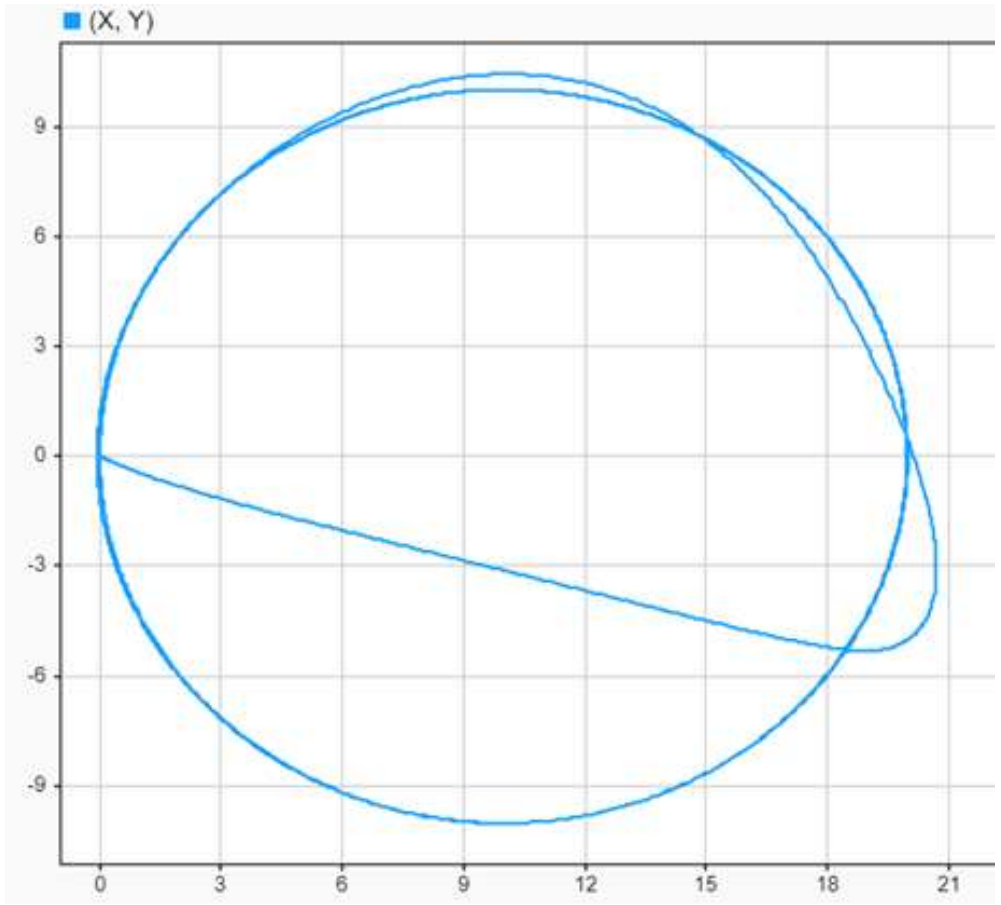

**Figure 7.** The trajectory.

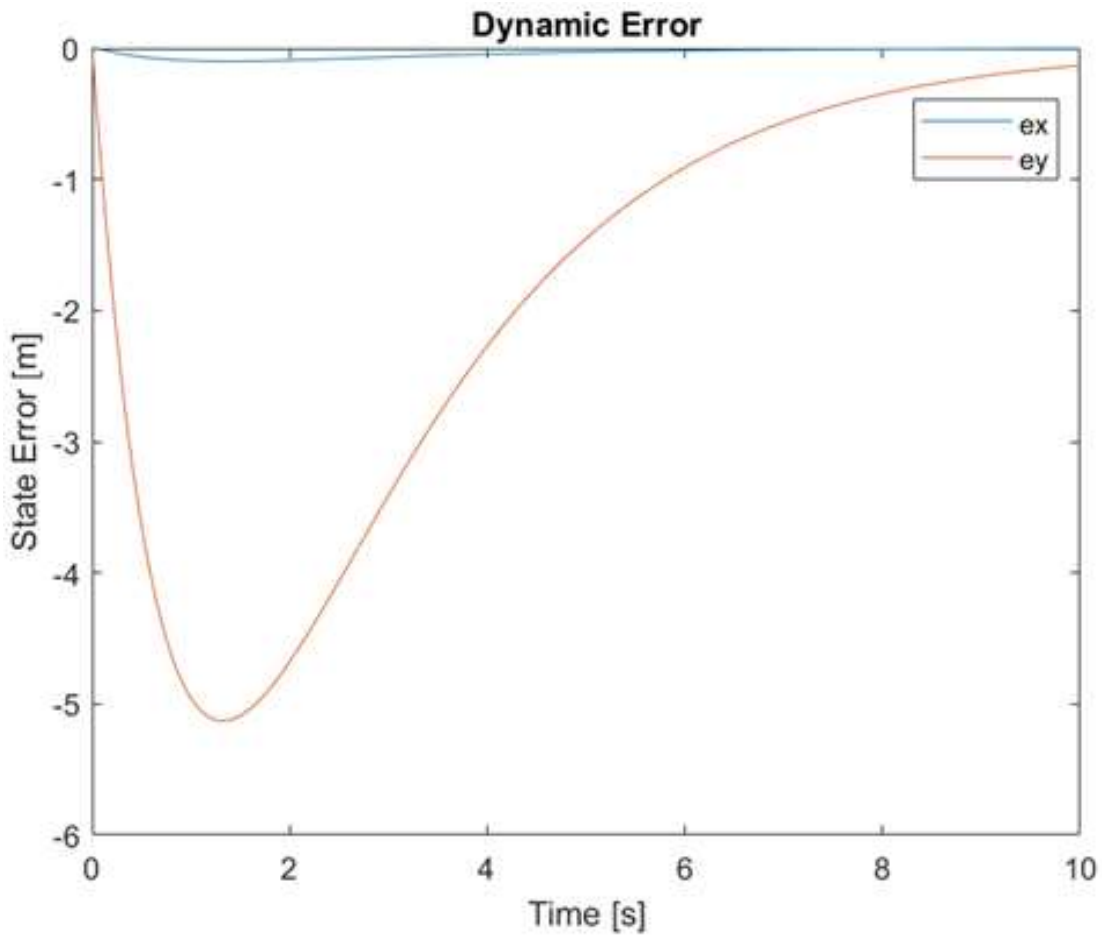

**Figure 8.** The dynamic state error.

Set the simulation time 10 seconds. The results are illustrated in Figure 7—10.

In Figure 7, the vehicle moves on a biased circular trajectory. The dynamic state error can be checked in Figure 8. Both the dynamic state errors in $x$ and $y$ directions decrease to zero.

Figure 9 shows the square of the input signals. Notice that two inputs almost overlap each other in Figure 9, especially at the beginning. The input signals seem to diverge/separate in the later time.

Figure 10 plots the information related to the direction of the thrust. The direction of the middle of two thrusts can be calculated in Equation (15).

$$\alpha + \psi \tag{15}$$

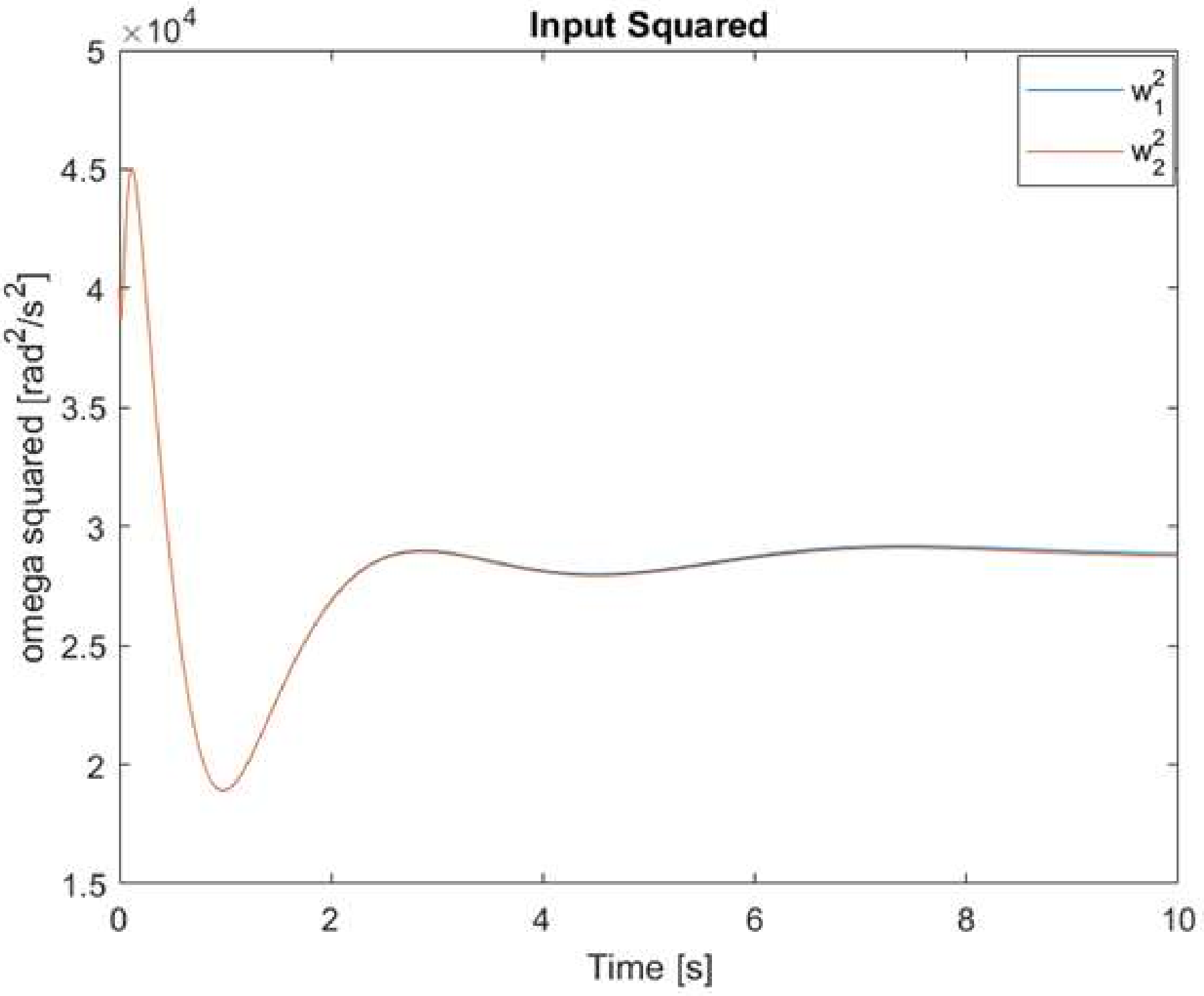

**Figure 9.** Input squared.

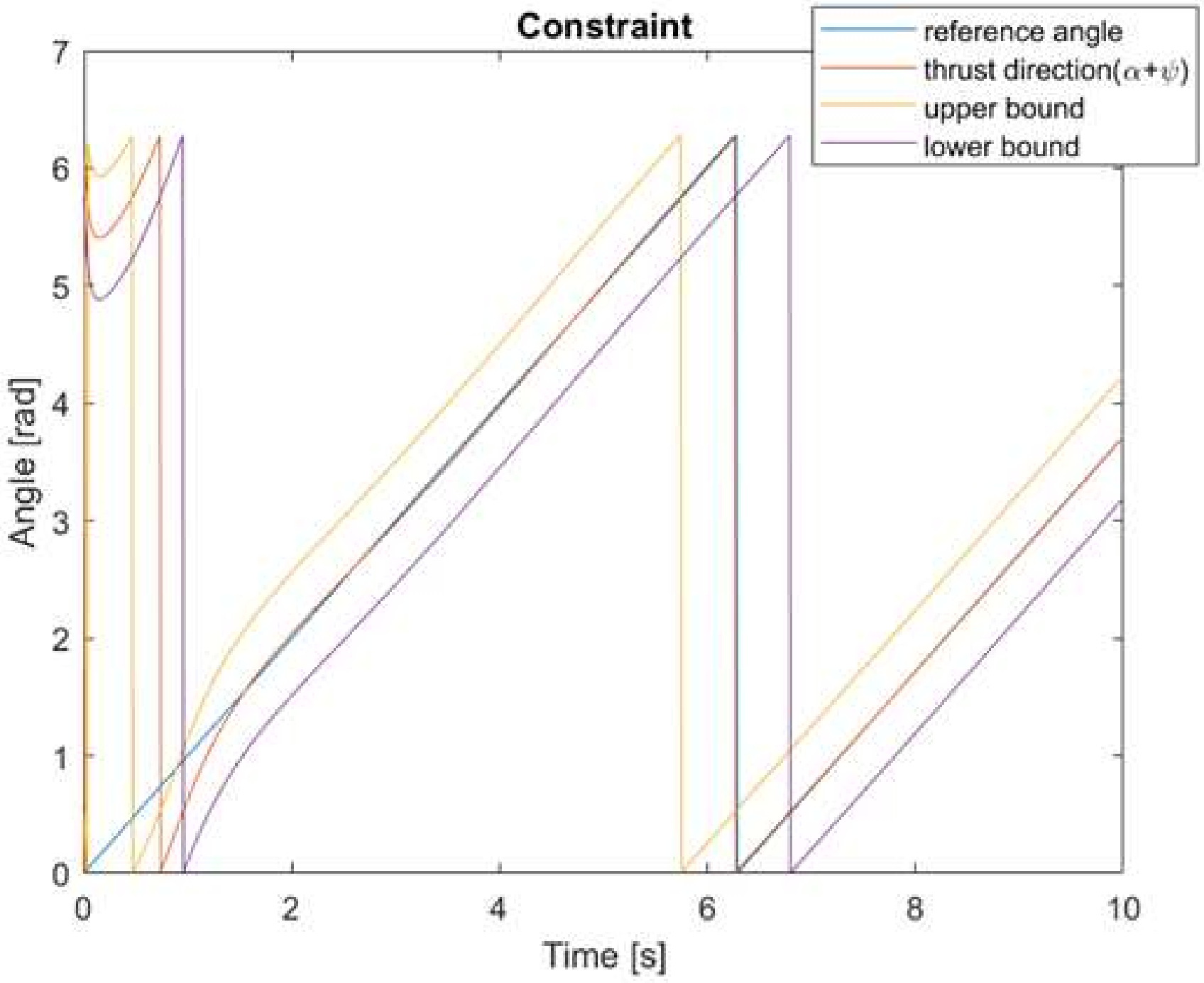

**Figure 10.** The direction of the thrust.

The orange line in Figure 10 represents the direction of the middle of two thrusts. The yellow line (upper bound) and the purple line (lower bound) represent the admissible direction bound of the vehicle. The potential direction for generating the acceleration is between the lower bound and the upper bound. The upper bound and lower bound are computed in Equation (16)—(17), respectively.

$$\alpha + \psi + \frac{\pi}{6} \tag{16}$$

$$\alpha + \psi - \frac{\pi}{6} \tag{17}$$

The blue line in Figure 10 is the direction pointing to the center of the circle (Figure 11). We can see that the direction of the middle of two thrusts of the vehicle (orange line) seems to stabilize at the direction pointing to the center of the circle (Figure 11) when time approaches 10 seconds.

Extending the simulation time to 2000 seconds, we receive the results in Figure 12—15.

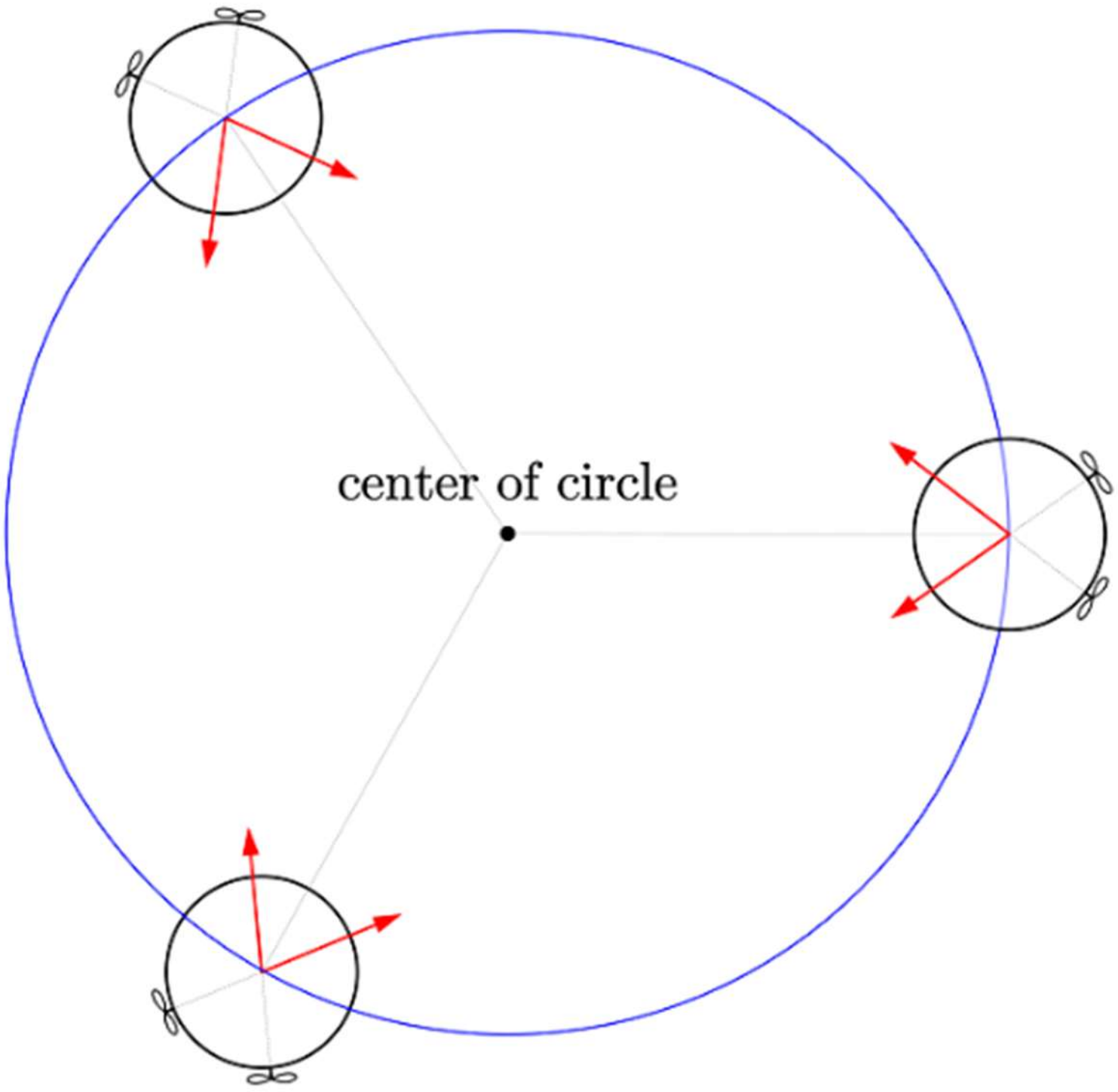

**Figure 11.** The direction pointing to the center of the circle

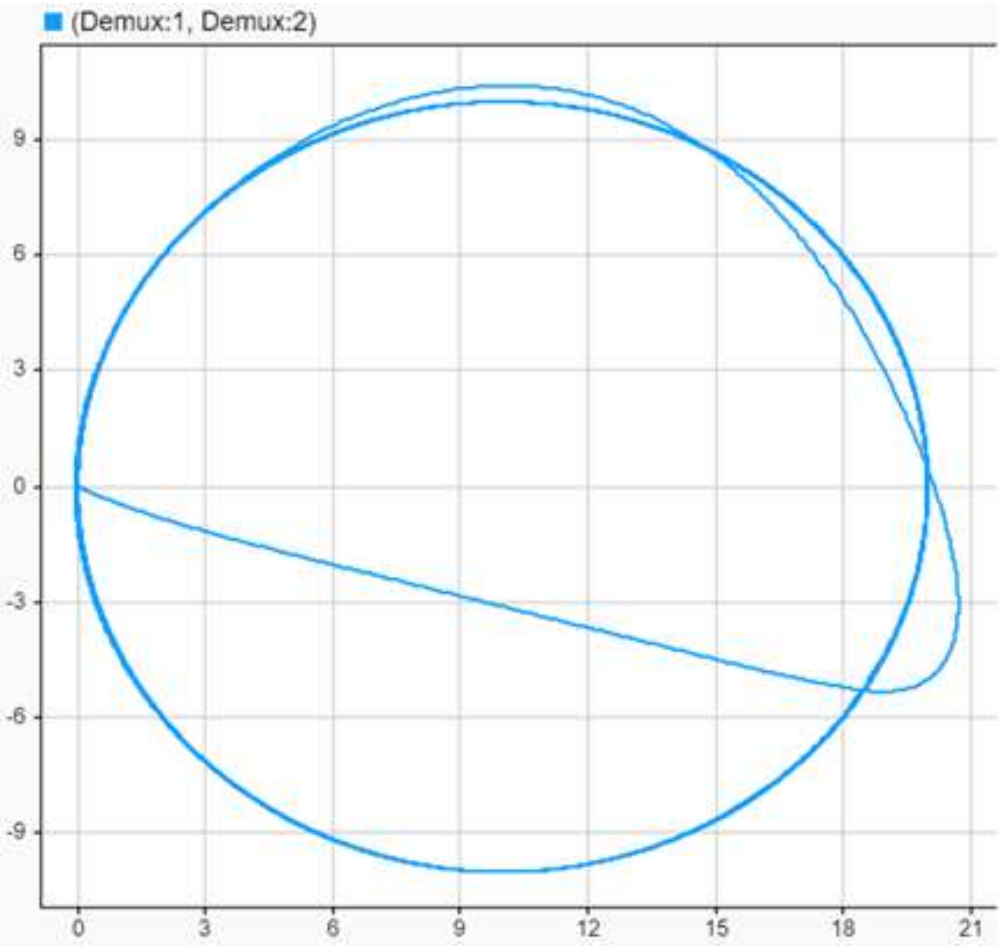

**Figure 12.** The trajectory.

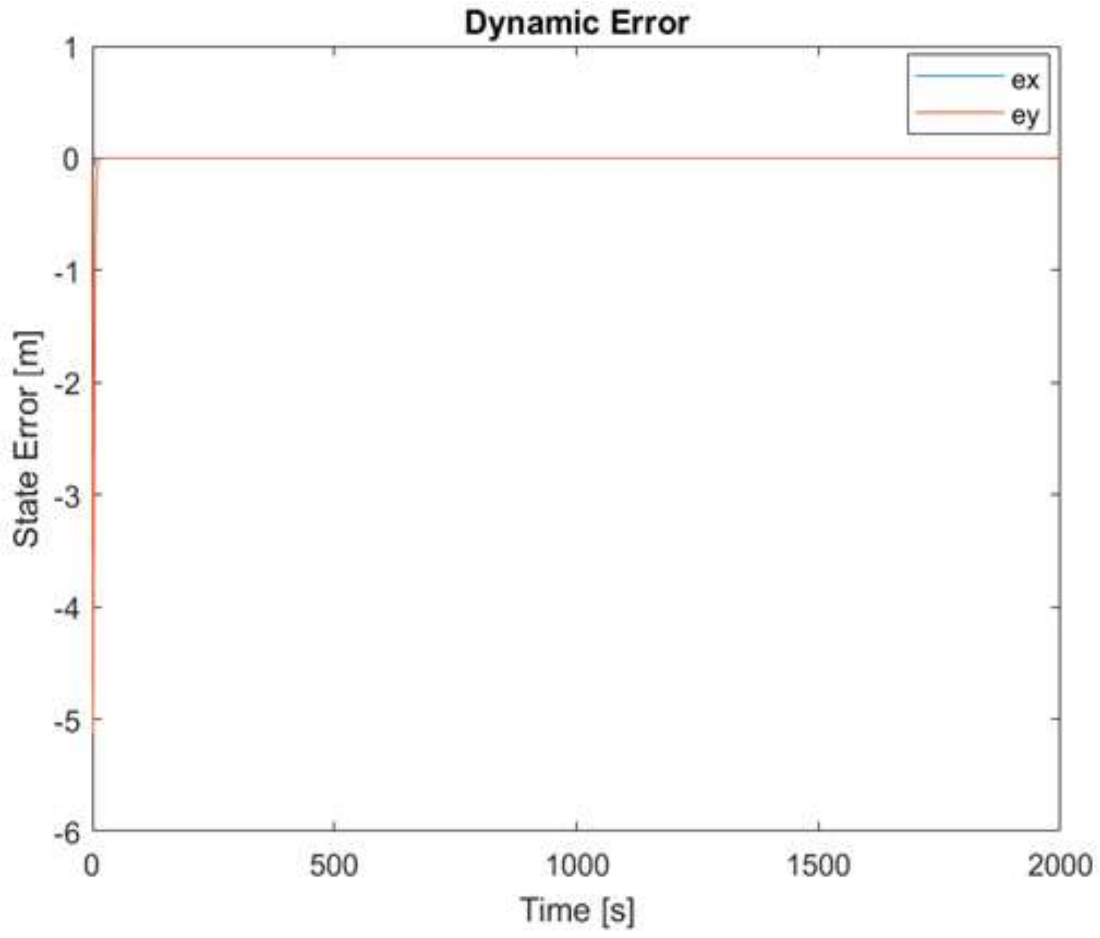

**Figure 13.** The dynamic state error.

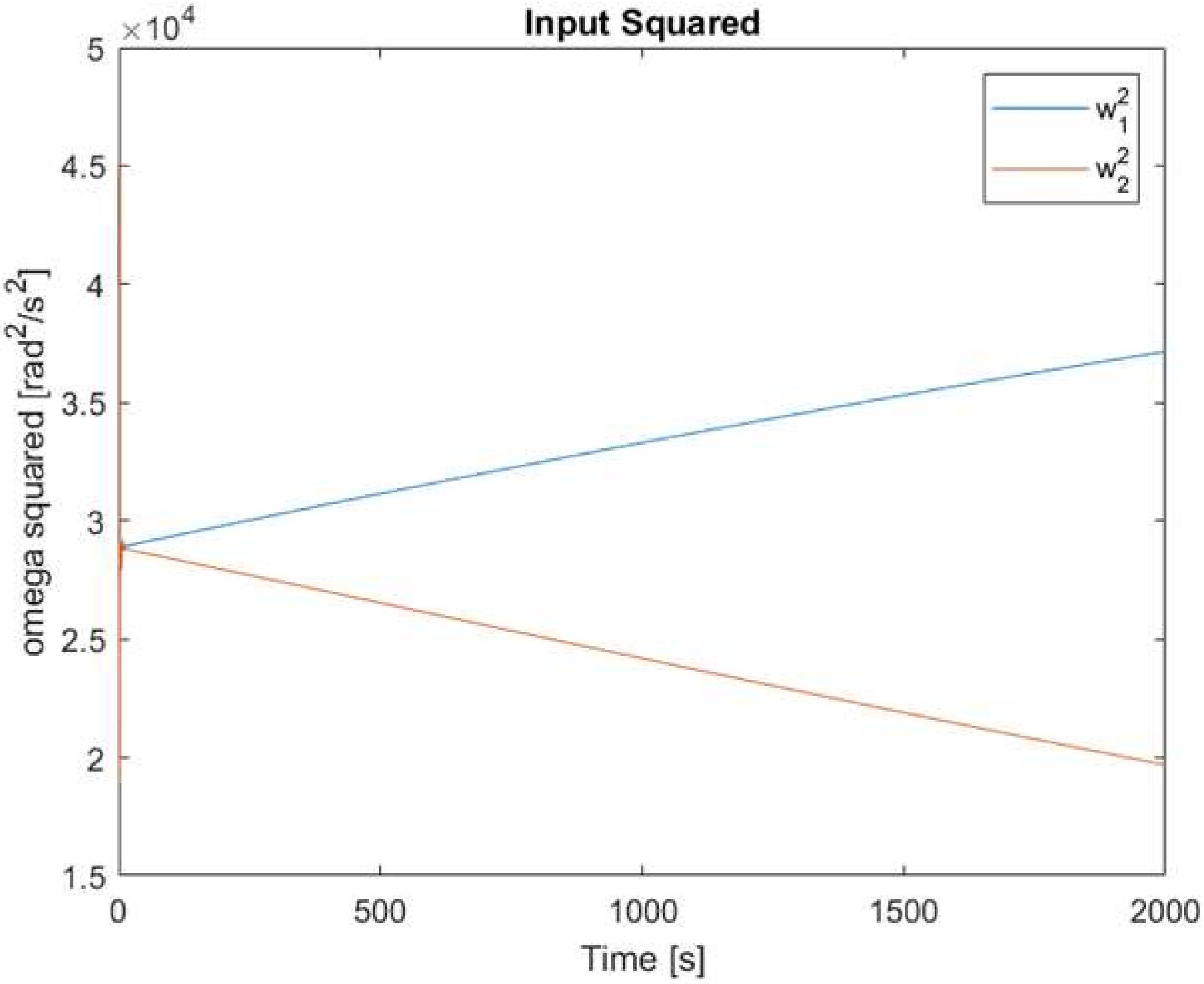

**Figure 14.** Input squared.

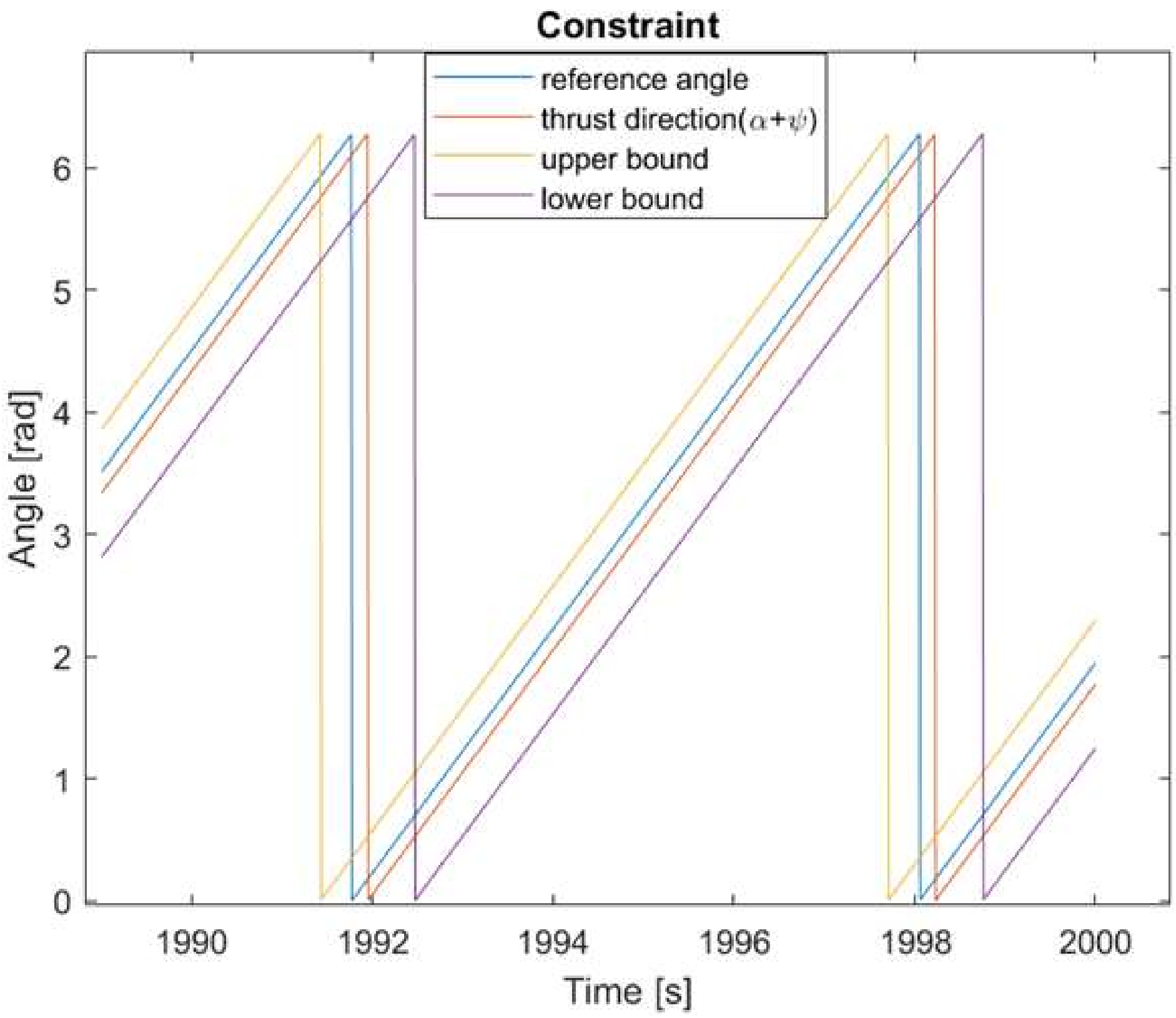

**Figure 15.** The direction of the thrust.

The trajectory follows the circle in high precision (Figure 12,13) after being controlled for sufficient long time. The dynamic state error remains close to zero, indicating the stability of the system. Figure 14 shows the squares of the two angular velocities. The inputs diverge in two opposite directions.

Figure 15 shows the acceleration constraint around 1990—2000 seconds. The blue line in Figure 15 is the direction pointing to the center of the circle (Figure 11). The orange line is the direction of the middle of two thrusts. It dramatically biases from the direction pointing to the center of the circle (blue line).

*4.2 State Drift*

The interesting phenomenon in the result is that the input changes dramatically after the vehicle is stable already. We define this phenomenon 'State Drift'. State Drift happens only in the over-actuated systems. The direct reason causing State Drift refers to Equation (12), (14). While deducing it mathematically in a systematic way can be hard or even impossible.

While the underlying reason causing State Drift in this vehicle can be obviously explained.

The centripetal force pointing to the center of the circle is required to follow the reference in Figure 15. While generating a specific centripetal force, the combination of the magnitudes of the two thrusts ($F_1$,$F_2$) is not unique (Figure 16—18). State Drift can be avoided if the number of the inputs is no more than the number of the output.

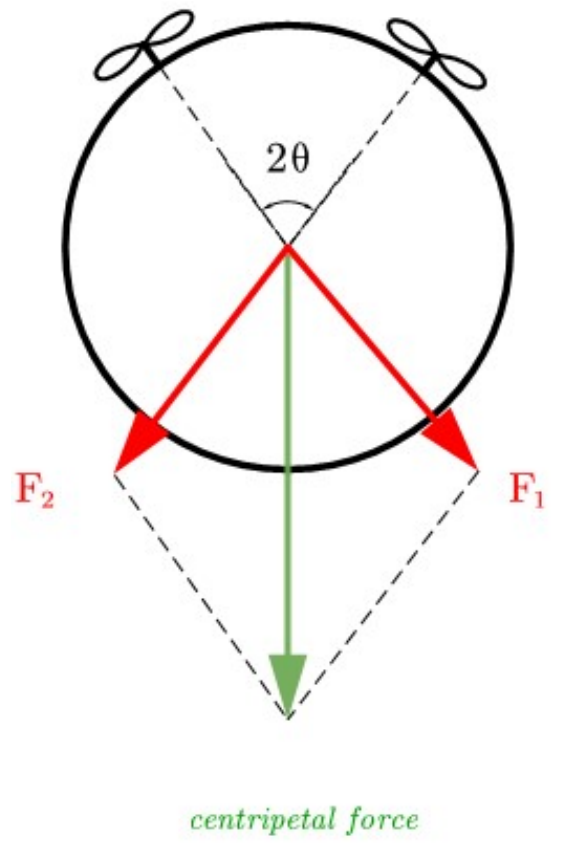

**Figure 16.** Symmetrical.

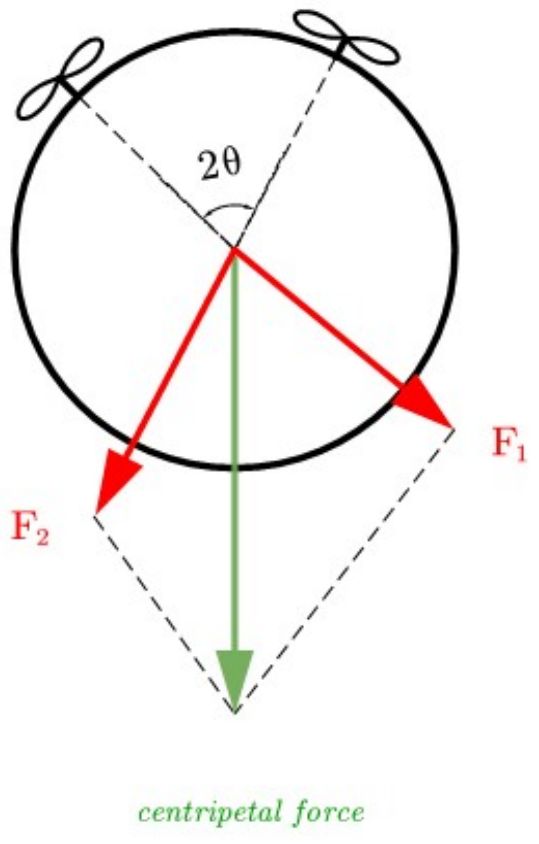

**Figure 17.** Symmetrical.

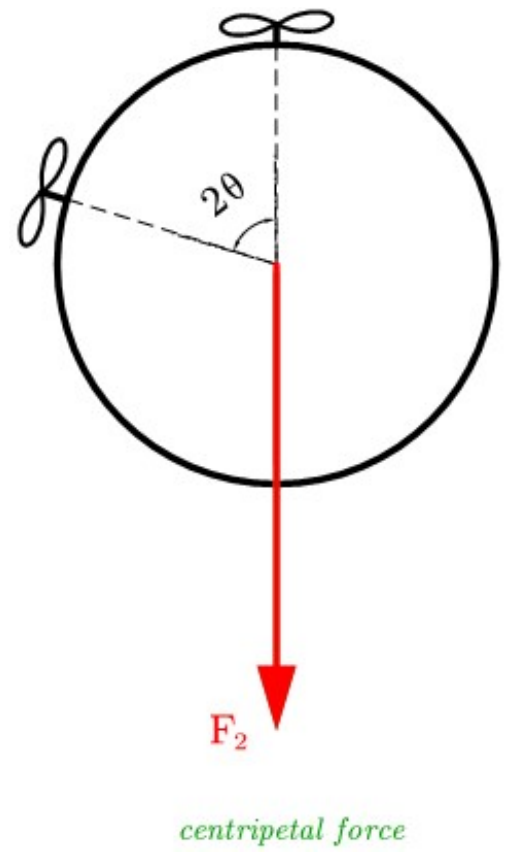

**Figure 18.** Degrade to one thrust.

## 5. Gait Plan

The approach to avoid the State Drift is to decrease the number of inputs to be equal to the number of the outputs. In this research, we plan the direction of the middle of two thrusts in advance and discard the tilt angle, $\alpha$, from the input. The only two inputs then are the angular velocities of the two propellers, $\omega_1$ and $\omega_2$.

$\alpha$ is defined in Equation (18) such that the direction of the middle of two thrusts ($\psi + \alpha$) always points to the center of the circular reference in Figure 11 during the entire movement.

$$\alpha = \frac{1}{1-\frac{T_T}{T_B}} \cdot t \tag{18}$$

Having decided $\alpha$ in Equation (18), only two inputs, $\omega_1$ and $\omega_2$, are left to be defined.

Based on the dynamics in Equation (3), the dynamic inversion is completed in Equation (19).

$$\begin{bmatrix} {\omega_1}^2 \\ {\omega_2}^2 \end{bmatrix} = m \cdot \left(J_{\psi+\alpha} \cdot J_\theta\right)^{-1} \cdot \begin{bmatrix} \ddot{x} \\ \ddot{y} \end{bmatrix} \tag{19}$$

It is worth mentioning that Equation (19) can be received for the reason that the term $J_{\psi+\alpha} \cdot J_\theta$ is invertible.

Further, we develop the PD controllers in Equation (20).

$$\begin{bmatrix} {\omega_1}^2 \\ {\omega_2}^2 \end{bmatrix} = m \cdot \left(J_{\psi+\alpha} \cdot J_\theta\right)^{-1} \cdot \begin{bmatrix} \ddot{x}_r + k_{x_1} \cdot (\dot{x}_r - \dot{x}) + k_{x_2} \cdot (x_r - x) \\ \ddot{y}_r + k_{y_1} \cdot (\dot{y}_r - \dot{y}) + k_{y_2} \cdot (y_r - y) \end{bmatrix} \tag{20}$$

where $x_r$ and $y_r$ are the position reference. $k_{x_i}$ and $k_{y_i}(i=1,2)$ are the coefficients for the controller.

The same reference in Figure 5 is used for the tracking experiment to test the controller in Equation (20). The Simulink block diagram for this experiment is plotted in Figure 19. The dynamics, controller, and the reference parts are marked.

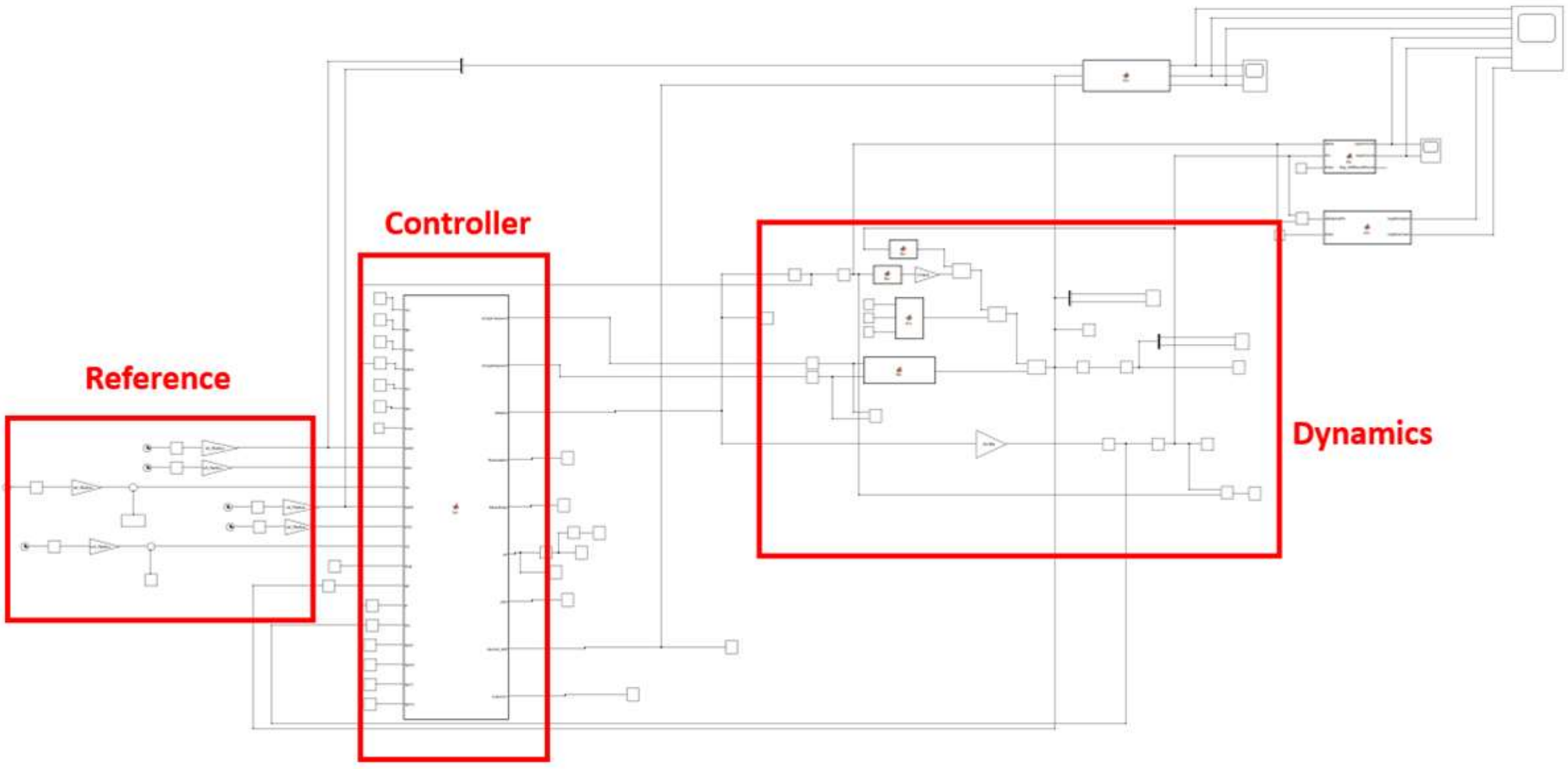

**Figure 19.** Simulink block diagram

The simulation time is 10 seconds. The sampling time is 0.01 second (ODE3 fixed-step solver). The result is plotted in Figures 20—23.

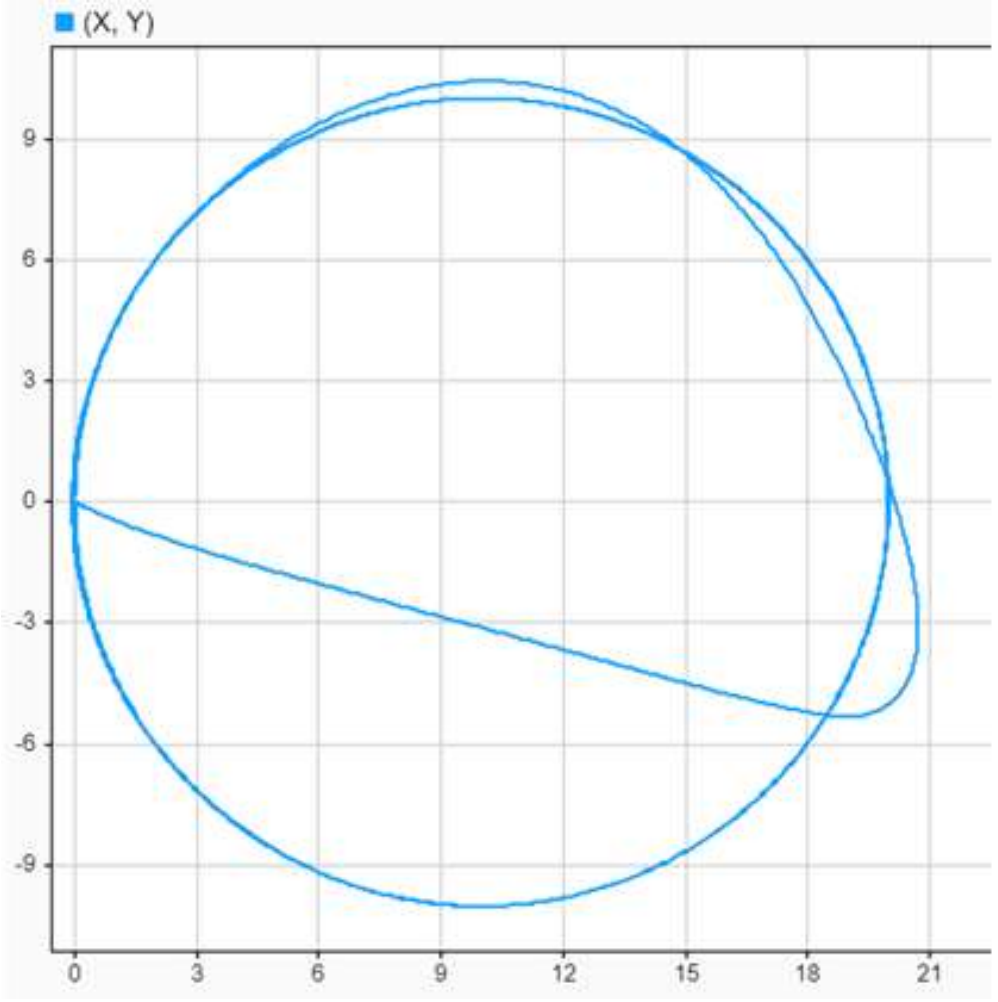

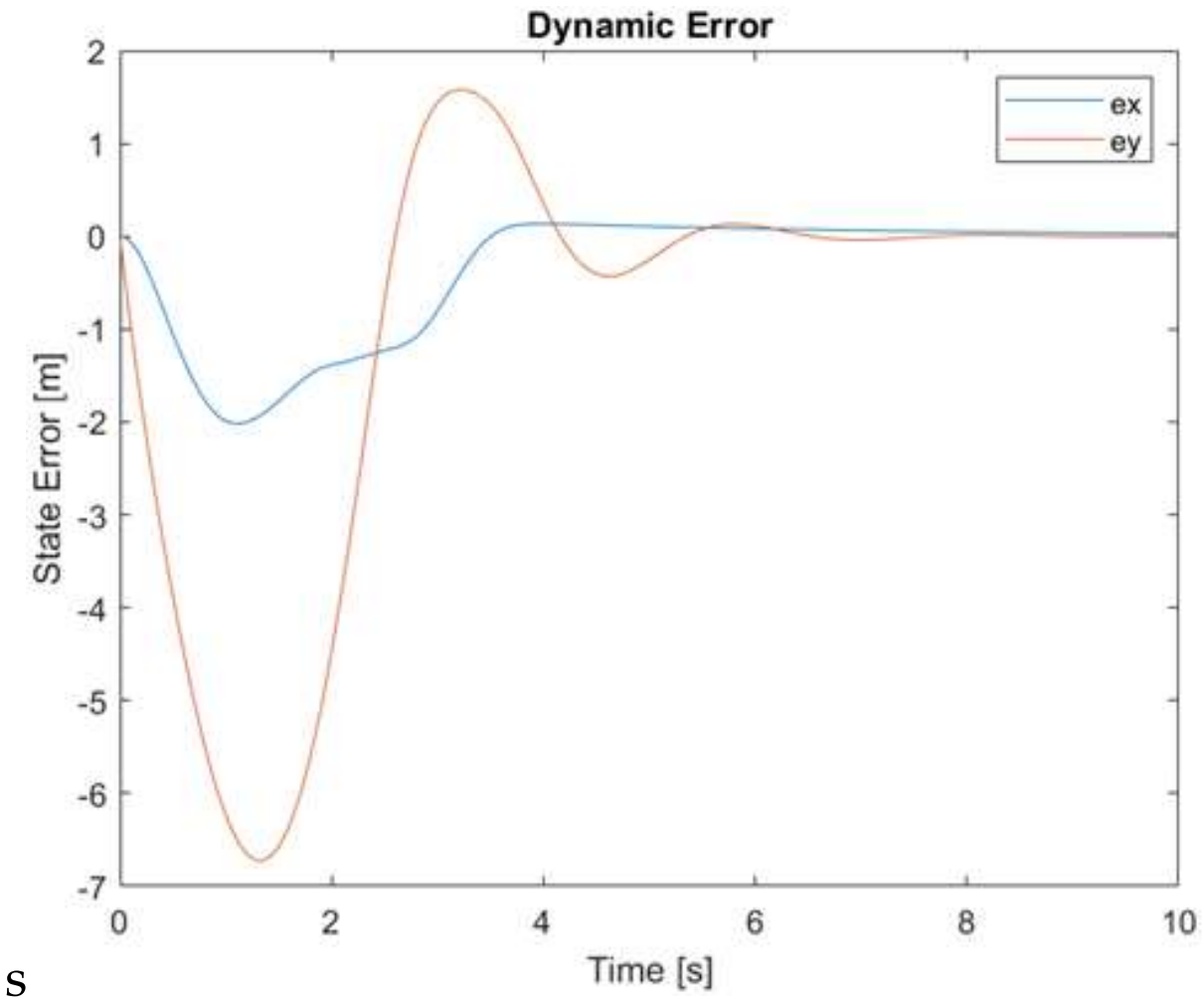

s

**Figure 20.** The trajectory.

**Figure 21**. The dynamic state error.

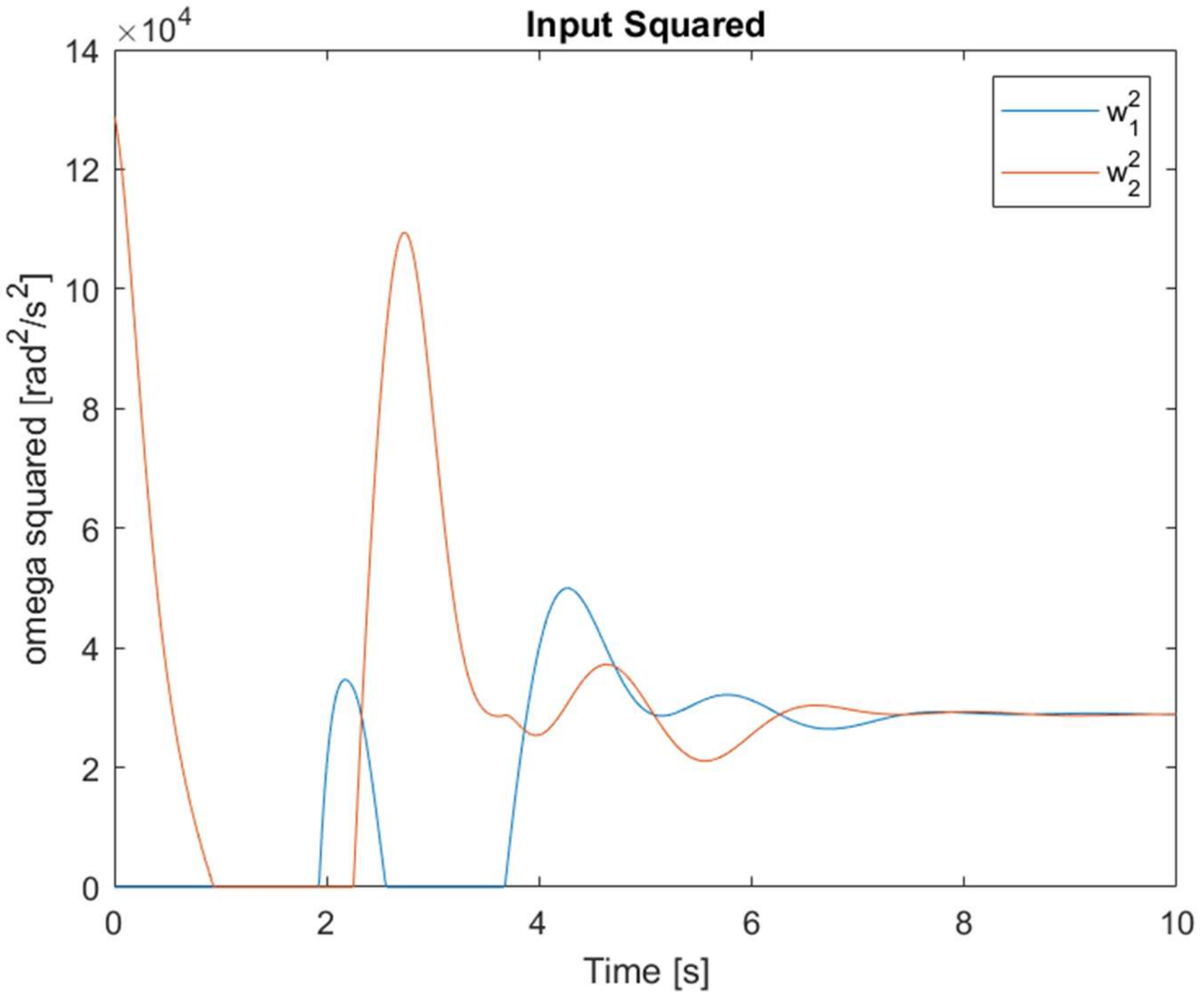

**Figure 22.** Input squared.

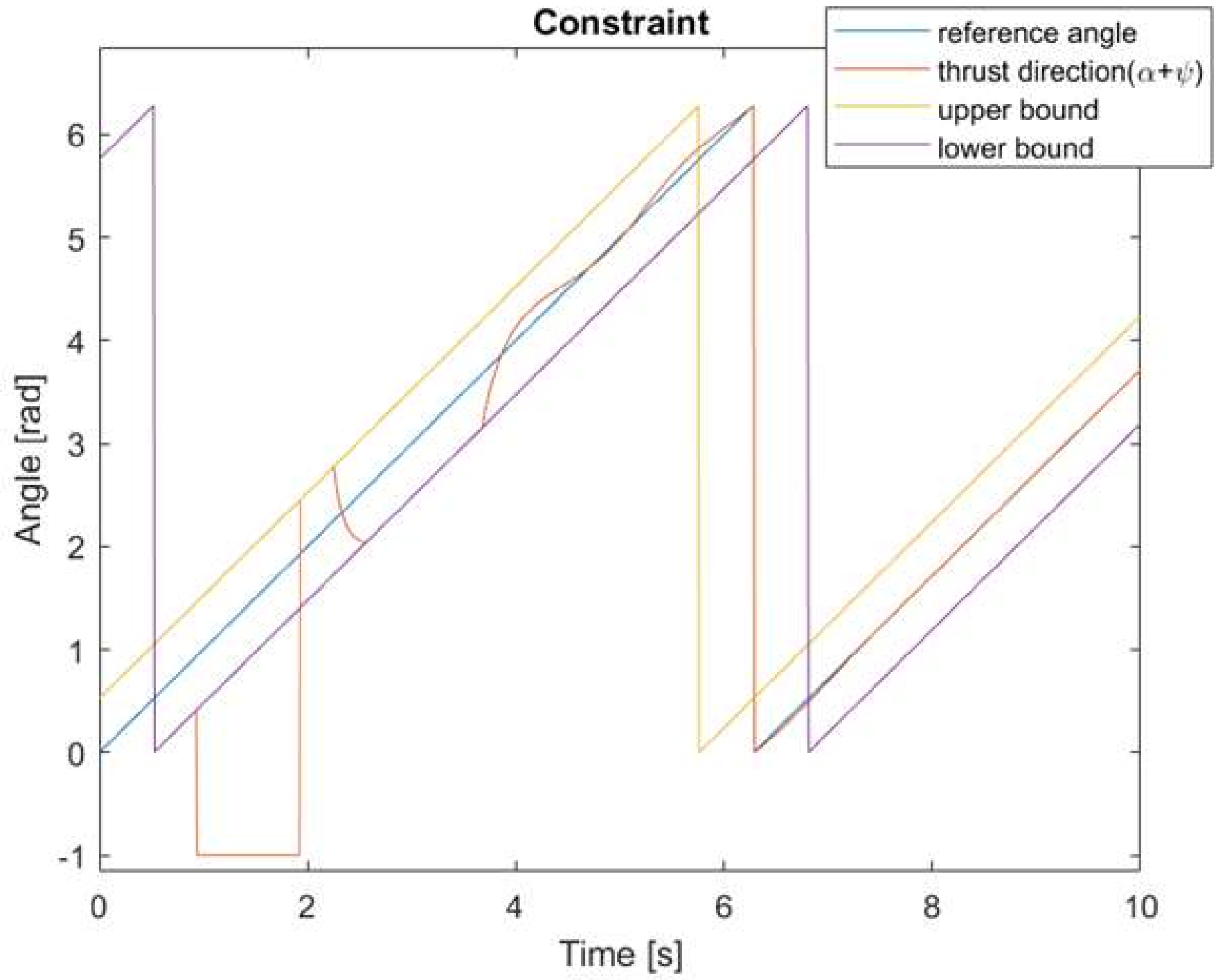

**Figure 23.** The direction of the thrust (angle -1 represents 0 acceleration).

Judged from Figure 20—21, the trajectory generally follows the reference as a circle after some time. In Figure 23, we can see that the inputs do not diverge in two opposite directions. Actually, they become identical in the end. This is because that the number of the inputs is equal to the number of the outputs; State Drift can happen only in an actuated system. For a system with an equal number of inputs and outputs, State Drift will not happen.

There is an undesired behavior in this controller: the saturation in inputs.

The feedback linearization transfers the nonlinear system to a linear system (as what we have done in Equation (19)). So that the linear controller can be applied. This approach is valid and solid if and only if the constraint is not effective; hitting the bound (e.g., positive thrust constraint) invalidates the feedback linearization method.

The nonlinear property of the original system should be considered if the saturation happens. However, there lacks stability proof for a general linear time invariant system. Only several categories of linear time variant systems with specific forms can be proved stable [19,20]. Seeking the stability proof for the rest linear time variant systems is still an open question.

Unfortunately, when saturation happens in this controller, Equation (20), with the circular reference in Figure 5, the linear time variant system lies in the category whose stability proof is not found. The detail of the deducing process is omitted in this paper, since the analysis of a nonlinear system is beyond this research.

The saturation can also be observed in Figure 23, the direction of the thrust. It can be seen that the input signals saturate at around 3 seconds.

## 6. Conclusions And discussions

The results in the over-actuated state feedback linearization show that the system is stabilized. The State Drift happens in the controlled over-actuated system, although the stability is not affected in this system.

Applying the gait plan to this system, the number of the inputs becomes equal to the number of the outputs. Subsequently, the State Drift phenomenon disappears. The saturation happens in this control method, which affects the stability proof.

Further steps can be deducing the stability proof for the control methods in gait plan with saturation and different gait patterns development.

In the experiments in this research, State Drift seems to only affect the allocation of the inputs. While the stability seems not be affected by this phenomenon. In this section, we will see a higher order feedback linearization controller. This controller is to stabilize the system in Equation (2)—(3). The State Drift in this example shows that the stability is affected.

### *6.1. Fourth Derivative Feedback Linearization*

Calculate the fourth derivative of the position by differentiating Equation (7).

The result yields:

$$\begin{bmatrix} \bar{x} \\ \bar{y} \end{bmatrix} = \frac{1}{m} \cdot J_\psi \cdot \ddot{J}_\alpha \cdot J_\theta \cdot \begin{bmatrix} {\omega_1}^2 \\ {\omega_2}^2 \end{bmatrix} + \frac{2}{m} \cdot J_\psi \cdot \dot{J}_\alpha \cdot J_\theta \cdot \begin{bmatrix} 2\omega_1 \cdot \dot{\omega}_1 \\ 2\omega_2 \cdot \dot{\omega}_2 \end{bmatrix} + \frac{1}{m} \cdot J_\psi \cdot J_\alpha \cdot J_\theta \cdot \begin{bmatrix} 2{\dot{\omega}_1}^2 + 2\omega_1 \cdot \ddot{\omega}_1 \\ 2{\dot{\omega}_2}^2 + 2\omega_2 \cdot \ddot{\omega}_2 \end{bmatrix} + \frac{2}{m} \cdot \dot{J}_\psi \cdot \dot{J}_\alpha \cdot J_\theta \cdot \begin{bmatrix} {\omega_1}^2 \\ {\omega_2}^2 \end{bmatrix} + \frac{2}{m} \cdot \dot{J}_\psi \cdot J_\alpha \cdot J_\theta \cdot \begin{bmatrix} 2\omega_1 \cdot \dot{\omega}_1 \\ 2\omega_2 \cdot \dot{\omega}_2 \end{bmatrix} + \frac{1}{m} \cdot \ddot{J}_\psi \cdot J_\alpha \cdot J_\theta \cdot \begin{bmatrix} {\omega_1}^2 \\ {\omega_2}^2 \end{bmatrix} \quad (21)$$

where

$$J_\psi = \begin{bmatrix} \cos(\psi) & -\sin(\psi) \\ \sin(\psi) & \cos(\psi) \end{bmatrix} \quad (22)$$

$$J_\alpha = \begin{bmatrix} \cos(\alpha) & -\sin(\alpha) \\ \sin(\alpha) & \cos(\alpha) \end{bmatrix} \quad (23)$$

Equation (21) can also be written in Equation (24).

$$\begin{bmatrix} \ddot{x} \\ \ddot{y} \end{bmatrix} = M_P + \Delta_P \cdot \begin{bmatrix} \ddot{\omega}_1 \\ \ddot{\omega}_2 \\ \ddot{\alpha} \end{bmatrix} \tag{24}$$

where $\Delta_P \cdot \begin{bmatrix} \ddot{\omega}_1 \\ \ddot{\omega}_2 \\ \ddot{\alpha} \end{bmatrix}$ includes all the terms containing $\ddot{\omega}_1$, $\ddot{\omega}_2$, or $\ddot{\alpha}$. $M_P$ includes all the remaining terms without containing $\ddot{\omega}_1$, $\ddot{\omega}_2$, or $\ddot{\alpha}$.

Specifically,

$$\Delta_P = \left[\Delta_{P_\omega}^{2\times2} \middle| \Delta_{P_\alpha}^{2\times1}\right] \tag{25}$$

where

$$\Delta_{P_\omega}^{2\times2} = \frac{1}{m} \cdot J_\psi \cdot J_\alpha \cdot J_\theta \cdot \begin{bmatrix} 2\omega_1 & 0 \\ 0 & 2\omega_2 \end{bmatrix} \tag{26}$$

$$\Delta_{P_\alpha}^{2\times1} = \frac{1}{m} \cdot J_\psi \cdot \begin{bmatrix} -\sin(\alpha) & -\cos(\alpha) \\ \cos(\alpha) & -\sin(\alpha) \end{bmatrix} \cdot J_\theta \cdot \begin{bmatrix} {\omega_1}^2 \\ {\omega_2}^2 \end{bmatrix} \tag{27}$$

Notice that when $\omega_1$ and $\omega_2$ are not both zero, we have

$$Rank\left(\Delta_{P_\omega}^{2\times2}\right) = 2 \tag{28}$$

Equation (28) indicates that $\Delta_P$ has full row rank when $\omega_1$ and $\omega_2$ are not both zero. With this condition, $pinv(\Delta_P)$ can be calculated based on Equation (12).

Separating the inputs in Equation (24) yields Equation (29).

$$\begin{bmatrix} \ddot{\omega}_1 \\ \ddot{\omega}_2 \\ \ddot{\alpha} \end{bmatrix} = pinv(\Delta_P) \cdot \left( \begin{bmatrix} \ddot{x} \\ \ddot{y} \end{bmatrix} - M_P \right) \tag{29}$$

PD controllers are built based on Equation (29).

The reference and the initial condition are identical to the settings in the previous experiments. They are initial $\omega_1 = 200$, initial $\omega_2 = 200$, and Figure 5. The dynamic state error and the input squared are in Figure 24 and 25, respectively.

The dynamic state error (Figure 24) decreases to an acceptably small value (0.2 meter) at the beginning. While it increases after around $3 \times 10^4$ seconds. Finally, the dynamic state error diverges from 0, potentially making this system unstable.

It is the State Drift that causes this. Judged from Figure 25, the State Drift happens so that the two inputs diverge in the two opposite directions. As we know from the previous experiments, it can not be a factor causing unstable.

However, after 3 seconds, the input (${\omega_2}^2$) hit the positive constraint (negative thrust is not allowed), destroying the underlying stability proof. The consequence is what we see in Figure 24; unstable system with increasing dynamic state error is reported.

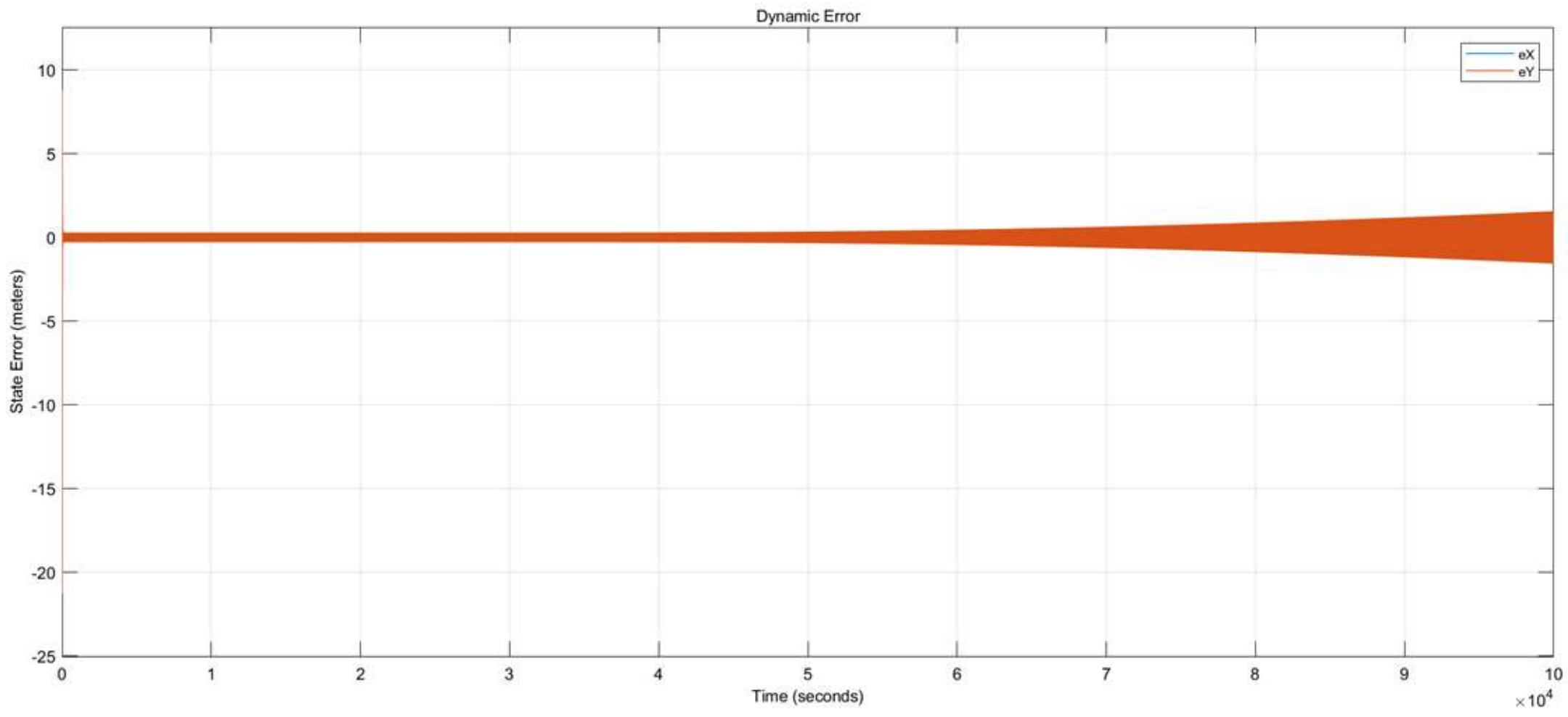

**Figure 24.** Dynamic State Error.

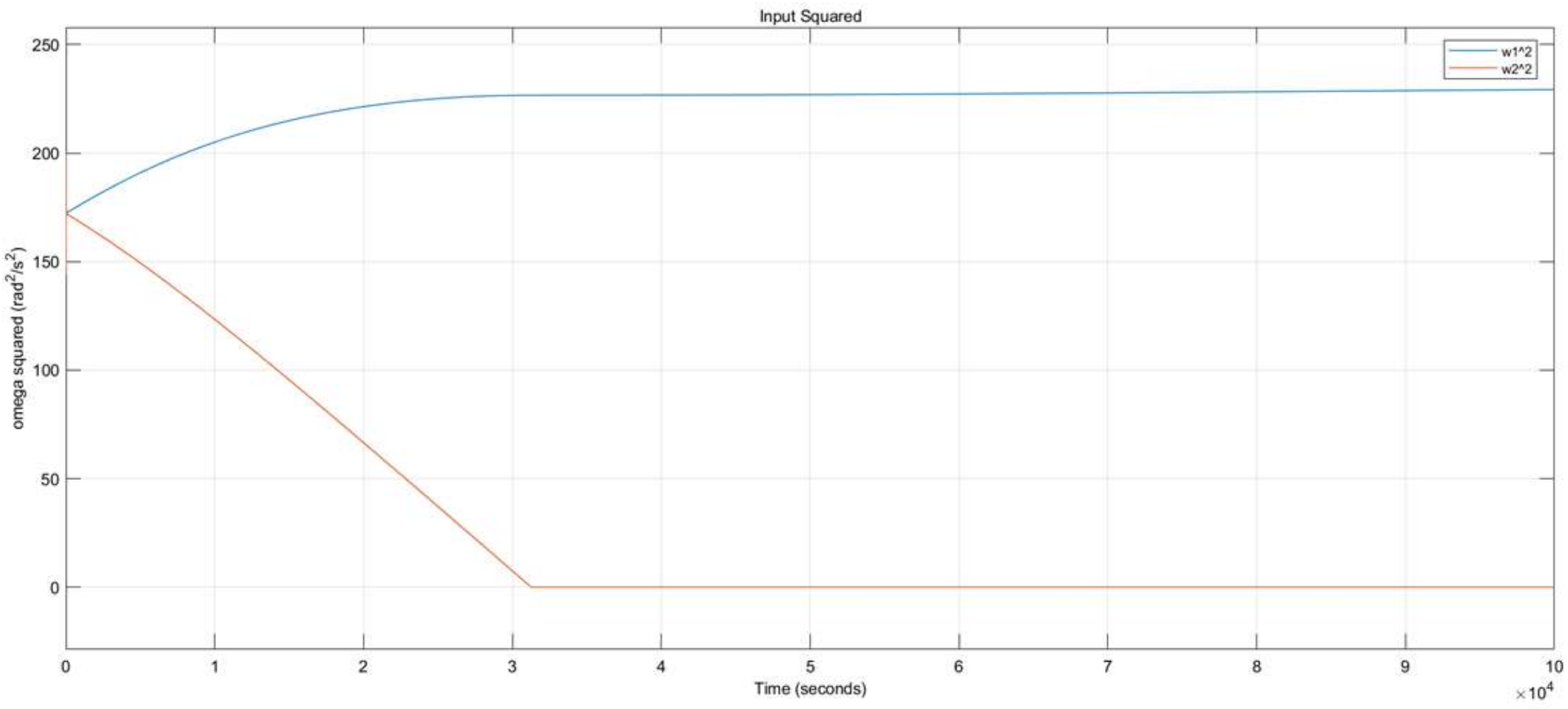

**Figure 25.** Input Squared.

*6.2. State Drift Phenomenon*

So far, State Drift phenomenon has been thoroughly reported by controlling a fictional system (tilt vehicle). The reason can be multiple equilibrium states (inputs), which is explained in Figure 16—18.

However, the mathematical cause for pulling the system from one equilibrium to another is complicated and beyond the scope of this research.

For example, the following question is extremely hard to reply: I understand that the system is stable even if State Drift happens, but why does State Drift happen?

The answer to this question can also be a challenged further step.

## References

1. Taniguchi, T., & Sugeno, M. (2017). Trajectory tracking controls for non-holonomic systems using dynamic feedback linearization based on piecewise multi-linear models. *IAENG International Journal of Applied Mathematics*, *47*(3), 339-351.
2. Mistler, V., Benallegue, A., & M'sirdi, N. K. (2001, September). Exact linearization and noninteracting control of a 4 rotors helicopter via dynamic feedback. In *Proceedings 10th IEEE international workshop on robot and human interactive communication. Roman 2001 (Cat. no. 01th8591)* (pp. 586-593). IEEE.

3. Mokhtari, A., & Benallegue, A. (2004, April). Dynamic feedback controller of Euler angles and wind parameters estimation for a quadrotor unmanned aerial vehicle. In *IEEE International Conference on Robotics and Automation, 2004. Proceedings. ICRA'04. 2004* (Vol. 3, pp. 2359-2366). IEEE.
4. Ryll, M., Bülthoff, H. H., & Giordano, P. R. (2012, May). Modeling and control of a quadrotor UAV with tilting propellers. In *2012 IEEE international conference on robotics and automation* (pp. 4606-4613). IEEE.
5. Bouabdallah, S., Noth, A., & Siegwart, R. (2004, September). PID vs LQ control techniques applied to an indoor micro quadrotor. In *2004 IEEE/RSJ International Conference on Intelligent Robots and Systems (IROS)(IEEE Cat. No. 04CH37566)* (Vol. 3, pp. 2451-2456). IEEE.
6. Lee, H., Kim, S., Ryan, T., & Kim, H. J. (2013, October). Backstepping control on se (3) of a micro quadrotor for stable trajectory tracking. In *2013 IEEE international conference on systems, man, and cybernetics* (pp. 4522-4527). IEEE.
7. Lee, T., Leok, M., & McClamroch, N. H. (2010, December). Geometric tracking control of a quadrotor UAV on SE (3). In *49th IEEE conference on decision and control (CDC)* (pp. 5420-5425). IEEE.
8. Xu, R., & Ozguner, U. (2006, December). Sliding mode control of a quadrotor helicopter. In *Proceedings of the 45th IEEE Conference on Decision and Control* (pp. 4957-4962). IEEE.
9. Bouabdallah, S., & Siegwart, R. (2005, April). Backstepping and sliding-mode techniques applied to an indoor micro quadrotor. In *Proceedings of the 2005 IEEE international conference on robotics and automation* (pp. 2247-2252). IEEE.
10. Das, A., Subbarao, K., & Lewis, F. (2008, September). Dynamic inversion of quadrotor with zero-dynamics stabilization. In *2008 IEEE International Conference on Control Applications* (pp. 1189-1194). IEEE.
11. Voos, H. (2009, April). Nonlinear control of a quadrotor micro-UAV using feedback-linearization. In *2009 IEEE International Conference on Mechatronics* (pp. 1-6). IEEE.
12. Shen, Z., & Tsuchiya, T. (2021). Singular Zone in Quadrotor Yaw-Position Feedback Linearization. *arXiv preprint arXiv:2110.07179*.
13. Şenkul, F., & Altuğ, E. (2013, May). Modeling and control of a novel tilt—Roll rotor quadrotor UAV. In *2013 International Conference on Unmanned Aircraft Systems (ICUAS)* (pp. 1071-1076). IEEE.
14. Nemati, A., Kumar, R., & Kumar, M. (2016, October). Stabilizing and control of tilting-rotor quadcopter in case of a propeller failure. In *Dynamic Systems and Control Conference* (Vol. 50695, p. V001T05A005). American Society of Mechanical Engineers.
15. Rajappa, S., Ryll, M., Bülthoff, H. H., & Franchi, A. (2015, May). Modeling, control and design optimization for a fully-actuated hexarotor aerial vehicle with tilted propellers. In *2015 IEEE international conference on robotics and automation (ICRA)* (pp. 4006-4013). IEEE.
16. Ryll, M., Bülthoff, H. H., & Giordano, P. R. (2014). A novel overactuated quadrotor unmanned aerial vehicle: Modeling, control, and experimental validation. *IEEE Transactions on Control Systems Technology*, *23*(2), 540-556.
17. Elfeky, M., Elshafei, M., Saif, A. W. A., & Al-Malki, M. F. (2013, June). Quadrotor helicopter with tilting rotors: Modeling and simulation. In *2013 world congress on computer and information technology (WCCIT)* (pp. 1-5). IEEE.
18. Zamani, A., Khorram, M., & Moosavian, S. A. A. (2011, October). Dynamics and stable gait planning of a quadruped robot. In *2011 11th International Conference on Control, Automation and Systems* (pp. 25-30). IEEE.
19. Wu, F., & Grigoriadis, K. M. (1997, December). LPV systems with parameter-varying time delays. In *Proceedings of the 36th IEEE Conference on Decision and Control* (Vol. 2, pp. 966-971). IEEE.
20. Zhou, B. (2016). On asymptotic stability of linear time-varying systems. *Automatica*, *68*, 266-276.

# Chapter 11

# Stability Analysis of Feedback Linearization with Saturation (Stability Analysis of a Feedback-Linearization-Based Controller with Saturation: A Tilt Vehicle with the Penguin-Inspired Gait Plan)

**This Chapter has been submitted to IEEE Transactions on Control Systems Technology as the first author. The rest co-author(s) contribute nothing other than supervising and/or minor edition.**

**Abstract:** In this paper, we apply the feedback linearization method to control a novel vehicle, tilt vehicle. The tilt vehicle is a fictional vehicle with tilting yaw. It is designed to help explore the controller design for a tilt rotor, a novel UAV. Penguin-inspired gait is adopted in this paper before controlling. Although the stability proof for a general feedback-linearization based controller is concrete in most cases, saturations in control signal can challenge this stability proof, even leading the system unstable for some cases. Thus, several approaches are established to avoid reaching the saturation bound. In the contrary, we utilize the property of saturation in control signal rather than discard it. The stability criteria do not hold for this case. Thus, the augmented stability proof is analyzed based on Lyapunov theory; the output is proved to be bounded, given the required initial condition. Simulation is conducted in Simulink, MATLAB. The result shows that the position is stabilized, maintaining the bounded oscillation for the gait with input saturation.

## 1. Introduction

A tilt vehicle is put forward [1] to aid the controller designation of the tilt rotor [4]. As reported, the gait plan is employed to avert the 'State Drift' phenomenon. Gait plan is originally adopted in four-legged robots [5-7], which was inspired by the four-legged animals. Parallel to the concept, we perform the gait plan for the tilt vehicle based on the moving pattern of penguins swinging to left and right periodically.

The procedure following gait plan is feedback linearization; this technique converts the nonlinear system to a linear system, providing the possibility of applying the linear controllers. This method is widely accepted in the studies [8-11] of the quadrotor control.

The generally adopted protocol in feedback linearization is averting the saturations. This is because that the system is no longer linearized when the states saturate. Adopting the unchanged control rule designed for the system without input saturation for the system with saturation challenges the stability proof, leading to unstable in some systems [1]. Several researches focus on avoiding activating the saturation constraints [12-16]. While the studies focusing on the stability proof of the dynamic inversion with saturation are rare.

In this research, we plan two gaits for the tilt vehicle. In the first gait, the feedback linearization with non-saturation states is assured. In the other gait, the saturation in states is inevitable, defying the previous stability proof (stability proof of the linear controller). To guarantee the stability of our controller, we provide the stability proof of the dynamic inversion with saturation for our system with the straight-line reference.

The Lyapunov criteria are commonly used in proving the asymptotically stable in UAV controls. Several well-known controllers are designed based on it: geometric control [17-20], backstepping control [21-23], sliding mode control [24-26], etc. Lyapunov candidates are found negative-definite in these studies, indicating the asymptotically stable [27,28]. However, seeking a Lyapunov candidate with such a requirement is hard or even impossible. Instead, we utilize the Lyapunov theorem to prove the relaxed definition of stability below.

$$\exists \epsilon, \exists \delta:$$

$$\text{If } |x(0)| < \delta,$$

$$\text{then } \left|x\big(t, x(0)\big)\right| < \epsilon, \forall t \geqslant 0.$$

This stability definition tells us: if the system is stable and the initial condition is within the admissible range, then the output is bounded.

The rest of the paper is organized as follows: Section 2 introduces the dynamics of the tilt vehicle. The penguin-gait-based gait plan is presented in Section 3. Section 4 designs the controller with the first gait, which avoids the saturation in states. The stability proof of this controller is also provided in this section. In Section 5, we adopt the same controller developed in Section 4. Since the gait in Section 5 leads to the saturation in states, the stability proof of this case is given in Section 6. The simulation results for this case are presented in Section 7. Section 8 makes conclusions and discussions.

## 2. Dynamics of The Tilt Vehicle

The tilt vehicle [1] to control is sketched in Figure 1—2. Two thrusts are provided by the propellers on the top disc. The top disc is able to tilt, giving the possibility of altering the direction of the thrusts. The friction between the bottom disc and the ground is neglected.

[1] analyzes its dynamics, which can be summarized in Equation (1).

$$\begin{bmatrix} \ddot{x} \\ \ddot{y} \end{bmatrix} = \frac{1}{m} \cdot J_\Lambda \cdot J_\theta \cdot \begin{bmatrix} {\omega_1}^2 \\ {\omega_2}^2 \end{bmatrix} \tag{1}$$

where $m$ is the mass of the entire vehicle. $J_\Lambda$ and $J_\theta$ are defined in Equation (2)—(3).

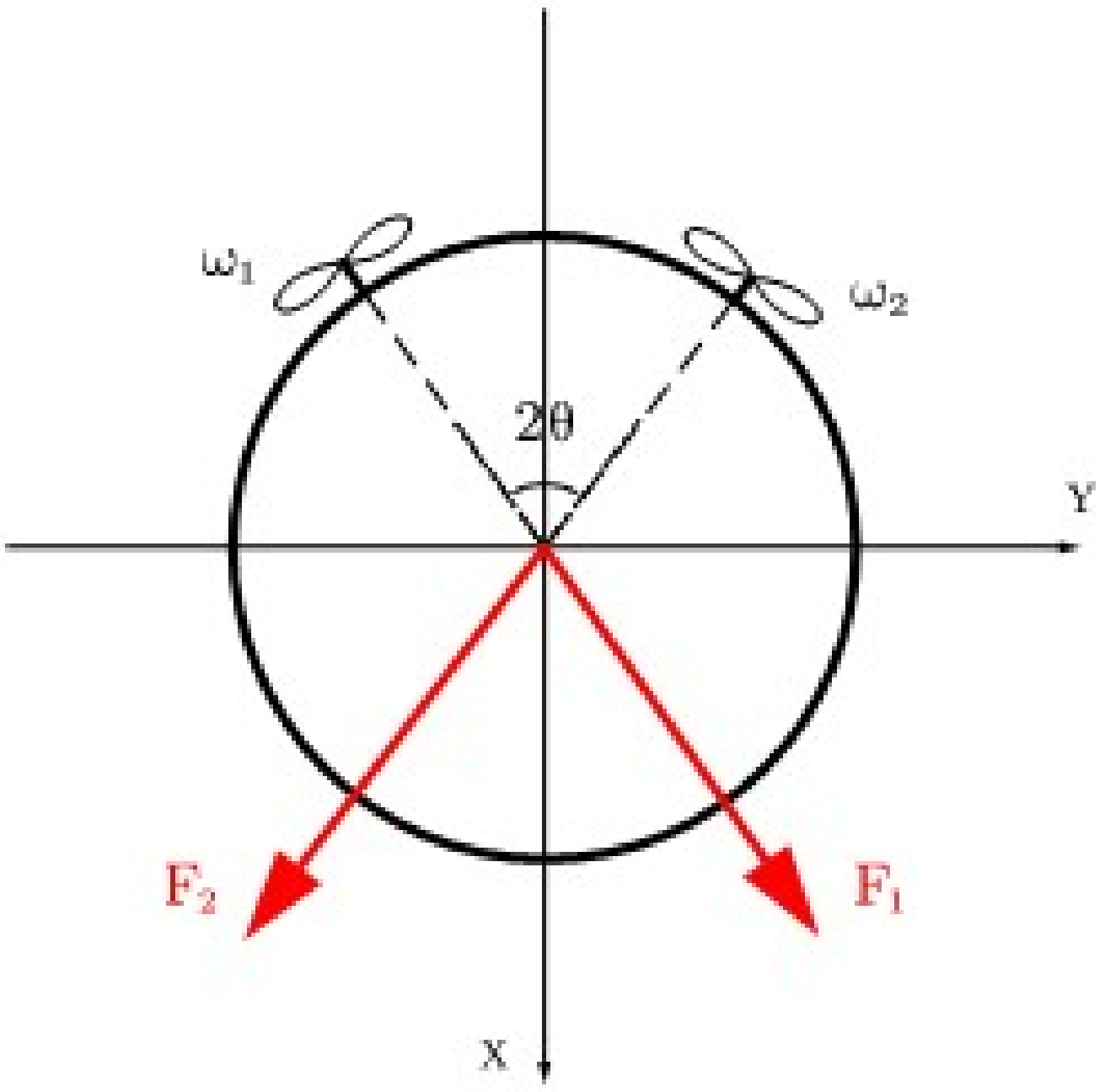

**Figure 1.** Tilt vehicle.

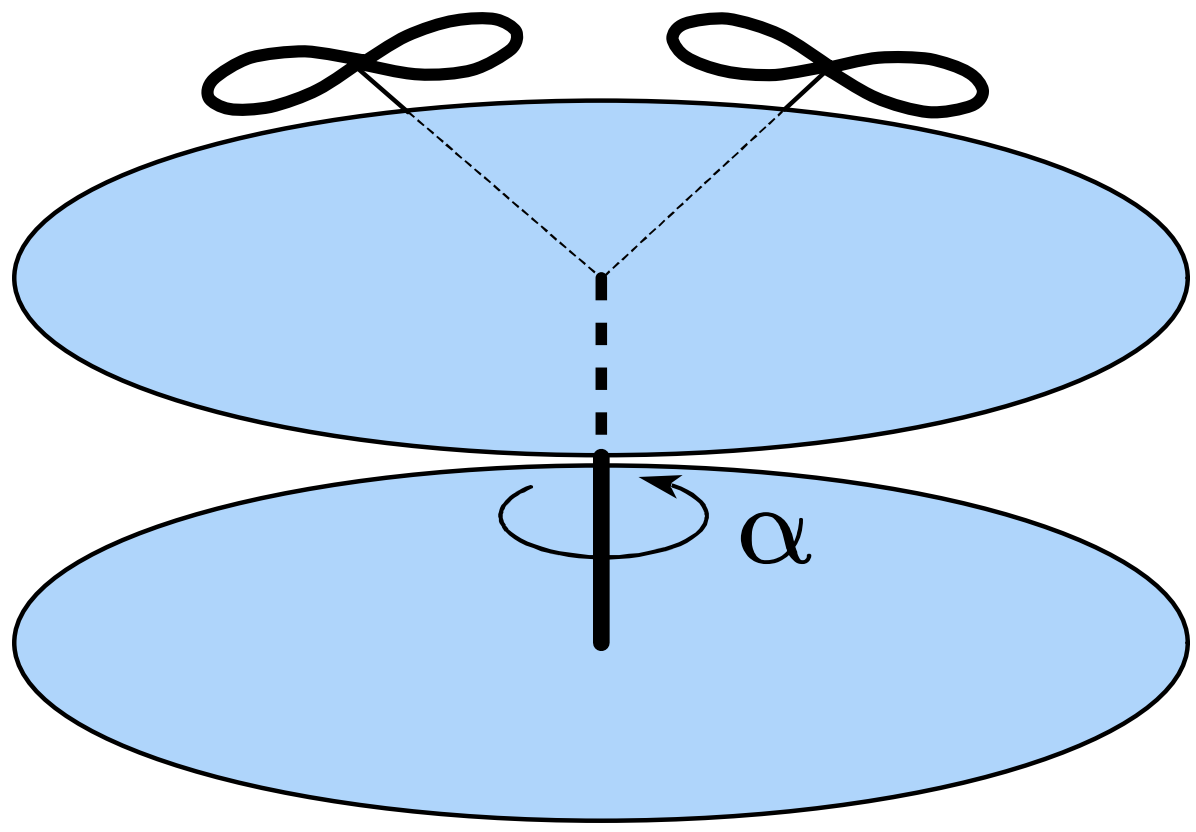

**Figure 2.** The sketch of the tilt vehicle.

$$J_{\Lambda} = \begin{bmatrix} \cos(\Lambda) & -\sin(\Lambda) \\ \sin(\Lambda) & \cos(\Lambda) \end{bmatrix} \tag{2}$$

$$J_{\theta} = \begin{bmatrix} \cos(\theta) & 0 \\ 0 & \sin(\theta) \end{bmatrix} \cdot \begin{bmatrix} K_{F_1} & K_{F_2} \\ K_{F_1} & -K_{F_2} \end{bmatrix} \tag{3}$$

where $\Lambda$ is yaw angle. $\theta$ is the half of the angle between two propellers (see Fig. 1). In our model, we design theta in Equation (4). $K_{F_i}$ is the thrust coefficient (Equation (5)).

$$\theta = \frac{\pi}{6} \tag{4}$$

$$K_{F_1} = K_{F_2} = K = 0.001 \tag{5}$$

A positive $\Lambda$ represents an anti-clockwise yaw (Figure 3). A negative $\Lambda$ represents a clockwise yaw (Figure 4).

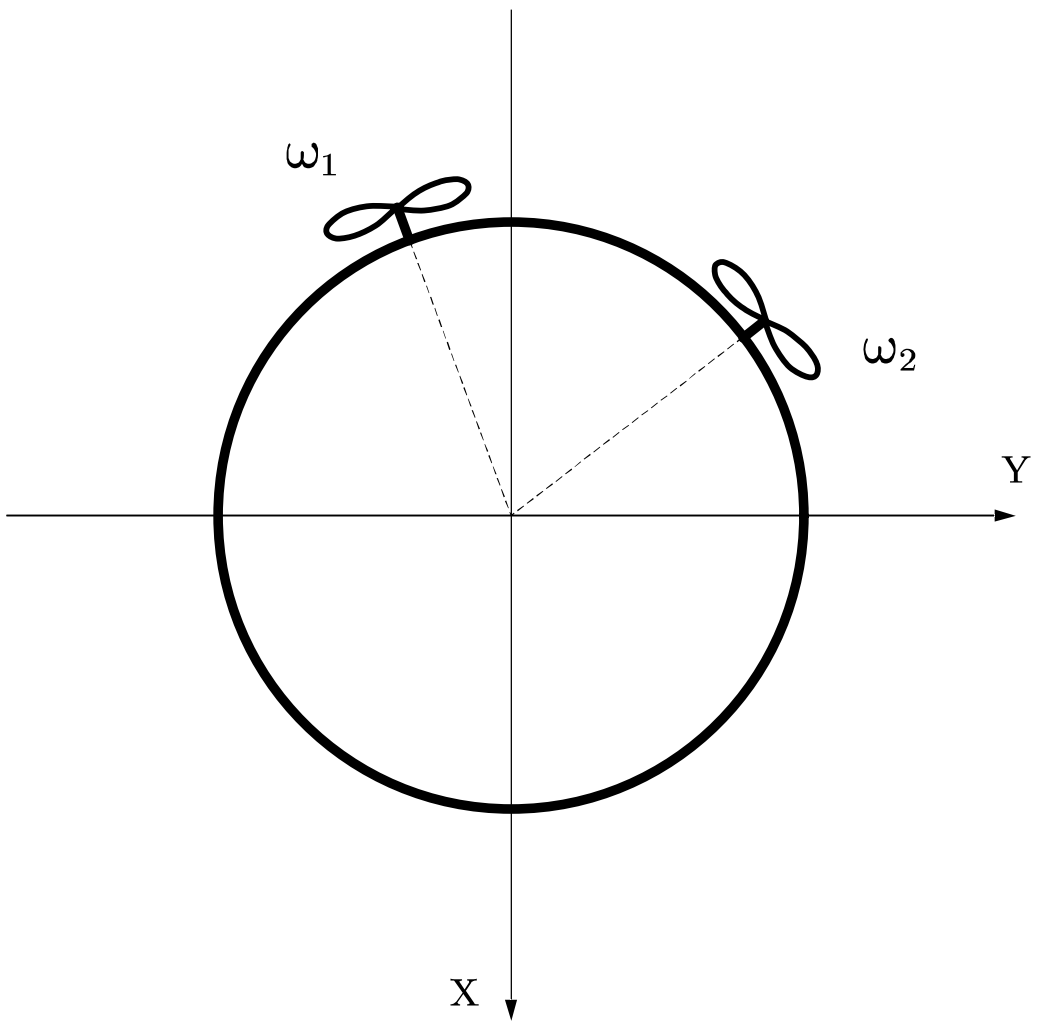

**Figure 3.** Tilt to right.

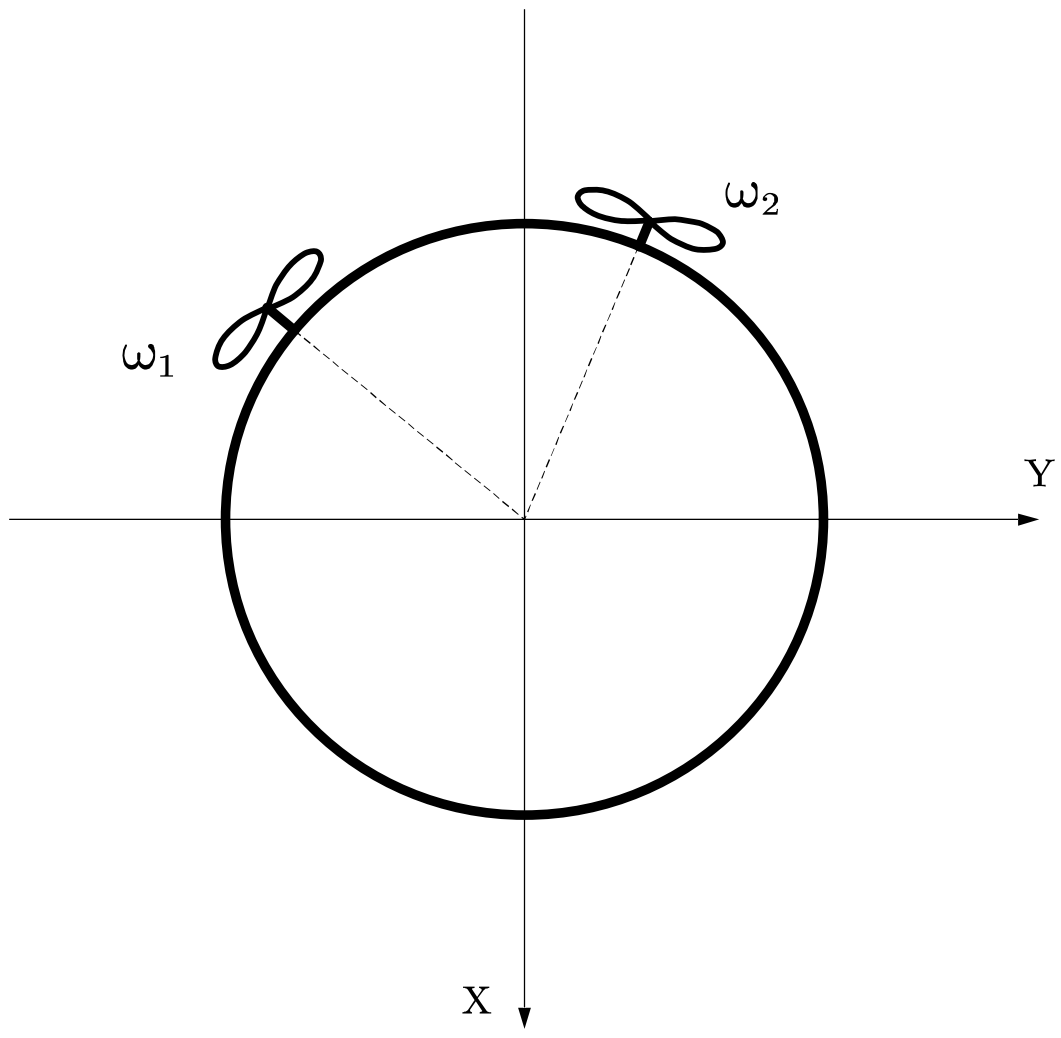

**Figure 4.** Tilt to left.

## 3. Reference and Penguin Gait Inspired Gait Plan

In this study, we aim to solve the straight-line tracking problem. The reference is a straight line with a constant acceleration. It is defined in Equation (6).

$$\begin{cases} x_r = \frac{1}{2} \cdot t^2 \\ y_r = 0 \end{cases} \quad (6)$$

To mimic the walking pattern of a penguin, we create two gaits planned in Figure 5 – Figure 6.

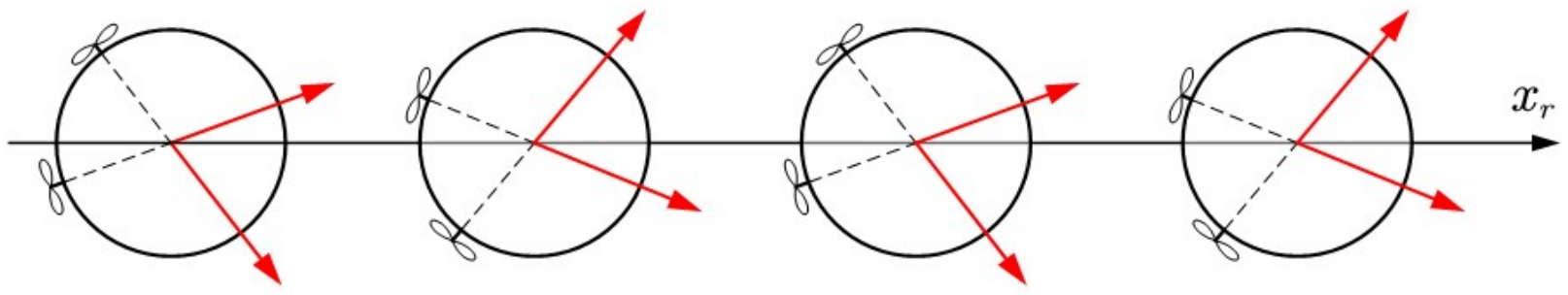

**Figure 5.** Penguin-inspired gait with small swing.

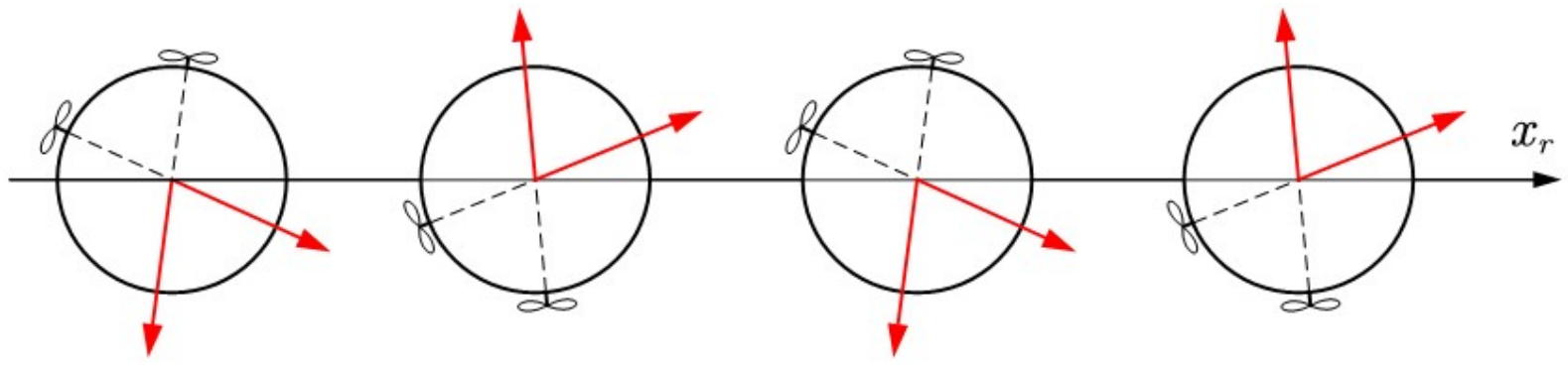

**Figure 6.** Penguin-inspired gait with large swing.

Both Figure 5 and Figure 6 plot the swinging-like gait; the yaw of the tilt vehicle changes periodically. The period of it is 2 seconds.

The changing yaw in Figure 5 changes relatively small, bringing the possibility of maintaining 0 acceleration along $Y - axis$. When changing yaw is large (Figure 6), it is impossible to receive 0 dynamic state error while maintaining 0 acceleration along $Y - axis$.

The $\Lambda - t$ (yaw-time) history for each gait is in Figure 7 ($|\Lambda| = \frac{\pi}{8}$) and Figure 8 ($|\Lambda| = \frac{\pi}{3}$), respectively.

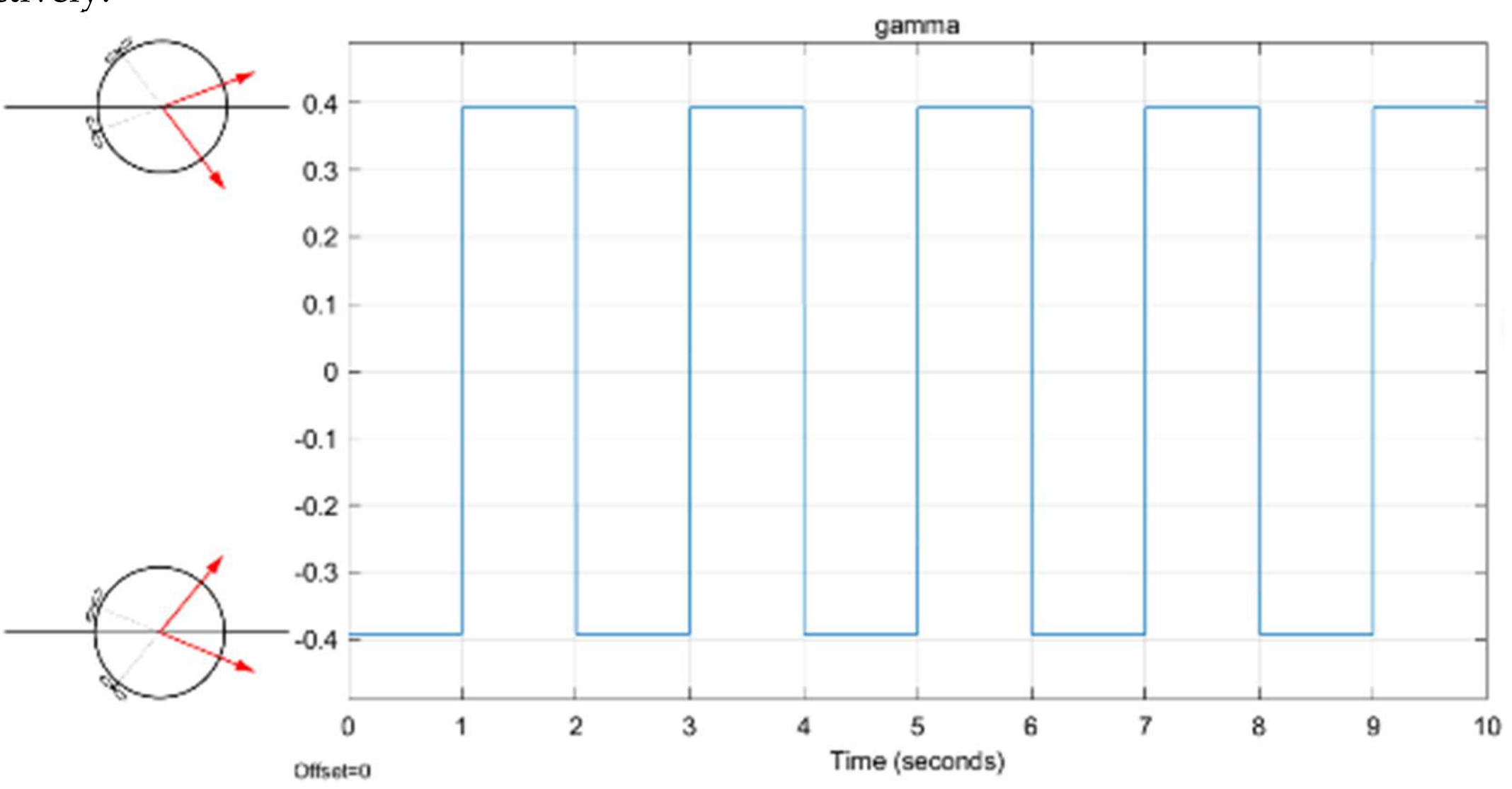

**Figure 7.** Small gait with $|\Lambda| = \frac{\pi}{8}$

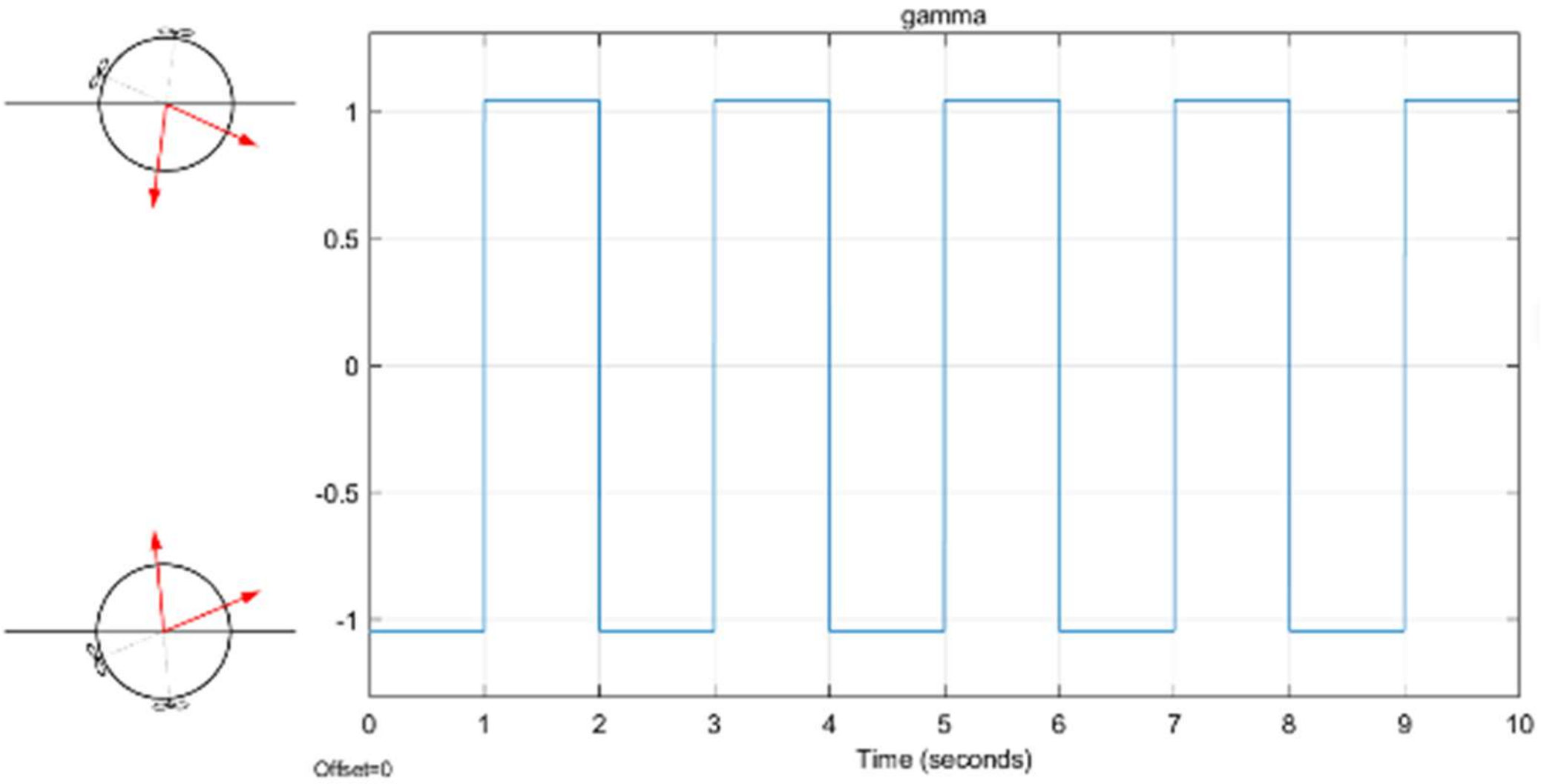

**Figure 8.** Large gait with $|\Lambda| = \frac{\pi}{3}$

The small-gait based controller is discussed in Section 4. While the large-gait based controller is discussed in Section 5 – Section 7.

## 4. Gait and Control without Saturation

In this section, the gait planned in Figure 7 ($|\Lambda| = \frac{\pi}{8}$) is employed. The feedback linearization is applied before applying PD controllers.

### *4.1. Feedback Linearization*

Notice that $J_\Lambda$ and $J_\theta$ in (1) are invertible. We can separate the inputs from the dynamics in (7).

$$\begin{bmatrix} {\omega_1}^2 \\ {\omega_2}^2 \end{bmatrix} = {J_\theta}^{-1} \cdot {J_\Lambda}^{-1} \cdot m \cdot \begin{bmatrix} \ddot{x} \\ \ddot{y} \end{bmatrix} \tag{7}$$

### *4.2. PD Controllers*

Substituting the desired accelerations into (7) yields Equation (8).

$$\begin{bmatrix} {\omega_1}^2 \\ {\omega_2}^2 \end{bmatrix} = {J_\theta}^{-1} \cdot {J_\Lambda}^{-1} \cdot m \cdot \begin{bmatrix} \ddot{x}_d \\ \ddot{y}_d \end{bmatrix} \tag{8}$$

We design the PD controllers based on Equation (8). The controllers are in Equation (9) – Equation (10).

$$\ddot{x}_d = \ddot{x}_r + K_{X_1} \cdot (\dot{x}_r - \dot{x}) + K_{X_2} \cdot (x_r - x) \tag{9}$$

$$\ddot{y}_d = \ddot{y}_r + K_{Y_1} \cdot (\dot{y}_r - \dot{y}) + K_{Y_2} \cdot (y_r - y) \tag{10}$$

where $K_{X_1} = 12$, $K_{X_2} = 6$, $K_{Y_1} = 9$, $K_{Y_2} = 18$.

Substitute Equation (9), (10) into Equation (8) yields Equation (11).

$$\begin{bmatrix}{\omega_1}^2\\{\omega_2}^2\end{bmatrix} = {J_\theta}^{-1}\cdot{J_\Lambda}^{-1}\cdot m\cdot\begin{bmatrix}\ddot{x}_r + K_{X_1}\cdot(\dot{x}_r - \dot{x}) + K_{X_2}\cdot(x_r - x)\\\ddot{y}_r + K_{Y_1}\cdot(\dot{y}_r - \dot{y}) + K_{Y_2}\cdot(y_r - y)\end{bmatrix} \tag{11}$$

However, the right side of Equation (11) is not assured to be nonnegative. While the left side of Equation (11) is the square of the angular velocity, which should be nonnegative.

Thus, rather than direct the right side of Equation (11) to the angular velocities squared, we require zero lower bounds (constraints) at zero. The specific control rule is in Equation (12), (13).

**Practical Control Law (with Saturation)**

$$\begin{bmatrix}squared\omega_1\\squared\omega_2\end{bmatrix} = {J_\theta}^{-1}\cdot{J_\Lambda}^{-1}\cdot m\cdot\begin{bmatrix}\ddot{x}_r + K_{X_1}\cdot(\dot{x}_r - \dot{x}) + K_{X_2}\cdot(x_r - x)\\\ddot{y}_r + K_{Y_1}\cdot(\dot{y}_r - \dot{y}) + K_{Y_2}\cdot(y_r - y)\end{bmatrix} \tag{12}$$

$$\begin{bmatrix}{\omega_1}^2\\{\omega_2}^2\end{bmatrix} = \begin{bmatrix}\max(squared\omega_1, 0)\\\max(squared\omega_2, 0)\end{bmatrix} \tag{13}$$

**Remark 1**

When the right side of Equation (12) is positive ($squared\omega_1 \geqslant 0$, $squared\omega_2 \geqslant 0$), Equation (12), (13) are equivalent to Equation (11).

The rest of the article applies the controller specified in Equation (12), (13).

*4.3. Stability Proof (without Saturation)*

This section presents the proof of the following fact:

**Proposition 1**

If the control law does not reach the saturation bound for the entire time, that is

$$\begin{cases}squared\omega_1(t) \geqslant 0\\squared\omega_2(t) \geqslant 0\end{cases} \quad (t \geqslant 0) \tag{14}$$

Then, $e_x$, $\dot{e}_x$, $e_y$, $\dot{e}_y$ are bounded applying the controller in Equation (12), (13).

**Proof**

Since the control law does not touch the saturation bound, the controller can be reduced to Equation (11).

Substituting Equation (11) into Equation (1) yields Equation (15).

$$\begin{bmatrix}\ddot{x}\\\ddot{y}\end{bmatrix} = \frac{1}{m}\cdot J_\Lambda\cdot J_\theta\cdot{J_\theta}^{-1}\cdot{J_\Lambda}^{-1}\cdot m\cdot\begin{bmatrix}\ddot{x}_r + K_{X_1}\cdot(\dot{x}_r - \dot{x}) + K_{X_2}\cdot(x_r - x)\\\ddot{y}_r + K_{Y_1}\cdot(\dot{y}_r - \dot{y}) + K_{Y_2}\cdot(y_r - y)\end{bmatrix} \tag{15}$$

Rearranging Equation (15) yields Equation (16).

$$\begin{cases}\ddot{e}_x + K_{X_1}\cdot\dot{e}_x + K_{X_2}\cdot e_x = 0\\\ddot{e}_y + K_{Y_1}\cdot\dot{e}_y + K_{Y_2}\cdot e_y = 0\end{cases} \tag{16}$$

where $e_x = x_r - x$, $e_y = y_r - y$.

Define the following Lyapunov candidate:

$$\mathcal{L}_y = \frac{1}{2} \cdot \dot{e}_y^{\,2} + \frac{1}{2} \cdot K_{Y_2} \cdot e_y^{\,2} \tag{17}$$

Differentiating Equation (17) and substituting Equation (16) yield Equation (18).

$$\dot{\mathcal{L}}_y = -K_{Y_1} \cdot \dot{e}_y^{\,2} \leqslant 0 \tag{18}$$

Similarly, define the following Lyapunov candidate:

$$\mathcal{L}_x = \frac{1}{2} \cdot \dot{e}_x^{\,2} + \frac{1}{2} \cdot K_{X_2} \cdot e_x^{\,2} \tag{19}$$

Calculate the derivative of (19). We receive (20).

$$\dot{\mathcal{L}}_x = -K_{X_1} \cdot \dot{e}_x^{\,2} \leqslant 0 \tag{20}$$

Thus, $e_x$, $\dot{e}_x$, $e_y$, $\dot{e}_y$ are bounded.

**Remark 2**

For a LTI system, BIBO stable is equivalent to asymptotic stable. While this result is unnecessary.

Moreover, there are several alternative ways to prove asymptotically stable for this LTI system. We use Lyapunov methods since the Lyapunov candidate in Equation (17) will facilitate to deal with the stability problem in Section 6.

*4.4. Simulation Result*

We simulate the process by Simulink. The Simulink diagram can be found in Figure 9, with several major parts marked.

The dynamic state error ($e_x$, $e_y$) is near zero (less than $10^{-1}$ ) all the time.

The inputs ($\omega_1{}^2$ and $\omega_2{}^2$) are plotted in Figure 10. The inputs change periodically according to the periodical change of the planned gait (yaw signal). Notice that both inputs are positive without touching the saturation bounds. The stability proof holds for this system.

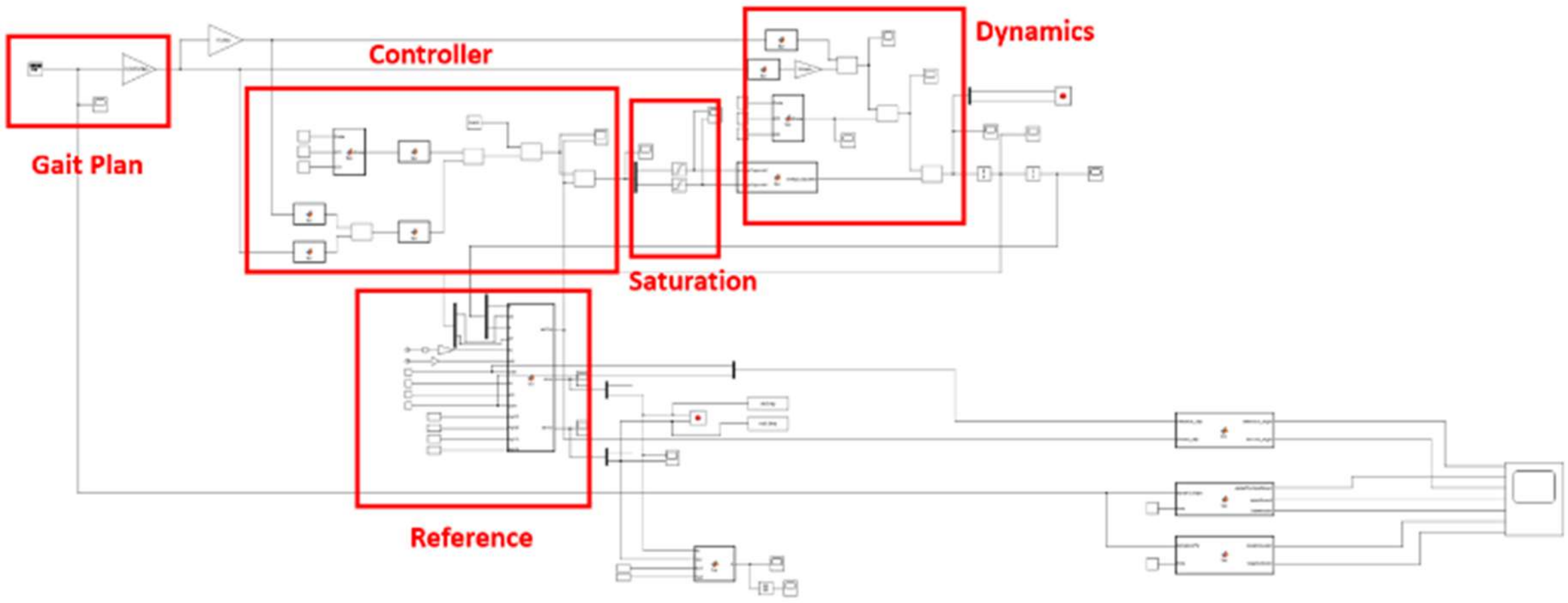

**Figure 9.** The Simulink block diagram.

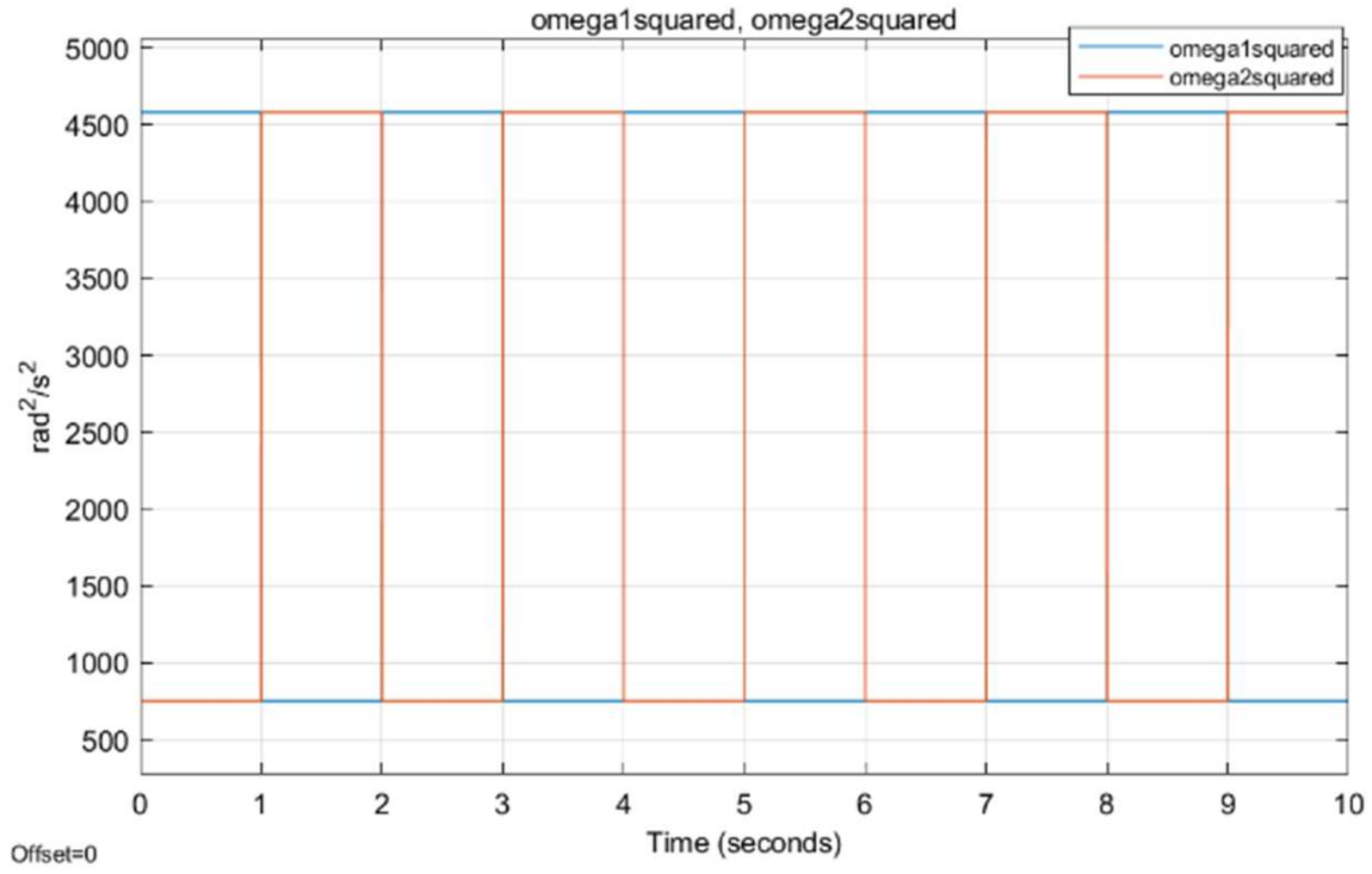

**Figure 10.** The inputs.

## 5. Gait and Control with Saturation

In this section, the gait planned in Figure 8 ($|\Lambda| = \frac{\pi}{3}$) is employed. As explained, this gait introduces the dynamic state error, even if the time is large enough.

We employ the same controller, Practical Control Law, illustrated in Equation (12), (13) to stabilize the tilt vehicle in this gait.

**Proposition 2**

If the gait planned in Figure 8 ($|\Lambda| = \frac{\pi}{3}$) is employed, then, $e_x$, $\dot{e}_x$, $e_y$, $\dot{e}_y$ are bounded applying the controller in Equation (12), (13).

Since the saturation bounds are touched in this gait, Proposition 1 is no longer applicable to this case. In other words, the stability is challenged when saturation appears.

In Section 6, we provide the proof of Proposition 2.

## 6. Proof of Proposition 2

**Lemma 1 (Switch Matrix Based Control Rule)**

The practical control rule described in (12), (13) is equivalent to Equation (21).

$$\begin{bmatrix} {\omega_1}^2 \\ {\omega_2}^2 \end{bmatrix} = S_{pq} \cdot {J_\theta}^{-1} \cdot {J_\Lambda}^{-1} \cdot m \cdot \begin{bmatrix} \ddot{x}_r + K_{X_1} \cdot (\dot{x}_r - \dot{x}) + K_{X_2} \cdot (x_r - x) \\ \ddot{y}_r + K_{Y_1} \cdot (\dot{y}_r - \dot{y}) + K_{Y_2} \cdot (y_r - y) \end{bmatrix} \tag{21}$$

where $S_{pq}$ is called 'Switch Matrix', which is specified in Equation (22).

$$S_{pq} = \begin{bmatrix} p & 0 \\ 0 & q \end{bmatrix} (\quad p{=}0{,}1 \qquad q{=}0{,}1) \tag{22}$$

where $p$ and $q$ are determined in (23), (24).

$$p = \begin{cases} 0, when\ squared\omega_1 \leqslant 0 \\ 1, when\ squared\omega_1 > 0 \end{cases} \tag{23}$$

$$q = \begin{cases} 0, when\ squared\omega_2 \leqslant 0 \\ 1, when\ squared\omega_2 > 0 \end{cases} \tag{24}$$

where $squared\omega_i$ ($i$=1,2) are specified in (12).

**Lemma 2 (Determine Switch Matrix)**

Define ($m$,$n$) in Equation (25).

$$\begin{bmatrix} m \\ n \end{bmatrix} = J_{\Lambda}{}^{-1} \cdot \begin{bmatrix} \ddot{x}_d \\ \ddot{y}_d \end{bmatrix} \tag{25}$$

where $J_{\Lambda}$ is determined in Equation (2). $J_{\Lambda}$ is the rotation transformation of $\Lambda$. In this section, we analyze $|\Lambda| = \frac{\pi}{3}$ (large gait). ($\ddot{x}_d$,$\ddot{y}_d$) is the desired acceleration, which is defined in Equation (26).

$$\begin{bmatrix} \ddot{x}_d \\ \ddot{y}_d \end{bmatrix} = \begin{bmatrix} \ddot{x}_r + K_{X_1} \cdot (\dot{x}_r - \dot{x}) + K_{X_2} \cdot (x_r - x) \\ \ddot{y}_r + K_{Y_1} \cdot (\dot{y}_r - \dot{y}) + K_{Y_2} \cdot (y_r - y) \end{bmatrix} \tag{26}$$

Then, $S_{pq}$ is determined based on ($m$,$n$). The relationship is in Figure 11.

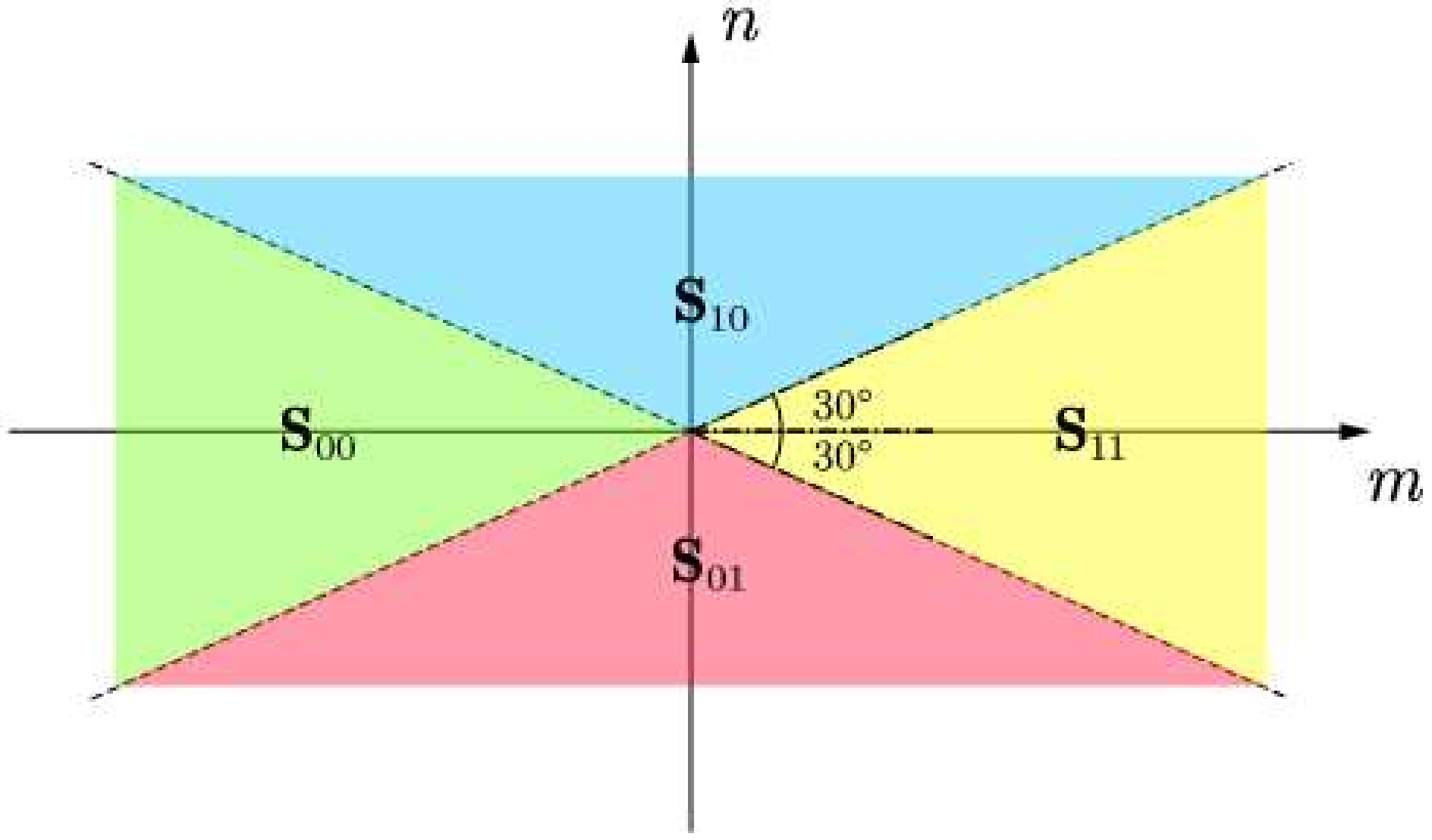

**Figure 11.** Relationship between ($m$,$n$) and $S_{pq}$.

**Proof**

We will determine $S_{pq}$ based on Equation (12), (23) and (24).
Substituting Equation (3) into Equation (12) yields Equation (27).

$$\begin{bmatrix} squared\omega_1 \\ squared\omega_2 \end{bmatrix} = \frac{m}{K} \cdot \begin{bmatrix} \frac{1}{2} & \frac{1}{2} \\ \frac{1}{2} & -\frac{1}{2} \end{bmatrix} \cdot \begin{bmatrix} \frac{1}{\cos(\theta)} & 0 \\ 0 & \frac{1}{\sin(\theta)} \end{bmatrix} \cdot J_{\Lambda}{}^{-1} \cdot \begin{bmatrix} \ddot{x}_d \\ \ddot{y}_d \end{bmatrix} \tag{27}$$

Define ($u$,$v$) in Equation (28).

$$\begin{bmatrix} u \\ v \end{bmatrix} = \begin{bmatrix} \frac{1}{\cos(\theta)} & 0 \\ 0 & \frac{1}{\sin(\theta)} \end{bmatrix} \cdot J_{\Lambda}{}^{-1} \cdot \begin{bmatrix} \ddot{x}_d \\ \ddot{y}_d \end{bmatrix} \tag{28}$$

The relationship between $(u,v)$ and $S_{pq}$ can be identified based on Equation (23), (24), (27) and (28). This result is plotted in Figure 12.

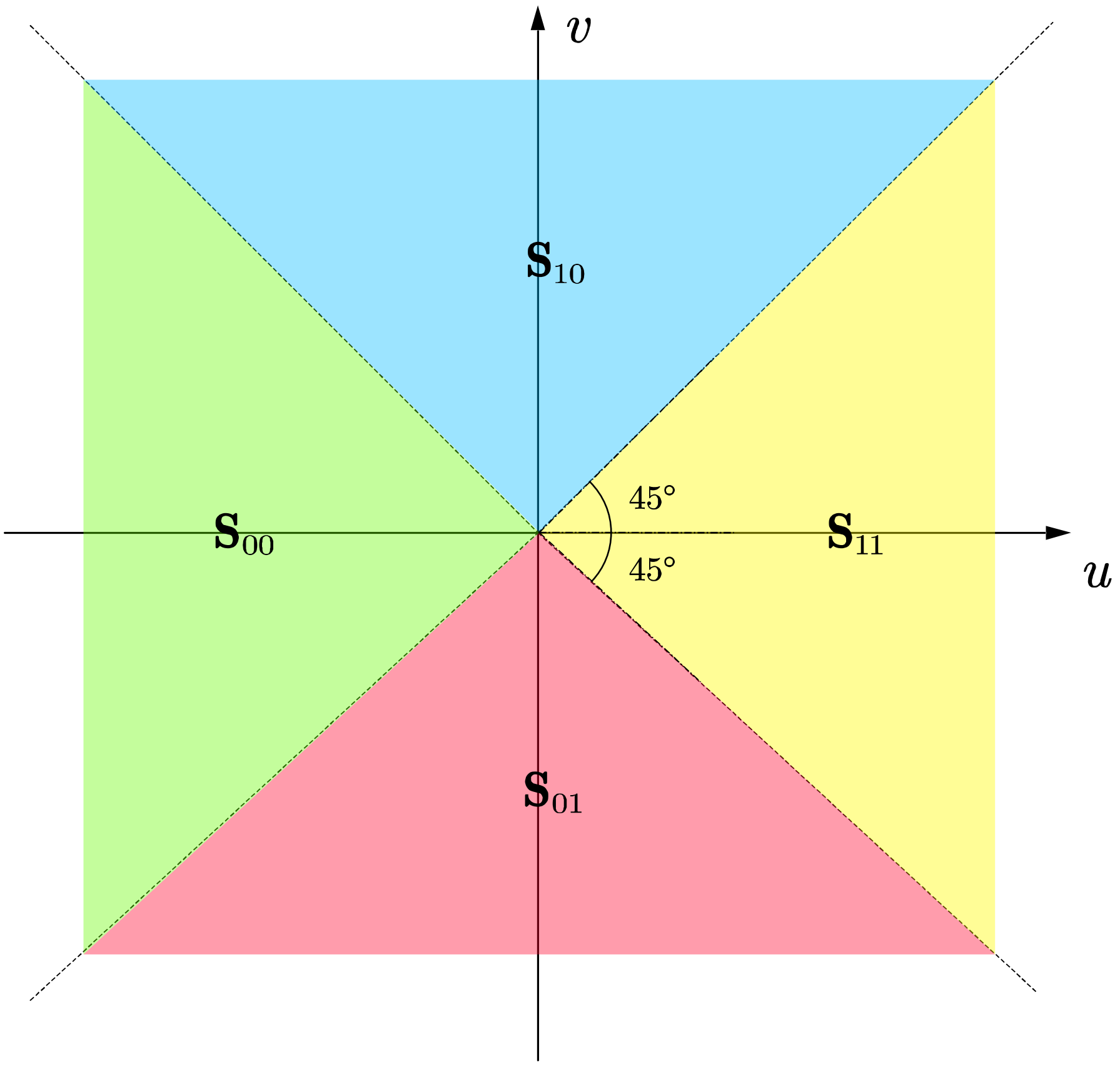

**Figure 12.** Relationship between $(u,v)$ and $S_{pq}$.

Notice the relationship between $(u,v)$ and $(m,n)$ in Equation (29).

$$\begin{bmatrix} u \\ v \end{bmatrix} = \begin{bmatrix} \frac{1}{\cos(\theta)} & 0 \\ 0 & \frac{1}{\sin(\theta)} \end{bmatrix} \cdot \begin{bmatrix} m \\ n \end{bmatrix} \tag{29}$$

Equation (29) is the extension transformation of the coordinates. Its result is exactly Figure 11.

**Lemma 3 (Sufficient Condition for $e_x \equiv 0$)**

$$If \begin{cases} e_x(0) = 0, \dot{e}_x(0) = 0 \\ when \Lambda(t) = -\frac{\pi}{3}, S_{pq} = \{S_{11}, S_{10}\} \\ when \Lambda(t) = \frac{\pi}{3}, S_{pq} = \{S_{11}, S_{01}\} \end{cases} \quad t \in \left[0, n \cdot \frac{T}{2}\right] \quad (n \in \boldsymbol{Z}^{+}, T = 2s),$$

$$then\ e_x \equiv 0, t \in \left[0, n \cdot \frac{T}{2}\right]$$

**Proof**

1° Calculate the dynamic state error in $S_{11}$

Substituting $S_{11}$ and Equation (21) into Equation (1) yields Equation (30).

$$\begin{cases} \ddot{e}_x = -K_{X_1} \cdot \dot{e}_x - K_{X_2} \cdot e_x \\ \ddot{e}_y = -K_{Y_1} \cdot \dot{e}_y - K_{Y_2} \cdot e_y \end{cases} \tag{30}$$

2° Calculate the dynamic state error in $S_{10}$

Substituting $S_{10}$, Equation (21), and $\Lambda(t) = -\frac{\pi}{3}$ into Equation (1) yields Equation (31).

$$\begin{cases} \ddot{e}_x = -K_{X_1} \cdot \dot{e}_x - K_{X_2} \cdot e_x \\ \ddot{e}_y = \frac{1}{\sqrt{3}} \cdot \left(K_{X_2} \cdot e_x + K_{X_1} \cdot \dot{e}_x + 1\right) \end{cases} \tag{31}$$

3° Calculate the dynamic state error in $S_{01}$

Substituting $S_{01}$, Equation (21), and $\Lambda(t) = \frac{\pi}{3}$ into Equation (1) yields Equation (32).

$$\begin{cases} \ddot{e}_x = -K_{X_1} \cdot \dot{e}_x - K_{X_2} \cdot e_x \\ \ddot{e}_y = -\frac{1}{\sqrt{3}} \cdot \left(K_{X_2} \cdot e_x + K_{X_1} \cdot \dot{e}_x + 1\right) \end{cases} \tag{32}$$

Based on the dynamics of $e_x$ in Equation (30) – (32), $e_x \equiv 0$ given that the initial condition is zero.

**Lemma 4 (Iteration in $S_{pq}$)**

$$If\ e_x \equiv 0\,,t \in \left[0{,}n \cdot \frac{T}{2}\right], \quad \left(n \in \boldsymbol{Z}^{+},\ \ T\ \ is\ \ the\ \ period\ \ (2s)\right)$$

$$then \begin{cases} when \Lambda(t) = -\dfrac{\pi}{3}, S_{pq} = \{S_{11}, S_{10}\} \\ when \Lambda(t) = \dfrac{\pi}{3}, S_{pq} = \{S_{11}, S_{01}\} \end{cases}, \qquad t \in \left[n \cdot \frac{T}{2},\ \ (n+1) \cdot \frac{T}{2}\right]$$

**Proof**

Substituting $e_x \equiv 0$ into Equation (26) yields (33).

$$\ddot{x}_d \equiv 1. \tag{33}$$

Notice that $J_\Lambda{}^{-1}$ in Equation (25) is a rotational matrix. It rotates $[\ddot{x}_d \quad \ddot{y}_d]'$ by $-\Lambda$ to $[m \quad n]'$. Equation (33) restricts the position of $[\ddot{x}_d \quad \ddot{y}_d]'$, before the rotation transformation. We present the possible $[\ddot{x}_d \quad \ddot{y}_d]'$ in Figure 13 (red dash line).

1° Find $S_{pq}$ when $\Lambda(t) = -\frac{\pi}{3}$

In this case, we need to rotate $[\ddot{x}_d \quad \ddot{y}_d]'$ by $\frac{\pi}{3}$ $(-\Lambda(t))$ to receive $[m \quad n]'$.
It can be found that $[m \quad n]'$ can only lie in $S_{10}$ or $S_{11}$.

2° Find $S_{pq}$ when $\Lambda(t) = \frac{\pi}{3}$

In this case, we need to rotate $[\ddot{x}_d \quad \ddot{y}_d]'$ by $-\frac{\pi}{3}$ $(-\Lambda(t))$ to receive $[m \quad n]'$.
It can be found that $[m \quad n]'$ can only lie in $S_{01}$ or $S_{11}$.

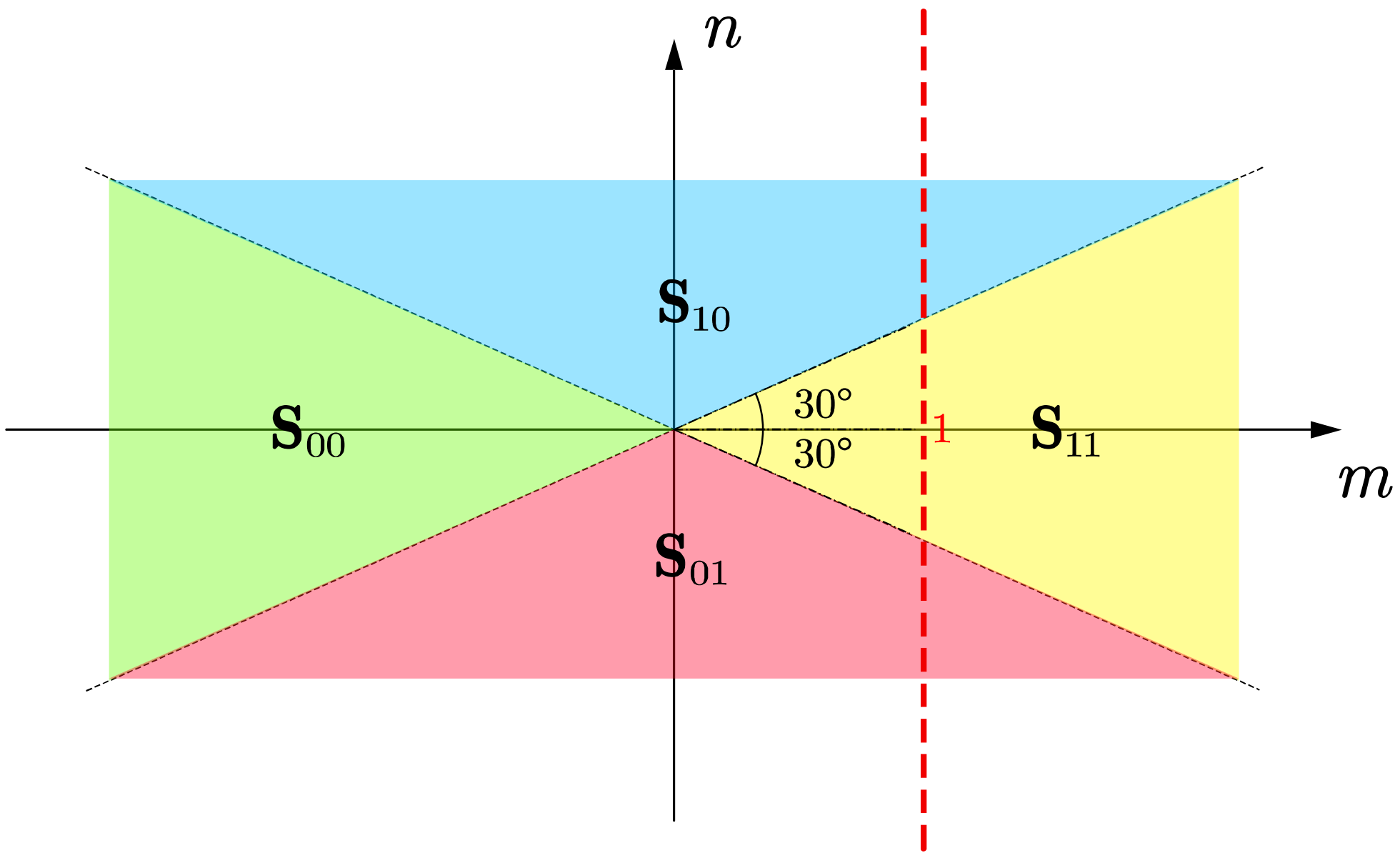

**Figure 13.** The possible positions of $[\ddot{x}_d \quad \ddot{y}_d]'$ are on the red dash line.

**Proposition 3**

$$e_x \equiv 0, \ddot{x}_d \equiv 1$$

**Proof**

Since $e_x(0) = 0, \dot{e}_x(0) = 0$, Proposition 3 is proved by Lemma 3 and Lemma 4.

**Proposition 4**

$$\Lambda(t) = -\frac{\pi}{3}\begin{cases} when K_{Y_1} \cdot \dot{e}_y + K_{Y_2} \cdot e_y \geqslant -\frac{1}{\sqrt{3}}, S_{pq} = S_{10} \\ when K_{Y_1} \cdot \dot{e}_y + K_{Y_2} \cdot e_y < -\frac{1}{\sqrt{3}}, S_{pq} = S_{11} \end{cases} \quad (34)$$

$$\Lambda(t) = \frac{\pi}{3}\begin{cases} when K_{Y_1} \cdot \dot{e}_y + K_{Y_2} \cdot e_y \leqslant \frac{1}{\sqrt{3}}, S_{pq} = S_{01} \\ when K_{Y_1} \cdot \dot{e}_y + K_{Y_2} \cdot e_y > \frac{1}{\sqrt{3}}, S_{pq} = S_{11} \end{cases} \quad (35)$$

**Proof**

1° Find $S_{pq}$ when $\Lambda(t) = -\frac{\pi}{3}$

To receive $S_{10}$, $[\ddot{x}_d \quad \ddot{y}_d]'$ must lie on the red dash line while within the yellow zone or blue zone. In other words, $[\ddot{x}_d \quad \ddot{y}_d]'$ should be above point $A$ while on the red dash line in Figure 14.

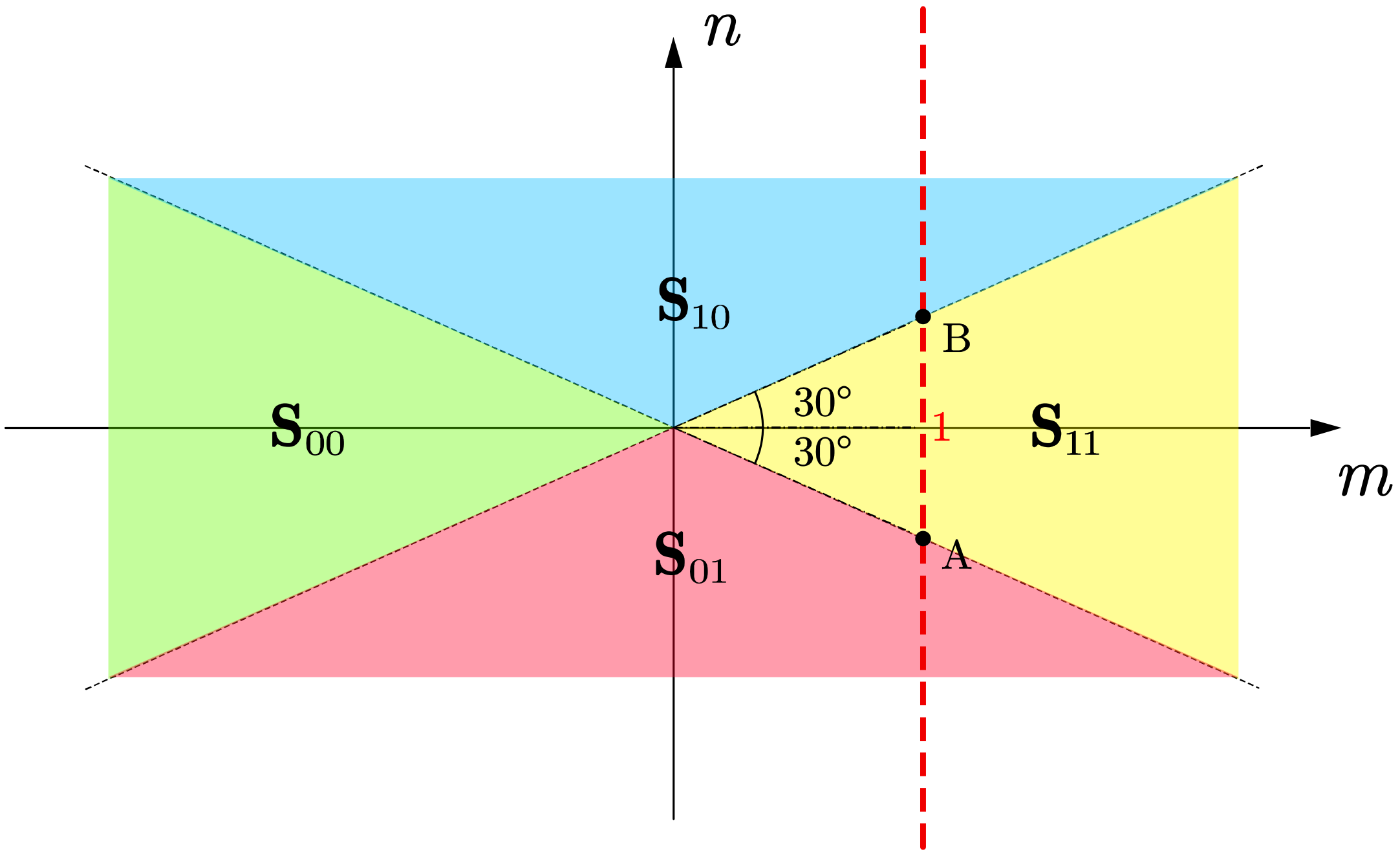

**Figure 14.** The possible positions of $[\ddot{x}_d \quad \ddot{y}_d]'$.

This requirement can be addressed in (36).

$$\ddot{y}_d \geqslant -\frac{1}{\sqrt{3}} \tag{36}$$

Notice that $\ddot{y}_r = 0$. Substituting Equation (26) into Equation (36) yields Equation (37).

$$K_{Y_1} \cdot \dot{e}_y + K_{Y_2} \cdot e_y \geqslant -\frac{1}{\sqrt{3}} \tag{37}$$

Similarly, $[\ddot{x}_d \quad \ddot{y}_d]'$ should be below point $A$ while on the red dash line in Figure 14 to receive $S_{11}$. It yields Equation (38).

$$K_{Y_1} \cdot \dot{e}_y + K_{Y_2} \cdot e_y < -\frac{1}{\sqrt{3}} \tag{38}$$

$2^{\circ}$ Find $S_{pq}$ when $\Lambda(t) = \frac{\pi}{3}$

Similar to the **proof** in $1^{\circ}$, the desired position of $[\ddot{x}_d \quad \ddot{y}_d]'$ now is above or below B in Figure 14. The detail is omitted.

**Remark 3**

Proposition 4 is visualized in Figure 15.

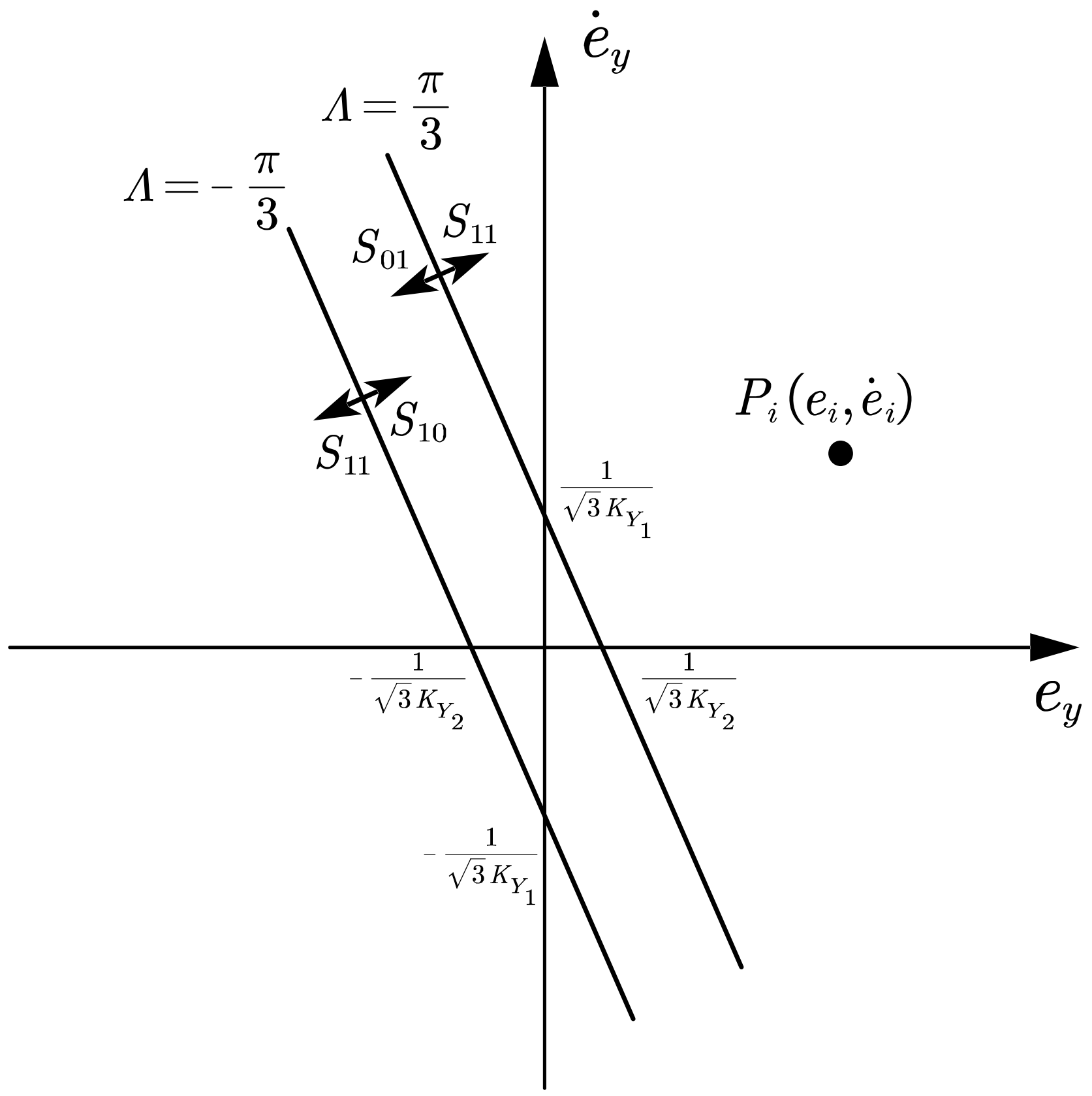

**Figure 15.** Proposition 4.

**Lemma 5**

If the system starts from the initial state point $P(e_y,\dot{e}_y) = P_i(e_i,\dot{e}_i)$ satisfying the conditions in (39)

$$\begin{cases} K_{Y_1} \cdot \dot{e}_i + K_{Y_2} \cdot e_i \geqslant \frac{1}{\sqrt{3}} \\ \dot{e}_i \geqslant 0, e_i \geqslant 0 \\ \Lambda = \frac{\pi}{3} \end{cases} \tag{39}$$

then the state $P(e_y,\dot{e}_y)$ will reach the bound specified in Equation (40) at $t_1$.

$$K_{Y_1} \cdot \dot{e}_y + K_{Y_2} \cdot e_y = \frac{1}{\sqrt{3}} \tag{40}$$

Further, $0 < t_1 < 1$.

**Proof**

Based on Equation (34), the dynamic state error satisfies Equation (41).

$$\ddot{e}_y = -K_{Y_1} \cdot \dot{e}_y - K_{Y_2} \cdot e_y \tag{41}$$

Considering the initial condition $P_i(e_i,\dot{e}_i)$, the solution to Equation (41) is (42).

$$\begin{cases} e_y = \left(2e_i+\frac{1}{3}\dot{e}_i\right) \cdot e^{-3t} + \left(-e_i-\frac{1}{3}\dot{e}_i\right) \cdot e^{-6t} \\ \dot{e}_y = (-6e_i-\dot{e}_i) \cdot e^{-3t} + (6e_i+2\dot{e}_i) \cdot e^{-6t} \end{cases} \tag{42}$$

Substituting Equation (42) into Equation (40) yields Equation (43).

$$\frac{1}{9\sqrt{3}} \cdot e^{6t} + \left(2e_i+\frac{1}{3}\dot{e}_i\right) \cdot e^{3t} - 4e_i - \frac{4}{3}\dot{e}_i = 0 \tag{43}$$

Thus, we have only one positive solution to $e^{3t_1}$. The result is in Equation (44).

$$e^{3t_1} = \frac{-2e_i-\frac{1}{3}\dot{e}_i+\sqrt{\left(2e_i+\frac{1}{3}\dot{e}_i\right)^2+\frac{4}{9\sqrt{3}}\cdot\left(4e_i+\frac{4}{3}\dot{e}_i\right)}}{\frac{2}{9\sqrt{3}}} \tag{44}$$

It can be checked that $0 < t_1 < 1$ if $\dot{e}_i \geqslant 0$, $e_i \geqslant 0$.

**Remark 4**

Since the half of the period $(\frac{T}{2})$ lasts 1 second, there is the extra time $t_2$ ($t_2 = \frac{T}{2} - t_1$) after reaching the bound in (40). The dynamic state error will obey the dynamics where $S_{pq} = S_{01}$ for $t_2$. It will not escape from $S_{pq} = S_{01}$ during $t_2$.

**Lemma 6**

If the system starts from the initial state point $P(e_y,\dot{e}_y) = P_i(e_i,\dot{e}_i)$ satisfying conditions in (45)

$$\begin{cases} K_{Y_1} \cdot \dot{e}_i + K_{Y_2} \cdot e_i \leqslant -\frac{1}{\sqrt{3}} \\ \dot{e}_i \leqslant 0, e_i \leqslant 0 \\ \Lambda = -\frac{\pi}{3} \end{cases} \tag{45}$$

then the state $P(e_y,\dot{e}_y)$ will reach the bound specified in Equation (46) at $t_3$.

$$K_{Y_1} \cdot \dot{e}_y + K_{Y_2} \cdot e_y = -\frac{1}{\sqrt{3}} \tag{46}$$

Further, $0 < t_3 < 1$.

**Proof**

Similar to the Proof of Lemma 5, $e^{3t_3}$ is calculated in Equation (47).

$$e^{3t_3} = \frac{2e_i+\frac{1}{3}\dot{e}_i+\sqrt{\left(2e_i+\frac{1}{3}\dot{e}_i\right)^2-\frac{4}{9\sqrt{3}}\cdot\left(4e_i+\frac{4}{3}\dot{e}_i\right)}}{\frac{2}{9\sqrt{3}}} \tag{47}$$

It can be checked that $0 < t_3 < 1$ if $\dot{e}_i \leqslant 0$, $e_i \leqslant 0$.

**Remark 5**

Since the half of the period $(\frac{T}{2})$ lasts 1 second, there is the extra time $t_4$ ($t_4 = \frac{T}{2} - t_3$) after reaching the bound in Equation (46). The dynamic state error will obey the dynamics where $S_{pq} = S_{10}$ for $t_4$. It will not escape from $S_{pq} = S_{10}$ during $t_4$.

**Lemma 7**

If the system starts from initial state $P(e_y, \dot{e}_y) = P_i(e_i, \dot{e}_i)$ satisfying conditions in (39), then the state $P(e_y, \dot{e}_y)$ will arrive at a state satisfying the conditions in (45) after half period (1 second).

**Proof**

We calculate the state after half period (1 second), $P(e_y, \dot{e}_y)\big|_{t=\frac{T}{2}}$, starting from the initial state $P_i(e_i, \dot{e}_i)$.

During $t_1$ in Equation (44), the dynamic state error obeys equation (30) ($S_{11}$). At time $t_1$, the state reaches the bound defined in Equation (40). After that, the dynamic state error follows Equation (32) ($S_{01}$), which lasts $t_2$ ($t_2 = 1 - t_1$).

The result is plotted in Figure 16. $z-axis$ represents the result of whether $P(e_y, \dot{e}_y)\big|_{t=\frac{T}{2}}$ satisfies the conditions in (45) in Quadrant III. It can be seen that all the initial states, $P_i(e_i, \dot{e}_i)$, satisfying the conditions in (39), meet this requirement.

**Lemma 8**

If the system starts from the initial state point $P(e_y, \dot{e}_y) = P_i(e_i, \dot{e}_i)$ satisfying conditions in (45), then the state $P(e_y, \dot{e}_y)$ will arrive at a state satisfying conditions in (39) after half period (1 second).

**Proof**

Similar to the Proof of Lemma 7, we find that all the initial states satisfying the condition in (45) will arrive at a state satisfying conditions in (39) after half period. It can be checked in Figure 17.

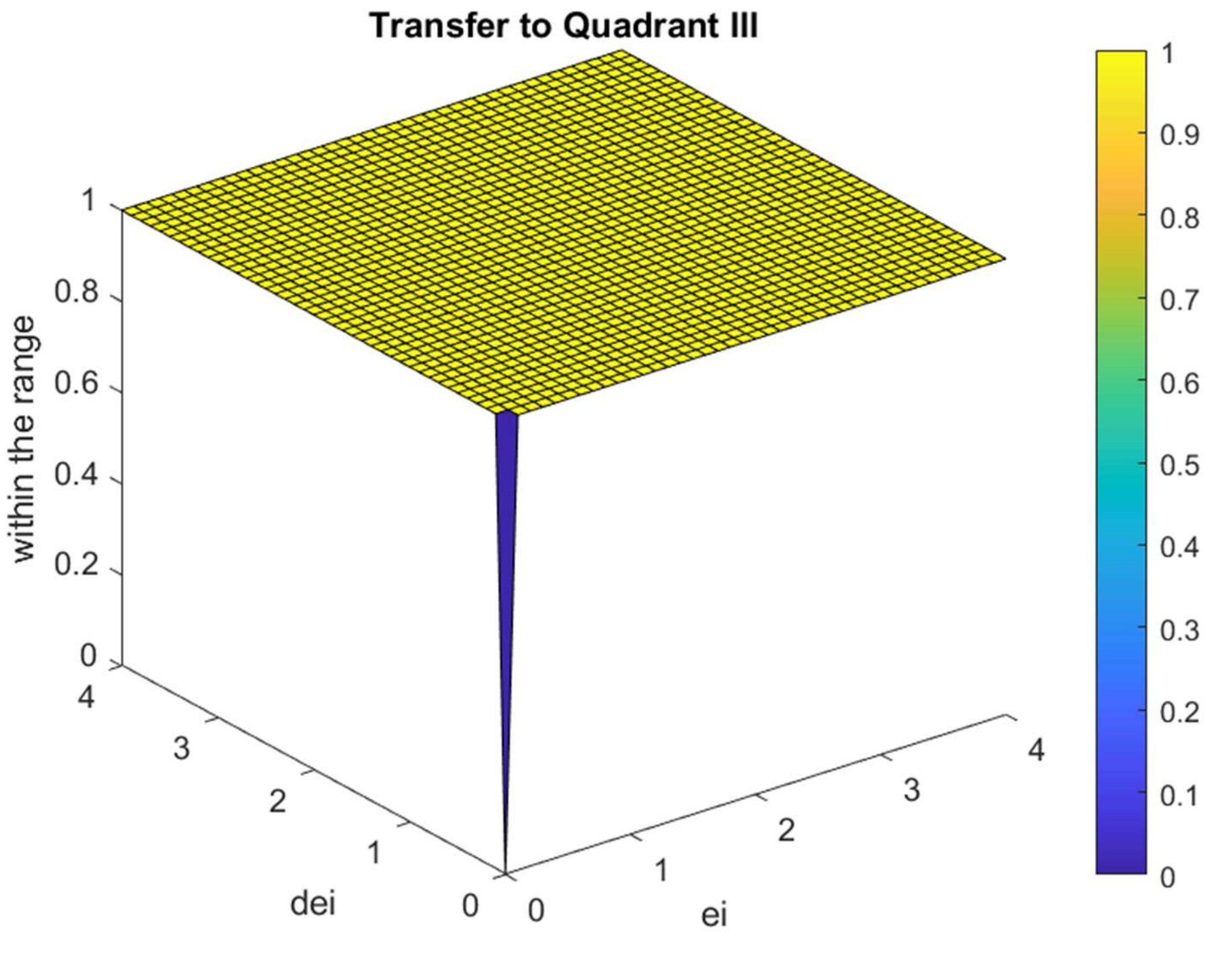

**Figure 16.** Proof of Lemma 7.

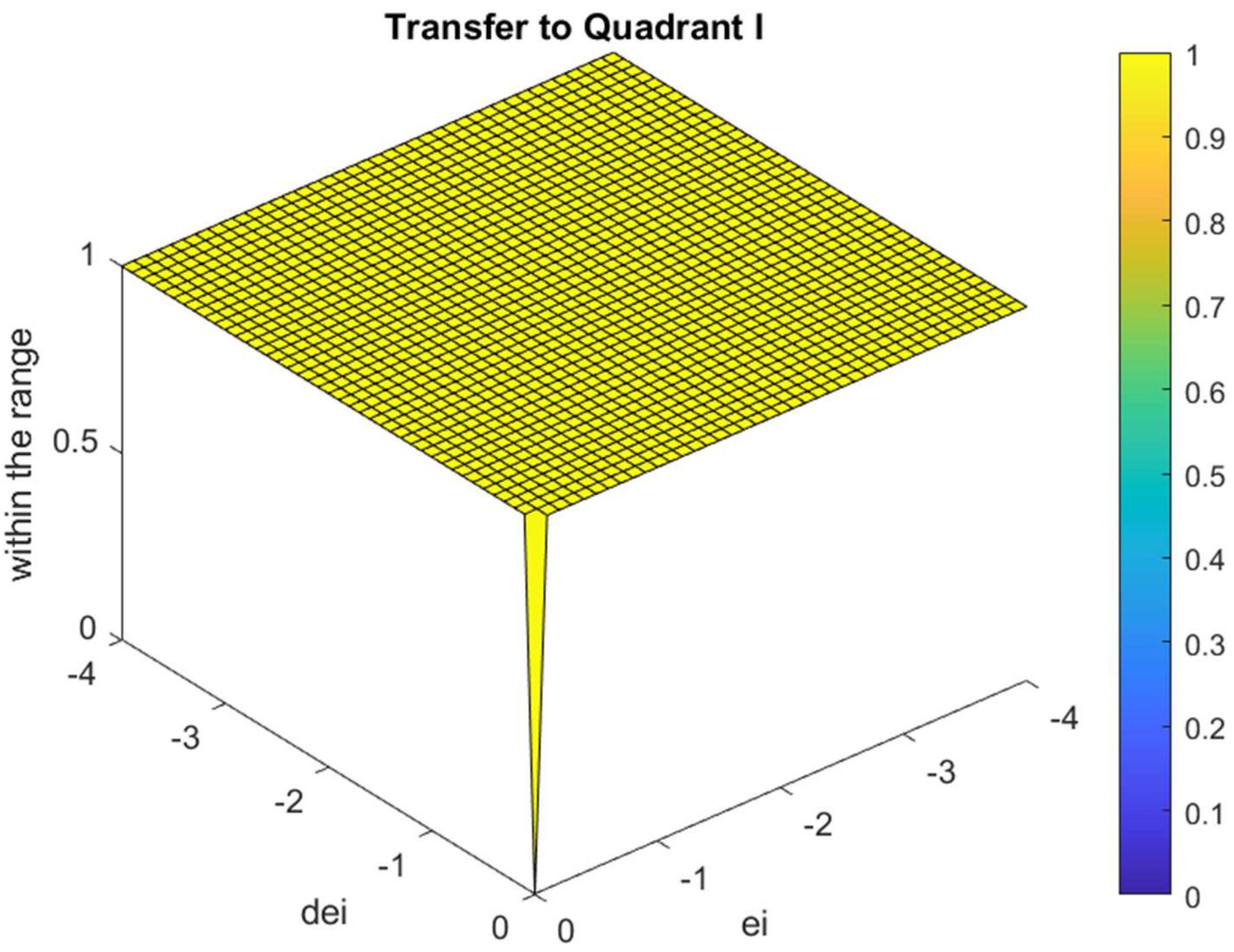

**Figure 17.** Proof of Lemma 8.

**Proposition 5**

If an initial state, $P_i(e_i,\dot{e}_i)$, satisfies the conditions in (48)

$$\begin{cases} K_{Y_1}\cdot\dot{e}_i + K_{Y_2}\cdot e_i \leqslant -\frac{1}{\sqrt{3}} \\ \dot{e}_i \leqslant 0, e_i \leqslant 0 \\ \Lambda = -\frac{\pi}{3} \end{cases} \quad or \quad \begin{cases} K_{Y_1}\cdot\dot{e}_i + K_{Y_2}\cdot e_i \geqslant \frac{1}{\sqrt{3}} \\ \dot{e}_i \geqslant 0, e_i \geqslant 0 \\ \Lambda = \frac{\pi}{3} \end{cases} \tag{48}$$

then $P(e_y,\dot{e}_y)\big|_{t=n\cdot\frac{T}{2}}$ $(n\in\boldsymbol{Z}^+)$ also satisfies the conditions in (48) applying the controller we set.

**Proof**

It can be proved from Lemma 7 and Lemma 8, recursively.

**Remark 6**

Proposition 5 can be explained in Figure 18. Once the initial state is from the yellow area or the blue area, the states after time $t = n\cdot\frac{T}{2}(n\in\boldsymbol{Z}^+)$ will also be inside yellow area or the blue area.

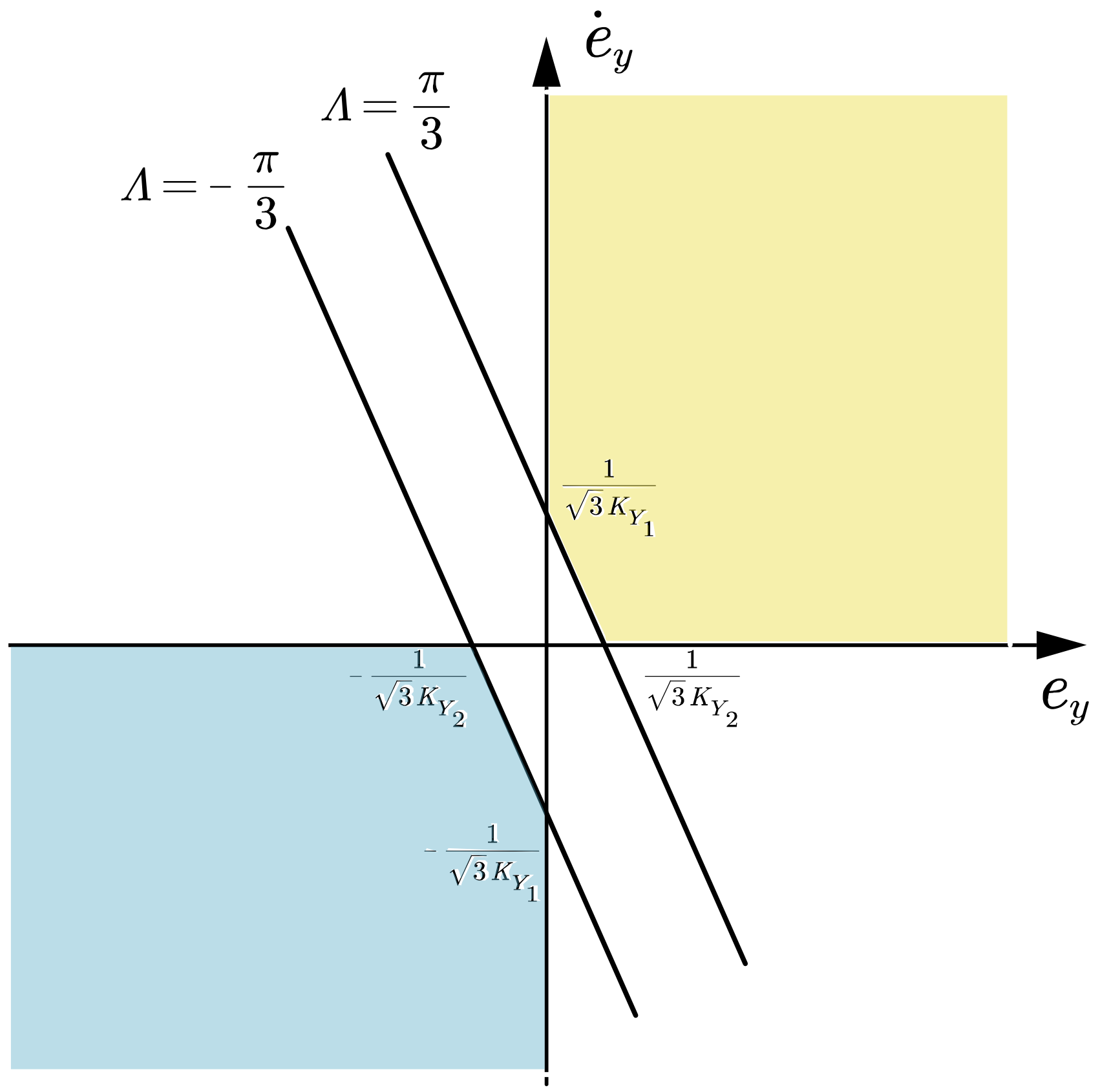

**Figure 18.** The area captures the states (Proposition 5).

**Lyapunov Candidate**

Define the Lyapunov candidate in Equation (49)

$$\mathcal{L} = \frac{1}{2} \cdot \dot{e}_y{}^2 + \frac{1}{2} \cdot K_{Y_2} \cdot e_y{}^2 \tag{49}$$

Instead of proving $\dot{\mathcal{L}}$ is negative/nonpositive (actually it is not), we will prove that $\mathcal{L}$ is bounded in the rest of this section.

**Property of The Defined Lyapunov Candidate**

We analyze the behavior of Lyapunov candidate for each $S_{pq}$.

$1^{\circ}$ $S_{pq} = S_{11}$

Similar to Equation (17), (18), we receive (50).

$$\dot{\mathcal{L}} = -K_{Y_1} \cdot \dot{e}_y{}^2 \leqslant 0 \tag{50}$$

$2^{\circ}$ $S_{pq} = S_{10}$

From (31), considering $e_x \equiv 0$, we receive (51).

$$\begin{cases} \ddot{e}_y &= \frac{1}{\sqrt{3}} \\ \dot{e}_y &= \frac{1}{\sqrt{3}} \cdot t + \dot{e}_{y_0} \\ e_y &= \frac{1}{2\sqrt{3}} \cdot t^2 + \dot{e}_{y_0} \cdot t + e_{y_0} \end{cases} \tag{51}$$

where $e_{y_0}$ and $\dot{e}_{y_0}$ are the initial dynamic state errors (constant).

Thus, the relevant Lyapunov candidate is in (52).

$$\mathcal{L} = \frac{1}{2} \cdot \left(\frac{1}{\sqrt{3}} \cdot t + \dot{e}_{y_0}\right)^2 + \frac{1}{2} \cdot K_{Y_2} \cdot \left(\frac{1}{2\sqrt{3}} \cdot t^2 + \dot{e}_{y_0} \cdot t + e_{y_0}\right)^2 \tag{52}$$

$3^\circ$ $S_{pq} = S_{01}$

From (32), considering $e_x \equiv 0$, we receive (53).

$$\begin{cases} \ddot{e}_y &= -\frac{1}{\sqrt{3}} \\ \dot{e}_y &= -\frac{1}{\sqrt{3}} \cdot t + \dot{e}_{y_0} \\ e_y &= -\frac{1}{2\sqrt{3}} \cdot t^2 + \dot{e}_{y_0} \cdot t + e_{y_0} \end{cases} \tag{53}$$

where $e_{y_0}$ and $\dot{e}_{y_0}$ are the initial dynamic state errors (constant).

Thus, the relevant Lyapunov candidate is in (54).

$$\mathcal{L} = \frac{1}{2} \cdot \left(-\frac{1}{\sqrt{3}} \cdot t + \dot{e}_{y_0}\right)^2 + \frac{1}{2} \cdot K_{Y_2} \cdot \left(-\frac{1}{2\sqrt{3}} \cdot t^2 + \dot{e}_{y_0} \cdot t + e_{y_0}\right)^2 \tag{54}$$

**Proposition 6**

If an initial state, $P_i(e_i, \dot{e}_i)$, satisfies the conditions defined in (55),

$$\begin{cases} K_{Y_1} \cdot \dot{e}_i + K_{Y_2} \cdot e_i \leqslant -\frac{1}{\sqrt{3}} \\ \dot{e}_i \leqslant 0, e_i \leqslant 0 \\ \Lambda = -\frac{\pi}{3} \end{cases} \quad or \quad \begin{cases} K_{Y_1} \cdot \dot{e}_i + K_{Y_2} \cdot e_i \geqslant \frac{1}{\sqrt{3}} \\ \dot{e}_i \geqslant 0, e_i \geqslant 0 \\ \Lambda = \frac{\pi}{3} \end{cases} \tag{55}$$

applying the controller we set, we have the result in (56).

$$\mathcal{L}(t) \leqslant \max\left\{\mathcal{L}\left(n \cdot \frac{T}{2}\right) \quad , \quad \mathcal{L}\left((n+1) \cdot \frac{T}{2}\right)\right\} \tag{56}$$

where $n \in \boldsymbol{Z}^+, t \in \left[n \cdot \frac{T}{2} \quad , \quad (n+1) \cdot \frac{T}{2}\right]$.

**Proof**

$1^\circ$ The initial state, $P_i(e_i, \dot{e}_i)$, satisfies the first condition set in (55)

The state $P(e_y, \dot{e}_y)$ is initially inside the area dominated by the switch matrix $S_{11}$. This phase lasts $t_3$ in (47). The change of the Lyapunov candidate obeys (50).

Thus, we have the result in (57).

$$\mathcal{L}(t) = \mathcal{L}\left(n \cdot \frac{T}{2}\right) + \int_{n \cdot \frac{T}{2}}^{t} \dot{\mathcal{L}}(t) \leqslant \mathcal{L}\left(n \cdot \frac{T}{2}\right) \tag{57}$$

where $n \in \boldsymbol{Z}^+, t \in \left[n \cdot \frac{T}{2} \quad , \quad n \cdot \frac{T}{2} + t_3\right]$

We can also find the minimum Lyapunov candidate during $t \in \left[n\cdot\frac{T}{2} \quad , \quad n\cdot\frac{T}{2}+t_3\right]$ in (58).

$$\min\{\mathcal{L}(t)\} = \mathcal{L}\left(n\cdot\frac{T}{2}+t_3\right), n \in \boldsymbol{Z}^{+}, t \in \left[n\cdot\frac{T}{2} \quad , \quad n\cdot\frac{T}{2}+t_3\right] \tag{58}$$

At time $n \cdot \frac{T}{2} + t_3$, the state $P\left(e_y,\dot{e}_y\right)$ reaches the bound and enters the area dominated by the switch matrix $S_{10}$. For the rest of the time ($t_4 = \frac{T}{2} - t_3$), the Lyapunov candidate follows the rule in (52).

Denote the state $P\left(e_y,\dot{e}_y\right)$ at $n \cdot \frac{T}{2} + t_3$ in (59).

$$\begin{cases} e_y\big|_{t=n\cdot\frac{T}{2}+t_3} = e_M \\ \dot{e}_y\big|_{t=n\cdot\frac{T}{2}+t_3} = \dot{e}_M \end{cases} \tag{59}$$

Substituting (59) into (46) yields (60).

$$\dot{e}_M + 2e_M = -\frac{1}{9\sqrt{3}} \tag{60}$$

Substituting (59), (60) into (54) and calculating its derivative yields (61).

$$\dot{\mathcal{L}} = 9 \cdot \left(\frac{1}{\sqrt{3}}\cdot t+\dot{e}_M\right) \cdot \left(\frac{1}{\sqrt{3}}\cdot t^2+2\cdot\dot{e}_M\cdot t-\dot{e}_M\right) \tag{61}$$

Based on the derivative in Equation (61), we receive the monotone of $\mathcal{L}$ below.

$a.\, when\ \dot{e}_M \in \left(-\infty,-\frac{1}{\sqrt{3}}\right) \cup \left(-\frac{1}{\sqrt{3}},0\right) \cup (0,+\infty)$

During $t_4$, $\mathcal{L}$ decreases monotonically/ $\mathcal{L}$ decreases for a while and increases later.

$b.\, when\ \dot{e}_M = 0$

During $t_4$, $\mathcal{L}$ increases monotonically.

$c.\, when\ \dot{e}_M = -\frac{1}{\sqrt{3}}$

During $t_4$, $\mathcal{L}$ decreases monotonically.

Thus, we make a conclusion in (62).

$$\mathcal{L}(t) \leqslant \max\left\{\mathcal{L}\left(n\cdot\frac{T}{2}+t_3\right) \quad , \quad \mathcal{L}\left((n+1)\cdot\frac{T}{2}\right)\right\} \tag{62}$$

where $n \in \boldsymbol{Z}^{+}, t \in \left[n\cdot\frac{T}{2}+t_3 \quad , \quad (n+1)\cdot\frac{T}{2}\right]$

Substituting the relationship in (58) into (62), we receive the result in (56).

$2^{\circ}$ The initial state, $P_i(e_i,\dot{e}_i)$, satisfies the second condition set in (55)

The state $P\left(e_y,\dot{e}_y\right)$ is initially inside the area dominated by the switch matrix $S_{11}$. This phase lasts $t_1$ in (44). The change of the Lyapunov candidate obeys (50).

Thus, we make a conclusion in (63).

$$\mathcal{L}(t) = \mathcal{L}\left(n\cdot\frac{T}{2}\right) + \int_{n\cdot\frac{T}{2}}^{t} \dot{\mathcal{L}}(t) \leqslant \mathcal{L}\left(n\cdot\frac{T}{2}\right) \tag{63}$$

where $n \in \boldsymbol{Z}^{+}, t \in \left[n\cdot\frac{T}{2} \quad , \quad n\cdot\frac{T}{2}+t_1\right]$.

We also find the minimum Lyapunov candidate during $t \in \left[n\cdot\frac{T}{2} \quad , \quad n\cdot\frac{T}{2}+t_1\right]$ in (64).

$$\min\{\mathcal{L}(t)\} = \mathcal{L}\left(n\cdot\frac{T}{2}+t_1\right) \tag{64}$$

where $n \in \boldsymbol{Z}^{+}, t \in \left[n\cdot\frac{T}{2} \quad , \quad n\cdot\frac{T}{2}+t_1\right]$

At time $n \cdot \frac{T}{2} + t_1$, the state $P\left(e_y,\dot{e}_y\right)$ hits the bound and enters the area dominated by the switch matrix $S_{01}$. For the rest of the time ($t_2 = \frac{T}{2} - t_1$), the Lyapunov candidate follows the rule in (54).

Denote the state $P\left(e_y,\dot{e}_y\right)$ at $n \cdot \frac{T}{2} + t_1$ in (65).

$$\begin{cases} e_y\big|_{t=n\cdot\frac{T}{2}+t_1} = e_m \\ \dot{e}_y\big|_{t=n\cdot\frac{T}{2}+t_1} = \dot{e}_m \end{cases} \tag{65}$$

Substituting (65) into (40) yields (66).

$$\dot{e}_m + 2e_m = \frac{1}{9\sqrt{3}} \tag{66}$$

Substituting (65), (66) into (52) and calculating its derivative yields (67).

$$\dot{\mathcal{L}} = 9 \cdot \left(\frac{1}{\sqrt{3}}\cdot t - \dot{e}_m\right) \cdot \left(\frac{1}{\sqrt{3}}\cdot t^2 - 2\cdot\dot{e}_m\cdot t + \dot{e}_m\right) \tag{67}$$

Based on the derivative in (67), we receive the monotone of $\mathcal{L}$ below.

$a. when\ \dot{e}_m \in (-\infty, 0) \cup \left(0,\frac{1}{\sqrt{3}}\right) \cup \left(\frac{1}{\sqrt{3}},+\infty\right)$

During $t_2$, $\mathcal{L}$ decreases monotonically / $\mathcal{L}$ decreases for a while and increases later.

$b. when\ \dot{e}_m = 0$

During $t_2$, $\mathcal{L}$ increases monotonically.

$c. when\ \dot{e}_m = \frac{1}{\sqrt{3}}$

During $t_2$, $\mathcal{L}$ decreases monotonically.

Thus, we find the following relationship in (68).

$$\mathcal{L}(t) \leqslant \max\left\{\mathcal{L}\left(n\cdot\frac{T}{2}+t_1\right) \quad , \quad \mathcal{L}\left((n+1)\cdot\frac{T}{2}\right)\right\} \tag{68}$$

where $n \in \boldsymbol{Z}^{+}, t \in \left[n\cdot\frac{T}{2}+t_1 \quad , \quad (n+1)\cdot\frac{T}{2}\right]$.

Substituting (64) into (68), we receive (56).

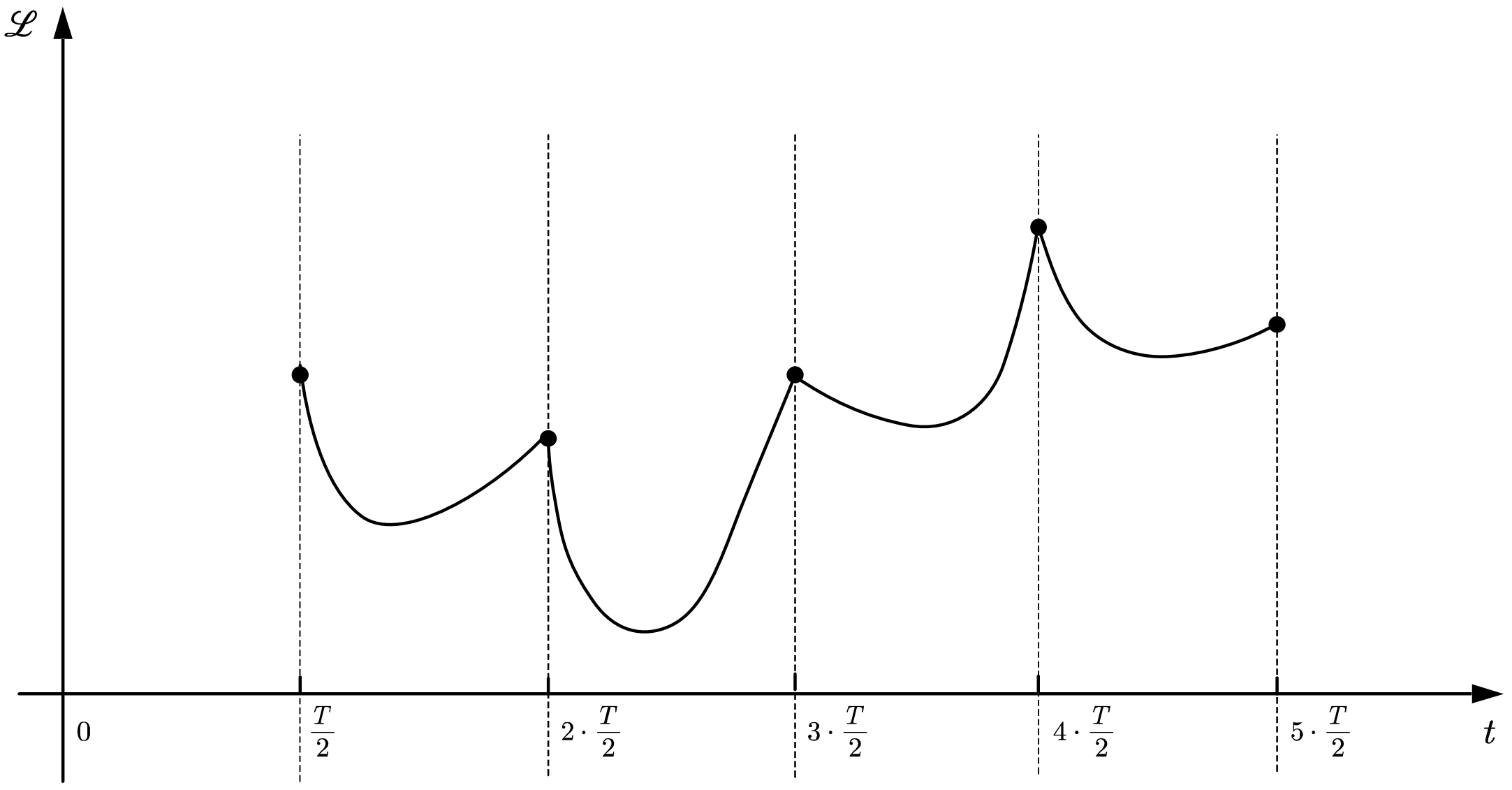

**Figure 19.** Lyapunov candidate at $t = n \cdot \frac{T}{2}$ $(n \in \boldsymbol{Z}^{+})$ is one-side locally maximized

**Remark 7**

Proposition 6 tells us that the Lyapunov candidate is locally maximized at $t = n \cdot \frac{T}{2} (n \in \boldsymbol{Z}^{+})$ if meeting the requirements in (55).

It can be demonstrated in Figure 19. When $t \geqslant \frac{T}{2}$, $\mathcal{L}(t)$ is bounded by $\mathcal{L}\left(n \cdot \frac{T}{2}\right)$, $(n \in \boldsymbol{Z}^{+})$.

**Remark 8 (First Half Period)**

The behavior of the state $P(e_y, \dot{e}_y)$ during $\left[0, \frac{T}{2}\right]$ follows (51), the rule for $S_{10}$, with the initial conditions $e_y\big|_{t=0} = 0$ and $\dot{e}_y\big|_{t=0} = 0$. The Lyapunov candidate in bounded during the first half period. The state $P(e_y, \dot{e}_y)$ is within space defined by (39) at $\frac{T}{2}$.

**Inference 1 (Stability Criteria)**

If $\mathcal{L}\left(n \cdot \frac{T}{2}\right)$, $(n \in \boldsymbol{Z}^{+})$, is upper bounded and $\mathcal{L}(t)$, $\left(t \in \left[0, \frac{T}{2}\right]\right)$, is upper bounded,
then $\mathcal{L}(t)$ is bounded.

**Inference 2 (Lyapunov Candidate Upper Bound)**

If $\mathcal{L}(t)$ is upper bounded,
then we can find the upper bound in (69).

$$\mathcal{L}(t) \leqslant \max_{n \in \boldsymbol{Z}^{+}, n \geqslant C} \left\{\mathcal{L}\left(n \cdot \frac{T}{2}\right)\right\} \tag{69}$$

where $t \geqslant C \cdot \frac{T}{2}$, $C$ is a positive constant.

**Lemma 9**

Define ${}_{\Delta}\mathcal{L}(n)$ in (70).

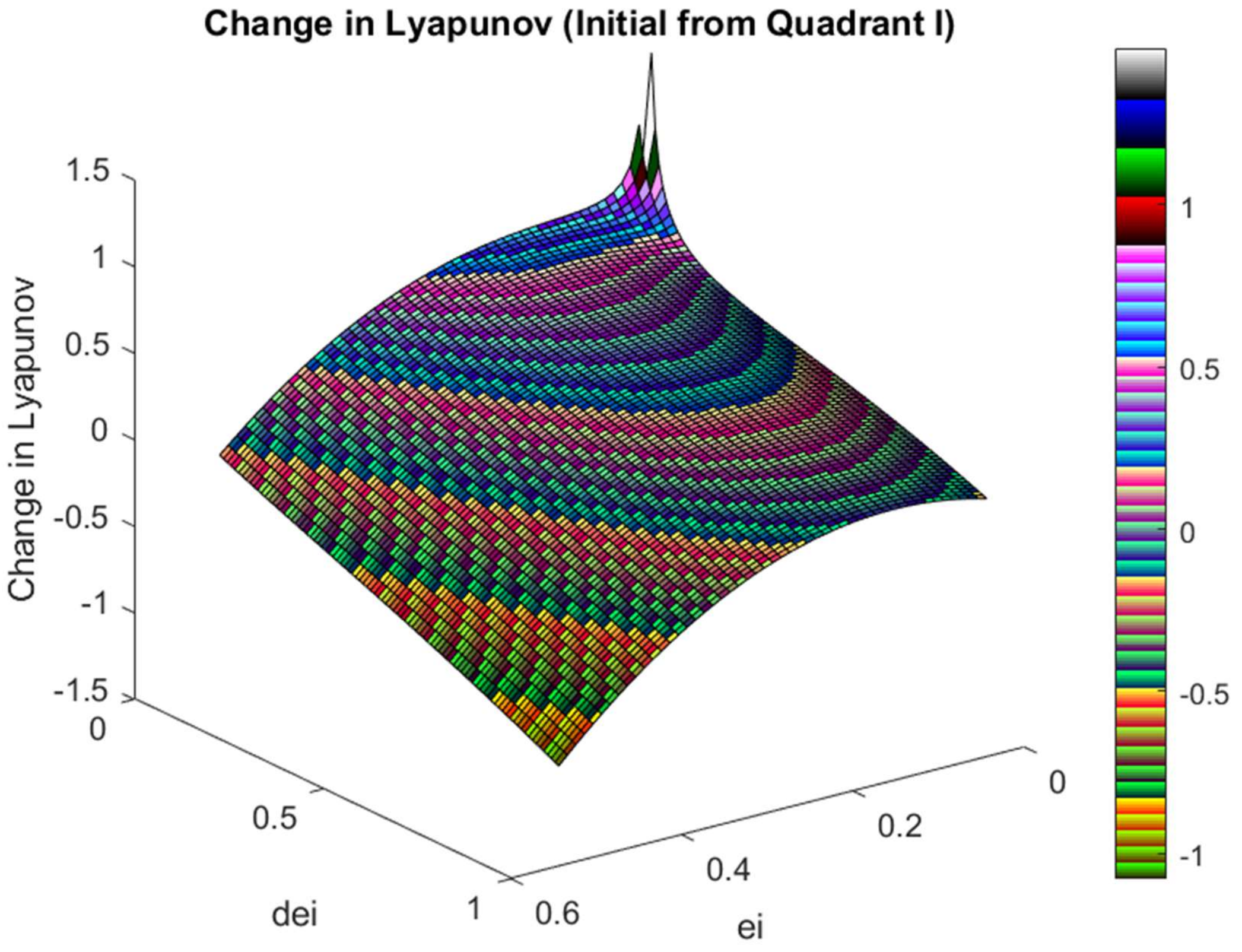

**Figure 20.** $(e_y, \dot{e}_y)\big|_{t=n\cdot\frac{T}{2}}$ in Quadrant I.

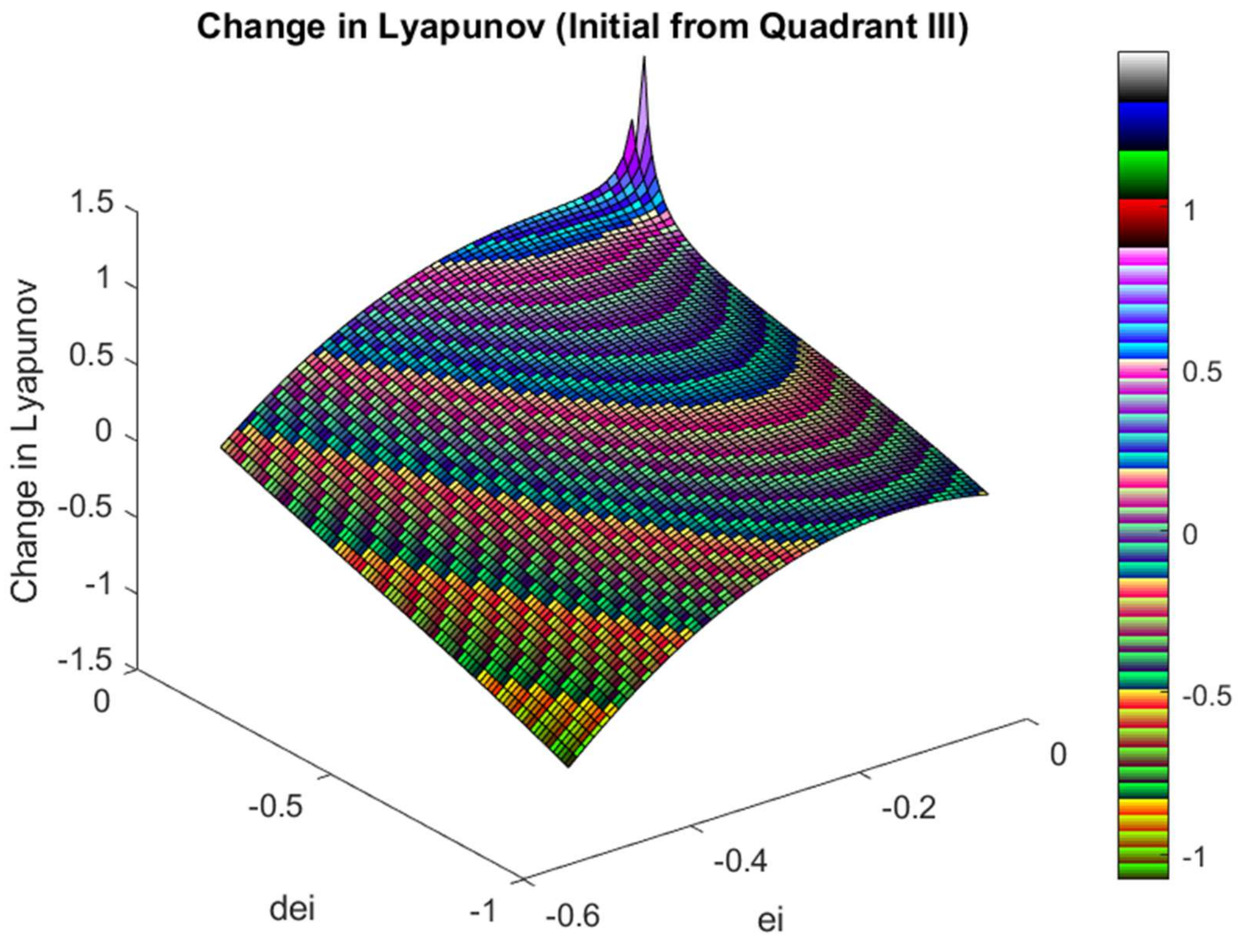

**Figure 21.** $(e_y, \dot{e}_y)\big|_{t=n\cdot\frac{T}{2}}$ in Quadrant III.

$$_{\Delta}\mathcal{L}(n) = \mathcal{L}\left((n+1)\cdot\frac{T}{2}\right) - \mathcal{L}\left(n\cdot\frac{T}{2}\right), (n\in\boldsymbol{Z}^{+}) \tag{70}$$

$_{\Delta}\mathcal{L}(n)$ is determined by $e_y\big|_{t=n\cdot\frac{T}{2}}$ and $\dot{e}_y\big|_{t=n\cdot\frac{T}{2}}$. The relationship between $(e_y,\dot{e}_y)\big|_{t=n\cdot\frac{T}{2}}$ and $_{\Delta}\mathcal{L}(n)$ is plotted in Figure 20, 21.

The cases where $(e_y,\dot{e}_y)\big|_{t=n\cdot\frac{T}{2}}$ is in Quadrant I and Quadrant III are demonstrated in Figure 20 and Figure 21, respectively.

**Remark 9**

Parts of the area in Figure 20 and 21 are unnecessary. The area we are interested in are the colored zones in Figure 18, defined in (48).

One might be depressed by finding positive $_{\Delta}\mathcal{L}(n)$ in some initial conditions $(e_y,\dot{e}_y)\big|_{t=n\cdot\frac{T}{2}}$ from Figure 20 and Figure 21. These initial conditions indicate the failure of proving asymptotic stable by this method. However, this system is not asymptotically stable.

**Remark 10 (Upper Bound of $_{\Delta}\boldsymbol{\mathcal{L}}(\boldsymbol{n})$)**

It can be concluded from Figure 20 and Figure 21 that the maximum value of $_{\Delta}\mathcal{L}(n)$ is either on the bound defined in (40) or on the bound defined in (46).

Calculating the maximum of $_{\Delta}\mathcal{L}(n)$, we receive the same result on either bound. That is (71).

$$_{\Delta}\mathcal{L}(n) \leqslant \frac{3}{4} \tag{71}$$

**Remark 11 (Graph of Positive $_{\Delta}\boldsymbol{\mathcal{L}}(\boldsymbol{n})$)**

We can also plot figure 20 and Figure 21 in 2 dimensions. The 2-D graph are in Figure 22 and 23.

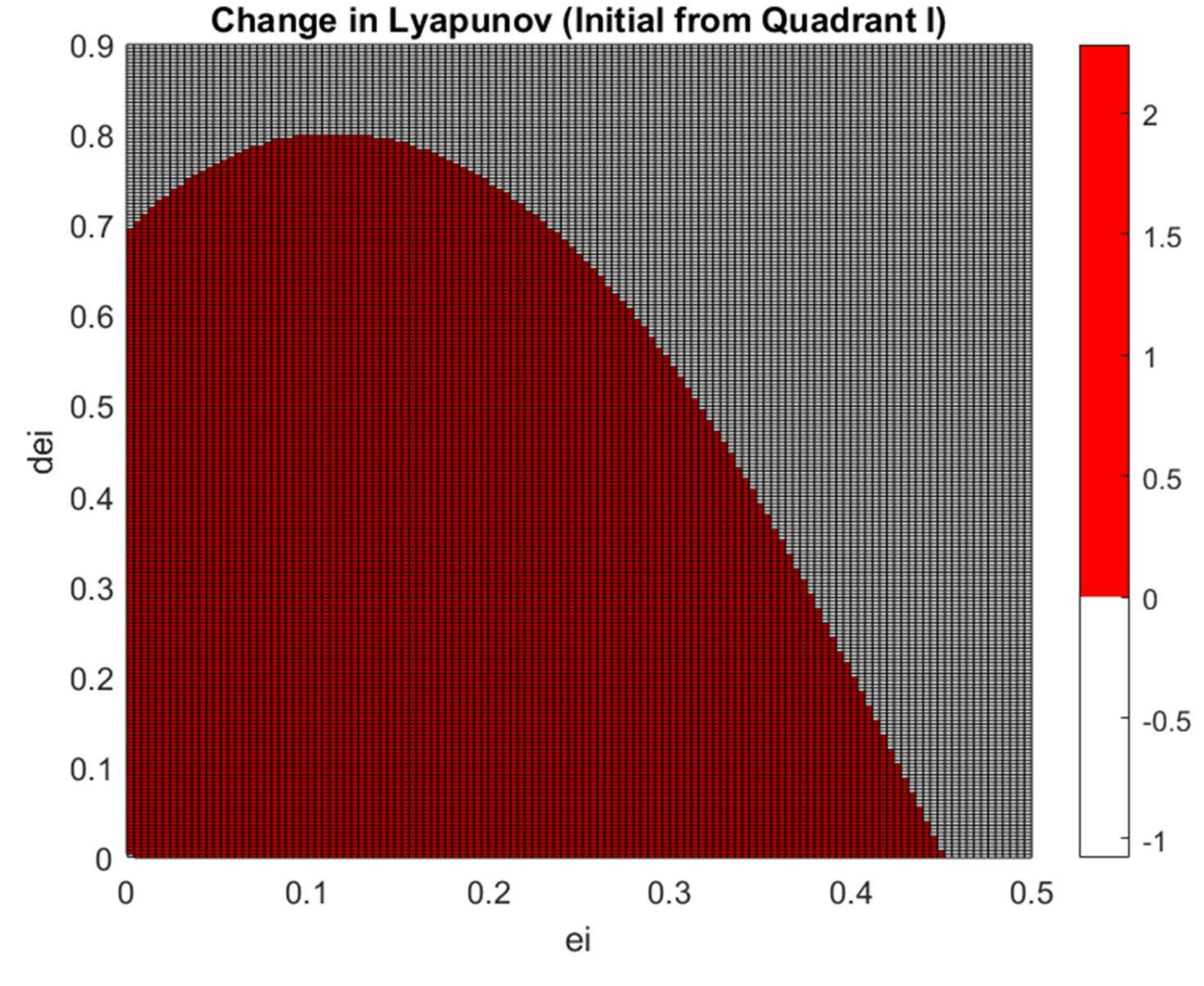

**Figure 22.** $(e_y,\dot{e}_y)\big|_{t=n\cdot\frac{T}{2}}$ in Quadrant I.

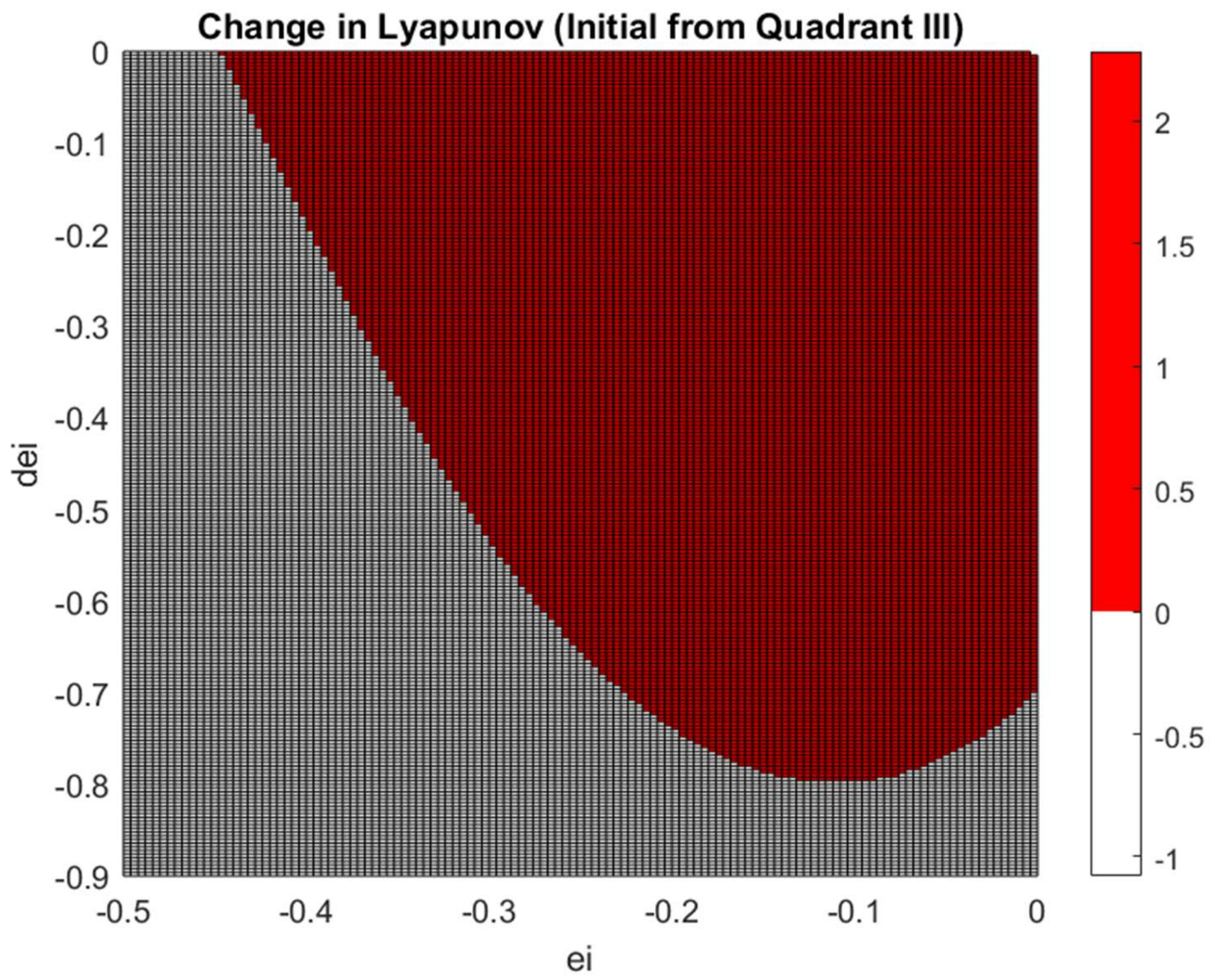

**Figure 23.** $(e_y,\dot{e}_y)|_{t=n\cdot\frac{T}{2}}$ in Quadrant III.

The red area in Figure 22 and 23 is the $(e_y,\dot{e}_y)|_{t=n\cdot\frac{T}{2}}$ receiving a positive ${}_{\Delta}\mathcal{L}(n)$, increasing $\mathcal{L}\left(n\cdot\frac{T}{2}\right),(n\in\boldsymbol{Z}^{+})$. The white area, on the other hand, receives a negative ${}_{\Delta}\mathcal{L}(n)$, decreasing $\mathcal{L}\left(n\cdot\frac{T}{2}\right),(n\in\boldsymbol{Z}^{+})$.

**Main Proof**

Since the first half period (See Remark 8) is bounded and lets the $(e_y,\dot{e}_y)|_{t=\frac{T}{2}}$ be inside the zone defined by (39), $(e_y,\dot{e}_y)|_{t=n\cdot\frac{T}{2}}$ will always be inside (See Remark 6) the zone defined by (55), which is the colored area in Figure 18.

The $\mathcal{L}\left(n\cdot\frac{T}{2}\right),(n\in\boldsymbol{Z}^{+})$ defined is bounded. The reason is given below:

$1^{\circ}$ When $(e_y,\dot{e}_y)|_{t=i\cdot\frac{T}{2}},(i\in\boldsymbol{Z}^{+})$ drops inside the white area in Figure 22 or Figure 23.

${}_{\Delta}\mathcal{L}(i)<0$ (See Remark 11). Thus, we have (72).

$$(e_y,\dot{e}_y)|_{t=i\cdot\frac{T}{2}}<(e_y,\dot{e}_y)|_{t=(i+1)\cdot\frac{T}{2}} \quad (72)$$

$2^{\circ}$ When $(e_y,\dot{e}_y)|_{t=i\cdot\frac{T}{2}},(i\in\boldsymbol{Z}^{+})$ drops inside the red area in Figure 22 or Figure 23.

${}_{\Delta}\mathcal{L}(i)>0$ (See Remark 11). It tries to push $(e_y,\dot{e}_y)|_{t=(i+1)\cdot\frac{T}{2}}$ outside the red area.

Notice that ${}_{\Delta}\mathcal{L}(i)$ has an upper bound (See Remark 10). It ensures that $(e_y,\dot{e}_y)|_{t=(i+1)\cdot\frac{T}{2}}$ is bounded for this case.

3° When $(e_y,\dot{e}_y)\big|_{t=i\cdot\frac{T}{2}}$, $(i\in\boldsymbol{Z}^+)$ drops inside the rest area where $_\Delta\mathcal{L}(i) = 0$.

$$(e_y,\dot{e}_y)\big|_{t=(i+1)\cdot\frac{T}{2}} = (e_y,\dot{e}_y)\big|_{t=i\cdot\frac{T}{2}}.$$

Since $(e_y,\dot{e}_y)\big|_{t=\frac{T}{2}}$ is bounded, $\mathcal{L}\left(n\cdot\frac{T}{2}\right)$, $(n\in\boldsymbol{Z}^+)$ is also bounded, proved recursively by 1°, 2°, and 3° above.

From Inference 1, we can make the conclusion — $\mathcal{L}(t)$ is bounded.
The proof is completed.

## 7. Simulation Results

The dynamic state error result is in Figure 24. The dynamic state error along $x-axis$ remains zero. The dynamic state error along $y-axis$ is not stabilized at zero. Although it is not asymptotically stable, the dynamic state error along $y-axis$ is bounded.

Figure 25 displays the history of inputs. It can be seen that the inputs change periodically in general. The period is identical to the period ($T$) of the gait planned. Saturation happens from time to time. While there are the time windows where no saturation appears.

Define the angle of acceleration $(\mathcal{U},\mathcal{V})$ in (73). Notice the angle is converted into the range $[0,2\pi]$.

$$angle\ of\ (\mathcal{U},\mathcal{V}) = 0\ to\ 2\pi\{\text{atan2}(\mathcal{V},\mathcal{U})\} \tag{73}$$

We can also check whether the saturation happens in Figure 26. The inputs for the tilt vehicle avoid the saturation if and only if the yellow curve, which represents the desired angle of acceleration, is between the purple curve, which represents the upper bound of the possible angle of acceleration, and the green curve, which represents the lower bound of the possible angle of acceleration.

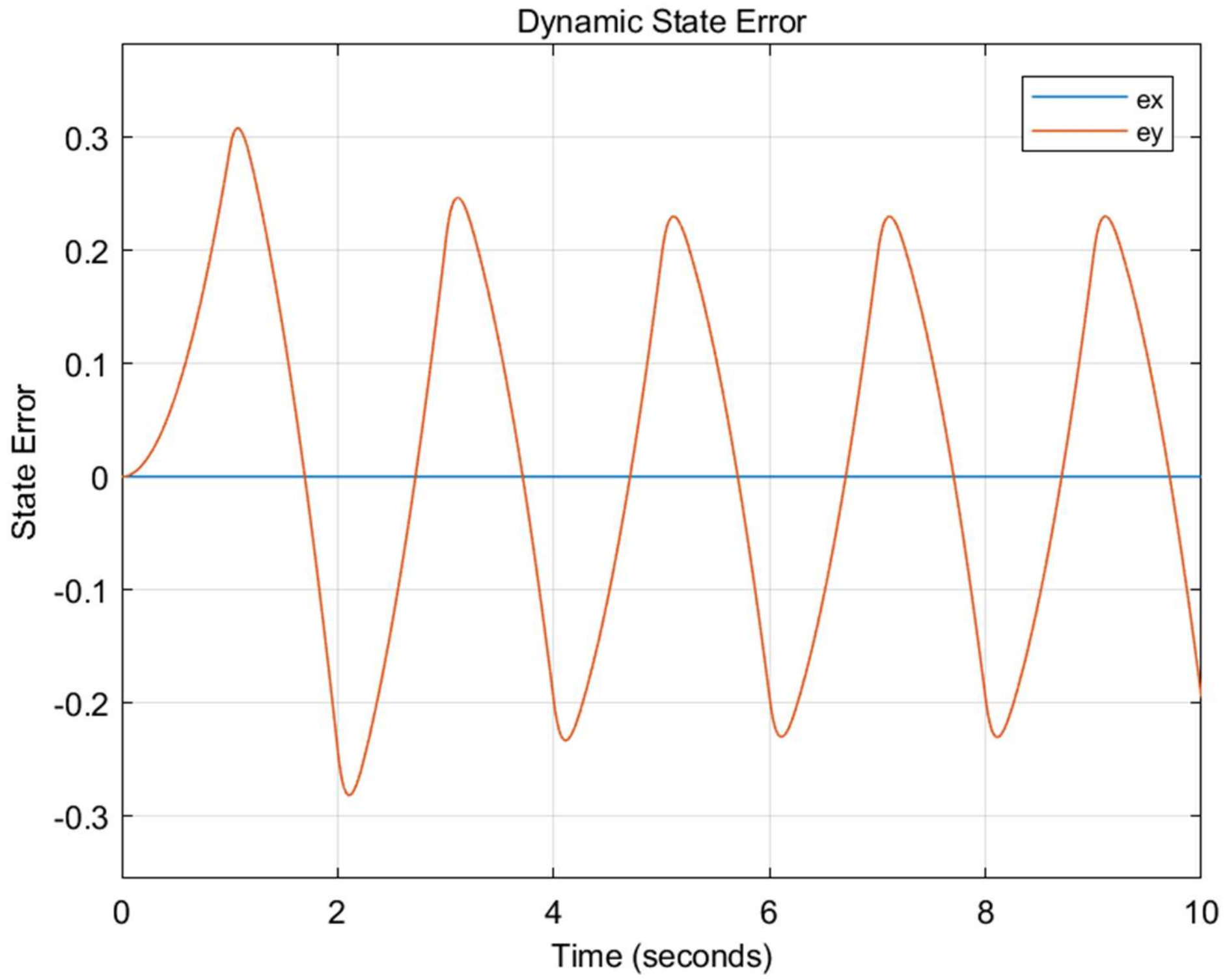

**Figure 24.** Dynamic state errors in 2 dimensions.

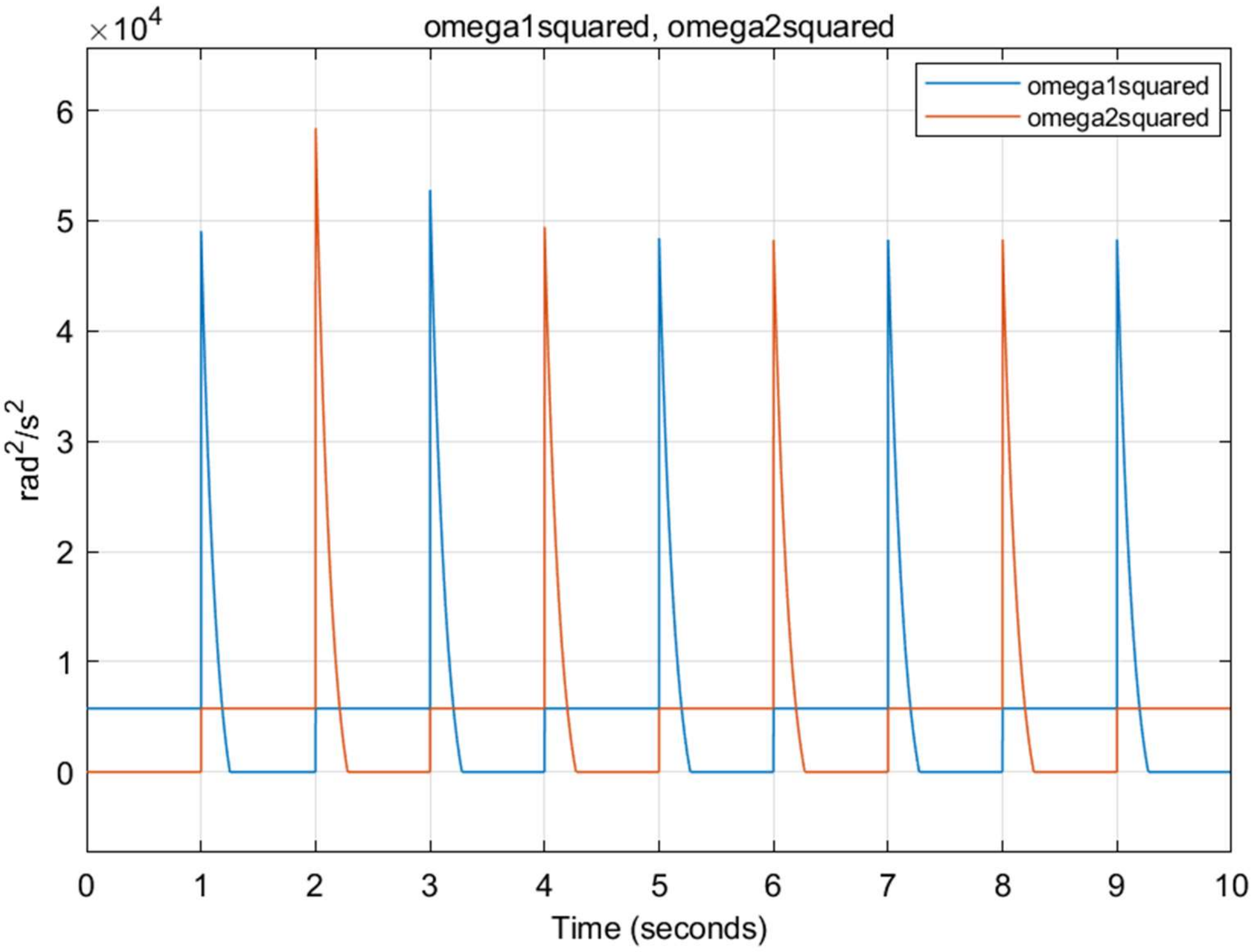

**Figure 25.** Inputs.

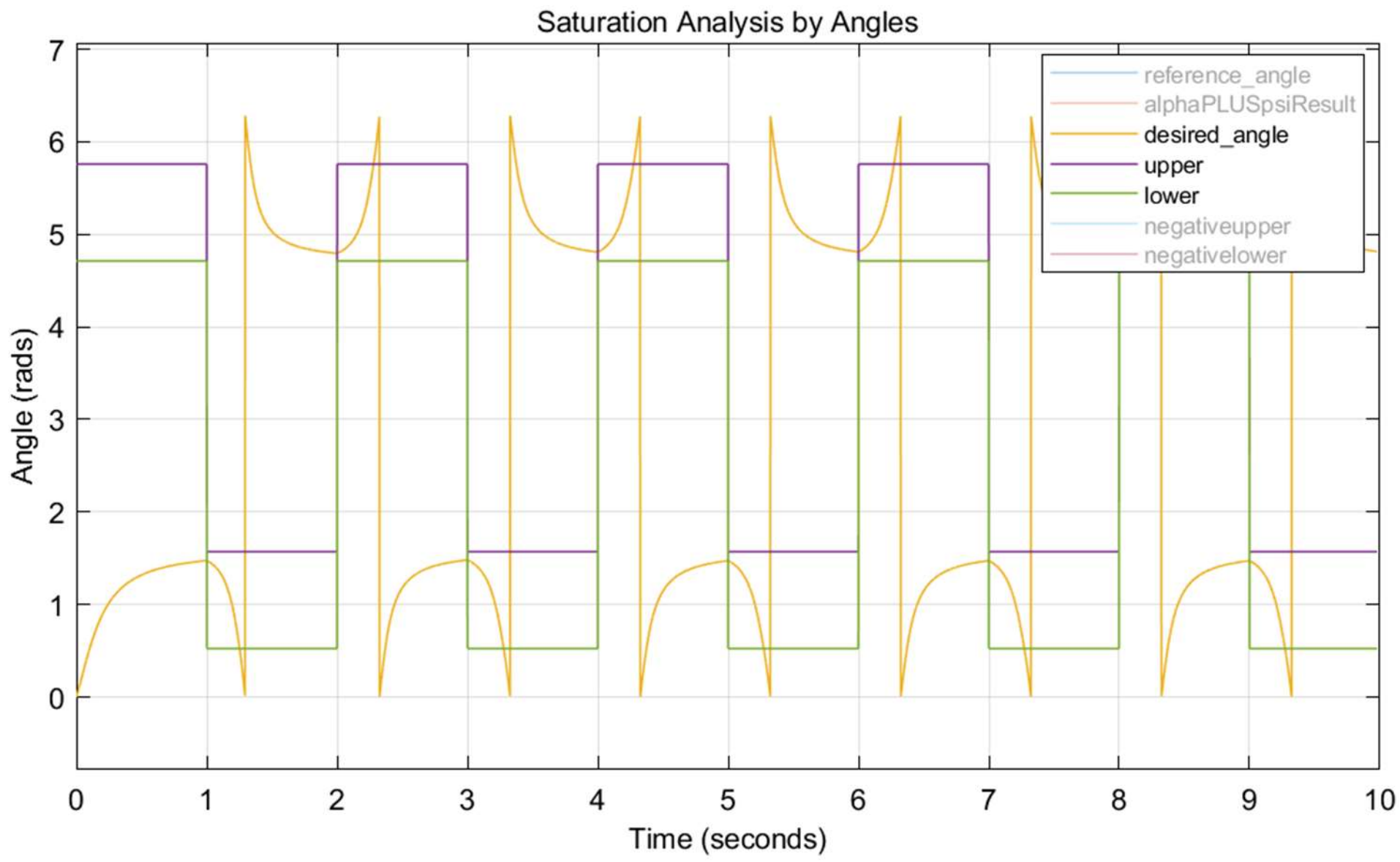

**Figure 26.** The angle of acceleration.

The history of the Lyapunov candidate is diagramed in Figure 27. It is not negative-definite, failing to prove the asymptotically stable. While the system is not asymptotically stable judged from Figure 24.

On the other hand, the Lyapunov candidate is bounded in Figure 27. The value of Lyapunov candidate is locally maximized at time $t = n \cdot \frac{T}{2}$, $(n \in \boldsymbol{Z}^{+})$. It is as expected in Remark 7.

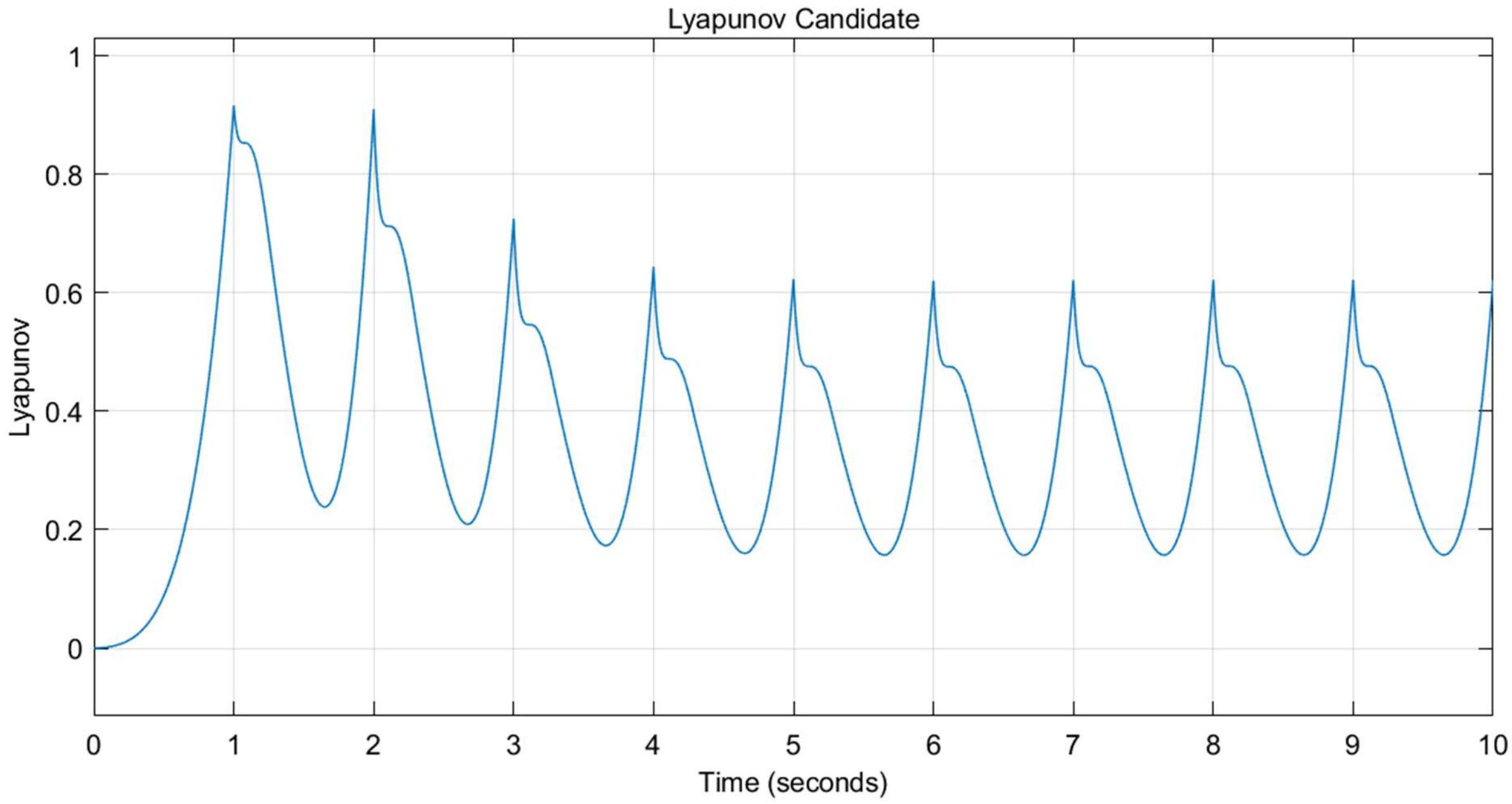

**Figure 27.** The record of the Lyapunov candidate.

## 8. Conclusions and Discussions

Two different penguin-inspired gaits are applied to facilitate to control the tilt vehicle in this research. The first gait introduces no saturation in control input, reaching the zero dynamic state error in the end. The controller in the second gait fails to asymptotically stabilize the tilt vehicle. Saturation happens in this case. While the state error in the second gait case is bounded.

The stability proofs for both cases are revealed. We picked the same Lyapunov candidate for both cases. It is proved semi-negative-definite for the first gait. While it is proved bounded for the second case.

In 'Main Proof' in Section 6, we proved that $\mathcal{L}(t)$ is bounded. It can be interesting to address further discussions on the upper bound of the Lyapunov candidate in the second case (gait causing saturation).

The Lyapunov candidate in (49) can be rewritten in (74), substituting $K_{Y_2}$.

$$\mathcal{L} = \frac{1}{2} \cdot \dot{e_y}^2 + 9 \cdot {e_y}^2 \quad (74)$$

(74) can be further written in (75).

$$\frac{{e_y}^2}{\left(\frac{\sqrt{\mathcal{L}}}{3}\right)^2} + \frac{\dot{e_y}^2}{\left(\sqrt{2} \cdot \sqrt{\mathcal{L}}\right)^2} = 1 \quad (75)$$

(75) defines an ellipse in Figure 28.

When we receive a particular $\mathcal{L}$, the corresponding $\left(e_y, \dot{e}_y\right)$ must fall on the edge of this ellipse. It is also worth noting that $\left(e_y, \dot{e}_y\right)\Big|_{t=n \cdot \frac{T}{2}}$, $(n \in \boldsymbol{Z}^+)$, can only appear on the edge of the ellipse in Quadrant I and Quadrant III (See Remark 6).

When we receive a larger value in Lyapunov candidate (e.g., $\mathcal{L}_1 > \mathcal{L}$), the ellipse is also expanded (See the $\mathcal{L}_1 - induced$ ellipse in Figure 29). When we receive a smaller Lyapunov (e.g., $\mathcal{L}_2 < \mathcal{L}$), the ellipse is also shrunk (See the $\mathcal{L}_2 - induced$ ellipse in Figure 29).

Define $\mathcal{L}_{critical}$ in (76).

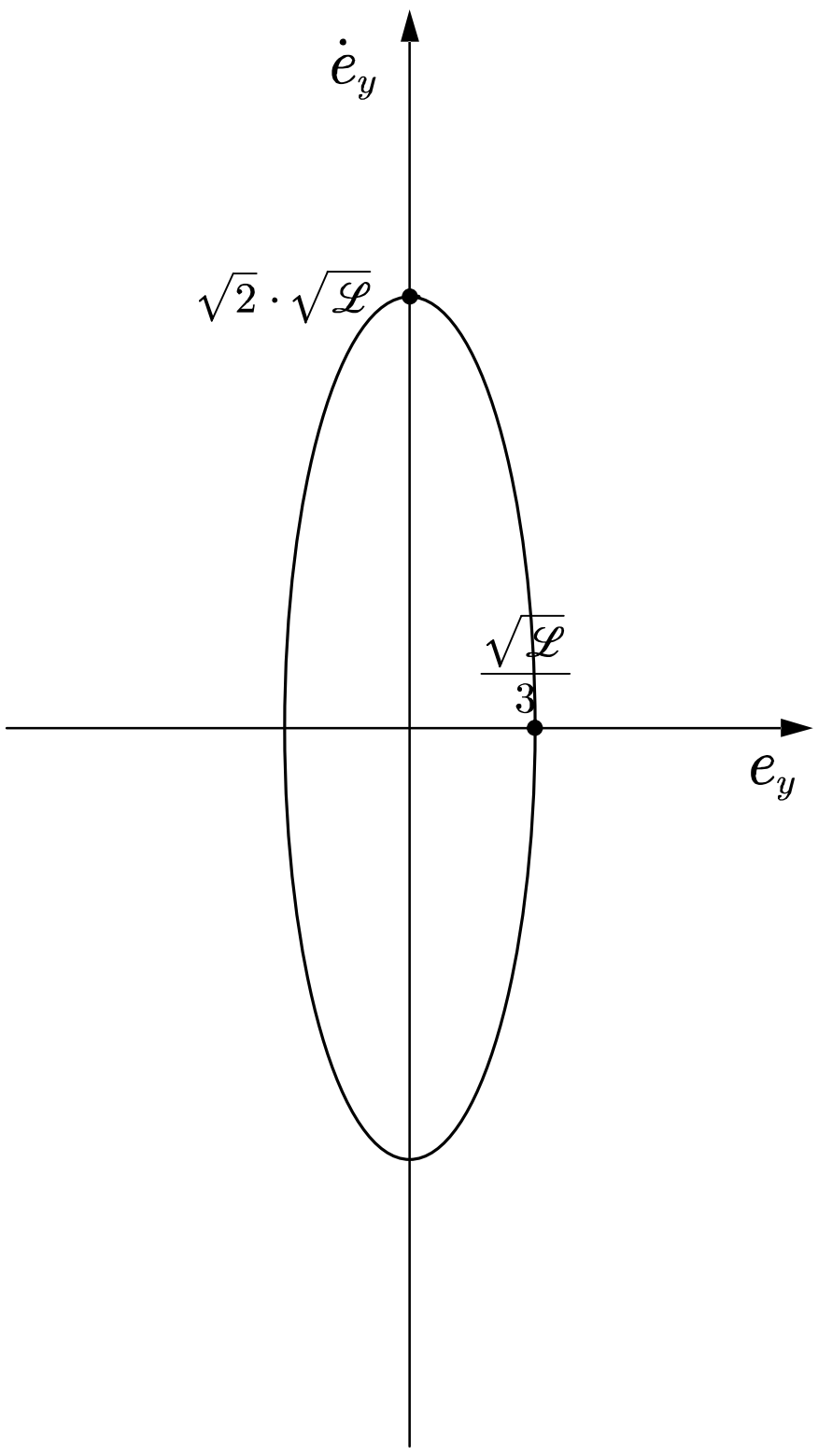

**Figure 28.** Possible $(e_y, \dot{e}_y)$.

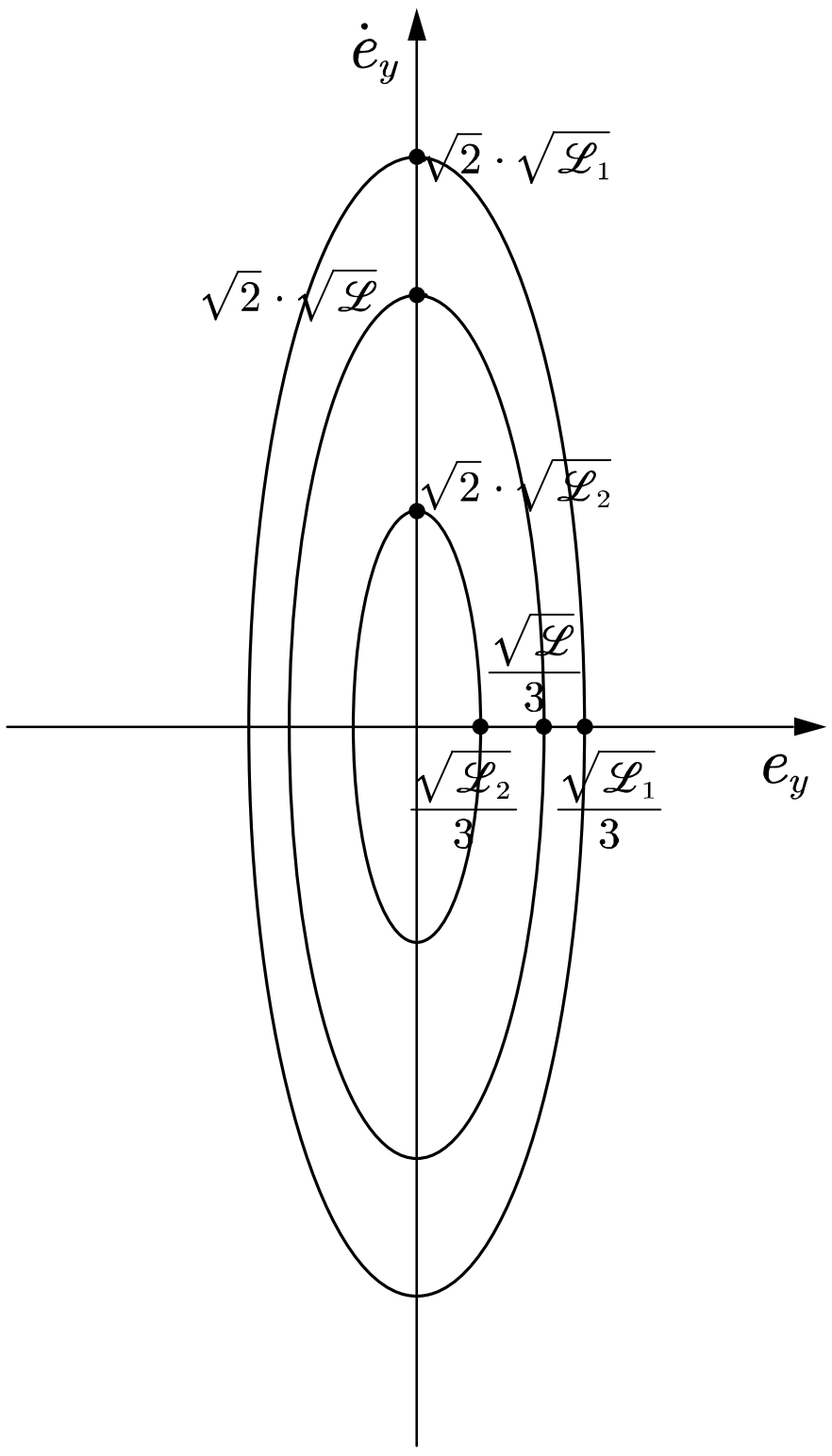

**Figure 29.** Different $\mathcal{L}$.

$$\begin{cases} \{\mathcal{L}_{critical}-induced \quad ellipse\} \cap \left\{ (e_y, \dot{e}_y) \right\}_{\Delta\mathcal{L}(n)\geqslant 0} \neq \varnothing \\ \forall \mathcal{L} > \mathcal{L}_{critical}: \\ \{\mathcal{L}-induced \quad ellipse\} \cap \left\{ (e_y, \dot{e}_y) \right\}_{\Delta\mathcal{L}(n)\geqslant 0} = \varnothing \end{cases} \tag{76}$$

where $\left\{ (e_y, \dot{e}_y) \right\}_{\Delta\mathcal{L}(n)\geqslant 0}$ is the collection of $(e_y, \dot{e}_y)$ satisfying $_\Delta\mathcal{L}(n) \geqslant 0$ (See Lemma 9).

Thus, $\left\{ (e_y, \dot{e}_y) \right\}_{\Delta\mathcal{L}(n)\geqslant 0}$ is (part of) the red area in Figure 22 and Figure 23 (See Remark 11).

Figure 30 plots the $\mathcal{L}_{critical} - induced$ ellipse.

**Proposition 7.**

$$supremum(\mathcal{L}(t)) \leqslant \mathcal{L}_{critical} + \tfrac{3}{4} \tag{77}$$

where $t \geqslant t_L$, $t_L$ is a finite constant.

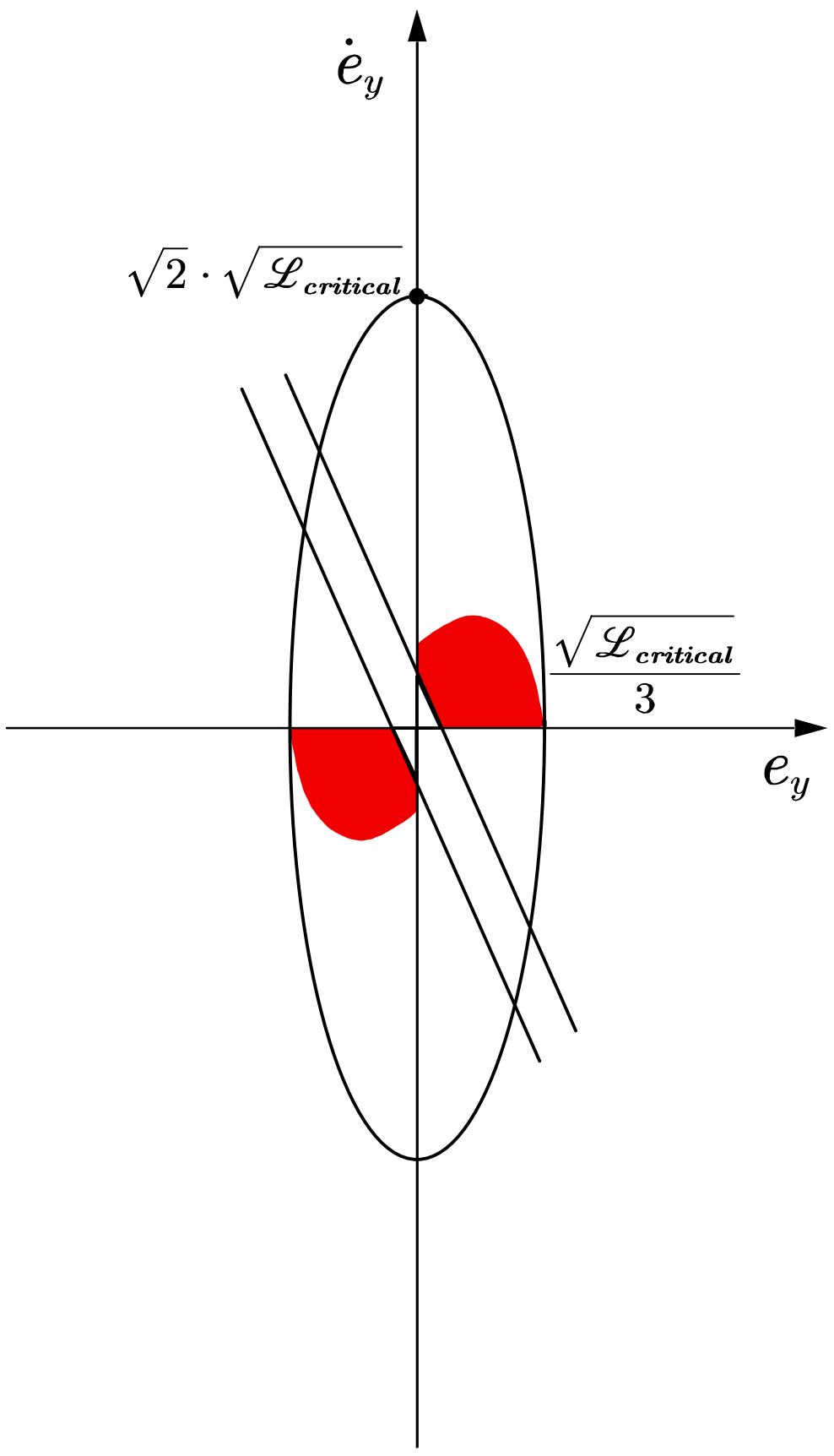

**Figure 30.** $\mathcal{L}_{critical} - induced$ ellipse.

**Proof**

We set $t_L$ sufficiently large so that $\left(e_y, \dot{e}_y\right)\big|_{t=\rho\cdot\frac{T}{2}}$, $\left(\rho\in \boldsymbol{Z}^+, \rho\cdot\frac{T}{2}<t_L\right)$, falls inside the red area defined in Figure 22 and Figure 23.

$1^\circ$ $\forall n > \rho$, $(n\in \boldsymbol{Z}^+)$, $\left(e_y, \dot{e}_y\right)\big|_{t=n\cdot\frac{T}{2}}$ is unable to enter the white area in Figure 22 and Figure 23

In this case, we have (78).

$$supremum\left(\mathcal{L}(t)\right) \leqslant \mathcal{L}_{critical} < \mathcal{L}_{critical} + \tfrac{3}{4}, (t\geqslant t_L) \tag{78}$$

$2^\circ$ $\exists m > \rho$, $(m\in \boldsymbol{Z}^+)$, $\left(e_y, \dot{e}_y\right)\big|_{t=m\cdot\frac{T}{2}}$ enters the white area while $\left(e_y, \dot{e}_y\right)\big|_{t=(m-1)\cdot\frac{T}{2}}$ is in the red area

In this case, we have (79).

$$\mathcal{L}\left(m\cdot\tfrac{T}{2}\right) = \mathcal{L}\left((m-1)\cdot\tfrac{T}{2}\right) + {}_{\Delta}\mathcal{L}(m-1) \leqslant \mathcal{L}_{critical} + \tfrac{3}{4} \tag{79}$$

After entering the white area, we have ${}_{\Delta}\mathcal{L} < 0$. We may have case $2^\circ a$ or case $2^\circ b$ subsequently.

$2^\circ a$. If $\forall q > m$, $(q\in \boldsymbol{Z}^+)$, $\left(e_y, \dot{e}_y\right)\big|_{t=q\cdot\frac{T}{2}}$ is unable to enter the red area.

$$\mathcal{L}\left(q\cdot\tfrac{T}{2}\right) < \mathcal{L}\left(m\cdot\tfrac{T}{2}\right) \leqslant \mathcal{L}_{critical} + \tfrac{3}{4} \tag{80}$$

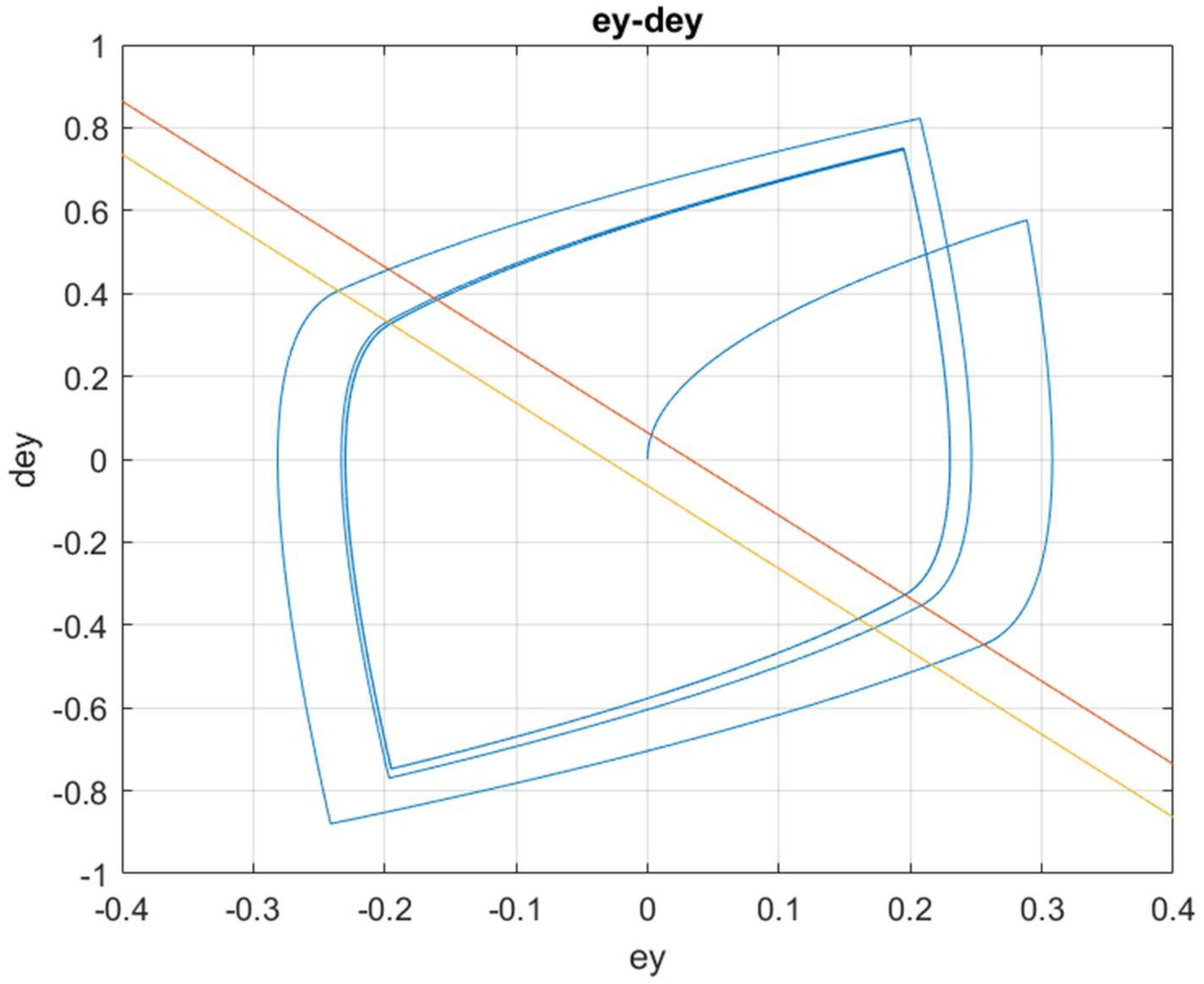

**Figure 31.** $(e_y, \dot{e}_y)$ history

Thus,

$$supremum(\mathcal{L}(t)) \leqslant \mathcal{L}\left(m \cdot \frac{T}{2}\right) < \mathcal{L}_{critical} + \frac{3}{4}, (t \geqslant t_L) \quad (81)$$

$2^{\circ}b$. $\exists w > m$, $(w \in \boldsymbol{Z}^{+})$, $(e_y, \dot{e}_y)\big|_{t=w\cdot\frac{T}{2}}$ enters the red area while $(e_y, \dot{e}_y)\big|_{t=(w-1)\cdot\frac{T}{2}}$ is in the white area

$(e_y, \dot{e}_y)\big|_{t=w\cdot\frac{T}{2}}$ returns to the red area where $(e_y, \dot{e}_y)\big|_{t=\rho\cdot\frac{T}{2}}$ is in. This case is recursively proved by $1^{\circ}$ and $2^{\circ}a$.

It is worth mentioning that Proposition 7 provides a relaxed upper bound for Lyapunov candidate. Further approaching the supremum is beyond the main scope of this research.

The actual $(e_y, \dot{e}_y)$ history in the result is in Figure 31.

Another point worth mentioning is that both the PD control coefficients (coefficients in (9) and (10)) and the planned gait (period, $T$, and the amplitude, $\Lambda$) can affect the stability. For some controller settings, the stability proof forwarded in this research can be challenged (some cases even cause the system unstable. That is partially the reason we avoid deducing the stability proof in a totally analytical way.

One further step can be advocating the controller to a more complicated reference (e.g., circular). Another is to generalize our stability proof to be applicable to wider control parameters.

## References

1. Z. Shen, and T. Tsuchiya, “State Drift and Gait Plan in Feedback Linearization Control of A Tilt Vehicle,” *arXiv preprint* arXiv:2111.04307, 2021.
2. W. Dunham, C. Petersen, and I. Kolmanovsky, “Constrained Control for soft landing on an asteroid with gravity model uncertainty,” *2016 American Control Conference (ACC)*, 2016.
3. K. McDonough and I. Kolmanovsky, “Controller State and reference governors for discrete-time linear systems with pointwise-in-time state and control constraints,” *2015 American Control Conference (ACC)*, 2015.

4. M. Ryll, H. H. Bulthoff, and P. R. Giordano, "Modeling and control of a quadrotor UAV with tilting propellers," *2012 IEEE International Conference on Robotics and Automation*, 2012.
5. G. K. K. and P. M. Pathak, "Dynamic Modelling & Simulation of a four legged jumping robot with compliant legs," *Robotics and Autonomous Systems*, vol. 61, no. 3, pp. 221–228, 2013.
6. S. Chemova and M. Veloso, "An evolutionary approach to gait learning for four-legged robots," *2004 IEEE/RSJ International Conference on Intelligent Robots and Systems (IROS), 2004*.
7. T. Kato, K. Shiromi, M. Nagata, H. Nakashima, and K. Matsuo, "Gait pattern acquisition for four-legged mobile robot by genetic algorithm," *IECON 2015 - 41st Annual Conference of the IEEE Industrial Electronics Society*, 2015.
8. J. I. Giribet, C. D. Pose, A. S. Ghersin, and I. Mas, "Experimental validation of a fault tolerant hexacopter with tilted rotors," *International Journal of Electrical and Electronic Engineering & Telecommunications*., pp. 58–65, 2018.
9. H. Voos, "Nonlinear control of a quadrotor micro-UAV using feedback-linearization," *2009 IEEE International Conference on Mechatronics*, 2009.
10. Y. Mutoh and S. Kuribara, "Control of quadrotor unmanned aerial vehicles using exact linearization technique with the static state feedback," *Journal of Automation and Control Engineering*, pp. 340–346, 2016.
11. T. Taniguchi, L. Eciolaza, and M. Sugeno, "Tracking control for a non-holonomic car-like robot using dynamic feedback linearization based on piecewise bilinear models," *2014 IEEE International Conference on Fuzzy Systems (FUZZ-IEEE)*, 2014.
12. E. Garone and M. M. Nicotra, "Explicit reference governor for Constrained Nonlinear Systems," *IEEE Transactions on Automatic Control*, vol. 61, no. 5, pp. 1379–1384, 2016.
13. Garone, E., Di Cairano, S. and Kolmanovsky, I., Reference and command governors for systems with constraints: A survey on theory and applications. *Automatica*, 75, pp.306-328, 2017.
14. B. Convens, K. Merckaert, M. M. Nicotra, R. Naldi, and E. Garone, "Control of fully actuated unmanned aerial vehicles with actuator saturation," *IFAC-PapersOnLine*, vol. 50, no. 1, pp. 12715–12720, 2017.
15. M. M. Nicotra and E. Garone, "Explicit reference governor for continuous time nonlinear systems subject to convex constraints," *2015 American Control Conference (ACC)*, 2015.
16. A. Cotorruelo, M. M. Nicotra, D. Limon, and E. Garone, "Explicit reference governor toolbox (ERGT)," *2018 IEEE 4th International Forum on Research and Technology for Society and Industry (RTSI)*, 2018.
17. T. Lee, M. Leok, and N. H. McClamroch, "Geometric tracking control of a quadrotor UAV on SE(3)," *49th IEEE Conference on Decision and Control (CDC)*, 2010.
18. F. A. Goodarzi, D. Lee, and T. Lee, "Geometric adaptive tracking control of a quadrotor unmanned aerial vehicle on SE(3) for agile maneuvers," *Journal of Dynamic Systems, Measurement, and Control*, vol. 137, no. 9, 2015.
19. W. Craig, D. Yeo, and D. A. Paley, "Geometric attitude and position control of a quadrotor in wind," *Journal of Guidance, Control, and Dynamics*, vol. 43, no. 5, pp. 870–883, 2020.
20. K. Gamagedara, M. Bisheban, E. Kaufman, and T. Lee, "Geometric controls of a quadrotor UAV with decoupled yaw control," *2019 American Control Conference (ACC)*, 2019.
21. H. Lee, S. Kim, T. Ryan, and H. J. Kim, "Backstepping control on SE(3) of a micro Quadrotor for stable trajectory tracking," *2013 IEEE International Conference on Systems, Man, and Cybernetics*, 2013.
22. S. Bouabdallah and R. Siegwart, "Backstepping and sliding-mode techniques applied to an indoor micro quadrotor," *Proceedings of the 2005 IEEE International Conference on Robotics and Automation*, 2005.
23. T. Madani and A. Benallegue, "Backstepping control for a quadrotor helicopter," *2006 IEEE/RSJ International Conference on Intelligent Robots and Systems*, 2006.
24. R. Xu and U. Ozguner, "Sliding mode control of a quadrotor helicopter," *Proceedings of the 45th IEEE Conference on Decision and Control*, 2006.
25. D. Lee, H. Jin Kim, and S. Sastry, "Feedback linearization vs. Adaptive Sliding Mode control for a quadrotor helicopter," *International Journal of Control, Automation and Systems*, vol. 7, no. 3, pp. 419–428, 2009.
26. K. Runcharoon and V. Srichatrapimuk, "Sliding mode control of quadrotor," *2013 The International Conference on Technological Advances in Electrical, Electronics and Computer Engineering (TAEECE)*, 2013.
27. E. Hendricks, O. Jannerup, and P. H. Sørensen, *Linear systems control: deterministic and stochastic methods*. Springer Science & Business Media, 2008.
28. Khalil, Hassan K., *Nonlinear systems third edition*, Patience Hall 115, 2002.

# Chapter 12

## Discussions (The Limitation of This Research: Numerically-Deduced Result)

In deducing the Two Color Map Theorem (Chapter 6) and the generalized Two Color Map Theorem (Chapter 8), the numerical solver is applied to receive the approximate roots. Consequently, the results highly depend on the parameters of the interested tiltrotor.

Here, the parameters of the tiltrotor model adopted in the previous chapters (Chapter 3 – Chapter 9 [1–7]) are inherited from AR Drone 2 [8,9]. One can wonder the effects of the parameter changes in the tiltrotor to the Two Color Map Theorem and the generalized Two Color Map Theorem, e.g., one may ask "what is result of the Two Color Map accommodating to the parameters of the updated tiltrotor?"

With this concern, this chapter (Chapter 13) addresses the discussion on the Two Color Map Theorem resulted from the tiltrotor with parameters inherited from another popular drone model [10].

Note that, judging from Equation (18) in Chapter 3, there are three parameters (the length of the arm, $L$, the coefficient of the thrust, $K_f$, the coefficient of the drag moment, $K_m$) of the tiltrotor model that influences the result of the Two Color Map (Chapter 6 and Chapter 8); the differences in these three parameters in the different tiltrotor models result in different decoupling matrices, which consequentially lead to the unique corresponding Two Color Map.

These three interested updated parameters of the new tiltrotor model [10] are:

$$L = 0.225m \tag{1}$$

$$K_f = 2.980 \times 10^{-6} N \cdot s^2/rad^2 \tag{2}$$

$$K_m = 1.140 \times 10^{-7} N \cdot m \cdot s^2/rad^2 \tag{3}$$

The relevant proposition corresponding to this new tiltrotor is subsequently asserted:

**Proposition 1.** *The decoupling matrix is invertible if, and only if:*

$$\begin{aligned}
&1.025 \cdot c1 \cdot c2 \cdot c3 \cdot s4 \cdot s\theta - 1.025 \cdot c1 \cdot c3 \cdot c4 \cdot s2 \cdot s\theta - 2.927 \\
&\cdot c1 \cdot c2 \cdot s3 \cdot s4 \cdot s\theta + 2.927 \cdot c1 \cdot c4 \cdot s2 \cdot s3 \cdot s\theta - 2.927 \cdot c2 \\
&\cdot c3 \cdot s1 \cdot s4 \cdot s\theta + 2.927 \cdot c3 \cdot c4 \cdot s1 \cdot s2 \cdot s\theta - 1.025 \cdot c2 \cdot s1 \\
&\cdot s3 \cdot s4 \cdot s\theta + 1.025 \cdot c4 \cdot s1 \cdot s2 \cdot s3 \cdot s\theta + 4.100 \cdot c1 \cdot c2 \cdot c3 \\
&\cdot c4 \cdot c\phi \cdot c\theta + 5.680 \cdot c1 \cdot c2 \cdot c3 \cdot s4 \cdot c\phi \cdot c\theta - 5.680 \cdot c1 \cdot c2 \\
&\cdot c4 \cdot s3 \cdot c\phi \cdot c\theta + 5.680 \cdot c1 \cdot c3 \cdot c4 \cdot s2 \cdot c\phi \cdot c\theta - 5.680 \cdot c2 \\
&\cdot c3 \cdot c4 \cdot s1 \cdot c\phi \cdot c\theta + 1.025 \cdot c1 \cdot c2 \cdot c4 \cdot s3 \cdot c\theta \cdot s\phi \\
&+ 0.9954 \cdot c1 \cdot c2 \cdot s3 \cdot s4 \cdot c\phi \cdot c\theta - 2.050 \cdot c1 \cdot c3 \cdot s2 \cdot s4 \\
&\cdot c\phi \cdot c\theta + 0.9954 \cdot c1 \cdot c4 \cdot s2 \cdot s3 \cdot c\phi \cdot c\theta - 1.025 \cdot c2 \cdot c3 \\
&\cdot c4 \cdot s1 \cdot c\theta \cdot s\phi + 0.9954 \cdot c2 \cdot c3 \cdot s1 \cdot s4 \cdot c\phi \cdot c\theta - 2.050 \\
&\cdot c2 \cdot c4 \cdot s1 \cdot s3 \cdot c\phi \cdot c\theta + 0.9954 \cdot c3 \cdot c4 \cdot s1 \cdot s2 \cdot c\phi \cdot c\theta \\
&+ 2.927 \cdot c1 \cdot c2 \cdot s3 \cdot s4 \cdot c\theta \cdot s\phi + 2.927 \cdot c1 \cdot c4 \cdot s2 \cdot s3 \cdot c\theta \\
&\cdot s\phi - 0.1743 \cdot c1 \cdot s2 \cdot s3 \cdot s4 \cdot c\phi \cdot c\theta - 2.927 \cdot c2 \cdot c3 \cdot s1 \cdot s4 \\
&\cdot c\theta \cdot s\phi + 0.1743 \cdot c2 \cdot s1 \cdot s3 \cdot s4 \cdot c\phi \cdot c\theta - 2.927 \cdot c3 \cdot c4 \\
&\cdot s1 \cdot s2 \cdot c\theta \cdot s\phi - 0.1743 \cdot c3 \cdot s1 \cdot s2 \cdot s4 \cdot c\phi \cdot c\theta + 0.1743 \\
&\cdot c4 \cdot s1 \cdot s2 \cdot s3 \cdot c\phi \cdot c\theta - 1.025 \cdot c1 \cdot s2 \cdot s3 \cdot s4 \cdot c\theta \cdot s\phi \\
&+ 1.025 \cdot c3 \cdot s1 \cdot s2 \cdot s4 \cdot c\theta \cdot s\phi
\end{aligned} \tag{4}$$

$$\neq 0,$$

where $s\Lambda = \sin(\Lambda)$ and $c\Lambda = \cos(\Lambda)$. $\phi$, $\theta$, and $\psi$ are the roll angle, pitch angle, and yaw angle, respectively. The tilting angles $\alpha = [\alpha_1 \quad \alpha_2 \quad \alpha_3 \quad \alpha_4]$. $si = \sin(\alpha_i)$, $ci = \cos(\alpha_i)$, $(i = 1,2,3,4)$.

In comparison, the coefficients in Equation (4) in this chapter differ greatly from the ones in Equation (17) in Chapter 3. This updated proposition subsequently leads to the updated result of the linearization at the state of zero roll and pitch angles in Equation (5).

$$R_\phi(\alpha_1,\alpha_2,\alpha_3,\alpha_4)\cdot\phi + R_\theta(\alpha_1,\alpha_2,\alpha_3,\alpha_4)\cdot\theta + R(\alpha_1,\alpha_2,\alpha_3,\alpha_4) \neq 0 \tag{5}$$

where

$$R_\phi(\alpha_1,\alpha_2,\alpha_3,\alpha_4) = 1.025\cdot c1\cdot c2\cdot s3\cdot c4 - 1.025\cdot s1\cdot c2\cdot c3\cdot c4 + 2.927\cdot c1\cdot c2\cdot s3\cdot s4 + 2.927\cdot c1\cdot s2\cdot s3\cdot c4 - 2.927\cdot s1\cdot c2\cdot c3\cdot s4 - 2.927\cdot s1\cdot s2\cdot c3\cdot c4 - 1.025\cdot c1\cdot s2\cdot s3\cdot s4 + 1.025\cdot s1\cdot s2\cdot c3\cdot s4, \tag{6}$$

$$R_\theta(\alpha_1,\alpha_2,\alpha_3,\alpha_4) = 1.025\cdot c1\cdot c2\cdot c3\cdot s4 - 1.025\cdot c1\cdot s2\cdot c3\cdot c4 - 2.927\cdot c1\cdot c2\cdot s3\cdot s4 + 2.927\cdot c1\cdot s2\cdot s3\cdot c4 - 2.927\cdot s1\cdot c2\cdot c3\cdot s4 + 2.927\cdot s1\cdot s2\cdot c3\cdot c4 - 1.025\cdot s1\cdot c2\cdot s3\cdot s4 + 1.025\cdot s1\cdot s2\cdot s3\cdot c4, \tag{7}$$

$$R(\alpha_1,\alpha_2,\alpha_3,\alpha_4) = 4.100\cdot c1\cdot c2\cdot c3\cdot c4 + 5.680\cdot c1\cdot c2\cdot c3\cdot s4 - 5.680\cdot c1\cdot c2\cdot s3\cdot c4 + 5.680\cdot c1\cdot s2\cdot c3\cdot c4 - 5.680\cdot s1\cdot c2\cdot c3\cdot c4 + 0.9954\cdot c1\cdot c2\cdot s3\cdot s4 - 2.050\cdot c1\cdot s2\cdot c3\cdot s4 + 0.9954\cdot c1\cdot s2\cdot s3\cdot c4 + 0.9954\cdot s1\cdot c2\cdot c3\cdot s4 - 2.050\cdot s1\cdot c2\cdot s3\cdot c4 + 0.9954\cdot s1\cdot s2\cdot c3\cdot c4 - 0.1743\cdot c1\cdot s2\cdot s3\cdot s4 + 0.1743\cdot s1\cdot c2\cdot s3\cdot s4 - 0.1743\cdot s1\cdot s2\cdot c3\cdot s4 + 0.1743\cdot s1\cdot s2\cdot s3\cdot c4. \tag{8}$$

Define

$$\mathcal{L}: \tilde{R}(\alpha_1,\alpha_2) = 0, \tag{9}$$

where

$$\tilde{R}(\alpha_1,\alpha_2) = 4.100\cdot c1\cdot c2\cdot c1\cdot c2 + 5.680\cdot c1\cdot c2\cdot c1\cdot s2 - 5.680\cdot c1\cdot c2\cdot s1\cdot c2 + 5.680\cdot c1\cdot s2\cdot c1\cdot c2 - 5.680\cdot s1\cdot c2\cdot c1\cdot c2 + 0.9954\cdot c1\cdot c2\cdot s1\cdot s2 - 2.050\cdot c1\cdot s2\cdot c1\cdot s2 + 0.9954\cdot c1\cdot s2\cdot s1\cdot c2 + 0.9954\cdot s1\cdot c2\cdot c1\cdot s2 - 2.050\cdot s1\cdot c2\cdot s1\cdot c2 + 0.9954\cdot s1\cdot s2\cdot c1\cdot c2 - 0.1743\cdot c1\cdot s2\cdot s1\cdot s2 + 0.1743\cdot s1\cdot c2\cdot s1\cdot s2 - 0.1743\cdot s1\cdot s2\cdot c1\cdot s2 + 0.1743\cdot s1\cdot s2\cdot s1\cdot c2, \tag{10}$$

where $si = \sin(\alpha_i)$, $ci = \cos(\alpha_i)$, $(i = 1,2)$.

Following the procedures proposed in Chapter 6 and Chapter 8, a new Two Color Map, which is similar to the map in Chapter 8, is deduced and depicted in Figure 1.

Note that these results are highly resemble to the ones reported in Chapter 8. However, comparing with the findings in Chapter 8, several minor differences are observed in this chapter. Firstly, the orange point $(\alpha_1, \alpha_2)$ in Figure 1 slightly changes its location. Secondly, $\mathcal{L}$, defined in Equation (10), shifts its shape in Two Color Map; the coefficients of Equation (10) differ from Equation (7) in Chapter 8.

Moreover, the $(\alpha_3, \alpha_4)$ corresponding to two kinds of $(\alpha_1, \alpha_2)$ in different colors (red and blue) are also updated as below:

The $(\alpha_3, \alpha_4)$ corresponding to red $(\alpha_1, \alpha_2)$ satisfy

$$\begin{cases}\alpha_3 = \alpha_1 \\ \alpha_4 = \alpha_2\end{cases}. \tag{11}$$

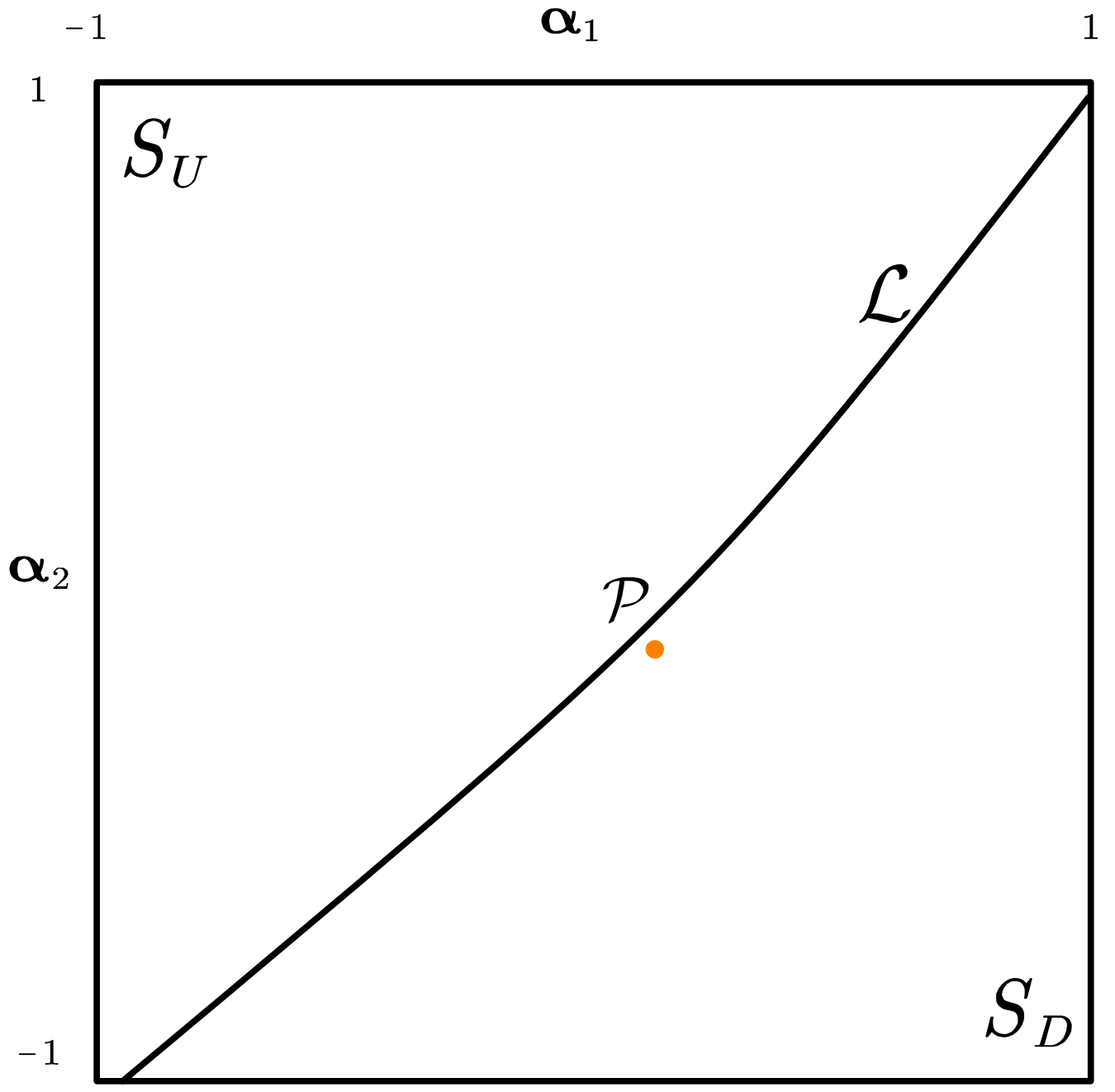

**Figure 1.** Curve $\mathcal{L}$ in $\alpha_1 - \alpha_2$ diagram. It divides the whole space into two subspaces, one upper space above $\mathcal{L}$ and one lower space below $\mathcal{L}$. The orange point is at $(\alpha_1, \alpha_2) = (0.168421, -0.168421)$. Note that this point is a little below $\mathcal{L}$ rather than on $\mathcal{L}$.

And, the $(\alpha_3, \alpha_4)$ corresponding to blue $(\alpha_1, \alpha_2)$ satisfy

$$\begin{cases} \alpha_3 = -\alpha_1 + 0.336842 \\ \alpha_4 = -\alpha_2 - 0.336842 \end{cases}. \tag{12}$$

Note that Equation (12) is distinct to the relevant ones, Equation (12) in Chapter 8. Further explanations on the Two Color Map Theorem are omitted since it is identical to the Chapter 8.

In conclusion, the change of the parameters (the length of the arm, $L$, the coefficient of the thrust, $K_f$, the coefficient of the drag moment, $K_m$) of the tiltrotor tends to influence the Two Color Map theorem. Therefore, it can be encouraged to re-deduce the Two Color Map to accommodate the parameters of the interested tiltrotor.

**References**

1. Shen, Z.; Tsuchiya, T. Gait Analysis for a Tiltrotor: The Dynamic Invertible Gait. *Robotics* **2022**, *11*, 33, doi:10.3390/robotics11020033.
2. Shen, Z.; Ma, Y.; Tsuchiya, T. Feedback Linearization-Based Tracking Control of a Tilt-Rotor with Cat-Trot Gait Plan. *International Journal of Advanced Robotic Systems* **2022**, *19*, 17298806221109360, doi:10.1177/17298806221109360.
3. Shen, Z.; Tsuchiya, T. Cat-Inspired Gaits for a Tilt-Rotor—From Symmetrical to Asymmetrical. *Robotics* **2022**, *11*, 60, doi:10.3390/robotics11030060.
4. Shen, Z.; Ma, Y.; Tsuchiya, T. Four-Dimensional Gait Surfaces for a Tilt-Rotor—Two Color Map Theorem. *Drones* **2022**, *6*, 103, doi:10.3390/drones6050103.
5. Shen, Z.; Tsuchiya, T. The Robust Gait of a Tilt-Rotor and Its Application to Tracking Control -- Application of Two Color Map Theorem 2022.
6. Shen, Z.; Ma, Y.; Tsuchiya, T. Generalized Two Color Map Theorem -- Complete Theorem of Robust Gait Plan for a Tilt-Rotor 2022.
7. Shen, Z.; Tsuchiya, T. Tracking Control for a Tilt-Rotor with Input Constraints by Robust Gaits. 7.

8. Li, Q. Masters Thesis: Grey-Box System Identification of a Quadrotor Unmanned Aerial Vehicle. **2014**.
9. Doukhi, O.; Fayjie, A.R.; Lee, D.J. Intelligent Controller Design for Quad-Rotor Stabilization in Presence of Parameter Variations. *Journal of Advanced Transportation* **2017**, *2017*, 1–10, doi:10.1155/2017/4683912.
10. Luukkonen, T. Modelling and Control of Quadcopter. *Independent research project in applied mathematics, Espoo* **2011**, *22*, 22.

# Chapter 13

## Conclusions and Future Works

We proved that, in the conventional quadrotors, the decoupling matrix in feedback linearization is not always invertible in the entire attitude zone for the yaw–position output combination (Chapter 2). Thus, the relevant controller risks encountering the singular decoupling matrix that causes the failure in controlling; the attitude causing the singularity in the decoupling matrix was presented analytically and visualized in this research.

Therefore, the feedback linearization with the yaw-position output combination is abandoned in stabilizing the tiltrotor. Whereas, we also proved that the attitude-altitude output combination in feedback linearization can still introduce the singular decoupling matrix (Chapter 3). The necessary conditions to develop an invertible decoupling matrix for the attitude–altitude-based feedback linearization method have been determined and visualized for a tiltrotor in the same chapter.

With this endeavor, the over-intensive change control signal (tilting angles) can be manipulated in a milder way and receives higher quality adopting the proper gait (Chapter 3).

To tracking the time-specified position reference, the relationship between position and attitude of the tilt-rotor is elucidated. The modified attitude-position decoupler is invented for position-tracking problem for a tilt-rotor. It significantly reduces the dynamic state error (or steady state error for the point reference) comparing with the conventional attitude-position decoupler (Chapter 4).

The unacceptable attitude curve proposed in Chapter 5 facilitates to evaluate the quality (robustness) of the gait of the tiltrotor effectively. The scaling method (Chapter 5) in the gait modification is proven to be feasible in finding a valid gait, liable to lead to the invertible decoupling matrix for the first time; this scaling method tends to enlarge the acceptable attitude zone in the roll-pitch diagram, indicating that this method strengthens the relevant gait.

Although scaling has been proven to be a valid approach to modifying an unacceptable gait, which results in a singular decoupling matrix, partially scaling may not increase the robustness and even may have the opposite effect for some gaits (Chapter 6).

Since no research guides the gait plan, considering the robustness, before Chapter 6, without adopting an existing gait. Indeed, deducing the explicit relationship between the attitude and the singularity of the decoupling matrix can be cumbersome since they are tangled in a highly nonlinear way.

Therefore, Chapter 6 deduces the Two Color Map which helps design the robust gaits in a more general way; the proposed four-dimensional gait surface helps plan the gait robust to the attitude change. The gaits on the four-dimensional gait surface show a wider region of acceptable attitudes compared with the relevant gait biased by partially scaling.

Interestingly, Two Color Map Theorem assists in finding the continuous gait on the four-dimensional gait surface. Multiple gaits can be found on the four-dimensional gait surface without violating this theorem (Chapter 6). The success in the tracking control adopting the gait obeying Two Color Map Theorem is witnessed in Chapter 7.

Comparing with the Two Color Map Theorem, the generalized theorem proposed in Chapter 8 provides the robust gaits which are ignored by the former theorem. Three ignored gaits are evaluated and found their strong robustness to the attitude change. Besides, the robust gaits with color switch are evaluated for the first time in this paper.

Several regions of the Two Color Map have been explored and found feasible in the trajectory-tracking tasks in Chapter 9. The admissible region of the accelerations while compensating the gravity has been explored. The result demonstrates a rectangular region of these accelerations, satisfying the input constraints.

In analyzing a mobile robot with the tilt-structure (Chapter 10), a disadvantage called state drift in feedback linearization is reported. Also, the stability is challenged due to the activation of the constraint of saturation. Chapter 11 successfully finds the stability proof to back the feedback linearization with saturation. It is also the first time that the robot controlled by feedback linearization with the activated saturation is proved stable.

In one word, the gait plan and its further generalized theorem put forward in this dissertation highly improved the quality of the control signals; the advanced feedback linearization control for tiltrotor UAVs avoided the over-intensive changes in the tilting angles, which is more desirable to the application. Also, it is the first time that the stability is proved for the tilt vehicle stabilized by feedback linearization in terms of the saturation.

Further works include the tracking simulations on the reference of higher complexity, the application on the real tiltrotor rather than the simulator, the deduction of the attitude-position decoupler with higher precision, the pursue of the generalized stability proof for feedback linearization to accommodate the real tiltrotor model with the existence of the saturation.

Furthermore, a deeper study on fault tolerant control for tiltrotors is also underway. The tiltrotor possesses eight inputs, exceeding the number of degrees of freedom. Therefore, fully or partially losing the specific number of inputs may not be critical in a flight; the tiltrotor may still hover even for the case of the malfunction in generating the desired magnitudes and/or directions of the thrusts. It is still an open question in finding the relationship between the faults in the inputs and the flight capability of the tiltrotor.

Moreover, this study has yet to clarify the connection between the control signal's frequency and the flight's quality, leaving room for further investigation. Neglecting the aerodynamic effects in the application of the actual models could potentially compromise the stability proof. Consequently, a comprehensive analysis of this issue may be essential.

# Bibliography

1. Ryll, M.; Bülthoff, H.H.; Giordano, P.R. Modeling and Control of a Quadrotor UAV with Tilting Propellers. In Proceedings of the Proceedings - IEEE International Conference on Robotics and Automation; 2012.
2. Senkul, F.; Altug, E. Adaptive Control of a Tilt-Roll Rotor Quadrotor UAV. In Proceedings of the 2014 International Conference on Unmanned Aircraft Systems, ICUAS 2014 - Conference Proceedings; 2014.
3. Senkul, F.; Altug, E. Modeling and Control of a Novel Tilt - Roll Rotor Quadrotor UAV. In Proceedings of the 2013 International Conference on Unmanned Aircraft Systems, ICUAS 2013 - Conference Proceedings; 2013.
4. Nemati, A.; Kumar, M. Modeling and Control of a Single Axis Tilting Quadcopter. In Proceedings of the Proceedings of the American Control Conference; 2014.
5. Junaid, A. Bin; Sanchez, A.D.D.C.; Bosch, J.B.; Vitzilaios, N.; Zweiri, Y. Design and Implementation of a Dual-Axis Tilting Quadcopter. Robotics 2018, 7, doi:10.3390/robotics7040065.
6. Andrade, R.; Raffo, G. V.; Normey-Rico, J.E. Model Predictive Control of a Tilt-Rotor UAV for Load Transportation. In Proceedings of the 2016 European Control Conference, ECC 2016; 2016.
7. Nemati, A.; Kumar, R.; Kumar, M. Stabilizing and Control of Tilting-Rotor Quadcopter in Case of a Propeller Failure. In Proceedings of the ASME 2016 Dynamic Systems and Control Conference, DSCC 2016; 2016; Vol. 1.
8. Anderson, R.B.; Marshall, J.A.; L'Afflitto, A. Constrained Robust Model Reference Adaptive Control of a Tilt-Rotor Quadcopter Pulling an Unmodeled Cart. IEEE Trans. Aerosp. Electron. Syst. 2021, 57, 39–54, doi:10.1109/TAES.2020.3008575.
9. Shen, Z.; Tsuchiya, T. Gait Analysis for a Tiltrotor: The Dynamic Invertible Gait. Robotics 2022, 11, 33, doi:10.3390/robotics11020033.
10. Ryll, M.; Bulthoff, H.H.; Giordano, P.R. A Novel Overactuated Quadrotor Unmanned Aerial Vehicle: Modeling, Control, and Experimental Validation. IEEE Trans. Contr. Syst. Technol. 2015, 23, 540–556, doi:10.1109/TCST.2014.2330999.
11. Jin, S.; Bak, J.; Kim, J.; Seo, T.; Kim, H.S. Switching PD-Based Sliding Mode Control for Hovering of a Tilting-Thruster Underwater Robot. PLOS ONE 2018, 13, e0194427, doi:10.1371/journal.pone.0194427.
12. Kumar, R.; Sridhar, S.; Cazaurang, F.; Cohen, K.; Kumar, M. Reconfigurable Fault-Tolerant Tilt-Rotor Quadcopter System.; American Society of Mechanical Engineers: Atlanta, Georgia, USA, September 30 2018; p. V003T37A008.
13. Invernizzi, D.; Giurato, M.; Gattazzo, P.; Lovera, M. Comparison of Control Methods for Trajectory Tracking in Fully Actuated Unmanned Aerial Vehicles. IEEE Transactions on Control Systems Technology 2021, 29, 1147–1160, doi:10.1109/TCST.2020.2992389.
14. Invernizzi, D.; Lovera, M. Trajectory Tracking Control of Thrust-Vectoring UAVs. Automatica 2018, 95, 180–186, doi:10.1016/j.automatica.2018.05.024.
15. Imamura, A.; Miwa, M.; Hino, J. Flight Characteristics of Quad Rotor Helicopter with Thrust Vectoring Equipment. Journal of Robotics and Mechatronics 2016, 28, 334–342, doi:10.20965/jrm.2016.p0334.
16. Ryll, M.; Bulthoff, H.H.; Giordano, P.R. First Flight Tests for a Quadrotor UAV with Tilting Propellers. In Proceedings of the 2013 IEEE International Conference on Robotics and Automation; IEEE: Karlsruhe, Germany, May 2013; pp. 295–302.
17. Shen, Z.; Tsuchiya, T. Cat-Inspired Gaits for a Tilt-Rotor—From Symmetrical to Asymmetrical. Robotics 2022, 11, 60, doi:10.3390/robotics11030060.
18. Ícaro Bezerra Viana; Luiz Manoel Santos Santana; Raphael Ballet; Davi Antônio dos Santos; Luiz Carlos Sandoval Góes Experimental Validation of a Trajectory Tracking Control Using the AR.Drone Quadrotor.; Fortaleza, Ceará, Brasil, 2016.
19. Merheb, A.-R.; Noura, H.; Bateman, F. Emergency Control of AR Drone Quadrotor UAV Suffering a Total Loss of One Rotor. IEEE/ASME Trans. Mechatron. 2017, 22, 961–971, doi:10.1109/TMECH.2017.2652399.
20. Lee, T.; Leok, M.; McClamroch, N.H. Geometric Tracking Control of a Quadrotor UAV on SE(3). In Proceedings of the 49th IEEE Conference on Decision and Control (CDC); IEEE: Atlanta, GA, December 2010; pp. 5420–5425.
21. Luukkonen, T. Modelling and Control of Quadcopter. Independent research project in applied mathematics, Espoo 2011, 22, 22.
22. Horla, D.; Hamandi, M.; Giernacki, W.; Franchi, A. Optimal Tuning of the Lateral-Dynamics Parameters for Aerial Vehicles With Bounded Lateral Force. IEEE Robot. Autom. Lett. 2021, 6, 3949–3955, doi:10.1109/LRA.2021.3067229.

23. Franchi, A.; Carli, R.; Bicego, D.; Ryll, M. Full-Pose Tracking Control for Aerial Robotic Systems With Laterally Bounded Input Force. IEEE Trans. Robot. 2018, 34, 534–541, doi:10.1109/TRO.2017.2786734.
24. Badr, S.; Mehrez, O.; Kabeel, A.E. A Design Modification for a Quadrotor UAV: Modeling, Control and Implementation. Advanced Robotics 2019, 33, doi:10.1080/01691864.2018.1556116.
25. Bhargavapuri, M.; Patrikar, J.; Sahoo, S.R.; Kothari, M. A Low-Cost Tilt-Augmented Quadrotor Helicopter : Modeling and Control. In Proceedings of the 2018 International Conference on Unmanned Aircraft Systems, ICUAS 2018; 2018.
26. Badr, S.; Mehrez, O.; Kabeel, A.E. A Novel Modification for a Quadrotor Design. In Proceedings of the 2016 International Conference on Unmanned Aircraft Systems, ICUAS 2016; 2016.
27. Jiang, X. ying; Su, C. li; Xu, Y. peng; Liu, K.; Shi, H. yuan; Li, P. An Adaptive Backstepping Sliding Mode Method for Flight Attitude of Quadrotor UAVs. Journal of Central South University 2018, 25, doi:10.1007/s11771-018-3765-0.
28. Jin, S.; Kim, J.; Kim, J.W.; Bae, J.H.; Bak, J.; Kim, J.; Seo, T.W. Back-Stepping Control Design for an Underwater Robot with Tilting Thrusters. In Proceedings of the Proceedings of the 17th International Conference on Advanced Robotics, ICAR 2015; 2015.
29. Kadiyam, J.; Santhakumar, M.; Deshmukh, D.; Seo, T.W. Design and Implementation of Backstepping Controller for Tilting Thruster Underwater Robot. In Proceedings of the International Conference on Control, Automation and Systems; 2018; Vol. 2018-October.
30. Scholz, G.; Popp, M.; Ruppelt, J.; Trommer, G.F. Model Independent Control of a Quadrotor with Tiltable Rotors: IEEE/ION PLANS 2016, April 11-14, Savannah, Georgia, United States of America. In Proceedings of the Proceedings of the IEEE/ION Position, Location and Navigation Symposium, PLANS 2016; 2016.
31. Phong Nguyen, N.; Kim, W.; Moon, J. Observer-Based Super-Twisting Sliding Mode Control with Fuzzy Variable Gains and Its Application to Overactuated Quadrotors. In Proceedings of the Proceedings of the IEEE Conference on Decision and Control; 2019; Vol. 2018-December.
32. Kumar, R.; Nemati, A.; Kumar, M.; Sharma, R.; Cohen, K.; Cazaurang, F. Tilting-Rotor Quadcopter for Aggressive Flight Maneuvers Using Differential Flatness Based Flight Controller. In Proceedings of the ASME 2017 Dynamic Systems and Control Conference, DSCC 2017; 2017; Vol. 3.
33. Saif, A.-W.A. Feedback Linearization Control of Quadrotor with Tiltable Rotors under Wind Gusts. International Journal of Advanced and Applied Sciences 2017, 4, 150–159, doi:10.21833/ijaas.2017.010.021.
34. Offermann, A.; Castillo, P.; De Miras, J.D. Control of a PVTOL* with Tilting Rotors*. In Proceedings of the 2019 International Conference on Unmanned Aircraft Systems, ICUAS 2019; 2019.
35. Rajappa, S.; Bulthoff, H.H.; Odelga, M.; Stegagno, P. A Control Architecture for Physical Human-UAV Interaction with a Fully Actuated Hexarotor. In Proceedings of the IEEE International Conference on Intelligent Robots and Systems; 2017; Vol. 2017-September.
36. Scholz, G.; Trommer, G.F. Model Based Control of a Quadrotor with Tiltable Rotors. Gyroscopy and Navigation 2016, 7, doi:10.1134/S2075108716010120.
37. Elfeky, M.; Elshafei, M.; Saif, A.W.A.; Al-Malki, M.F. Quadrotor Helicopter with Tilting Rotors: Modeling and Simulation. In Proceedings of the 2013 World Congress on Computer and Information Technology, WCCIT 2013; 2013.
38. Park, S.; Lee, J.; Ahn, J.; Kim, M.; Her, J.; Yang, G.H.; Lee, D. ODAR: Aerial Manipulation Platform Enabling Omnidirectional Wrench Generation. IEEE/ASME Transactions on Mechatronics 2018, 23, doi:10.1109/TMECH.2018.2848255.
39. Magariyama, T.; Abiko, S. Seamless 90-Degree Attitude Transition Flight of a Quad Tilt-Rotor UAV under Full Position Control. In Proceedings of the IEEE/ASME International Conference on Advanced Intelligent Mechatronics, AIM; 2020; Vol. 2020-July.
40. Falanga, D.; Kleber, K.; Mintchev, S.; Floreano, D.; Scaramuzza, D. The Foldable Drone: A Morphing Quadrotor That Can Squeeze and Fly. IEEE Robotics and Automation Letters 2019, 4, doi:10.1109/LRA.2018.2885575.
41. Lu, D.; Xiong, C.; Zeng, Z.; Lian, L. Adaptive Dynamic Surface Control for a Hybrid Aerial Underwater Vehicle with Parametric Dynamics and Uncertainties. IEEE Journal of Oceanic Engineering 2020, 45, doi:10.1109/JOE.2019.2903742.
42. Antonelli, G.; Cataldi, E.; Arrichiello, F.; Giordano, P.R.; Chiaverini, S.; Franchi, A. Adaptive Trajectory Tracking for Quadrotor MAVs in Presence of Parameter Uncertainties and External Disturbances. IEEE Transactions on Control Systems Technology 2018, 26, doi:10.1109/TCST.2017.2650679.
43. Lee, D.; Jin Kim, H.; Sastry, S. Feedback Linearization vs. Adaptive Sliding Mode Control for a Quadrotor Helicopter. Int. J. Control Autom. Syst. 2009, 7, 419–428, doi:10.1007/s12555-009-0311-8.
44. Al-Hiddabi, S.A. Quadrotor Control Using Feedback Linearization with Dynamic Extension. In Proceedings of the 2009 6th International Symposium on Mechatronics and its Applications; Sharjah, United Arab Emirates, March 2009; pp. 1–3.

45. Mukherjee, P.; Waslander, S. Direct Adaptive Feedback Linearization for Quadrotor Control. In Proceedings of the AIAA Guidance, Navigation, and Control Conference; Minneapolis, Minnesota, August 13 2012.
46. Rajappa, S.; Ryll, M.; Bulthoff, H.H.; Franchi, A. Modeling, Control and Design Optimization for a Fully-Actuated Hexarotor Aerial Vehicle with Tilted Propellers. In Proceedings of the Proceedings - IEEE International Conference on Robotics and Automation; 2015; Vol. 2015-June.
47. Chen, C.-C.; Chen, Y.-T. Feedback Linearized Optimal Control Design for Quadrotor With Multi-Performances. IEEE Access 2021, 9, 26674–26695, doi:10.1109/ACCESS.2021.3057378.
48. Lian, S.; Meng, W.; Lin, Z.; Shao, K.; Zheng, J.; Li, H.; Lu, R. Adaptive Attitude Control of a Quadrotor Using Fast Nonsingular Terminal Sliding Mode. IEEE Transactions on Industrial Electronics 2022, 69, 1597–1607, doi:10.1109/TIE.2021.3057015.
49. Chang, D.E.; Eun, Y. Global Chartwise Feedback Linearization of the Quadcopter With a Thrust Positivity Preserving Dynamic Extension. IEEE Trans. Automat. Contr. 2017, 62, 4747–4752, doi:10.1109/TAC.2017.2683265.
50. Martins, L.; Cardeira, C.; Oliveira, P. Feedback Linearization with Zero Dynamics Stabilization for Quadrotor Control. J Intell Robot Syst 2021, 101, 7, doi:10.1007/s10846-020-01265-2.
51. Kuantama, E.; Tarca, I.; Tarca, R. Feedback Linearization LQR Control for Quadcopter Position Tracking. In Proceedings of the 2018 5th International Conference on Control, Decision and Information Technologies (CoDIT); IEEE: Thessaloniki, April 2018; pp. 204–209.
52. Taniguchi, T.; Sugeno, M. Trajectory Tracking Controls for Non-Holonomic Systems Using Dynamic Feedback Linearization Based on Piecewise Multi-Linear Models. IAENG Int. J. Appl. Math. 2017, 47, 339–351.
53. Mutoh, Y.; Kuribara, S. Control of Quadrotor Unmanned Aerial Vehicles Using Exact Linearization Technique with the Static State Feedback. J. Autom. Control Eng 2016, 4, 340–346, doi:10.18178/joace.4.5.340-346.
54. Shen, Z.; Tsuchiya, T. Singular Zone in Quadrotor Yaw–Position Feedback Linearization. Drones 2022, 6, 20, doi:doi.org/10.3390/drones6040084.
55. Voos, H. Nonlinear Control of a Quadrotor Micro-UAV Using Feedback-Linearization. In Proceedings of the 2009 IEEE International Conference on Mechatronics; IEEE: Malaga, Spain, 2009; pp. 1–6.
56. Mistler, V.; Benallegue, A.; M'Sirdi, N.K. Exact Linearization and Noninteracting Control of a 4 Rotors Helicopter via Dynamic Feedback. In Proceedings of the Proceedings 10th IEEE International Workshop on Robot and Human Interactive Communication. ROMAN 2001 (Cat. No.01TH8591); IEEE: Paris, France, September 2001; pp. 586–593.
57. Bolandi, H.; Rezaei, M.; Mohsenipour, R.; Nemati, H.; Smailzadeh, S.M. Attitude Control of a Quadrotor with Optimized PID Controller. ICA 2013, 04, 335–342, doi:10.4236/ica.2013.43039.
58. Bouabdallah, S.; Noth, A.; Siegwart, R. PID vs LQ Control Techniques Applied to an Indoor Micro Quadrotor. In Proceedings of the 2004 IEEE/RSJ International Conference on Intelligent Robots and Systems (IROS) (IEEE Cat. No.04CH37566); IEEE: Sendai, Japan, 2004; Vol. 3, pp. 2451–2456.
59. Wang, S.; Polyakov, A.; Zheng, G. Quadrotor Stabilization under Time and Space Constraints Using Implicit PID Controller. Journal of the Franklin Institute 2022, 359, 1505–1530, doi:10.1016/j.jfranklin.2022.01.002.
60. Bouabdallah, S.; Siegwart, R. Backstepping and Sliding-Mode Techniques Applied to an Indoor Micro Quadrotor. In Proceedings of the Proceedings of the 2005 IEEE International Conference on Robotics and Automation; IEEE: Barcelona, Spain, 2005; pp. 2247–2252.
61. Madani, T.; Benallegue, A. Backstepping Control for a Quadrotor Helicopter. In Proceedings of the 2006 IEEE/RSJ International Conference on Intelligent Robots and Systems; IEEE: Beijing, China, October 2006; pp. 3255–3260.
62. Chen, F.; Lei, W.; Zhang, K.; Tao, G.; Jiang, B. A Novel Nonlinear Resilient Control for a Quadrotor UAV via Backstepping Control and Nonlinear Disturbance Observer. Nonlinear Dyn 2016, 85, 1281–1295, doi:10.1007/s11071-016-2760-y.
63. Liu, P.; Ye, R.; Shi, K.; Yan, B. Full Backstepping Control in Dynamic Systems With Air Disturbances Optimal Estimation of a Quadrotor. IEEE Access 2021, 9, 34206–34220, doi:10.1109/ACCESS.2021.3061598.
64. Xu, R.; Ozguner, U. Sliding Mode Control of a Quadrotor Helicopter. In Proceedings of the Proceedings of the 45th IEEE Conference on Decision and Control; IEEE: San Diego, CA, USA, 2006; pp. 4957–4962.
65. Runcharoon, K.; Srichatrapimuk, V. Sliding Mode Control of Quadrotor. In Proceedings of the 2013 The International Conference on Technological Advances in Electrical, Electronics and Computer Engineering (TAEECE); IEEE: Konya, Turkey, May 2013; pp. 552–557.
66. Luque-Vega, L.; Castillo-Toledo, B.; Loukianov, A.G. Robust Block Second Order Sliding Mode Control for a Quadrotor. Journal of the Franklin Institute 2012, 349, 719–739, doi:10.1016/j.jfranklin.2011.10.017.
67. Xu, L.; Shao, X.; Zhang, W. USDE-Based Continuous Sliding Mode Control for Quadrotor Attitude Regulation: Method and Application. IEEE Access 2021, 9, 64153–64164, doi:10.1109/ACCESS.2021.3076076.

68. Ganga, G.; Dharmana, M.M. MPC Controller for Trajectory Tracking Control of Quadcopter. In Proceedings of the 2017 International Conference on Circuit ,Power and Computing Technologies (ICCPCT); IEEE: Kollam, India, April 2017; pp. 1–6.
69. Abdolhosseini, M.; Zhang, Y.M.; Rabbath, C.A. An Efficient Model Predictive Control Scheme for an Unmanned Quadrotor Helicopter. J Intell Robot Syst 2013, 70, 27–38, doi:10.1007/s10846-012-9724-3.
70. Alexis, K.; Nikolakopoulos, G.; Tzes, A. Model Predictive Control Scheme for the Autonomous Flight of an Unmanned Quadrotor. In Proceedings of the 2011 IEEE International Symposium on Industrial Electronics; June 2011; pp. 2243–2248.
71. Torrente, G.; Kaufmann, E.; Föhn, P.; Scaramuzza, D. Data-Driven MPC for Quadrotors. IEEE Robotics and Automation Letters 2021, 6, 3769–3776, doi:10.1109/LRA.2021.3061307.
72. Ahmed, A.M.; Katupitiya, J. Modeling and Control of a Novel Vectored-Thrust Quadcopter. Journal of Guidance, Control, and Dynamics 2021, 44, 1399–1409, doi:10.2514/1.G005467.
73. Xu, J.; D'Antonio, D.S.; Saldaña, D. H-ModQuad: Modular Multi-Rotors with 4, 5, and 6 Controllable DOF. In Proceedings of the 2021 IEEE International Conference on Robotics and Automation (ICRA); Xi'an, China, May 2021; pp. 190–196.
74. Convens, B.; Merckaert, K.; Nicotra, M.M.; Naldi, R.; Garone, E. Control of Fully Actuated Unmanned Aerial Vehicles with Actuator Saturation. IFAC-PapersOnLine 2017, 50, 12715–12720, doi:10.1016/j.ifacol.2017.08.1823.
75. Cotorruelo, A.; Nicotra, M.M.; Limon, D.; Garone, E. Explicit Reference Governor Toolbox (ERGT). In Proceedings of the 2018 IEEE 4th International Forum on Research and Technology for Society and Industry (RTSI); IEEE: Palermo, Italy, September 2018; pp. 1–6.
76. Dunham, W.; Petersen, C.; Kolmanovsky, I. Constrained Control for Soft Landing on an Asteroid with Gravity Model Uncertainty. In Proceedings of the 2016 American Control Conference (ACC); IEEE: Boston, MA, USA, July 2016; pp. 5842–5847.
77. Hosseinzadeh, M.; Garone, E. An Explicit Reference Governor for the Intersection of Concave Constraints. IEEE Trans. Automat. Contr. 2020, 65, 1–11, doi:10.1109/TAC.2019.2906467.
78. Shen, Z.; Ma, Y.; Tsuchiya, T. Stability Analysis of a Feedback-Linearization-Based Controller with Saturation: A Tilt Vehicle with the Penguin-Inspired Gait Plan. arXiv preprint 2021.
79. Shen, Z.; Tsuchiya, T. State Drift and Gait Plan in Feedback Linearization Control of A Tilt Vehicle. In Proceedings of the Computer Science & Information Technology (CS & IT); Academy & Industry Research Collaboration Center (AIRCC): Vienna, Austria, March 19 2022; Vol. 12, pp. 1–17.
80. Hamandi, M.; Usai, F.; Sablé, Q.; Staub, N.; Tognon, M.; Franchi, A. Design of Multirotor Aerial Vehicles: A Taxonomy Based on Input Allocation. The International Journal of Robotics Research 2021, 40, 1015–1044, doi:10.1177/02783649211025998.
81. Das, A.; Subbarao, K.; Lewis, F. Dynamic Inversion of Quadrotor with Zero-Dynamics Stabilization. In Proceedings of the 2008 IEEE International Conference on Control Applications, San Antonio, TX, USA, 3–5 Sptember 2008; pp. 1189–1194.
82. Ghandour, J.; Aberkane, S.; Ponsart, J.-C. Feedback Linearization Approach for Standard and Fault Tolerant Control: Application to a Quadrotor UAV Testbed. *J. Phys. Conf. Ser.* **2014**, *570*, 082003. https://doi.org/10.1088/1742-6596/570/8/082003.
83. Cowling, I.; Yakimenko, O.; Whidborne, J.; Cooke, A. Direct Method Based Control System for an Autonomous Quadrotor. *J. Intell. Robot. Syst.* **2010**, *60*, 285–316. https://doi.org/10.1007/s10846-010-9416-9.
84. Wie, B. New Singularity Escape/Avoidance Steering Logic for Control Moment Gyro Systems. In Proceedings of the AIAA Guidance, Navigation, and Control Conference and Exhibit; American Institute of Aeronautics and Astronautics: Austin, TX, USA, 11 August 2003.
85. Sands, T.; Kim, J.; Agrawal, B. Singularity Penetration with Unit Delay (SPUD). *Mathematics* **2018**, *6*, 23. https://doi.org/10.3390/math6020023.
86. Senkul, F.; Altug, E. Modeling and Control of a Novel Tilt-Roll Rotor Quadrotor UAV. In Proceedings of the 2013 International Conference on Unmanned Aircraft Systems, ICUAS 2013, Atlanta, GA, USA, 28–31 May 2013.
87. McDonough, K.; Kolmanovsky, I. Controller State and Reference Governors for Discrete-Time Linear Systems with Pointwise-in-Time State and Control Constraints. In Proceedings of the American Control Conference, Chicago, IL, USA, 1–3 July 2015; Volume 2015.
88. Kolmanovsky, I.; Kalabić, U.; Gilbert, E. Developments in Constrained Control Using Reference Governors. *IFAC Proc.* **2012**, *4*, 282–290.
89. Bemporad, A. Reference Governor for Constrained Nonlinear Systems. *IEEE Trans. Automat. Contr.* **1998**, *43*, 415–419. https://doi.org/10.1109/9.661611.
90. Chernova, S.; Veloso, M. An Evolutionary Approach to Gait Learning for Four-Legged Robots. In Proceedings of the 2004 IEEE/RSJ International Conference on Intelligent Robots and Systems (IROS), Sendai, Japan, 28 September–2 October 2004; Volume 3.

91. Bennani, M.; Giri, F. Dynamic Modelling of a Four-Legged Robot. *J. Intell. Robot. Syst. Theory Appl.* **1996**, *17*, 419–428. https://doi.org/10.1007/BF00571701.
92. Talebi, S.; Poulakakis, I.; Papadopoulos, E.; Buehler, M. Quadruped Robot Running with a Bounding Gait. In *Experimental Robotics VII.*; Rus, D., Singh, S., Eds.; Springer: Berlin/Heidelberg, Germany, 2007.
93. Lewis, M.A.; Bekey, G.A. Gait Adaptation in a Quadruped Robot. *Auton. Robot.* **2002**, *12*, 301–312.
94. Hirose, S. A Study of Design and Control of a Quadruped Walking Vehicle. *Int. J. Robot. Res.* **1984**, *3*, 113–133. https://doi.org/10.1177/027836498400300210.
95. Goodarzi, F.A.; Lee, D.; Lee, T. Geometric Adaptive Tracking Control of a Quadrotor Unmanned Aerial Vehicle on SE(3) for Agile Maneuvers. *J. Dyn. Syst. Measur. Control* **2015**, *137*, 091007. https://doi.org/10.1115/1.4030419.
96. Shi, X.-N.; Zhang, Y.-A.; Zhou, D. A Geometric Approach for Quadrotor Trajectory Tracking Control. *Int. J. Control* **2015**, *88*, 2217–2227. https://doi.org/10.1080/00207179.2015.1039593.
97. Lee, T.; Leok, M.; McClamroch, N.H. Nonlinear Robust Tracking Control of a Quadrotor UAV on SE(3): Nonlinear Robust Tracking Control of a Quadrotor UAV. *Asian J. Control* **2013**, *15*, 391–408. https://doi.org/10.1002/asjc.567.
98. Shen, Z.; Tsuchiya, T. Singular Zone in Quadrotor Yaw-Position Feedback Linearization. *arXiv* **2021**, arXiv:2110.07179.
99. Das, A.; Subbarao, K.; Lewis, F. Dynamic Inversion with Zero-Dynamics Stabilisation for Quadrotor Control. *IET Control Theory Appl.* **2009**, *3*, 303–314. https://doi.org/10.1049/iet-cta:20080002.
100. Mokhtari, A.; Benallegue, A. Dynamic Feedback Controller of Euler Angles and Wind Parameters Estimation for a Quadrotor Unmanned Aerial Vehicle. In Proceedings of the IEEE International Conference on Robotics and Automation, 2004. Proceedings. ICRA ’04. 2004; IEEE: New Orleans, LA, USA, 2004; pp. 2359-2366 Vol.3.
101. Nemati, A.; Kumar, M. Non-Linear Control of Tilting-Quadcopter Using Feedback Linearization Based Motion Control.; American Society of Mechanical Engineers: San Antonio, Texas, USA, October 22 2014; p. V003T48A005.
102. Mokhtari, A.; M’Sirdi, N.K.; Meghriche, K.; Belaidi, A. Feedback Linearization and Linear Observer for a Quadrotor Unmanned Aerial Vehicle. *Advanced Robotics* **2006**, *20*, 71–91, doi:10.1163/156855306775275495.
103. Ansari, U.; Bajodah, A.H.; Hamayun, M.T. Quadrotor Control Via Robust Generalized Dynamic Inversion and Adaptive Non-Singular Terminal Sliding Mode. *Asian Journal of Control* **2019**, *21*, 1237–1249, doi:10.1002/asjc.1800.
104. Verdugo, M.R.; Rahal, S.C.; Agostinho, F.S.; Govoni, V.M.; Mamprim, M.J.; Monteiro, F.O. Kinetic and Temporospatial Parameters in Male and Female Cats Walking over a Pressure Sensing Walkway. *BMC Veterinary Research* **2013**, *9*, 129, doi:10.1186/1746-6148-9-129.
105. Wisleder, D.; Zernicke, R.F.; Smith, J.L. Speed-Related Changes in Hindlimb Intersegmental Dynamics during the Swing Phase of Cat Locomotion. *Exp Brain Res* **1990**, *79*, doi:10.1007/BF00229333.
106. Vilensky, J.A.; Njock Libii, J.; Moore, A.M. Trot-Gallop Gait Transitions in Quadrupeds. *Physiology & Behavior* **1991**, *50*, 835–842, doi:10.1016/0031-9384(91)90026-K.
107. Chovancová, A.; Fico, T.; Chovanec, Ľ.; Hubinsk, P. Mathematical Modelling and Parameter Identification of Quadrotor (a Survey). *Procedia Engineering* **2014**, *96*, 172–181, doi:10.1016/j.proeng.2014.12.139.
108. Abas, N.; Legowo, A.; Akmeliawati, R. Parameter Identification of an Autonomous Quadrotor. In Proceedings of the 2011 4th International Conference on Mechatronics (ICOM); May 2011; pp. 1–8.
109. Altug, E.; Ostrowski, J.P.; Mahony, R. Control of a Quadrotor Helicopter Using Visual Feedback. In Proceedings of the Proceedings 2002 IEEE International Conference on Robotics and Automation (Cat. No.02CH37292); Washington, DC, USA, May 2002; Vol. 1, pp. 72–77.
110. Zhou, Q.-L.; Zhang, Y.; Rabbath, C.-A.; Theilliol, D. Design of Feedback Linearization Control and Reconfigurable Control Allocation with Application to a Quadrotor UAV. In Proceedings of the 2010 Conference on Control and Fault-Tolerant Systems (SysTol); IEEE: Nice, France, October 2010; pp. 371–376.
111. Shen, Z.; Tsuchiya, T. The Pareto-Frontier-Based Stiffness of A Controller: Trade-off between Trajectory Plan and Controller Design. *arXiv:2108.08667 [cs, eess]* **2021**.
112. Michael, N.; Mellinger, D.; Lindsey, Q.; Kumar, V. The GRASP Multiple Micro-UAV Testbed. *IEEE Robotics Automation Magazine* **2010**, *17*, 56–65, doi:10.1109/MRA.2010.937855.
113. Michieletto, G.; Cenedese, A.; Zaccarian, L.; Franchi, A. Hierarchical nonlinear control for multi-rotor asymptotic stabilization based on zero-moment direction. *Automatica* **2020**, *117*, 108991. https://doi.org/10.1016/j.automatica.2020.108991.
114. Shen, Z.; Ma, Y.; Tsuchiya, T. Feedback linearization based tracking control of a tilt-rotor with cat-trot gait plan. *ArXiv* **2022**, ArXiv*:* 2202.02926 *CsRO*.
115. Appel, K.; Haken, W. The Solution of the Four-Color-Map Problem. *Sci. Am.* **1977**, *237*, 108–121.
116. Franklin, P. The Four Color Problem. *Am. J. Math.* **1922**, *44*, 225–236. https://doi.org/10.2307/2370527.

117. Shen, Z.; Ma, Y.; Tsuchiya, T. Feedback Linearization-Based Tracking Control of a Tilt-Rotor with Cat-Trot Gait Plan. *International Journal of Advanced Robotic Systems* **2022**, *19*, 17298806221109360, doi:10.1177/17298806221109360.
118. Shen, Z.; Ma, Y.; Tsuchiya, T. Four-Dimensional Gait Surfaces for a Tilt-Rotor—Two Color Map Theorem. *Drones* **2022**, *6*, 103, doi:10.3390/drones6050103.
119. Oosedo, A.; Abiko, S.; Narasaki, S.; Kuno, A.; Konno, A.; Uchiyama, M. Flight Control Systems of a Quad Tilt Rotor Unmanned Aerial Vehicle for a Large Attitude Change. In Proceedings of the 2015 IEEE International Conference on Robotics and Automation (ICRA); IEEE: Seattle, WA, USA, May 2015; pp. 2326–2331.
120. Blandino, T.; Leonessa, A.; Doyle, D.; Black, J. Position Control of an Omni-Directional Aerial Vehicle for Simulating Free-Flyer In-Space Assembly Operations. In Proceedings of the ASCEND 2021; American Institute of Aeronautics and Astronautics: Las Vegas, Nevada & Virtual, November 15 2021.
121. Shen, Z.; Tsuchiya, T. The Robust Gait of a Tilt-Rotor and Its Application to Tracking Control -- Application of Two Color Map Theorem 2022.
122. Hameduddin, I.; Bajodah, A.H. Nonlinear Generalised Dynamic Inversion for Aircraft Manoeuvring Control. *International Journal of Control* **2012**, *85*, 437–450, doi:10.1080/00207179.2012.656143.
123. Mellinger, D.; Kumar, V. Minimum Snap Trajectory Generation and Control for Quadrotors. In Proceedings of the 2011 IEEE International Conference on Robotics and Automation; IEEE: Shanghai, China, May 2011; pp. 2520–2525.
124. Ryll, M.; Muscio, G.; Pierri, F.; Cataldi, E.; Antonelli, G.; Caccavale, F.; Franchi, A. 6D Physical Interaction with a Fully Actuated Aerial Robot. In Proceedings of the 2017 IEEE International Conference on Robotics and Automation (ICRA); May 2017; pp. 5190–5195.
125. Shen, Z.; Ma, Y.; Tsuchiya, T. Generalized Two Color Map Theorem -- Complete Theorem of Robust Gait Plan for a Tilt-Rotor 2022.
126. Ducard, G.J.J.; Hua, M.-D. Discussion and Practical Aspects on Control Allocation for a Multi-Rotor Helicopter. *Int. Arch. Photogramm. Remote Sens. Spatial Inf. Sci.* **2012**, *XXXVIII-1/C22*, 95–100, doi:10.5194/isprsarchives-XXXVIII-1-C22-95-2011.
127. Mochida, S.; Matsuda, R.; Ibuki, T.; Sampei, M. A Geometric Method of Hoverability Analysis for Multirotor UAVs With Upward-Oriented Rotors. *IEEE Transactions on Robotics* **2021**, *37*, 1765–1779, doi:10.1109/TRO.2021.3064101.
128. Michieletto, G.; Ryll, M.; Franchi, A. Fundamental Actuation Properties of Multirotors: Force–Moment Decoupling and Fail–Safe Robustness. *IEEE Trans. Robot.* **2018**, *34*, 702–715, doi:10.1109/TRO.2018.2821155.
129. Grupo de Procesamiento de Señales, Identificación y Control (GPSIC), Departamento de Ingeniería Electrónica, Facultad de Ingeniería Universidad de Buenos Aires (FIUBA), Argentina; Giribet, J.I.; Pose, C.D.; Ghersin, A.S.; Mas, I. Experimental Validation of a Fault Tolerant Hexacopter with Tilted Rotors. *IJEETC* **2018**, 58–65, doi:10.18178/ijeetc.7.2.58-65.
130. Lee, H., Kim, S., Ryan, T., & Kim, H. J. (2013, October). Backstepping control on se (3) of a micro quadrotor for stable trajectory tracking. In *2013 IEEE international conference on systems, man, and cybernetics* (pp. 4522-4527). IEEE.
131. Şenkul, F., & Altuğ, E. (2013, May). Modeling and control of a novel tilt—Roll rotor quadrotor UAV. In *2013 International Conference on Unmanned Aircraft Systems (ICUAS)* (pp. 1071-1076). IEEE.
132. Zamani, A., Khorram, M., & Moosavian, S. A. A. (2011, October). Dynamics and stable gait planning of a quadruped robot. In *2011 11th International Conference on Control, Automation and Systems* (pp. 25-30). IEEE.
133. Wu, F., & Grigoriadis, K. M. (1997, December). LPV systems with parameter-varying time delays. In *Proceedings of the 36th IEEE Conference on Decision and Control* (Vol. 2, pp. 966-971). IEEE.
134. Zhou, B. (2016). On asymptotic stability of linear time-varying systems. *Automatica*, *68*, 266-276.
135. G. K. K. and P. M. Pathak, "Dynamic Modelling & Simulation of a four legged jumping robot with compliant legs," *Robotics and Autonomous Systems*, vol. 61, no. 3, pp. 221–228, 2013.
136. T. Kato, K. Shiromi, M. Nagata, H. Nakashima, and K. Matsuo, "Gait pattern acquisition for four-legged mobile robot by genetic algorithm," *IECON 2015 - 41st Annual Conference of the IEEE Industrial Electronics Society*, 2015.
137. T. Taniguchi, L. Eciolaza, and M. Sugeno, "Tracking control for a non-holonomic car-like robot using dynamic feedback linearization based on piecewise bilinear models," *2014 IEEE International Conference on Fuzzy Systems (FUZZ-IEEE)*, 2014.
138. E. Garone and M. M. Nicotra, "Explicit reference governor for Constrained Nonlinear Systems," *IEEE Transactions on Automatic Control*, vol. 61, no. 5, pp. 1379–1384, 2016.
139. Garone, E., Di Cairano, S. and Kolmanovsky, I., Reference and command governors for systems with constraints: A survey on theory and applications. *Automatica*, 75, pp.306-328, 2017.
140. M. M. Nicotra and E. Garone, "Explicit reference governor for continuous time nonlinear systems subject to convex constraints," *2015 American Control Conference (ACC)*, 2015.

141. W. Craig, D. Yeo, and D. A. Paley, "Geometric attitude and position control of a quadrotor in wind," *Journal of Guidance, Control, and Dynamics*, vol. 43, no. 5, pp. 870–883, 2020.
142. K. Gamagedara, M. Bisheban, E. Kaufman, and T. Lee, "Geometric controls of a quadrotor UAV with decoupled yaw control," *2019 American Control Conference (ACC)*, 2019.
143. H. Lee, S. Kim, T. Ryan, and H. J. Kim, "Backstepping control on SE(3) of a micro Quadrotor for stable trajectory tracking," *2013 IEEE International Conference on Systems, Man, and Cybernetics*, 2013.
144. E. Hendricks, O. Jannerup, and P. H. Sørensen, *Linear systems control: deterministic and stochastic methods*. Springer Science & Business Media, 2008.
145. Khalil, Hassan K., *Nonlinear systems third edition*, Patience Hall 115, 2002.
146. Shen, Z.; Tsuchiya, T. Tracking Control for a Tilt-Rotor with Input Constraints by Robust Gaits. 7.
147. Li, Q. Masters Thesis: Grey-Box System Identification of a Quadrotor Unmanned Aerial Vehicle. **2014**.
148. Doukhi, O.; Fayjie, A.R.; Lee, D.J. Intelligent Controller Design for Quad-Rotor Stabilization in Presence of Parameter Variations. *Journal of Advanced Transportation* **2017**, *2017*, 1–10, doi:10.1155/2017/4683912.